\documentclass{article}

\usepackage{microtype}
\usepackage{graphicx}
\usepackage{subfigure}
\usepackage{booktabs} 

\usepackage{hyperref}

\usepackage{amsmath}
\usepackage{amsthm}
\usepackage{amsfonts}

\usepackage{ulem}
\usepackage{float}

\usepackage{multirow}

\usepackage[dvipsnames]{xcolor}
\definecolor{Mycolor1}{HTML}{00F9DE}

\newtheorem{theorem}{Theorem}
\newtheorem{definition}{Definition}
\newtheorem{proposition}{Proposition}

\usepackage[accepted]{icml2021}

\icmltitlerunning{Interpretable performance analysis towards offline reinforcement learning: A dataset perspective}

\begin{document}

\twocolumn[
\icmltitle{{\color{black}{
{Interpretable performance analysis towards offline \\reinforcement learning: A dataset perspective}
}}}




\begin{icmlauthorlist}
	\icmlauthor{Chenyang Xi}{ali,bit}
	\icmlauthor{Bo Tang}{ali}
	\icmlauthor{Jiajun Shen}{purdu,ali}
	\icmlauthor{Xinfu Liu}{bit}
	\icmlauthor{Feiyu Xiong}{ali}
	\icmlauthor{Xueying Li}{ali}
\end{icmlauthorlist}

\icmlaffiliation{ali}{Alibaba Group, Hangzhou, China}
\icmlaffiliation{bit}{School of Aerospace Engineering, Beijing Institute of Technology, Beijing, China}
\icmlaffiliation{purdu}{School of Electrical and Computer Engineering, Purdue University, Indiana, USA}

\icmlcorrespondingauthor{Xinfu Liu}{lau.xinfu@gmail.com}


\vskip 0.3in
]



\printAffiliationsAndNotice{}


\begin{abstract}
Offline reinforcement learning (RL) has increasingly become the focus of the artificial intelligent research due to its wide real-world applications where the collection of data may be difficult, time-consuming, or costly. In this paper, we first propose a two-fold taxonomy for existing offline RL algorithms from the perspective of exploration and exploitation tendency. Secondly, we derive the explicit expression of the upper bound of extrapolation error and explore the correlation between the performance of different types of algorithms and the distribution of actions under states. Specifically, we relax the strict assumption on the sufficiently large amount of state-action tuples. Accordingly, we provably explain why batch constrained Q-learning (BCQ)  performs better than other existing techniques. Thirdly, after identifying the weakness of BCQ on dataset of low mean episode returns, we propose a modified variant based on top return selection mechanism, which is proved to be able to gain the state-of-the-art performance on various datasets. Lastly, we create a benchmark platform on the Atari domain, entitled RL easy go (RLEG), at an estimated cost of more than 0.3 million dollars. We make it open-source for fair and comprehensive competitions between offline RL algorithms with complete datasets and checkpoints being provided. 
\end{abstract}

\section{Introduction}
\label{Introduction}

\subsection{Background and motivations}
Reinforcement learning (RL) tries to figure out how intelligent agent ought to take actions under the interaction with environment such that the accumulative reward could be maximized, and becomes increasingly popular due to its wide real-world applications where data collection may be difficult, time-consuming, and costly.

In most studies concerned with reinforcement learning (RL) algorithms \cite{sutton2018reinforcement-book}, authors assume that an agent interacts with an online environment or simulator and learns from the ''dynamic'' date set generated by updated policy of its own. However, when facing complex real world problems, it is a totally different case due to the extremely large data (including states and actions), which limits the applicability of online methodologies. As a consequence, offline RL (also known as batch RL in some researches) algorithms are well and rapidly developed especially in many practical scenarios where the explorations (actions of trial and errors) are extremely costly, e.g., robotics, E-commercials, manufactures. Especially, in E-commercial case, RL has been widely applied in different and challenging business scenarios, e.g., coupons delivery \cite{xiao2019AntFin}, search engine \cite{hu2018RL_EcommerceSearchEgine}, recommendations \cite{zhao2018RL_EcommerceRecommendation}, impression allocation \cite{cai2018RL_Ecommerce_ImpressionAllocation}, etc. Nonetheless, each update and iteration of algorithm already deployed online would introduce uncertainties to production system, which possibly contributes to an asset loss accident. In addition, performance improvement cannot be guaranteed since the training process of online algorithms (e.g. Nature DQN) is time-consuming.

{\color{black}{
However, we notice that there {are no conclusive investigations and reviews for the effectiveness and applicability of existing offline RL algorithms}, which no doubtedly would make readers confused when choosing algorithms given an offline dataset. For example, both batch-constrained Q-learning (BCQ) and random ensemble mixture (REM) are claimed to preform better than each other. However, the offline dataset of their experiments are basically different. In BCQ experiments, the offline dataset is generated by a partially trained DDPG (i.e. a medium oracle), while that of REM is generated by Nature DQN (i.e. a combination of starter, medium and complete oracle). Consequently, it is of great necessity to figure out the underlying principles such that fair comparisons could be made.

To address the above problems, we first propose a taxonomy,  which divides the existed offline RL algorithms into two categories, exploitation-tentative algorithms (e.g. BCQ, \cite{Fujimoto:C-BCQ}, \cite{Fujimoto:D-BCQ}, best action imitation learning (BAIL) \cite{chen2020bail}, bootstrapping error accumulation reduction (BEAR) \cite{Kumar:BEAR}, safe Policy improvement with baseline bootstrapping (SPIBB) \cite{laroche2019SPIBB}), and exploration-tentative algorithms (e.g. ensemble DQN, \cite{fausser2015EnsembleDQN}, \cite{anschel2017EnsembleDQN}, Bootstrapped-DQN, \cite{osband2016BootstrapDQN},  C51, \cite{bellemare2017C51}, Quantile Regression(QR)-DQN \cite{dabney2017QR-DQN}, REM \cite{agarwal2020REM}).
}}

\subsection{Literature Review}
\subsubsection{Exploitation-tentative algorithms}
{\color{black}{
In \cite{Fujimoto:C-BCQ}, authors claim that most off-policy algorithms would fail in offline setting due to extrapolation errors caused by erroneously estimating the unseen state-action pairs, and therefore proposed BCQ. In BCQ, when selecting actions that maximize Q value, they further eliminate actions which are unlikely to be selected by behavioral policy using a generative model. Experiments are made on offline dataset generated by deep deterministic policy gradient (DDPG).
}}

The counterpart of \cite{Fujimoto:C-BCQ} on discrete action space is \cite{Fujimoto:D-BCQ} where discrete BCQ is claimed to be the optimal offline RL algorithm. However, its performance can only achieve to be equivalent to or a little bit higher than the one of noiseless policy, which is obtained based on the data set generated by partially trained DQN, i.e., a quite lower level of performance. Under the circumstance of limited data in offline setting, BCQ acts more likely to robust imitation learning algorithm (\cite{Wang:RobustIL}). {In contrast to the investigation in \cite{agarwal2020REM} where the scale of offline dataset is assumed to be large enough, the authors conclude that common off-policy Deep RL algorithms are not well suitable for offline learning tasks.}

Similar to BCQ of continuous version, \cite{Kumar:BEAR} also {imposes the  constraints on the distribution of continuous action space in off-policy Q-learning cases}. Authors identify bootstrapping error as key source of instability in existing off-policy RL algorithms, {\color{black}{the performance of which could not be elevated merely through {scaling} up the the batch.}}

{\color{black}{
Different from imposing strict constraints on distributional similarities in BCQ, the basic mindset of BEAR is to make trade-off between concentrability coefficient (i.e., the parameter quantifying the degree to which current states and actions are out of distribution generated by behavioral policy) and suboptimality constant (i.e., the parameter quantifying the distance between the current policy and the optimal one). {Compared} with BCQ, the visitation distribution generated by current policies would not be too much similar to the batch data distribution in BEAR. Thus, BEAR can be treated as a robust variant of BCQ.}} {\color{black}{However, it is not readily extended to the cases of discrete action space.}}

{\color{black}{
Instead focusing on generating similar state-action visitations, \cite{chen2020bail} tries to crack the problem from a imitation-learning perspective by selecting ''valuable'' state-action pairs and episodes that contain enough information for learning a relatively optimal strategy with regards to higher returns. {Supervised learning methodology has been applied for the derivation of an upper envelope where high return data  lie nearby}. {Accordingly, the optimal strategy is obtained directly through imitation.}
}} 

In \cite{laroche2019SPIBB}, authors judge the value of state-action pairs based on the number of occurrences. For a specified state-action pair $(s,a)$, it will be accepted for further imitation learning process (in a greedy way) only if the number of occurrences $N(s,a)$ {\color{black}{is more than a fixed threshold $N_{\wedge}(s,a)$ (calculated based on \cite{petrik2016safe} and \cite{weissman2003inequalities})}}, otherwise  behavioral policy would serves as a baseline.

Our discussion and review on exploitation-tentative offline RL algorithms can be concluded in {\color{black}{Table. \ref{Table: exploitation-tentative}}}.

\begin{table*}[!htb]
\centering
\vspace{-0.4cm}
\caption{Comparison of exploitation-tentative offline RL algorithms}
\label{Table: exploitation-tentative}
\begin{tabular}{p{2.5cm}p{2.5cm}p{2.5cm}p{2.5cm}p{2cm}p{2cm}}
 \hline
     & C-BCQ & BAIL & D-BCQ & BEAR & SPIBB \\ 
 \hline
Scenario & Continuous action space & Continuous action space & Discrete action space & Continuous action space & Countable $(s, a)$ pairs \\ 
Constraints on (s,a) quantity & N/A & YES & YES & N/A & YES \\

Main idea & Maximizing the similarities between behavioral and trained policy & Imitate (s,a) pairs with high return & Extension of C-BCQ to discrete action space & Relax the constraints on distributional similarities & Safely improved based on behavioral policy \\ 
 
Pros & SOTA among exploitation-tentative algorithms & Same as C-BCQ & Better performance on random/less Oracle dataset & Readily implement & Safely bounded performance \\

Cons & Rely on Oracle dataset & Same as C-BCQ & Suboptimal performance &  & Hard to count N(s,a) \\
 
 \hline
\end{tabular}
\vspace{-0.4cm}
\end{table*}

\subsubsection{Exploration-tentative algorithm}
{\color{black}{
C51 \cite{bellemare2017C51}, as a comb form methodology, extends the Q-value to Q-distribution where value function is defined as the expectation of value distribution with multiple peaks. Ensemble-DQN \cite{anschel2017EnsembleDQN} is a simple extension of DQN that approximates
the Q-values via an ensemble of parameterized Q-functions, i.e., multiple heads. It should be noted that each head independently estimates the Q-value with huber loss. The final loss is derived by simply taking average of all heads.

Bootstrapped-DQN \cite{osband2016BootstrapDQN} uses one of the Q-value estimates in each episode to improve the depth of exploration. The authors claim that bootstrapped neural networks are able to produce reasonable posterior estimates for regression. The basic mindset of REM \cite{agarwal2020REM} borrows from dropout mechanism. For five different outputs (Q-networks generated by a shared neural network), authors randomly assign the weights with sampling performance of the algorithm being lifted. In order to underscore the importance of randomness, the authors also make comparisons between their random mechanism and the average one, which is proved to be less optimal. 
}}

{\color{black}{
Among most of off-policy algorithms, QR-DQN is provably to be the best rather still underperform the policy with noise \cite{dabney2017QR-DQN}. Again, it should be noted that although QR-DQN is not exclusively developed for offline setting, it is still able to achieve high performance given sufficiently large and complete data set.

Our discussion and review on exploration-tentative offline RL algorithms can be concluded in Table. \ref{Table: exploration-tentative}.
}}

\begin{table*}[!htb]
\centering
\vspace{-0.4cm}
\caption{Comparison of exploration-tentative off-policy RL algorithms}
\label{Table: exploration-tentative}
\begin{tabular}{p{1.5cm}p{2cm}p{3cm}p{3cm}p{2cm}p{3cm}}
 \hline
     & Ensemble DQN & Bootstrapped DQN & C51 & QR-DQN & REM \\ 
 \hline
Main idea & Taking average of single heads & Randomly select optimal strategy of higher possibility & Extend Q-value to Q-distribution & Same as C51 & Convex combination of single heads \\ 
Pros & Easy to compute & Efficient training time & Innovated idea & Theoretical proof based & SOTA Offline RL methodology under Atari environment \\  
Cons & Training process is not stable & N/A & No theoretical support & N/A & No theoretical support\\
 \hline
\end{tabular}
\vspace{-0.4cm}
\end{table*}

Noticed that in all imitation-based offline RL algorithms, a strong assumption of the amount of $(s,a)$ pairs, i.e., $N(s,a) > N_{\wedge}(s,a)$, $\forall(s,a) \in \mathcal{S} \times \mathcal{A}$, is a necessity. In fact the assumption is unreal in certain practical scenarios especially video games due to the high cost as we have discussed before. 

\subsection{Main contribution}

From the above literature reviews, we notice that existing offline RL techniques have not been well concluded, and the applicability of them under various datasets has not been clearly stated either. Thus, readers might be confused about how to select the most appropriate algorithm when facing a brand new dataset generated by different behavioral polices or even unknown ones. In practical scenarios, trial and error is often costly and time consuming. Besides, the argue between self-claimed SOTA algorithms has not been well resolved due to the totally different behavior policies. To address all the mentioned concerns, we list our four-fold contributions as:

(1) We propose a taxonomy for existing offline RL algorithms from the perspective of exploration and exploitation tendency.

(2) Based on the derivation of upper bounded extrapolation error, we provably investigate the applicability of both types of algorithms on different datasets (in terms of different action distributions under states) and explain why BCQ performs better than existing techniques.

(3) We identify the limitation of BCQ, as its weak performance on datasets with low mean episode return. To bridge the gap, we propose a modified version by introducing a return-based data selection mechanism, which reaches better performance on various datasets.

(4) A benchmark of Atari domain is open sourced and most existing offline RL algorithms are included. We spend more than $0.3$ million dollars on the experiments and collect all intermediate results incuding various datasets and the intermediate model checkpoints. The benchmark could be used for fair and comprehensive competitions between existing and future offline RL algorithms. 

%
%
%
%
%

{\color{black}{
{\color{black}{
\section{Extrapolation error-based applicability analysis and comparisons}
}}
In this section, we would analyze the applicability of both exploration and exploitation-tentative algorithms on different datasets from a perspective of extrapolation error.

{Due to the inability of interacting with the environment}, both types of offline RL algorithms we conclude before {have to be suffered from the failure of learning as well as online ones}. More specifically, when testing the offline trained model in real world,  the mismatch between offline dataset and practical state-action visitations of the current policy would give rise to the extrapolation error, which introduces the performance gap between offline and online RL algorithms.

Besides, intuitively, due to {he vital role that behavioral policy plays} in offline RL training, the discrepancies of various offline datasets would contribute to the different extrapolation errors and therefore distinct performance of offline RL algorithms. The corresponding argue with respect to the state-of-the-art performance comes up referring to the one between above-mentioned REM and BCQ. How to provably explain and make a fair comparison is undoubtedly a challenging task, which would be resolved in this section. 
}}
{\color{black}{
\subsection{Can extrapolation error be completely eliminated?}
In existing exploitation-tentative algorithms, researchers try to eliminate the extrapolation error by placing strict constraints on the distribution of (state-action) tuples with the help of generative model (e.g. \cite{Fujimoto:C-BCQ}, \cite{Fujimoto:D-BCQ}, \cite{Kumar:BEAR}), simply discarding the tuples of insufficient amount of occurrences (e.g. \cite{chen2020bail}) or replacing these tuples with behavioral policy as a safe baseline (e.g. \cite{laroche2019SPIBB}). However, as we conclude in Table \ref{Table: exploitation-tentative}, a strong assumption on the sufficient visitations of selected tuples is necessary.

For exploration-tentative algorithms, extrapolation error exists as well but claims to completely vanish by assuming a {big enough} offline dataset with adequate diversity (e.g. 50 million tuples per game from Nature DQN \cite{agarwal2020REM}).

However, we notice that the assumptions concerned with the size of tuples in the offline dataset are too strong to be satisfied in practical scenarios (especially for some costly application such as E-commercial, robotics, etc). In addition, for large continuous state or action space, it is impossible to accurately count the amount of occurrences for each tuple.

Thus, in the following context, we start with deriving an explicit upper bound on the extrapolation error, and investigate how it is affected by behavioral policy and corresponding offline dataset.


\subsection{Preliminaries}

We first introduce following useful lemmas and necessary assumptions for the ensuing proofs.
}}

\newtheorem{lemma}{Lemma}[section]
\begin{lemma} 
    \label{lemma: upper bound of Q}
\cite{weissman2003inequalities} For $\forall {(s,a)} \in \mathcal{S} \times \mathcal{A}$, $Q(s, a) \leq  \frac{R_{max}}{{1 - \gamma}}$
\end{lemma}

\begin{lemma} 
    \label{lemma: e(s,a) expression}
\cite{petrik2016safe} For $\forall {(s,a)} \in \mathcal{S} \times \mathcal{A}$, $||p_1(.|s, a) - p_2(.|s, a)||_1 \leq e(s, a)$ where, $e(s, a) = \sqrt{\frac{2}{N(s, a)}log{\frac{|\mathcal{S}||\mathcal{A}|2^{|\mathcal{S}|}}{\delta}}}$
\end{lemma}

{\color{black}{
\newtheorem{assumption}{Assumption}[section]
\begin{assumption}
    \label{assumption: state-action pair occurence number}
In offline dataset to be fed to training process, the number of all state-action pair occurrences satisfy $ N(s) = \sum_{a\in \mathcal{A}} N(s, a) = N$, $\forall s \in \mathcal{S}$, i.e., although for each state $s \in \mathcal{S}$, the distribution of action $a \in \mathcal{A}$ may vary, the total amount of occurrence of each state would be the same.
\end{assumption}

{Compared with the strong assumption that $N(s,a) < N_{\wedge}(s,a)$, $\forall(s,a) \in \mathcal{S} \times \mathcal{A}$ made in most previous research}, the assumption \ref{assumption: state-action pair occurence number} is much relaxed and therefore more realistic to be satisfied in practical scenarios, which also indicates that our work is not a trivial extension of previous work.

\begin{assumption}
	\label{assumption: agent's rationality}
The agent is rational (i.e., aiming for a relatively better result or higher total rewards following the rules of environment) such that pure random policy (i.e., under each state, choosing each action with same probability) is always the worst one (i.e., the mean episode return is the lowest) among all candidate policies, while the policies of biased action distribution under each state would result in a better achievement. In addition, the more biased the distribution is, the higher reward is achieved.
\end{assumption}

It is noted that we use \textit{randomness} for capturing the degree of distributional bias in the following context, which would be further discussed in the ensuing sections. Assumption \ref{assumption: agent's rationality} reveals our insights about some practical scenarios (e.g., video games, E-commercial, robotics, etc) for applying RL-based methodologies, which reflects an underlying mindset that Oracle policy are mostly deterministic. We admit that there might be cases against this mind as shown in the following four-quadrant figure, through which we try to explain Assumption \ref{assumption: agent's rationality} from a perspective of the correlation between randomness and returns.

\begin{figure}[!htb]
	\centering
	\includegraphics[width=2.0in]{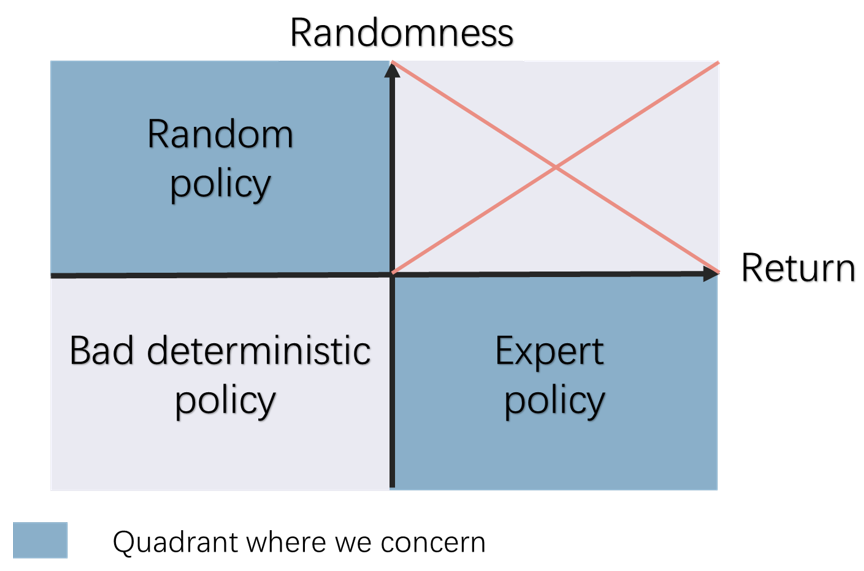}\\
	\centering
	\caption{\textbf{Correlation between randomness and returns}}
	\label{fig:four-quadrant figure}
\end{figure}

In Fig. \ref{fig:four-quadrant figure}, we take all possible cases into consideration. The 2nd and 4th quadrants conform the Assumption \ref{assumption: agent's rationality}. Nonetheless, some cases might fall in 3rd quadrant due to some inaccurate description of fundamental elements, e.g., dynamics, rewards. Game players may misunderstand the goal and take completely opposite actions to optimal ones such that worst return is achieved. These special cases are not included in this paper and would be discussed separately.

%

\begin{assumption}
If a dataset is generated by a given policy $ \pi_b$, then for $\forall {s} \in \mathcal{S}$, $N(s, a) = N(s)\pi_b(a|s)$ 
\end{assumption}
}}

In the following context, for notational convenience, we omit the arguments of 
$N(s)$, and simply denote it as $N$.


{\color{black}{
\subsection{How does dataset affect the extrapolation error of {\color{black}{exploration-tentative}} algorithms?}

\begin{proposition}
    \label{Proposition: upper bound of extrapolaton error}
    The extrapolation error $\epsilon_{s,a}$ is upper bounded by

        \begin{equation}
        \begin{aligned}
            \overline{\epsilon}_{s, a} &= (2log(\frac{|\mathcal{S}||\mathcal{A}|2^{|\mathcal{S}|}}{\delta}))^{1/2} \frac{R_{max}}{({1 - \gamma})}{N}^{-1/2} \\
            &[
            ({\pi_{b}{(a|s)}})^{-1/2}\\
                        & +\gamma \sum_{s'}p_1(s'|s, a) \sum_{a'} \pi(a'|s')
                        (\pi_{b}{(a'|s')})^{-1/2}\\
                        & + ... + {\gamma}^{(n)} \sum_{s'}p_1(s'|s, a) \sum_{a'} \pi(a'|s') \sum_{s''} ...\\ 
                        & \sum_{s^{(n)}}p_1(s^{(n)}|s^{(n-1)}, a^{(n-1)}) \\
                        & \sum_{a^{(n)}} \pi(a^{(n)}|s^{(n)})(\pi_b{(a^{n}|s^{n})})^{-1/2}+...]\\
        \end{aligned}
    \end{equation}
    
\end{proposition}

Inspired by the results of Proposition \ref{Proposition: upper bound of extrapolaton error}, we define the randomness of dataset as follow.

\begin{definition}
    \label{definition: dataset quality}
    The randomness of dataset is defined as
    \begin{equation}
        \label{eq: dataquality}
        q = \frac{1}{|\mathcal{S}|}\sum_{s\in\mathcal{S}}\sum_{i\in\mathcal{A}}\frac{1}{\sqrt{\pi_{b}(a_{i}|s)}}
    \end{equation}
\end{definition}
}}

{\color{black}{
According to Definition \ref{definition: dataset quality}, a pure random behavior policy {having a highest value of $q$ indicates the highest randomness of the corresponding datasets}, i.e., uniform distribution of actions under each state. 
}}



\begin{assumption}
    \label{assumption: dirichlet distribution}
The distribution of $\pi$ is uniform, i.e., the possibilities of any type of trainer or player are same. Intuitively, it is also noted that the distribution of $\pi$ is independent of the distribution of $\pi_{b}$.
\end{assumption}

Now, we are ready to propose the following theorem. 

{\color{black}{
\begin{theorem}
    \label{theorem: main theorem}
    Given the distribution of $\pi(a|s)$ follows assumption \ref{assumption: dirichlet distribution}, the general term of extrapolation error, equation (\ref{equation: general term of extrapolation error}),  reaches its minimum when $\pi_{b}{(a|s)}$ is pure random. Also, the more even the behavior policy $\pi_{b}{(a|s)}$ under each state is, the less value it would be.
    
    \begin{equation}
    \label{equation: general term of extrapolation error}
        \sum_{a} \pi(a|s)\frac{R_{max}}{({1 - \gamma})}
                        N^{-1/2}(\pi_{b}{(a|s)})^{-1/2}, {\forall}{s }\in \mathcal{S}
    \end{equation}
    
    Equivalently, the optimal behavioral policy is
    
     $\pi^{*}{(a|s)} = $
     $\frac{1}{\vert \mathcal{A} \vert}$ = $\mathop{argmin} \limits_{{\pi_{b}} \in \Pi} \mathbb{E}_{\pi}[{\sum_{a} \pi(a|s) ({\pi_{b}}{(a|s)})^{-\frac{1}{2}}}]$, \ $\forall (a,s) \in \mathcal{A} \times \mathcal{S}$
     
\end{theorem}
}}

{\color{black}{
According to Theorem \ref{theorem: main theorem}, the performance of exploration-tentative algorithms would best when offline dataset is of the highest randomness. Besides, the returns would decrease along with the randomness descent of dataset.
}}

%
%


{\color{black}{
\subsection{How does dataset affect the extrapolation error of exploitation-tentative algorithms?}

In this section, we continue to investigate influence of dataset on exploitation-tentative algorithms. As we have discussed in Section 1, most  exploitation-tentative algorithms fall into two underlying mindsets represented by BCQ and BAIL respectively, i.e., generating similar state-action distribution, and selecting targeted state-action tuples. Thus, the following investigation is two-fold.


\subsubsection{BCQ-like algorithms}


For BCQ-like algorithms, actions chosen for offline optimization must satisfy the constraint $G(a|s) > \tau$, where $G$ is the generative model for selecting batch-constrained actions
and quantitatively depends on the number of occurrences of tuple $(s,a)$ in dataset. Given Assumption \ref{assumption: state-action pair occurence number} being satisfied, we have $N(s, a) > N\tau$, where $(s, a)$ are batch-constrained tuples.

Thus, we derive the upper bounded extrapolation error for BCQ-like algorithms, based on which we further give the sufficient condition for ensuring a lower upper boundary compared with exploration-tentative methods through Theorem \ref{Theorem: compare the upper bound}.

%

\begin{proposition}
    \label{Proposition: upper bound of extrapolaton error for BCQ}
    The extrapolation error for BCQ, $\epsilon_{s,a}$, is upper bounded by

        \begin{equation}
        \label{equation: upper bound of extrapolaton error for BCQ}
        \begin{aligned}
            &\overline{\epsilon}_{s, a} = (2log(\frac{|\mathcal{S}||\mathcal{A}|2^{|\mathcal{S}|}}{\delta}))^{1/2} \frac{R_{max}}{({1 - \gamma})}{(N\tau)}^{-1/2} \\
            &\quad [1 + \gamma
                         + ... + {\gamma}^{(n)}
                         +...]
        \end{aligned}
    \end{equation}
    
    
\end{proposition}

{\color{black}{
\begin{theorem}
    \label{Theorem: compare the upper bound}
    When $\tau > \frac{1}{|A|}$, The upper bound of extrapolation error for BCQ-like algorithms is strictly less than exploration-tentative methods.
\end{theorem}
}}
}
}



{
    \color{black}{
    
\subsubsection{BAIL-like algorithms }
    }
}

{\color{black}{
Instead of choosing $(S, A)$ pairs with $N(S, A) > N\tau$, the underlying mechanism of BAIL is to select tuples with top returns given under each state for offline training. The data selection scheme can be either percentile-based or difference-based. As claimed by the author, the first one performs better, and thus is considered in this paper. For notational convenience and easier understanding, the percentile is still notated by $\tau$ for BAIL-like algorithms. The dataset after percentile-based selection is denoted as $\mathcal{\hat{D}}$, with the corresponding MDP $\hat{M}$. The number of occurrence of tuple $(s,a)$ is denoted as $\hat{N}(S, A)$.

Intuitively, the distribution of $\hat{N}(s, a)$, $\forall (s,a) \in \mathcal{S} \times \mathcal{A}$ would be of greater variance and can barely be lower bounded by a specified value, such as $N\tau$ in BCQ-like case, due to the selection merely based on episode return. Thus, it seems quite {challenged} to derive an explicit expression as same as equation (\ref{equation: upper bound of extrapolaton error for BCQ}).

\begin{assumption}
	\label{assumption: expectation of N(s,a)}
	The expectation of each $\hat{N}(s,a)$ is $N(s,a)\tau$.
\end{assumption}

However, given assumption \ref{assumption: expectation of N(s,a)}, we are able to derive the upper bounded extrapolation error and its corresponding minimum for BAIL-like algorithms.
}}

\begin{proposition}
    \label{Proposition: upper bound of extrapolaton error for BAIL}
    {\color{black}{
    The expectation of extrapolation error for BAIL-like algorithms,}} $\epsilon_{s,a}$, is upper bounded by

    \begin{equation}
    \setlength{\abovedisplayskip}{-2pt}
        \label{equation: general term of extrapolation error for BAIL}
        \begin{aligned}\mathbb{E}(\overline{\epsilon}_{s, a}) &= (2log(\frac{|\mathcal{S}||\mathcal{A}|2^{|\mathcal{S}|}}{\delta}))^{1/2} \frac{R_{max}}{({1 - \gamma})}{(N\tau)^{-1/2}} \\
        & [ {(\pi_b(a|s)}^{-1/2} +\gamma \sum_{s'}p_1(s'|s, a) \sum_{a'} {\pi_b(a'|s')}^{1/2}\\& + ... + {\gamma}^{(n)} \sum_{s'}p_1(s'|s, a) \sum_{a'} \pi_b(a'|s') \sum_{s''} ...\\& \sum_{s^{(n)}}p_1(s^{(n)}|s^{(n-1)}, a^{(n-1)}) \\
        &\sum_{a^{(n)}} {\pi_b(a^{n}|s^{n})}^{1/2} +...]\\\end{aligned}
    \setlength{\abovedisplayskip}{-2pt}
    \end{equation}
    
    
\end{proposition}

\begin{proposition}
    \label{proposition: correlation between dataset quality and extrapoliation error for BAIL}
    The general term of extrapolation error, equation (\ref{equation: general term of extrapolation error for BAIL}),  reaches its minimum when{\color{black}{ $\pi_{b}{(a|s)}$ is uniform, i.e., a pure random policy.}} Also, the more even the action distribution under each state, $\pi_{b}{(a|s)}$, is, the less value it would be.
    
    {\color{black}{
    Specifically, the corresponding optimal value of extrapolation error would be
    \begin{equation}
    \begin{aligned} \mathbb{E}^{*}(\overline{\epsilon}_{s, a}) &= (2log(\frac{|\mathcal{S}||\mathcal{A}|2^{|\mathcal{S}|}}{\delta}))^{1/2} \frac{R_{max}}{({1 - \gamma})}{(N\tau)}^{-1/2} \\
    &[|\mathcal{A}|^{-1/2} +\gamma |\mathcal{A}|^{1/2}+ ... + {\gamma}^{(n)}|\mathcal{A}|^{1/2} +...]
    \end{aligned}
    \end{equation}
    }}
    {\color{black}{
     On the other hand, when $\pi_{b}$ is deterministic,}} the optimal value of extrapolation error is
    
    {\color{black}{
     \begin{equation}
    \begin{aligned}
     \mathbb{E}^{*}({\overline{\epsilon}_{s, a}}) &= (2log(\frac{|\mathcal{S}||\mathcal{A}|2^{|\mathcal{S}|}}{\delta}))^{1/2} \frac{R_{max}}{({1 - \gamma})}{(N\tau)}^{-1/2}\\
                        & [1+\gamma 
                        + ... + {\gamma}^{(n)}
                        +...]
    \end{aligned}
    \end{equation}
    }}
\end{proposition}



{\color{black}{
\subsection{Comparisons among algorithms}
 Compared with exploration-tentative algorithms, exploitation-tentative algorithms are most likely to have less extrapolation error. As for exploitation-tentative algorithms, from Proposition \ref{proposition: correlation between dataset quality and extrapoliation error for BAIL}, we notice that the best case of BAIL-like algorithm is the same as the general case of BCQ-like algorithms. From the perspective of extrapolation error, BCQ-like algorithms provably performs better than the others.
}}

{\color{black}{
\section{Top Return Batch Constrained Q-learning (TR-BCQ)}
{\color{black}{
According to the results of Section 2, exploitation-tentative algorithms would achieves less extrapolation error. However, less extrapolation error does not equivalent to be better overall performance (i.e., estimate Q-value $\pm$ extrapolation error). The red{estimated} Q-value is critical as well. Compared with BAIL-like algorithms, BCQ-like algorithms adopt extra off-policy optimization techniques for a higher estimated Q-value, and thus is expected to have a better overall performance. Besides, for the exploitation-tentative algorithms, a good ``teacher'' is of great importance due to their imitation-based essentials. Thus, we explore the weakness of BCQ as it would suffer from the dataset generated by the behavioral policy of low mean episode returns. To fix this shortcoming, we propose a variant of BCQ, named top return batch constrained Q-learning (TR-BCQ). For easy understanding and avoiding confusions, we will use ``low-quality dataset'' or directly ``low dataset'' for the same meaning of ``dataset generated by the behavioral policy of low mean episode returns'' in the following context.}}

The proposed TR-BCQ is basically consisted of two phases:
}}

{\color{black}{
\textbf{Phase 1 -- Top return data selection}: In this phase, we select tuples with high episode return based on a percentile parameter, $\zeta$;

\textbf{Phase 2 -- Tuple visitation constrained optimization}: In this phase, we proceed off-policy optimization on a visitation constrained batch of tuples.
}}

Please refer to the pseudocode in the appendix.

%

    
{\color{black}{
We further explain advantages of our algorithm over original BCQ not merely on low-quality dataset, starting with the investigation of  extrapolation errors as follow

\begin{equation}
\vspace{-0.2cm}
  \label{equation: bound for TRCQ1}
  \vert Q_{M}^{\pi_{BCQ}}  -  Q_{M^*}^{\pi_{BCQ}} \vert \le \overline{\epsilon}_{1}, \vert Q_{{\hat{M}}}^{\pi_{TRCQ}}  - Q_{M^
  	*}^{\pi_{TRCQ}} \vert \le \overline{\epsilon}_{2} 
\end{equation}


where $M$ indicates the underlying MDP of original offline dataset, $M^{*}$ is the MDP in the realistic world, ${\hat{M}}$ represents the MDP of top-return selected offline dataset.

\begin{proposition}
    \label{Proposition: error bound for selected dataset of TRCQ}
    Using same off-policy optimization techniques, the relationship between the upper bound of extrapolation error generated on $\hat{M}$ and $M$ is as follows:
    \begin{equation}
    \label{equation: upper bound of extrapolaton error for TR-BCQ}
    \mathbb{E}({{\hat{\overline{\epsilon}}}_2}) =\mathbb{E}({{\overline{\epsilon}}_1}){\zeta}^{-1/2}
    \end{equation}
    
\end{proposition}
}}





{\color{black}{
Additionally, since imitating the selected tuples with higher episode returns, the policy derived by TR-BCQ would achieve higher {estimated Q-value compared with the original BCQ, i.e., $ Q_{M}^{\pi_{BCQ}} \textless Q_{\hat{M}}^{\pi_{TRCQ}}$.}

Nevertheless, due to the uncertainty introduced by $\overline{\epsilon}_2$, TR-BCQ might not {gain} a better performance. Meanwhile, we notice that the extrapolation error rather changes slightly along with $\zeta$, e.g., $ \mathbb{E}({{\hat{\overline{\epsilon}}}_{2}}) =1.29 \mathbb{E}({{\overline{\epsilon}}_{1}}) $ when $\zeta = 60\%$, i.e., tuples of top $40\%$ episode returns are selected. It indicates that we are able to achieve higher overall performance (i.e., estimate Q-value $\pm$ extrapolation error) by increasing extrapolation error a little bit in return for higher estimate Q-value.

It should be noted that although the name of proposed algorithm contains ``BCQ'', BCQ is not the only option for Phase 2. More advanced extrapolation error-insensitive techniques are applicable for further improvement, yet which is out of the scope of this paper.  
}}

\begin{figure}[htbp]
\centering
\centering
\includegraphics[width=2.7in]{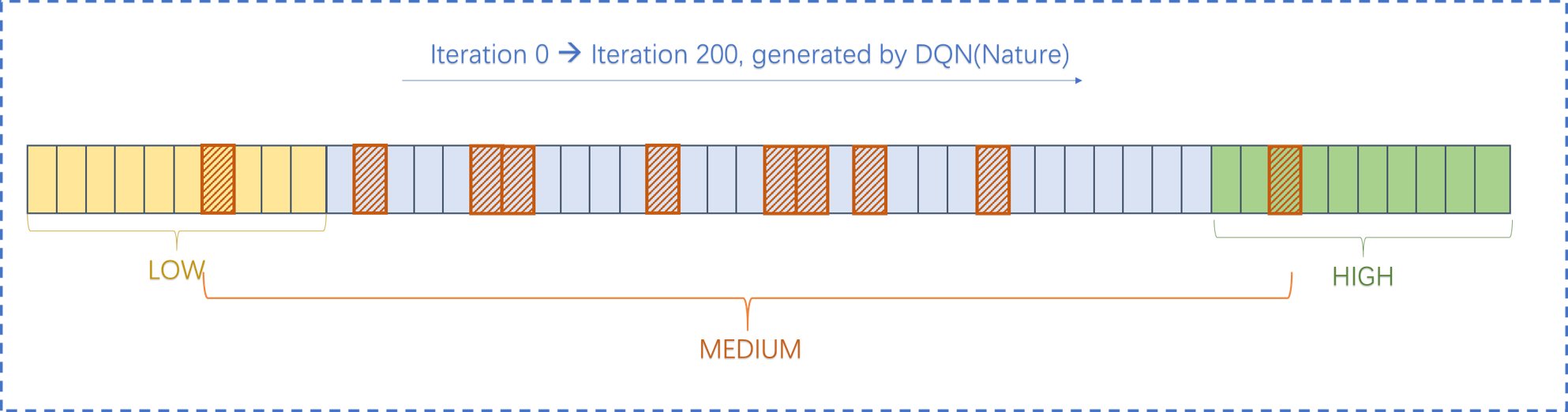}\\
\centering
\caption{\textbf{Datasets of Tri-level Quality}}
\vspace{-0.4cm}
\label{fig:datasets of tri-level quality}
\vspace{-0.3cm}
\end{figure}

\begin{table*}[!htb]
    \vspace{-0.4cm}
    \centering
    \footnotesize
    \caption{Comparison on dataset quality}
    \begin{tabular}{llllllll}
    \hline \noalign{\smallskip}
    \multicolumn{2}{l}{\multirow{2}{*}{}}    & \multicolumn{6}{c}{Algorithms} \\
    \multicolumn{2}{l}{}                                           &  BCQ  & BAIL&   DQN    &  MultiHeadDQN  & Quantile  &  REM \\ \hline
    \multirow{1}{*}{Trend of Performance} &  Increase &  $41.667\%$  &$43.333\%$ & $15.000\%$   & $6.667\%$  & $5.000\%$  &  $6.667\%$ \\
    {with Dataset Quality}&    Decrease      &  $21.667\%$  &$21.667\%$&  $56.667\%$  &  $66.667\%$  & $48.333\%$  & $53.333\%$  \\ 
    \multirow{3}{*}{Num of Best Scores}  & Low &  $11.667\%$ &  $16.667\%$ & $0.000\%$ &  $10.000\%$  & $55.000\%$  &  $6.667\%$ \\
                                                    & Medium & $25.000\%$&$25.000\%$& $0.000\%$& $5.000\%$& $41.667\%$& $3.333\%$\\
                                                    & High & $33.667\%$& $20.000\%$&$1.667\%$& $8.333\%$& $30.000\%$& $3.333\%$\\\hline
    \end{tabular}
    \label{table: SOTA algo on different dataset qualities}
    \vspace{-0.4cm}
  
\end{table*}

\begin{figure*}[htbp]
\centering

\subfigure[Alien]{
    \begin{minipage}[t]{0.20\linewidth}
        \centering
        \includegraphics[width=1.55in]{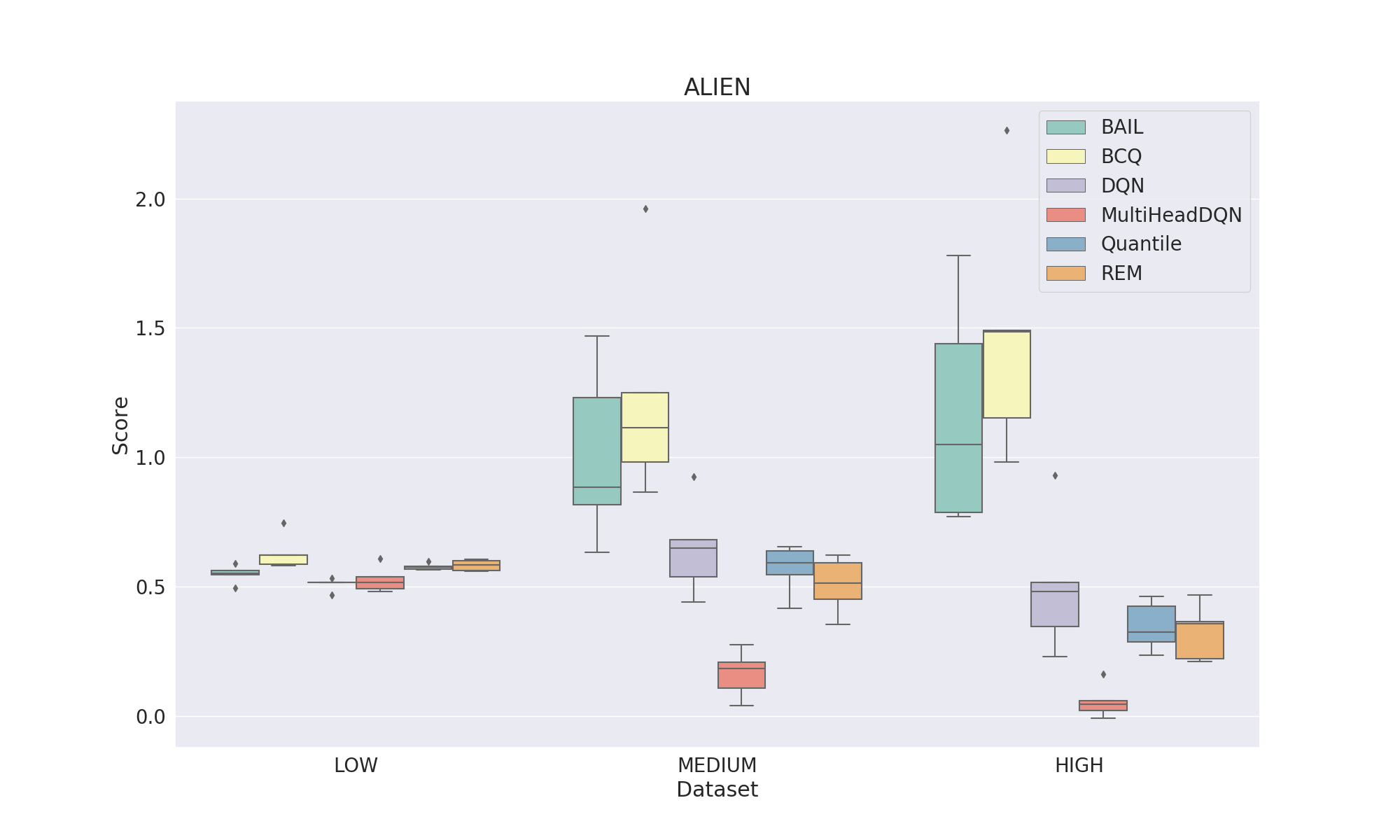}\\
        \vspace{0.01cm}
    \end{minipage}%
}%
\subfigure[Amidar]{
    \begin{minipage}[t]{0.20\linewidth}
        \centering
        \includegraphics[width=1.55in]{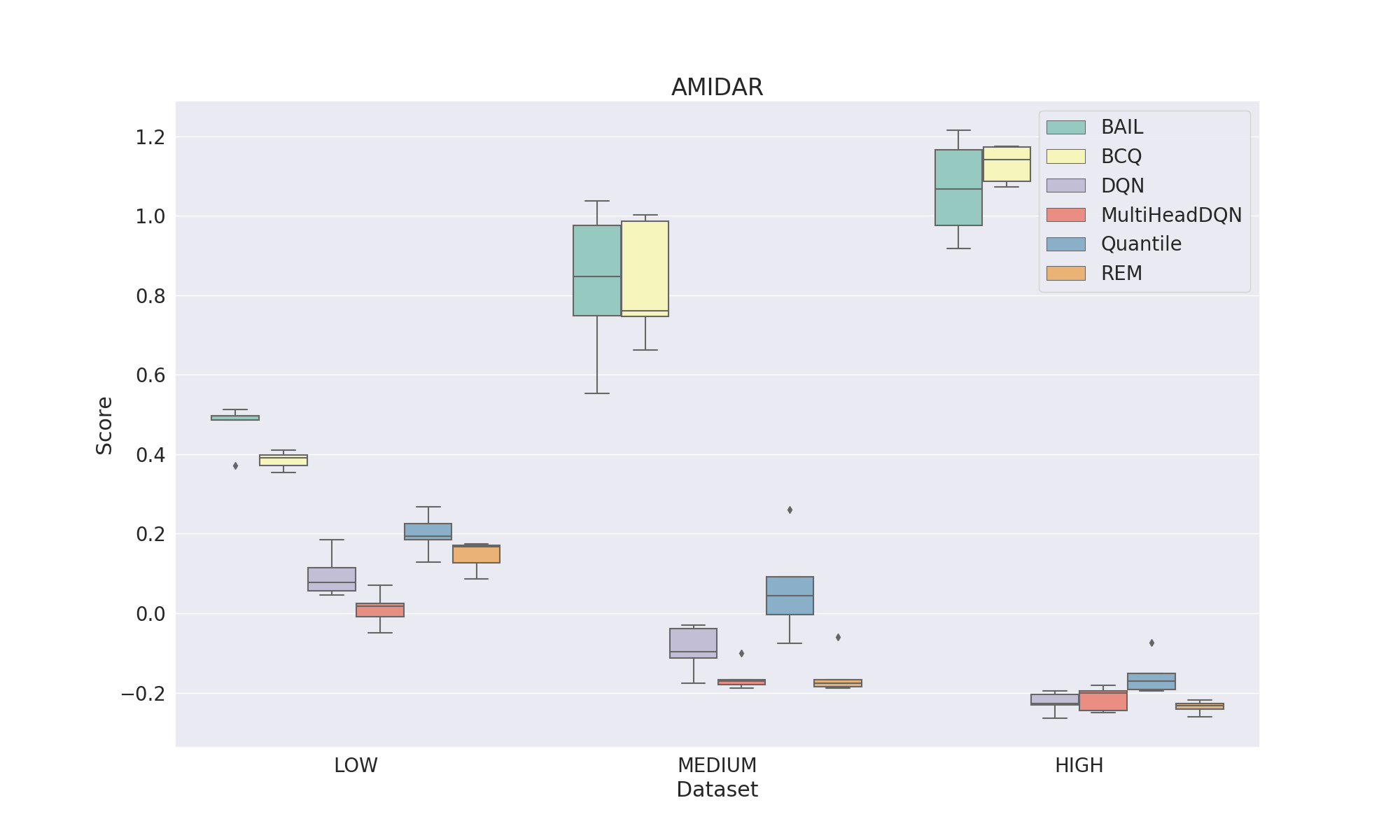}\\
        \vspace{0.01cm}
    \end{minipage}%
}%
\subfigure[Atlantis]{
    \begin{minipage}[t]{0.20 \linewidth}
        \centering
        \includegraphics[width=1.55in]{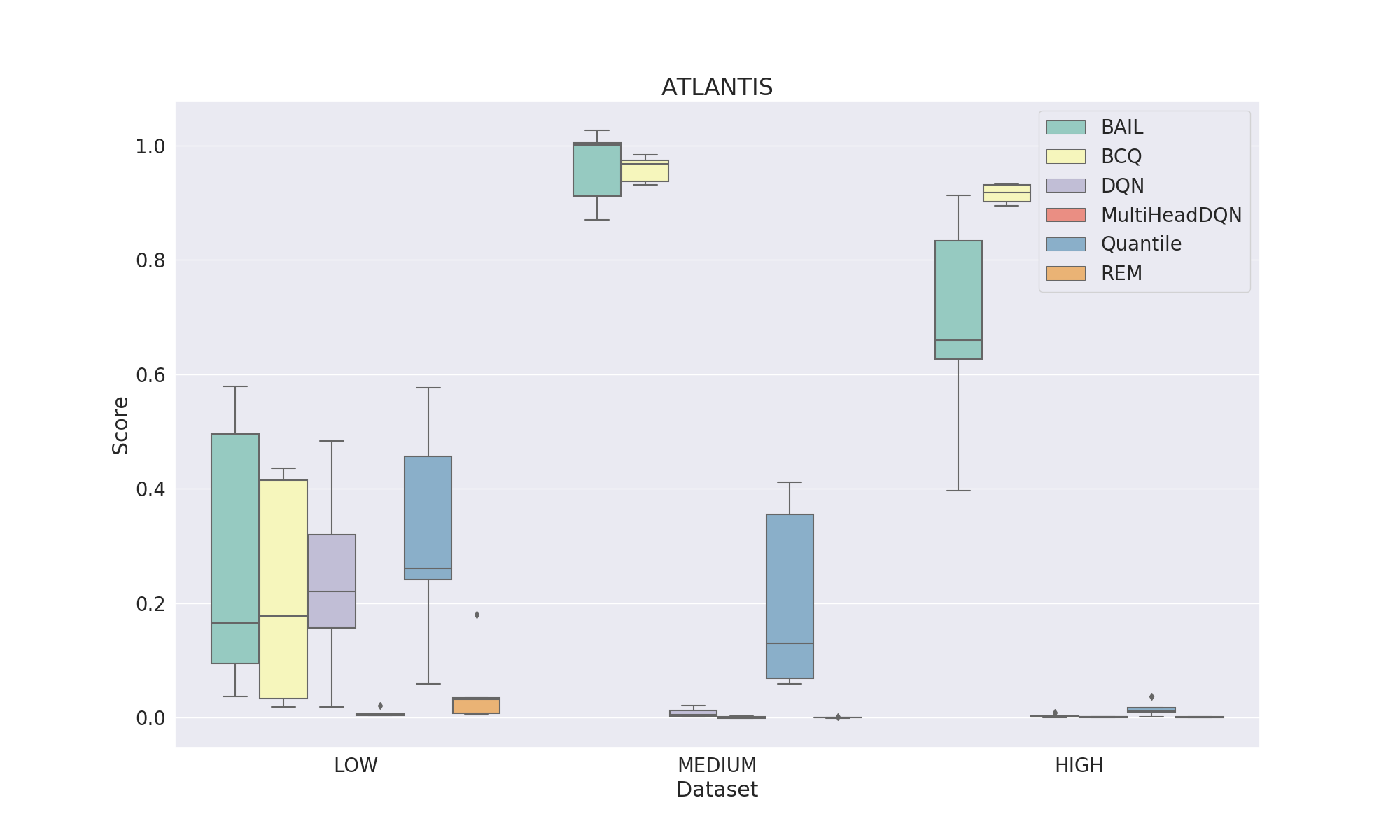}\\
        \vspace{0.01cm}
    \end{minipage}%
}%
\subfigure[Boxing]{
    \begin{minipage}[t]{0.20\linewidth}
        \centering
        \includegraphics[width=1.55in]{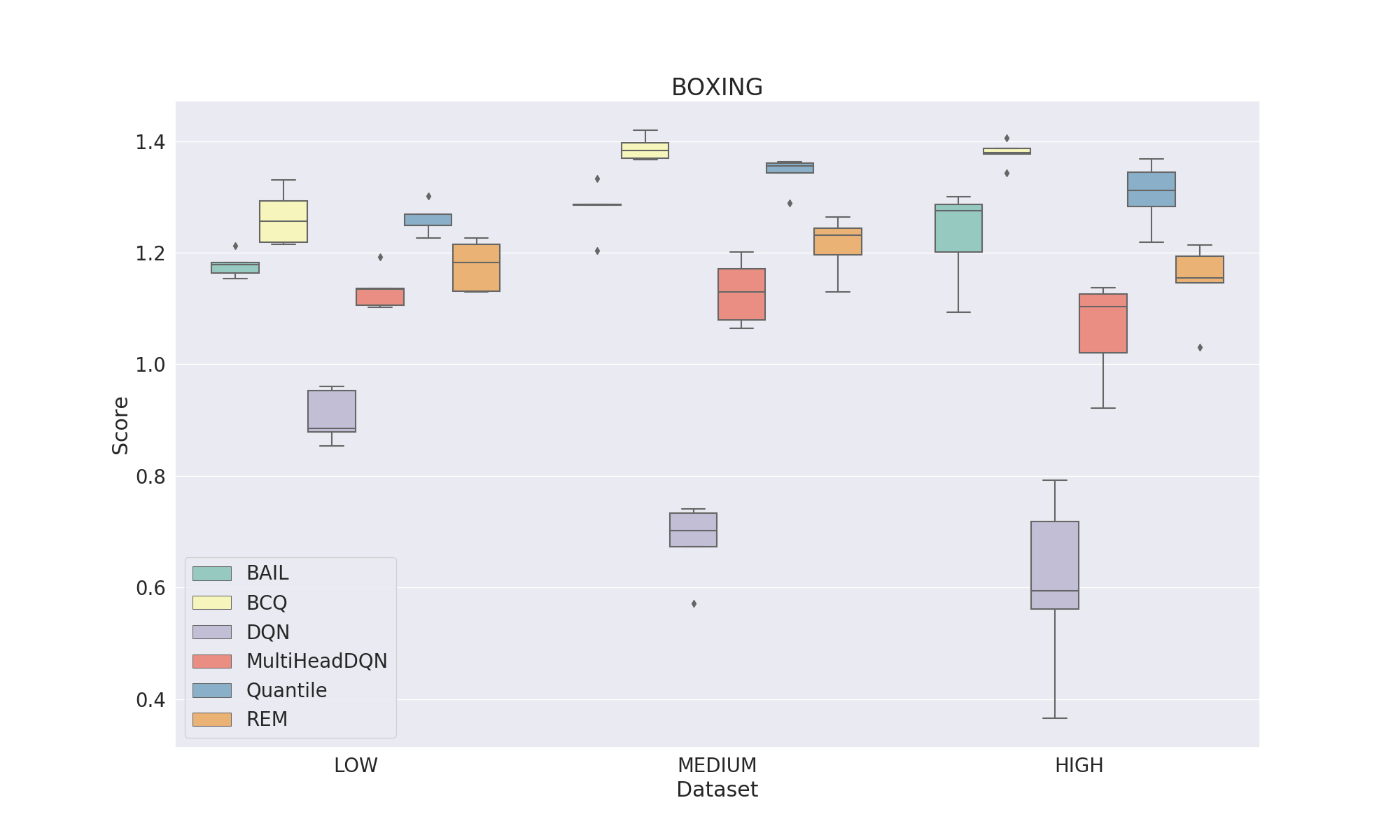}\\
        \vspace{0.01cm}
    \end{minipage}%
}%
\subfigure[Kangaroo]{
    \begin{minipage}[t]{0.20\linewidth}
        \centering
        \includegraphics[width=1.55in]{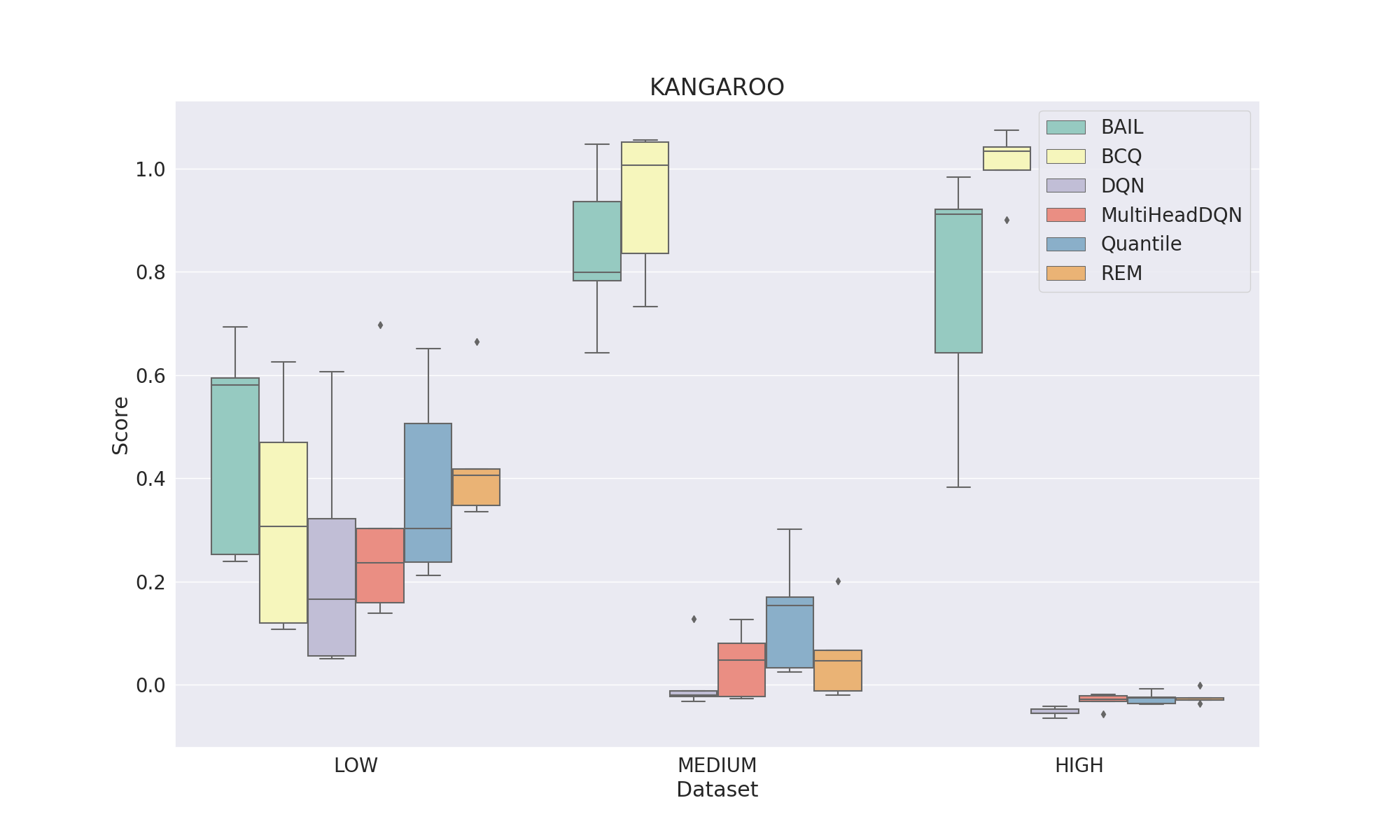}\\
        \vspace{0.01cm}
    \end{minipage}%
}%

\vspace{-0.4cm}
\subfigure[Krull]{
    \begin{minipage}[t]{0.20\linewidth}
        \centering
        \includegraphics[width=1.55in]{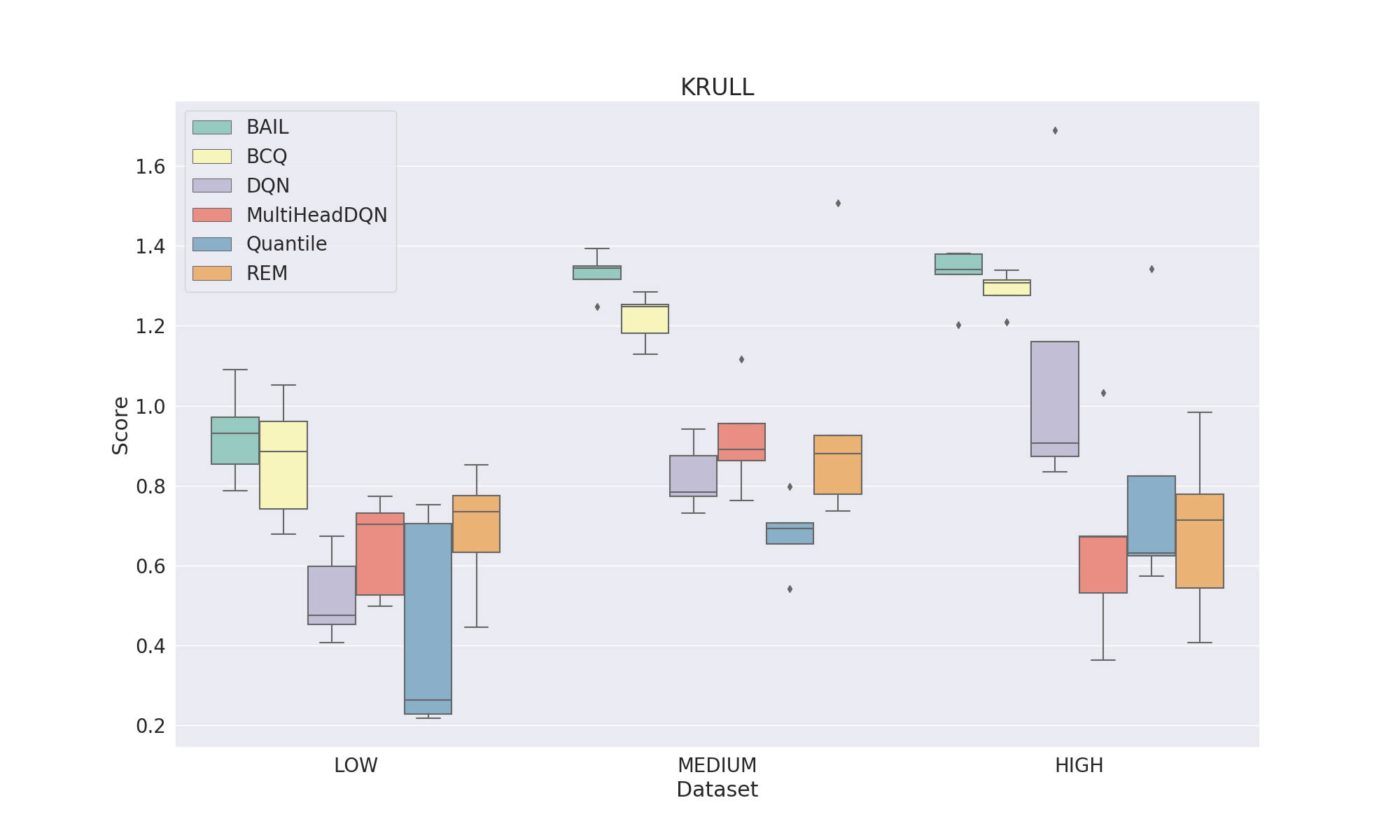}\\
        \vspace{0.01cm}
    \end{minipage}%
}%
\subfigure[Phoenix]{
    \begin{minipage}[t]{0.20\linewidth}
        \centering
        \includegraphics[width=1.55in]{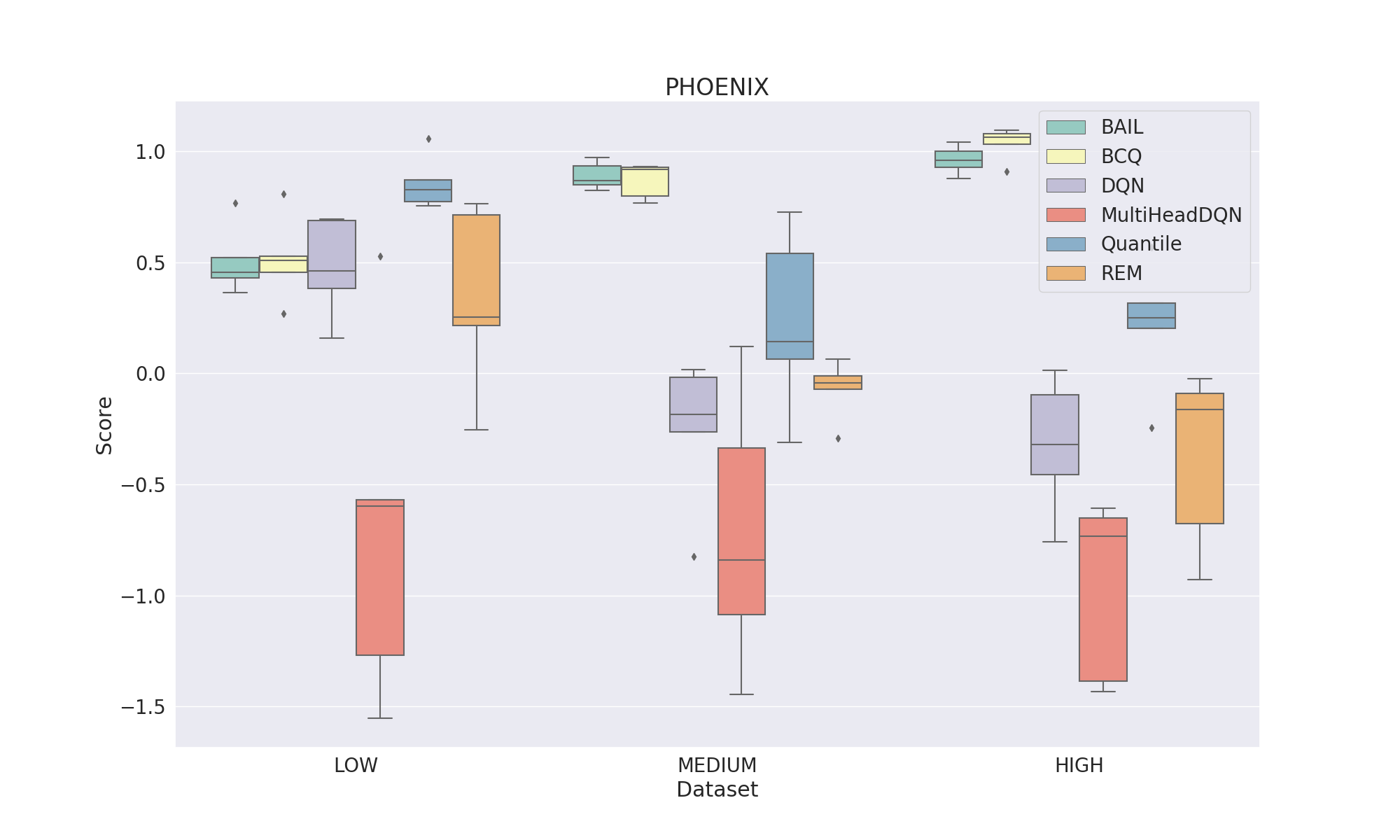}\\
        \vspace{0.01cm}
    \end{minipage}%
}%
\subfigure[Pong]{
    \begin{minipage}[t]{0.20\linewidth}
        \centering
        \includegraphics[width=1.55in]{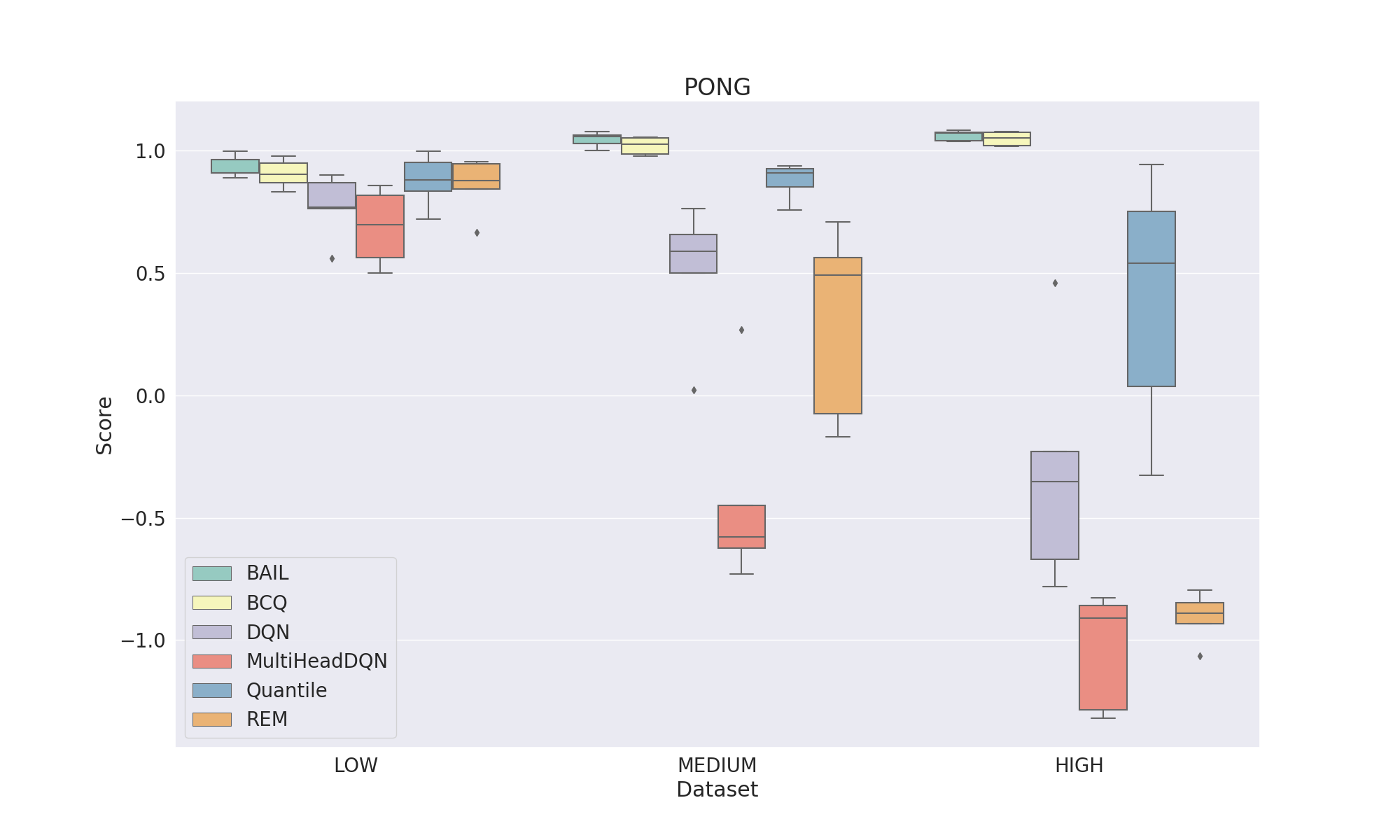}\\
        \vspace{0.01cm}
    \end{minipage}%
}%
\subfigure[Qbert]{
    \begin{minipage}[t]{0.20\linewidth}
        \centering
        \includegraphics[width=1.55in]{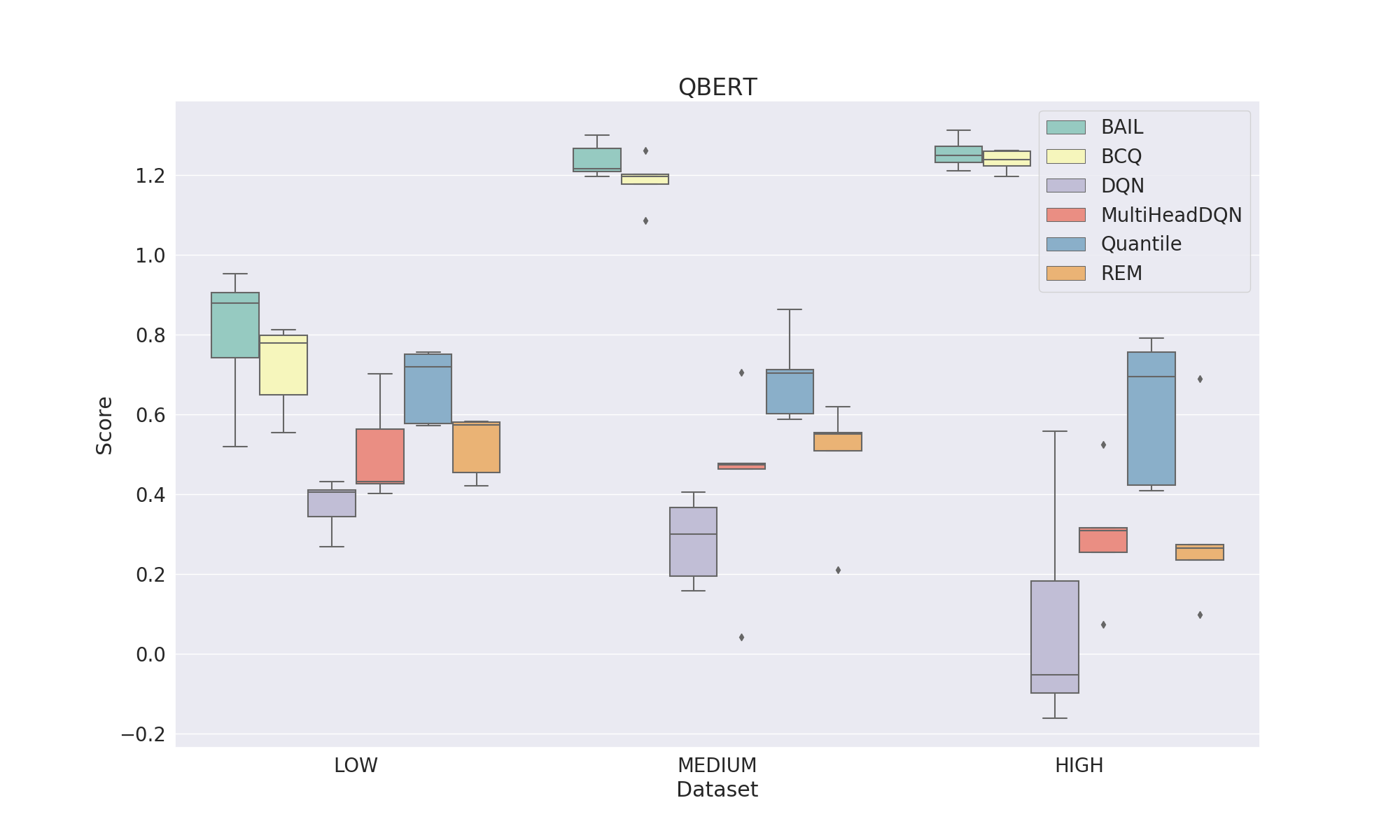}\\
        \vspace{0.01cm}
    \end{minipage}%
}%
\subfigure[StarGunner]{
    \begin{minipage}[t]{0.20\linewidth}
        \centering
        \includegraphics[width=1.55in]{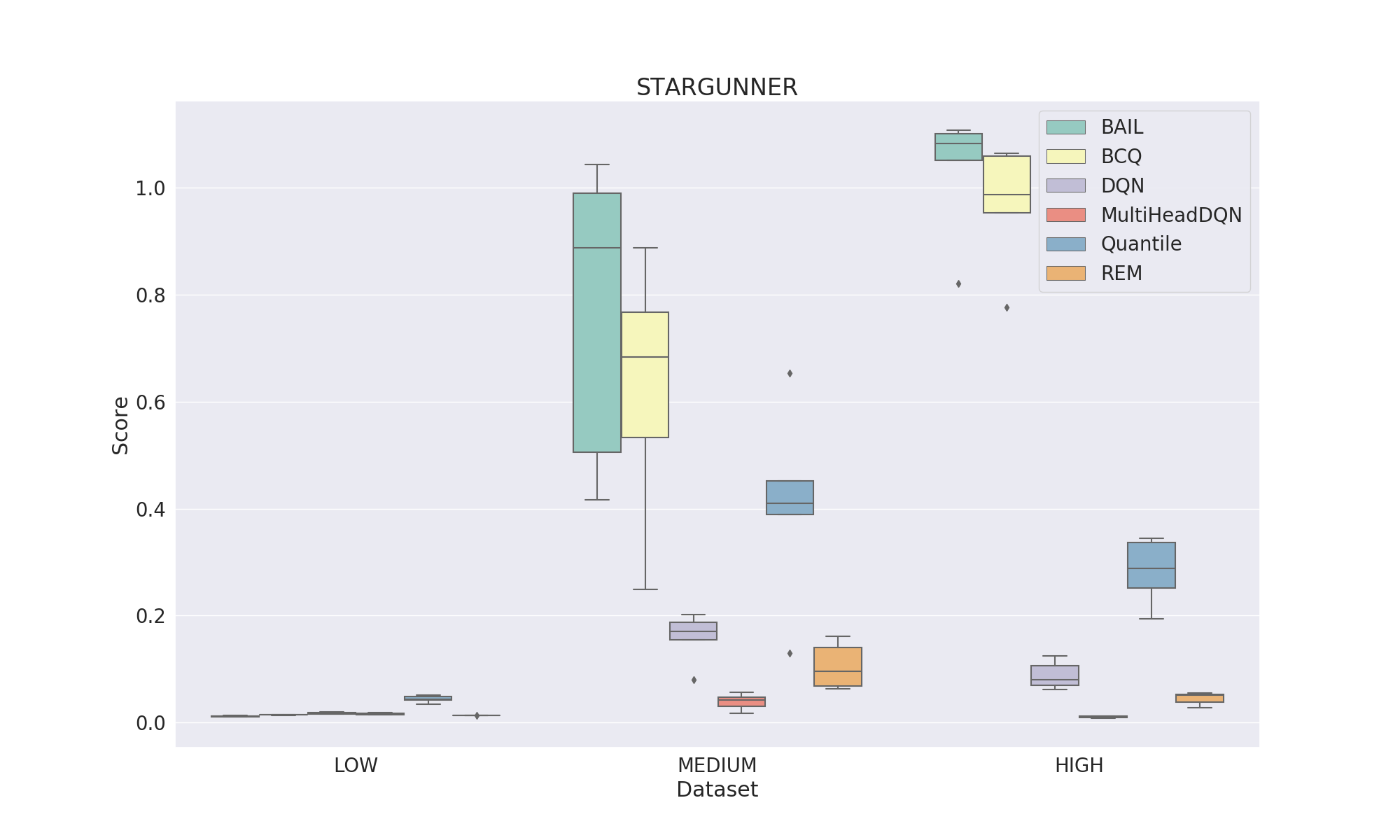}\\
        \vspace{0.01cm}
    \end{minipage}%
}%
\centering
\vspace{-0.4cm}
\caption{\textbf{Comparison for Different Dataset Qualities.} The mean score of top $20$ iterations are used for fair comparison. The boxplot indicates scenarios of $5$ different random seeds.}
\vspace{-0.4cm}


\label{fig:SOTA algo on different dataset qualities}
\end{figure*}

\begin{figure*}[!htb]
\centering

\subfigure[Alien]{
    \begin{minipage}[t]{0.20\linewidth}
        \centering
        \includegraphics[width=1.55in]{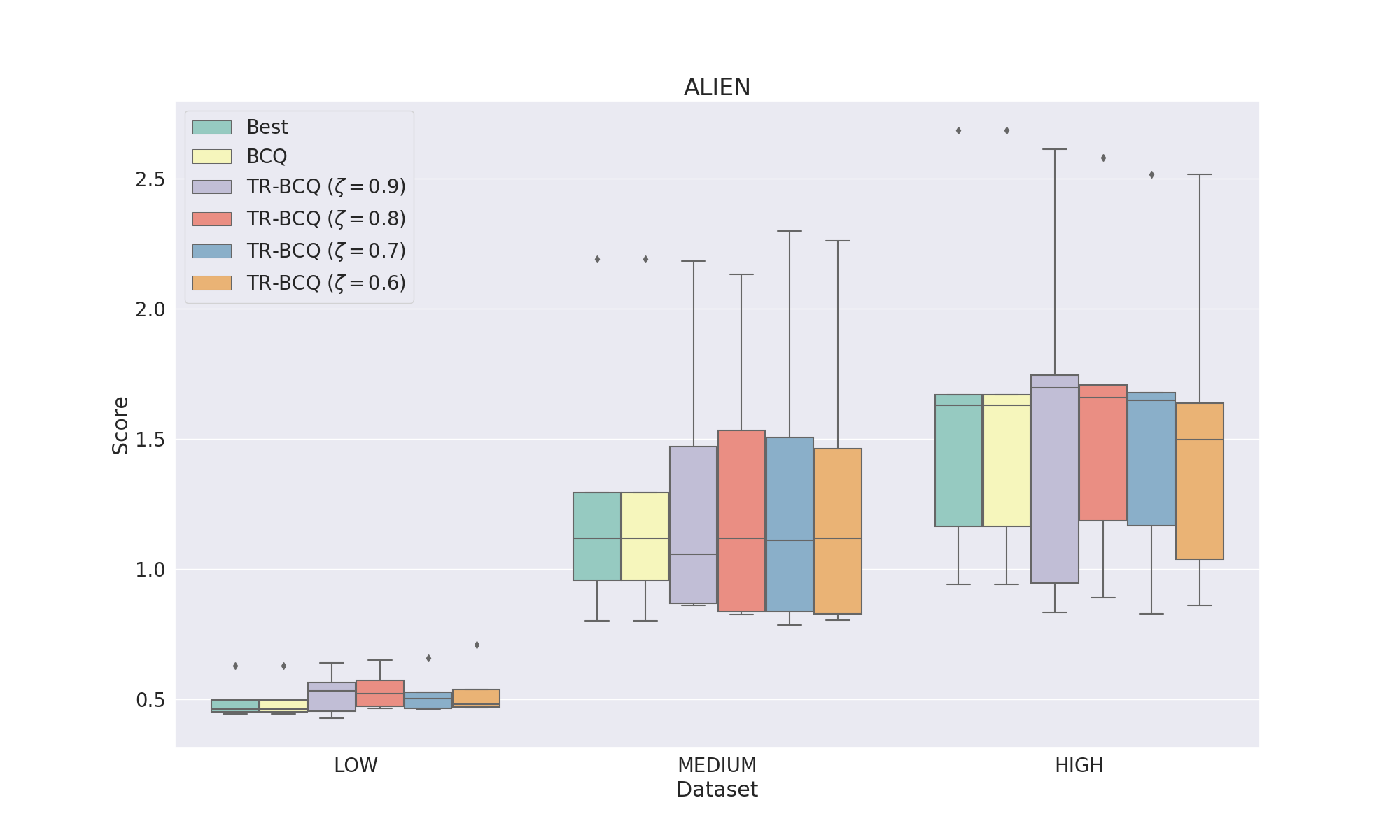}\\
        \vspace{0.01cm}
    \end{minipage}%
}%
\subfigure[Amidar]{
    \begin{minipage}[t]{0.20\linewidth}
        \centering
        \includegraphics[width=1.55in]{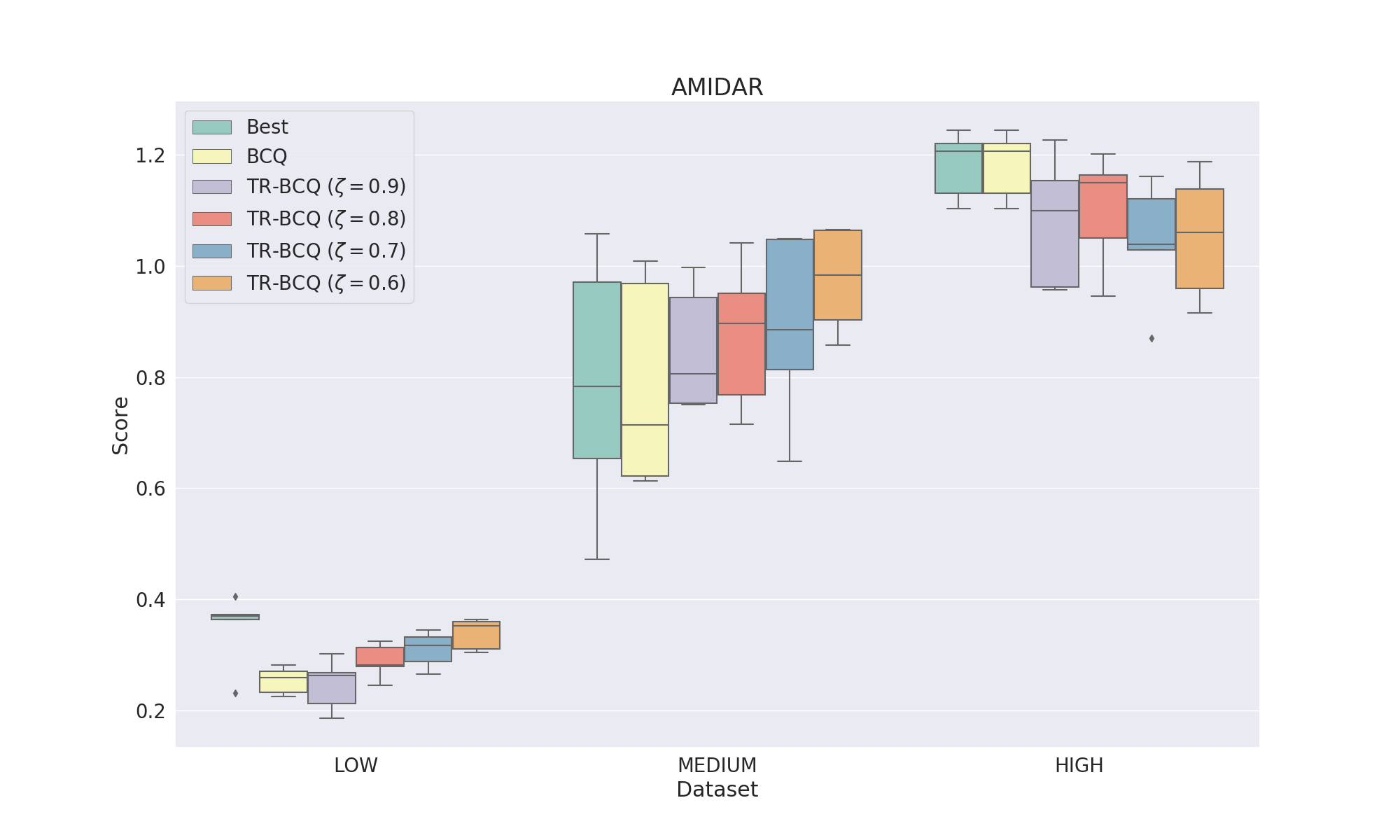}\\
        \vspace{0.01cm}
    \end{minipage}%
}%
\subfigure[Atlantis]{
    \begin{minipage}[t]{0.20\linewidth}
        \centering
        \includegraphics[width=1.55in]{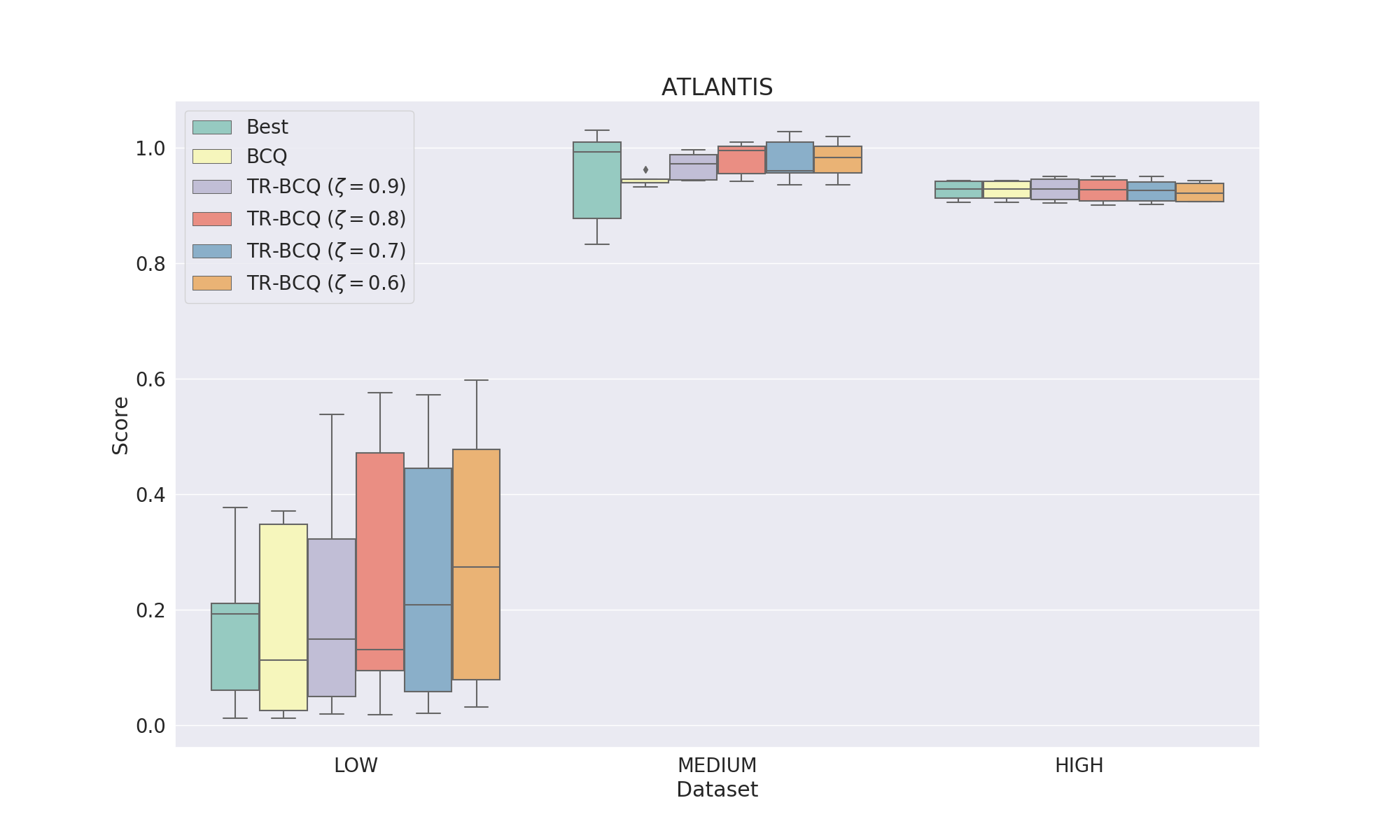}\\
        \vspace{0.01cm}
    \end{minipage}%
}%
\subfigure[Boxing]{
    \begin{minipage}[t]{0.20\linewidth}
        \centering
        \includegraphics[width=1.55in]{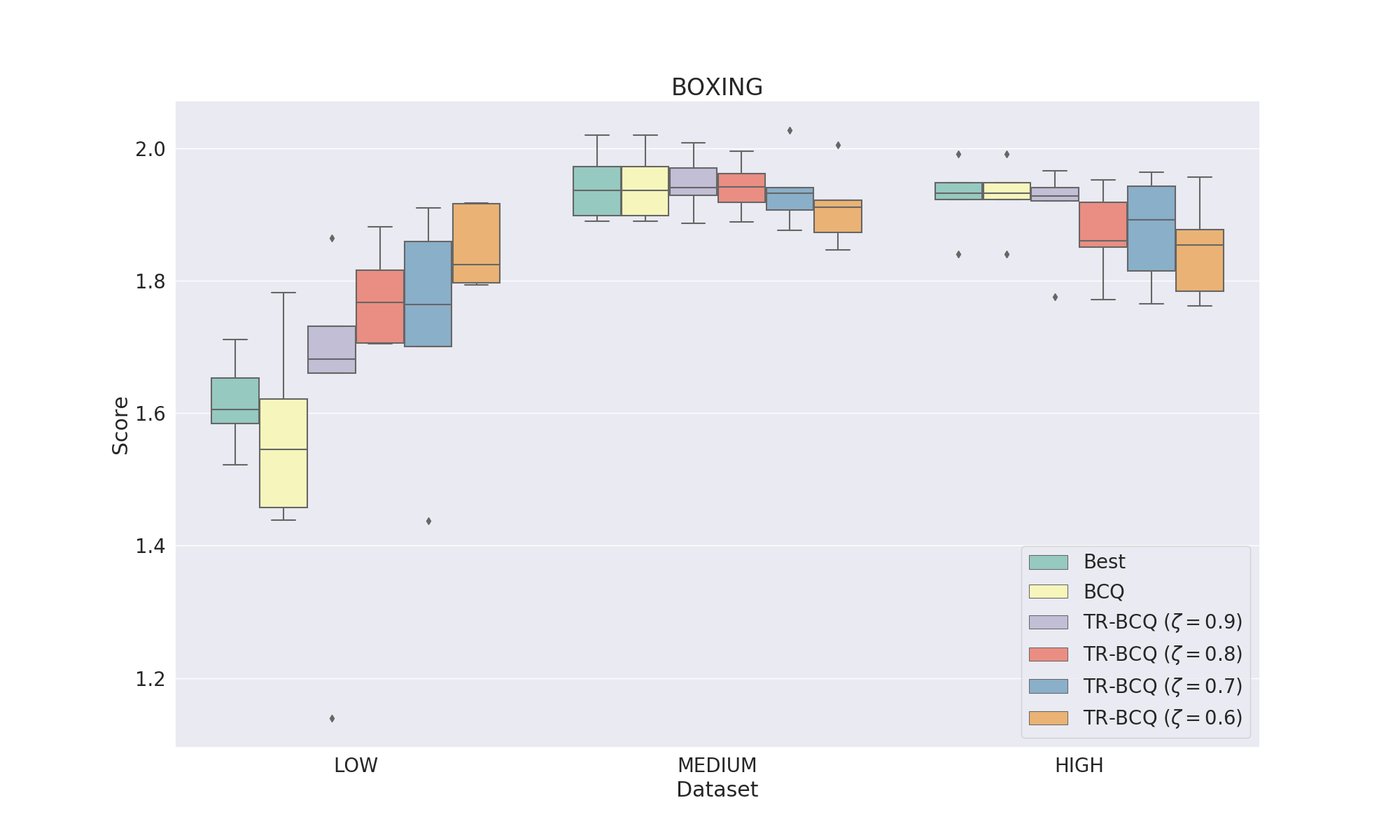}\\
        \vspace{0.01cm}
    \end{minipage}%
}%
\subfigure[Kangaroo]{
    \begin{minipage}[t]{0.20\linewidth}
        \centering
        \includegraphics[width=1.55in]{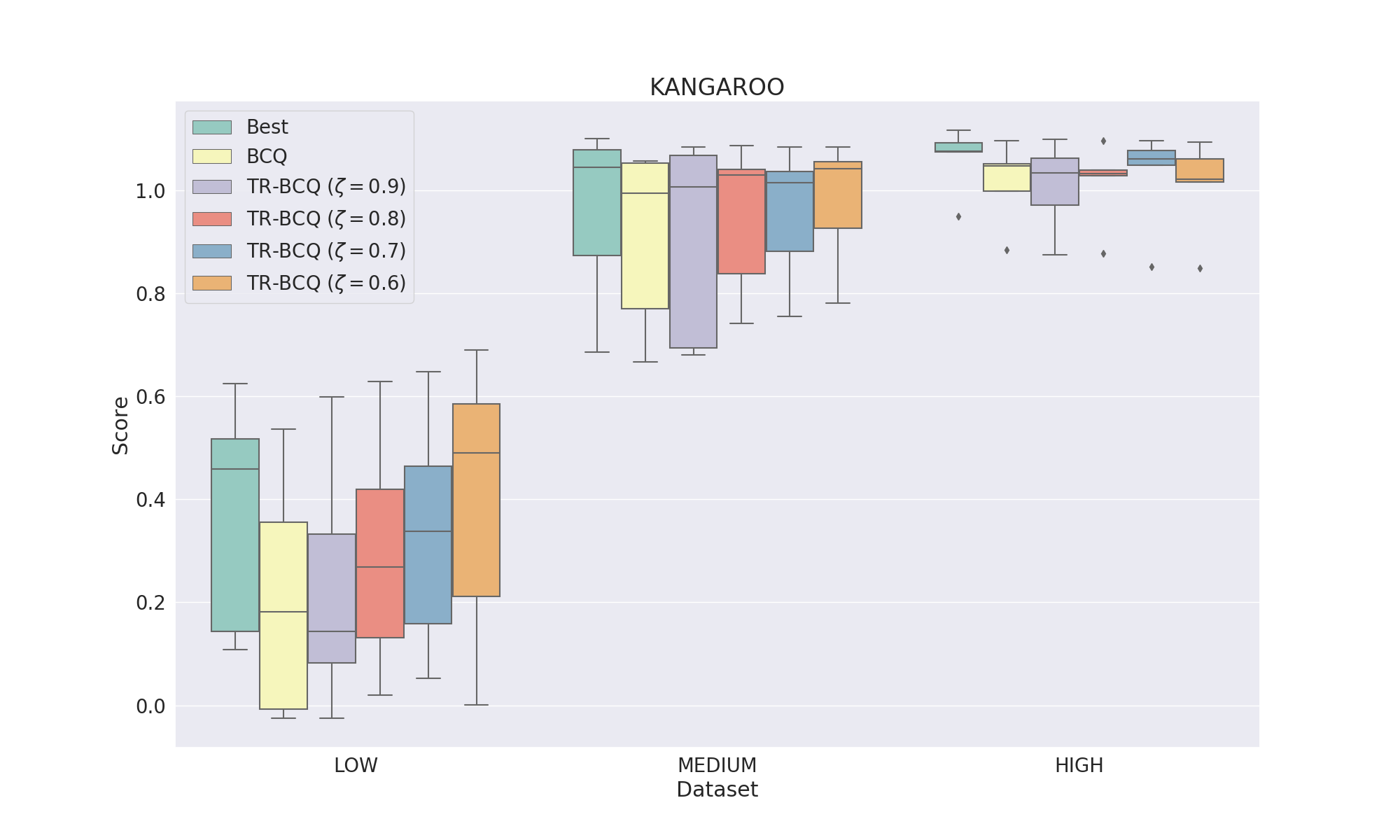}\\
        \vspace{0.01cm}
    \end{minipage}%
}%

\vspace{-0.4cm}
\subfigure[Krull]{
    \begin{minipage}[t]{0.20\linewidth}
        \centering
        \includegraphics[width=1.55in]{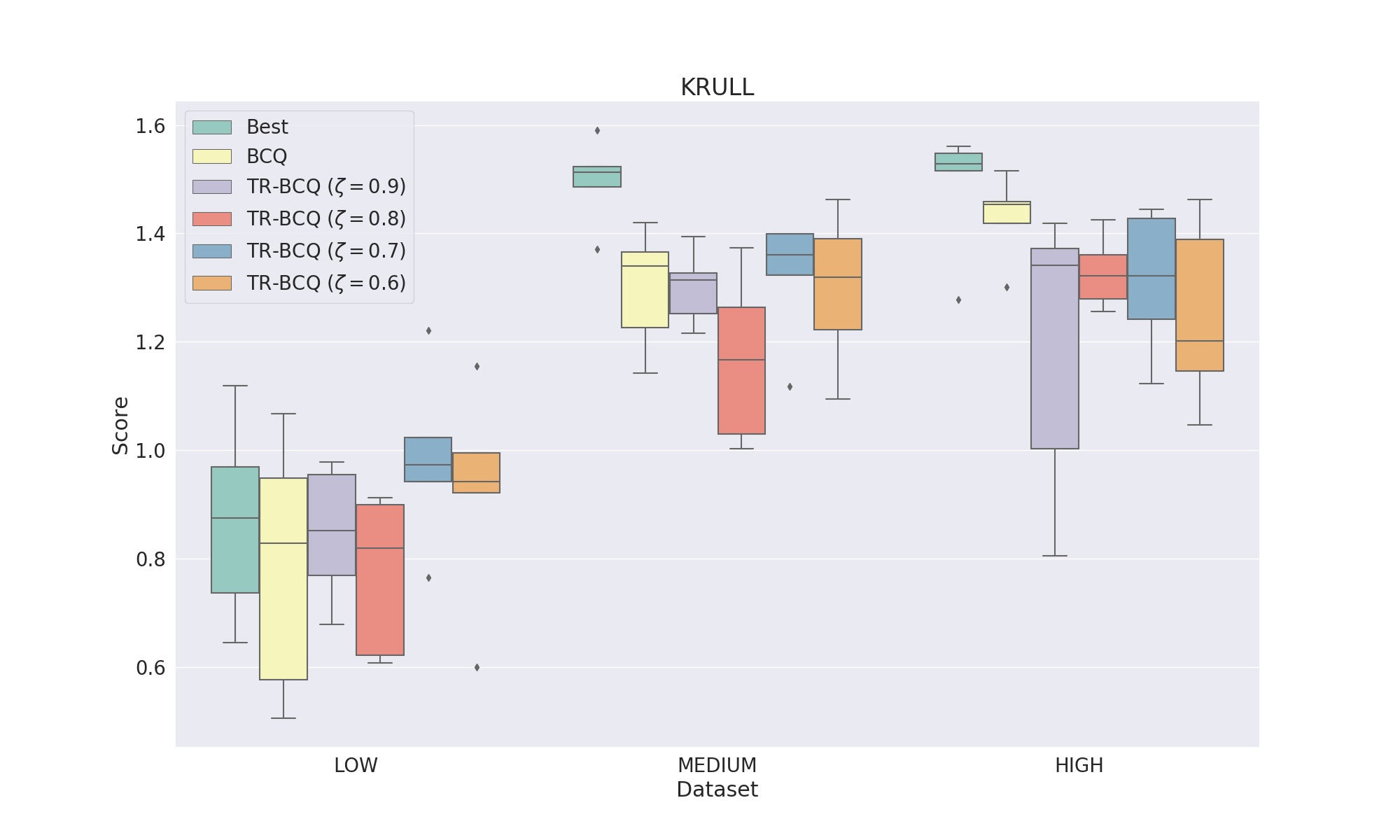}\\
        \vspace{0.01cm}
    \end{minipage}%
}%
\subfigure[Phoenix]{
    \begin{minipage}[t]{0.20\linewidth}
        \centering
        \includegraphics[width=1.55in]{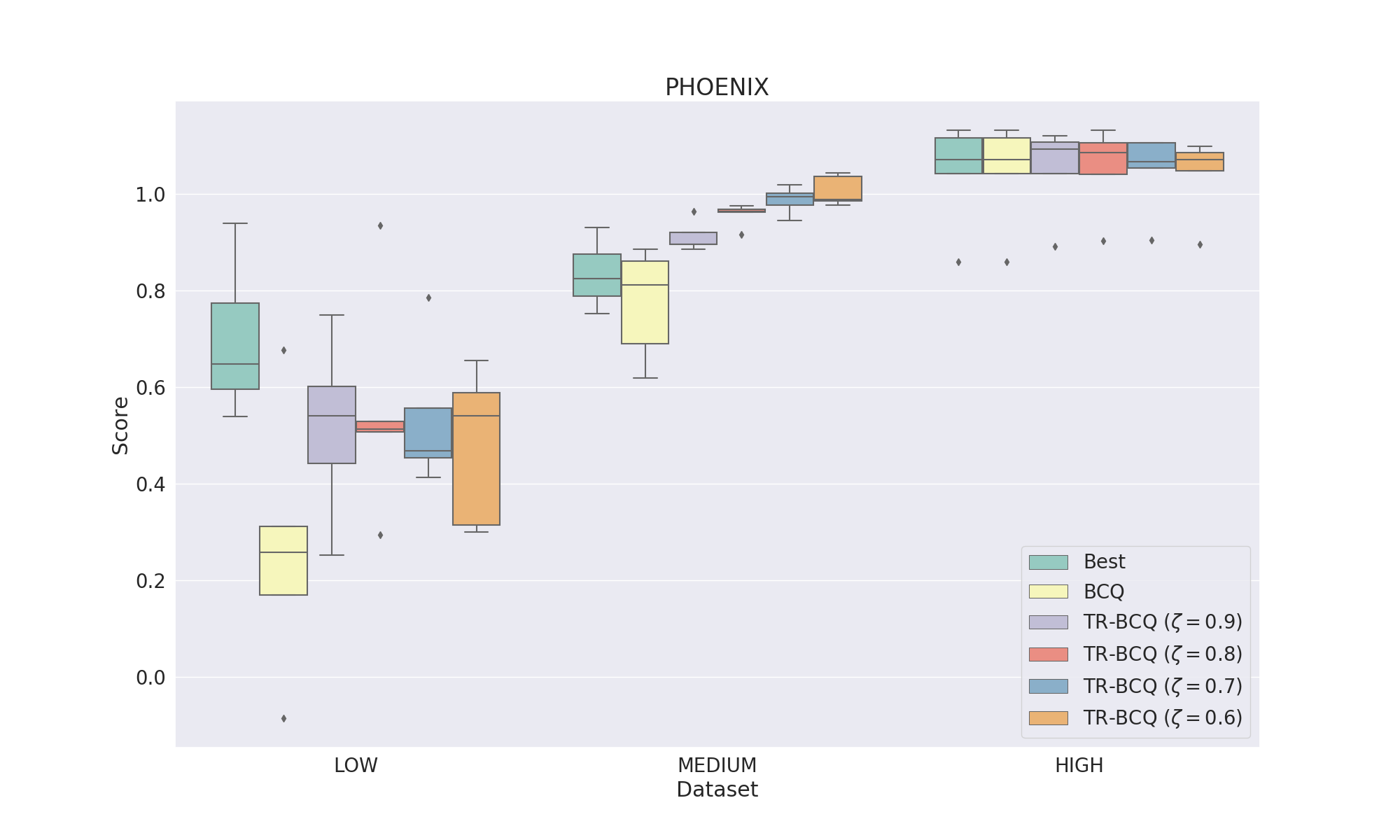}\\
        \vspace{0.01cm}
    \end{minipage}%
}%
\subfigure[Pong]{
    \begin{minipage}[t]{0.20\linewidth}
        \centering
        \includegraphics[width=1.55in]{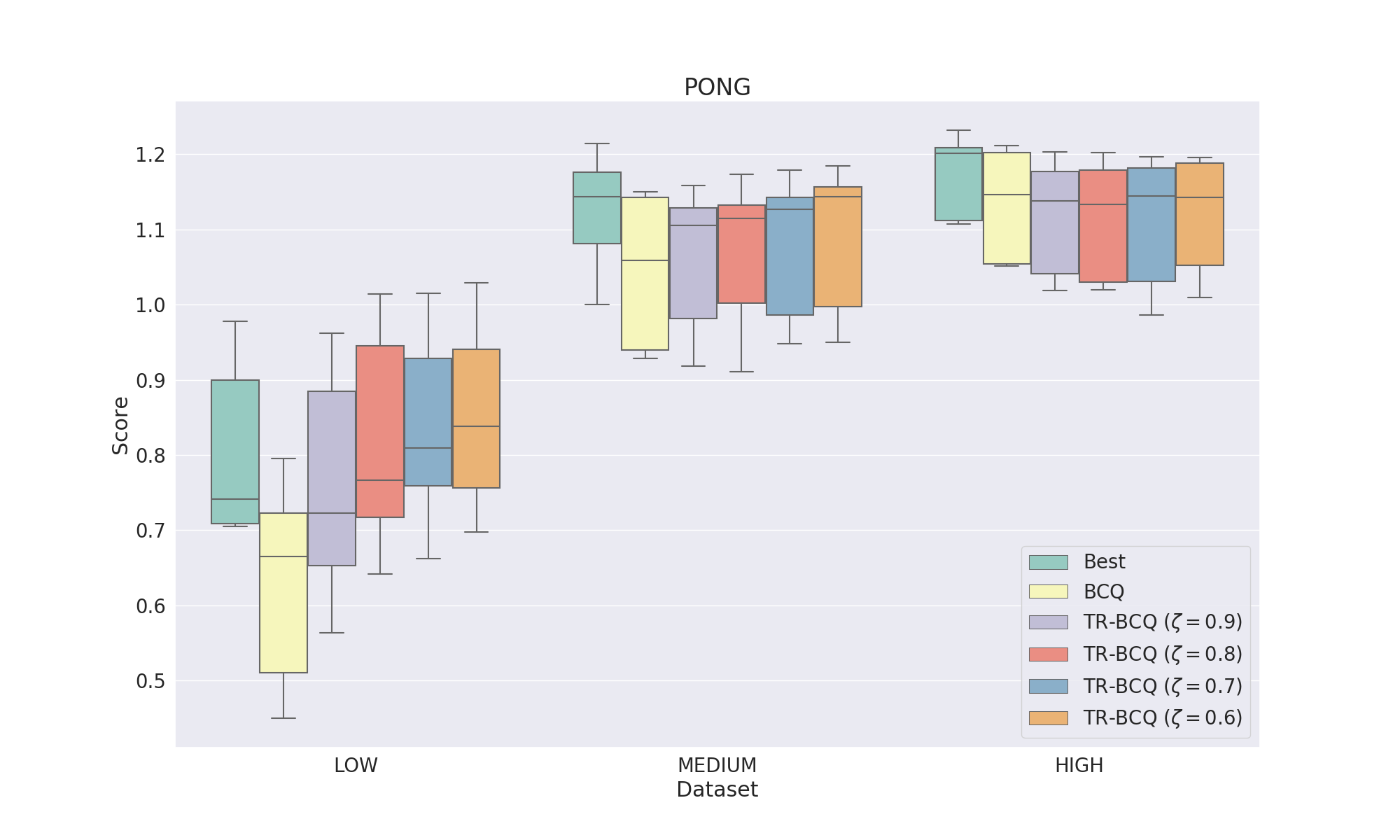}\\
        \vspace{0.01cm}
    \end{minipage}%
}%
\subfigure[Qbert]{
    \begin{minipage}[t]{0.20\linewidth}
        \centering
        \includegraphics[width=1.55in]{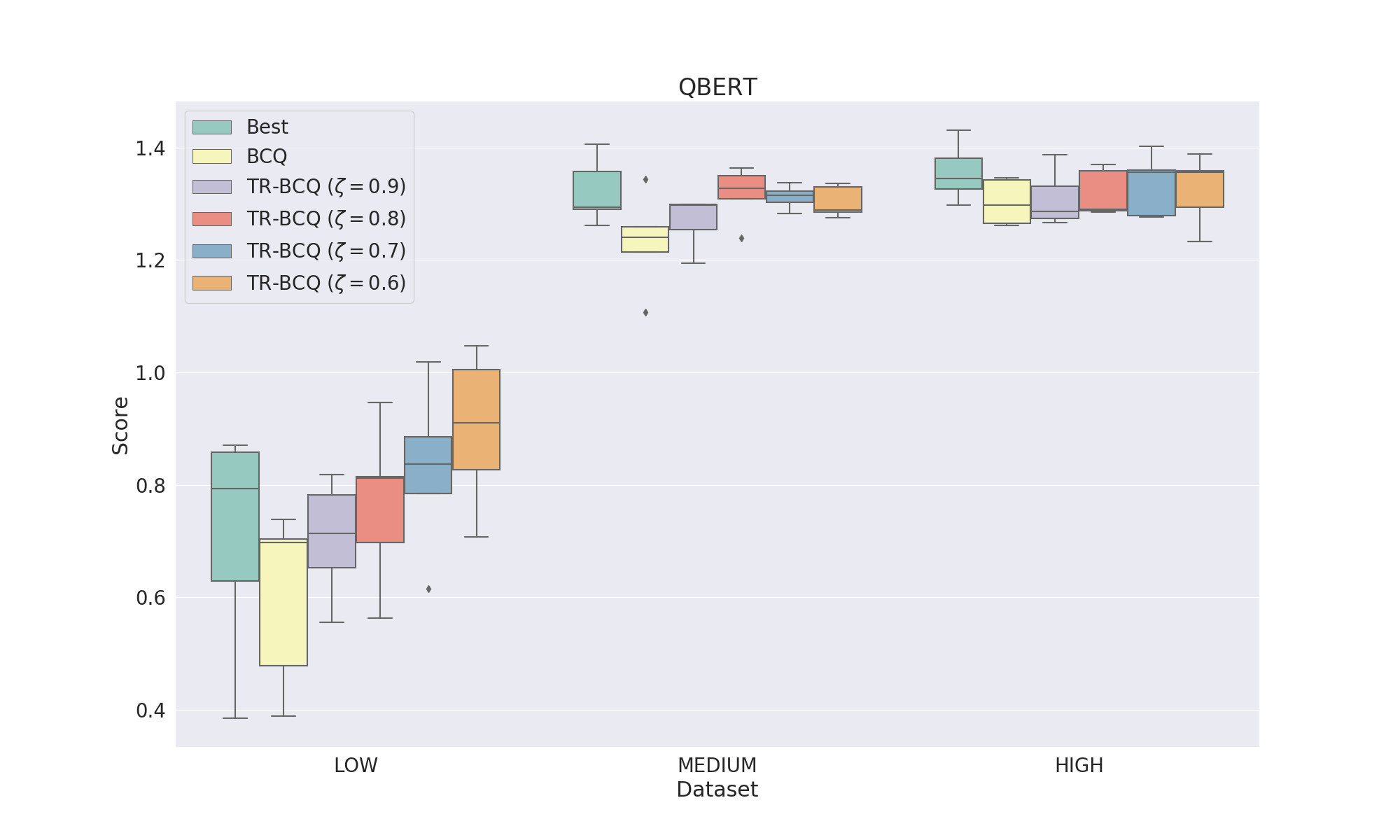}\\
        \vspace{0.01cm}
    \end{minipage}%
}%
\subfigure[StarGunner]{
    \begin{minipage}[t]{0.20\linewidth}
        \centering
        \includegraphics[width=1.55in]{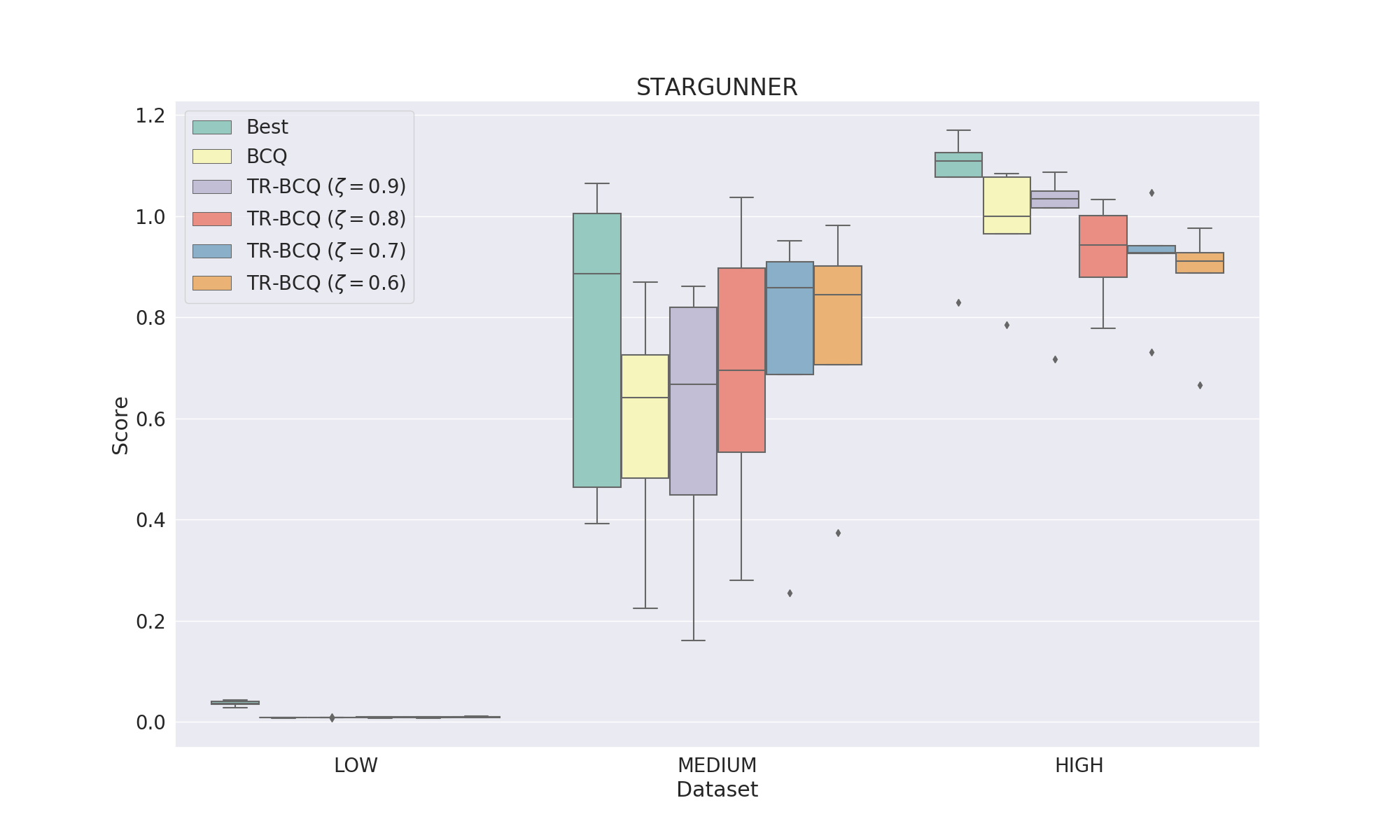}\\
        \vspace{0.01cm}
    \end{minipage}%
}%
\vspace{-0.4cm}
\centering
\caption{\textbf{\color{black}{Performance of TR-BCQ and SOTA algorithms on datasets of different qualities}} In low, meidum, and high dataset, TR-BCQ performs best on $70\%$, $50\%$, and $30\%$ of the games respectively. 
}
\vspace{-0.4cm}
\label{fig: Comparison between TRCQ and SOTA algo}

\end{figure*}

\section{Experiments on Atari 2600 Games}

\subsection{Experiment setup}
\vspace{-0.2cm}


{\color{black}{
We generate the dataset through an online DQN agent from scratch in the Atari 2600 Games. Considering better understanding of readers and space limitation, we list experimental results towards ten randomly selected games in this section. {\color{black}{More results can refer to Appendix.}}
}}

{\color{black}{
\vspace{-0.2cm}
\subsection{The benchmark platform: RL easy go}
Fig. \ref{fig:datasets of tri-level quality} indicates the iterations of different episode returns during the whole training process, and how we divide the dataset into three subsets, named ``low'', ``medium'', and ``high'' according to their mean episode returns. We named our benchmark platform RL easy go (RLEG), which enables a lighter and faster evaluation of off line RL experiments. {\color{black}{Given Assumption \ref{assumption: agent's rationality} being satisfied, we are able to derive an equivalence between ``low return'' with ``high randomness''}}

{\color{black}{
The reasons why we do not directly quantify randomness is that since considering continuous state space, we are exposed with uncountable states. Thus, the results of performance would strongly rely on the granularity of discretized states. Although the more fine granularity is, the more accurate the result would be, and corresponding computational overhead would be undesirable.
}}

\subsection{Performance of existing offline RL algorithms on various datasets}
\vspace{-0.2cm}

{\color{black}{
In this section, we will show the experimental results of exploration and exploitation-tentative algorithms on ``low'', ``medium'' and ``high'' quality datasets.
}}


In Fig. \ref{fig:SOTA algo on different dataset qualities}, we can clearly distinguish the monotonically trend of performance, which are deceasing and increasing for exploration-tentative and exploitation-tentative algorithms respectively, along with dataset randomness (represented in the form of return). {\color{black}{Refer to the statistical comparison results in table \ref{table: SOTA algo on different dataset qualities} for a more clear insight. We notice that exploration-tentative algorithms perform better on ``low'' dataset, i.e., high randomness and low-quality dataset, which conforms to the analysis in section 2.3. }} {\color{black}{ Besides, exploitation-tentative algorithms outperform on ``high'' dataset than exploration-tentative candidates. On ``medium'' datasets, exploitation-tentative algorithms behaves best on half games. {\color{black}{Based on the above results, we would recommend giving top priority to exploitation-tentative algorithms on high-quality dataset, while trying both types of algorithms on low-quality dataset. }}Admittedly, defining the quality of an offline dataset is empirical, {\color{black}{our suggestion is to take both the mean episode return and the action distribution into accounts.}}
}}

{\color{black}{
\subsection{Comparison between TR-BCQ with the best of existing algorithms}
}}
\vspace{-0.2cm}


Fig. \ref{fig: Comparison between TRCQ and SOTA algo} shows the performance of TR-BCQ under different percentages of data selection in Phase 1, which is one of the most critical hyperparameters and can be set in some heuristic ways. As shown in Fig. \ref{fig: Comparison between TRCQ and SOTA algo}, it is not always better to train with as much data as possible.
{{\color{black}{By choosing an appropriate $\zeta$, {the proposed algorithm is able to achieve higher online performance.}}}
From Fig.  \ref{fig: Comparison between TRCQ and SOTA algo}, TR-BCQ outperforms original BCQ a lot in ``low'' dataset, while less in ``high'' dataset.
{\color{black}
{
This implies that for dataset with lower quality, TR-BCQ is an indeed method with controllable extrapolation error, even if the number of data sets is greatly reduced. 
On the other hand, with the {\color{black}{mean episode return}} increasing along with the dataset quality, {\color{black}{the extrapolation error is of greater importance in dataset with higher quality.
}
}
}}

}

Overall, {\color{black}{we recommend TR-BCQ for ``low'' and ``medium'' dataset and any  exploitation-tentative or simply imitation-based algorithms for ``high'' dataset.}}

}}

\section{Conclusion and Future Work}

{\color{black}{
In this paper, we start with proposing a two-fold taxonomy for existing offline RL methodologies, i.e., exploration and exploitation-tentative algorithms. Then, with the help of derived upper bound of extrapolation error, we explore and prove the dependence of algorithm performance on dataset, especially for the action distribution for each state. From such a dataset perspective, although BCQ is provably better than the other, we identify its weak performance on dataset of low mean episode returns. Accordingly, we propose a modified BCQ based on a top return-based data selection mechanism. Our experimental results indicates that our algorithm could reach the best performance on various datasets. At last, a benchmark platform is created on the Atari domain, where we further open-source all datasets and checkpoints as a fair and comprehensive environment for competition between offline RL techniques.

Along with our top return-based data selection, some tuples are discarded and therefore is not used for constructing the underlying MDP. {How to make full use of these data would be worth to investigated}. Besides, for generalizing the proposed algorithms, we will extend it to the case of continuous state space and action space.
}}

\nocite{langley00}

\bibliography{TRCQ}
\bibliographystyle{icml2021}

\onecolumn
\appendix
\section{Introduction of open-source benchmark: RL easy go (RLEG)}
	
	\subsection{Workflow}
	The open sourced RLEG is a benchmark platform for offline reinforcement learning reproduction and evaluation, which consists of the following parts:
	
	\begin{itemize}
	\item Codes: We integrate all the source code of existing SOTA algorithms including BCQ, REM, Quantile, DQN, MultiHeadDQN, and BAIL into two frameworks based on our two-fold taxonomy (i.e., exploration- and exploitation-tentative). Besides, we open-source the source code of TR-BCQ under exploitation-tentative framework. We appreciate the open-source code of REM, BCQ, and BAIL. We build our code based on their work.
	
	\item Datasets:
	(i) All the checkpoints generated during the experiments;
	(ii) The tuples of the form $(s, a, r, s', a', r', t)$ for low, medium, high-quality dataset.
	(iii) The mean episode return logs for data generation process and offline reinforcement learning process on the aforementioned algorithms.  Researchers can check their own off line reinforcement learning algorithms on various dataset directly without running the costly data generation process.
	
	\end{itemize}
	

	{\color{black}{						
\section{Missing proofs}
}}
\subsection{Proof of Proposition 1}

{\color{black}{
First, we express the extrapolation error as:
}}

    \begin{equation}
        \begin{aligned}
            \epsilon_{s, a} &= Q_{\pi}^{MDP_1}(s, a) - Q_{\pi}^{MDP_2}(s, a) \\
                &= \sum_{s'}[(p_1(s'|s, a) - p_2(s'|s, a)]\color{black}{r(s,a,s')} + \sum_{s'}p_1(s'|s, a) \gamma \sum_{a'} \pi(a'|s')(Q_1(s', a') - Q_2(s', a'))
                + \\
                &\quad \sum_{s'}(p_1(s'|s, a) - p_2(s'|s, a)) \gamma \sum_{a'} \pi(a'|s')\color{black}{Q_2(s', a')} \\
                &\leq {\color{black}{R_{max}}}\sum_{s'} [(p_1(s'|s, a) - p_2(s'|s, a)] +  \sum_{s'}p_1(s'|s, a) \gamma \sum_{a'} \pi(a'|s')(Q_1(s', a') - Q_2(s', a'))
                + \\
                &\quad \sum_{s'}(p_1(s'|s, a) - p_2(s'|s, a)) \gamma \sum_{a'} \pi(a'|s')\color{black}{\frac{R_{max}}{({1 - \gamma})}}\\
                    &= \frac{R_{max}}{({1 - \gamma})}e(s, a)
                    + \gamma \sum_{s'}p_1(s'|s, a) \sum_{a'} \pi(a'|s')\frac{R_{max}}{({1 - \gamma})}e(s', a')
                    +...
                    +\\
                    &\quad  {\gamma}^{(n)} \sum_{s'}p_1(s'|s, a) \sum_{a'} \pi(a'|s') \sum_{s''} ... \sum_{s^{(n)}}p_1(s^{(n)}|s^{(n-1)}, a^{(n-1)}) \sum_{a^{(n)}} \pi(a^{(n)}|s^{(n)})
                    \frac{R_{max}}{({1 - \gamma})}e(s^{(n)}, a^{(n)})
                    +\\
                    &\quad ...
                \end{aligned}
            \end{equation}

{\color{black}{
    Given Lemma $2.2$ and Assumption $2.1$ to be satisfied,  equation (1) could be readily derived.
}}

\subsection{Proof of Theorem 1}
        
        The main problem we focus now turns to be the following expression:
        
        $$\sum_{a} \pi(a|s)\frac{R_{max}}{({1 - \gamma})}
        N^{-1/2}(\pi_{b}{(a|s)})^{-1/2}, {\forall}{s }\in \mathcal{S}$$
        
        We need to prove that no matter what distribution $\pi(a|s)$ is, the expression reaches its minimum when $\pi_{b}{(a|s)}$ is uniform. Also, the more even the distribution $\pi_{b}{(a|s)}$ is, the less expression value would be.
                
                The result for $\pi_b$ when $|\mathcal{A}|=2$ will be given firstly. Then the generalized proof will be presented.
                
                Case: $n = 2$
                \begin{equation}
                    \begin{aligned}
                        & \min \quad \mathbb{E}{[\sum_{a\in\mathcal{A}}{\frac{p_i}{\sqrt{q_i}}}]} \\
                        & \quad s.t. \quad q_1 + q_2 = 1
                    \end{aligned}
                \end{equation}
                
                To solve this problem, 
                \begin{equation}
                    \begin{aligned}
                        \label{Case for Theorem1}
                        q &=
                        \arg \min \frac{1}{(n-1)!}|_{n=2} \int_{0}^{1} 
                        \frac{p_1}{\sqrt{q_1}} + \frac{1 - p_1}{\sqrt{q_2}}
                        dp_1 \\
                        &=
                        \arg \min \frac{1}{2\sqrt{q1}} + \frac{1}{2\sqrt{q2}}
                        \\
                        & \quad s.t. \quad q_1 + q_2 = 1
                    \end{aligned}
                \end{equation}
                
                The solution is $q_1 = q_2 = \frac{1}{2}$ for Equation \ref{Case for Theorem1}.
                

                
                
                Consider the most general case as follow:
        \begin{equation}
            \label{eq: case n = n}
            \begin{aligned}
                \mathbb{E}(\pi | \pi_{b}) 
                &\quad = \frac{1}{(n-1)!} \int_{0}^{1} dp_1 \int_{0}^{1-p_1} dp_2 ... \int_{0}^{1 - p_1 - p_2 - ... -p_{n-2}} \sum_{i=1}^{n-1}\frac{p_i}{\sqrt{q_i}} + \frac{p_n}{\sqrt{q_{n}}} dp_{n-1} \\
                &\quad = \frac{1}{(n-1)!} \int_{0}^{1} dp_1 \int_{0}^{1-p_1} dp_2 ... \int_{0}^{1 - p_1 - p_2 - ... -p_{n-2}} \sum_{i=1}^{n-1}\frac{p_i}{\sqrt{q_i}} + \frac{1 - p_1 - p_2 - ... - p_{n-1}}{\sqrt{q_{n}}} dp_{n-1}
            \end{aligned}
        \end{equation}
        
        We rewrite the above equation as
        \begin{equation}
            \label{eq: rewrite case n = n}
            \mathbb{E}(\pi | \pi_{b}) = \sum_{i=1}^{n}\phi(p_i)\frac{1}{\sqrt{q_{i}}}
        \end{equation}    
        
        where $\phi(p_i) = \frac{1}{(n-1)!} \int_{0}^{1}dp_{1} \int_{0}^{1-p_{1}}dp_{2} \cdots \int_{0}^{1 - p_{1} - p_{2} - ... - p_{n-2}}p_{i} dp_{n-1}$.
        
        We notice that $\phi(p_i) = \int_{0}^{1}dp_1 \int_{0}^{1-p_1} \cdots \int_{0}^{1 - p_1 - p_2 - \cdots - p_{i-1}} \frac{(n-i)!}{(1 - p_1 - p_2 - \cdots - p_{i-1})^{n-i}} dp_i$, and for each $i \in [1, n]$, $\phi(p_i)$ are same. This fact is also intuitively understandable, since all $p_i$, $i=1,2, \cdots, n$ are independent and identically distributed (i.i.d), we have $\phi(p_1) = \phi(p_2) = \cdots = \phi(p_n) = \mathbb{E}(p_i)$.

\subsection{Proof of Proposition 2}

{\color{black}{
        In order to derive the upper bound of extrapolation error for BCQ, $\overline{\epsilon}_{s,a}$, we adopt zoom method using the result of Proposition $1$ and $N(s,a) \textgreater N\tau$ as
}}
        
        \begin{equation}
            \label{equation: upper bound of extrapolaton error for BCQ}
            \begin{aligned}
                {\epsilon}_{s, a} &= 
                (
                \frac{R_{max}}{({1 - \gamma})}e(s, a)
                + \gamma \sum_{s'}p_1(s'|s, a) \sum_{a'} \pi(a'|s')\frac{R_{max}}{({1 - \gamma})}e(s', a')
                + ...
                +\\
                &\quad  {\gamma}^{(n)} \sum_{s'}p_1(s'|s, a) \sum_{a'} \pi(a'|s') \sum_{s''} ... \sum_{s^{(n)}}p_1(s^{(n)}|s^{(n-1)}, a^{(n-1)}) \sum_{a^{(n)}} \pi(a^{(n)}|s^{(n)})
                \frac{R_{max}}{({1 - \gamma})}e(s^{(n)}, a^{(n)})
                + \\
                &\quad ... )\arrowvert_{N(s, a) > N\tau}  \\ 
                & \leq
                (2log(\frac{|\mathcal{S}||\mathcal{A}|2^{|\mathcal{S}|}}{\delta}))^{1/2}[ \frac{R_{max}}{({1 - \gamma})}
                {(N\tau)}^{-1/2}  +\gamma \sum_{s'}p_1(s'|s, a) \sum_{a'} \pi(a'|s')\frac{R_{max}}{({1 - \gamma})}
                {(N\tau)}^{-1/2}  + ... +\\
                &\quad {\gamma}^{(n)} \sum_{s'}p_1(s'|s, a) \sum_{a'} \pi(a'|s') \sum_{s''} ...  \sum_{s^{(n)}}p_1(s^{(n)}|s^{(n-1)}, a^{(n-1)}) \sum_{a^{(n)}} \pi(a^{(n)}|s^{(n)})
                \frac{R_{max}}{({1 - \gamma})}{(N\tau)}^{-1/2}
        +...]\\
                &= (2log(\frac{|\mathcal{S}||\mathcal{A}|2^{|\mathcal{S}|}}{\delta}))^{1/2} \frac{R_{max}}{({1 - \gamma})}[
                {(N\tau)}^{-1/2}  +\gamma
                {(N\tau)}^{-1/2}
                + ... + {\gamma}^{(n)}
                {(N\tau)}^{-1/2}
                +...]\\
                &= (2log(\frac{|\mathcal{S}||\mathcal{A}|2^{|\mathcal{S}|}}{\delta}))^{1/2} \frac{R_{max}}{({1 - \gamma})}{(N\tau)}^{-1/2}[1
                    +\gamma
                + ... + {\gamma}^{(n)}
                +...]
            \end{aligned}
        \end{equation}
        

\subsection{Proof of Theorem 2}
Given $\tau > \frac{1}{|A|}$, the extrapolation bound for BCQ satisfies that 

            \begin{equation}
            \label{A-7}
            \begin{aligned}
                \overline{{\epsilon}}_{s, a}
                &\leq (2log(\frac{|\mathcal{S}||\mathcal{A}|2^{|\mathcal{S}|}}{\delta}))^{1/2} \frac{R_{max}}{({1 - \gamma})}{(\frac{N}{|A|})}^{-1/2}[1
                    +\gamma
                + ... + {\gamma}^{(n)}
                +...]
            \end{aligned}
        \end{equation}
From Theorem 1, the minimum bound of exploration-tentative algorithm is achieved when $\pi_b$ is uniform, i.e., $\pi(\cdot|s) = \frac{1}{|A|}$. Thus, the minimal extrapolation bound for exploration-tentative algorithm is 

    \begin{equation}
    \label{A-8}
    \begin{aligned}
        \overline{\epsilon}_{s, a} &= (2log(\frac{|\mathcal{S}||\mathcal{A}|2^{|\mathcal{S}|}}{\delta}))^{1/2} \frac{R_{max}}{({1 - \gamma})}{N}^{-1/2} \\
        &[
        ({\frac{1}{|A|}})^{-1/2} +\gamma \sum_{s'}p_1(s'|s, a) \sum_{a'} \pi(a'|s')
                    ({\frac{1}{|A|}})^{-1/2} + ... + {\gamma}^{(n)} \sum_{s'}p_1(s'|s, a) \sum_{a'} \pi(a'|s') \sum_{s''} ...\\ 
                    & \sum_{s^{(n)}}p_1(s^{(n)}|s^{(n-1)}, a^{(n-1)}) \sum_{a^{(n)}} \pi(a^{(n)}|s^{(n)})({\frac{1}{|A|}})^{-1/2}+...]\\
    \end{aligned}
\end{equation}

Note that not all terms of equation \ref{A-8} exists in equation \ref{A-7}, because only $(s, a)$ pairs that satisfy $N(s, a) > \frac{1}{|A|}$ are selected in BCQ. Therefore, Theorem 2 holds.




\subsection{Proof of Proposition 5}

{\color{black}{
    In the first phase, because we process the data selection, for any $(s, a)$ pair under the batch constraint, $\hat{N}(s, a) \le N(s, a)$. Probabilistically, $\mathbb{E}{(\hat{N}(s, a))} = \mathbb{E}{(N(s, a))} \zeta $

    Thus, for every term in equation (\ref{equation: upper bound of extrapolaton error for BCQ}), the extrapolation error would increase after data selection, the expectation of which turns out to be
}}
\begin{equation}
        \label{equation: upper bound of extrapolaton error for TR-BCQ}
        \begin{aligned}
            &\mathbb{E}({{\hat{\overline{\epsilon}}}_{s, a}}) = \mathbb{E}({{\overline{\epsilon}}_{s, a}}){\zeta}^{-1/2}
        \end{aligned}
    \end{equation}

\section{Pseudocode of TR-BCQ}
Please refer Algorithm 1.

\begin{algorithm}[t]
    \caption{TR-BCQ Algorithm}
    \label{pseudocode: Top return constrained Q-learning Algorithm}
    
    {\color{black}{
            \textbf{Input:} Offline dataset tuples $(\mathcal{S}, \mathcal{A}, \mathcal{S}', \mathcal{R}, \mathcal{T}, \mathcal{G})$ , data selection percentile $\zeta$, and number of iterations $T$;
        }}

        \textbf{Output:} Policy $\pi$;
        
        {\color{black}{
                \textbf{Initialization:} Q-network $Q_\theta$, generative model $G_\omega$ and target network $Q_{\theta^{'}}$.

                \begin{center}
                    \textbf{Phase 1: Top Return-based Data selection}\\
                \end{center}
            }}
            
            
            
            
            
            
            
            
            {\color{black}{
                    \emph{a) Sort the tuples by $\mathcal{G}$.}

                    \emph{b) Select top $(1 - \zeta)$ percentage of tuples.}

                    \begin{center}
                        \textbf{Phase 2: tuple visitation constrained Q-learning}\\
                    \end{center}
                }}
                
                {\color{black}{
                        %
                        
                        
                        \textit{For} $t=1$ to $T$ \{
                    }}
                    
                    \emph{a) Selecting the max valued action with $Q_\theta$}
                    
                    $$a^{'} = \mathop{argmax}\limits_{a^{'}|G_{\omega}(a^{'}|s^{'})/max\hat{a}G_{\omega}(\hat{a}|s^{'}) > \tau}Q_{\theta}(s^{'},a^{'})$$
                    
                    \emph{b) Evaluating with  $Q_{\theta^{'}}$}
                    
                    $$\theta \leftarrow \mathop{argmin}\limits_{\theta}\sum_{\{s,a,r,s^{'}\}\in \{\mathcal{S}, \mathcal{A}, 
            \mathcal{R},
            \mathcal{S'}\}} \mathcal{L}(\theta)$$
                    
                    where $\mathcal{L}(\theta) = l_{\mathcal{K}}(r + \gamma Q_{\theta^{'}}(s^{'},a^{'}) - Q_{\theta}(s,a))$
                    
                    \emph{c) Behavioral cloning with  $G_{\omega}$}
                    
                        $$\omega \leftarrow \mathop{argmin}\limits_{\omega} - \sum\limits_{(s,a) \in \tau-\text{constrained dataset}} log G_{\omega}(s|a)$$
                    \} 

                        \emph{d) Update target network $Q_{\theta^{'}}$ with $\theta^{'} \leftarrow \theta$}
                        
                        
\end{algorithm}

\section{Additional Experiments}
	
	
	

	\subsection{Data Generation}
\begin{figure*}[!htb]
		\centering
		\subfigure{
			\begin{minipage}[t]{0.166\linewidth}
				\centering
				\includegraphics[width=1.05in]{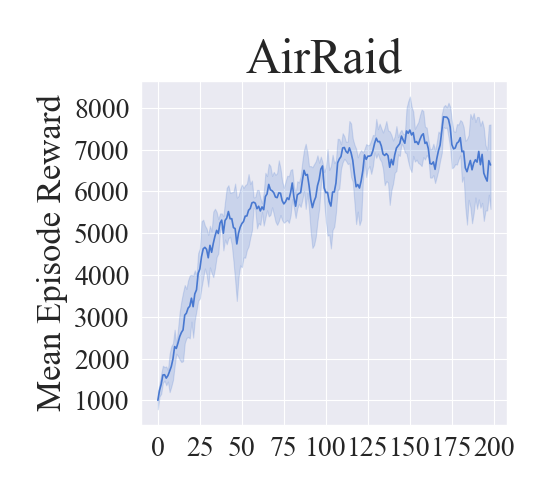}\\
			\end{minipage}%
		}%
		\subfigure{
			\begin{minipage}[t]{0.166\linewidth}
				\centering
				\includegraphics[width=1.05in]{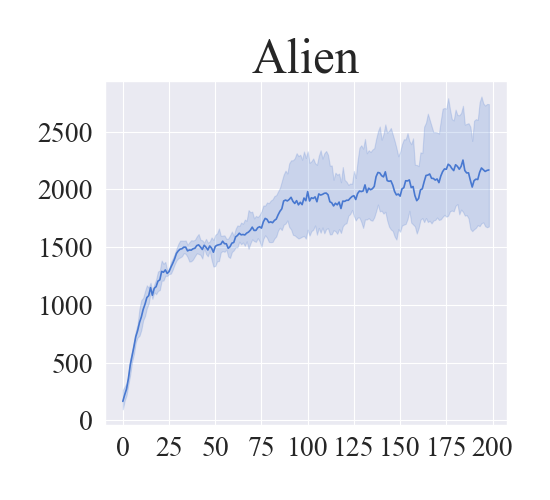}\\
			\end{minipage}%
		}%
		\subfigure{
			\begin{minipage}[t]{0.166\linewidth}
				\centering
				\includegraphics[width=1.05in]{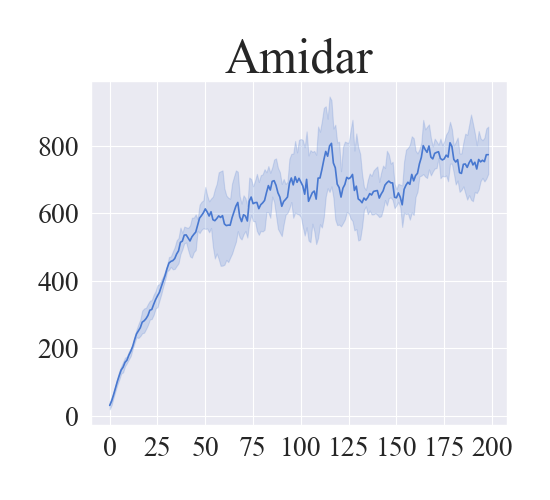}\\
			\end{minipage}%
		}%
		\subfigure{
			\begin{minipage}[t]{0.166\linewidth}
				\centering
				\includegraphics[width=1.05in]{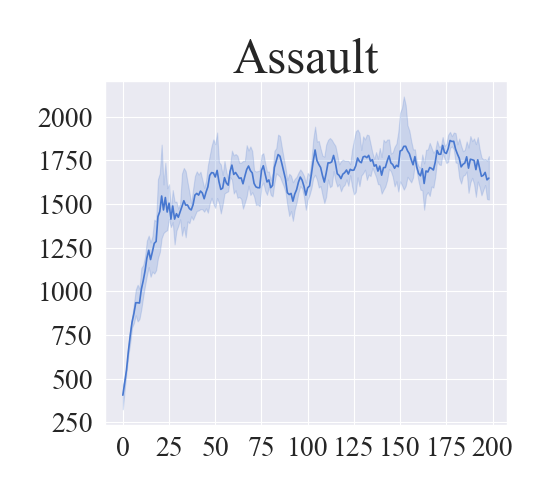}\\
			\end{minipage}%
		}%
		\subfigure{
			\begin{minipage}[t]{0.166\linewidth}
				\centering
				\includegraphics[width=1.05in]{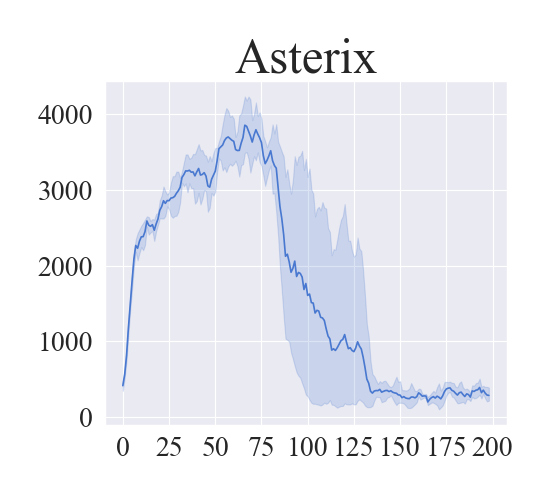}\\
			\end{minipage}%
		}%
		\subfigure{
			\begin{minipage}[t]{0.166\linewidth}
				\centering
				\includegraphics[width=1.05in]{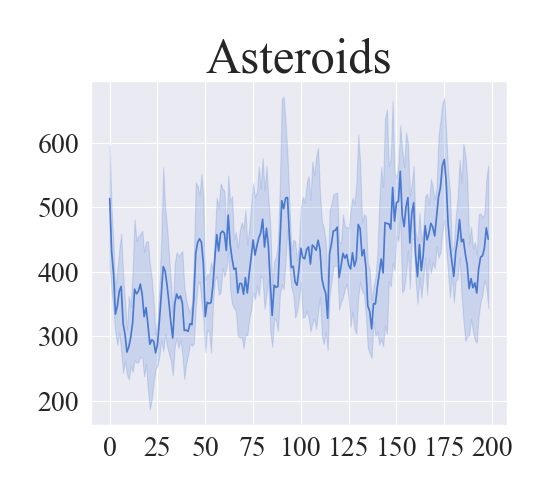}\\\
			\end{minipage}%
		}%
		\vspace{-1.0cm}
		
		\subfigure{
			\begin{minipage}[t]{0.166\linewidth}
				\centering
				\includegraphics[width=1.05in]{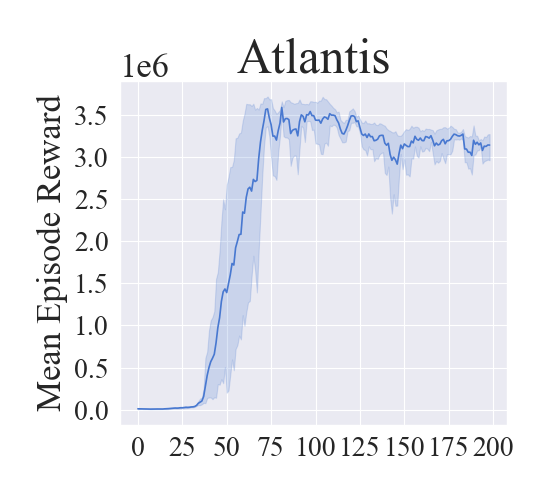}\\
			\end{minipage}%
		}%
		\subfigure{
			\begin{minipage}[t]{0.166\linewidth}
				\centering
				\includegraphics[width=1.05in]{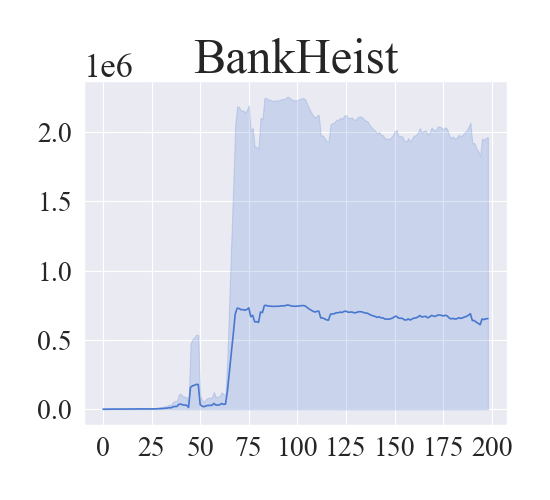}\\
			\end{minipage}%
		}%
		\subfigure{
			\begin{minipage}[t]{0.166\linewidth}
				\centering
				\includegraphics[width=1.05in]{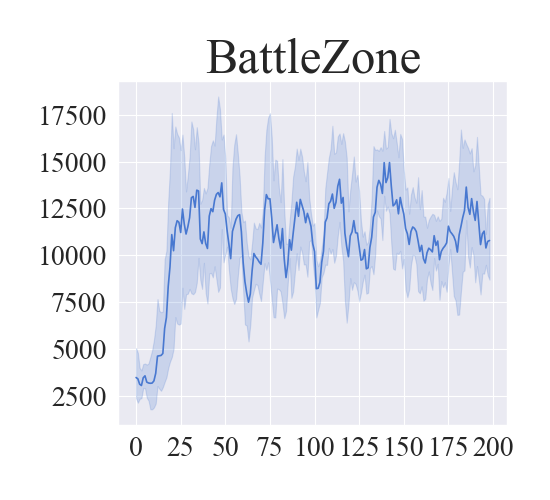}\\
			\end{minipage}%
		}%
		\subfigure{
			\begin{minipage}[t]{0.166\linewidth}
				\centering
				\includegraphics[width=1.05in]{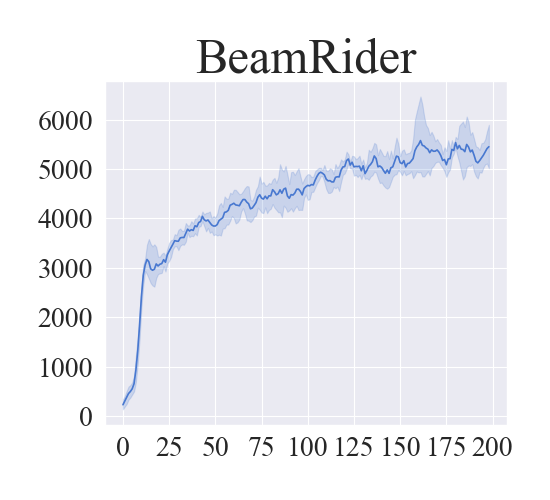}\\
			\end{minipage}%
		}%
		\subfigure{
			\begin{minipage}[t]{0.166\linewidth}
				\centering
				\includegraphics[width=1.05in]{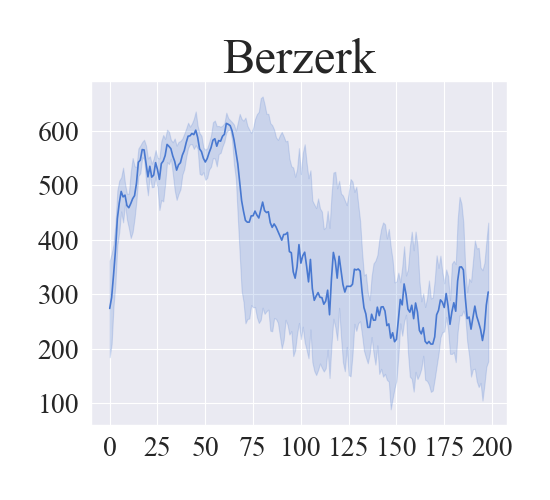}\\
			\end{minipage}%
		}%
		\subfigure{
			\begin{minipage}[t]{0.166\linewidth}
				\centering
				\includegraphics[width=1.05in]{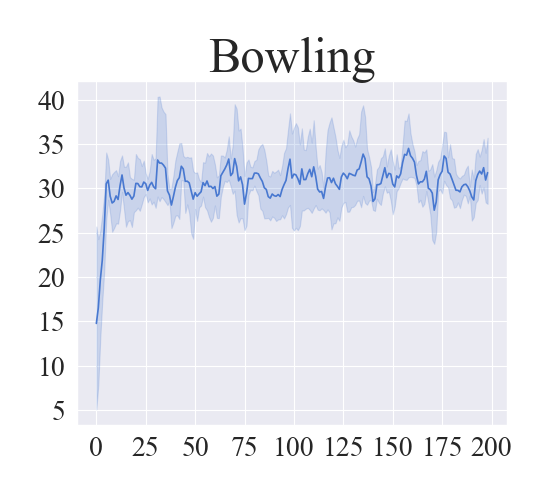}\\
			\end{minipage}%
		}%
		\vspace{-0.6cm}
		
		\subfigure{
			\begin{minipage}[t]{0.166\linewidth}
				\centering
				\includegraphics[width=1.05in]{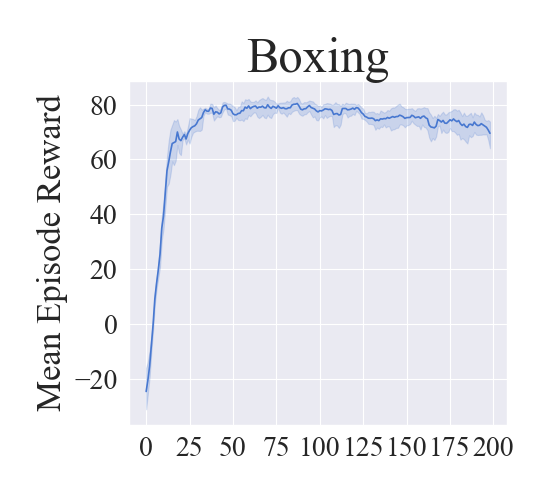}\\
			\end{minipage}%
		}%
		\subfigure{
			\begin{minipage}[t]{0.166\linewidth}
				\centering
				\includegraphics[width=1.05in]{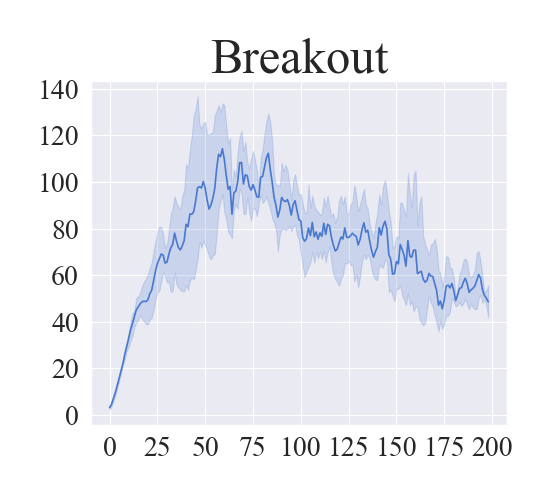}\\
			\end{minipage}%
		}%
		\subfigure{
			\begin{minipage}[t]{0.166\linewidth}
				\centering
				\includegraphics[width=1.05in]{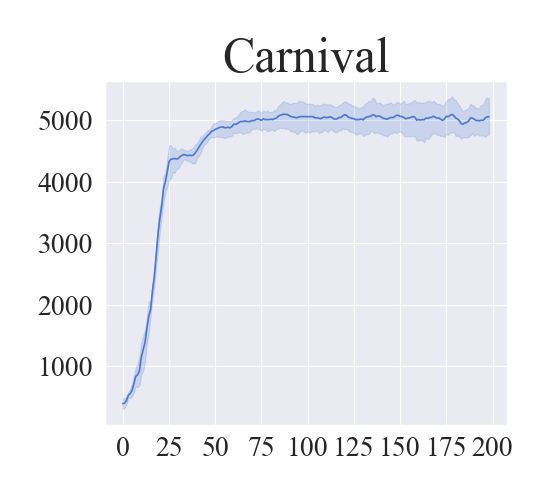}\\
			\end{minipage}%
		}%
		\subfigure{
			\begin{minipage}[t]{0.166\linewidth}
				\centering
				\includegraphics[width=1.05in]{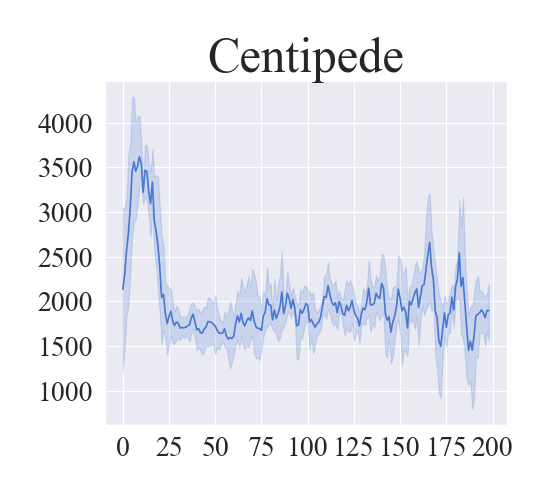}\\
			\end{minipage}%
		}%
		\subfigure{
			\begin{minipage}[t]{0.166\linewidth}
				\centering
				\includegraphics[width=1.05in]{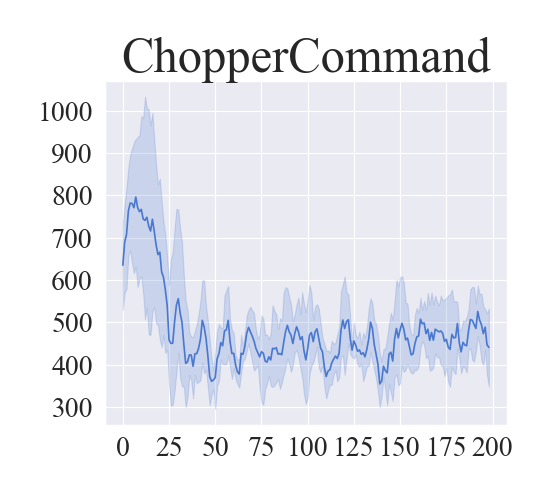}\\
			\end{minipage}%
		}%
		\subfigure{
			\begin{minipage}[t]{0.166\linewidth}
				\centering
				\includegraphics[width=1.05in]{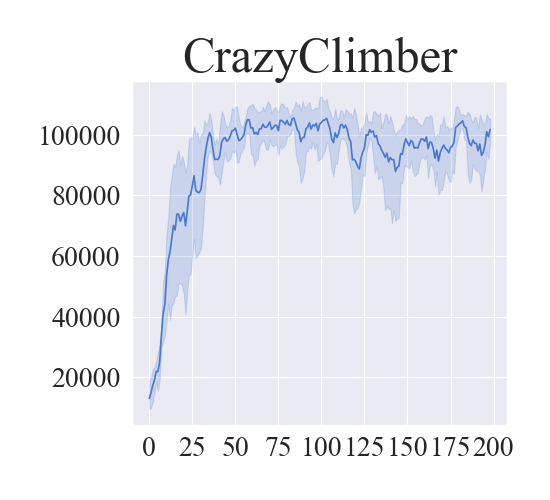}\\
			\end{minipage}%
		}%
		\vspace{-0.6cm}
		
		\subfigure{
			\begin{minipage}[t]{0.166\linewidth}
				\centering
				\includegraphics[width=1.05in]{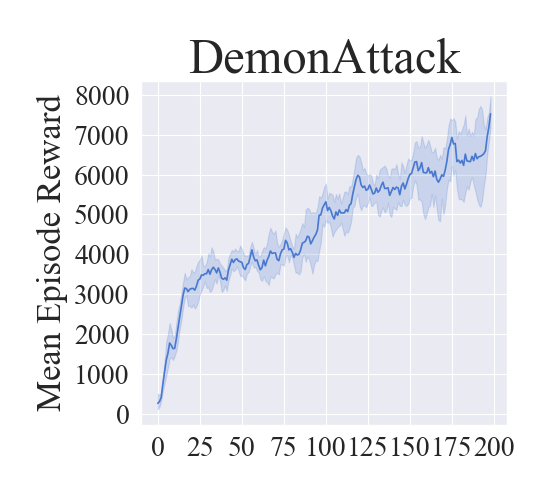}\\
			\end{minipage}%
		}%
		\subfigure{
			\begin{minipage}[t]{0.166\linewidth}
				\centering
				\includegraphics[width=1.05in]{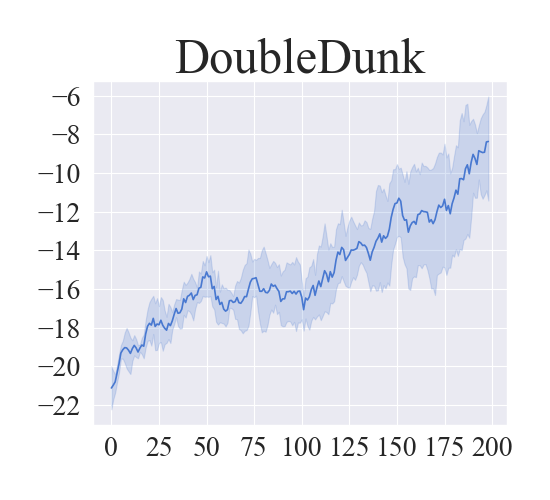}\\
			\end{minipage}%
		}%
		\subfigure{
			\begin{minipage}[t]{0.166\linewidth}
				\centering
				\includegraphics[width=1.05in]{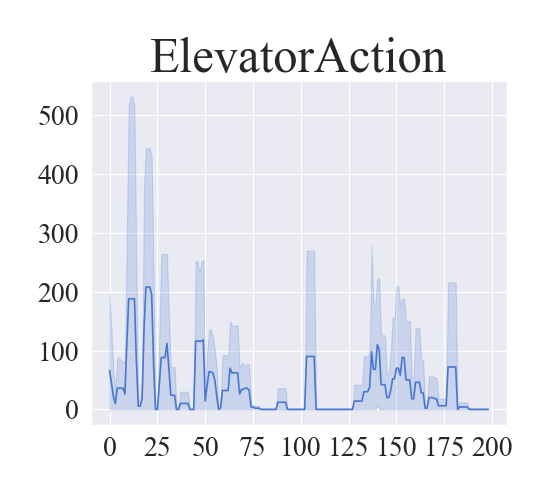}\\
			\end{minipage}%
		}%
		\subfigure{
			\begin{minipage}[t]{0.166\linewidth}
				\centering
				\includegraphics[width=1.05in]{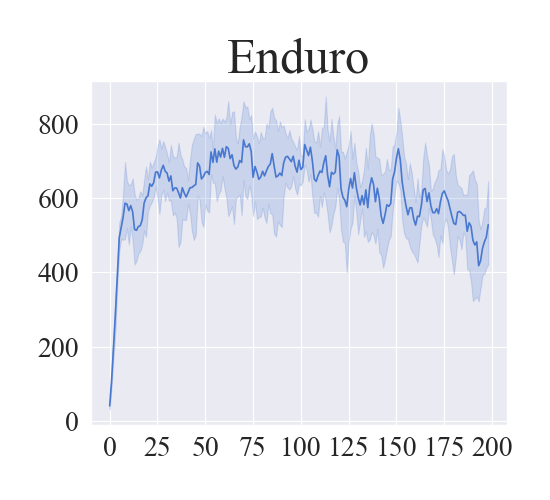}\\
			\end{minipage}%
		}%
		\subfigure{
			\begin{minipage}[t]{0.166\linewidth}
				\centering
				\includegraphics[width=1.05in]{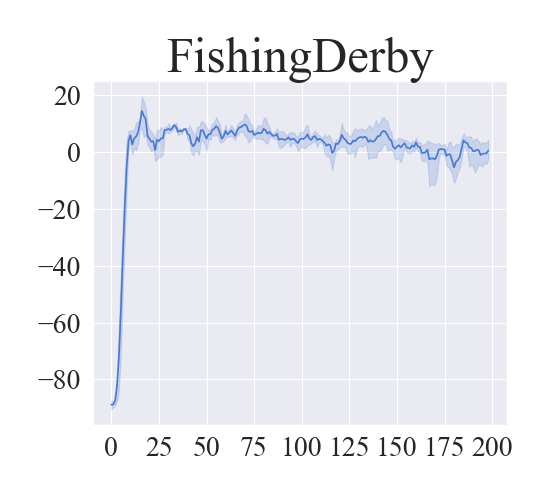}\\
			\end{minipage}%
		}%
		\subfigure{
			\begin{minipage}[t]{0.166\linewidth}
				\centering
				\includegraphics[width=1.05in]{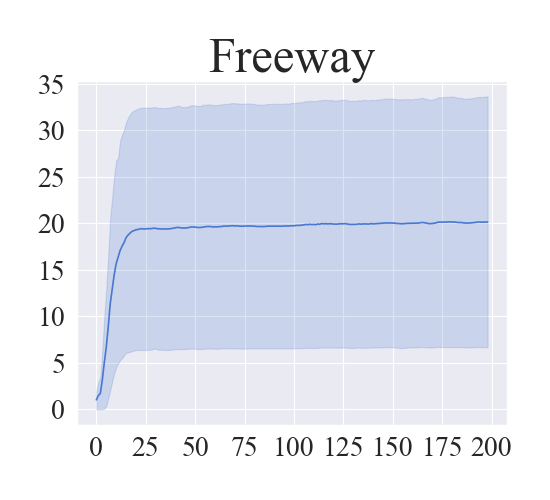}\\
			\end{minipage}%
		}%
		\vspace{-0.6cm}
		
		\subfigure{
			\begin{minipage}[t]{0.166\linewidth}
				\centering
				\includegraphics[width=1.05in]{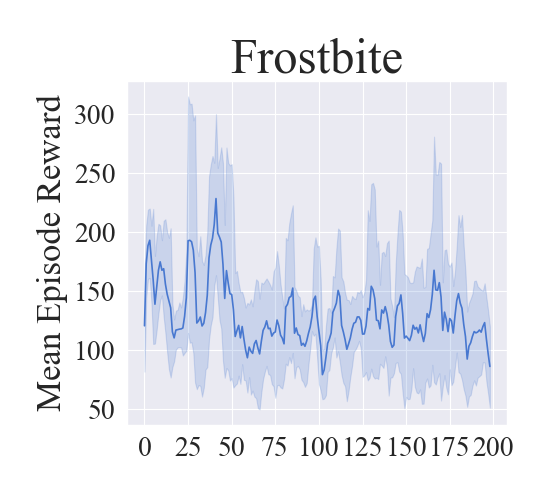}\\
			\end{minipage}%
		}%
		\subfigure{
			\begin{minipage}[t]{0.166\linewidth}
				\centering
				\includegraphics[width=1.05in]{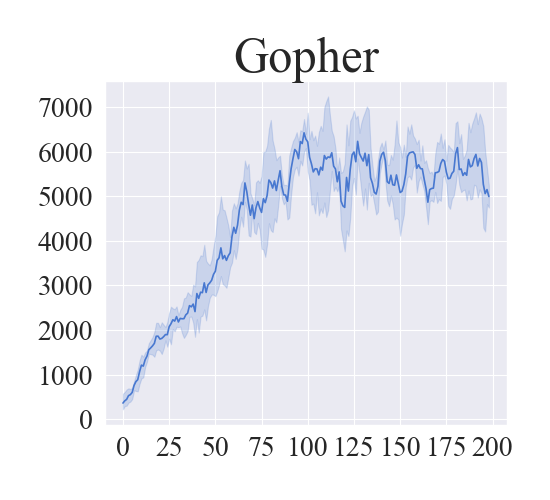}\\
			\end{minipage}%
		}%
		\subfigure{
			\begin{minipage}[t]{0.166\linewidth}
				\centering
				\includegraphics[width=1.05in]{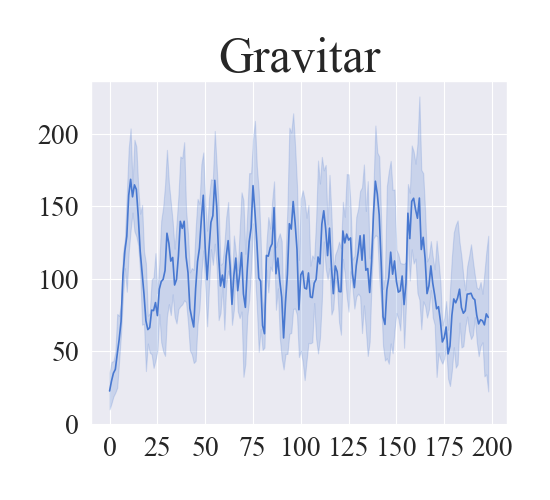}\\
			\end{minipage}%
		}%
		\subfigure{
			\begin{minipage}[t]{0.166\linewidth}
				\centering
				\includegraphics[width=1.05in]{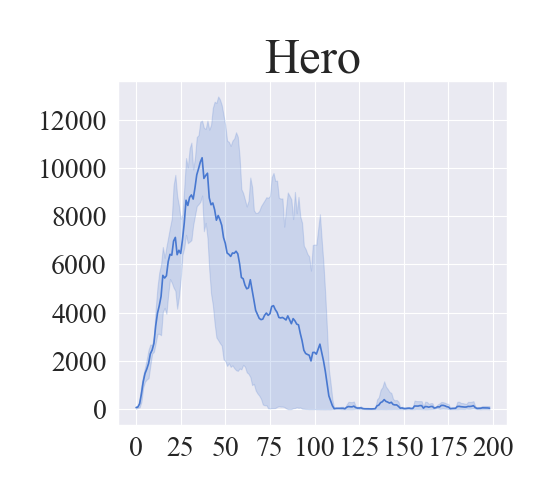}\\
			\end{minipage}%
		}%
		\subfigure{
			\begin{minipage}[t]{0.166\linewidth}
				\centering
				\includegraphics[width=1.05in]{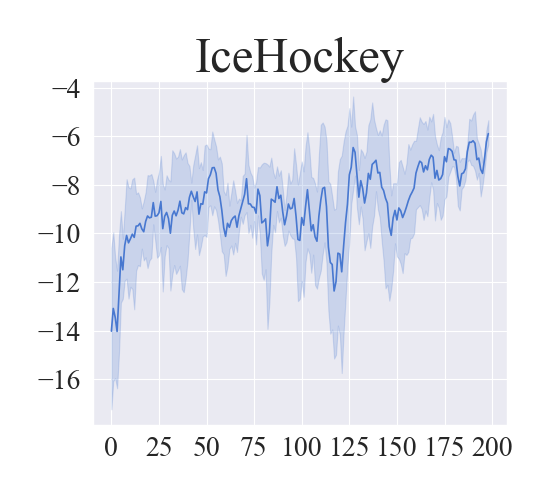}\\
			\end{minipage}%
		}%
		\subfigure{
			\begin{minipage}[t]{0.166\linewidth}
				\centering
				\includegraphics[width=1.05in]{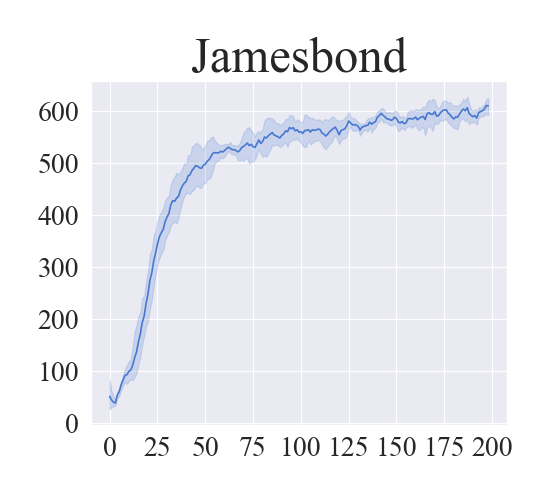}\\
			\end{minipage}%
		}%
		\vspace{-0.6cm}
		
		\subfigure{
			\begin{minipage}[t]{0.166\linewidth}
				\centering
				\includegraphics[width=1.05in]{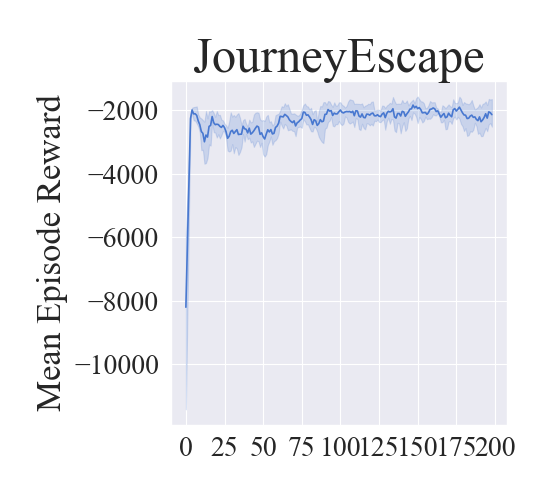}\\
			\end{minipage}%
		}%
		\subfigure{
			\begin{minipage}[t]{0.166\linewidth}
				\centering
				\includegraphics[width=1.05in]{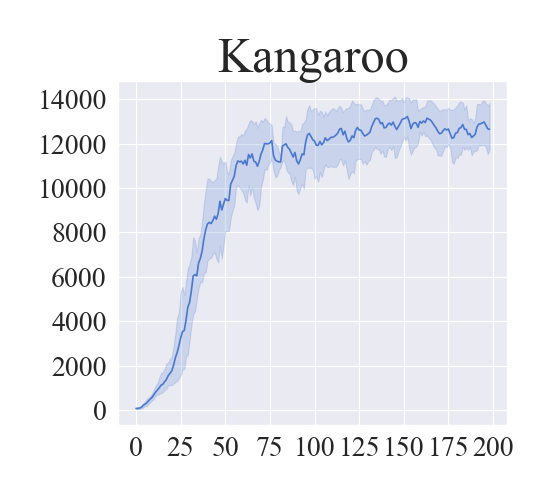}\\
			\end{minipage}%
		}%
		\subfigure{
			\begin{minipage}[t]{0.166\linewidth}
				\centering
				\includegraphics[width=1.05in]{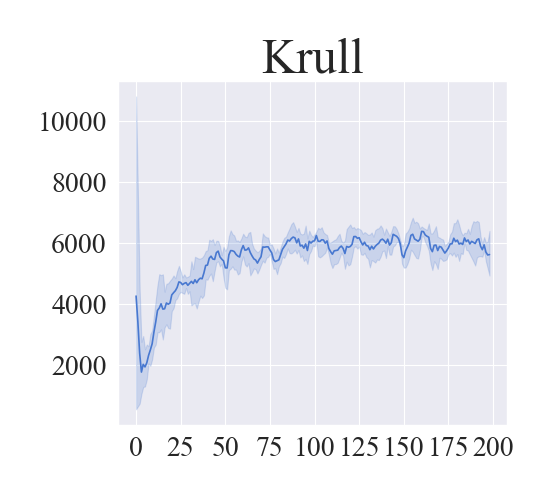}\\
			\end{minipage}%
		}%
		\subfigure{
			\begin{minipage}[t]{0.166\linewidth}
				\centering
				\includegraphics[width=1.05in]{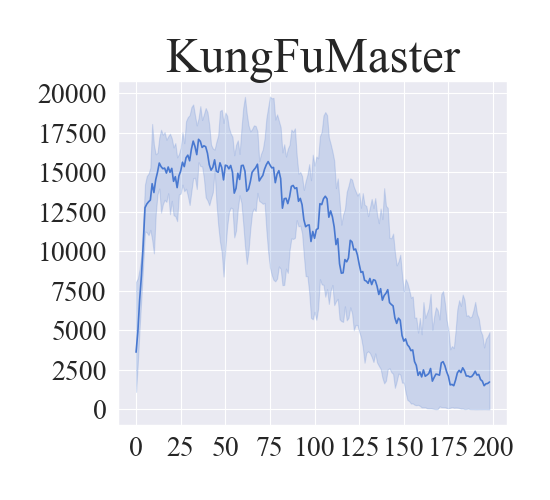}\\
			\end{minipage}%
		}%
		\subfigure{
			\begin{minipage}[t]{0.166\linewidth}
				\centering
				\includegraphics[width=1.05in]{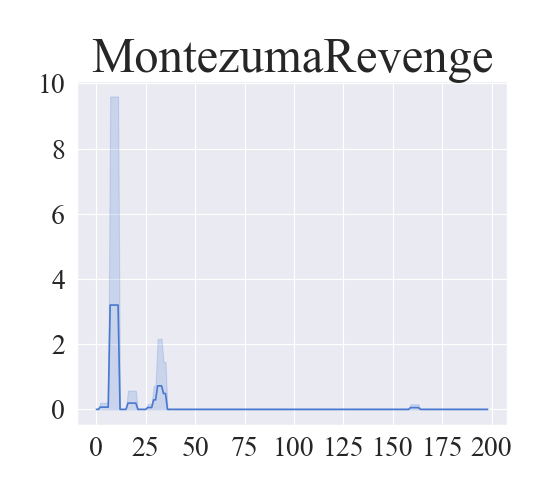}\\
			\end{minipage}%
		}%
		\subfigure{
			\begin{minipage}[t]{0.166\linewidth}
				\centering
				\includegraphics[width=1.05in]{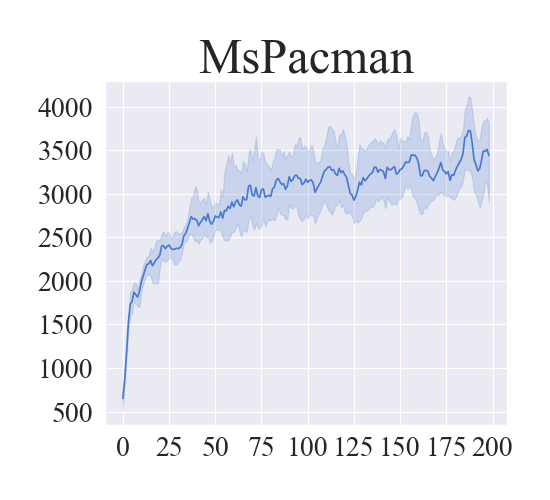}\\
			\end{minipage}%
		}%
		\vspace{-0.6cm}
		
		\subfigure{
			\begin{minipage}[t]{0.166\linewidth}
				\centering
				\includegraphics[width=1.05in]{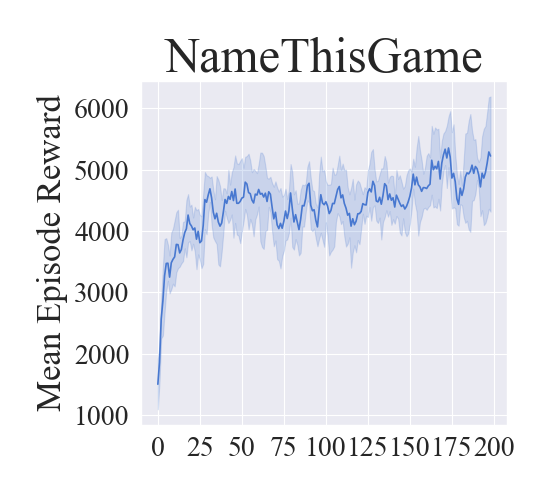}\\
			\end{minipage}%
		}%
		\subfigure{
			\begin{minipage}[t]{0.166\linewidth}
				\centering
				\includegraphics[width=1.05in]{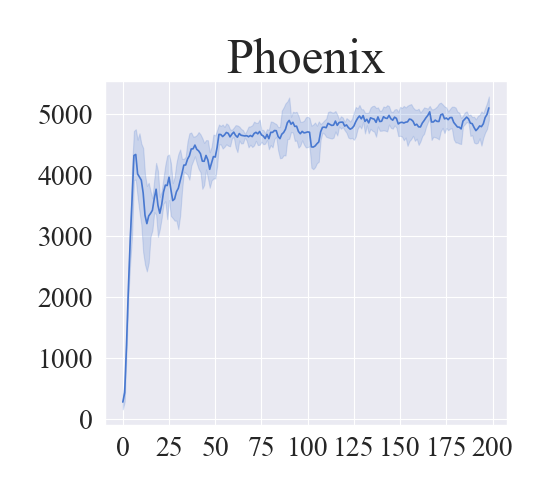}\\
			\end{minipage}%
		}%
		\subfigure{
			\begin{minipage}[t]{0.166\linewidth}
				\centering
				\includegraphics[width=1.05in]{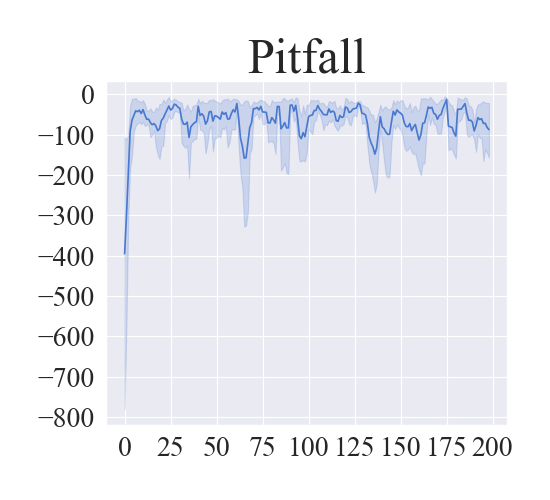}\\
			\end{minipage}%
		}%
		\subfigure{
			\begin{minipage}[t]{0.166\linewidth}
				\centering
				\includegraphics[width=1.05in]{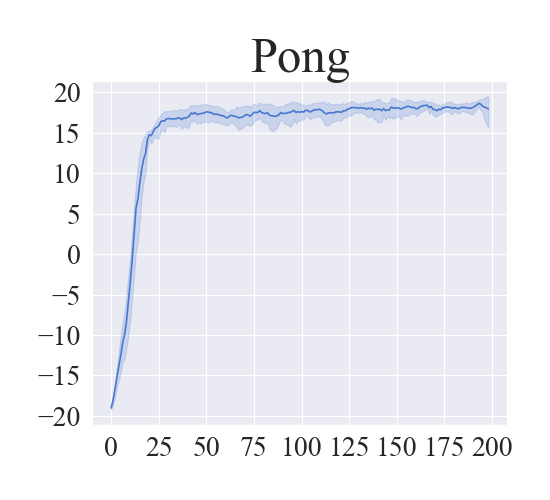}\\
			\end{minipage}%
		}%
		\subfigure{
			\begin{minipage}[t]{0.166\linewidth}
				\centering
				\includegraphics[width=1.05in]{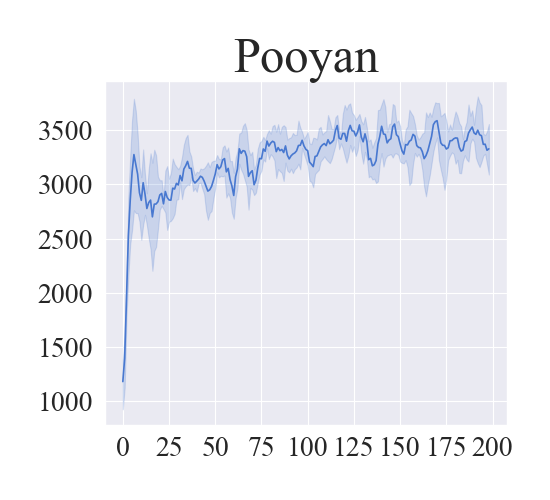}\\
			\end{minipage}%
		}%
		\subfigure{
			\begin{minipage}[t]{0.166\linewidth}
				\centering
				\includegraphics[width=1.05in]{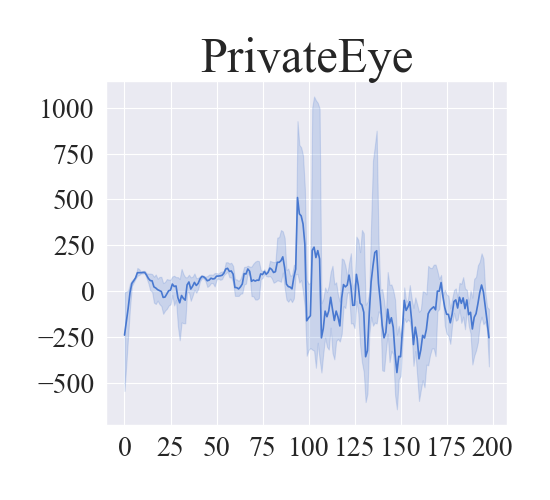}\\
			\end{minipage}%
		}%
		\vspace{-0.6cm}
		
		\subfigure{
			\begin{minipage}[t]{0.166\linewidth}
				\centering
				\includegraphics[width=1.05in]{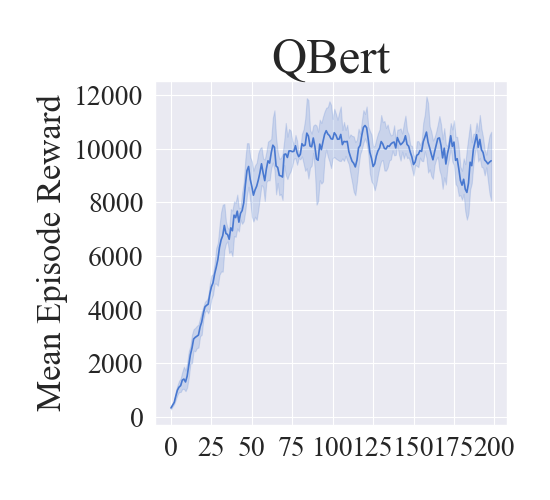}\\
			\end{minipage}%
		}%
		\subfigure{
			\begin{minipage}[t]{0.166\linewidth}
				\centering
				\includegraphics[width=1.05in]{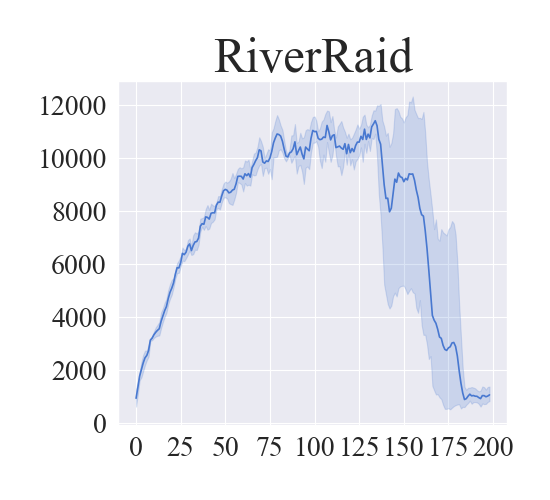}\\
			\end{minipage}%
		}%
		\subfigure{
			\begin{minipage}[t]{0.166\linewidth}
				\centering
				\includegraphics[width=1.05in]{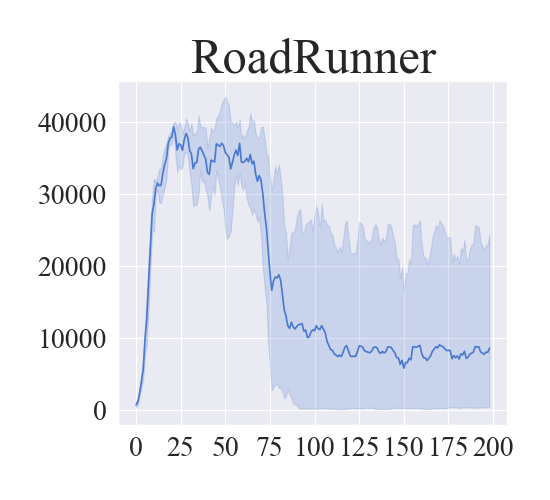}\\
			\end{minipage}%
		}%
		\subfigure{
			\begin{minipage}[t]{0.166\linewidth}
				\centering
				\includegraphics[width=1.05in]{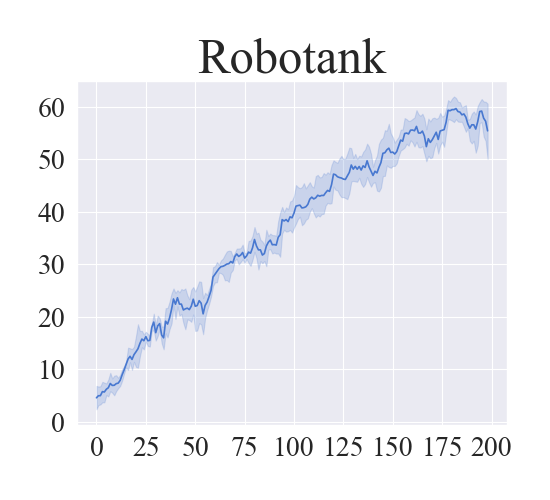}\\
			\end{minipage}%
		}%
		\subfigure{
			\begin{minipage}[t]{0.166\linewidth}
				\centering
				\includegraphics[width=1.05in]{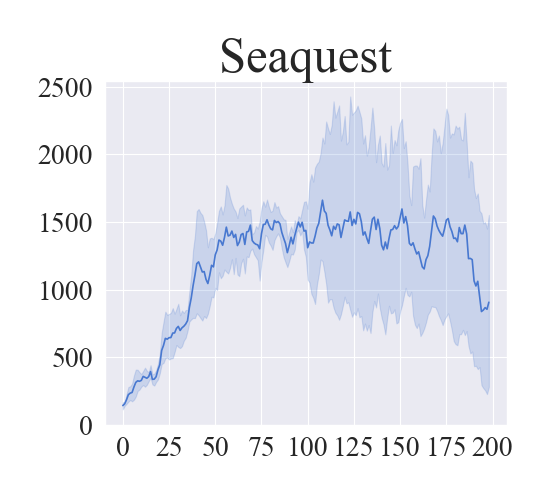}\\
			\end{minipage}%
		}%
		\subfigure{
			\begin{minipage}[t]{0.166\linewidth}
				\centering
				\includegraphics[width=1.05in]{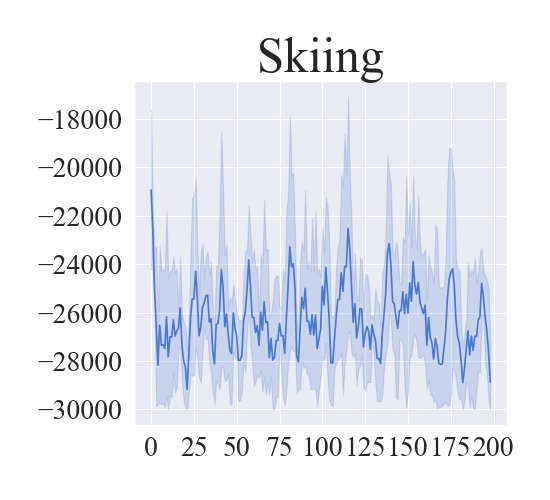}\\
			\end{minipage}%
		}%
		\vspace{-0.6cm}
		
		\subfigure{
			\begin{minipage}[t]{0.166\linewidth}
				\centering
				\includegraphics[width=1.05in]{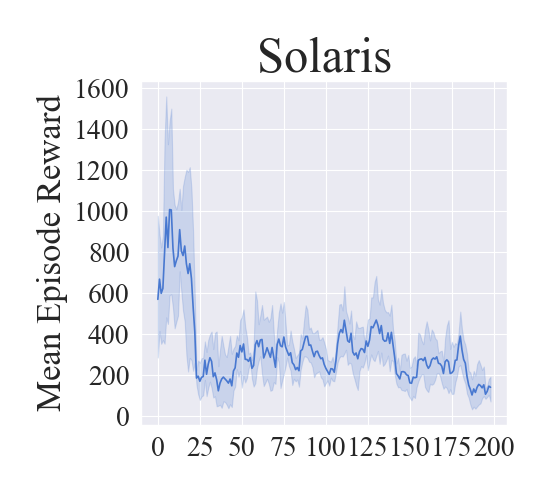}\\
			\end{minipage}%
		}%
		\subfigure{
			\begin{minipage}[t]{0.166\linewidth}
				\centering
				\includegraphics[width=1.05in]{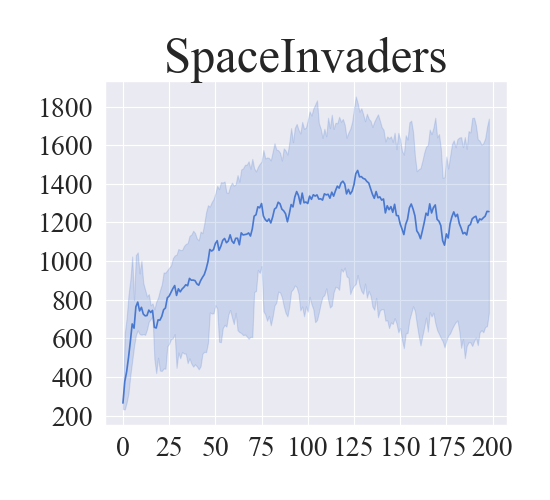}\\
			\end{minipage}%
		}%
		\subfigure{
			\begin{minipage}[t]{0.166\linewidth}
				\centering
				\includegraphics[width=1.05in]{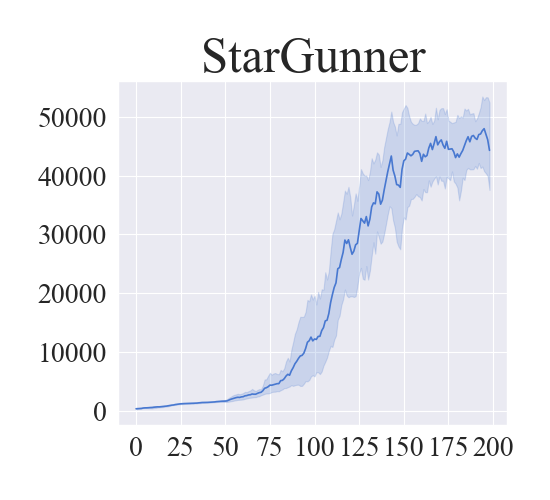}\\
			\end{minipage}%
		}%
		\subfigure{
			\begin{minipage}[t]{0.166\linewidth}
				\centering
				\includegraphics[width=1.05in]{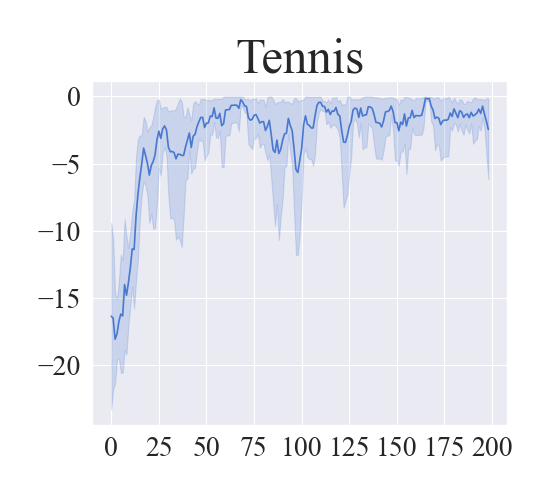}\\
			\end{minipage}%
		}%
		\subfigure{
			\begin{minipage}[t]{0.166\linewidth}
				\centering
				\includegraphics[width=1.05in]{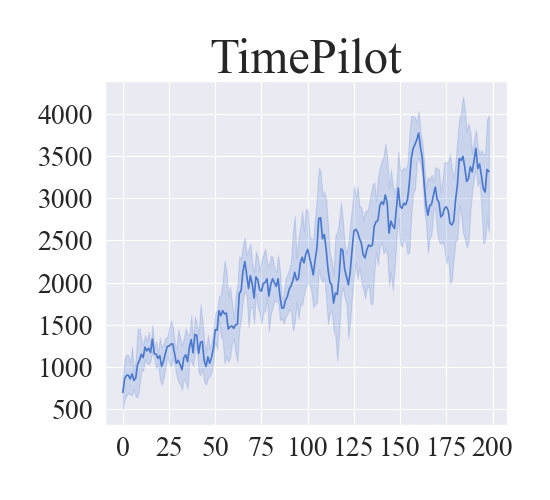}\\
			\end{minipage}%
		}%
		\subfigure{
			\begin{minipage}[t]{0.166\linewidth}
				\centering
				\includegraphics[width=1.05in]{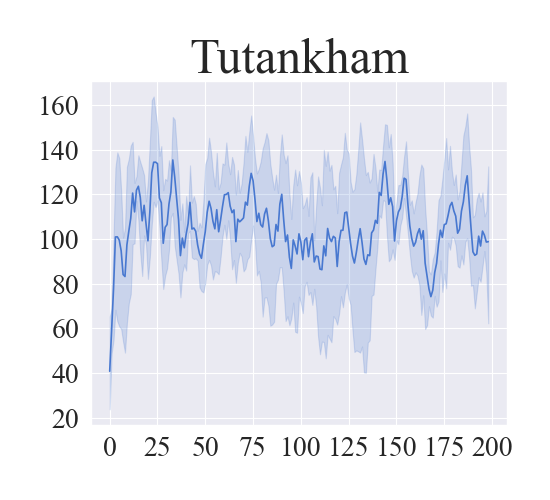}\\
			\end{minipage}%
		}%
		\vspace{-0.6cm}
		
		\subfigure{
			\begin{minipage}[t]{0.166\linewidth}
				\centering
				\includegraphics[width=1.05in]{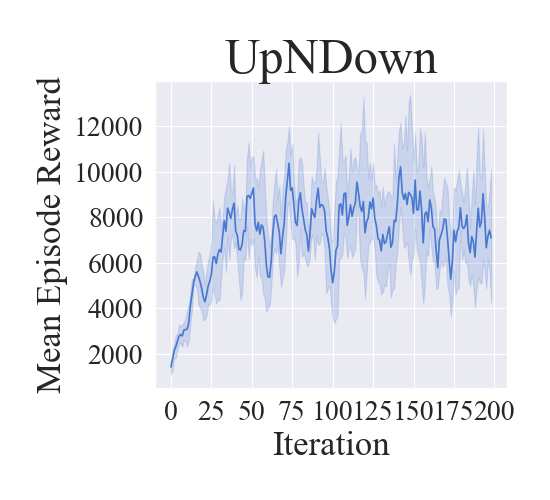}\\
			\end{minipage}%
		}%
		\subfigure{
			\begin{minipage}[t]{0.166\linewidth}
				\centering
				\includegraphics[width=1.05in]{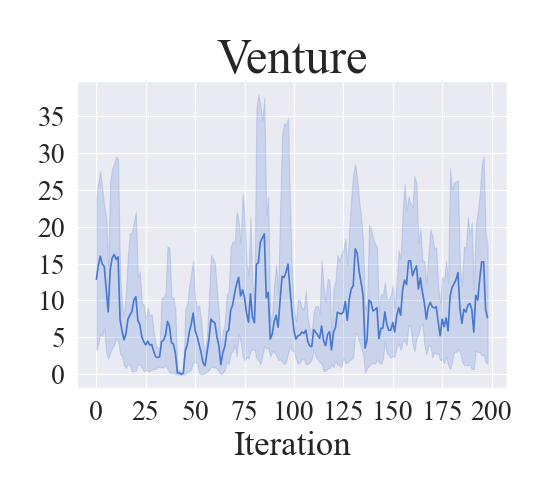}\\
			\end{minipage}%
		}%
		\subfigure{
			\begin{minipage}[t]{0.166\linewidth}
				\centering
				\includegraphics[width=1.05in]{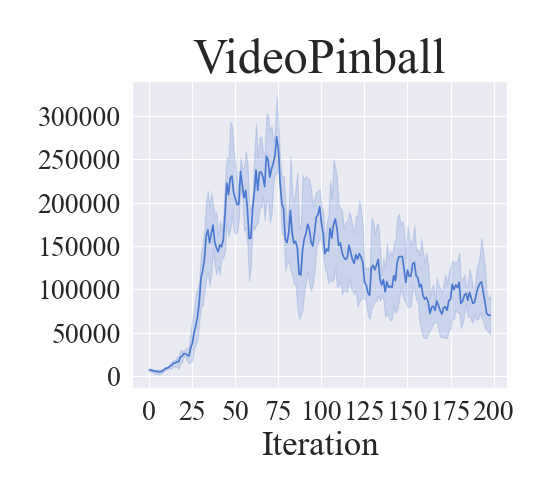}\\
			\end{minipage}%
		}%
		\subfigure{
			\begin{minipage}[t]{0.166\linewidth}
				\centering
				\includegraphics[width=1.05in]{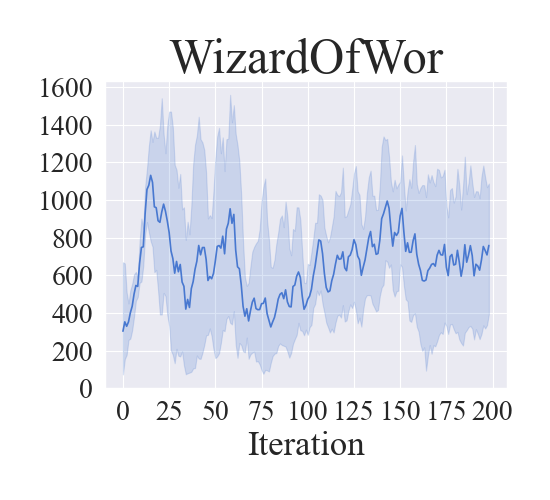}\\
			\end{minipage}%
		}%
		\subfigure{
			\begin{minipage}[t]{0.166\linewidth}
				\centering
				\includegraphics[width=1.05in]{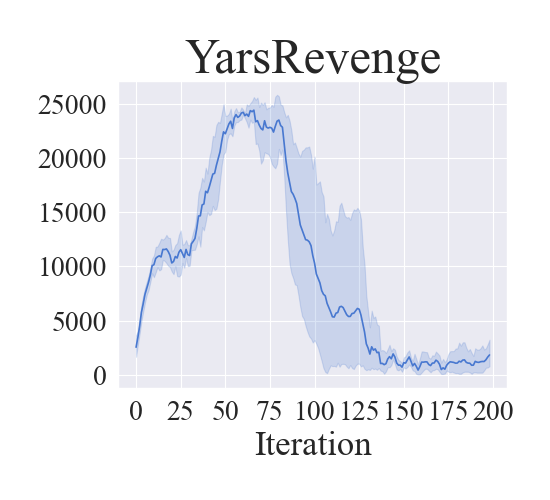}\\
			\end{minipage}%
		}%
		\subfigure{
			\begin{minipage}[t]{0.166\linewidth}
				\centering
				\includegraphics[width=1.05in]{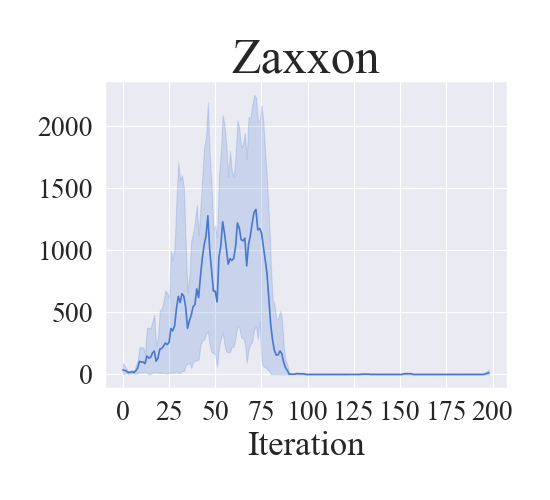}\\
			\end{minipage}%
		}%
		
		\centering
		\caption{\textbf{Data Generation (by DQN)}}
		\label{fig: Data Generation}
		
	\end{figure*}

	Please refer Fig. \ref{fig: Data Generation}.

    \subsection{Learning curves of all $60$ Atari $2600$ games on poor dataset}
\begin{figure*}[!htb]
		\centering

		\subfigure{
			\begin{minipage}[t]{\linewidth}
				\centering
				\includegraphics[width=4in]{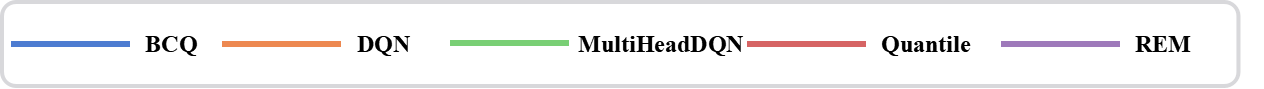}\\
			\end{minipage}%
		}%
		
		\subfigure{
			\begin{minipage}[t]{0.166\linewidth}
				\centering
				\includegraphics[width=1.05in]{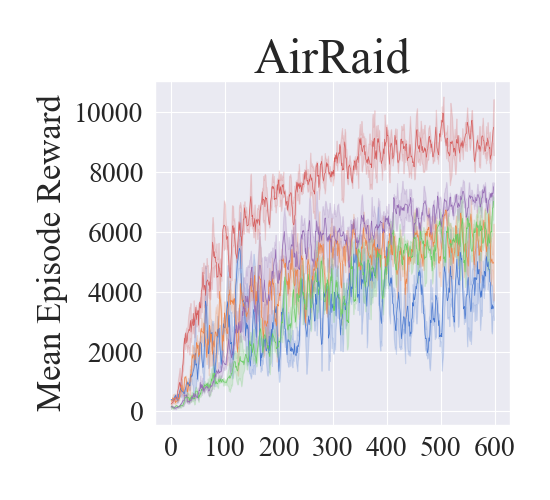}\\
			\end{minipage}%
		}%
		\subfigure{
			\begin{minipage}[t]{0.166\linewidth}
				\centering
				\includegraphics[width=1.05in]{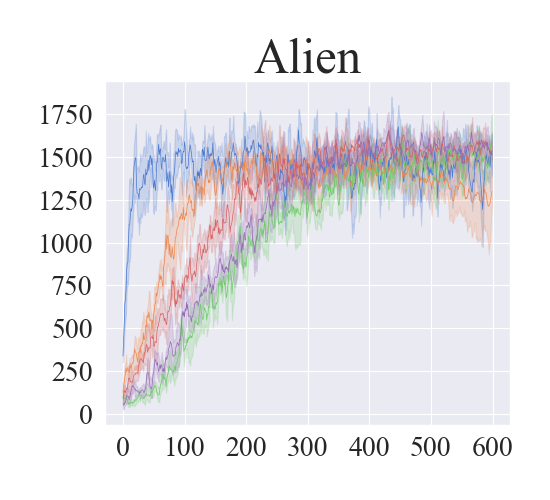}\\
			\end{minipage}%
		}%
		\subfigure{
			\begin{minipage}[t]{0.166\linewidth}
				\centering
				\includegraphics[width=1.05in]{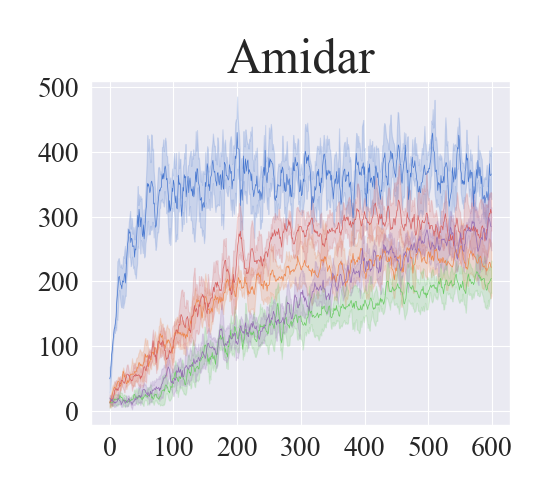}\\
			\end{minipage}%
		}%
		\subfigure{
			\begin{minipage}[t]{0.166\linewidth}
				\centering
				\includegraphics[width=1.05in]{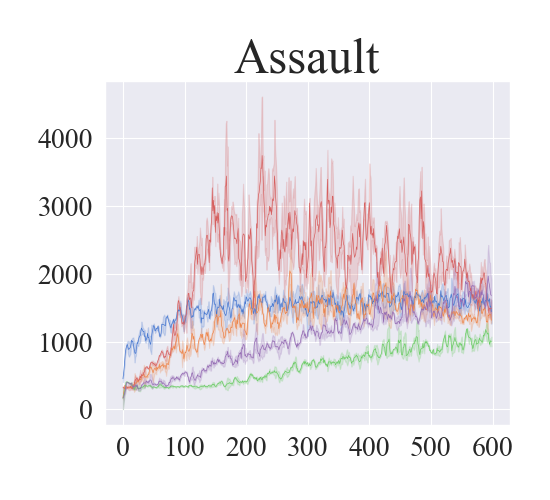}\\
			\end{minipage}%
		}%
		\subfigure{
			\begin{minipage}[t]{0.166\linewidth}
				\centering
				\includegraphics[width=1.05in]{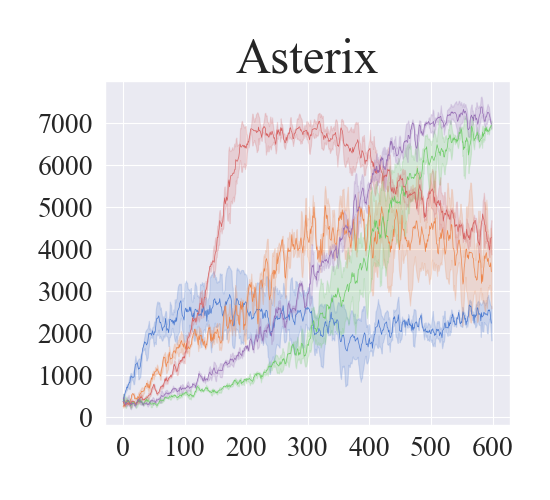}\\
			\end{minipage}%
		}%
		\subfigure{
			\begin{minipage}[t]{0.166\linewidth}
				\centering
				\includegraphics[width=1.05in]{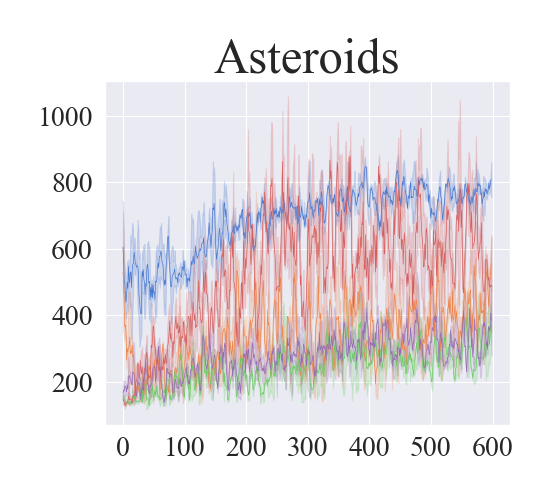}\\\
			\end{minipage}%
		}%
		\vspace{-1.0cm}
		
		\subfigure{
			\begin{minipage}[t]{0.166\linewidth}
				\centering
				\includegraphics[width=1.05in]{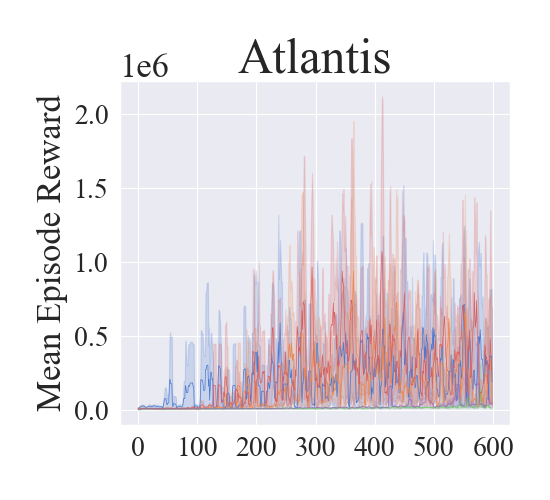}\\
			\end{minipage}%
		}%
		\subfigure{
			\begin{minipage}[t]{0.166\linewidth}
				\centering
				\includegraphics[width=1.05in]{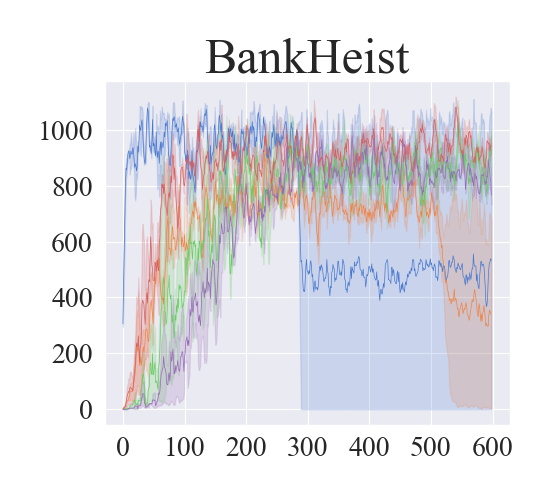}\\
			\end{minipage}%
		}%
		\subfigure{
			\begin{minipage}[t]{0.166\linewidth}
				\centering
				\includegraphics[width=1.05in]{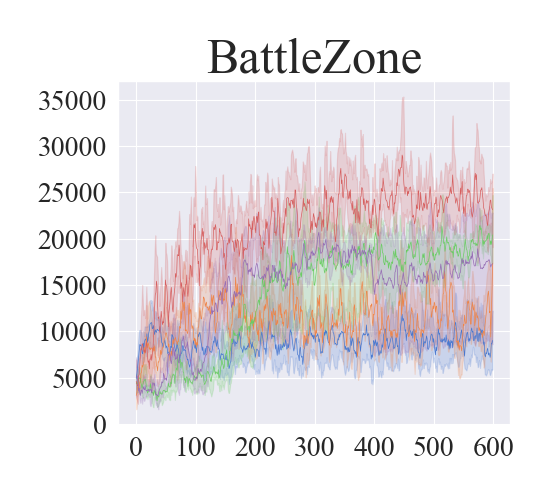}\\
			\end{minipage}%
		}%
		\subfigure{
			\begin{minipage}[t]{0.166\linewidth}
				\centering
				\includegraphics[width=1.05in]{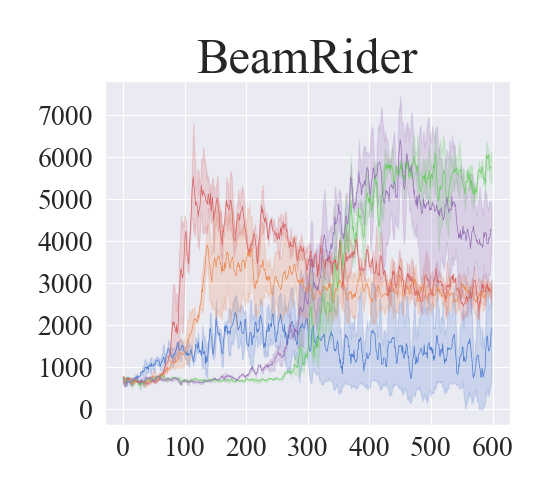}\\
			\end{minipage}%
		}%
		\subfigure{
			\begin{minipage}[t]{0.166\linewidth}
				\centering
				\includegraphics[width=1.05in]{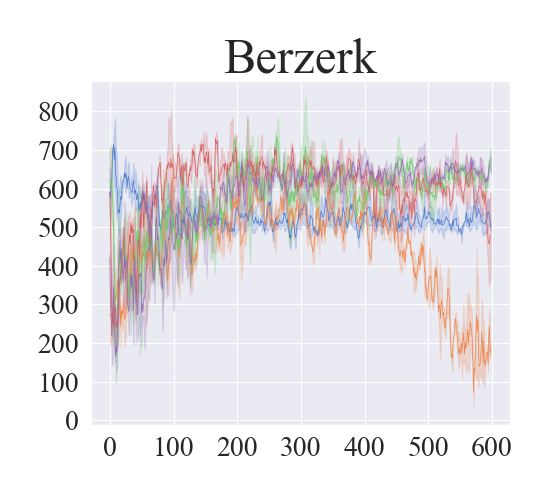}\\
			\end{minipage}%
		}%
		\subfigure{
			\begin{minipage}[t]{0.166\linewidth}
				\centering
				\includegraphics[width=1.05in]{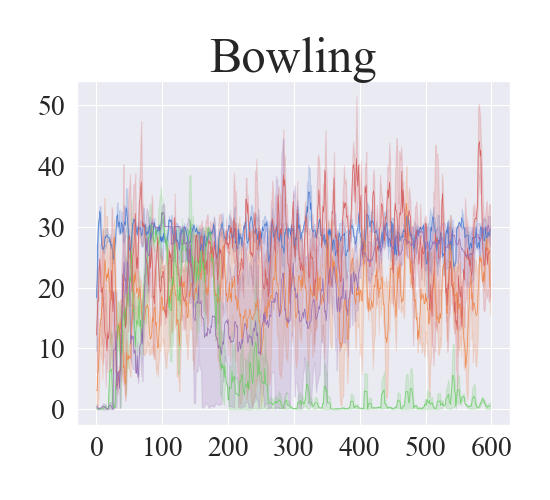}\\
			\end{minipage}%
		}%
		\vspace{-0.6cm}
		
		\subfigure{
			\begin{minipage}[t]{0.166\linewidth}
				\centering
				\includegraphics[width=1.05in]{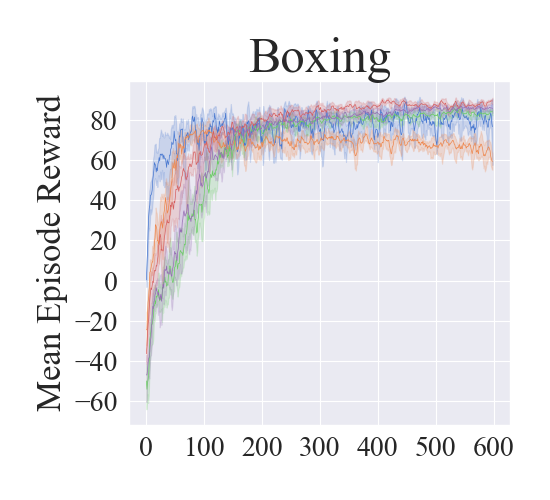}\\
			\end{minipage}%
		}%
		\subfigure{
			\begin{minipage}[t]{0.166\linewidth}
				\centering
				\includegraphics[width=1.05in]{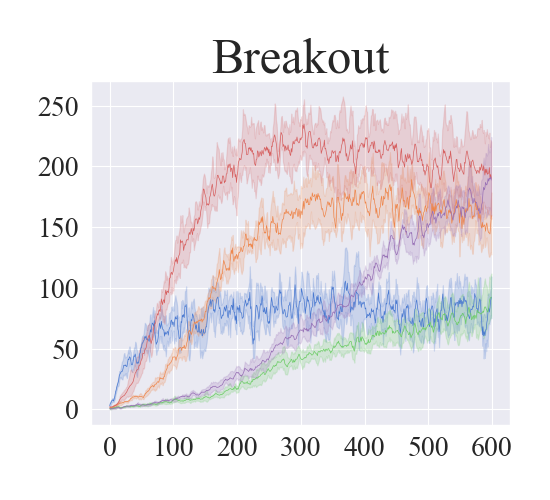}\\
			\end{minipage}%
		}%
		\subfigure{
			\begin{minipage}[t]{0.166\linewidth}
				\centering
				\includegraphics[width=1.05in]{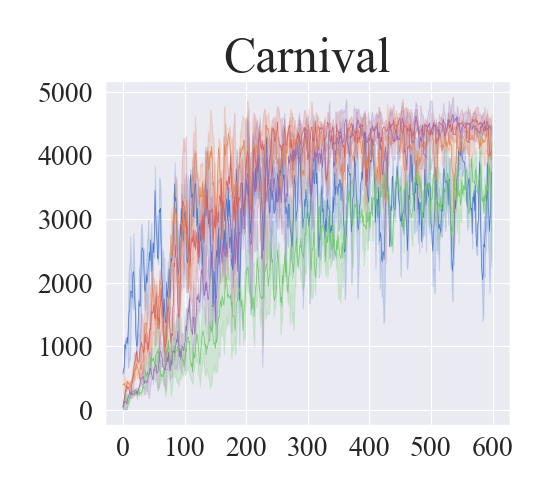}\\
			\end{minipage}%
		}%
		\subfigure{
			\begin{minipage}[t]{0.166\linewidth}
				\centering
				\includegraphics[width=1.05in]{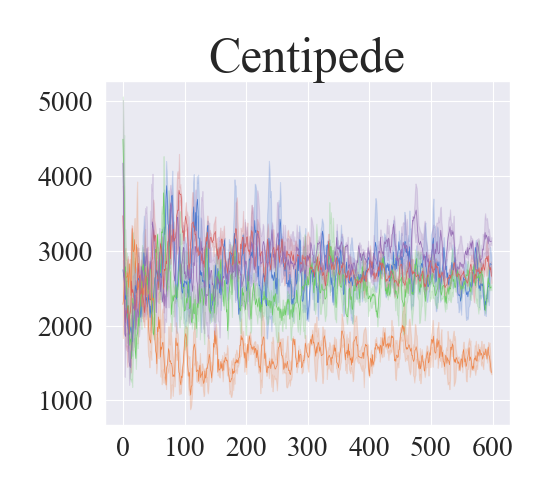}\\
			\end{minipage}%
		}%
		\subfigure{
			\begin{minipage}[t]{0.166\linewidth}
				\centering
				\includegraphics[width=1.05in]{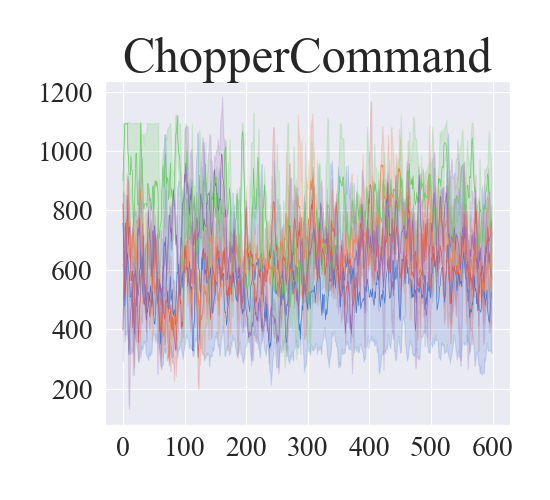}\\
			\end{minipage}%
		}%
		\subfigure{
			\begin{minipage}[t]{0.166\linewidth}
				\centering
				\includegraphics[width=1.05in]{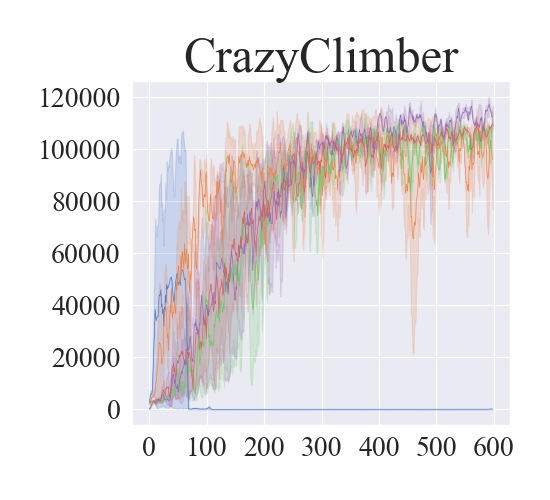}\\
			\end{minipage}%
		}%
		\vspace{-0.6cm}
		
		\subfigure{
			\begin{minipage}[t]{0.166\linewidth}
				\centering
				\includegraphics[width=1.05in]{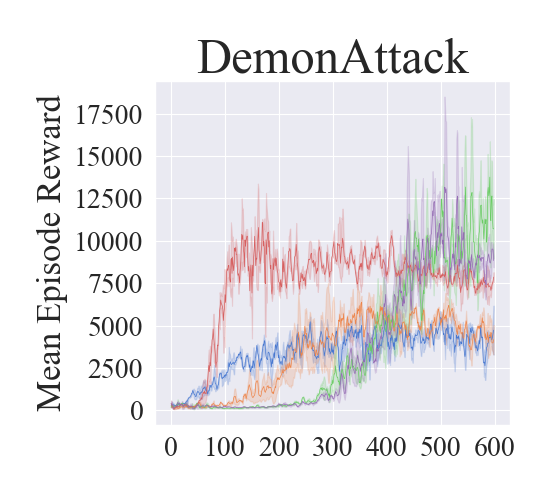}\\
			\end{minipage}%
		}%
		\subfigure{
			\begin{minipage}[t]{0.166\linewidth}
				\centering
				\includegraphics[width=1.05in]{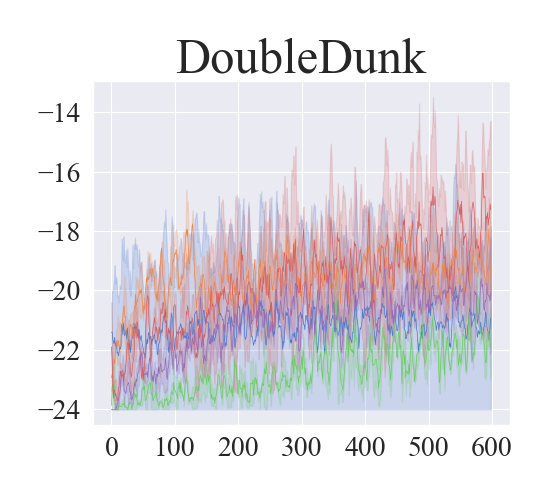}\\
			\end{minipage}%
		}%
		\subfigure{
			\begin{minipage}[t]{0.166\linewidth}
				\centering
				\includegraphics[width=1.05in]{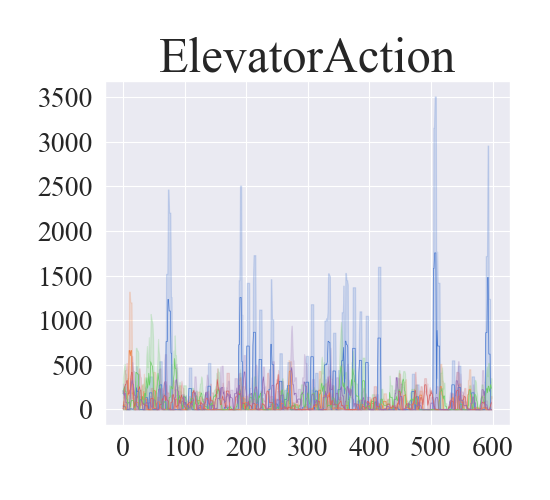}\\
			\end{minipage}%
		}%
		\subfigure{
			\begin{minipage}[t]{0.166\linewidth}
				\centering
				\includegraphics[width=1.05in]{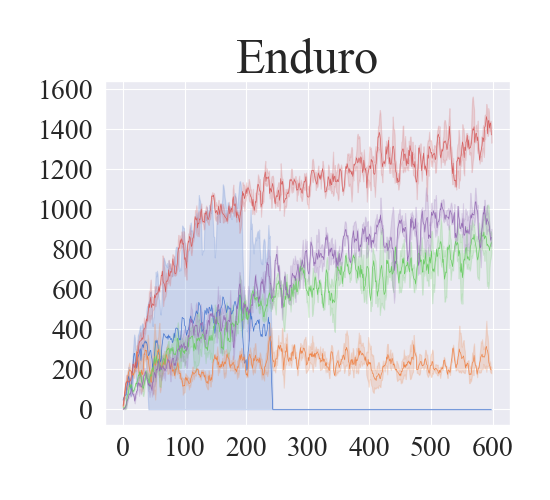}\\
			\end{minipage}%
		}%
		\subfigure{
			\begin{minipage}[t]{0.166\linewidth}
				\centering
				\includegraphics[width=1.05in]{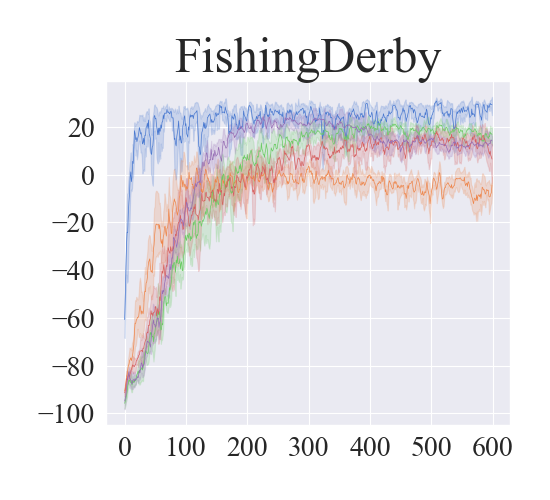}\\
			\end{minipage}%
		}%
		\subfigure{
			\begin{minipage}[t]{0.166\linewidth}
				\centering
				\includegraphics[width=1.05in]{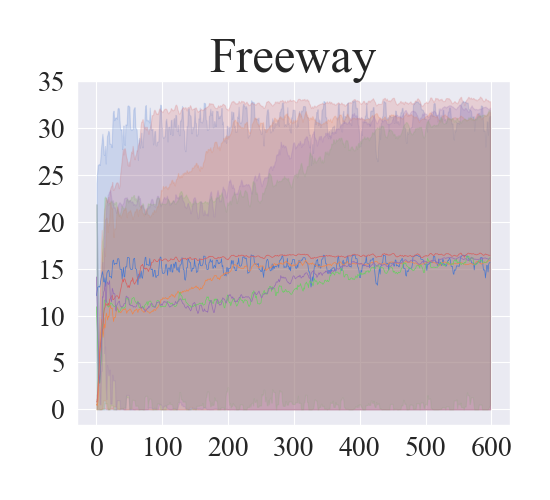}\\
			\end{minipage}%
		}%
		\vspace{-0.6cm}
		
		\subfigure{
			\begin{minipage}[t]{0.166\linewidth}
				\centering
				\includegraphics[width=1.05in]{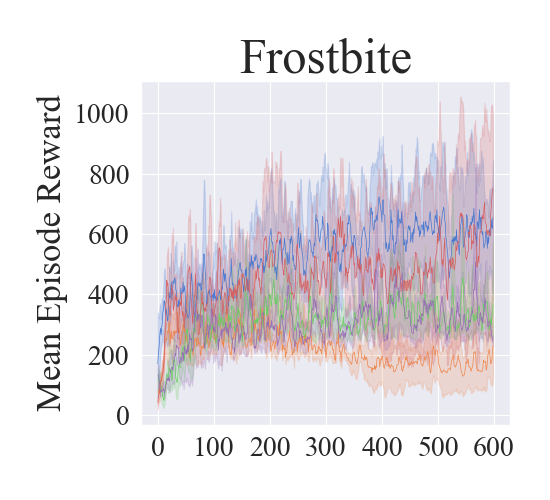}\\
			\end{minipage}%
		}%
		\subfigure{
			\begin{minipage}[t]{0.166\linewidth}
				\centering
				\includegraphics[width=1.05in]{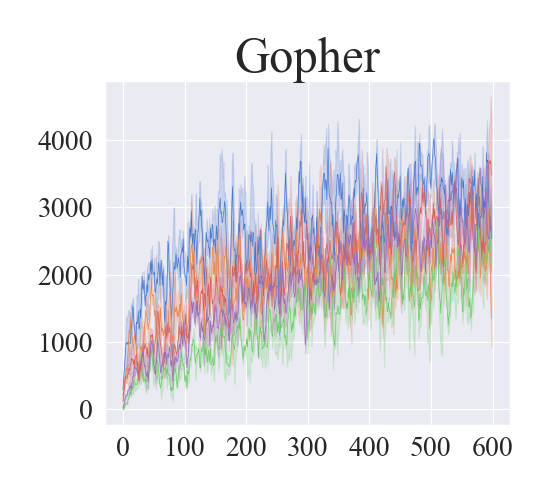}\\
			\end{minipage}%
		}%
		\subfigure{
			\begin{minipage}[t]{0.166\linewidth}
				\centering
				\includegraphics[width=1.05in]{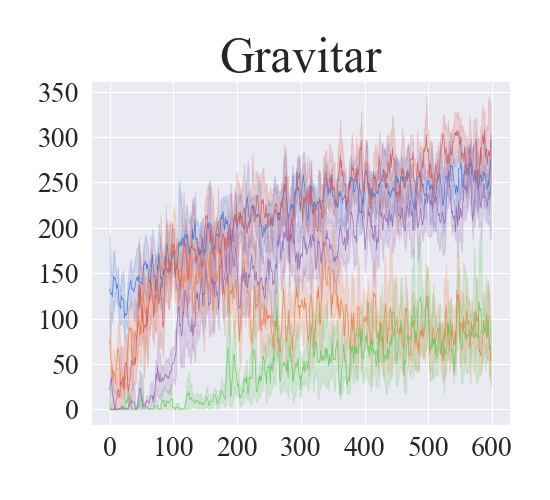}\\
			\end{minipage}%
		}%
		\subfigure{
			\begin{minipage}[t]{0.166\linewidth}
				\centering
				\includegraphics[width=1.05in]{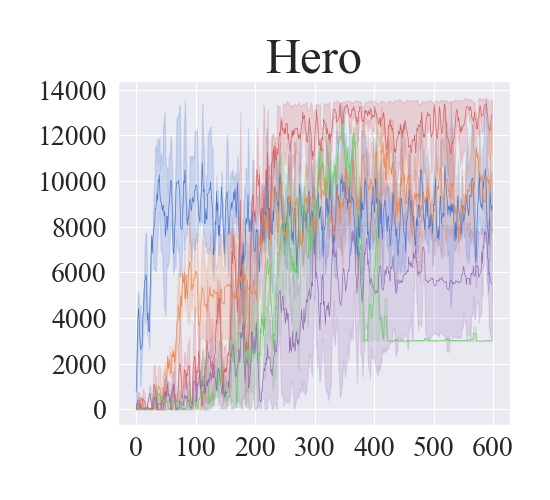}\\
			\end{minipage}%
		}%
		\subfigure{
			\begin{minipage}[t]{0.166\linewidth}
				\centering
				\includegraphics[width=1.05in]{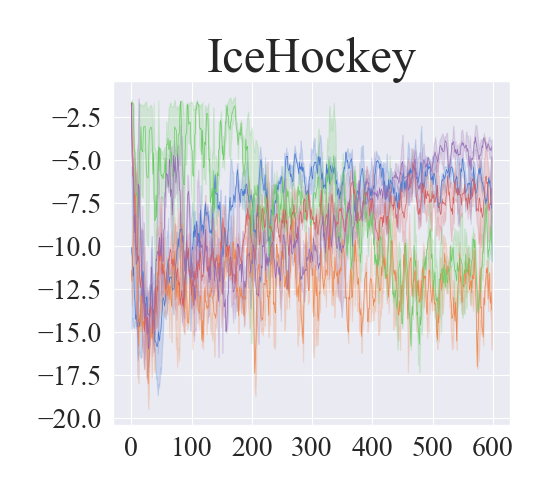}\\
			\end{minipage}%
		}%
		\subfigure{
			\begin{minipage}[t]{0.166\linewidth}
				\centering
				\includegraphics[width=1.05in]{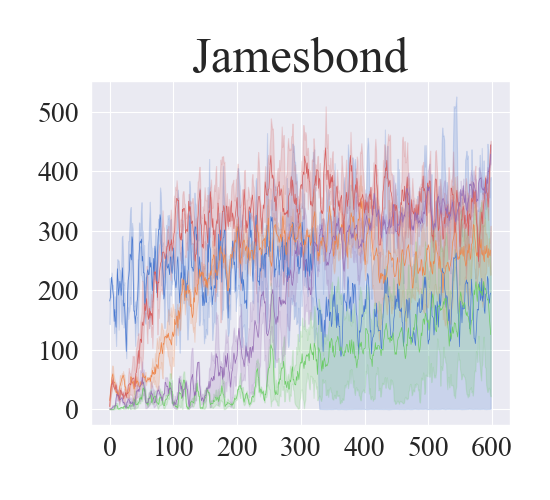}\\
			\end{minipage}%
		}%
		\vspace{-0.6cm}
		
		\subfigure{
			\begin{minipage}[t]{0.166\linewidth}
				\centering
				\includegraphics[width=1.05in]{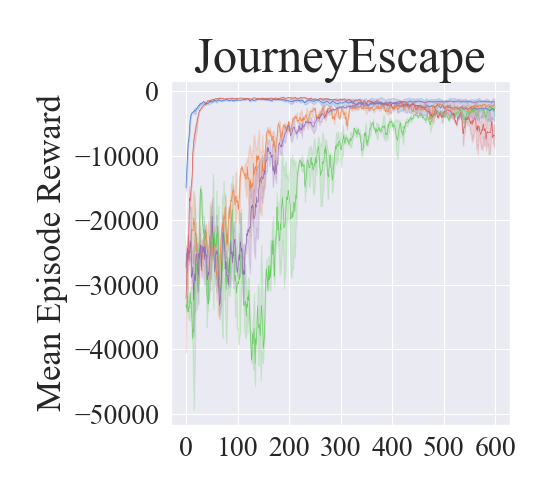}\\
			\end{minipage}%
		}%
		\subfigure{
			\begin{minipage}[t]{0.166\linewidth}
				\centering
				\includegraphics[width=1.05in]{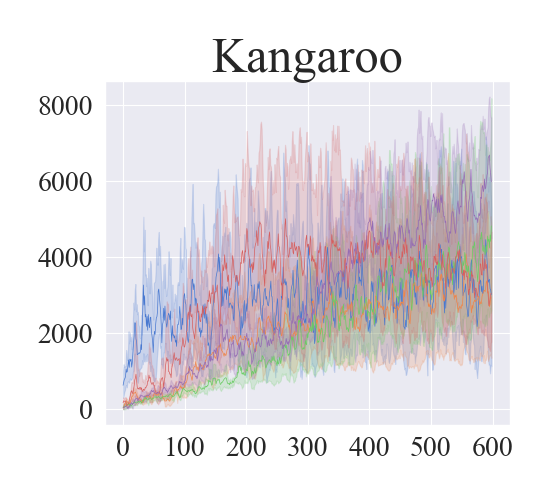}\\
			\end{minipage}%
		}%
		\subfigure{
			\begin{minipage}[t]{0.166\linewidth}
				\centering
				\includegraphics[width=1.05in]{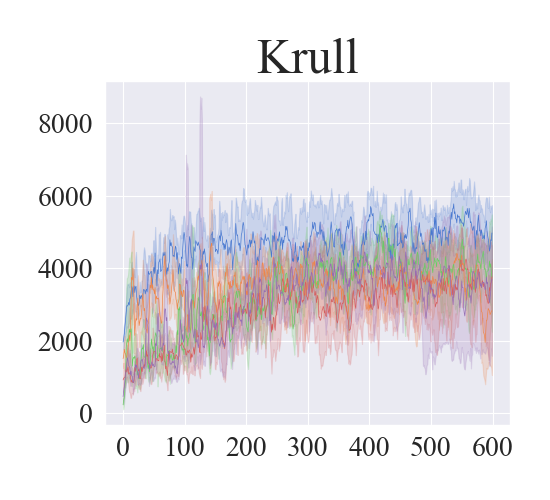}\\
			\end{minipage}%
		}%
		\subfigure{
			\begin{minipage}[t]{0.166\linewidth}
				\centering
				\includegraphics[width=1.05in]{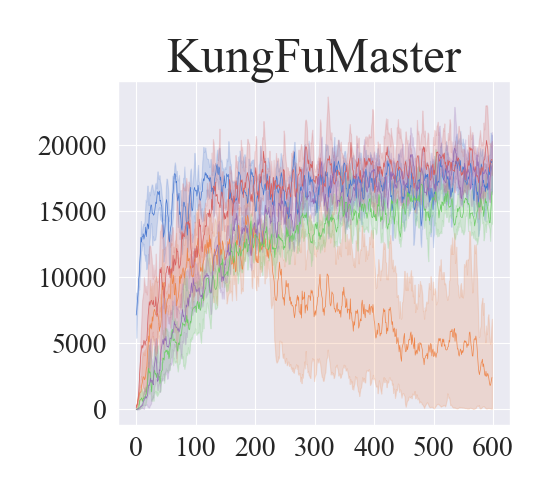}\\
			\end{minipage}%
		}%
		\subfigure{
			\begin{minipage}[t]{0.166\linewidth}
				\centering
				\includegraphics[width=1.05in]{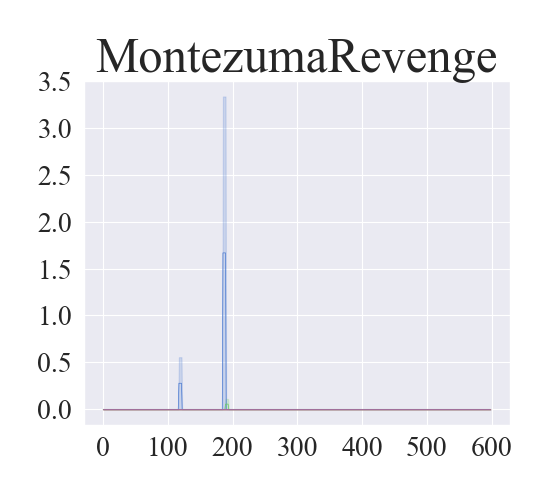}\\
			\end{minipage}%
		}%
		\subfigure{
			\begin{minipage}[t]{0.166\linewidth}
				\centering
				\includegraphics[width=1.05in]{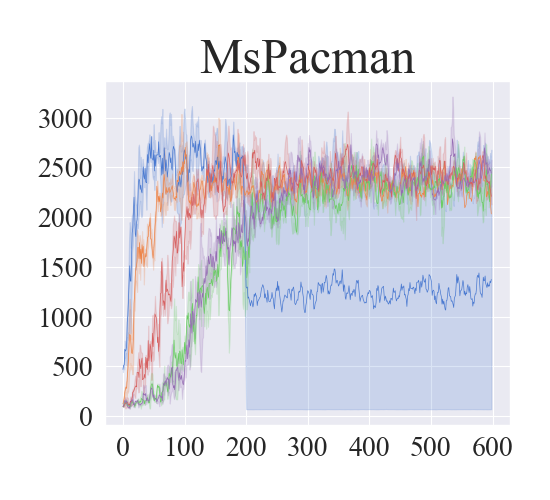}\\
			\end{minipage}%
		}%
		\vspace{-0.6cm}
		
		\subfigure{
			\begin{minipage}[t]{0.166\linewidth}
				\centering
				\includegraphics[width=1.05in]{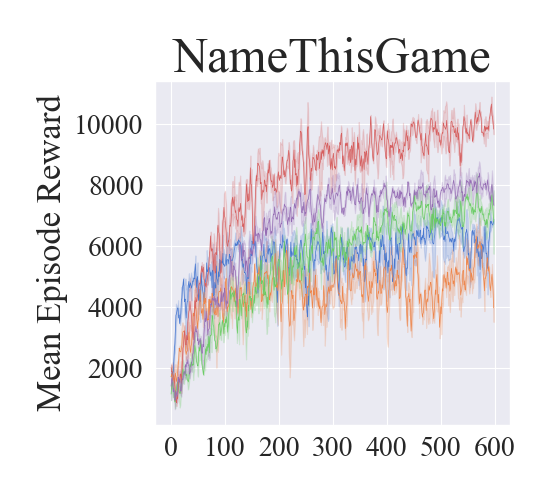}\\
			\end{minipage}%
		}%
		\subfigure{
			\begin{minipage}[t]{0.166\linewidth}
				\centering
				\includegraphics[width=1.05in]{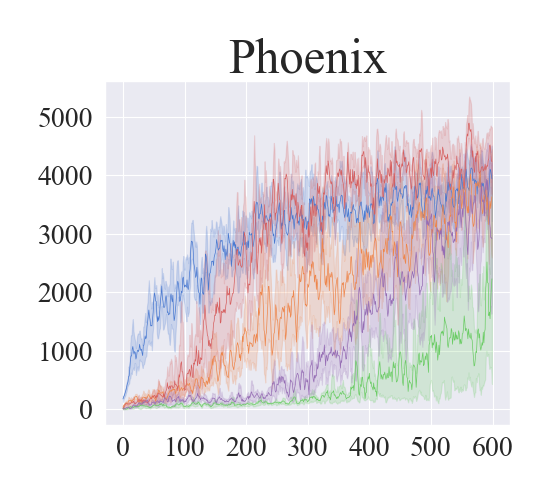}\\
			\end{minipage}%
		}%
		\subfigure{
			\begin{minipage}[t]{0.166\linewidth}
				\centering
				\includegraphics[width=1.05in]{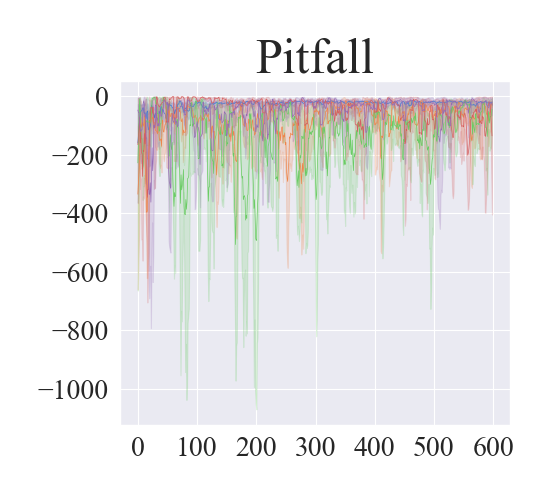}\\
			\end{minipage}%
		}%
		\subfigure{
			\begin{minipage}[t]{0.166\linewidth}
				\centering
				\includegraphics[width=1.05in]{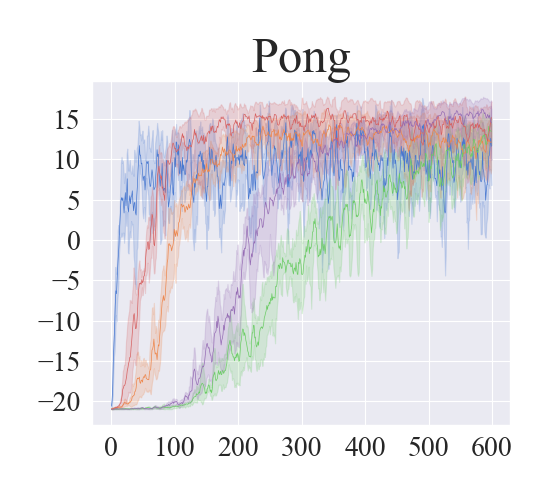}\\
			\end{minipage}%
		}%
		\subfigure{
			\begin{minipage}[t]{0.166\linewidth}
				\centering
				\includegraphics[width=1.05in]{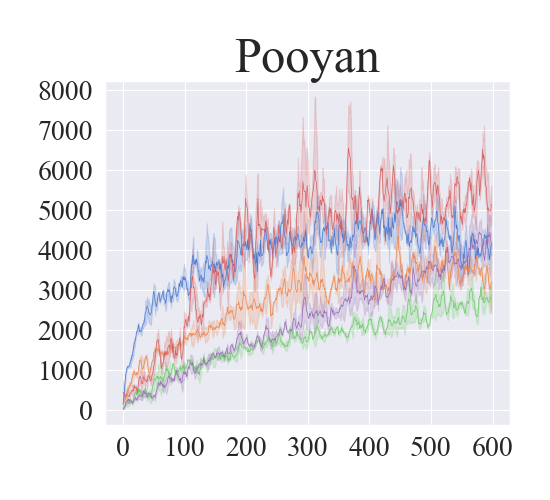}\\
			\end{minipage}%
		}%
		\subfigure{
			\begin{minipage}[t]{0.166\linewidth}
				\centering
				\includegraphics[width=1.05in]{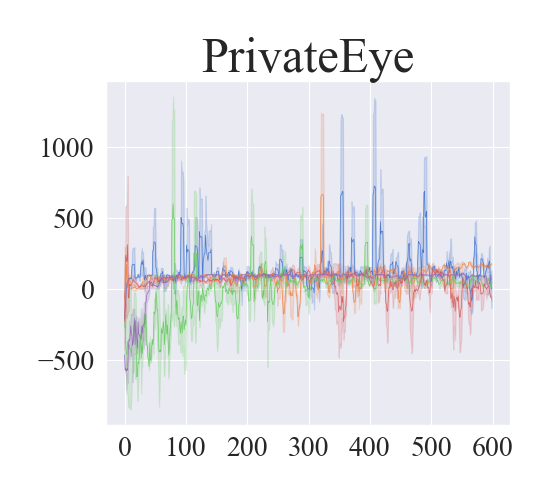}\\
			\end{minipage}%
		}%
		\vspace{-0.6cm}
		
		\subfigure{
			\begin{minipage}[t]{0.166\linewidth}
				\centering
				\includegraphics[width=1.05in]{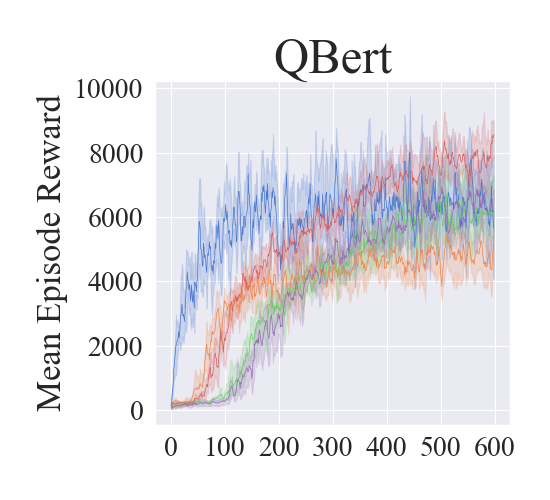}\\
			\end{minipage}%
		}%
		\subfigure{
			\begin{minipage}[t]{0.166\linewidth}
				\centering
				\includegraphics[width=1.05in]{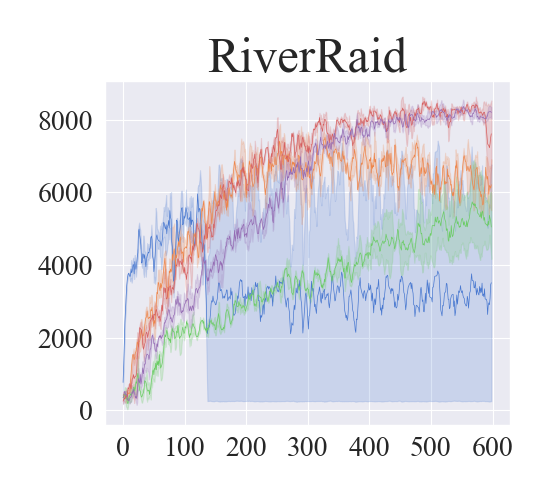}\\
			\end{minipage}%
		}%
		\subfigure{
			\begin{minipage}[t]{0.166\linewidth}
				\centering
				\includegraphics[width=1.05in]{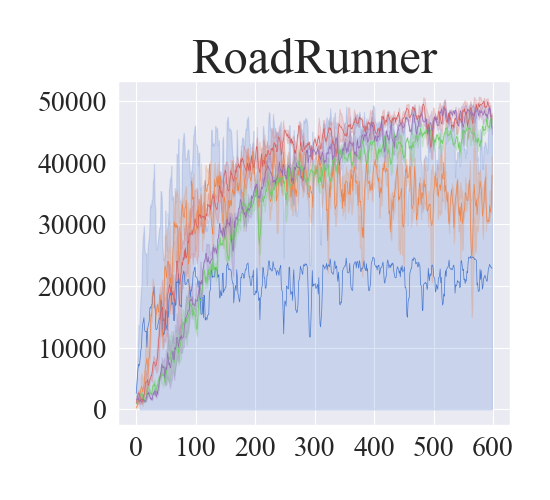}\\
			\end{minipage}%
		}%
		\subfigure{
			\begin{minipage}[t]{0.166\linewidth}
				\centering
				\includegraphics[width=1.05in]{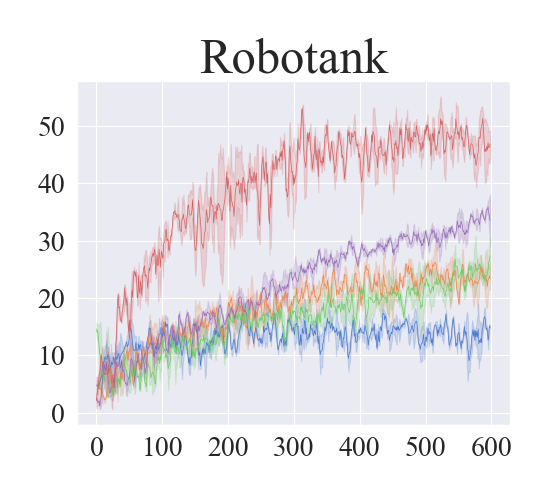}\\
			\end{minipage}%
		}%
		\subfigure{
			\begin{minipage}[t]{0.166\linewidth}
				\centering
				\includegraphics[width=1.05in]{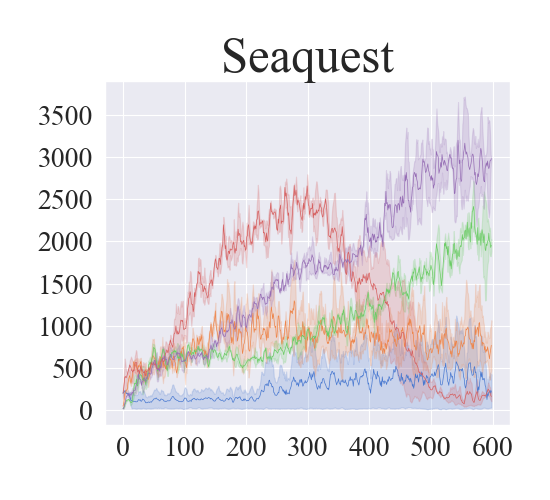}\\
			\end{minipage}%
		}%
		\subfigure{
			\begin{minipage}[t]{0.166\linewidth}
				\centering
				\includegraphics[width=1.05in]{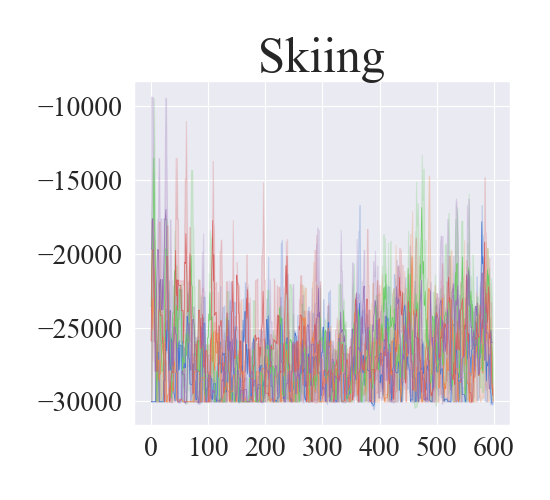}\\
			\end{minipage}%
		}%
		\vspace{-0.6cm}
		
		\subfigure{
			\begin{minipage}[t]{0.166\linewidth}
				\centering
				\includegraphics[width=1.05in]{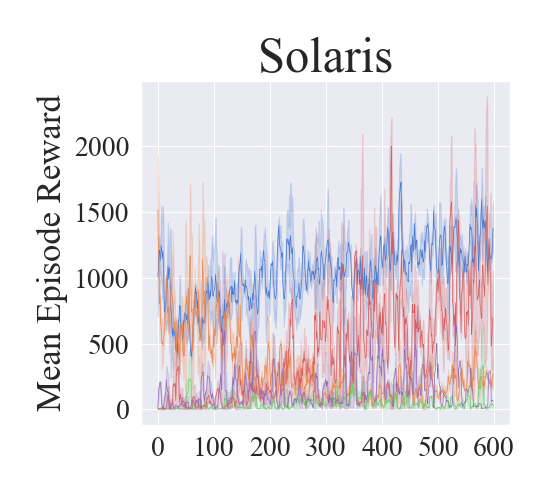}\\
			\end{minipage}%
		}%
		\subfigure{
			\begin{minipage}[t]{0.166\linewidth}
				\centering
				\includegraphics[width=1.05in]{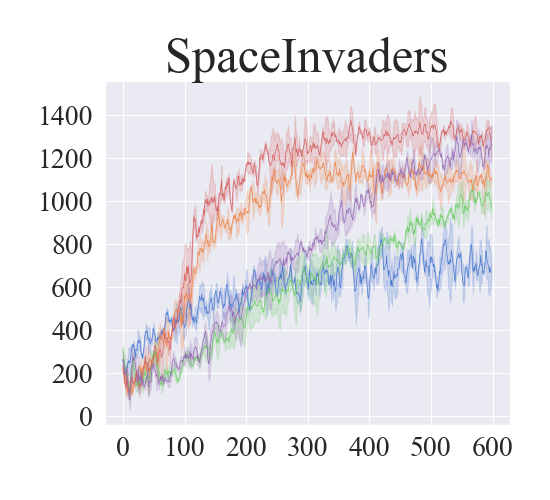}\\
			\end{minipage}%
		}%
		\subfigure{
			\begin{minipage}[t]{0.166\linewidth}
				\centering
				\includegraphics[width=1.05in]{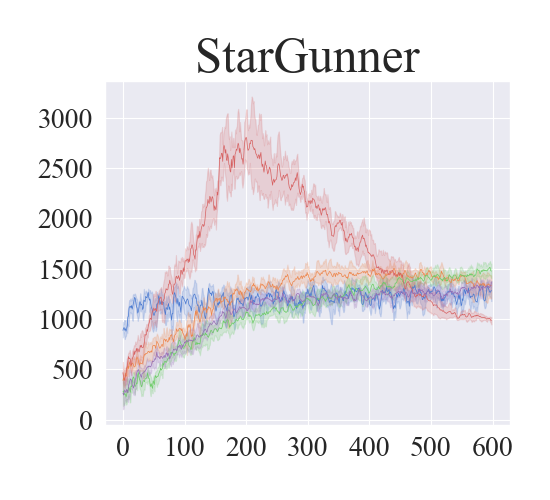}\\
			\end{minipage}%
		}%
		\subfigure{
			\begin{minipage}[t]{0.166\linewidth}
				\centering
				\includegraphics[width=1.05in]{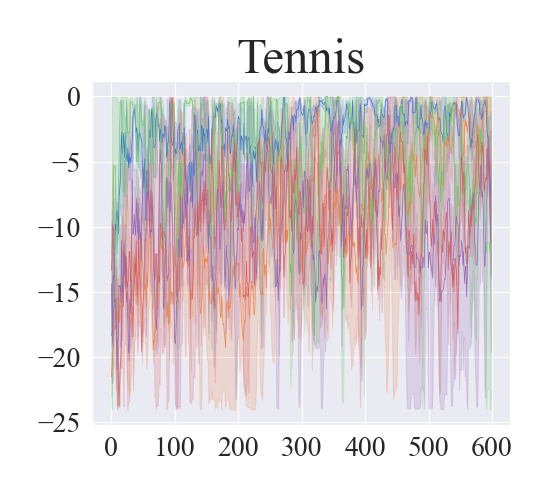}\\
			\end{minipage}%
		}%
		\subfigure{
			\begin{minipage}[t]{0.166\linewidth}
				\centering
				\includegraphics[width=1.05in]{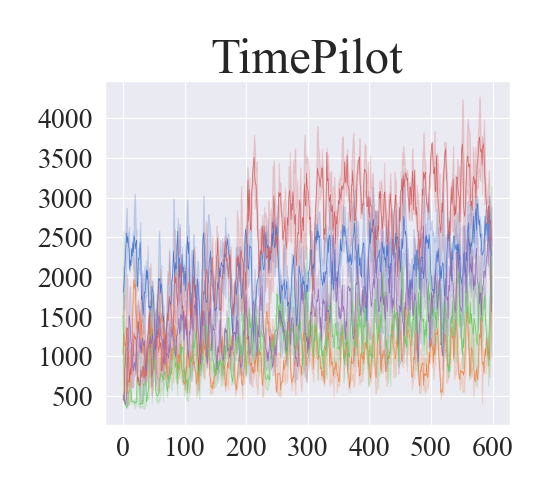}\\
			\end{minipage}%
		}%
		\subfigure{
			\begin{minipage}[t]{0.166\linewidth}
				\centering
				\includegraphics[width=1.05in]{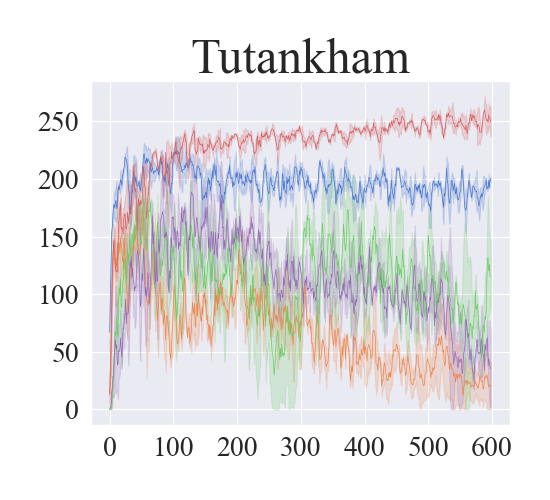}\\
			\end{minipage}%
		}%
		\vspace{-0.6cm}
		
		\subfigure{
			\begin{minipage}[t]{0.166\linewidth}
				\centering
				\includegraphics[width=1.05in]{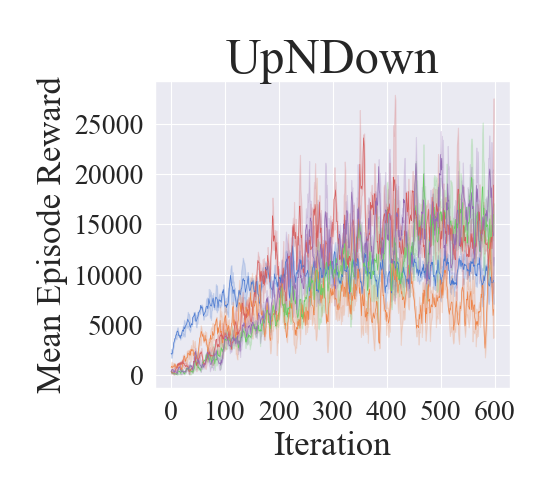}\\
			\end{minipage}%
		}%
		\subfigure{
			\begin{minipage}[t]{0.166\linewidth}
				\centering
				\includegraphics[width=1.05in]{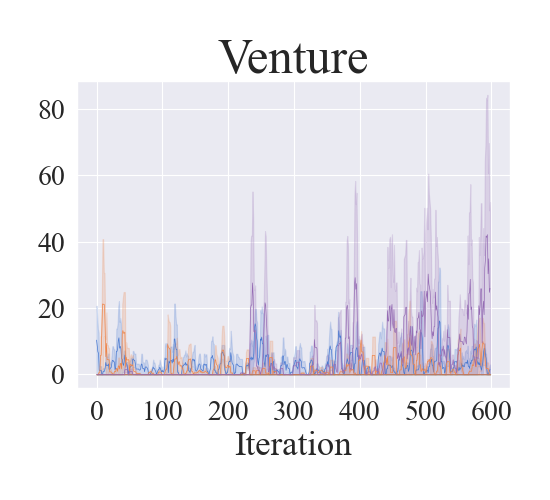}\\
			\end{minipage}%
		}%
		\subfigure{
			\begin{minipage}[t]{0.166\linewidth}
				\centering
				\includegraphics[width=1.05in]{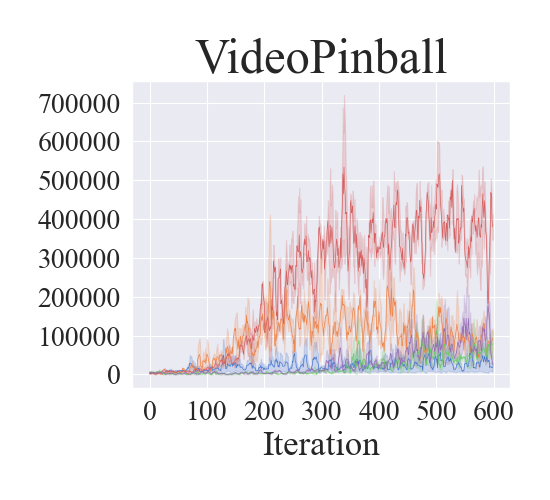}\\
			\end{minipage}%
		}%
		\subfigure{
			\begin{minipage}[t]{0.166\linewidth}
				\centering
				\includegraphics[width=1.05in]{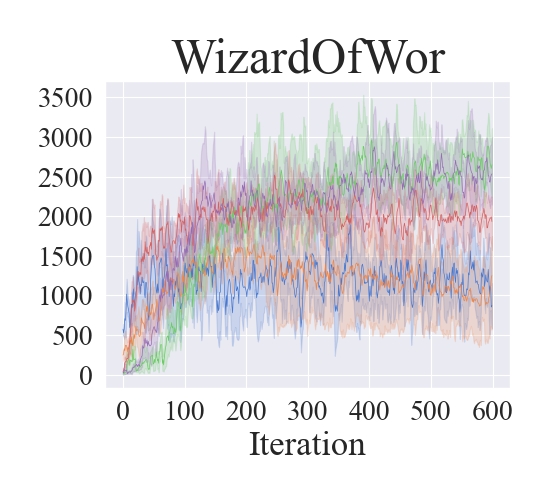}\\
			\end{minipage}%
		}%
		\subfigure{
			\begin{minipage}[t]{0.166\linewidth}
				\centering
				\includegraphics[width=1.05in]{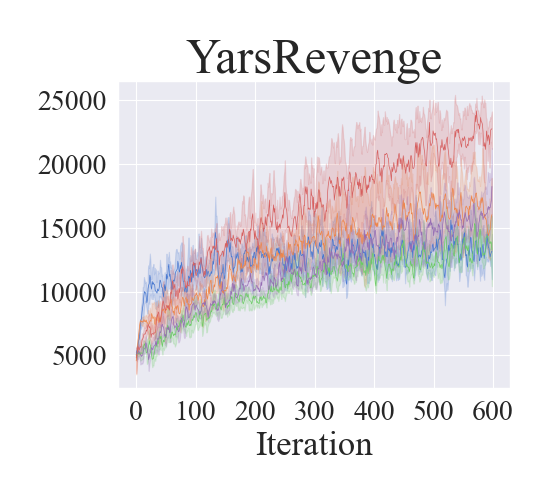}\\
			\end{minipage}%
		}%
		\subfigure{
			\begin{minipage}[t]{0.166\linewidth}
				\centering
				\includegraphics[width=1.05in]{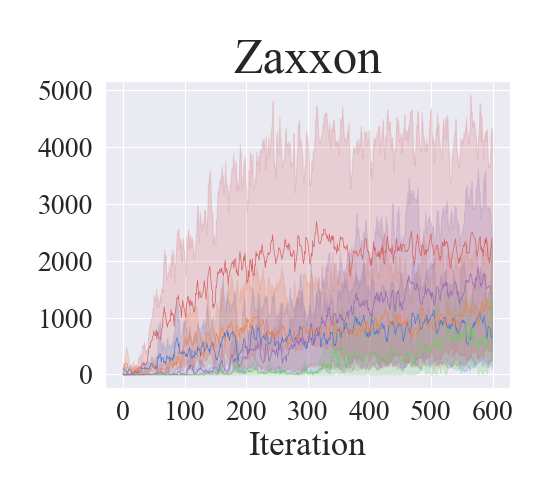}\\
			\end{minipage}%
		}%

		\centering
		\caption{\textbf{Learning curves of all $60$ Atari $2600$ games on poor dataset}}
		\label{fig: Learning curves of all $60$ Atari $2600$ games on poor dataset}
								
	\end{figure*}

\begin{figure*}[!htb]
		\centering

		\subfigure{
			\begin{minipage}[t]{\linewidth}
				\centering
				\includegraphics[width=4in]{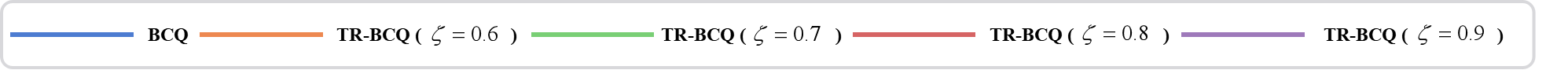}\\
			\end{minipage}%
		}%

		\subfigure{
			\begin{minipage}[t]{0.166\linewidth}
				\centering
				\includegraphics[width=1.05in]{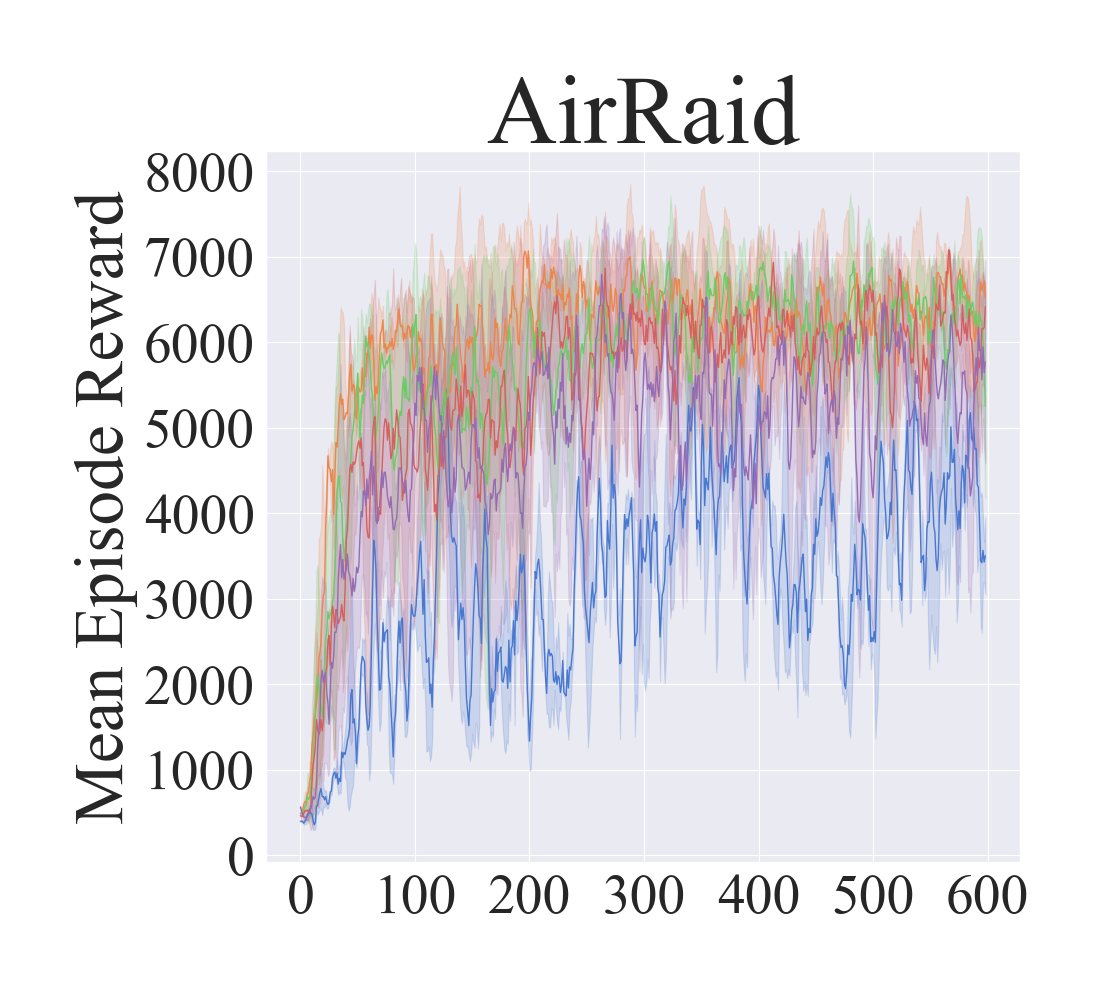}\\
			\end{minipage}%
		}%
		\subfigure{
			\begin{minipage}[t]{0.166\linewidth}
				\centering
				\includegraphics[width=1.05in]{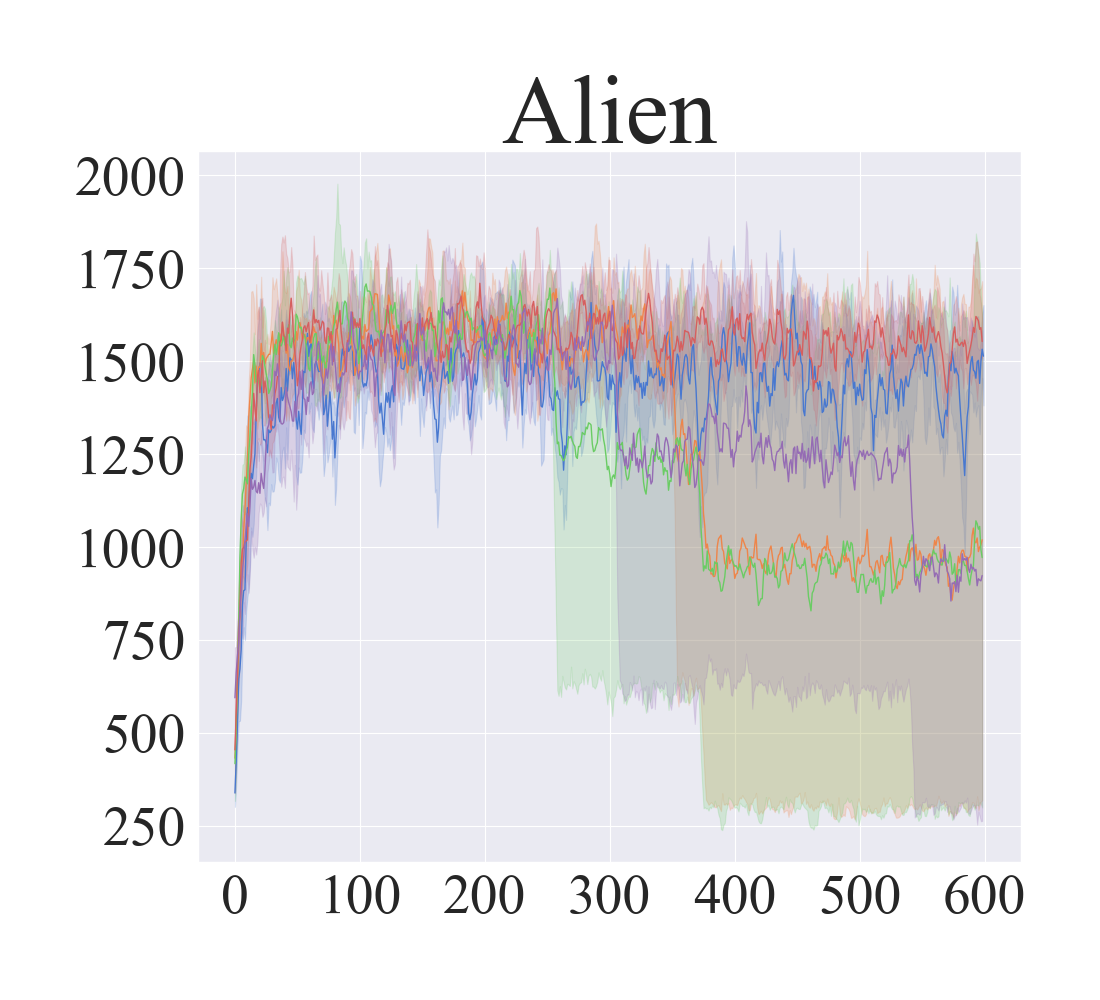}\\
			\end{minipage}%
		}%
		\subfigure{
			\begin{minipage}[t]{0.166\linewidth}
				\centering
				\includegraphics[width=1.05in]{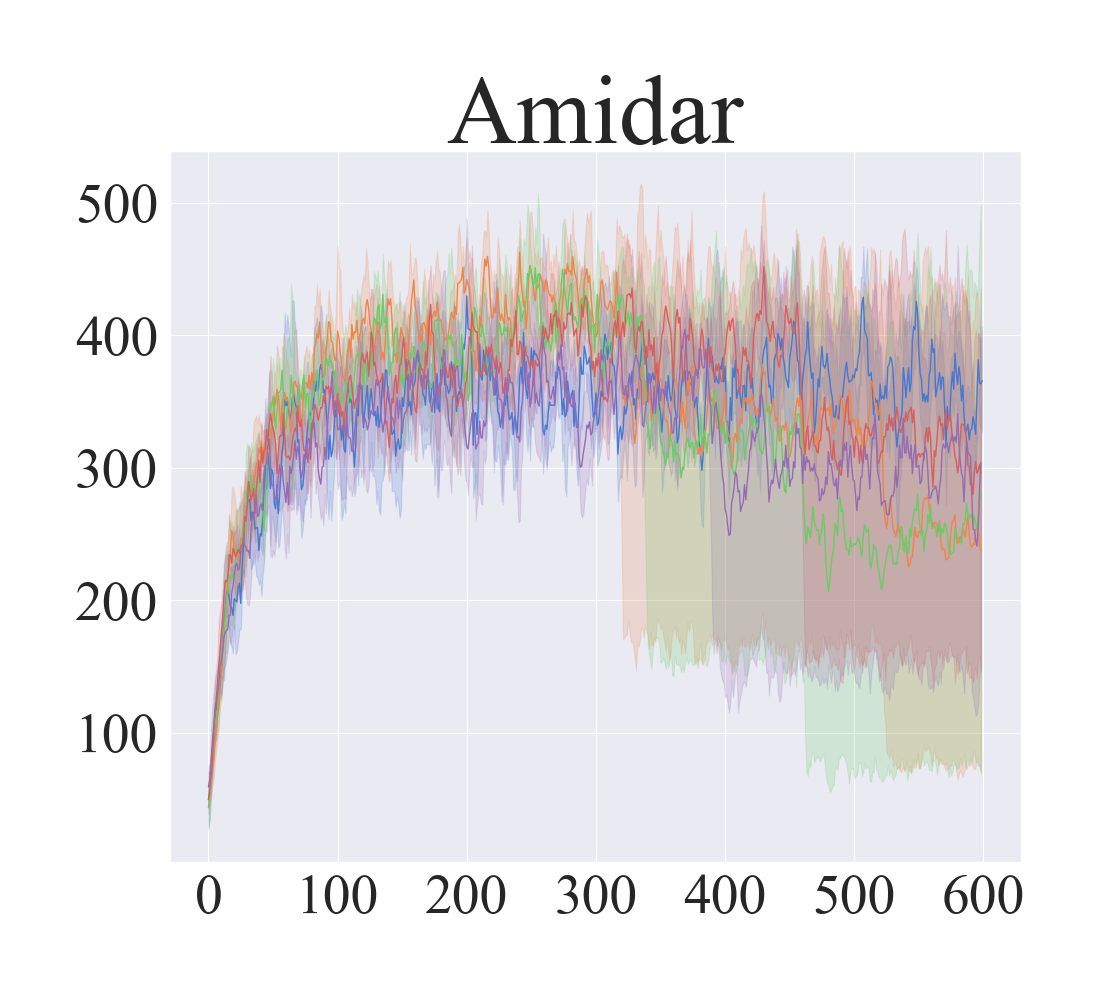}\\
			\end{minipage}%
		}%
		\subfigure{
			\begin{minipage}[t]{0.166\linewidth}
				\centering
				\includegraphics[width=1.05in]{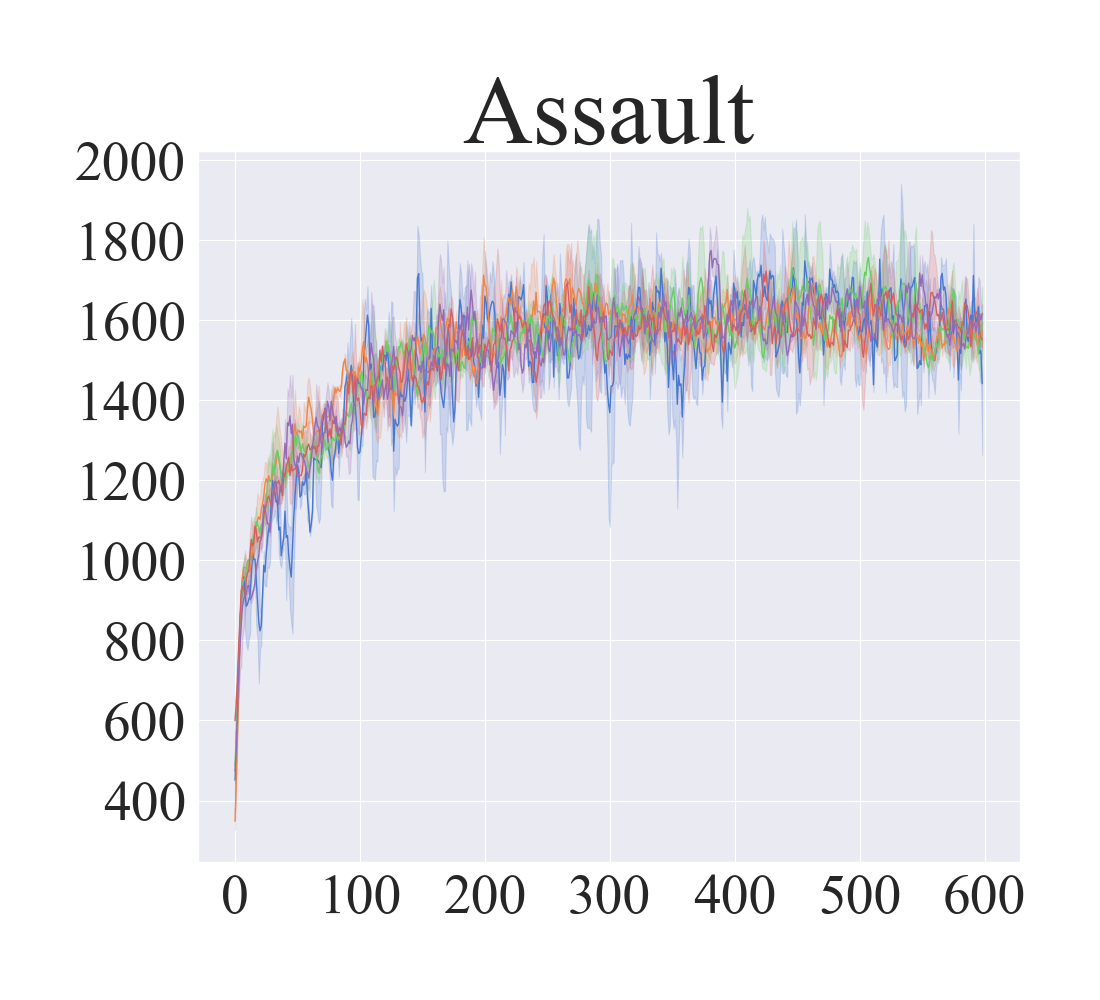}\\
			\end{minipage}%
		}%
		\subfigure{
			\begin{minipage}[t]{0.166\linewidth}
				\centering
				\includegraphics[width=1.05in]{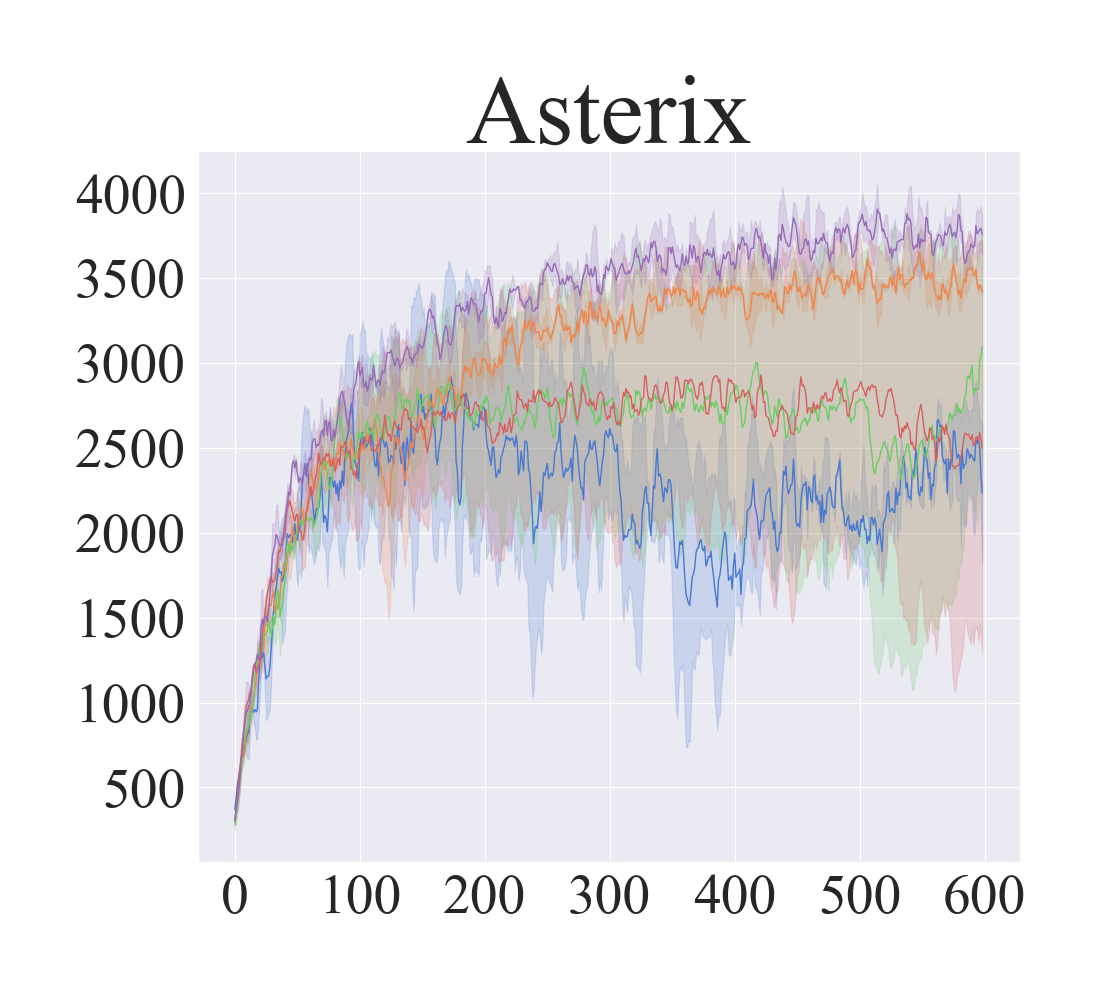}\\
			\end{minipage}%
		}%
		\subfigure{
			\begin{minipage}[t]{0.166\linewidth}
				\centering
				\includegraphics[width=1.05in]{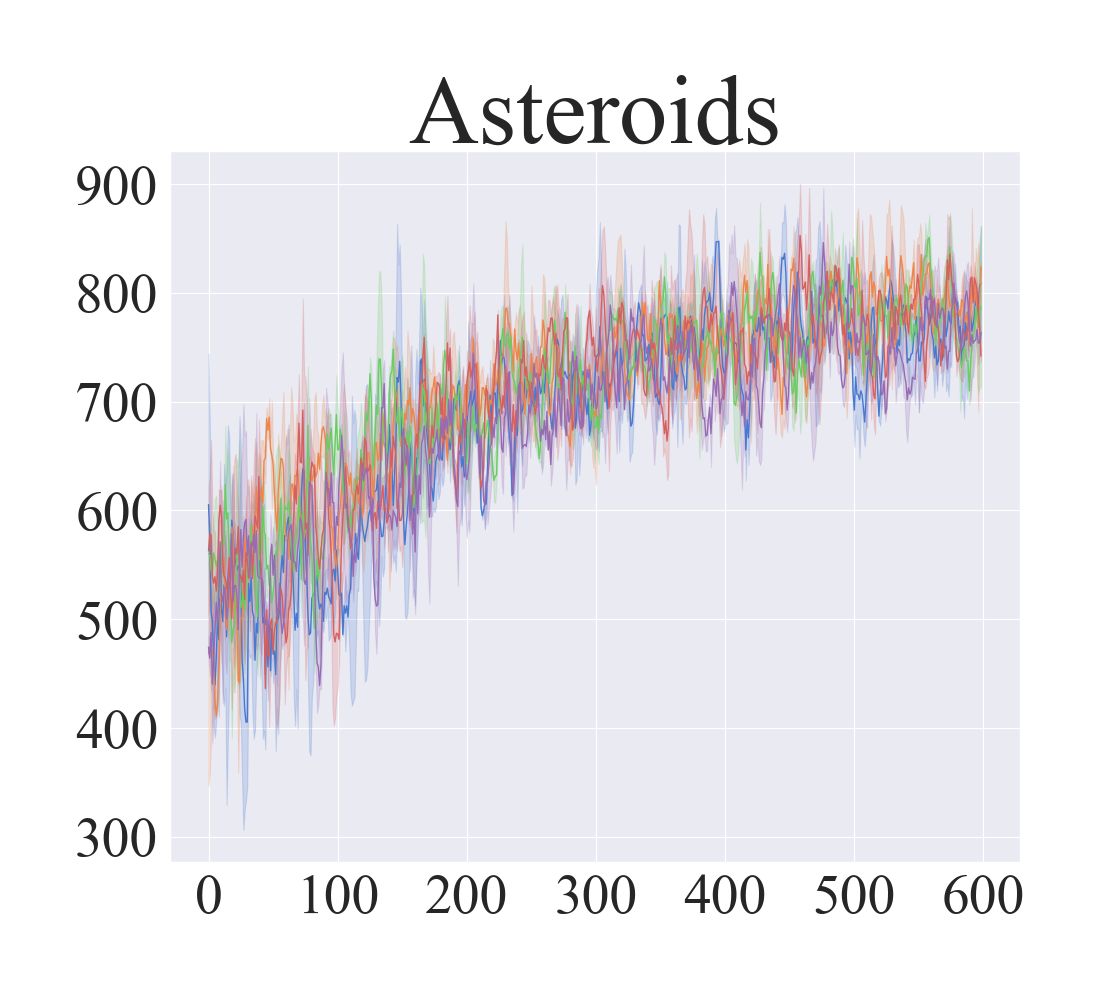}\\\
			\end{minipage}%
		}%
		\vspace{-1.0cm}
		
		\subfigure{
			\begin{minipage}[t]{0.166\linewidth}
				\centering
				\includegraphics[width=1.05in]{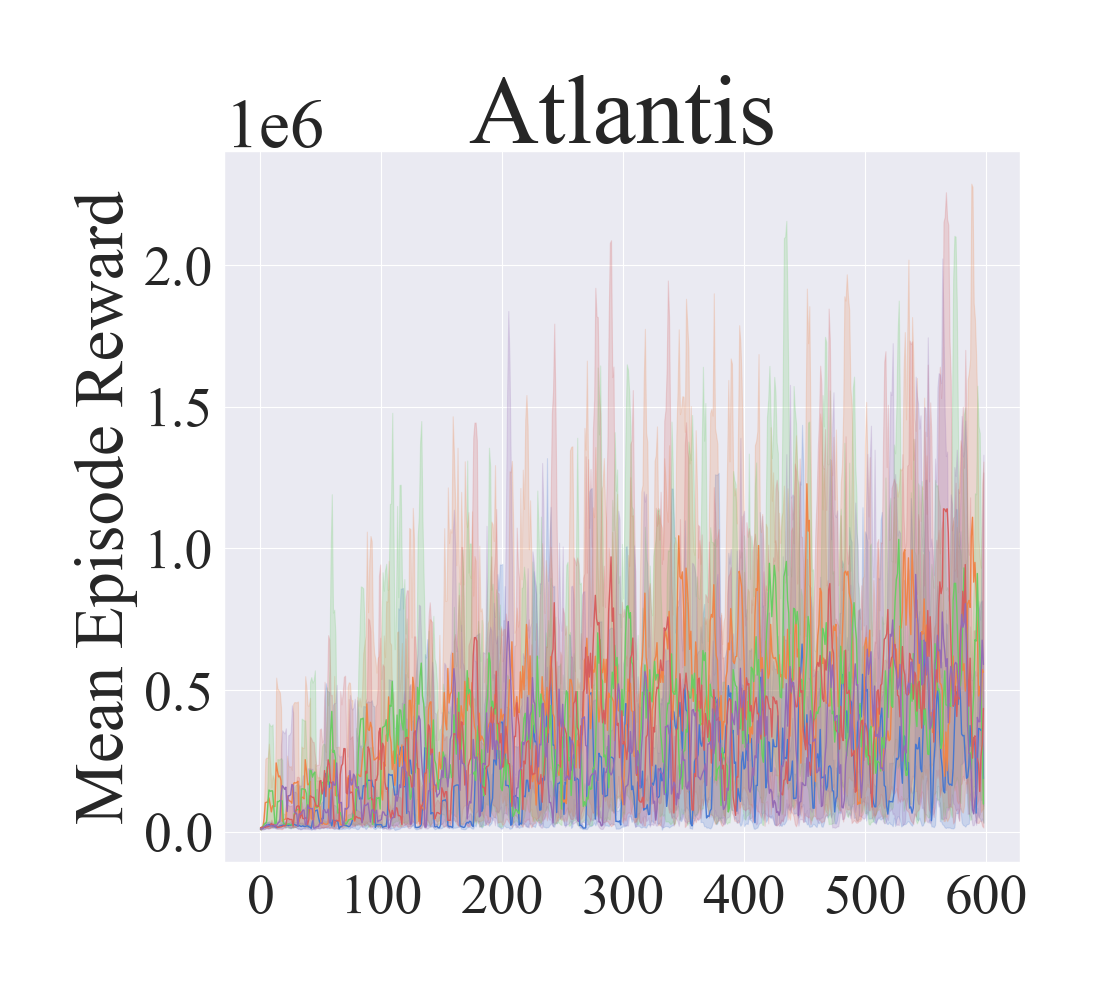}\\
			\end{minipage}%
		}%
		\subfigure{
			\begin{minipage}[t]{0.166\linewidth}
				\centering
				\includegraphics[width=1.05in]{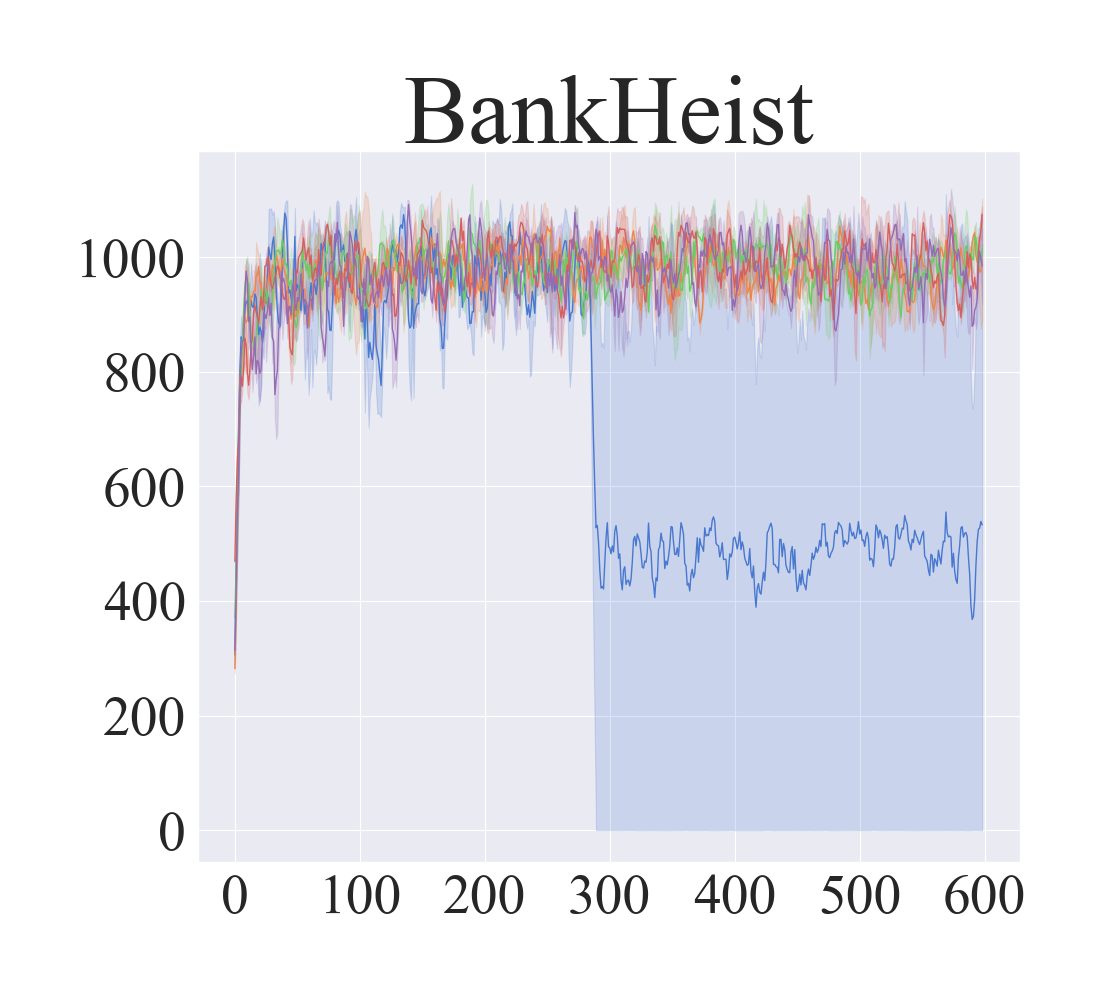}\\
			\end{minipage}%
		}%
		\subfigure{
			\begin{minipage}[t]{0.166\linewidth}
				\centering
				\includegraphics[width=1.05in]{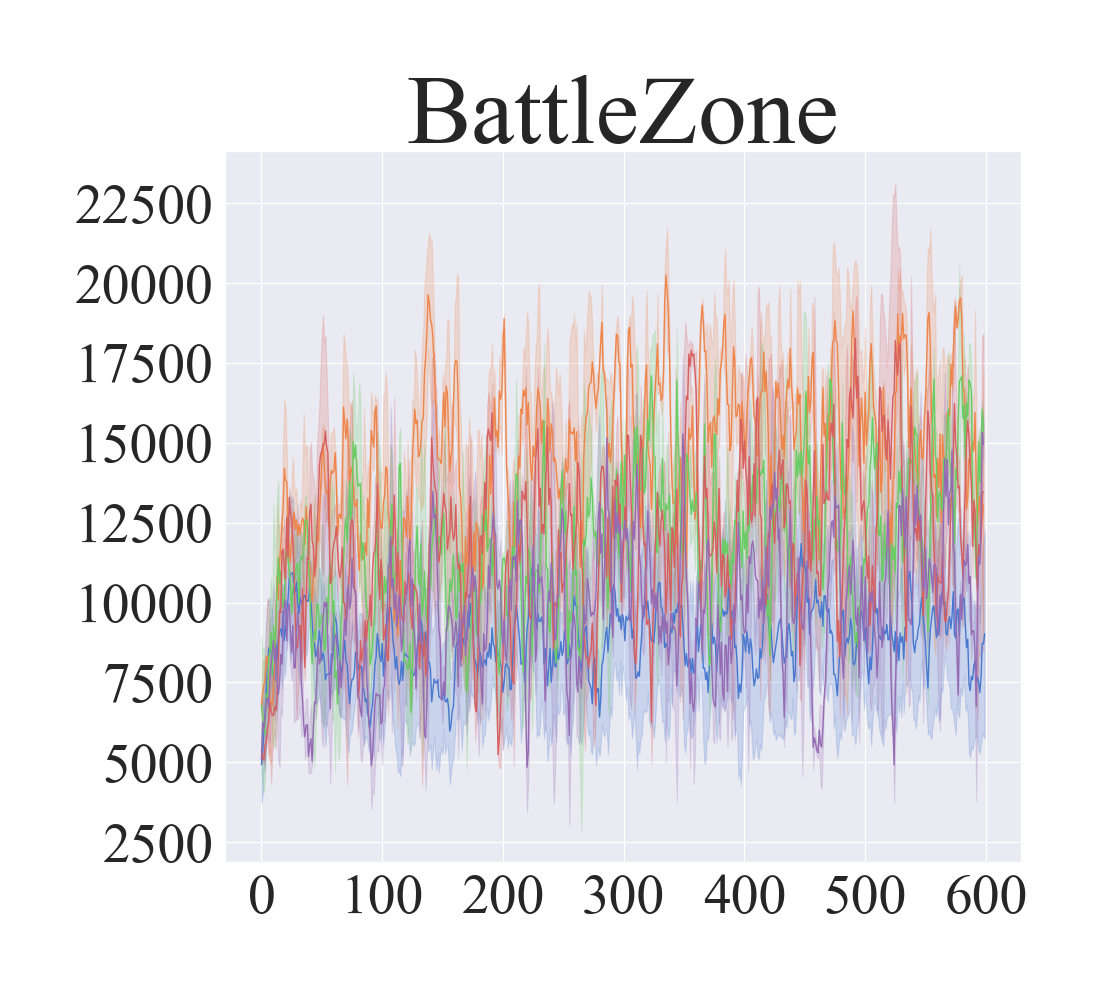}\\
			\end{minipage}%
		}%
		\subfigure{
			\begin{minipage}[t]{0.166\linewidth}
				\centering
				\includegraphics[width=1.05in]{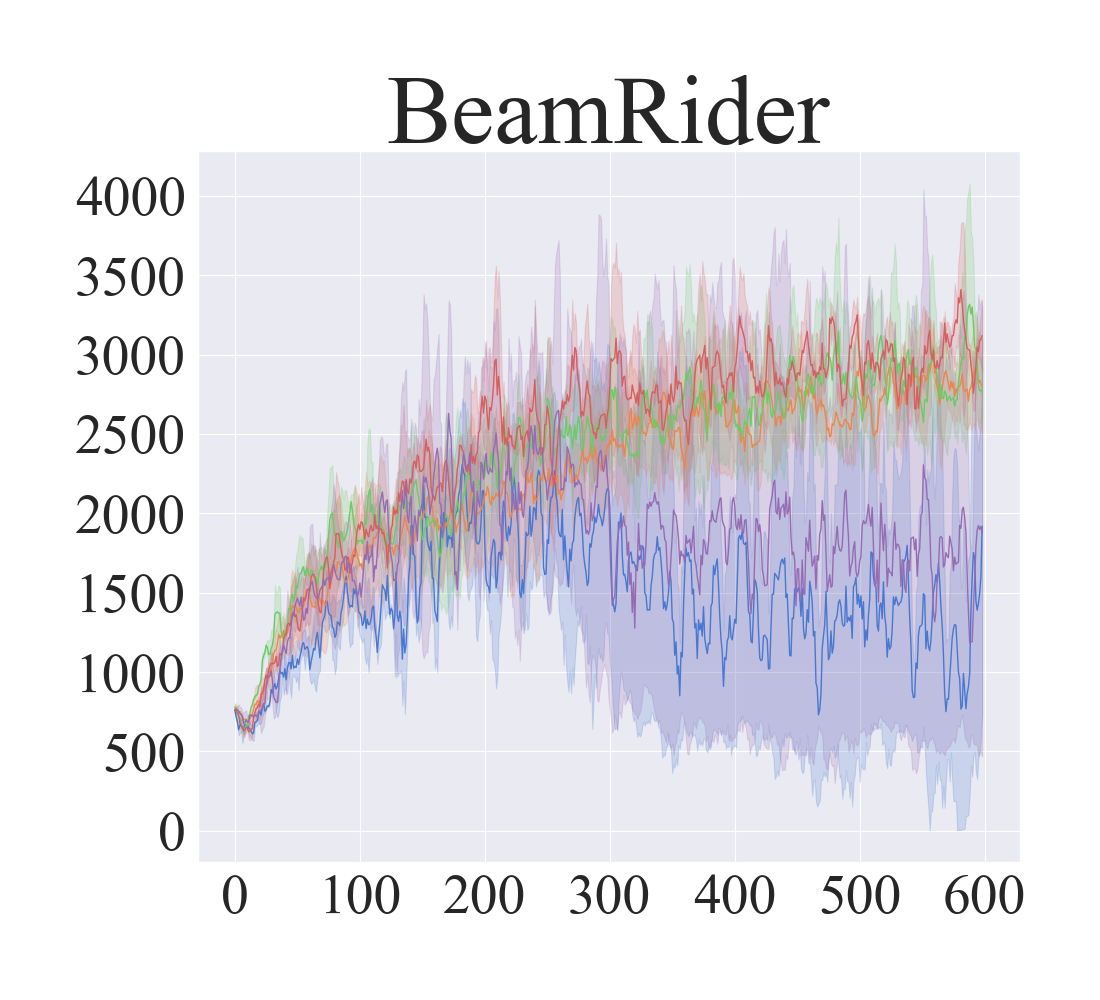}\\
			\end{minipage}%
		}%
		\subfigure{
			\begin{minipage}[t]{0.166\linewidth}
				\centering
				\includegraphics[width=1.05in]{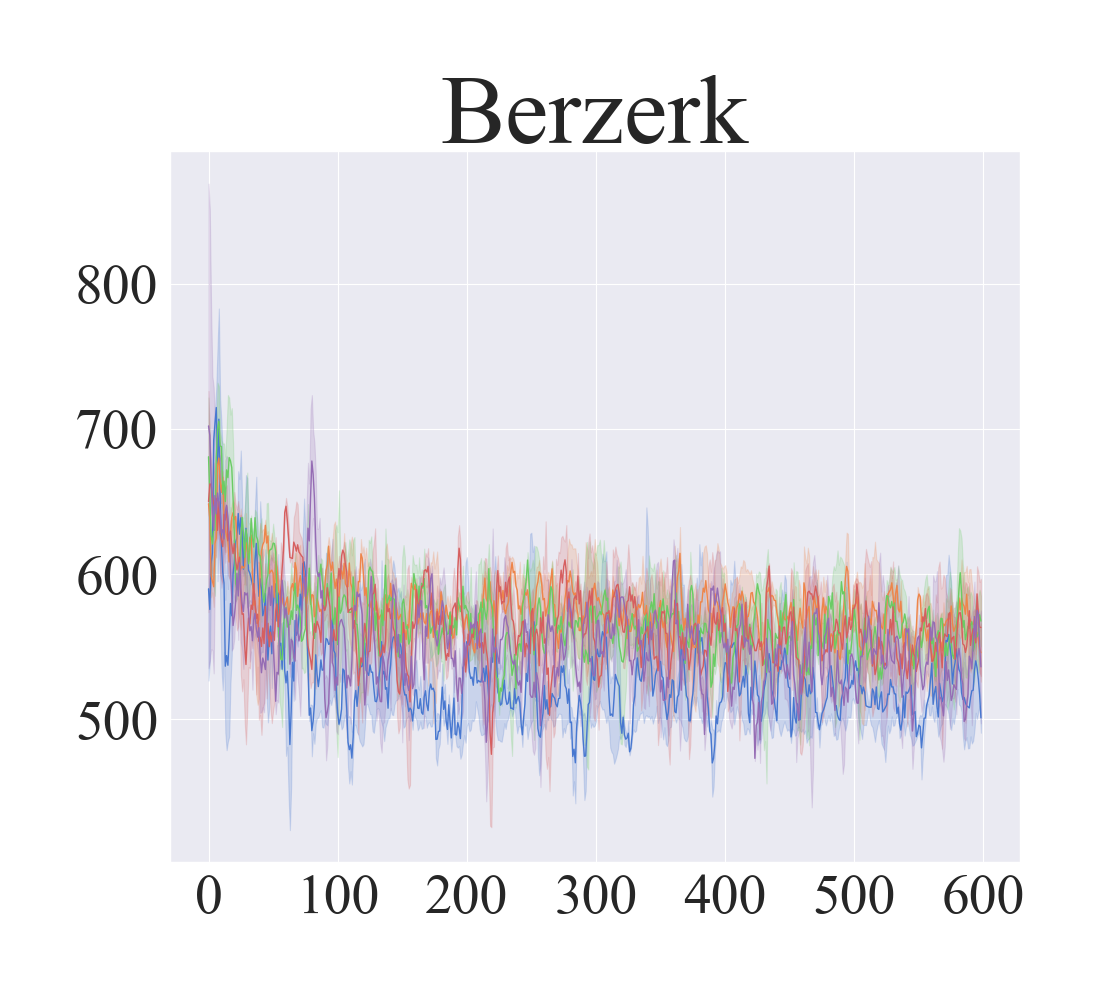}\\
			\end{minipage}%
		}%
		\subfigure{
			\begin{minipage}[t]{0.166\linewidth}
				\centering
				\includegraphics[width=1.05in]{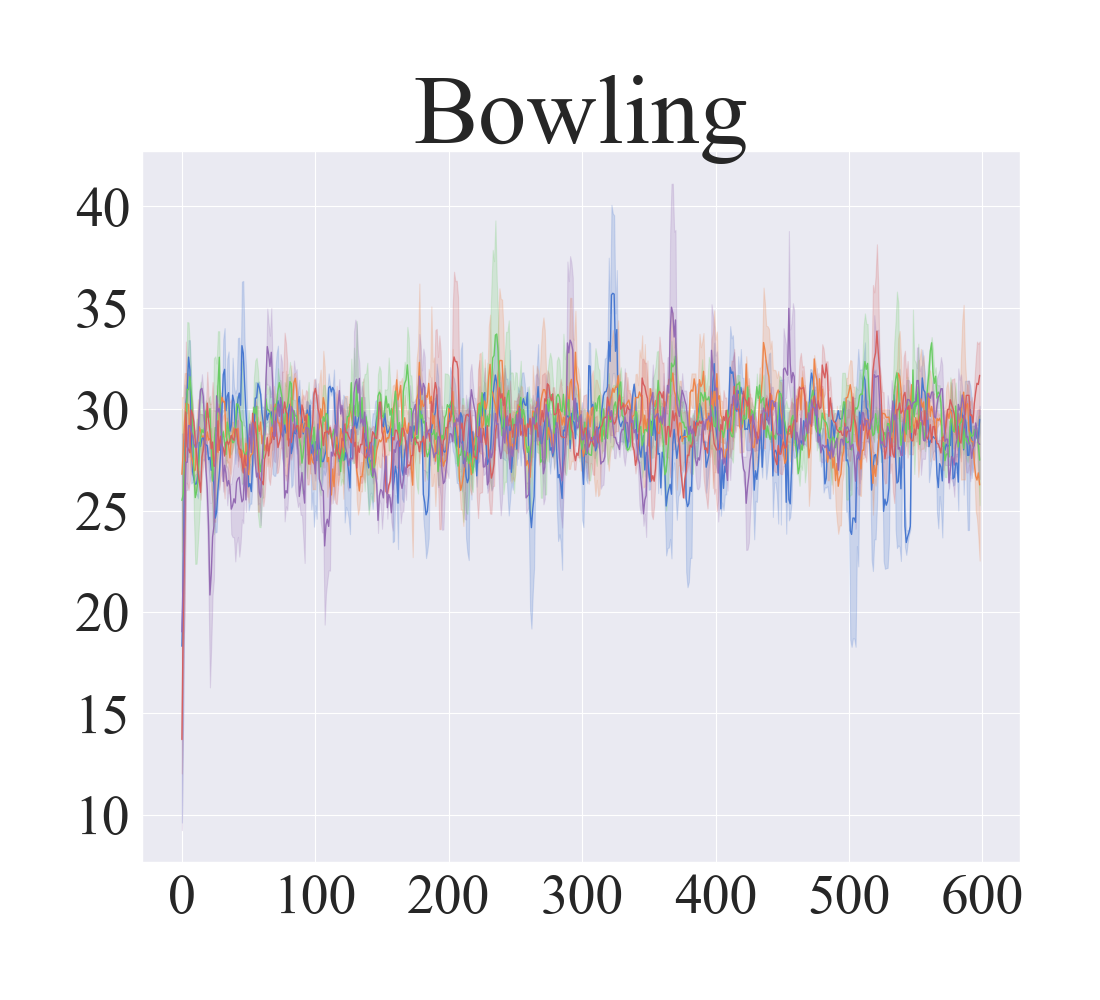}\\
			\end{minipage}%
		}%
		\vspace{-0.6cm}
		
		\subfigure{
			\begin{minipage}[t]{0.166\linewidth}
				\centering
				\includegraphics[width=1.05in]{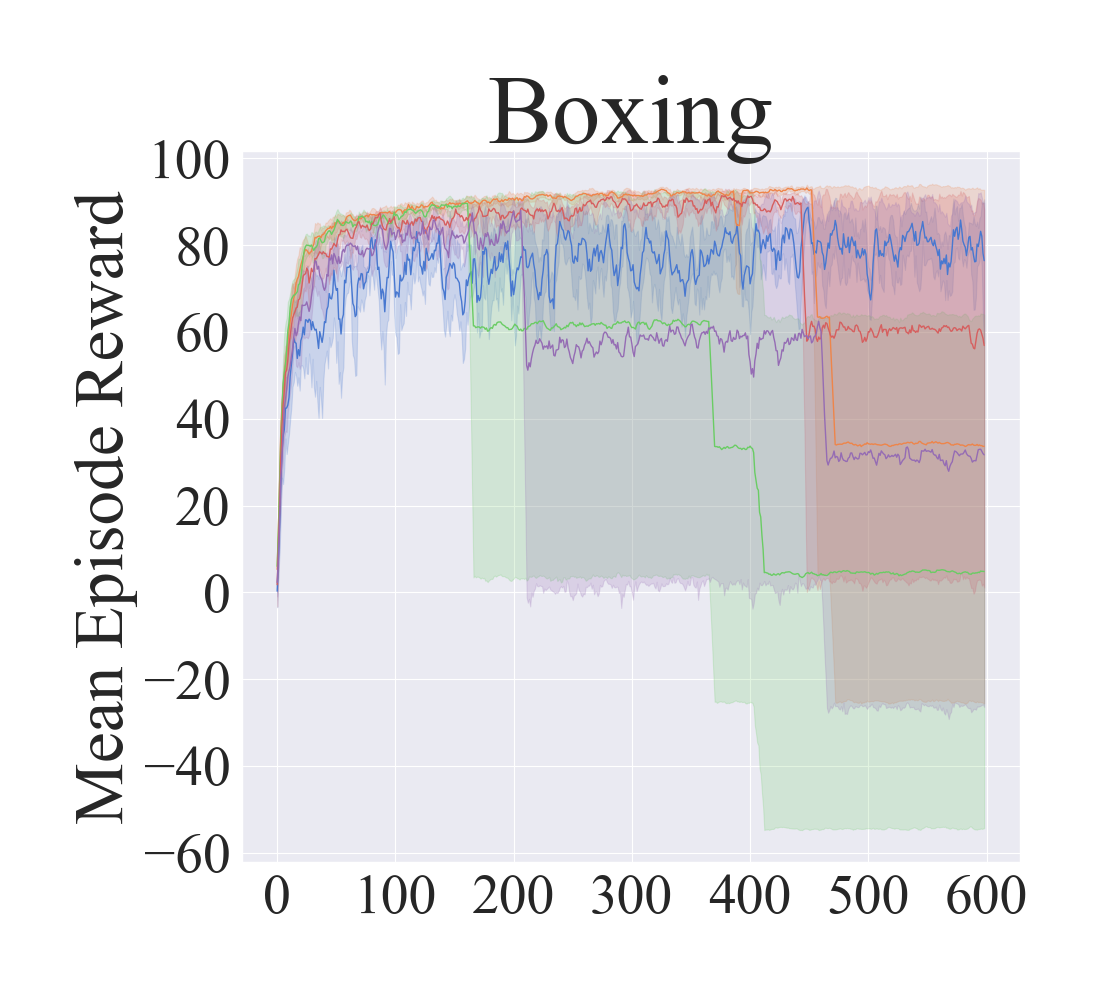}\\
			\end{minipage}%
		}%
		\subfigure{
			\begin{minipage}[t]{0.166\linewidth}
				\centering
				\includegraphics[width=1.05in]{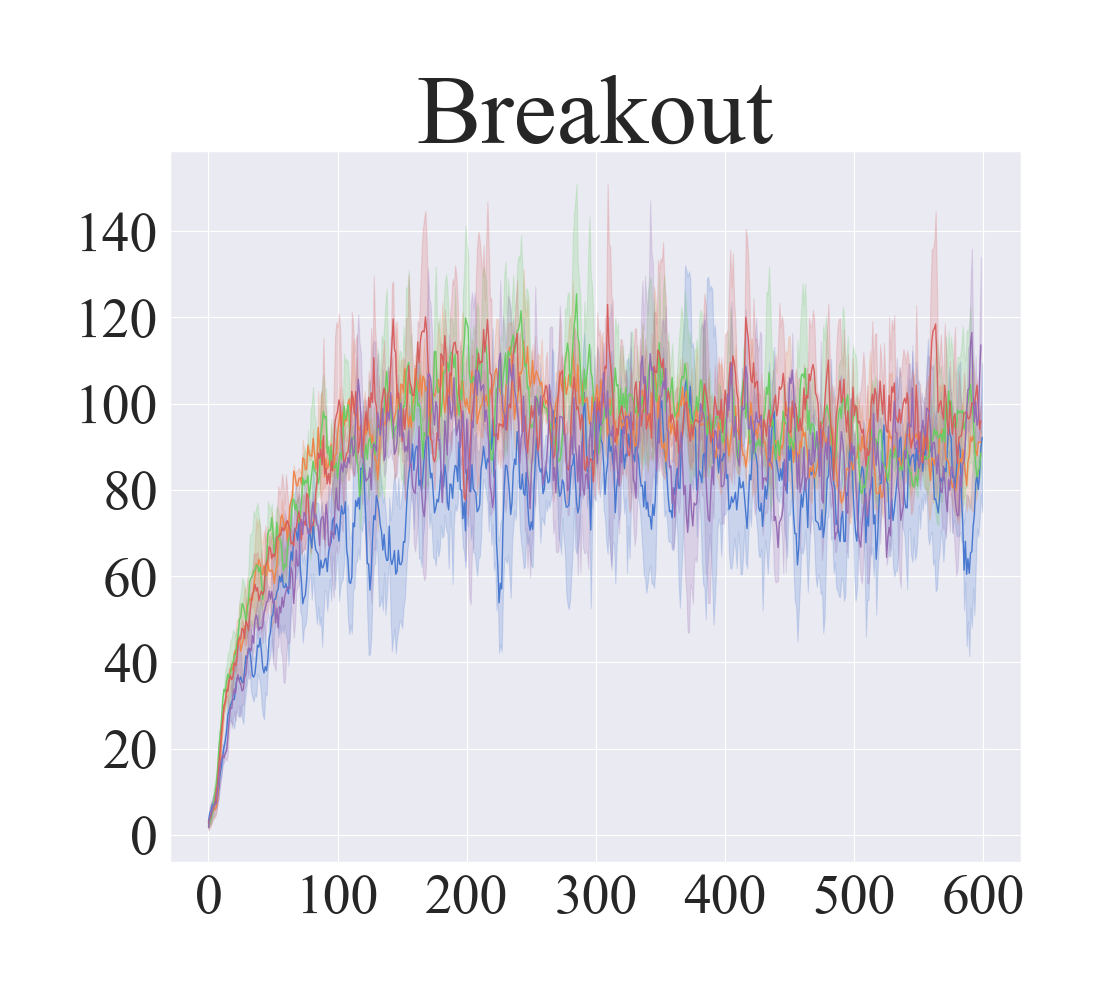}\\
			\end{minipage}%
		}%
		\subfigure{
			\begin{minipage}[t]{0.166\linewidth}
				\centering
				\includegraphics[width=1.05in]{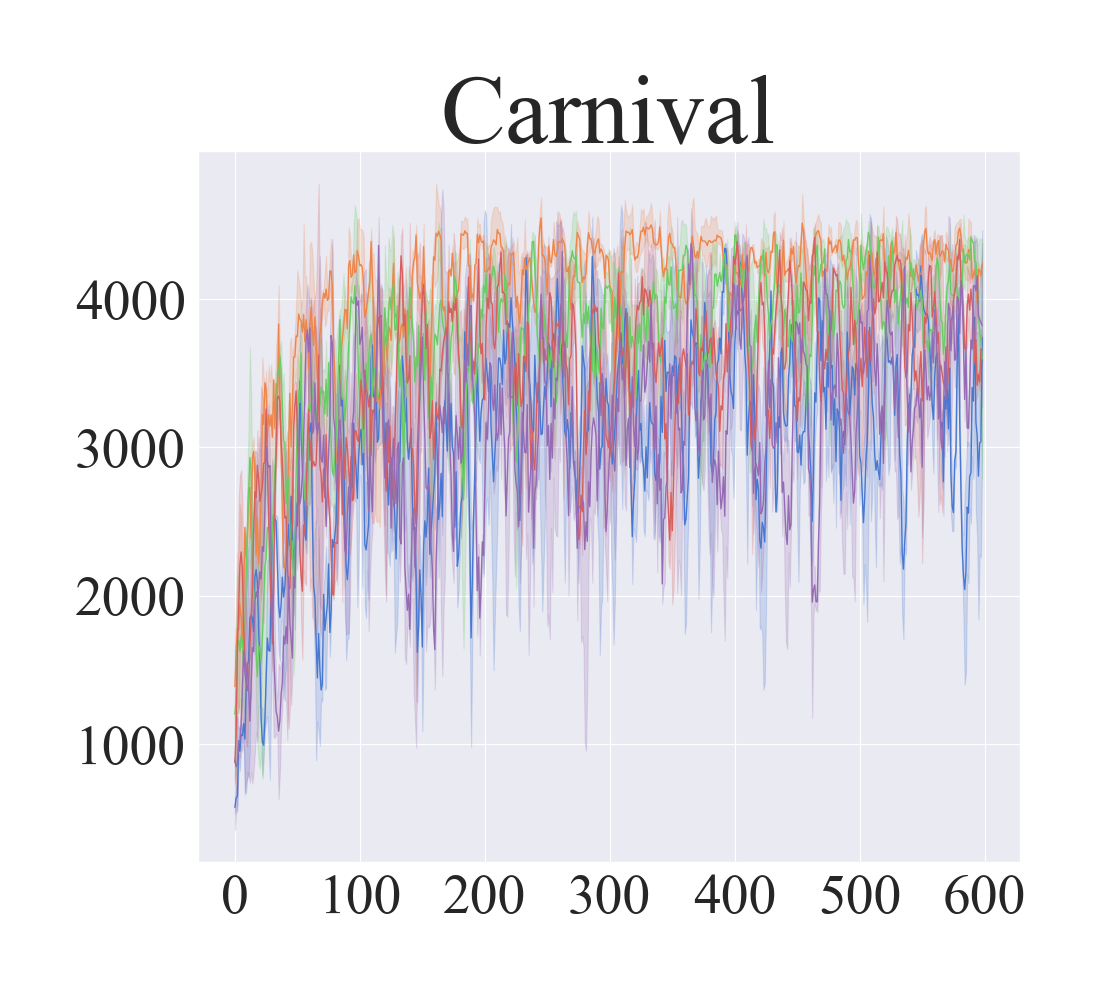}\\
			\end{minipage}%
		}%
		\subfigure{
			\begin{minipage}[t]{0.166\linewidth}
				\centering
				\includegraphics[width=1.05in]{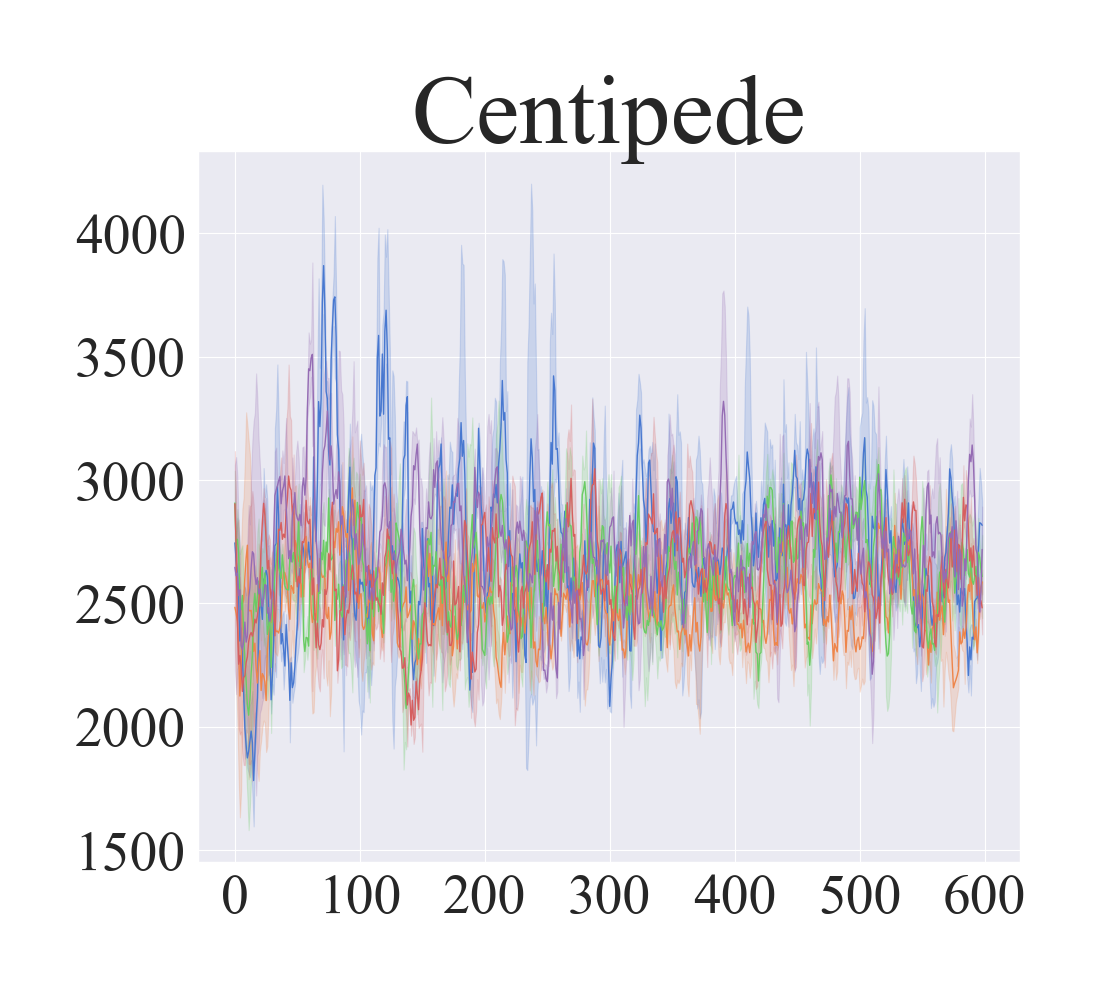}\\
			\end{minipage}%
		}%
		\subfigure{
			\begin{minipage}[t]{0.166\linewidth}
				\centering
				\includegraphics[width=1.05in]{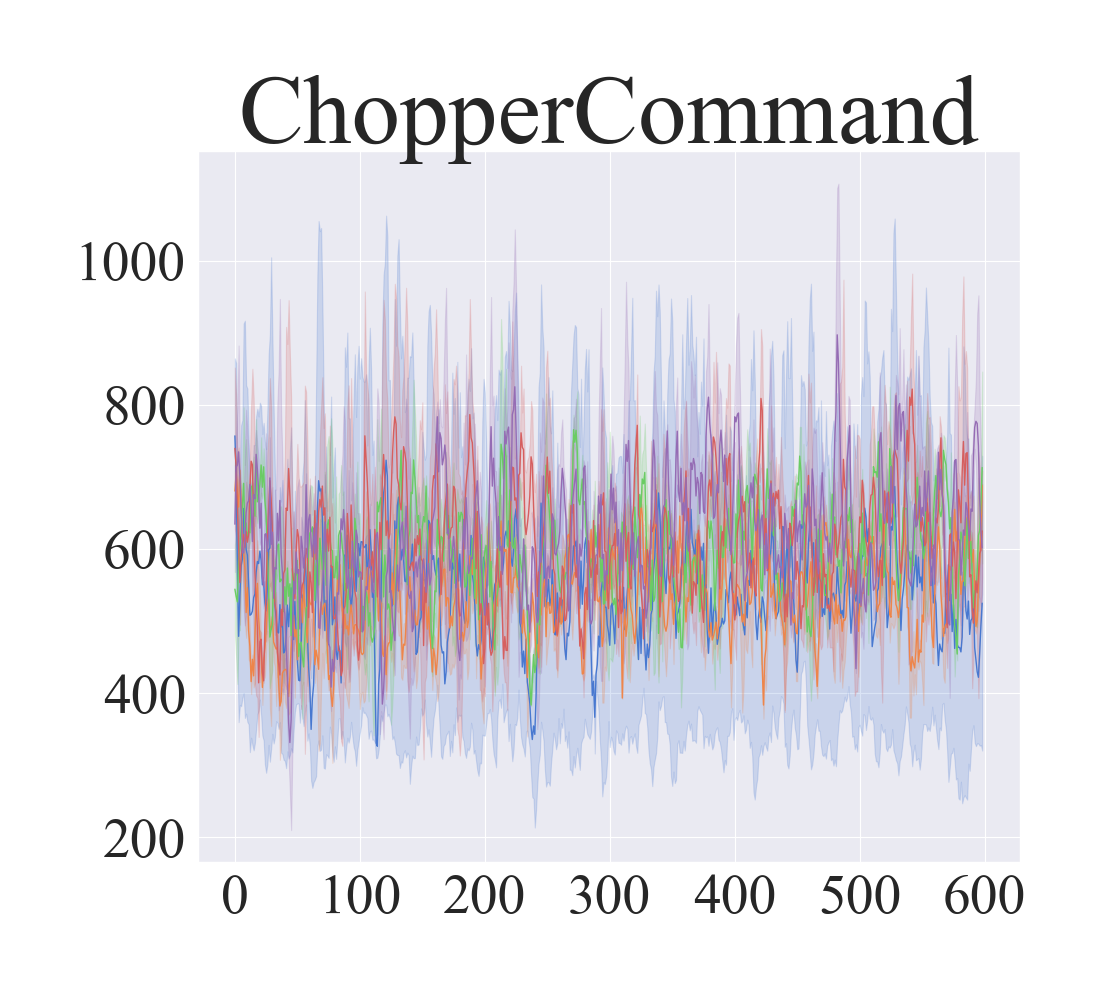}\\
			\end{minipage}%
		}%
		\subfigure{
			\begin{minipage}[t]{0.166\linewidth}
				\centering
				\includegraphics[width=1.05in]{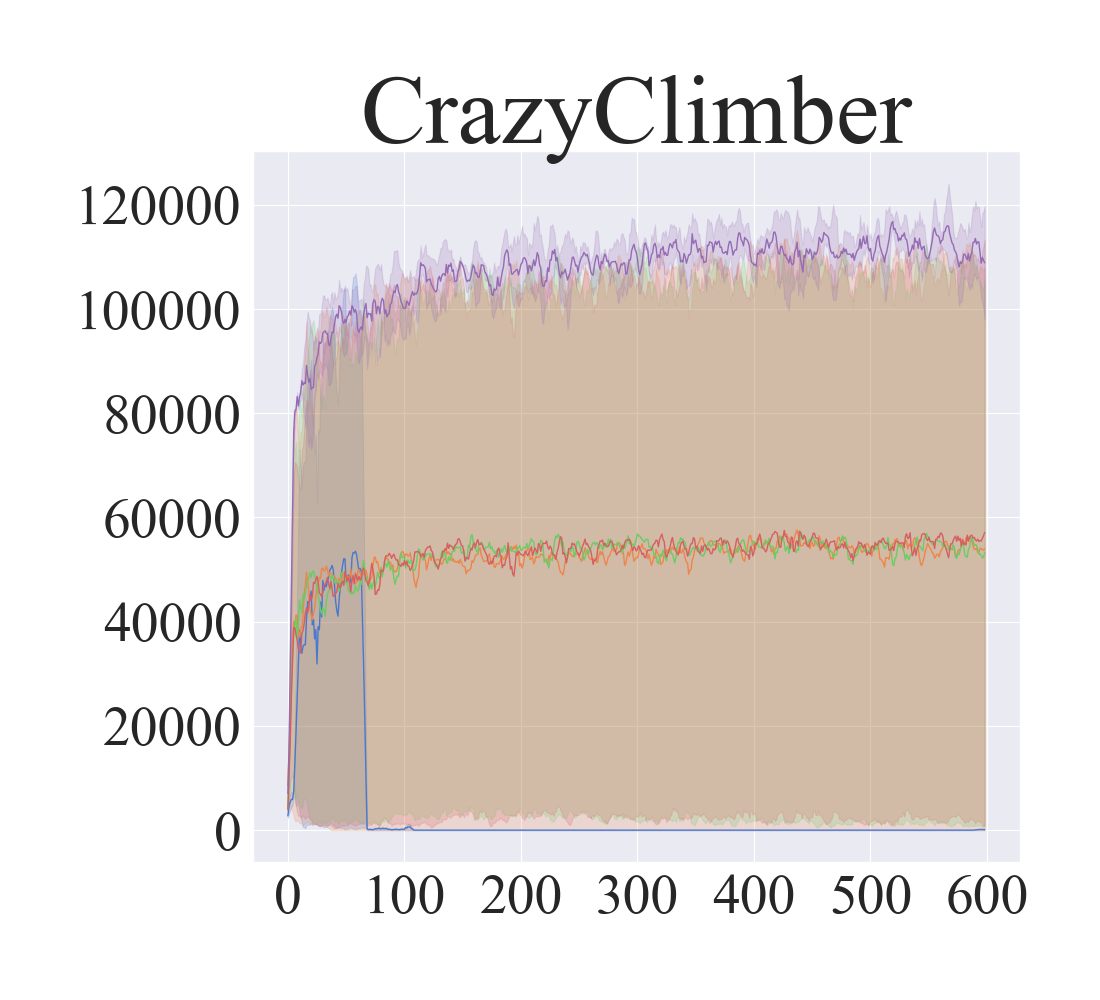}\\
			\end{minipage}%
		}%
		\vspace{-0.6cm}
		
		\subfigure{
			\begin{minipage}[t]{0.166\linewidth}
				\centering
				\includegraphics[width=1.05in]{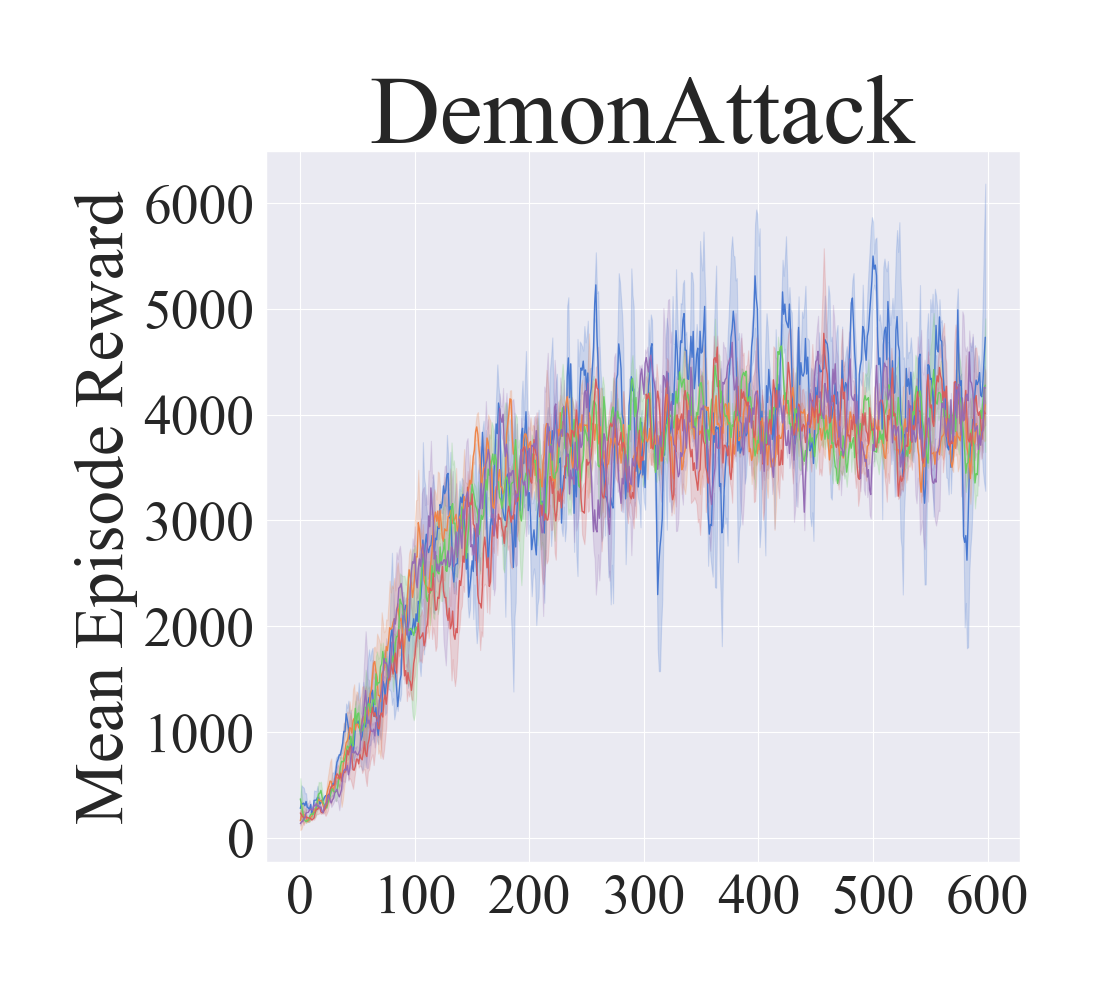}\\
			\end{minipage}%
		}%
		\subfigure{
			\begin{minipage}[t]{0.166\linewidth}
				\centering
				\includegraphics[width=1.05in]{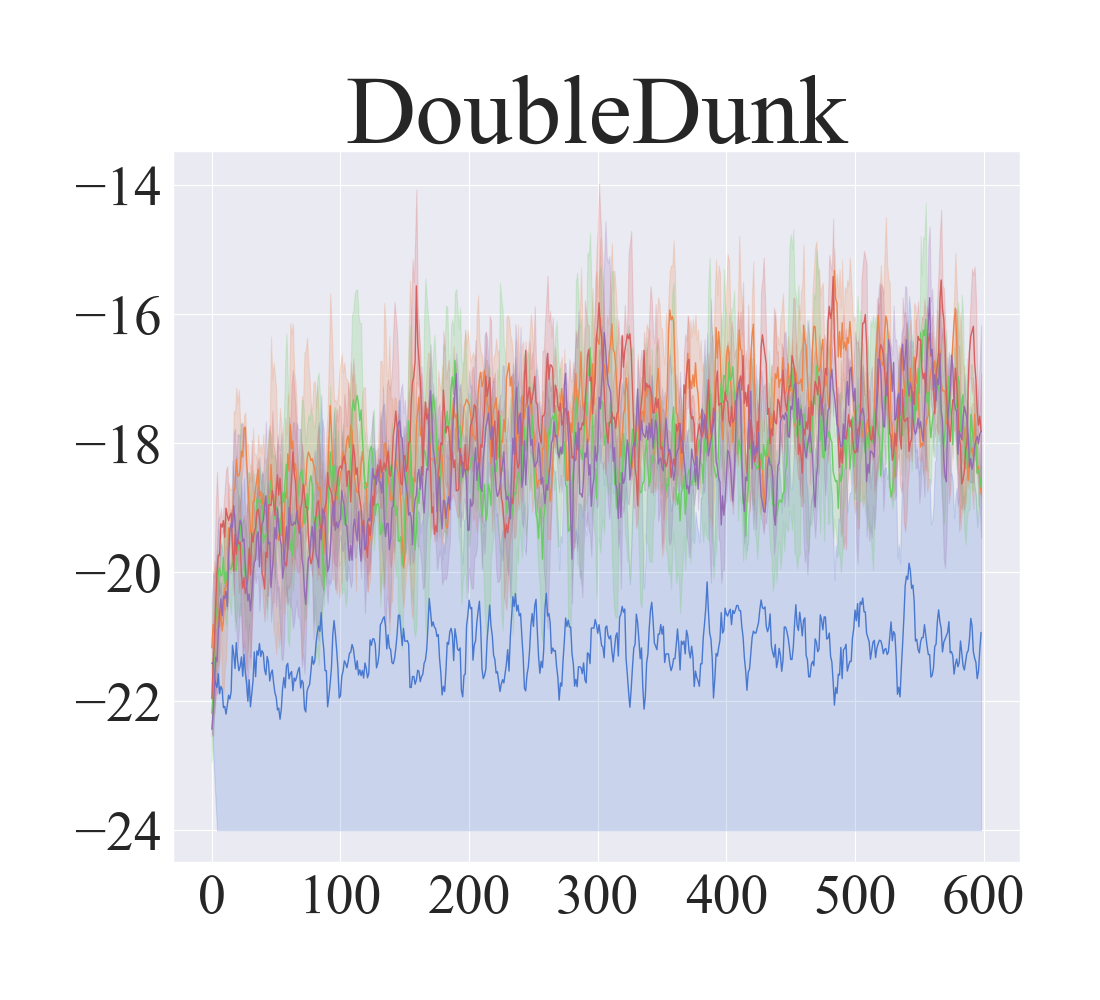}\\
			\end{minipage}%
		}%
		\subfigure{
			\begin{minipage}[t]{0.166\linewidth}
				\centering
				\includegraphics[width=1.05in]{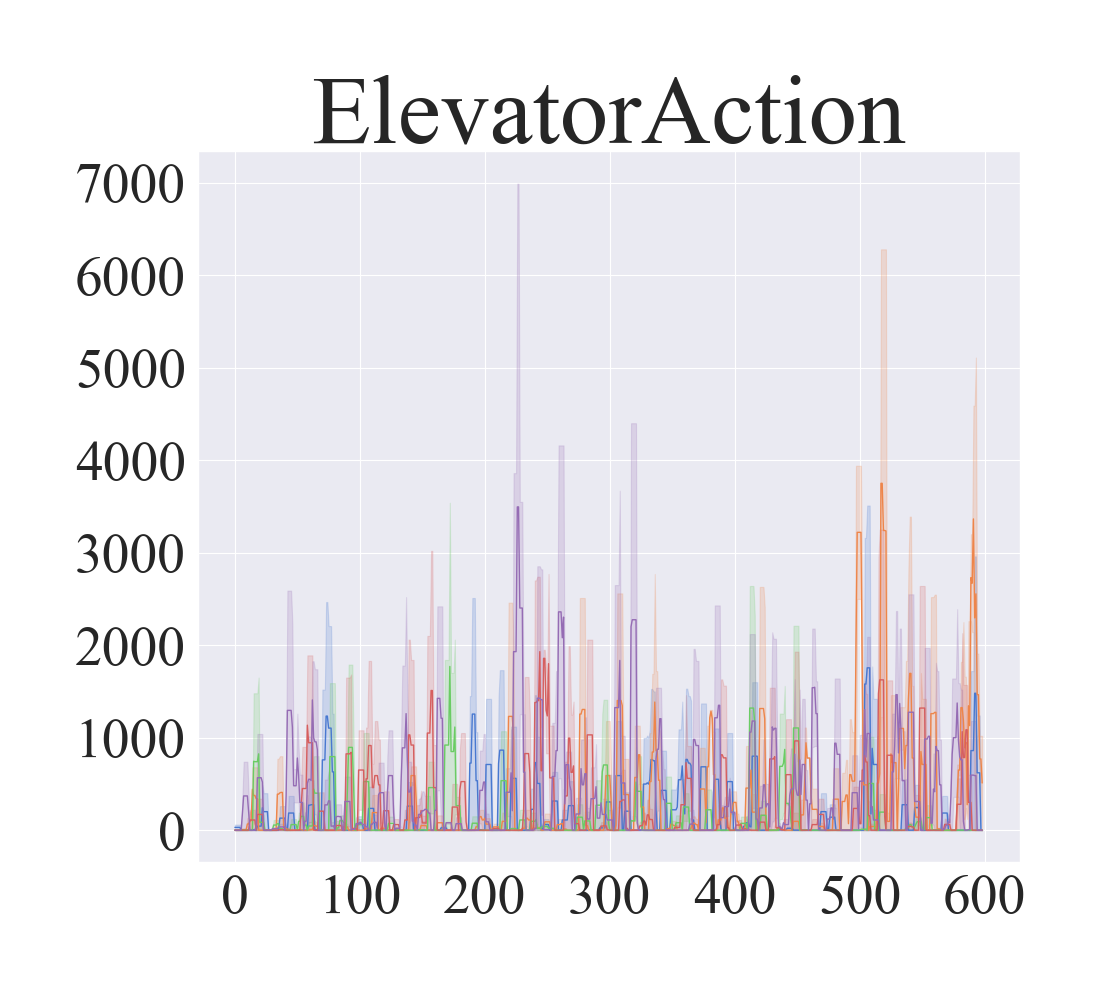}\\
			\end{minipage}%
		}%
		\subfigure{
			\begin{minipage}[t]{0.166\linewidth}
				\centering
				\includegraphics[width=1.05in]{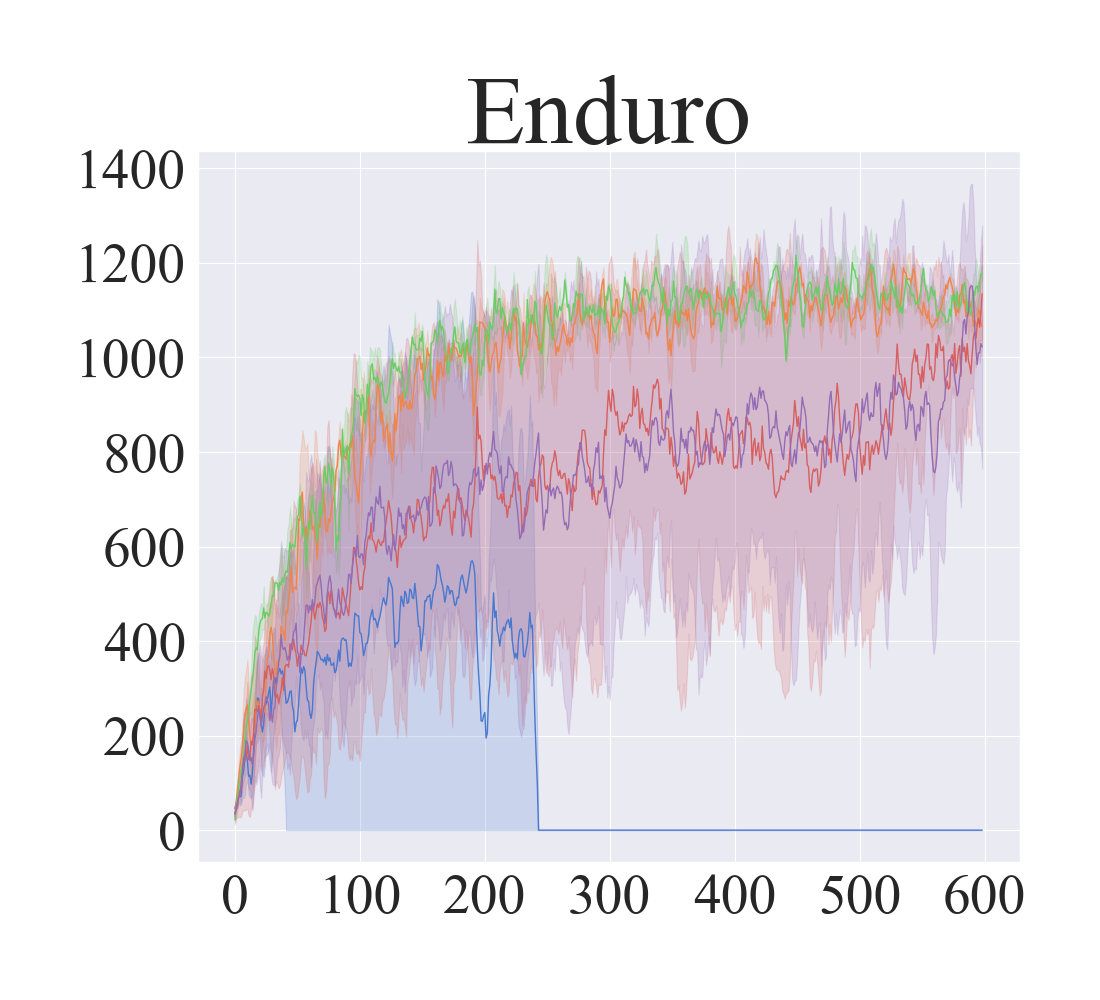}\\
			\end{minipage}%
		}%
		\subfigure{
			\begin{minipage}[t]{0.166\linewidth}
				\centering
				\includegraphics[width=1.05in]{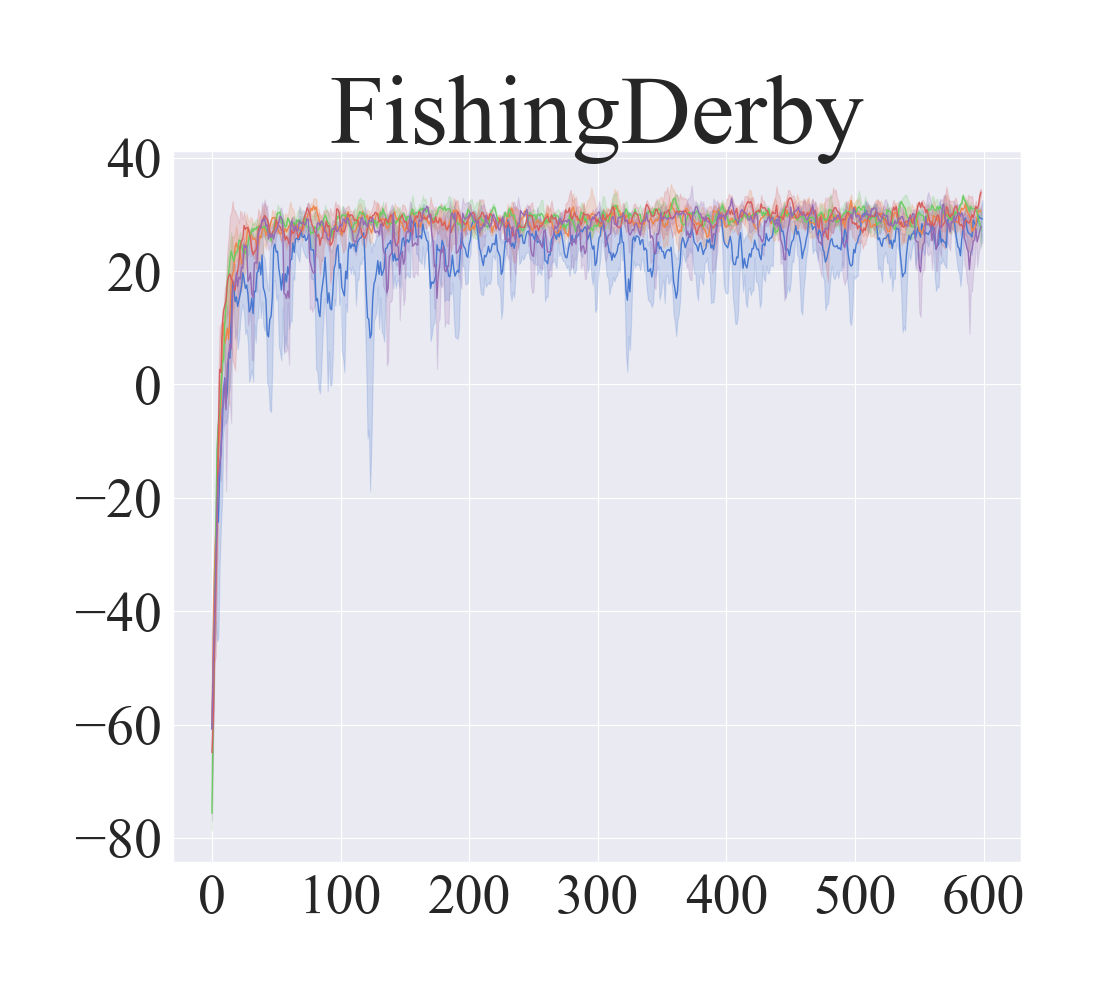}\\
			\end{minipage}%
		}%
		\subfigure{
			\begin{minipage}[t]{0.166\linewidth}
				\centering
				\includegraphics[width=1.05in]{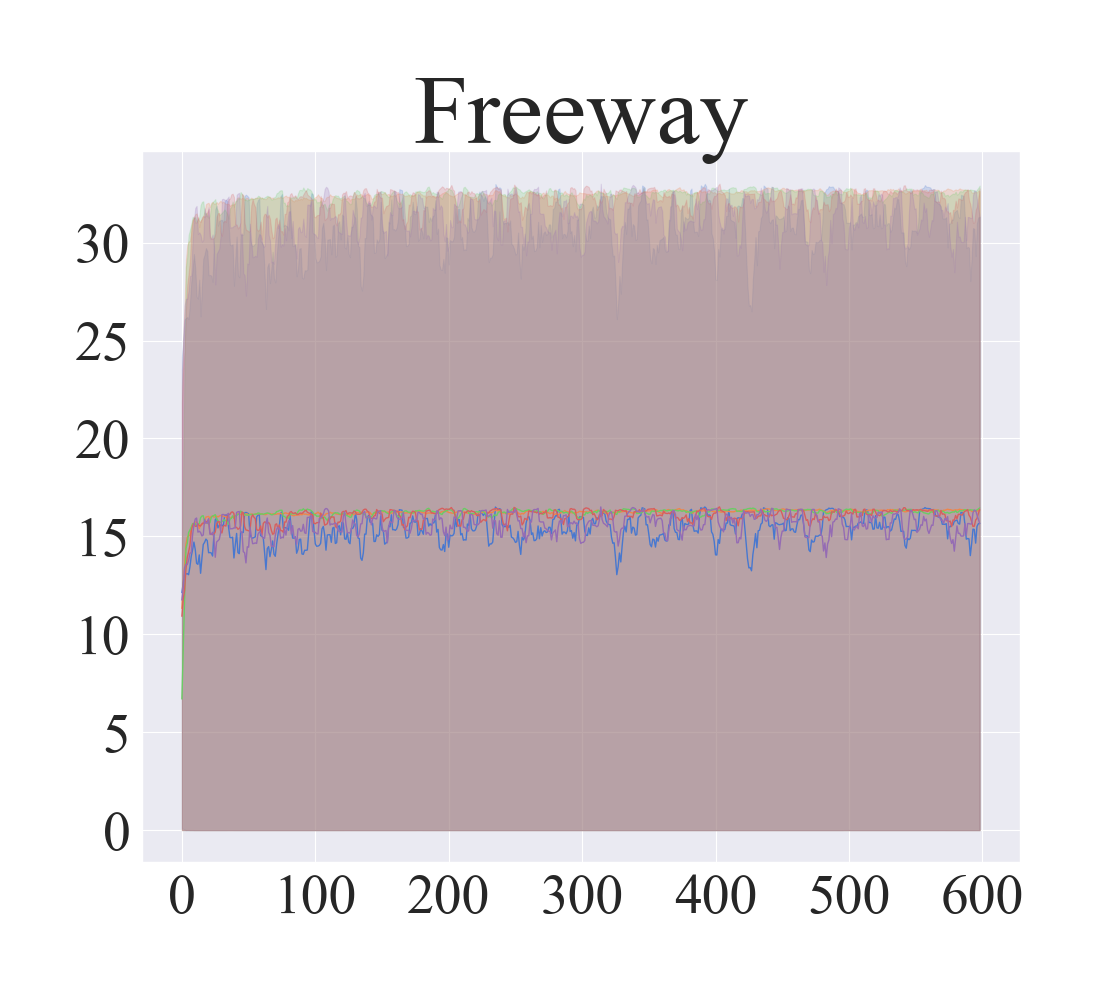}\\
			\end{minipage}%
		}%
		\vspace{-0.6cm}
		
		\subfigure{
			\begin{minipage}[t]{0.166\linewidth}
				\centering
				\includegraphics[width=1.05in]{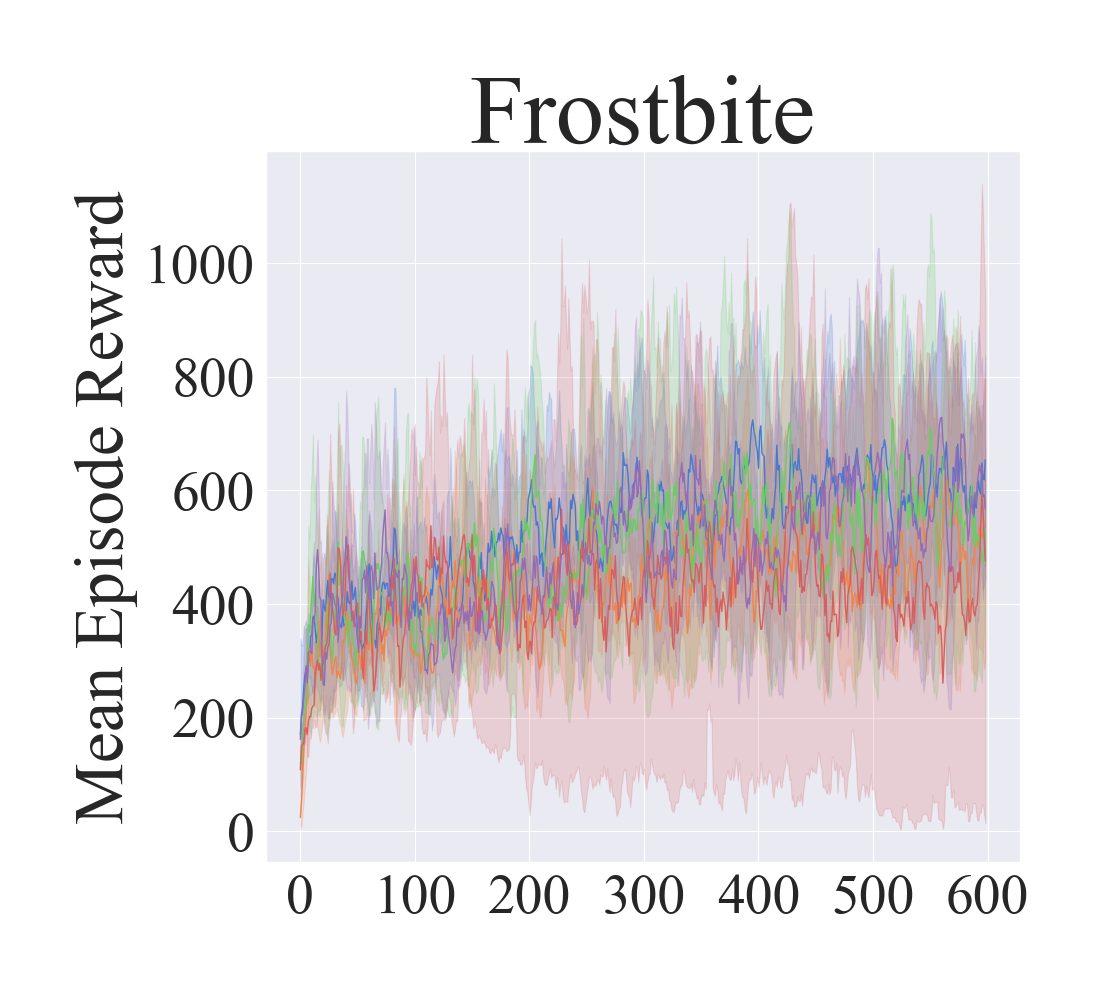}\\
			\end{minipage}%
		}%
		\subfigure{
			\begin{minipage}[t]{0.166\linewidth}
				\centering
				\includegraphics[width=1.05in]{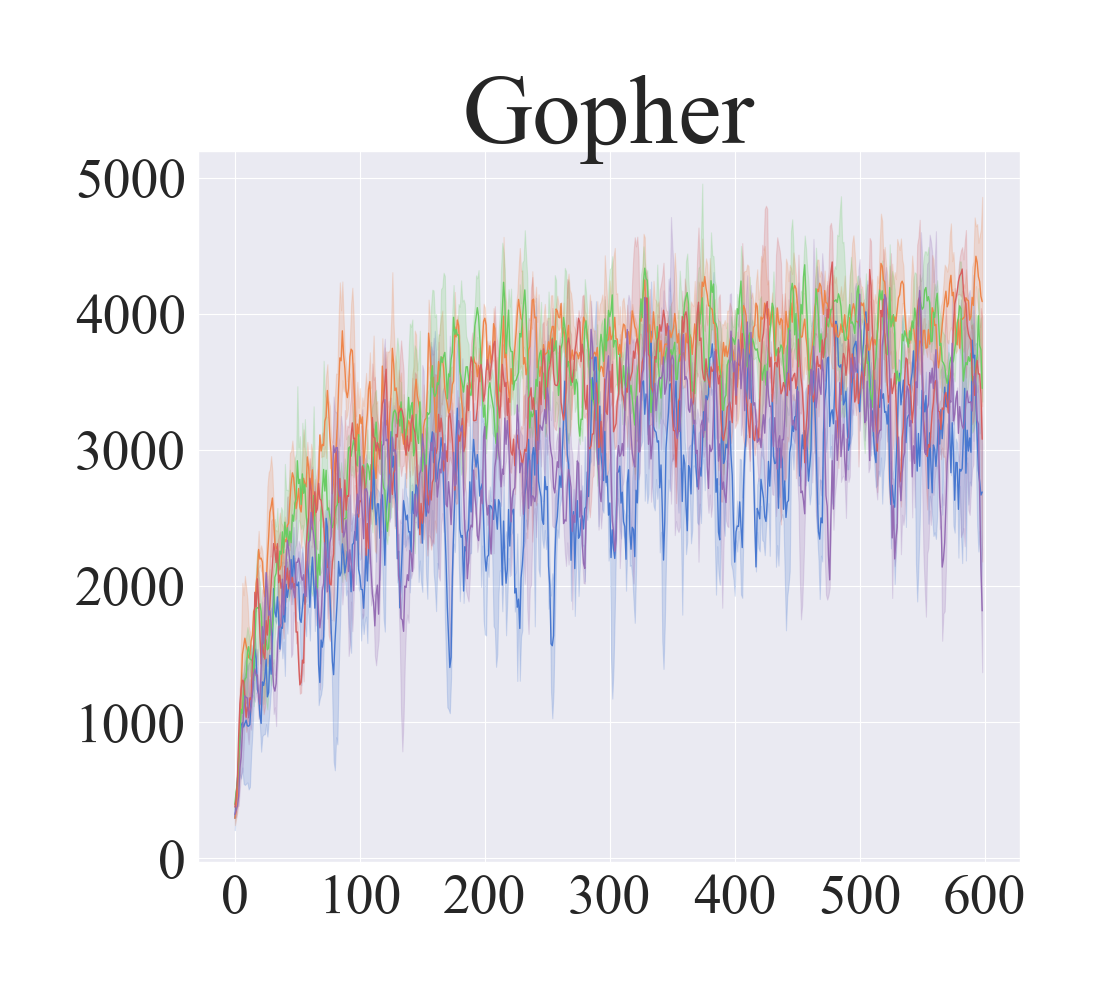}\\
			\end{minipage}%
		}%
		\subfigure{
			\begin{minipage}[t]{0.166\linewidth}
				\centering
				\includegraphics[width=1.05in]{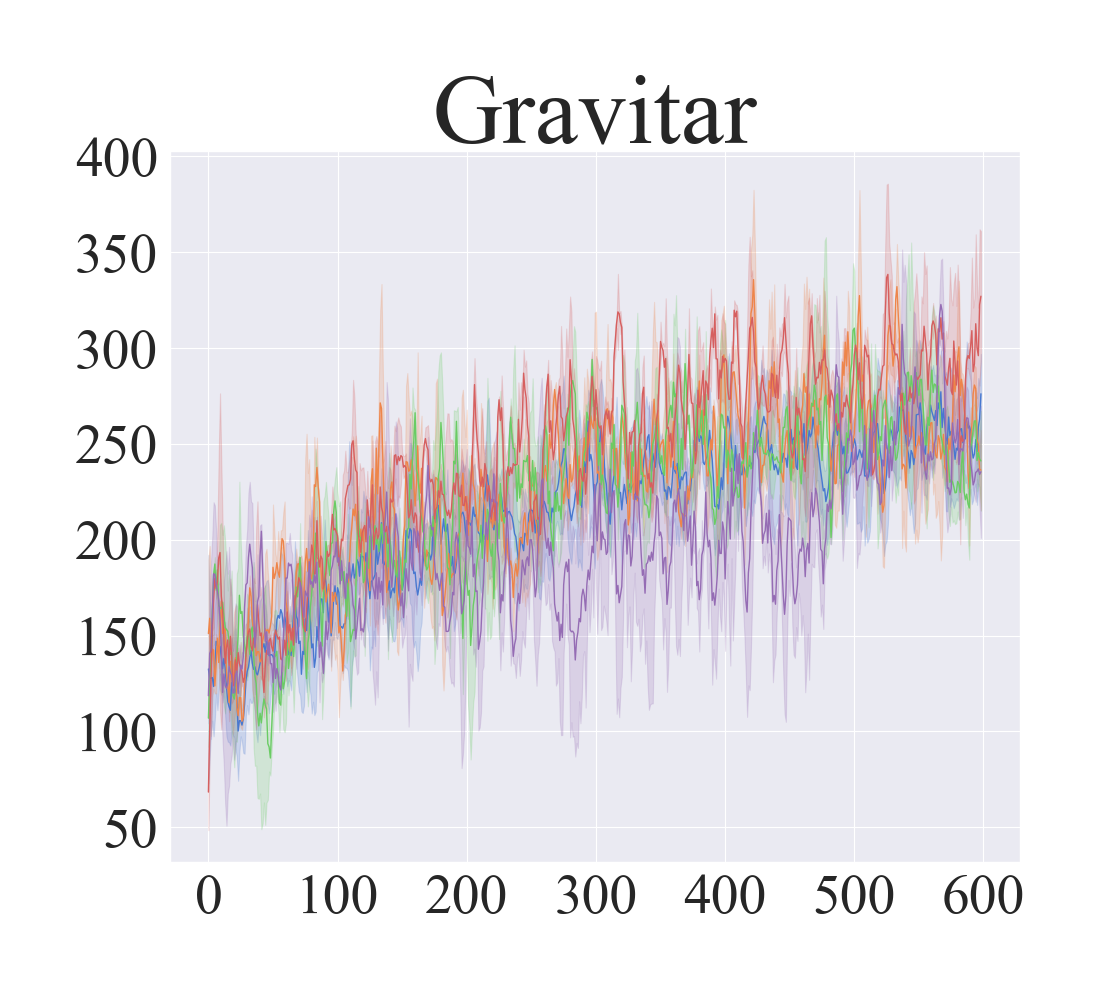}\\
			\end{minipage}%
		}%
		\subfigure{
			\begin{minipage}[t]{0.166\linewidth}
				\centering
				\includegraphics[width=1.05in]{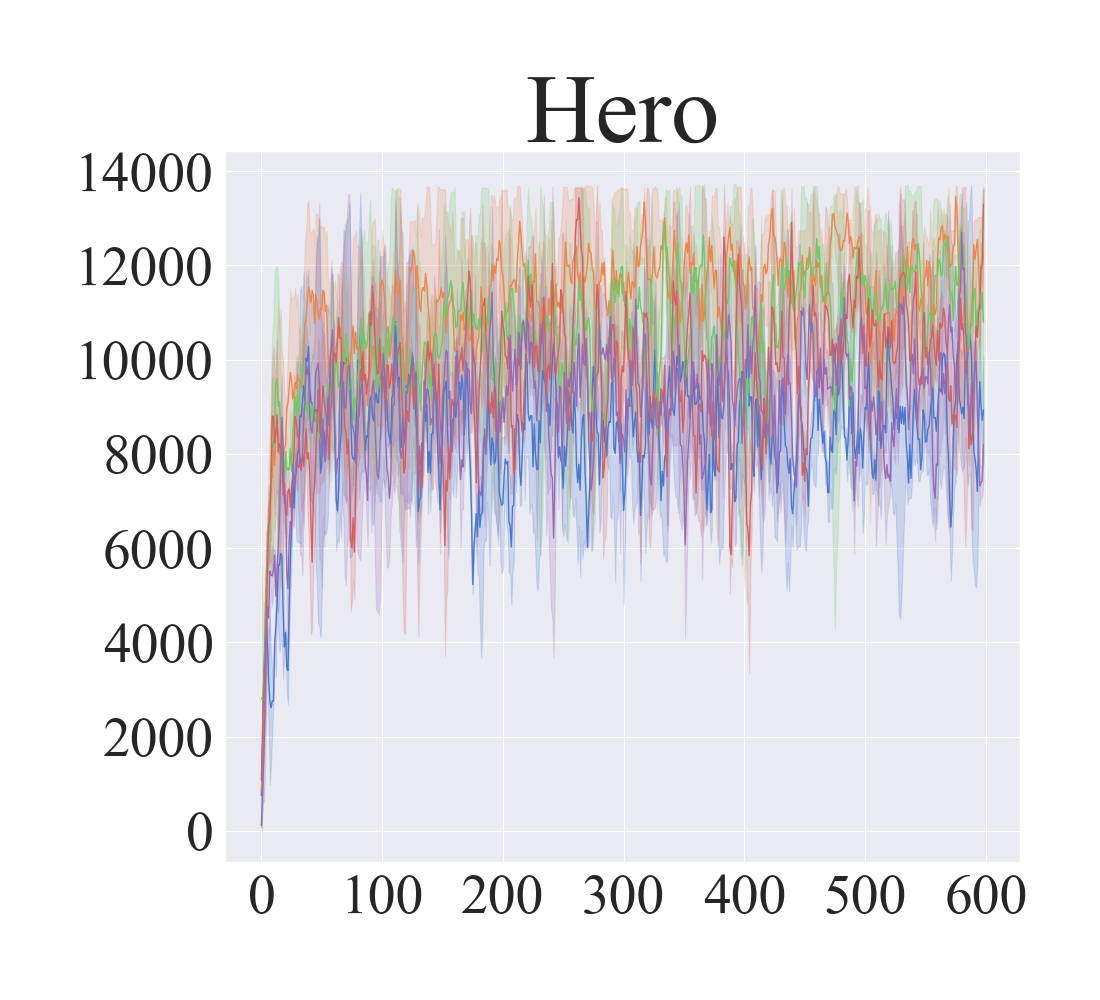}\\
			\end{minipage}%
		}%
		\subfigure{
			\begin{minipage}[t]{0.166\linewidth}
				\centering
				\includegraphics[width=1.05in]{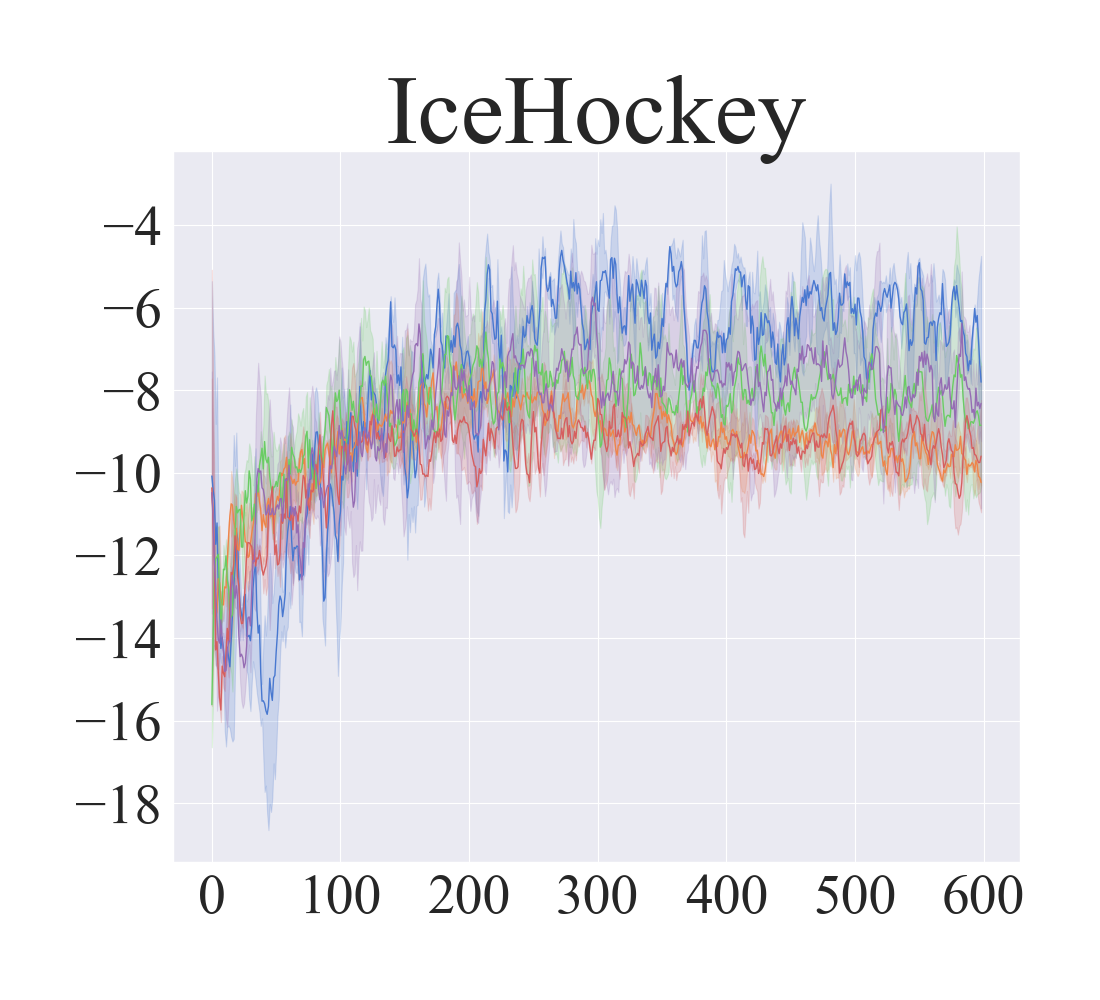}\\
			\end{minipage}%
		}%
		\subfigure{
			\begin{minipage}[t]{0.166\linewidth}
				\centering
				\includegraphics[width=1.05in]{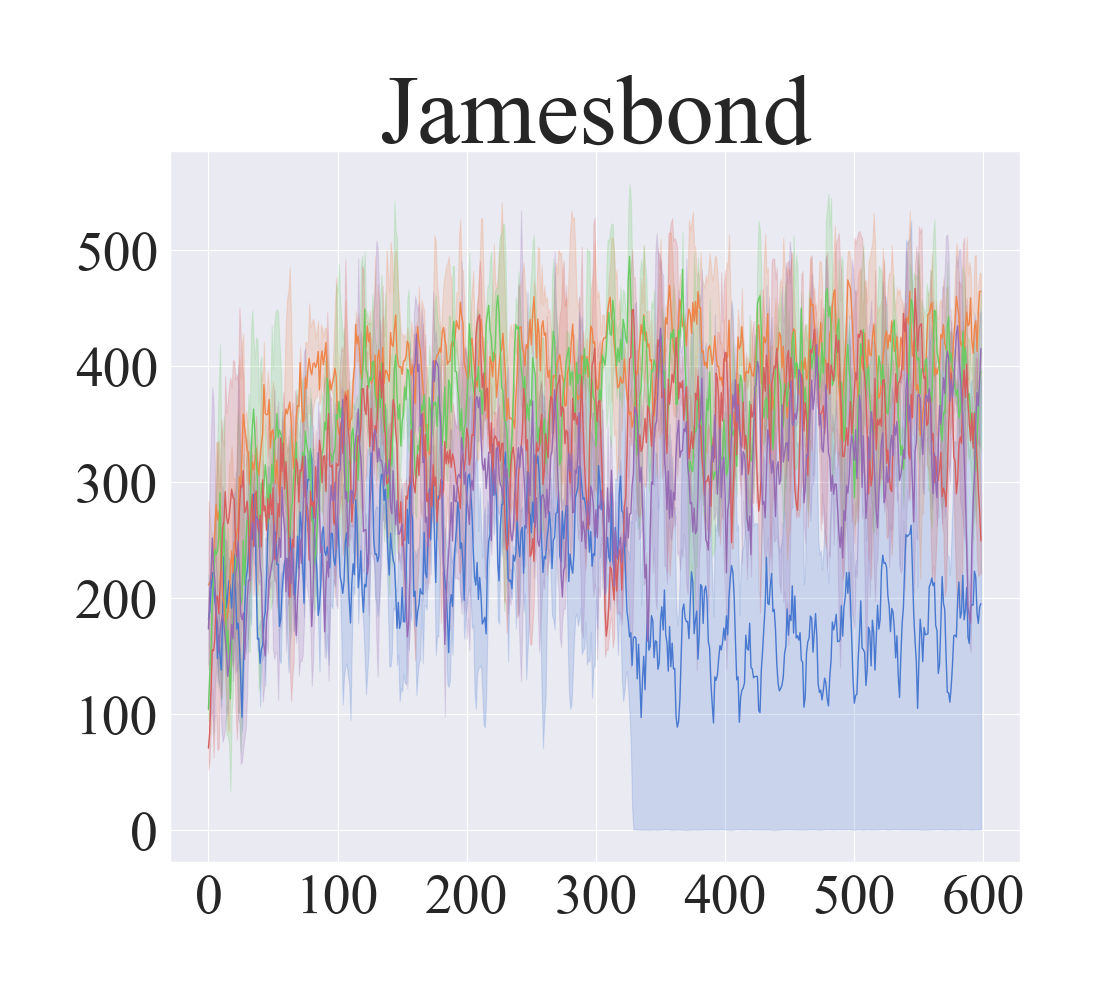}\\
			\end{minipage}%
		}%
		\vspace{-0.6cm}
		
		\subfigure{
			\begin{minipage}[t]{0.166\linewidth}
				\centering
				\includegraphics[width=1.05in]{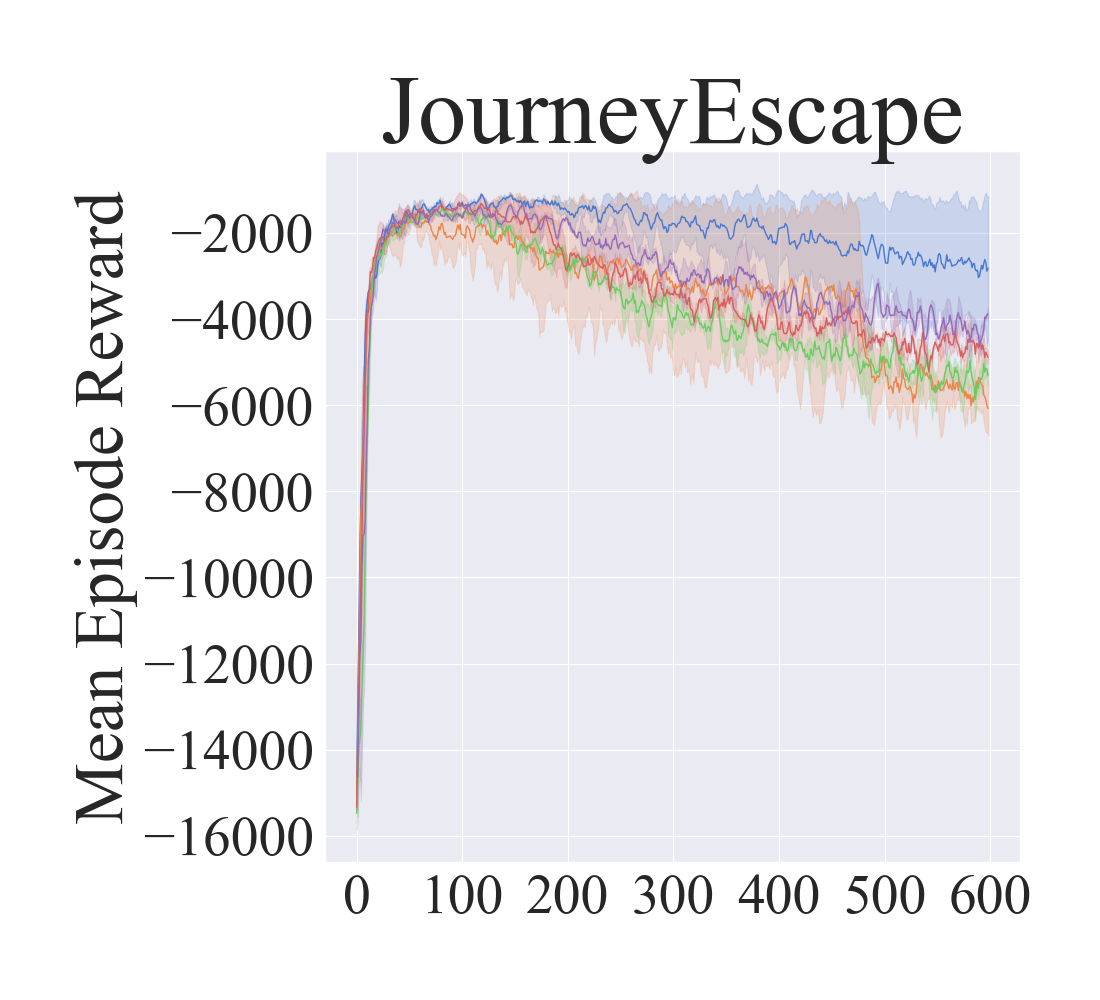}\\
			\end{minipage}%
		}%
		\subfigure{
			\begin{minipage}[t]{0.166\linewidth}
				\centering
				\includegraphics[width=1.05in]{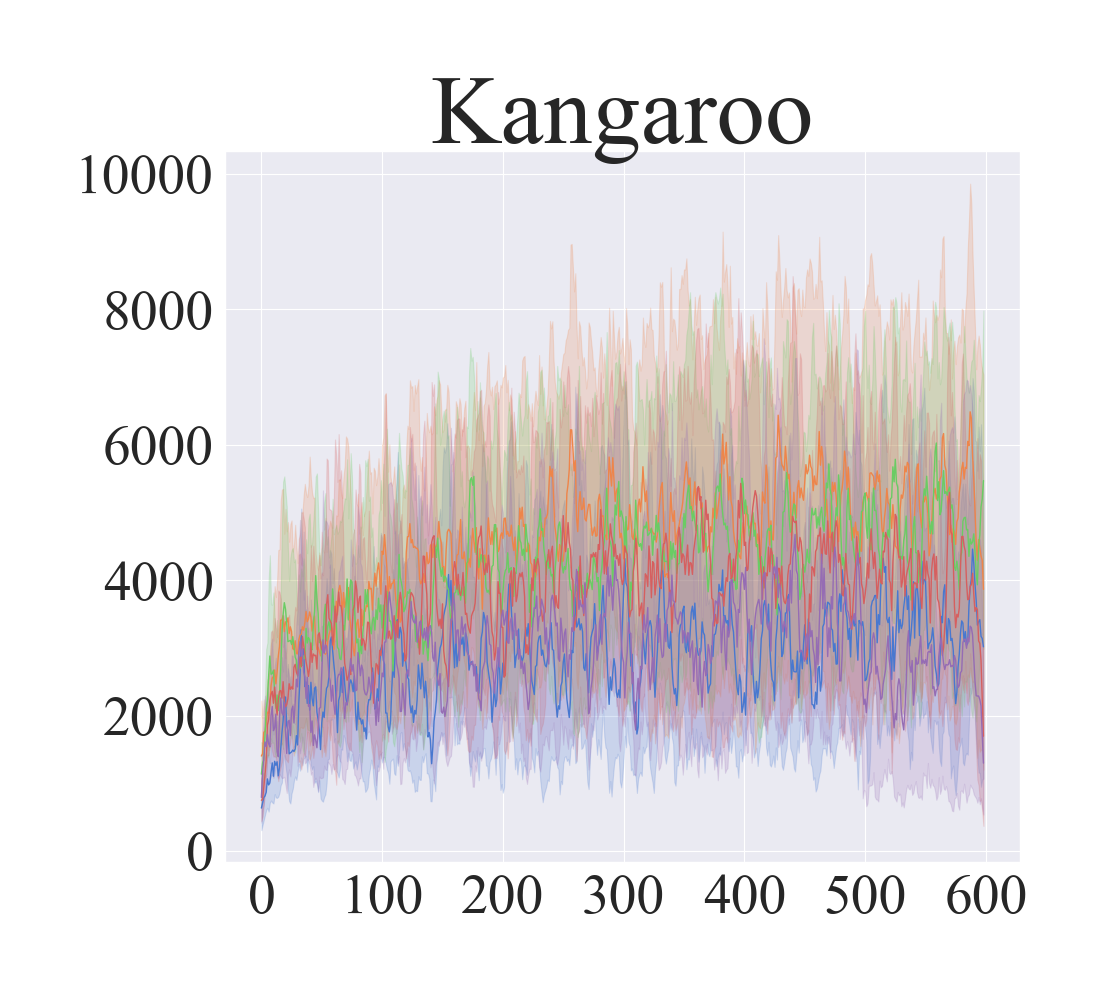}\\
			\end{minipage}%
		}%
		\subfigure{
			\begin{minipage}[t]{0.166\linewidth}
				\centering
				\includegraphics[width=1.05in]{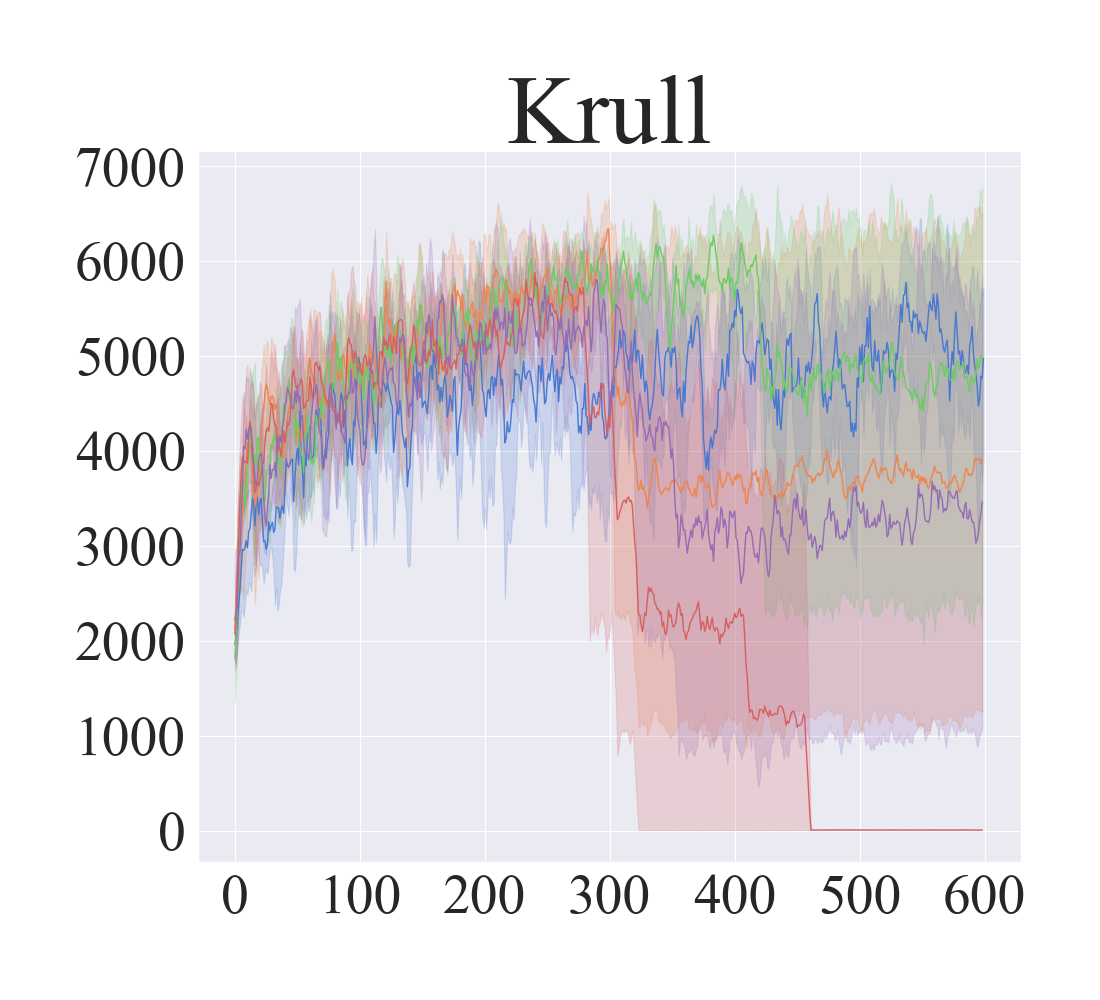}\\
			\end{minipage}%
		}%
		\subfigure{
			\begin{minipage}[t]{0.166\linewidth}
				\centering
				\includegraphics[width=1.05in]{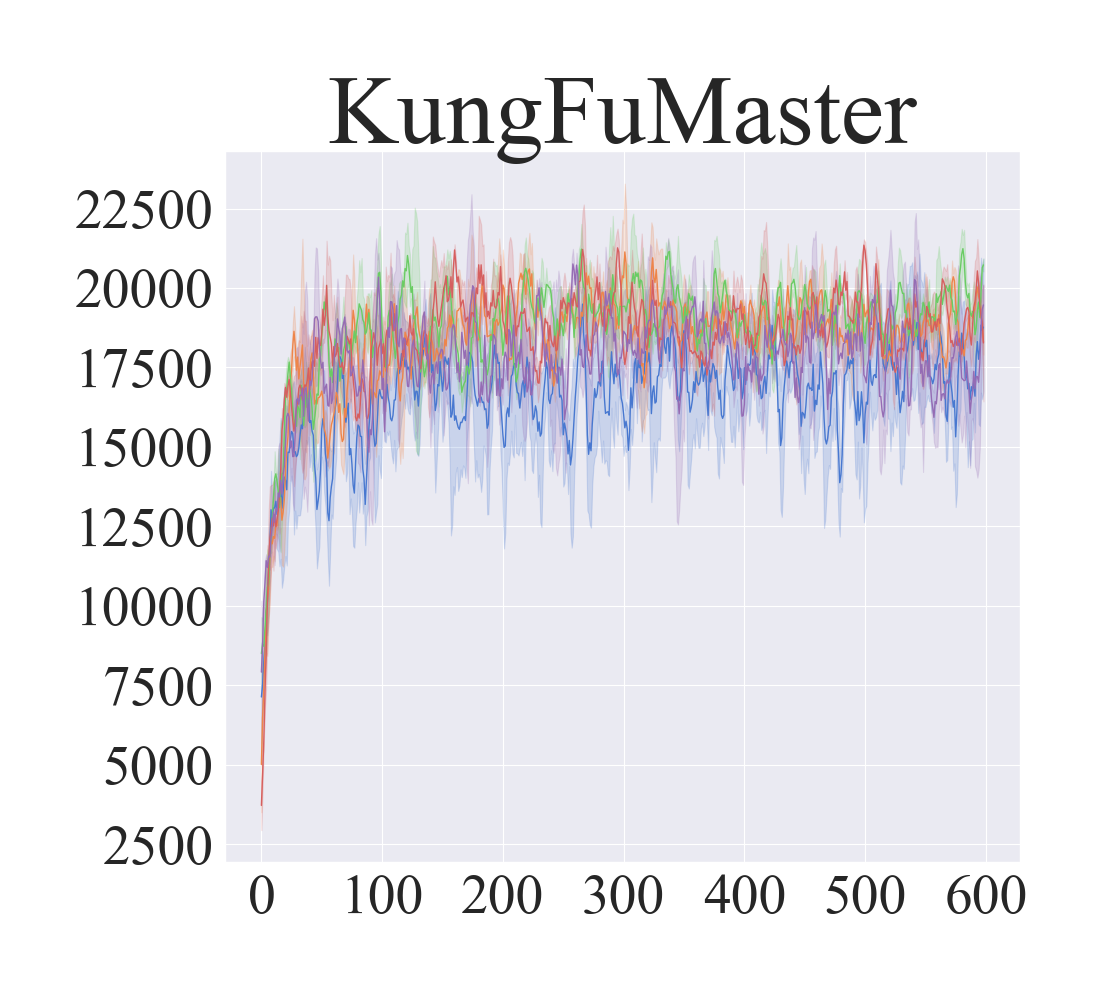}\\
			\end{minipage}%
		}%
		\subfigure{
			\begin{minipage}[t]{0.166\linewidth}
				\centering
				\includegraphics[width=1.05in]{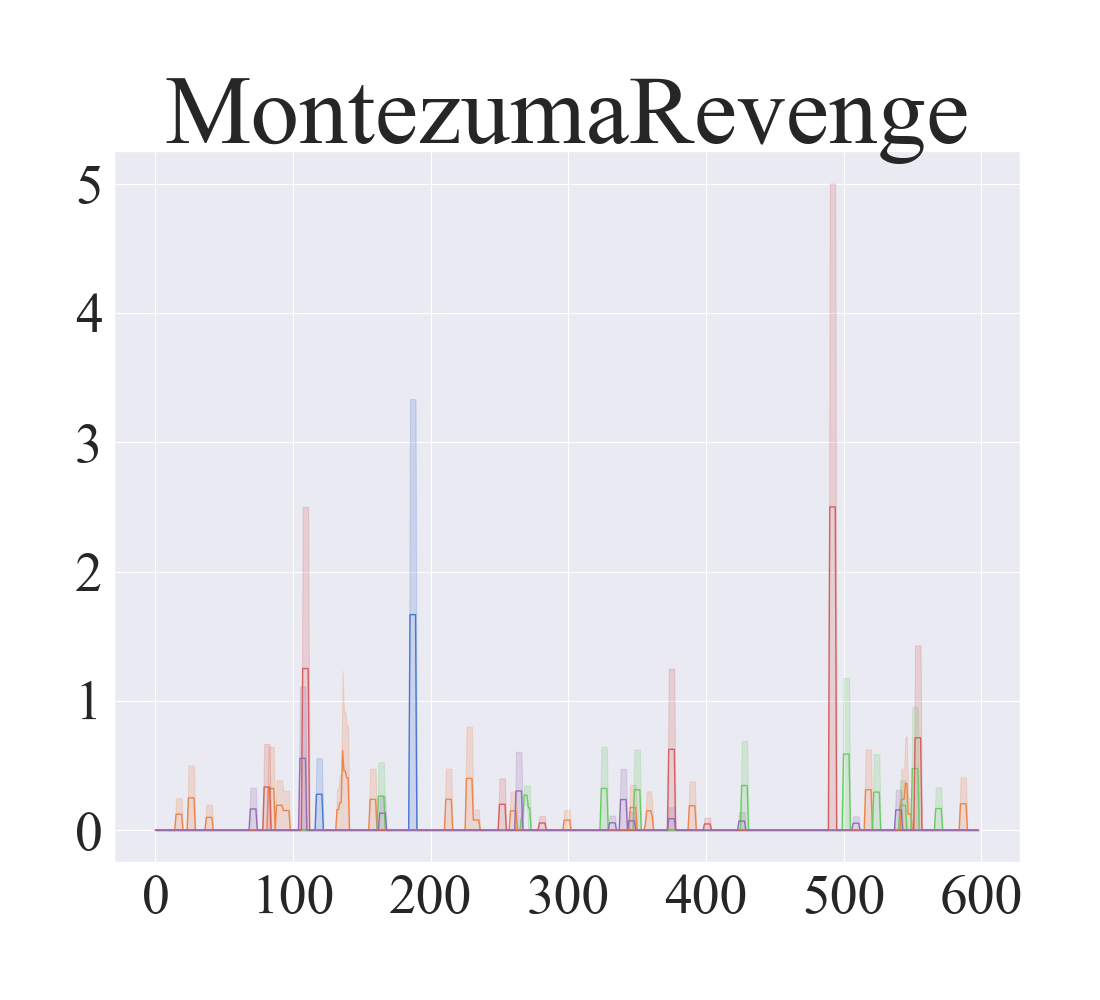}\\
			\end{minipage}%
		}%
		\subfigure{
			\begin{minipage}[t]{0.166\linewidth}
				\centering
				\includegraphics[width=1.05in]{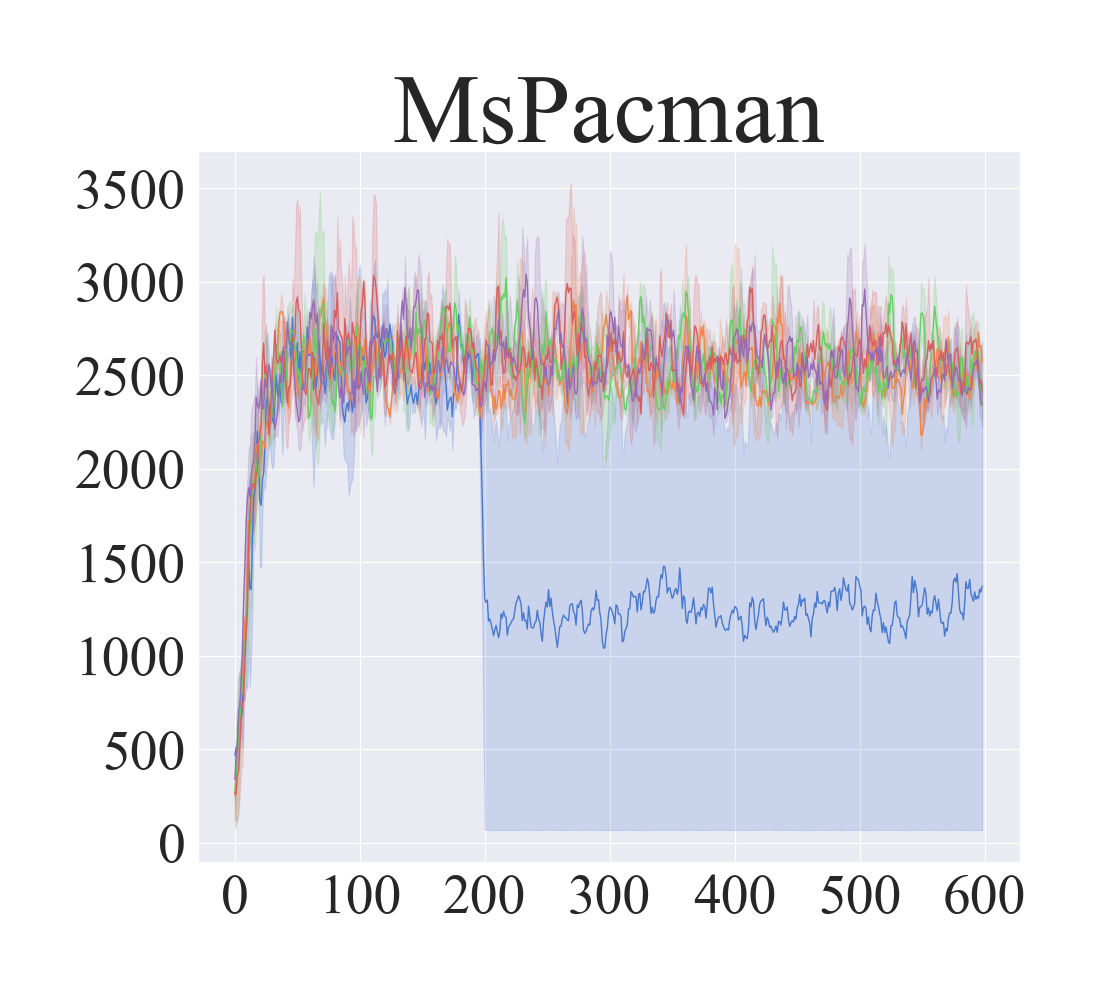}\\
			\end{minipage}%
		}%
		\vspace{-0.6cm}
		
		\subfigure{
			\begin{minipage}[t]{0.166\linewidth}
				\centering
				\includegraphics[width=1.05in]{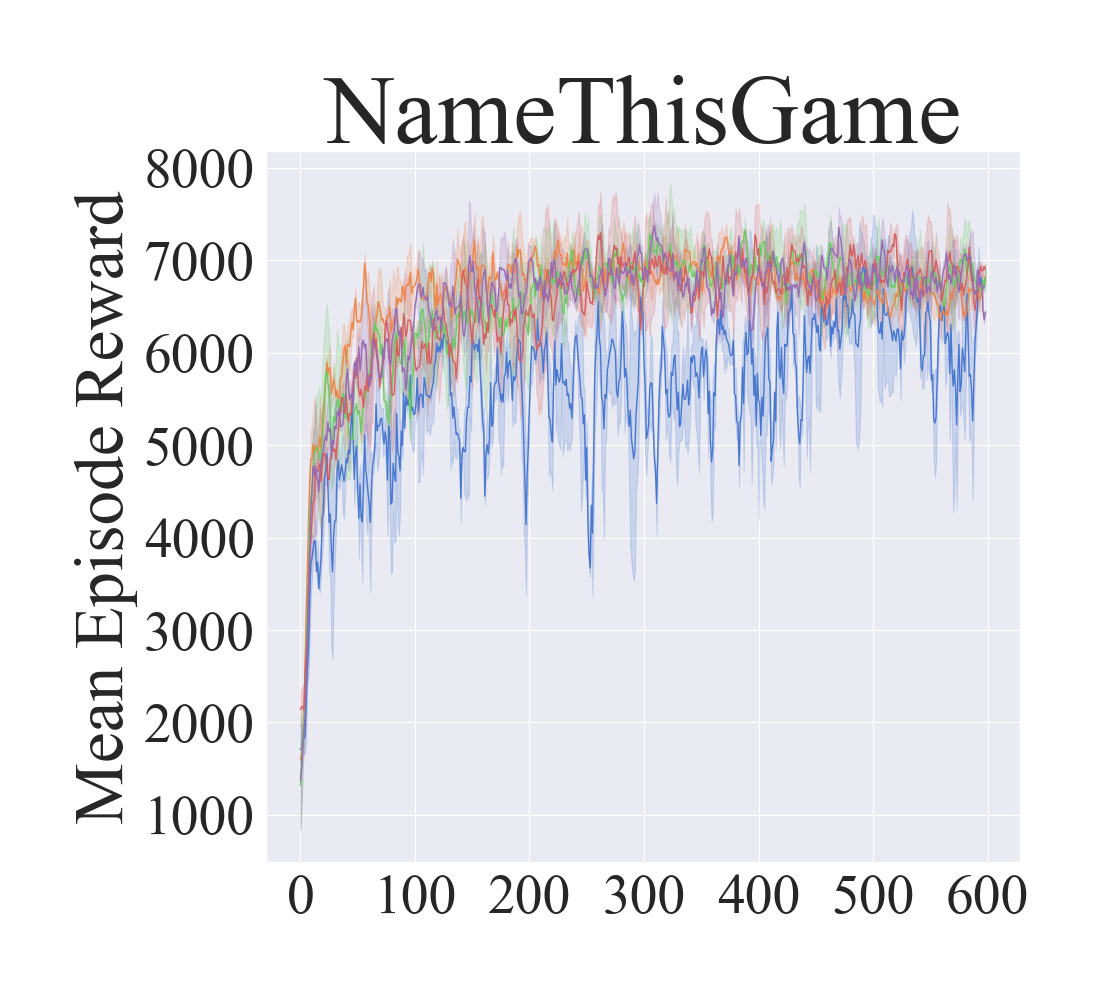}\\
			\end{minipage}%
		}%
		\subfigure{
			\begin{minipage}[t]{0.166\linewidth}
				\centering
				\includegraphics[width=1.05in]{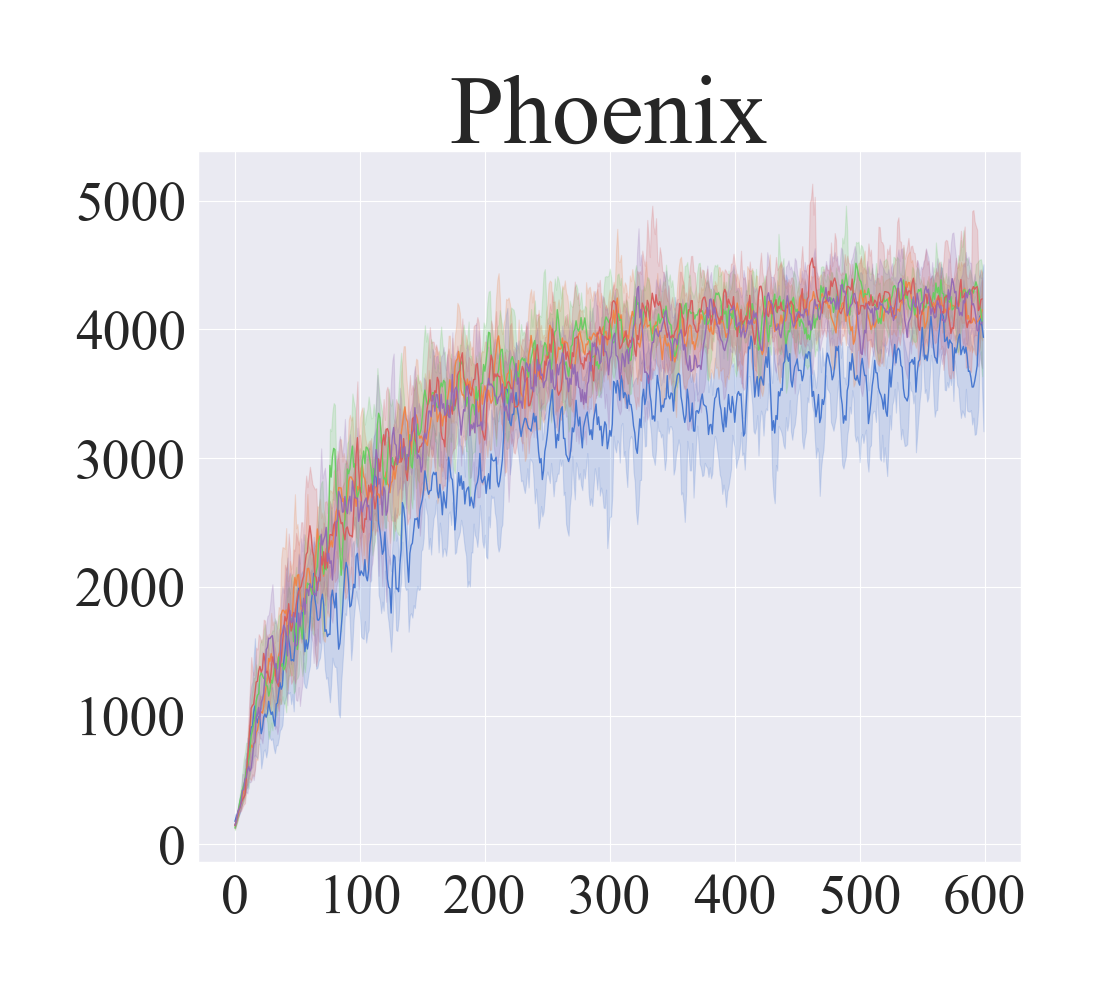}\\
			\end{minipage}%
		}%
		\subfigure{
			\begin{minipage}[t]{0.166\linewidth}
				\centering
				\includegraphics[width=1.05in]{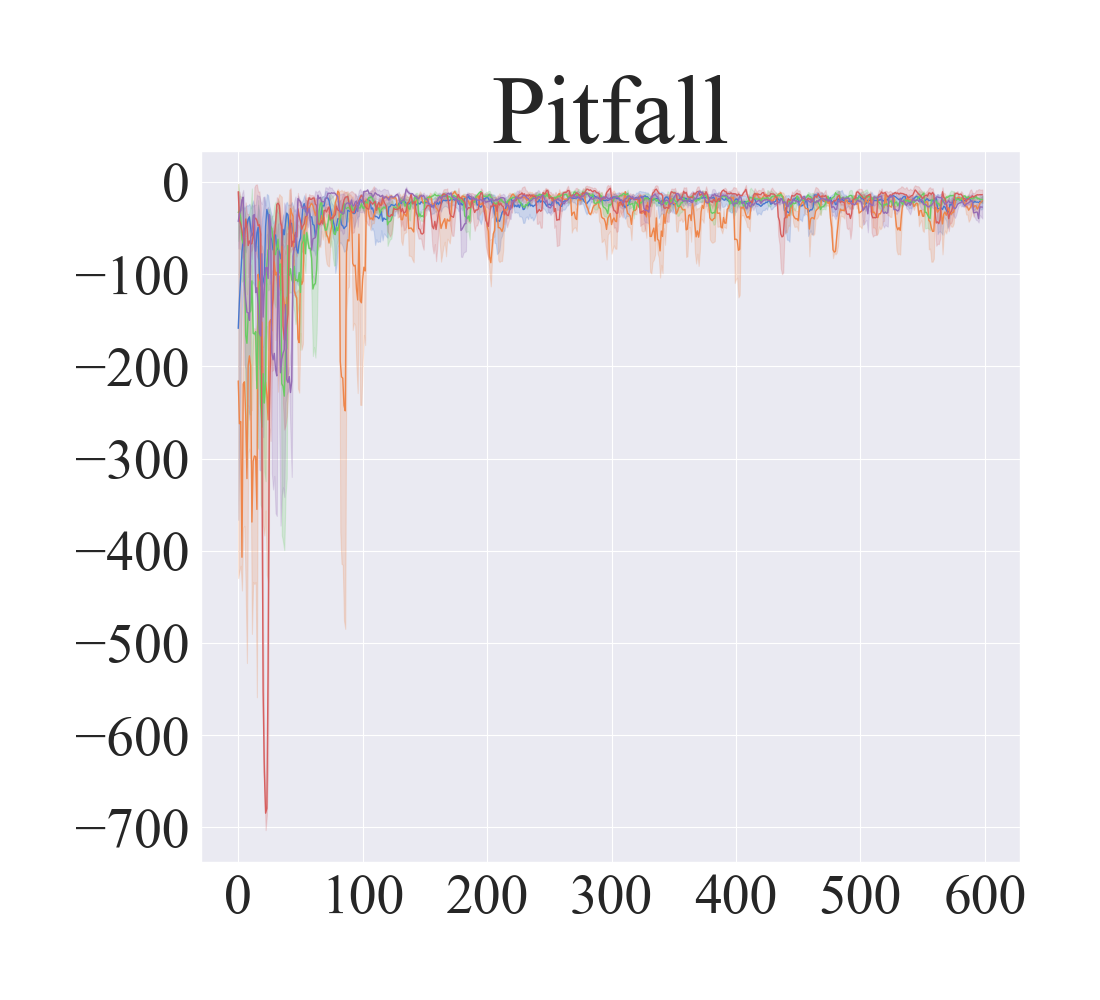}\\
			\end{minipage}%
		}%
		\subfigure{
			\begin{minipage}[t]{0.166\linewidth}
				\centering
				\includegraphics[width=1.05in]{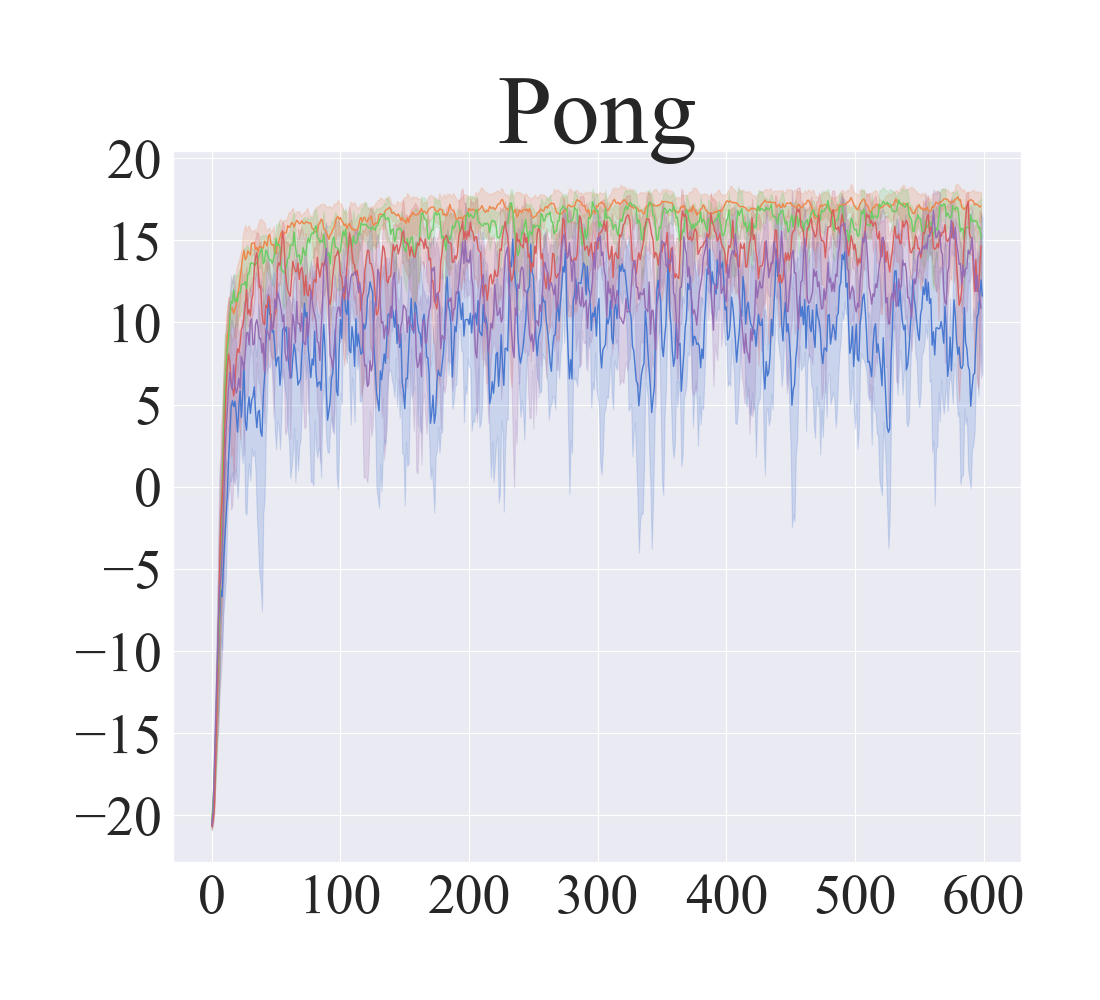}\\
			\end{minipage}%
		}%
		\subfigure{
			\begin{minipage}[t]{0.166\linewidth}
				\centering
				\includegraphics[width=1.05in]{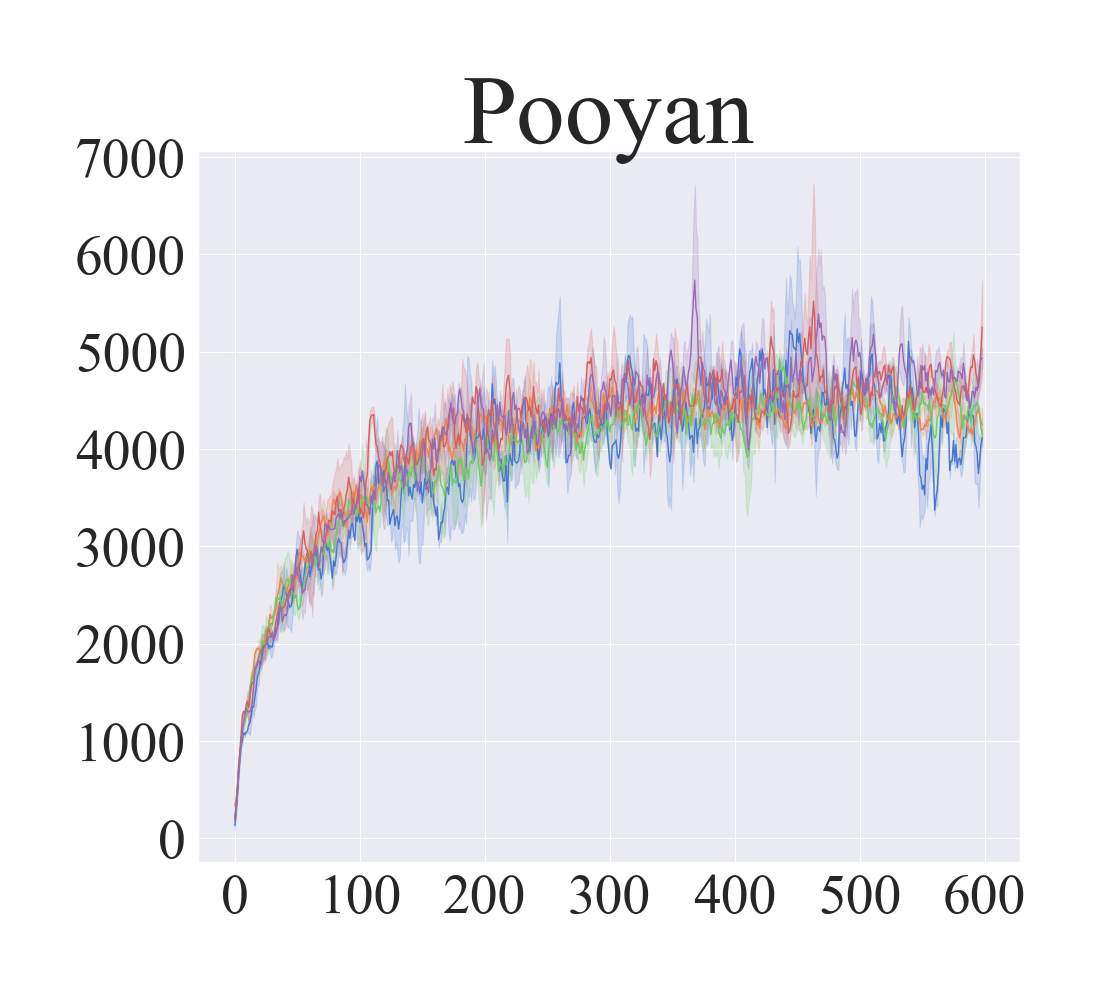}\\
			\end{minipage}%
		}%
		\subfigure{
			\begin{minipage}[t]{0.166\linewidth}
				\centering
				\includegraphics[width=1.05in]{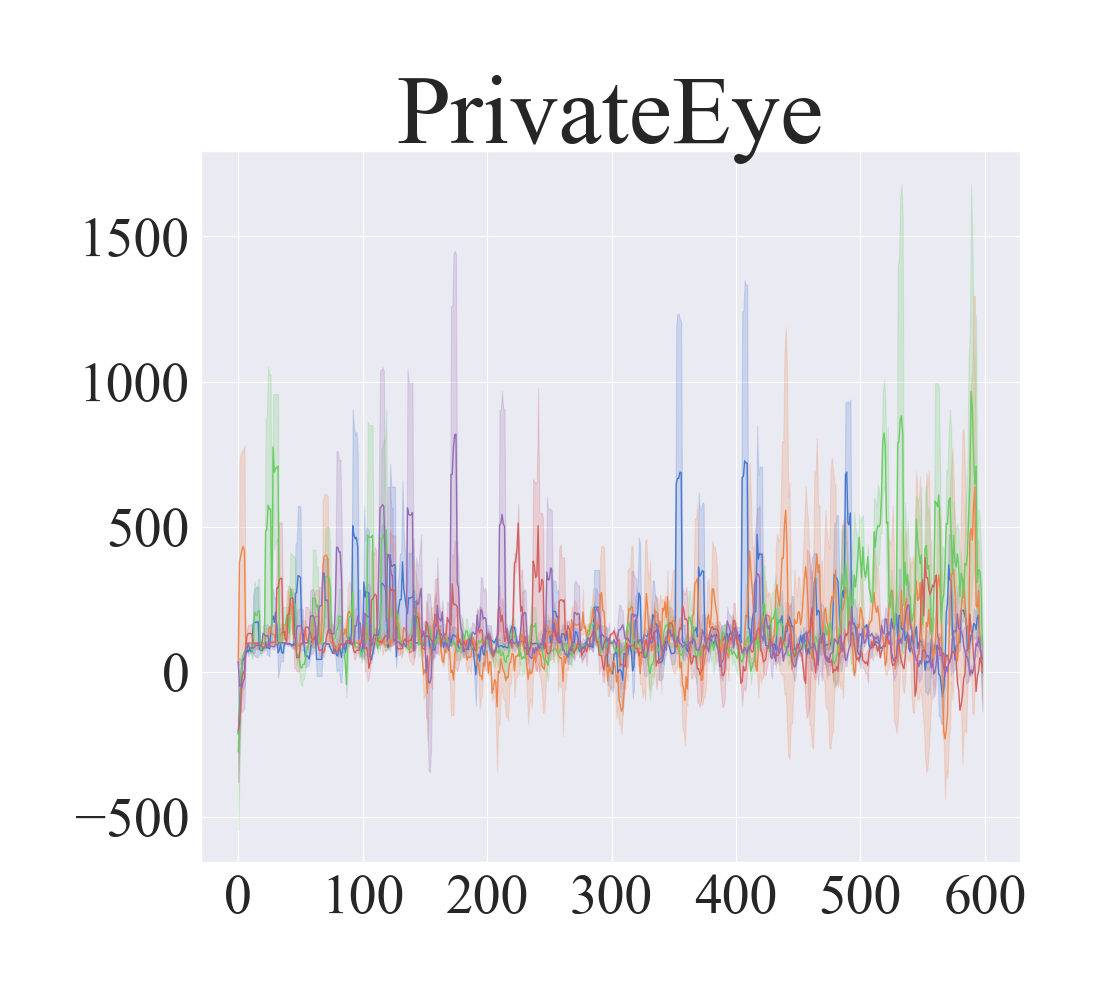}\\
			\end{minipage}%
		}%
		\vspace{-0.6cm}
		
		\subfigure{
			\begin{minipage}[t]{0.166\linewidth}
				\centering
				\includegraphics[width=1.05in]{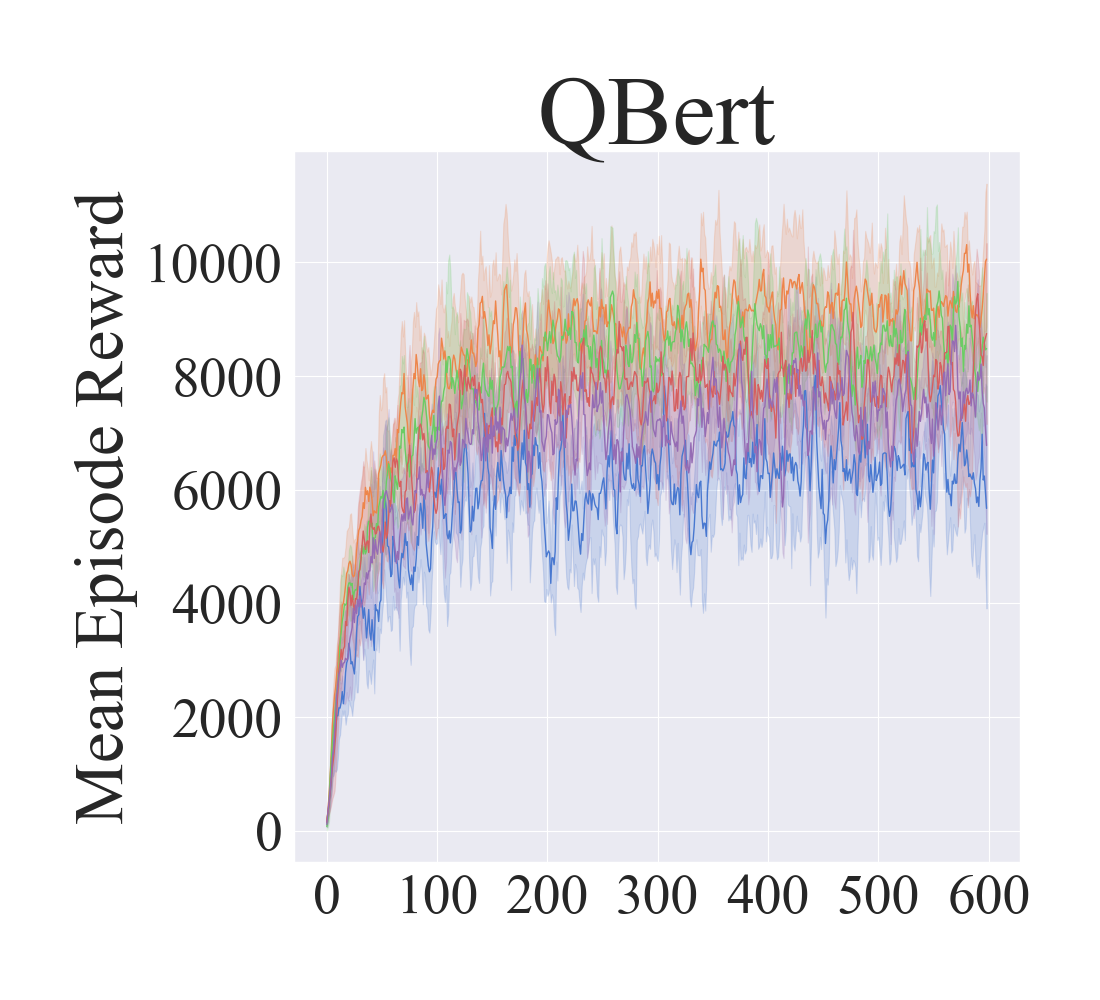}\\
			\end{minipage}%
		}%
		\subfigure{
			\begin{minipage}[t]{0.166\linewidth}
				\centering
				\includegraphics[width=1.05in]{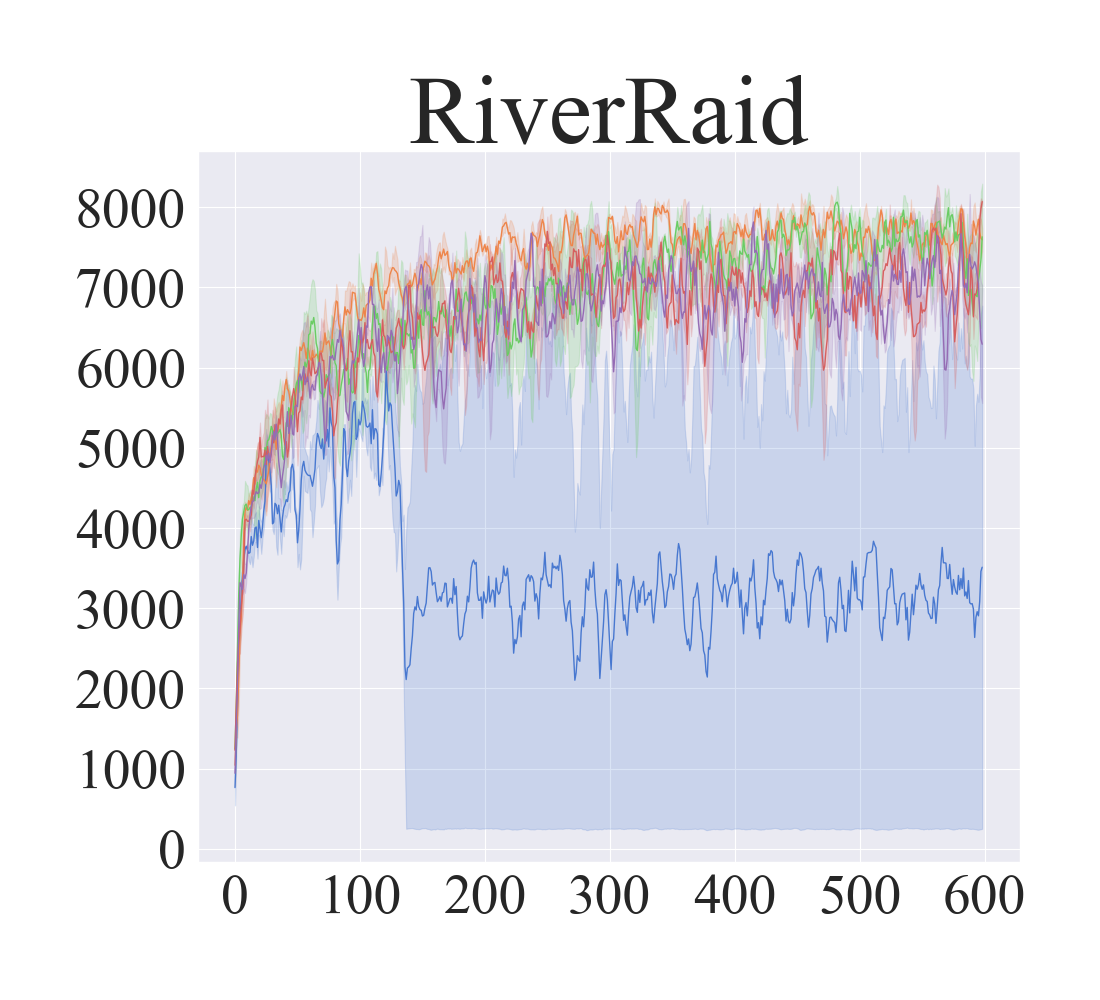}\\
			\end{minipage}%
		}%
		\subfigure{
			\begin{minipage}[t]{0.166\linewidth}
				\centering
				\includegraphics[width=1.05in]{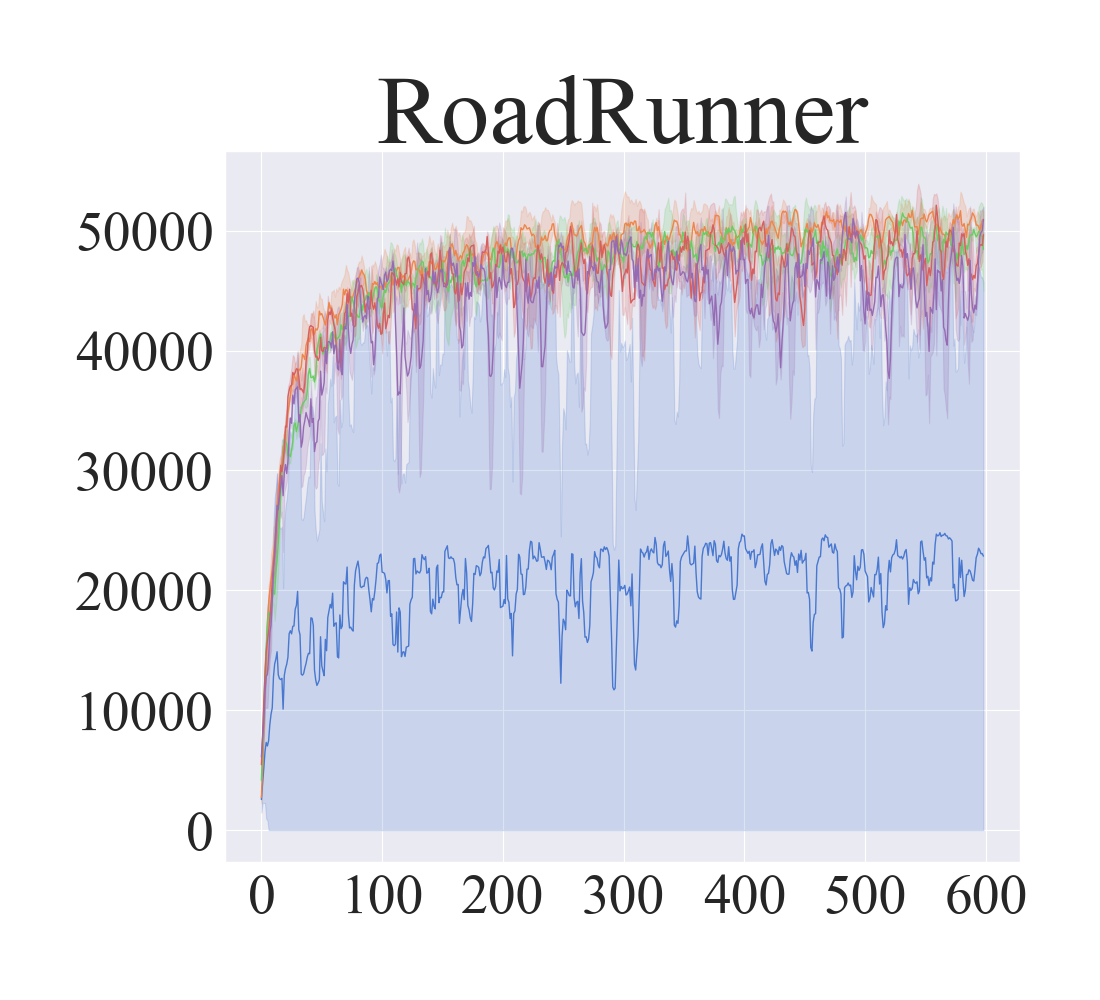}\\
			\end{minipage}%
		}%
		\subfigure{
			\begin{minipage}[t]{0.166\linewidth}
				\centering
				\includegraphics[width=1.05in]{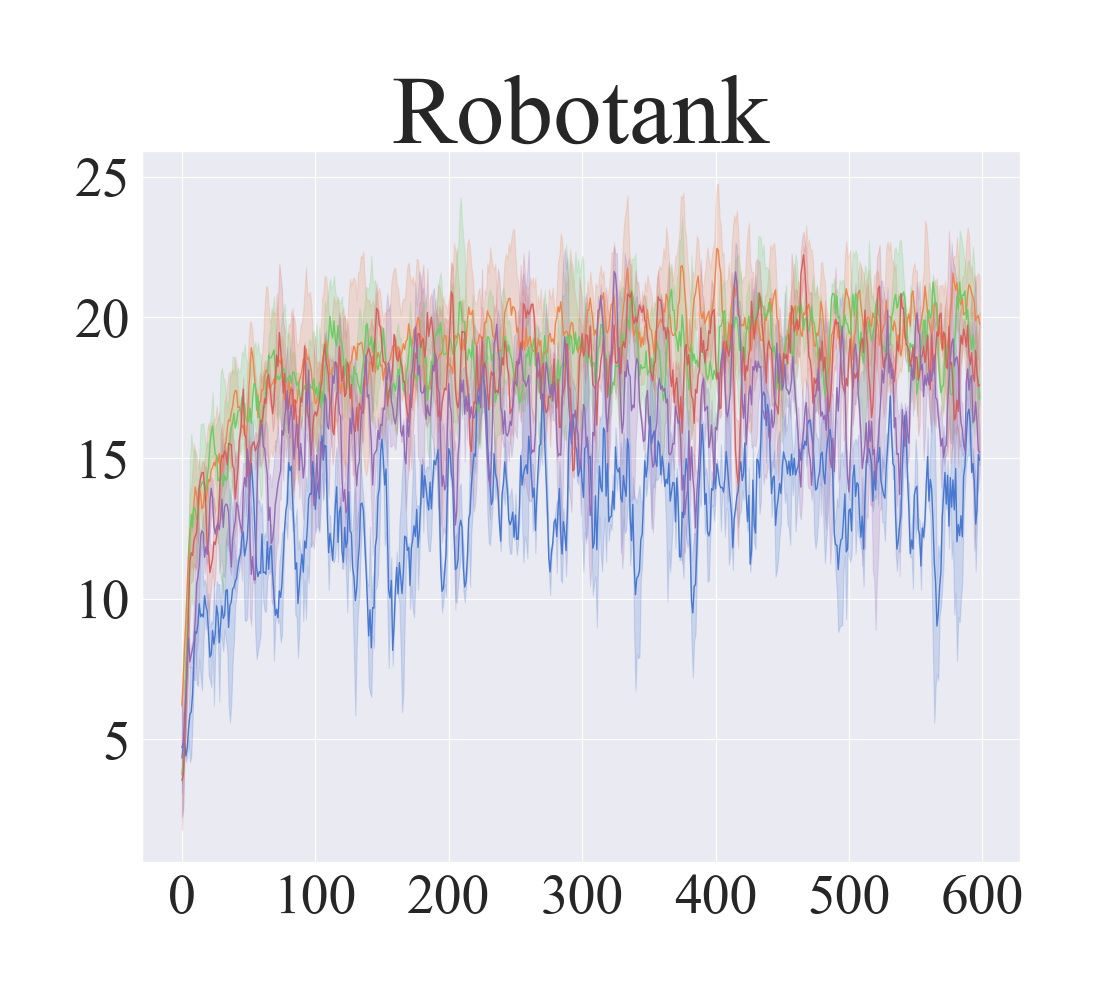}\\
			\end{minipage}%
		}%
		\subfigure{
			\begin{minipage}[t]{0.166\linewidth}
				\centering
				\includegraphics[width=1.05in]{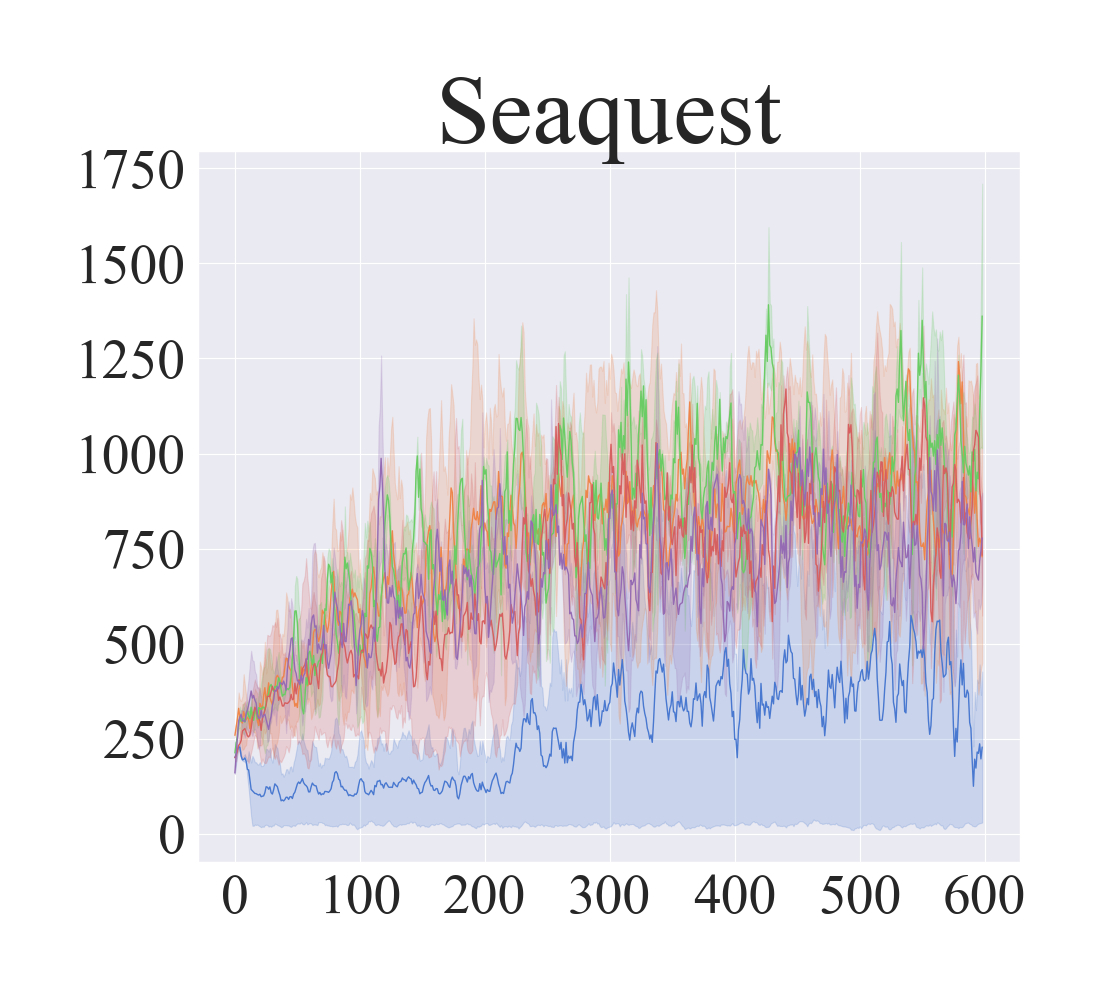}\\
			\end{minipage}%
		}%
		\subfigure{
			\begin{minipage}[t]{0.166\linewidth}
				\centering
				\includegraphics[width=1.05in]{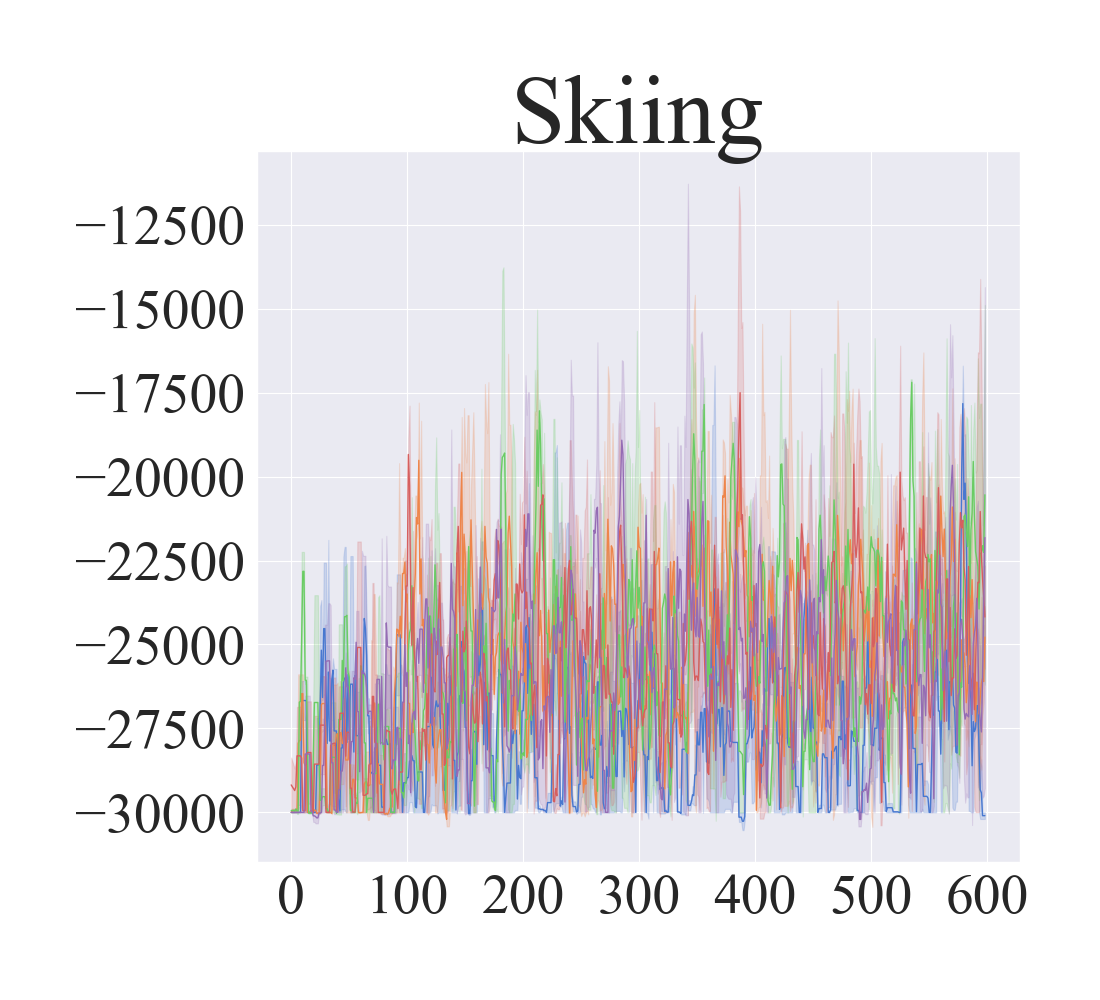}\\
			\end{minipage}%
		}%
		\vspace{-0.6cm}
		
		\subfigure{
			\begin{minipage}[t]{0.166\linewidth}
				\centering
				\includegraphics[width=1.05in]{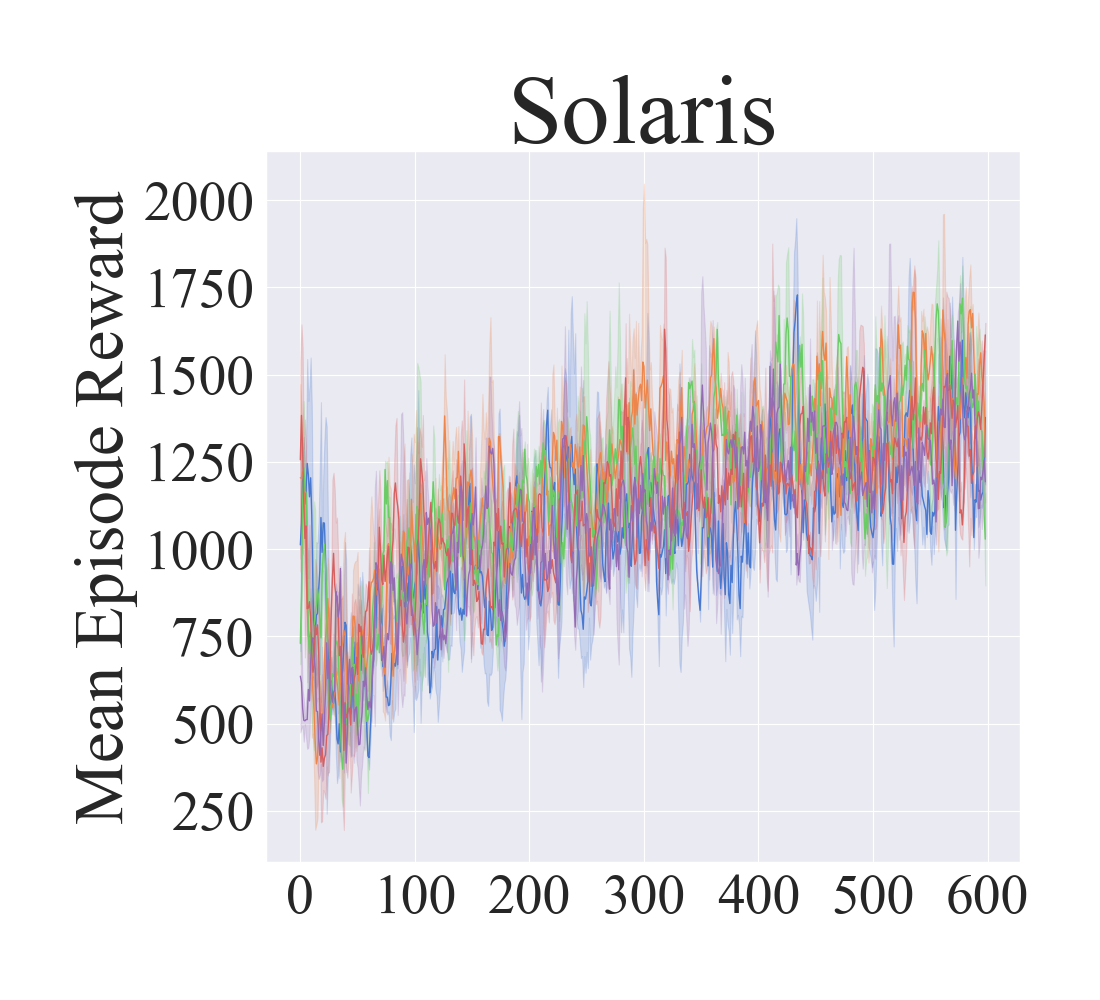}\\
			\end{minipage}%
		}%
		\subfigure{
			\begin{minipage}[t]{0.166\linewidth}
				\centering
				\includegraphics[width=1.05in]{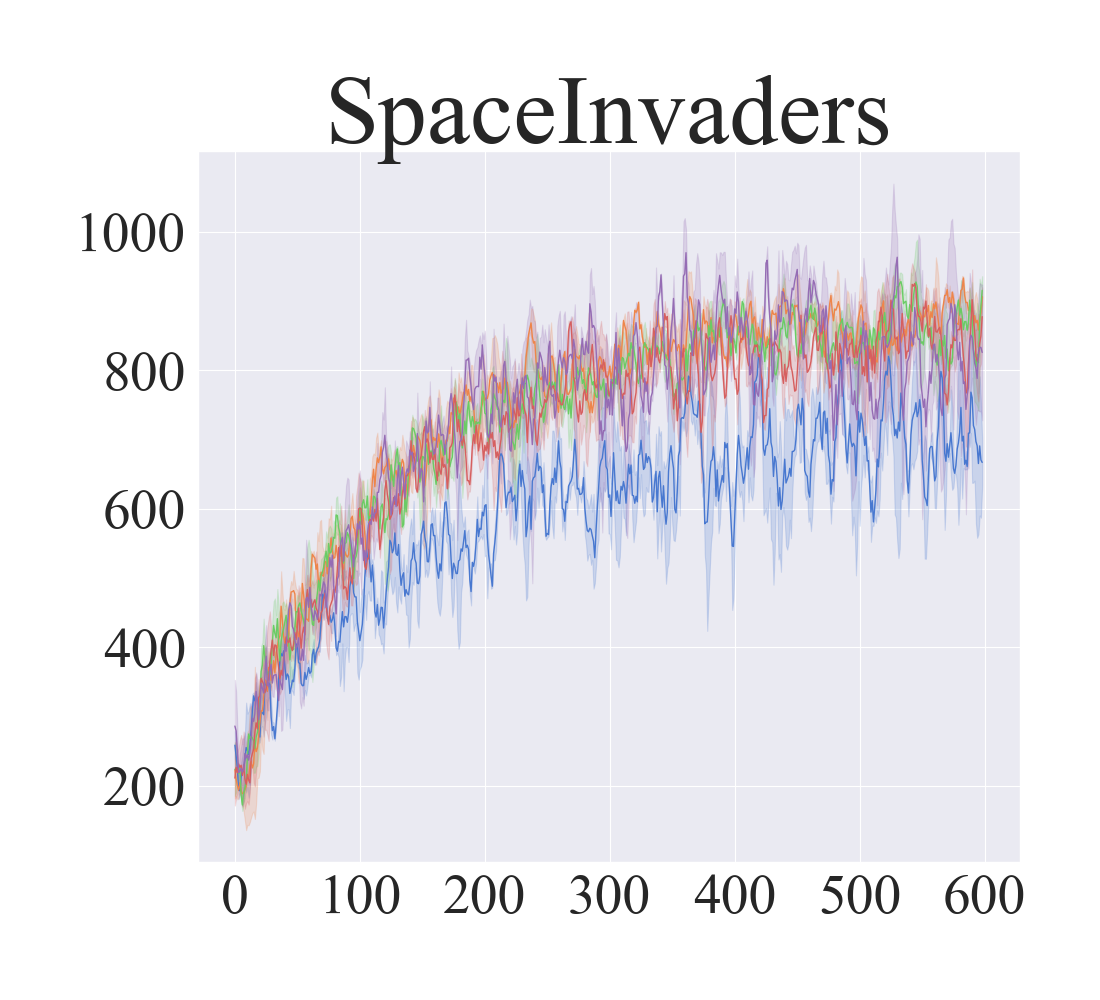}\\
			\end{minipage}%
		}%
		\subfigure{
			\begin{minipage}[t]{0.166\linewidth}
				\centering
				\includegraphics[width=1.05in]{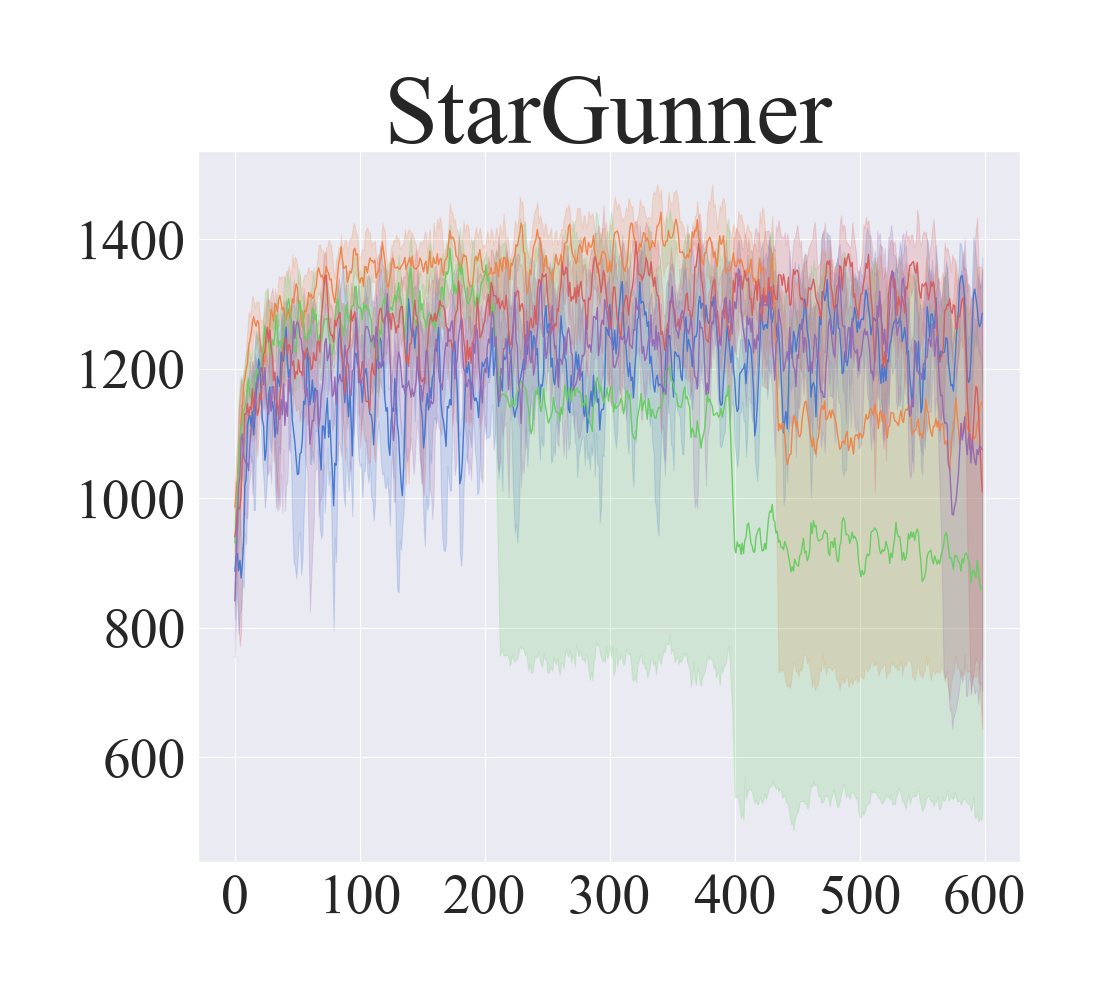}\\
			\end{minipage}%
		}%
		\subfigure{
			\begin{minipage}[t]{0.166\linewidth}
				\centering
				\includegraphics[width=1.05in]{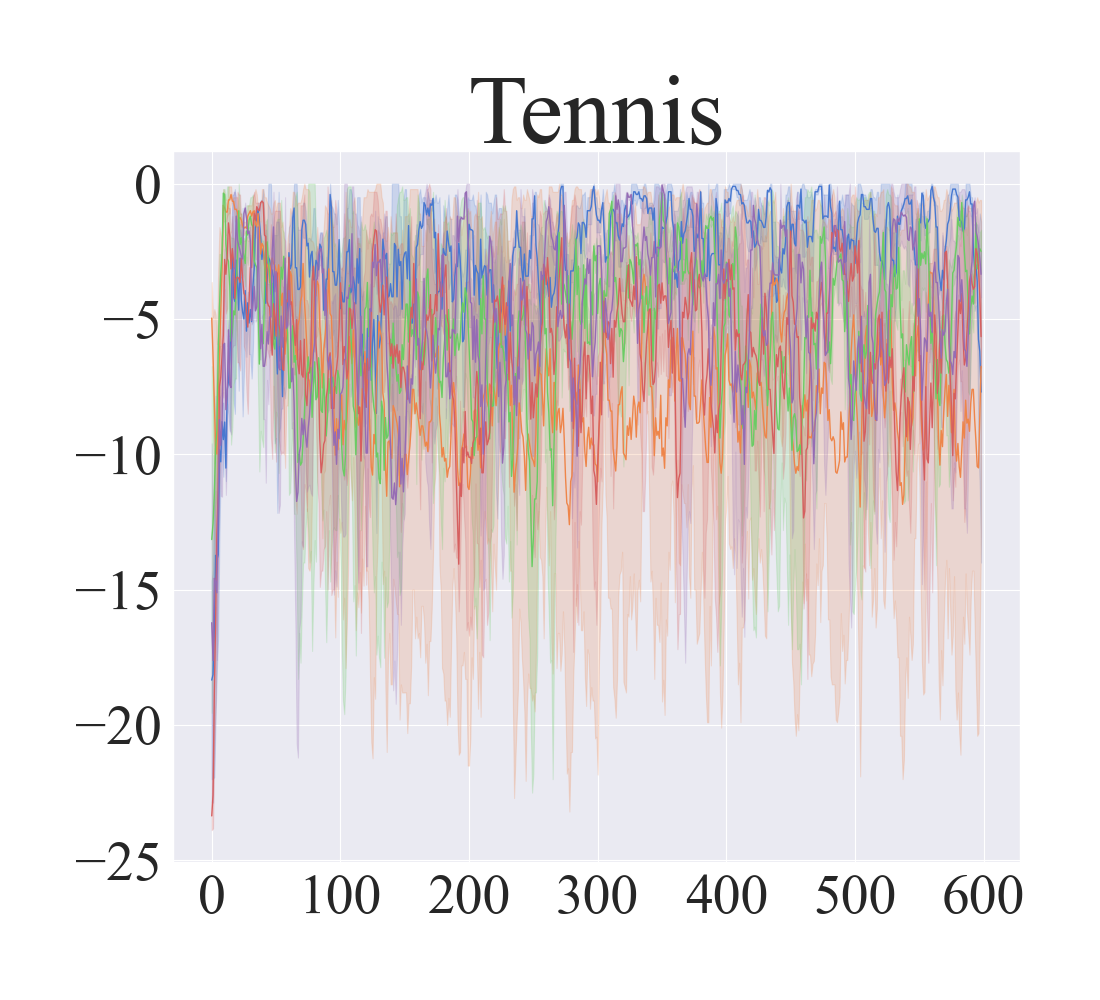}\\
			\end{minipage}%
		}%
		\subfigure{
			\begin{minipage}[t]{0.166\linewidth}
				\centering
				\includegraphics[width=1.05in]{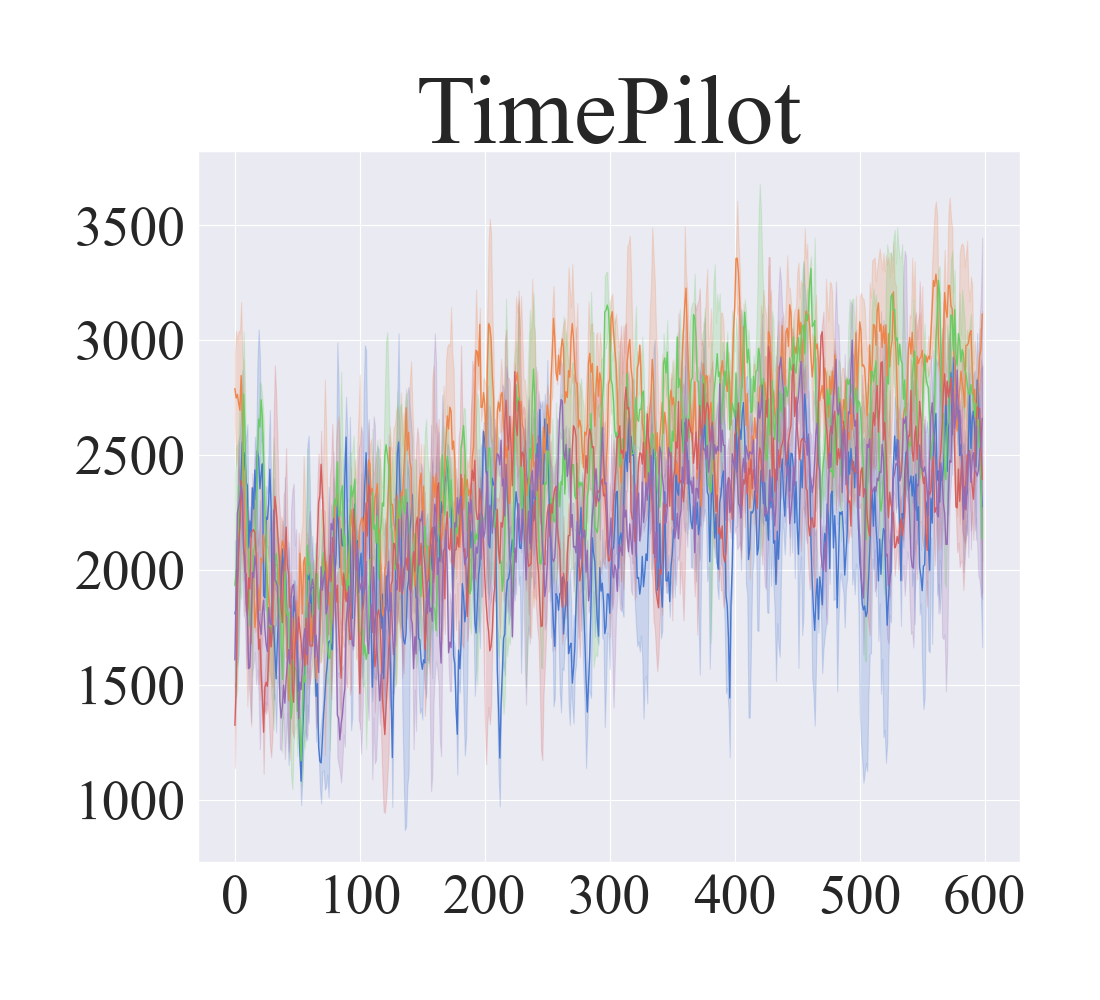}\\
			\end{minipage}%
		}%
		\subfigure{
			\begin{minipage}[t]{0.166\linewidth}
				\centering
				\includegraphics[width=1.05in]{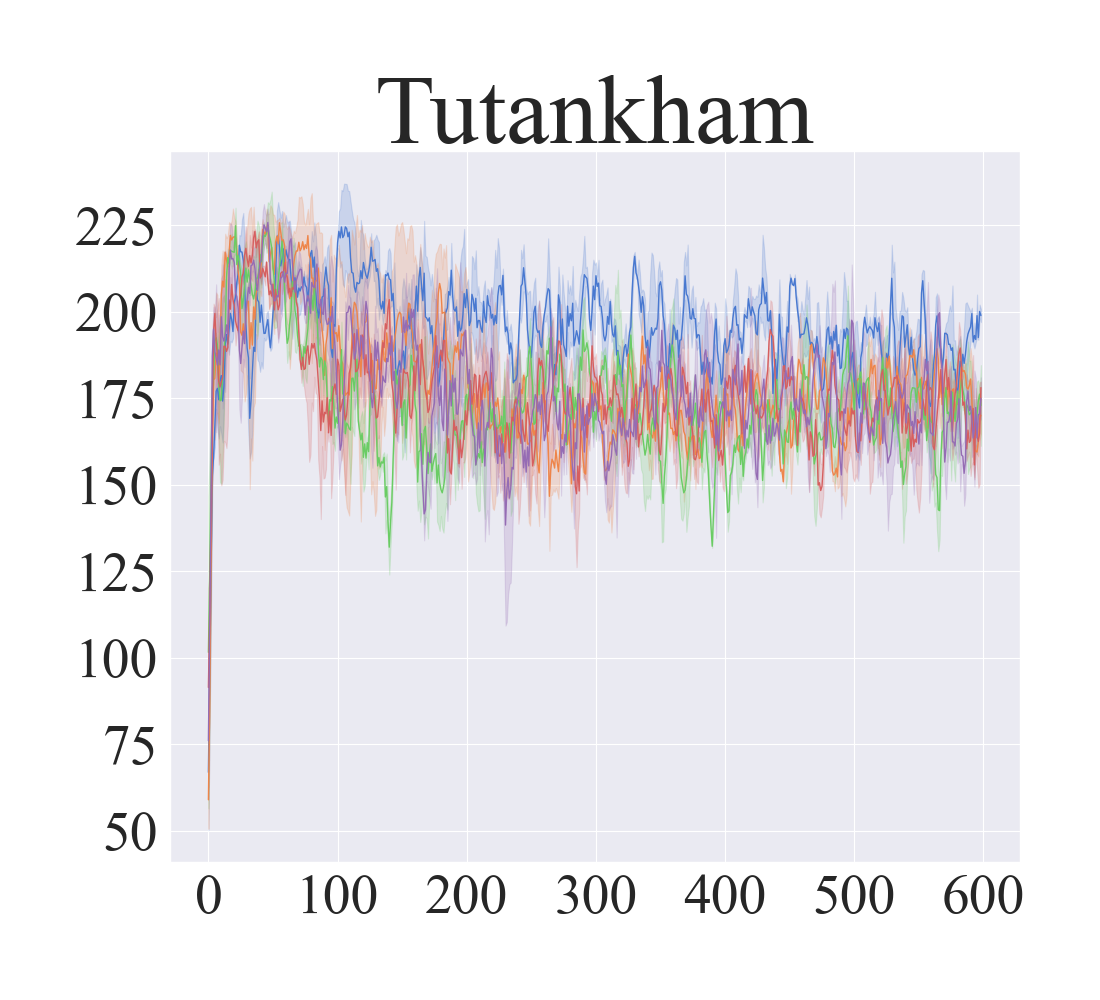}\\
			\end{minipage}%
		}%
		\vspace{-0.6cm}
		
		\subfigure{
			\begin{minipage}[t]{0.166\linewidth}
				\centering
				\includegraphics[width=1.05in]{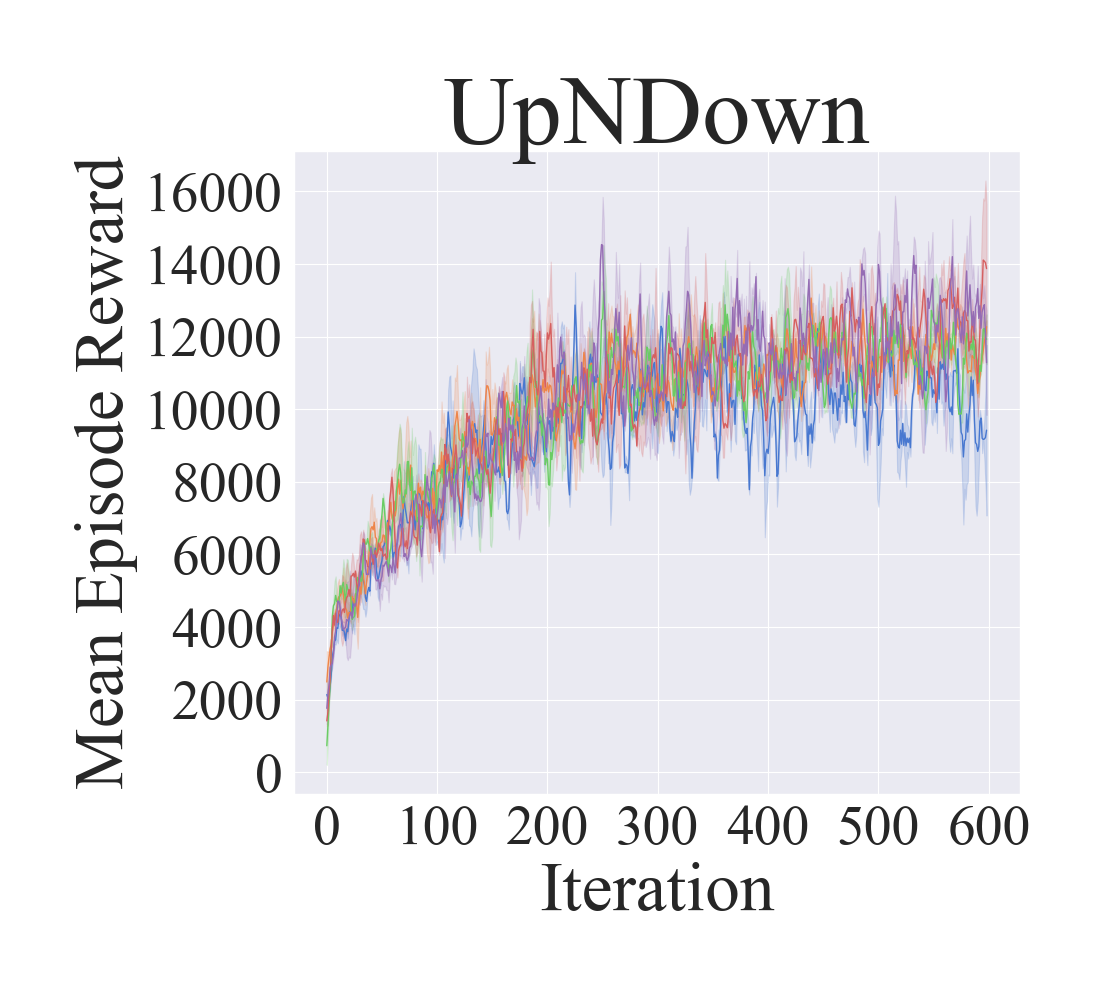}\\
			\end{minipage}%
		}%
		\subfigure{
			\begin{minipage}[t]{0.166\linewidth}
				\centering
				\includegraphics[width=1.05in]{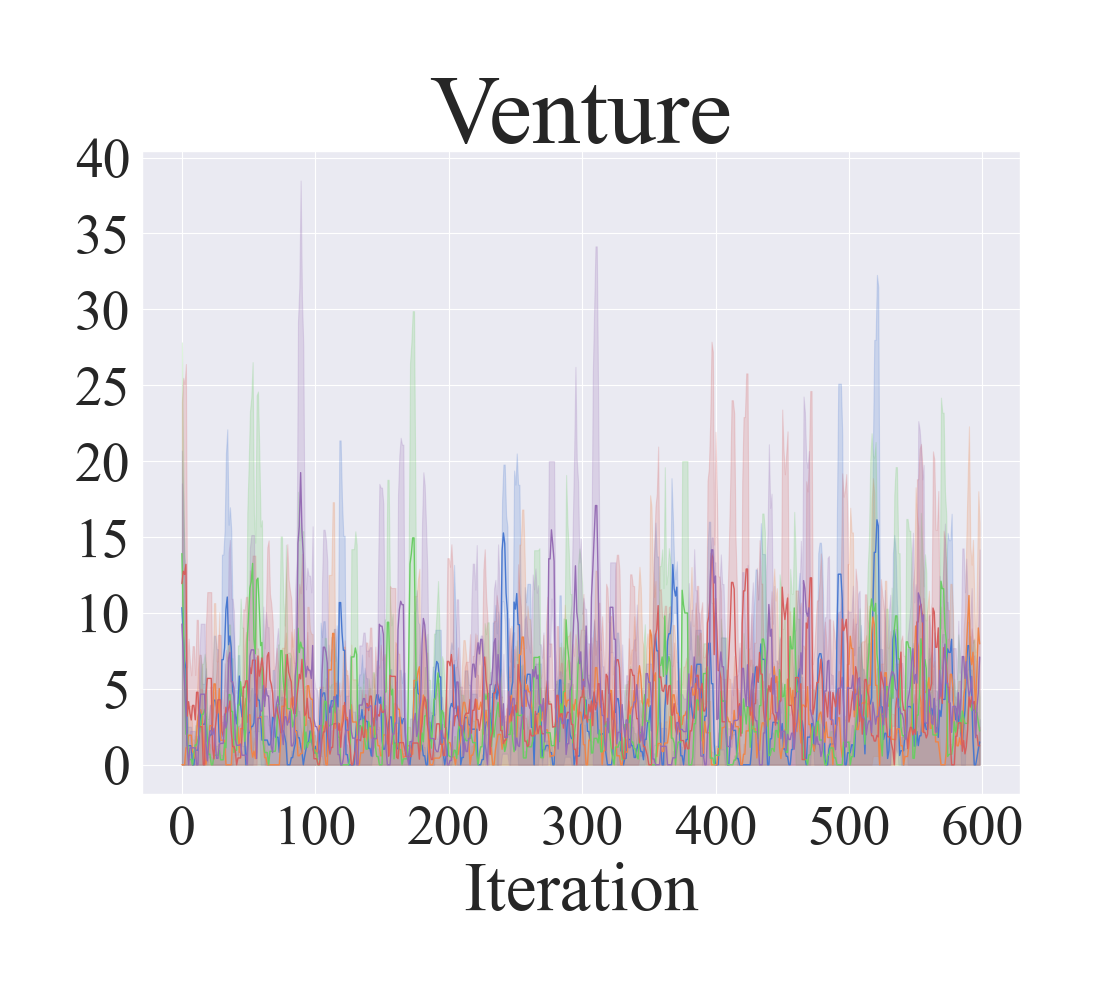}\\
			\end{minipage}%
		}%
		\subfigure{
			\begin{minipage}[t]{0.166\linewidth}
				\centering
				\includegraphics[width=1.05in]{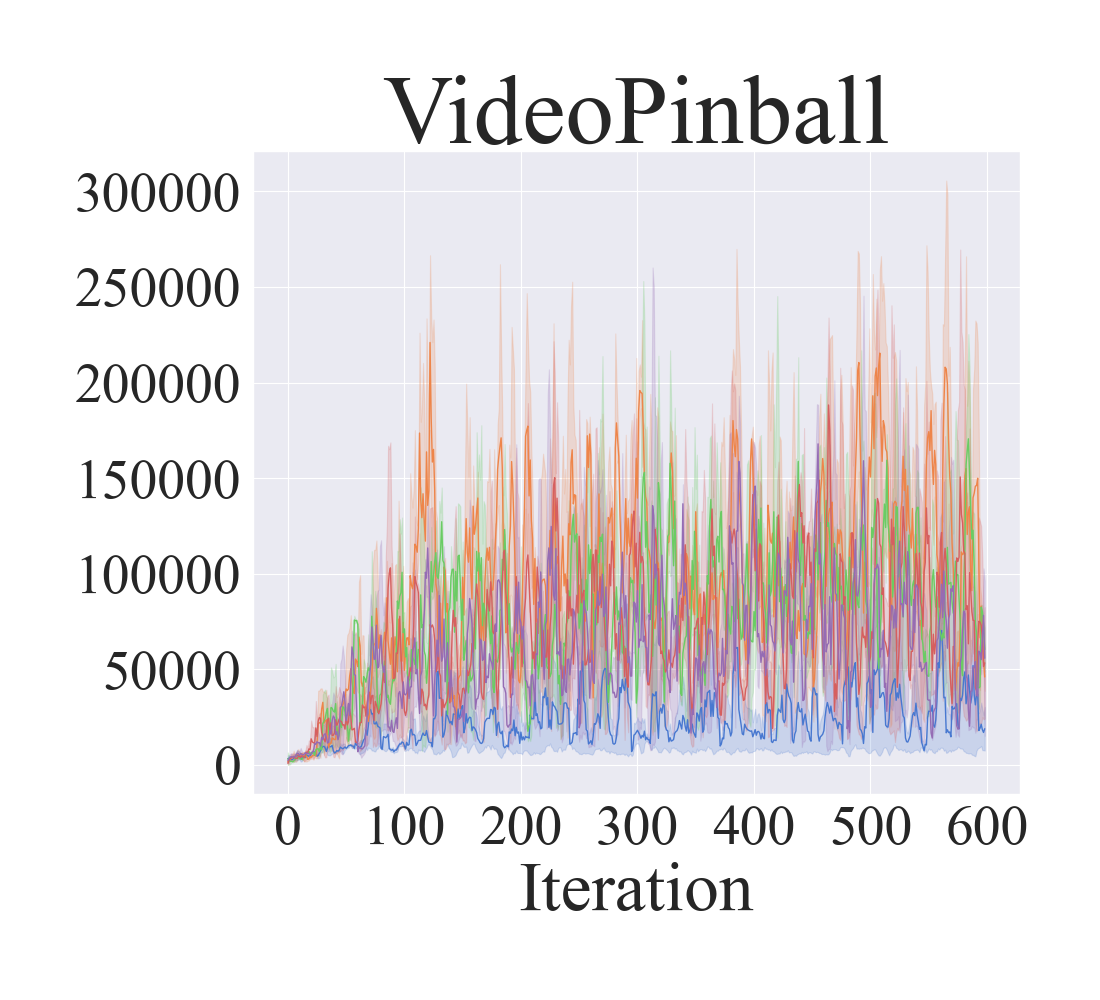}\\
			\end{minipage}%
		}%
		\subfigure{
			\begin{minipage}[t]{0.166\linewidth}
				\centering
				\includegraphics[width=1.05in]{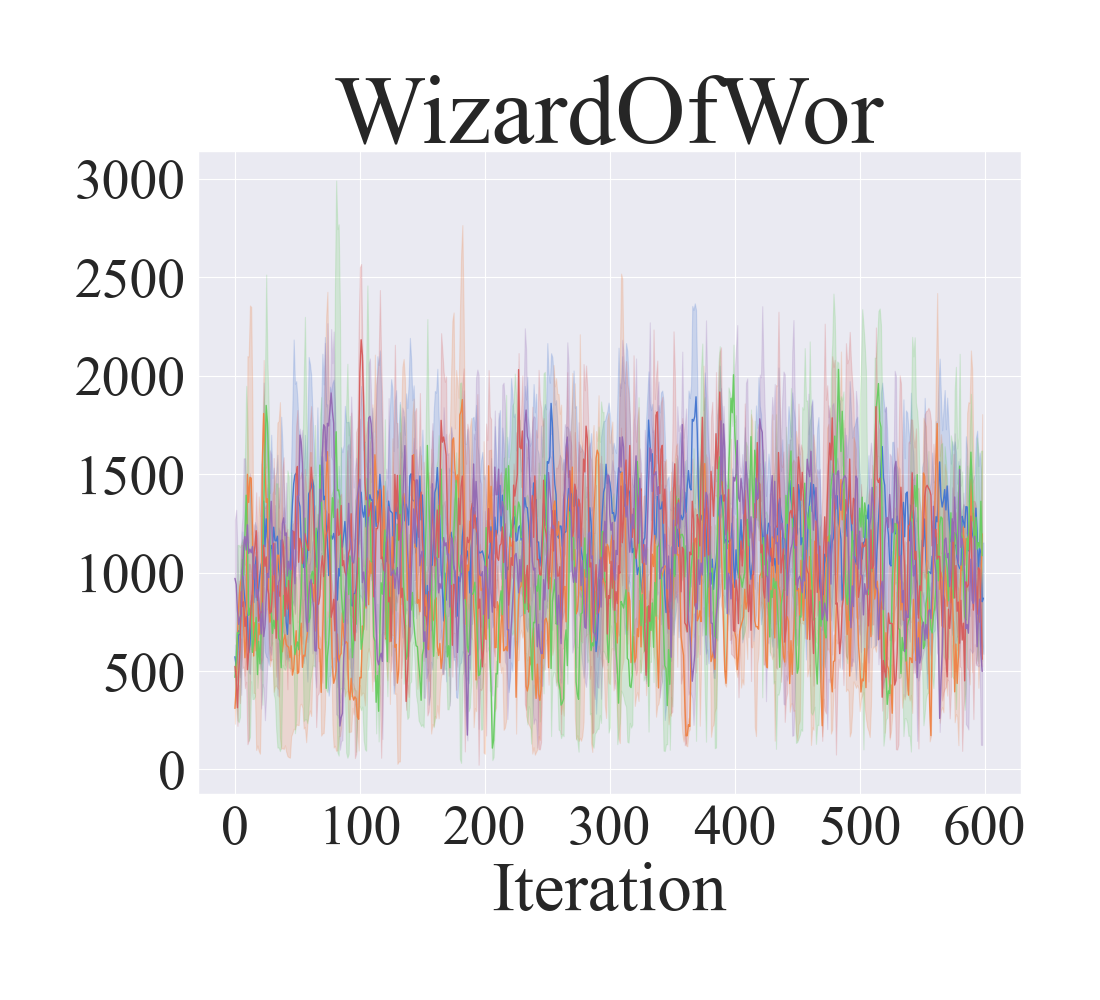}\\
			\end{minipage}%
		}%
		\subfigure{
			\begin{minipage}[t]{0.166\linewidth}
				\centering
				\includegraphics[width=1.05in]{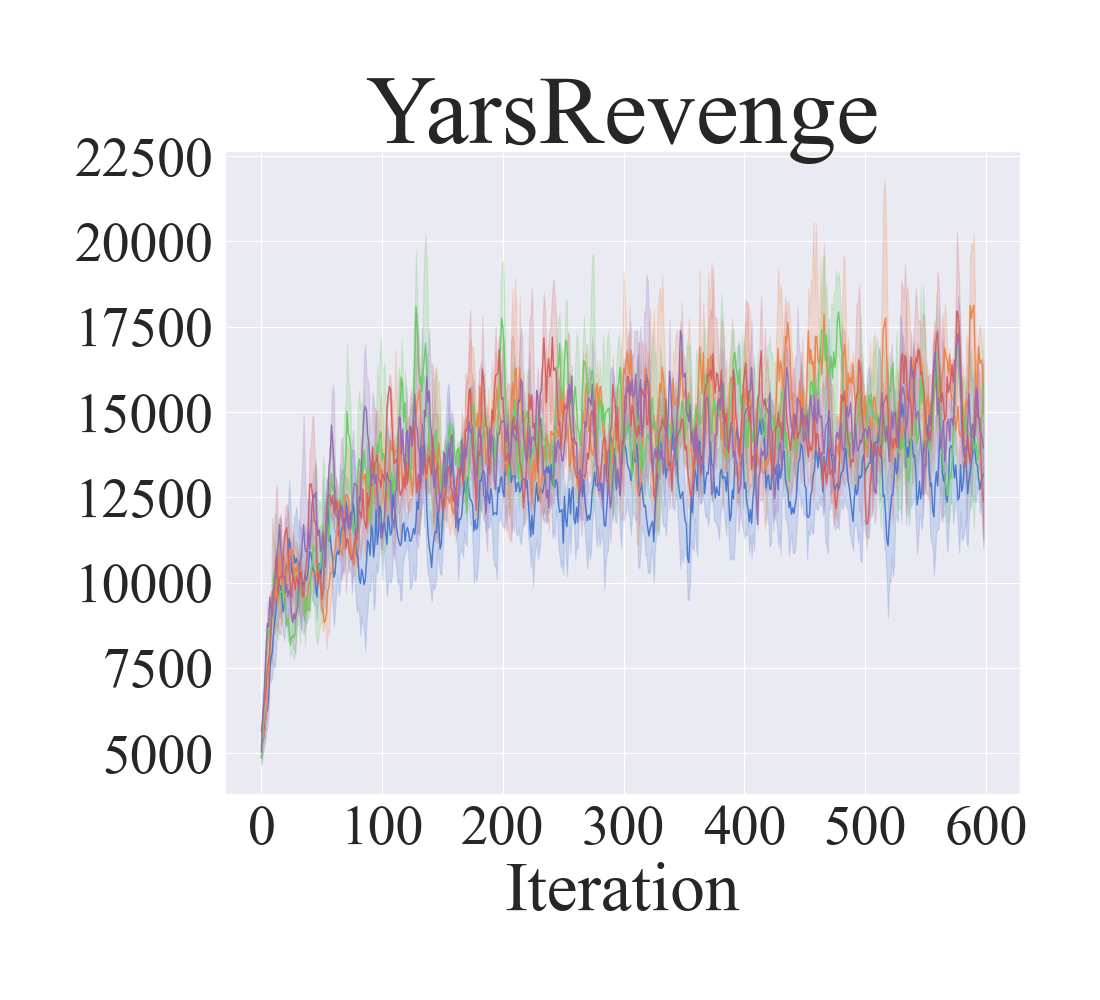}\\
			\end{minipage}%
		}%
		\subfigure{
			\begin{minipage}[t]{0.166\linewidth}
				\centering
				\includegraphics[width=1.05in]{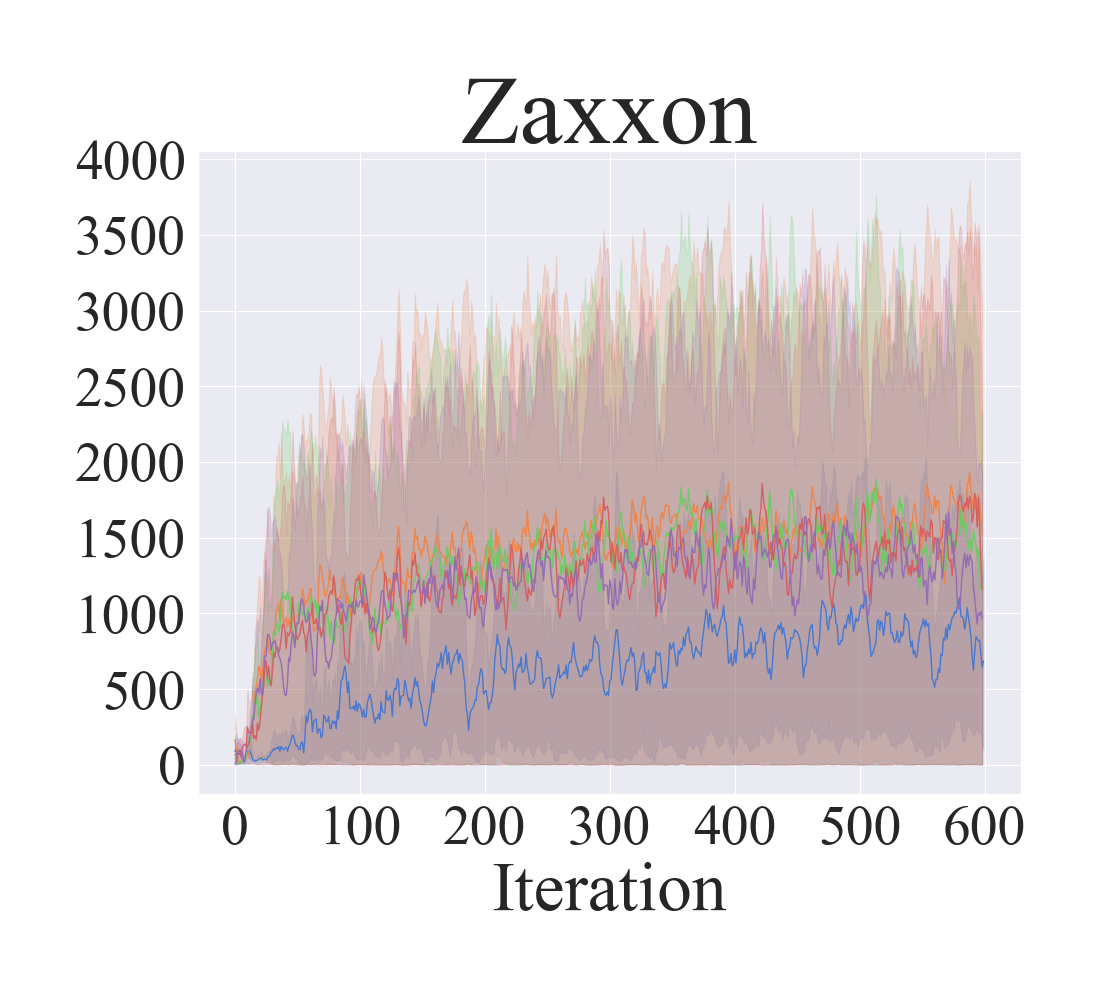}\\
			\end{minipage}%
		}%

		\centering
		\caption{\textbf{Learning curves of all $60$ Atari $2600$ games on poor dataset}}
		\label{fig: Learning curves of all $60$ Atari $2600$ games on poor dataset1}
								
	\end{figure*}

	Please refer Fig. \ref{fig: Learning curves of all $60$ Atari $2600$ games on poor dataset} and Fig. \ref{fig: Learning curves of all $60$ Atari $2600$ games on poor dataset1}.
								
	\subsection{Learning curves of all $60$ Atari $2600$ games on medium dataset}
\begin{figure*}[!htb]
		\centering
		
		\subfigure{
			\begin{minipage}[t]{\linewidth}
				\centering
				\includegraphics[width=4in]{legend.png}\\
			\end{minipage}%
		}%

		\subfigure{
			\begin{minipage}[t]{0.166\linewidth}
				\centering
				\includegraphics[width=1.05in]{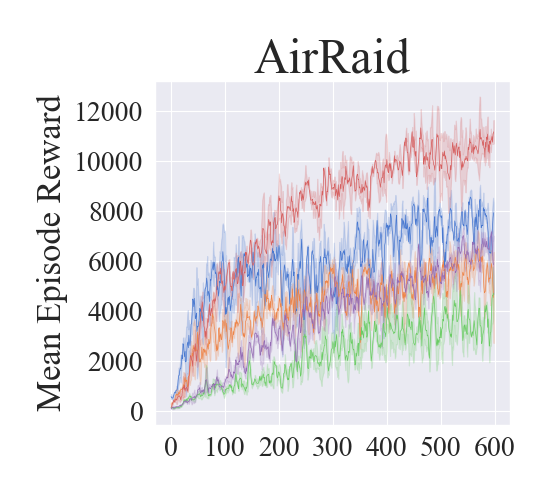}\\
			\end{minipage}%
		}%
		\subfigure{
			\begin{minipage}[t]{0.166\linewidth}
				\centering
				\includegraphics[width=1.05in]{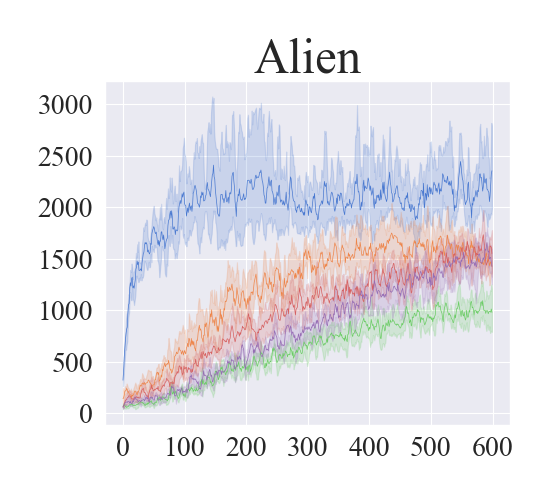}\\
			\end{minipage}%
		}%
		\subfigure{
			\begin{minipage}[t]{0.166\linewidth}
				\centering
				\includegraphics[width=1.05in]{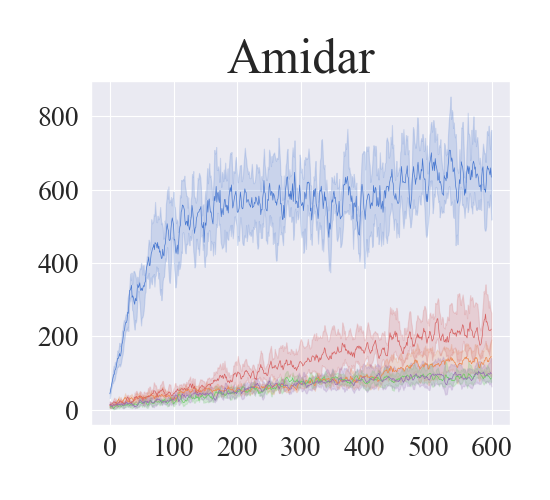}\\
			\end{minipage}%
		}%
		\subfigure{
			\begin{minipage}[t]{0.166\linewidth}
				\centering
				\includegraphics[width=1.05in]{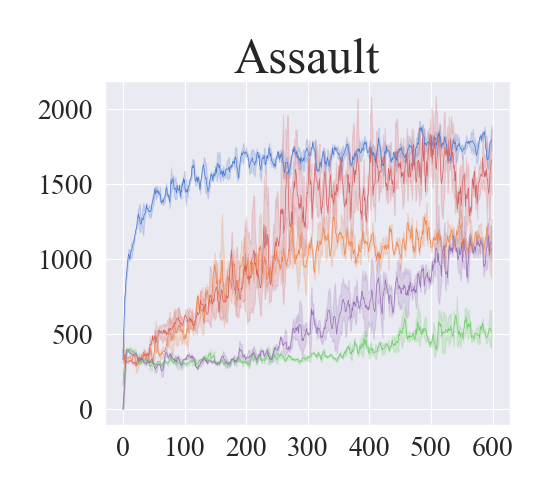}\\
			\end{minipage}%
		}%
		\subfigure{
			\begin{minipage}[t]{0.166\linewidth}
				\centering
				\includegraphics[width=1.05in]{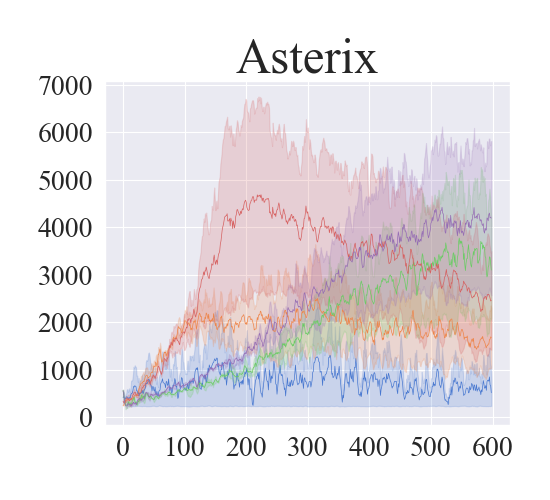}\\
			\end{minipage}%
		}%
		\subfigure{
			\begin{minipage}[t]{0.166\linewidth}
				\centering
				\includegraphics[width=1.05in]{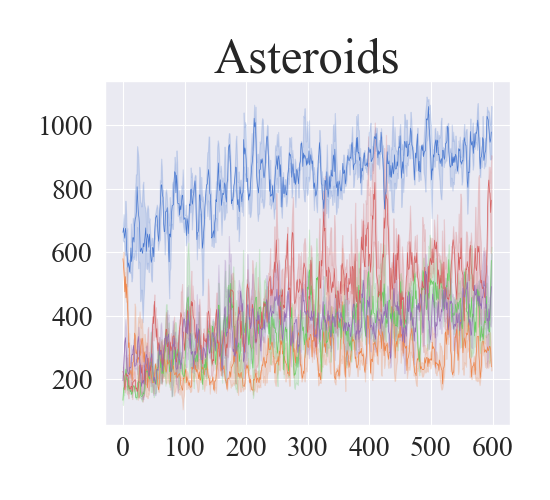}\\\
			\end{minipage}%
		}%
		\vspace{-1.0cm}
		
		\subfigure{
			\begin{minipage}[t]{0.166\linewidth}
				\centering
				\includegraphics[width=1.05in]{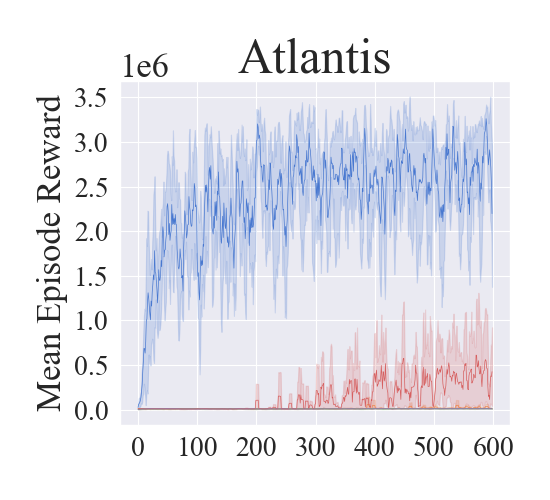}\\
			\end{minipage}%
		}%
		\subfigure{
			\begin{minipage}[t]{0.166\linewidth}
				\centering
				\includegraphics[width=1.05in]{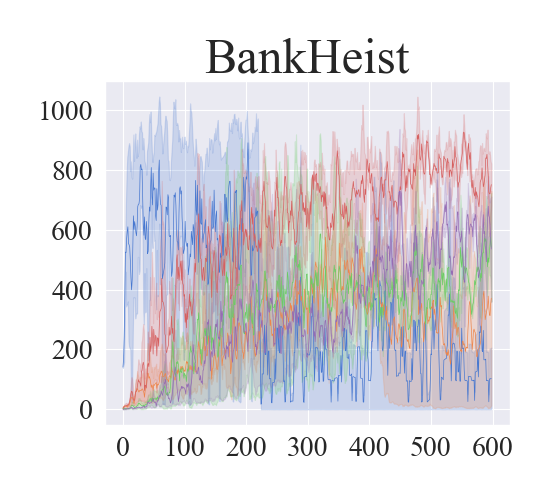}\\
			\end{minipage}%
		}%
		\subfigure{
			\begin{minipage}[t]{0.166\linewidth}
				\centering
				\includegraphics[width=1.05in]{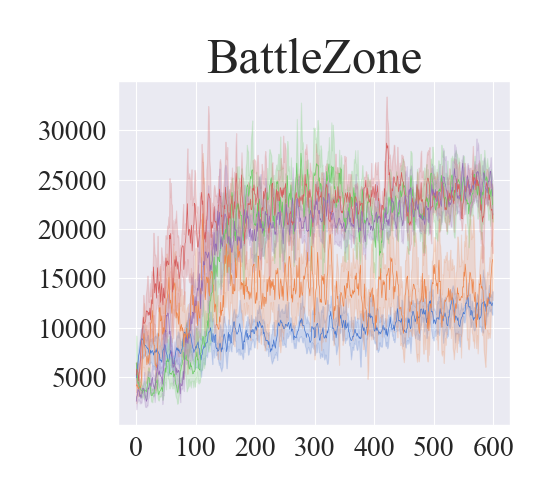}\\
			\end{minipage}%
		}%
		\subfigure{
			\begin{minipage}[t]{0.166\linewidth}
				\centering
				\includegraphics[width=1.05in]{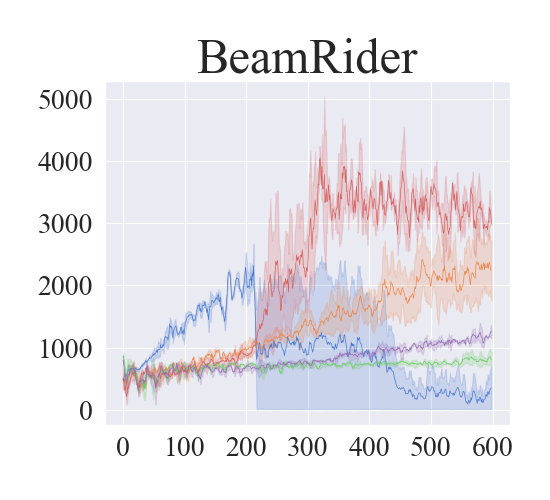}\\
			\end{minipage}%
		}%
		\subfigure{
			\begin{minipage}[t]{0.166\linewidth}
				\centering
				\includegraphics[width=1.05in]{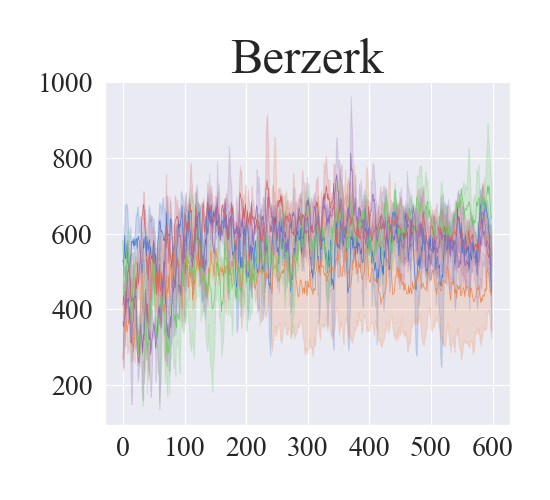}\\
			\end{minipage}%
		}%
		\subfigure{
			\begin{minipage}[t]{0.166\linewidth}
				\centering
				\includegraphics[width=1.05in]{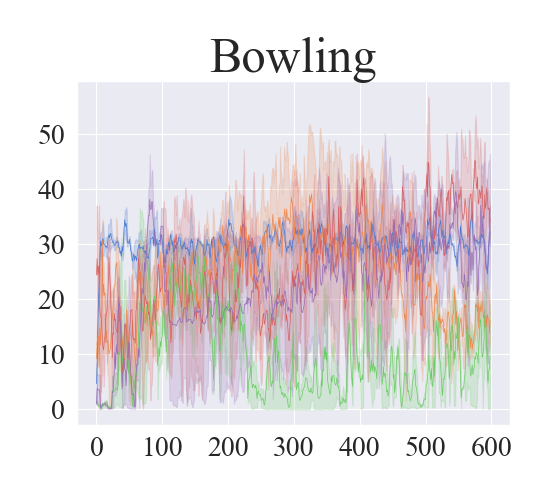}\\
			\end{minipage}%
		}%
		\vspace{-0.6cm}
		
		\subfigure{
			\begin{minipage}[t]{0.166\linewidth}
				\centering
				\includegraphics[width=1.05in]{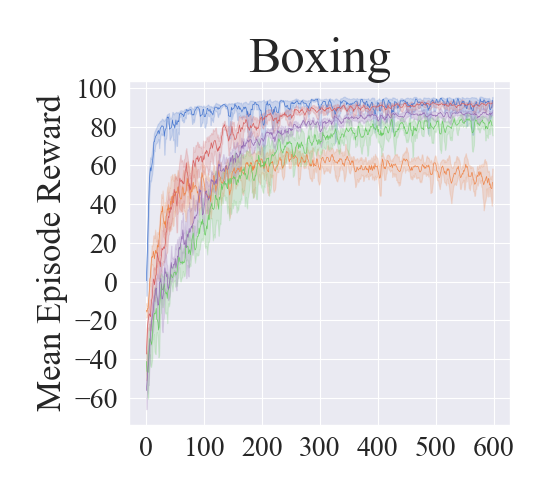}\\
			\end{minipage}%
		}%
		\subfigure{
			\begin{minipage}[t]{0.166\linewidth}
				\centering
				\includegraphics[width=1.05in]{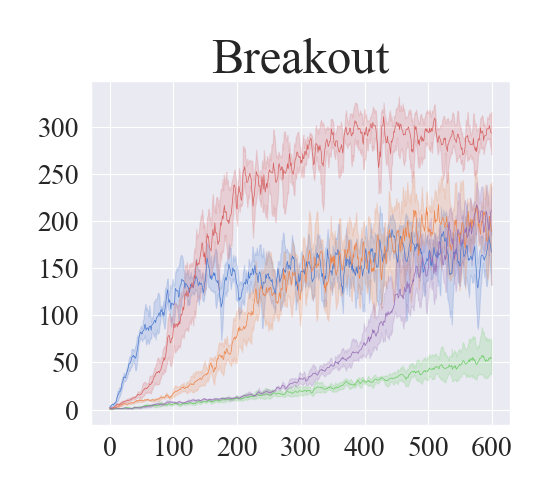}\\
			\end{minipage}%
		}%
		\subfigure{
			\begin{minipage}[t]{0.166\linewidth}
				\centering
				\includegraphics[width=1.05in]{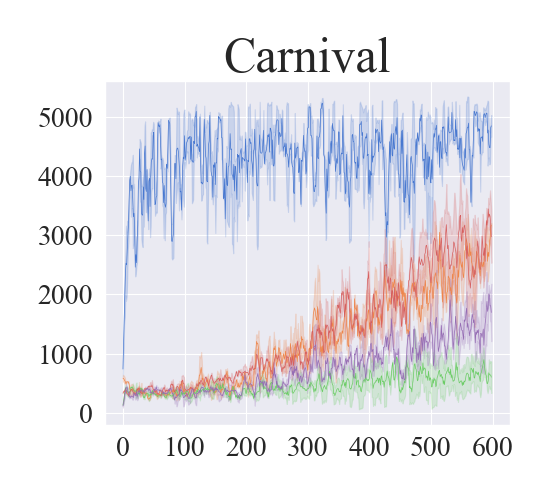}\\
			\end{minipage}%
		}%
		\subfigure{
			\begin{minipage}[t]{0.166\linewidth}
				\centering
				\includegraphics[width=1.05in]{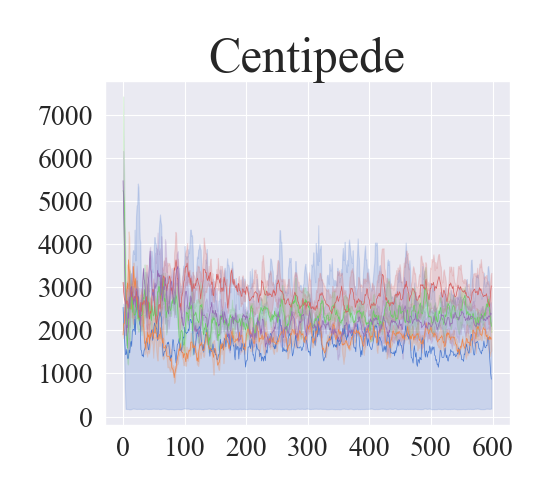}\\
			\end{minipage}%
		}%
		\subfigure{
			\begin{minipage}[t]{0.166\linewidth}
				\centering
				\includegraphics[width=1.05in]{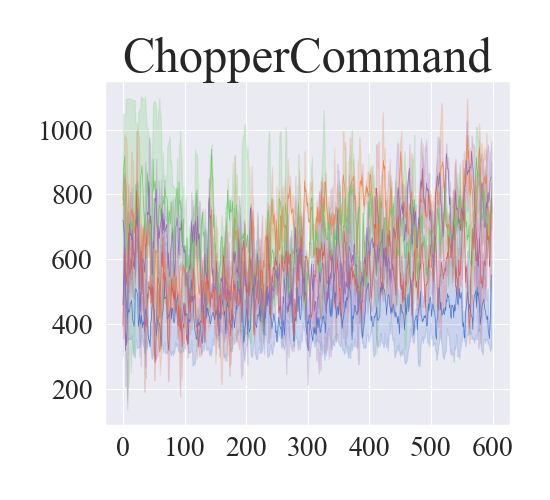}\\
			\end{minipage}%
		}%
		\subfigure{
			\begin{minipage}[t]{0.166\linewidth}
				\centering
				\includegraphics[width=1.05in]{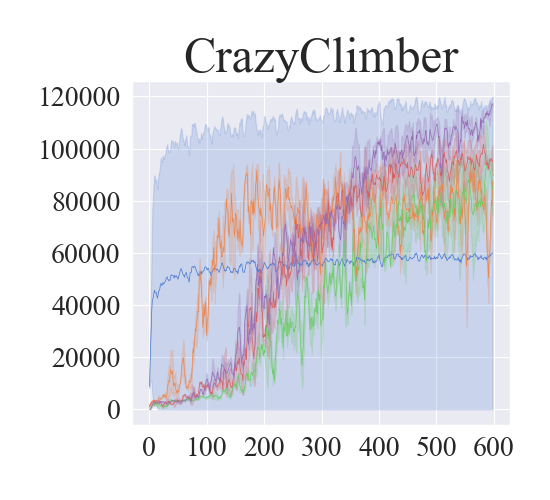}\\
			\end{minipage}%
		}%
		\vspace{-0.6cm}
		
		\subfigure{
			\begin{minipage}[t]{0.166\linewidth}
				\centering
				\includegraphics[width=1.05in]{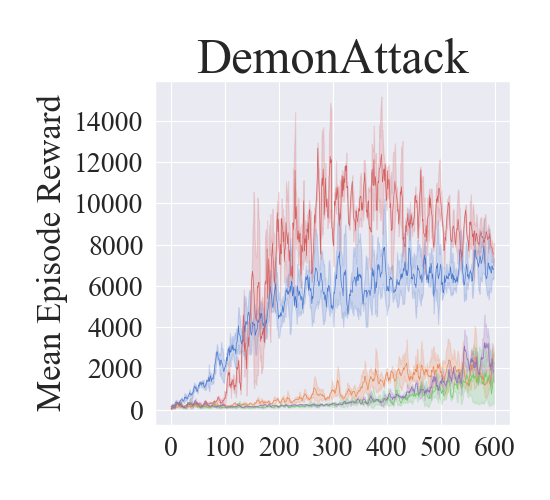}\\
			\end{minipage}%
		}%
		\subfigure{
			\begin{minipage}[t]{0.166\linewidth}
				\centering
				\includegraphics[width=1.05in]{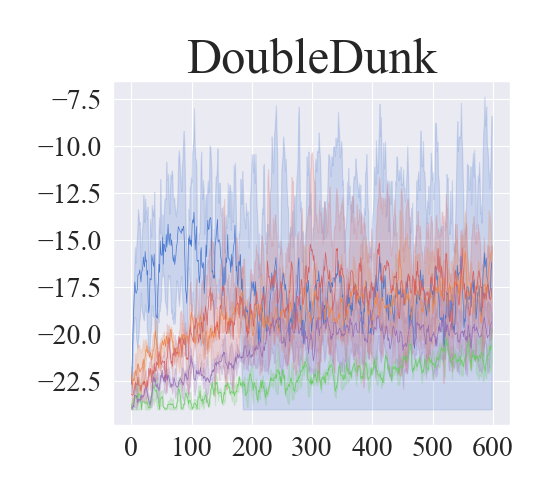}\\
			\end{minipage}%
		}%
		\subfigure{
			\begin{minipage}[t]{0.166\linewidth}
				\centering
				\includegraphics[width=1.05in]{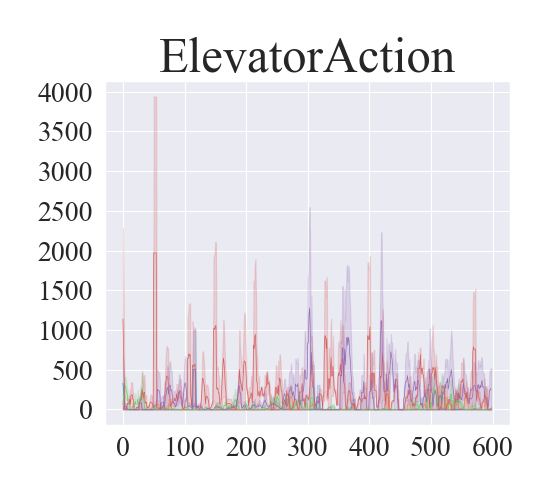}\\
			\end{minipage}%
		}%
		\subfigure{
			\begin{minipage}[t]{0.166\linewidth}
				\centering
				\includegraphics[width=1.05in]{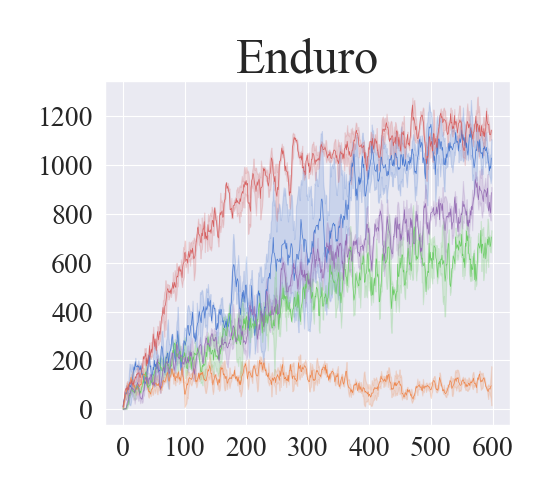}\\
			\end{minipage}%
		}%
		\subfigure{
			\begin{minipage}[t]{0.166\linewidth}
				\centering
				\includegraphics[width=1.05in]{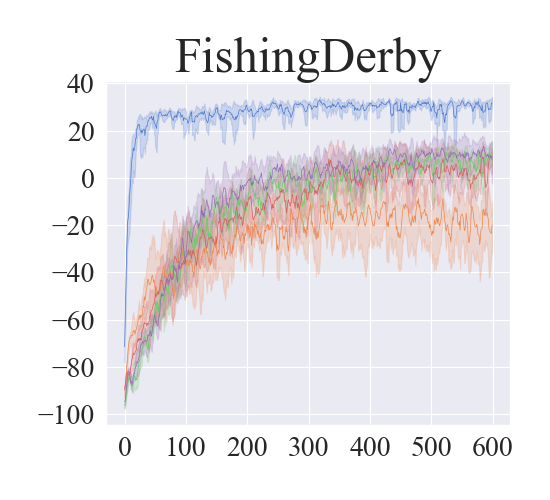}\\
			\end{minipage}%
		}%
		\subfigure{
			\begin{minipage}[t]{0.166\linewidth}
				\centering
				\includegraphics[width=1.05in]{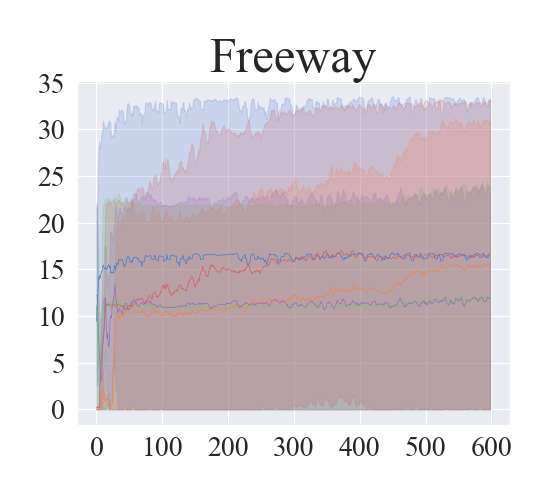}\\
			\end{minipage}%
		}%
		\vspace{-0.6cm}
		
		\subfigure{
			\begin{minipage}[t]{0.166\linewidth}
				\centering
				\includegraphics[width=1.05in]{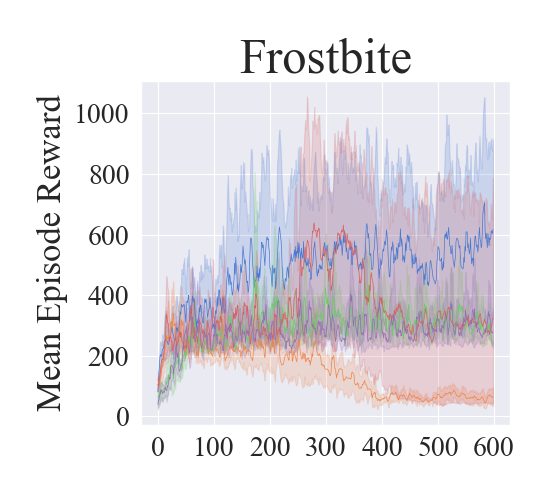}\\
			\end{minipage}%
		}%
		\subfigure{
			\begin{minipage}[t]{0.166\linewidth}
				\centering
				\includegraphics[width=1.05in]{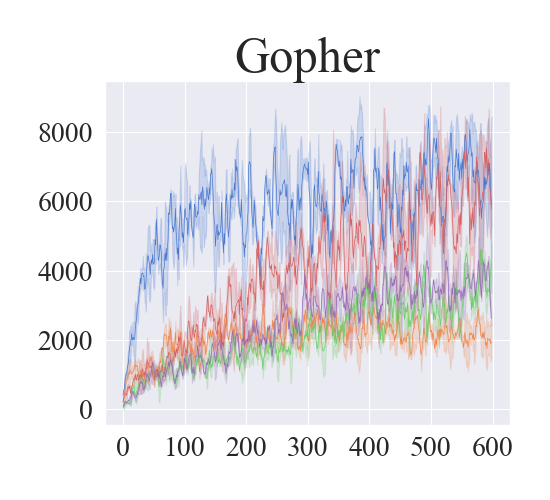}\\
			\end{minipage}%
		}%
		\subfigure{
			\begin{minipage}[t]{0.166\linewidth}
				\centering
				\includegraphics[width=1.05in]{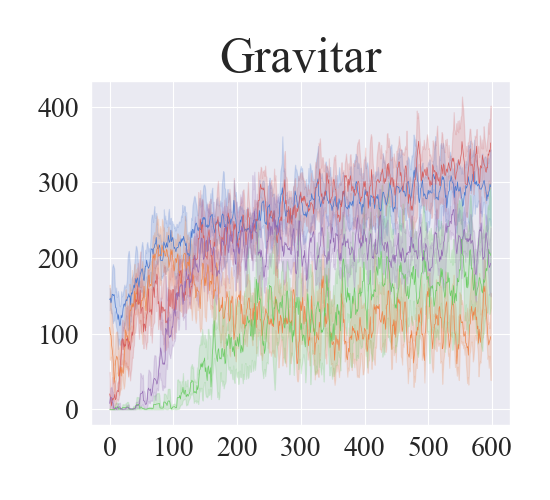}\\
			\end{minipage}%
		}%
		\subfigure{
			\begin{minipage}[t]{0.166\linewidth}
				\centering
				\includegraphics[width=1.05in]{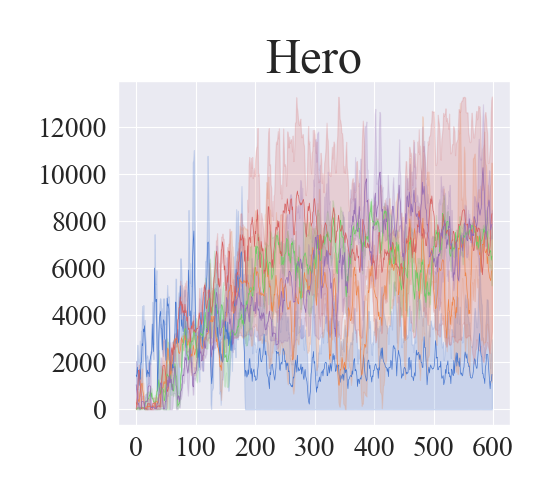}\\
			\end{minipage}%
		}%
		\subfigure{
			\begin{minipage}[t]{0.166\linewidth}
				\centering
				\includegraphics[width=1.05in]{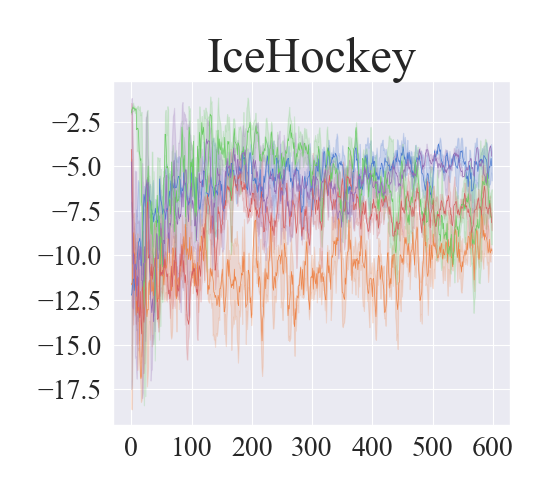}\\
			\end{minipage}%
		}%
		\subfigure{
			\begin{minipage}[t]{0.166\linewidth}
				\centering
				\includegraphics[width=1.05in]{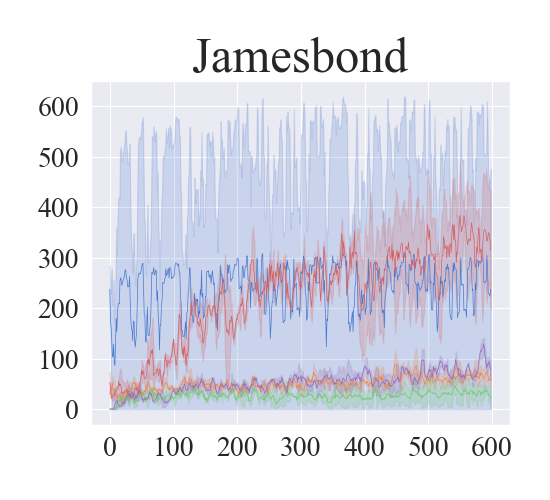}\\
			\end{minipage}%
		}%
		\vspace{-0.6cm}
		
		\subfigure{
			\begin{minipage}[t]{0.166\linewidth}
				\centering
				\includegraphics[width=1.05in]{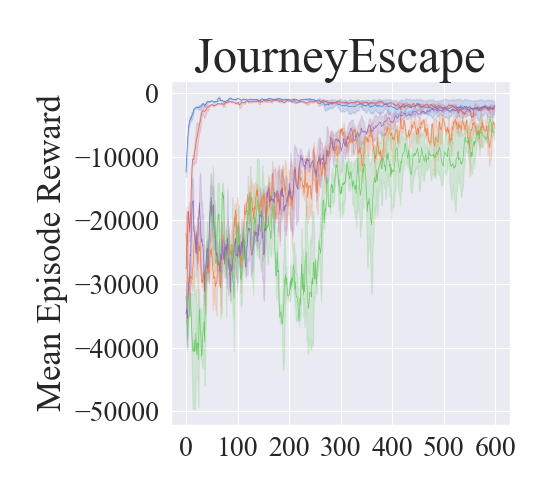}\\
			\end{minipage}%
		}%
		\subfigure{
			\begin{minipage}[t]{0.166\linewidth}
				\centering
				\includegraphics[width=1.05in]{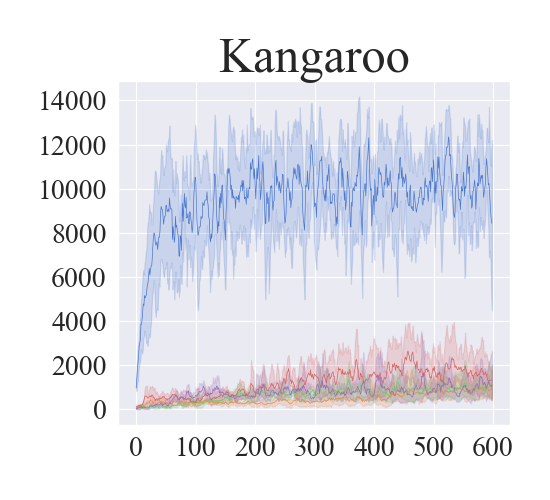}\\
			\end{minipage}%
		}%
		\subfigure{
			\begin{minipage}[t]{0.166\linewidth}
				\centering
				\includegraphics[width=1.05in]{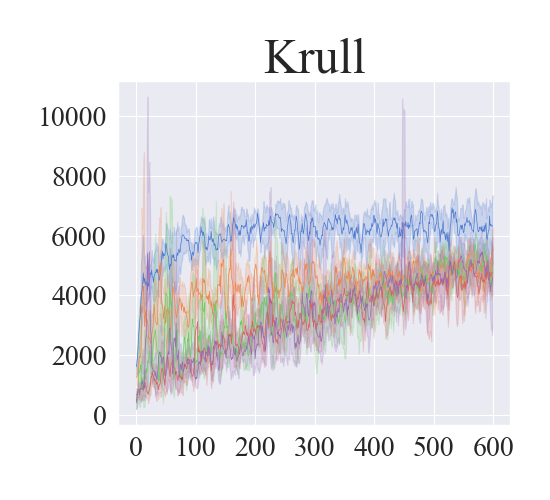}\\
			\end{minipage}%
		}%
		\subfigure{
			\begin{minipage}[t]{0.166\linewidth}
				\centering
				\includegraphics[width=1.05in]{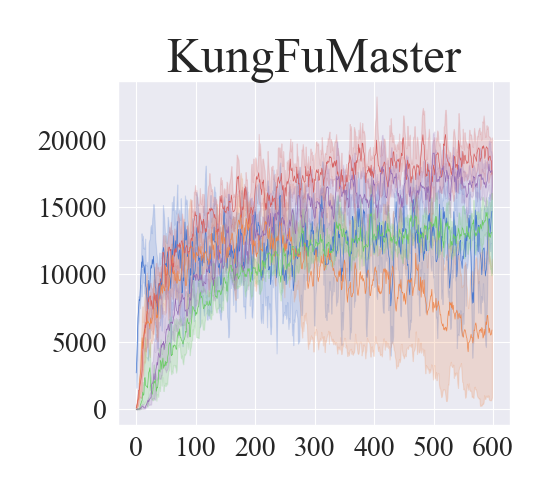}\\
			\end{minipage}%
		}%
		\subfigure{
			\begin{minipage}[t]{0.166\linewidth}
				\centering
				\includegraphics[width=1.05in]{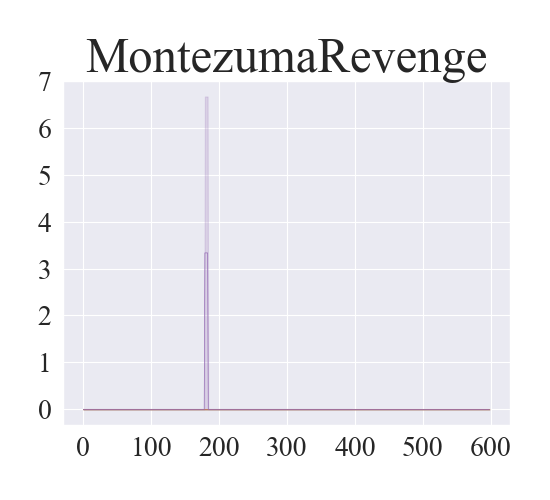}\\
			\end{minipage}%
		}%
		\subfigure{
			\begin{minipage}[t]{0.166\linewidth}
				\centering
				\includegraphics[width=1.05in]{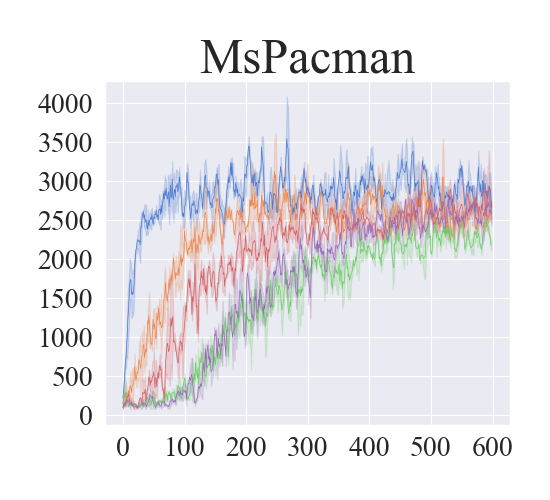}\\
			\end{minipage}%
		}%
		\vspace{-0.6cm}
		
		\subfigure{
			\begin{minipage}[t]{0.166\linewidth}
				\centering
				\includegraphics[width=1.05in]{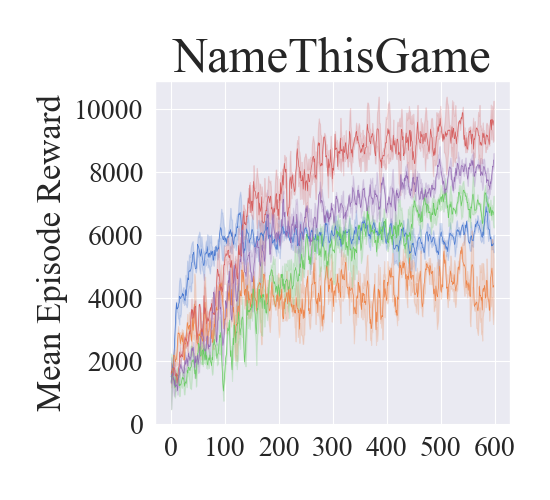}\\
			\end{minipage}%
		}%
		\subfigure{
			\begin{minipage}[t]{0.166\linewidth}
				\centering
				\includegraphics[width=1.05in]{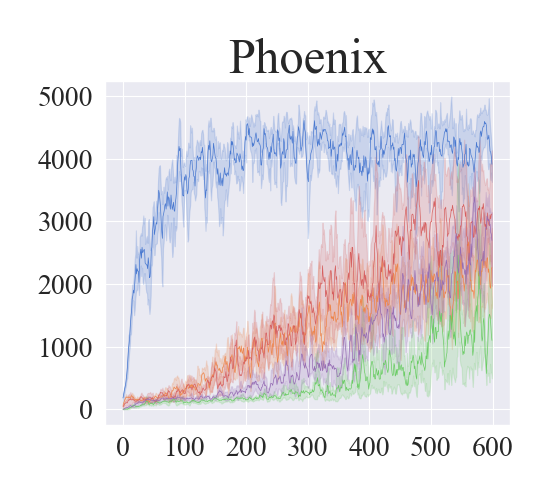}\\
			\end{minipage}%
		}%
		\subfigure{
			\begin{minipage}[t]{0.166\linewidth}
				\centering
				\includegraphics[width=1.05in]{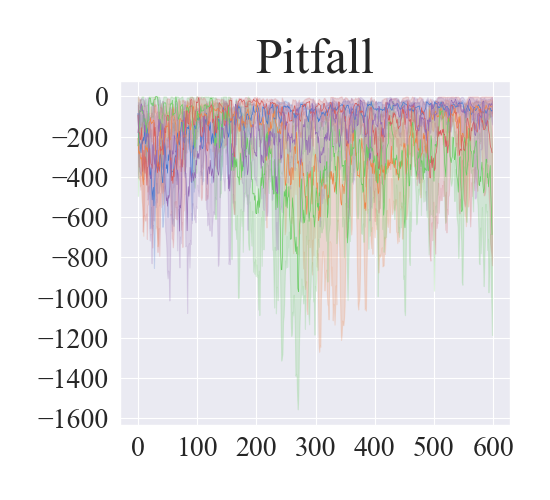}\\
			\end{minipage}%
		}%
		\subfigure{
			\begin{minipage}[t]{0.166\linewidth}
				\centering
				\includegraphics[width=1.05in]{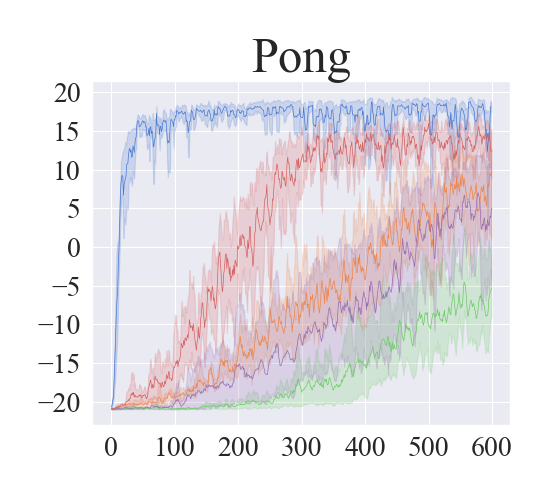}\\
			\end{minipage}%
		}%
		\subfigure{
			\begin{minipage}[t]{0.166\linewidth}
				\centering
				\includegraphics[width=1.05in]{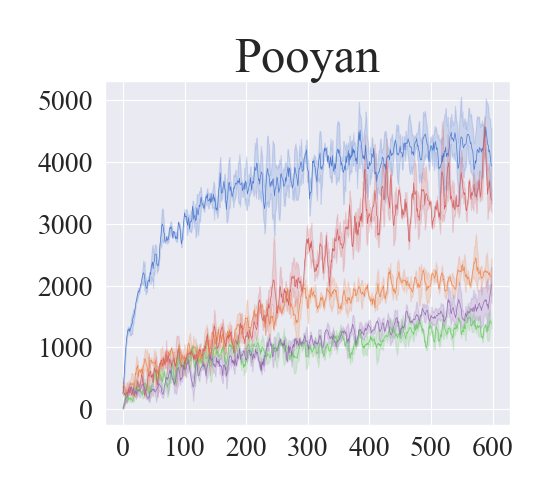}\\
			\end{minipage}%
		}%
		\subfigure{
			\begin{minipage}[t]{0.166\linewidth}
				\centering
				\includegraphics[width=1.05in]{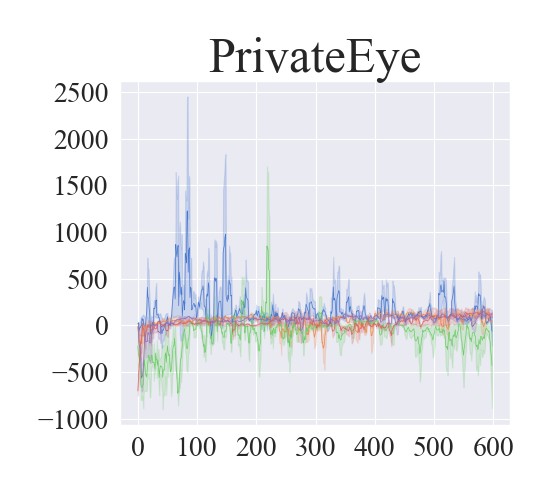}\\
			\end{minipage}%
		}%
		\vspace{-0.6cm}
		
		\subfigure{
			\begin{minipage}[t]{0.166\linewidth}
				\centering
				\includegraphics[width=1.05in]{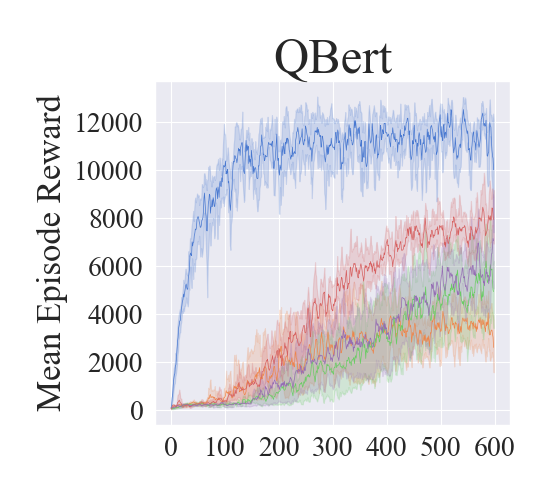}\\
			\end{minipage}%
		}%
		\subfigure{
			\begin{minipage}[t]{0.166\linewidth}
				\centering
				\includegraphics[width=1.05in]{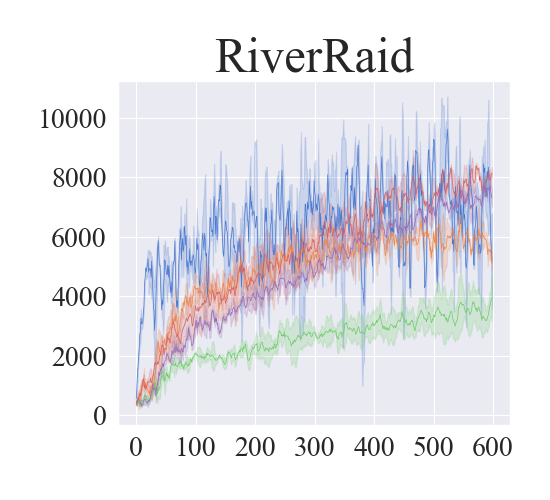}\\
			\end{minipage}%
		}%
		\subfigure{
			\begin{minipage}[t]{0.166\linewidth}
				\centering
				\includegraphics[width=1.05in]{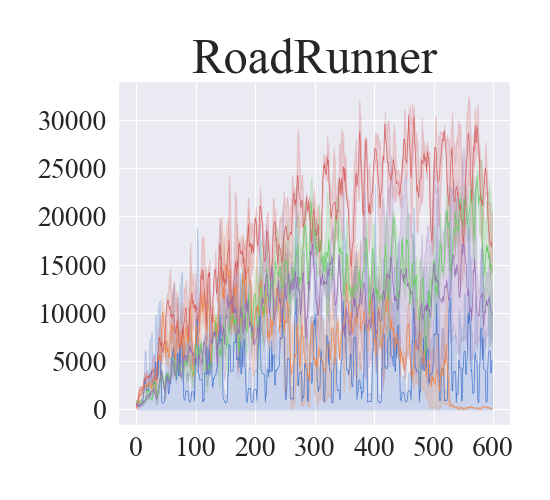}\\
			\end{minipage}%
		}%
		\subfigure{
			\begin{minipage}[t]{0.166\linewidth}
				\centering
				\includegraphics[width=1.05in]{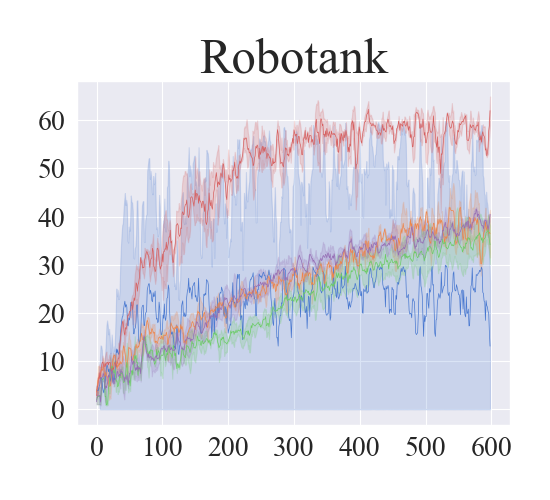}\\
			\end{minipage}%
		}%
		\subfigure{
			\begin{minipage}[t]{0.166\linewidth}
				\centering
				\includegraphics[width=1.05in]{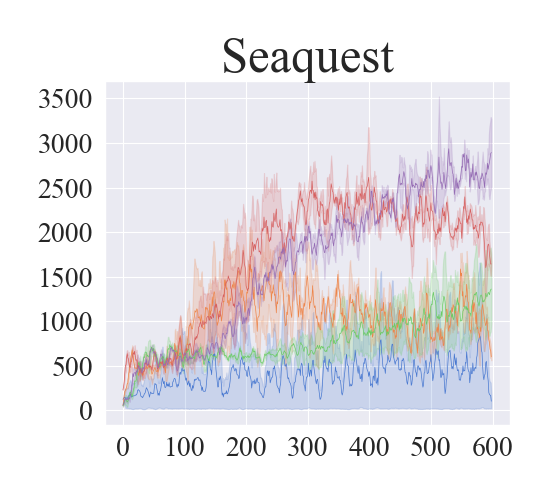}\\
			\end{minipage}%
		}%
		\subfigure{
			\begin{minipage}[t]{0.166\linewidth}
				\centering
				\includegraphics[width=1.05in]{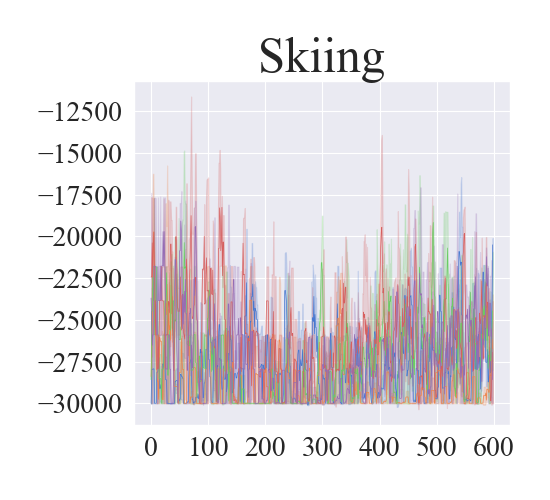}\\
			\end{minipage}%
		}%
		\vspace{-0.6cm}
		
		\subfigure{
			\begin{minipage}[t]{0.166\linewidth}
				\centering
				\includegraphics[width=1.05in]{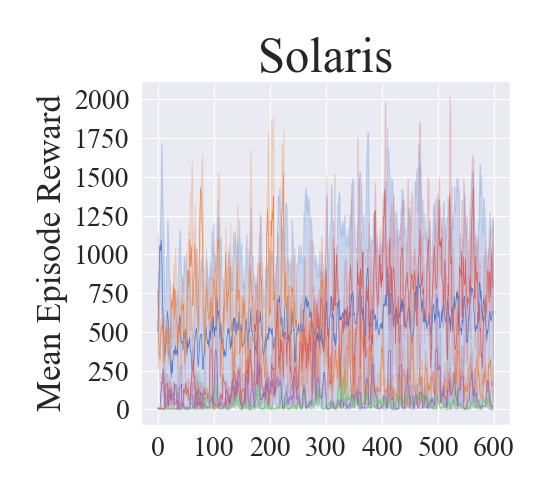}\\
			\end{minipage}%
		}%
		\subfigure{
			\begin{minipage}[t]{0.166\linewidth}
				\centering
				\includegraphics[width=1.05in]{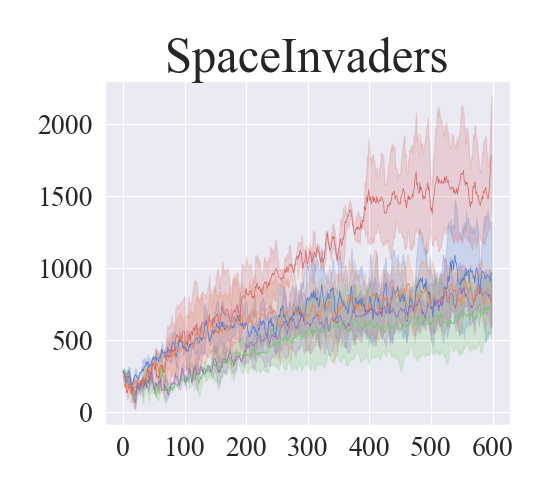}\\
			\end{minipage}%
		}%
		\subfigure{
			\begin{minipage}[t]{0.166\linewidth}
				\centering
				\includegraphics[width=1.05in]{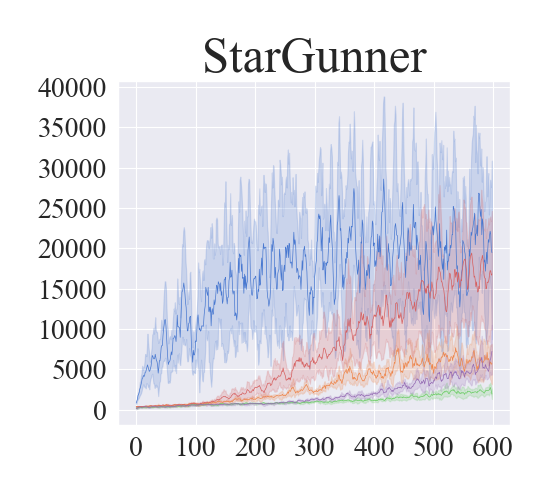}\\
			\end{minipage}%
		}%
		\subfigure{
			\begin{minipage}[t]{0.166\linewidth}
				\centering
				\includegraphics[width=1.05in]{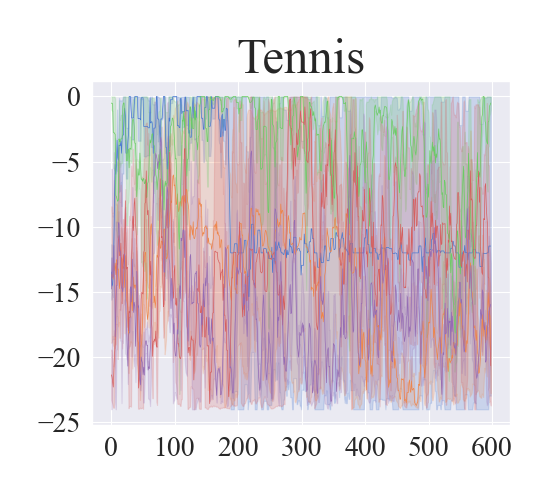}\\
			\end{minipage}%
		}%
		\subfigure{
			\begin{minipage}[t]{0.166\linewidth}
				\centering
				\includegraphics[width=1.05in]{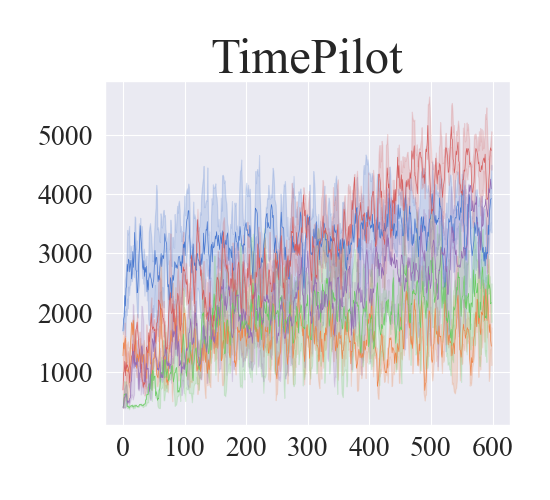}\\
			\end{minipage}%
		}%
		\subfigure{
			\begin{minipage}[t]{0.166\linewidth}
				\centering
				\includegraphics[width=1.05in]{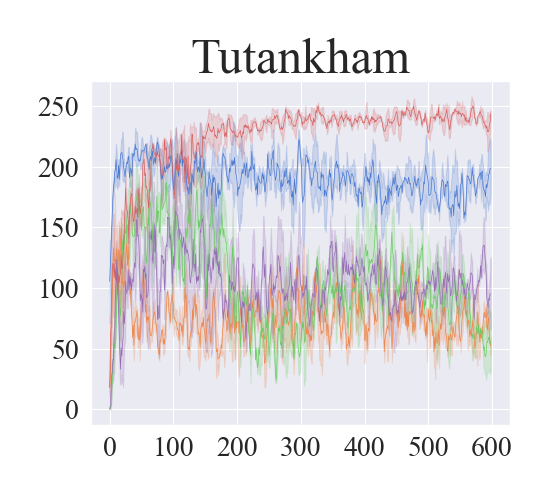}\\
			\end{minipage}%
		}%
		\vspace{-0.6cm}
		
		\subfigure{
			\begin{minipage}[t]{0.166\linewidth}
				\centering
				\includegraphics[width=1.05in]{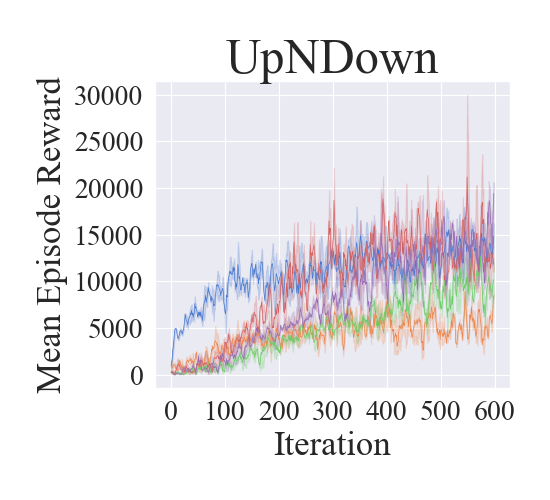}\\
			\end{minipage}%
		}%
		\subfigure{
			\begin{minipage}[t]{0.166\linewidth}
				\centering
				\includegraphics[width=1.05in]{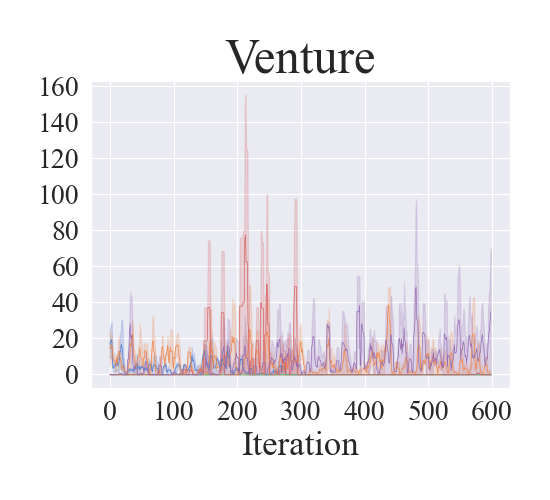}\\
			\end{minipage}%
		}%
		\subfigure{
			\begin{minipage}[t]{0.166\linewidth}
				\centering
				\includegraphics[width=1.05in]{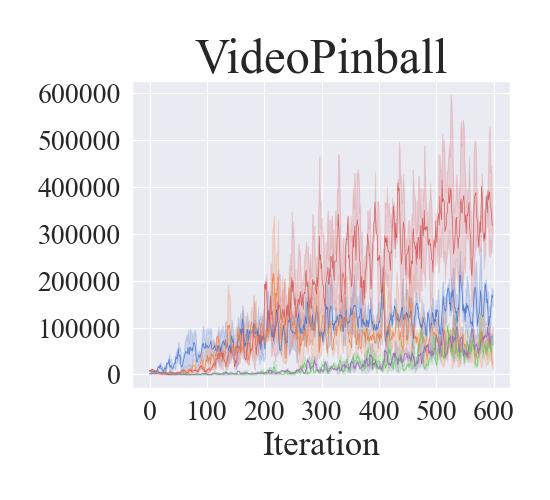}\\
			\end{minipage}%
		}%
		\subfigure{
			\begin{minipage}[t]{0.166\linewidth}
				\centering
				\includegraphics[width=1.05in]{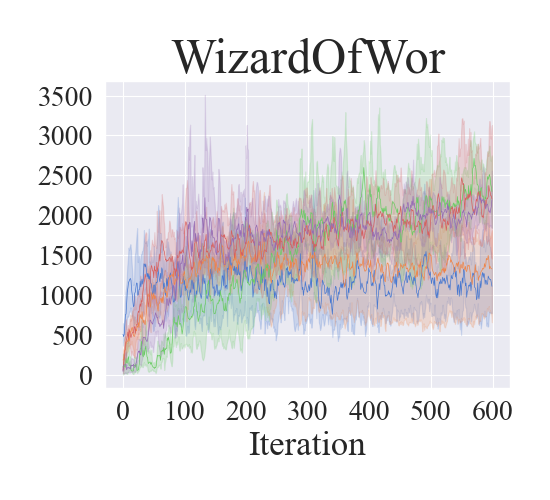}\\
			\end{minipage}%
		}%
		\subfigure{
			\begin{minipage}[t]{0.166\linewidth}
				\centering
				\includegraphics[width=1.05in]{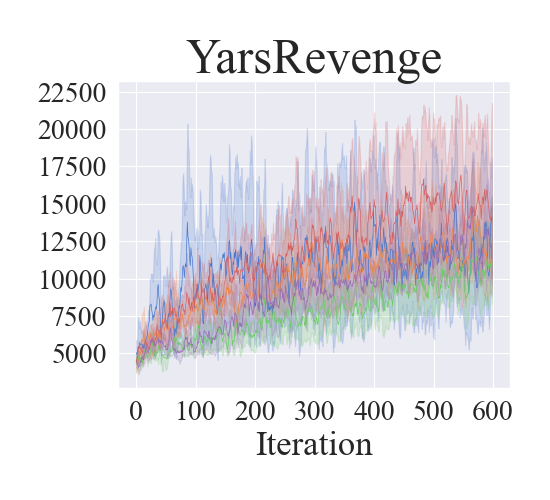}\\
			\end{minipage}%
		}%
		\subfigure{
			\begin{minipage}[t]{0.166\linewidth}
				\centering
				\includegraphics[width=1.05in]{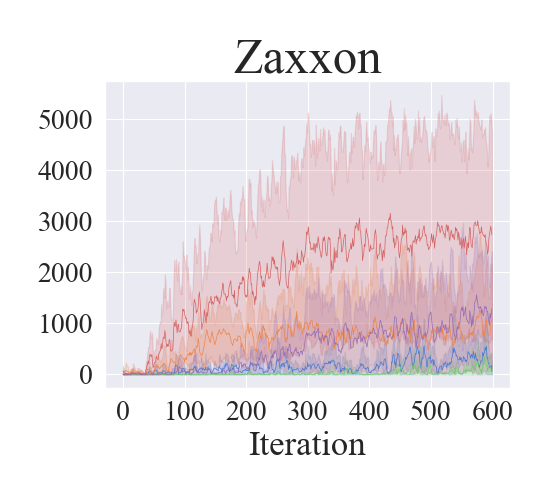}\\
			\end{minipage}%
		}%

		\centering
		\caption{\textbf{Learning curves of all $60$ Atari $2600$ games on medium dataset}}
		\label{fig:Learning curves of all $60$ Atari $2600$ games on medium dataset}
								
	\end{figure*}

\begin{figure*}[!htb]
		\centering

		\subfigure{
			\begin{minipage}[t]{\linewidth}
				\centering
				\includegraphics[width=4in]{legend2.png}\\
			\end{minipage}%
		}%
		
		\subfigure{
			\begin{minipage}[t]{0.166\linewidth}
				\centering
				\includegraphics[width=1.05in]{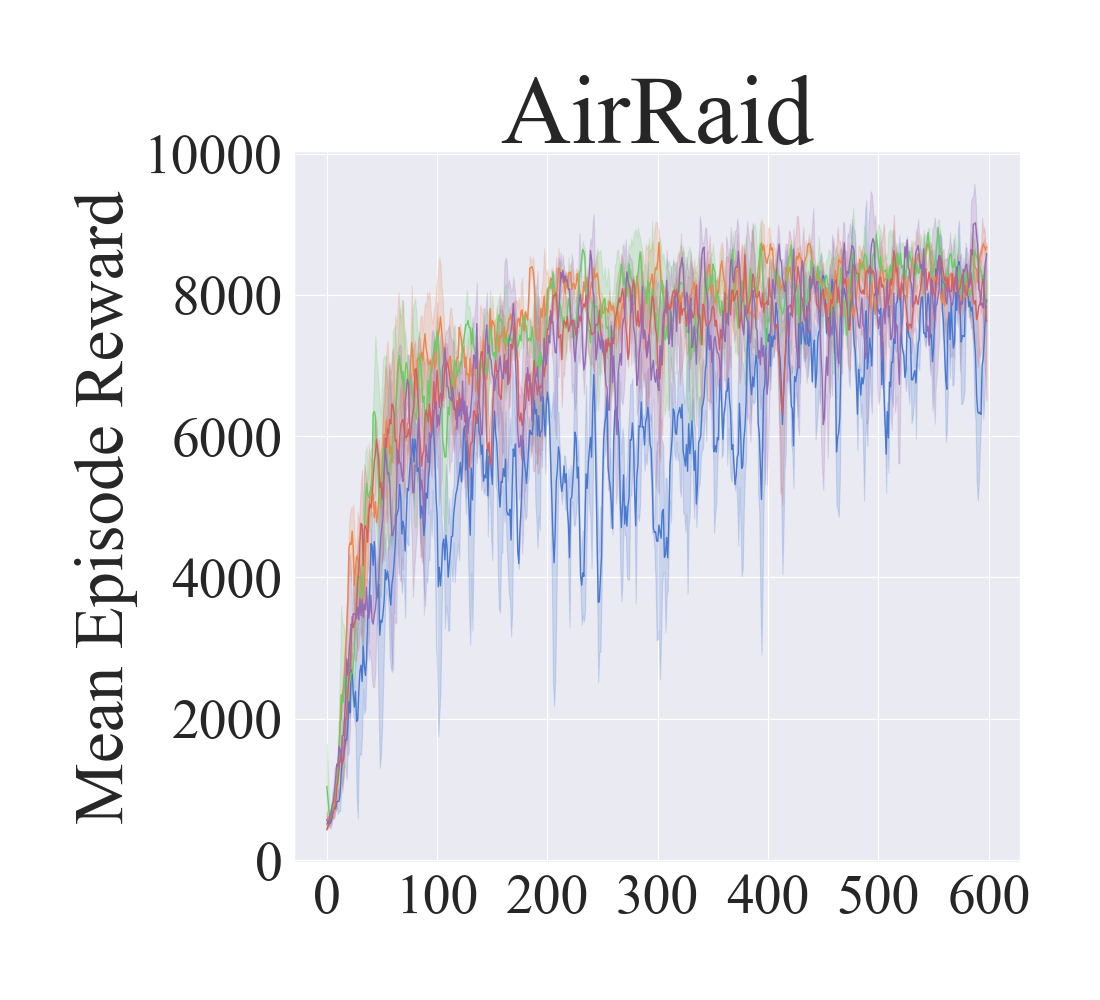}\\
			\end{minipage}%
		}%
		\subfigure{
			\begin{minipage}[t]{0.166\linewidth}
				\centering
				\includegraphics[width=1.05in]{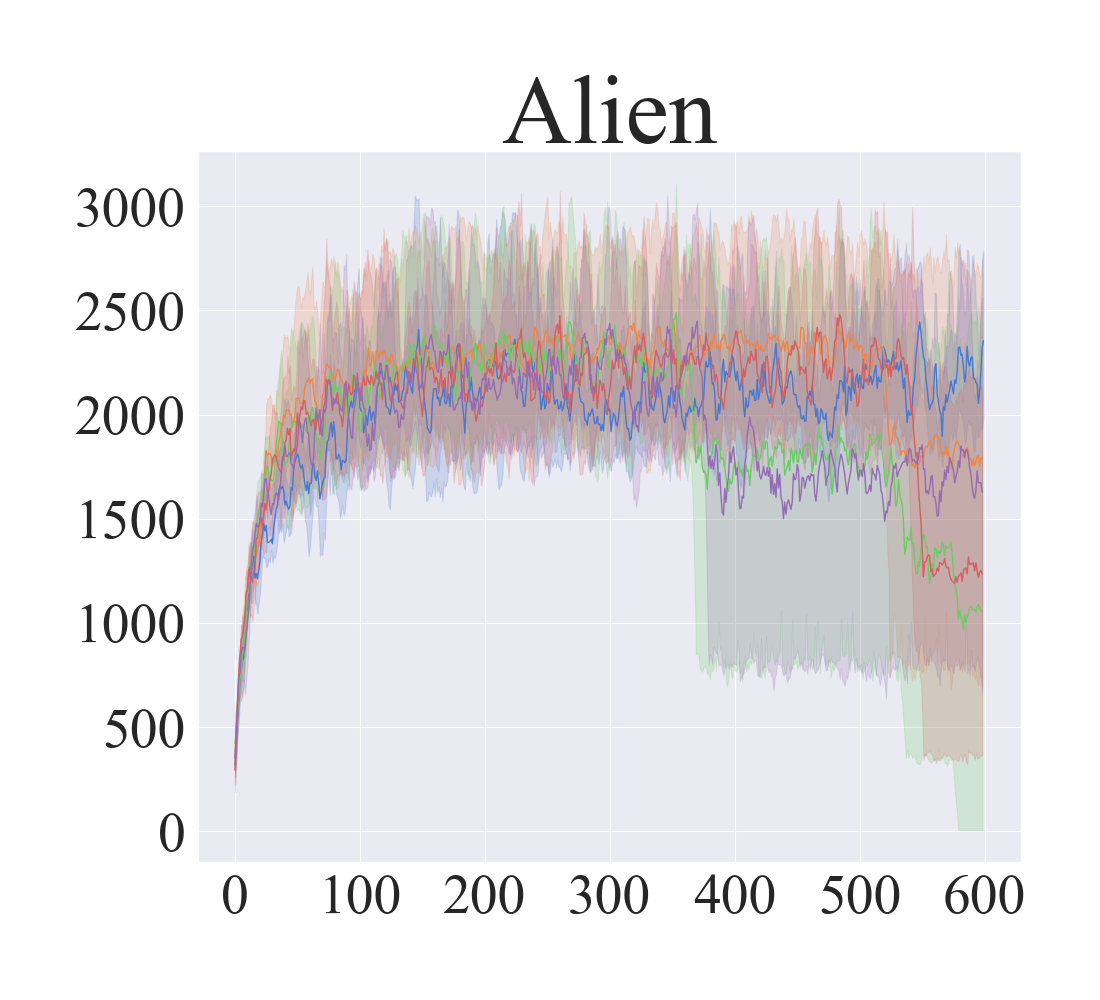}\\
			\end{minipage}%
		}%
		\subfigure{
			\begin{minipage}[t]{0.166\linewidth}
				\centering
				\includegraphics[width=1.05in]{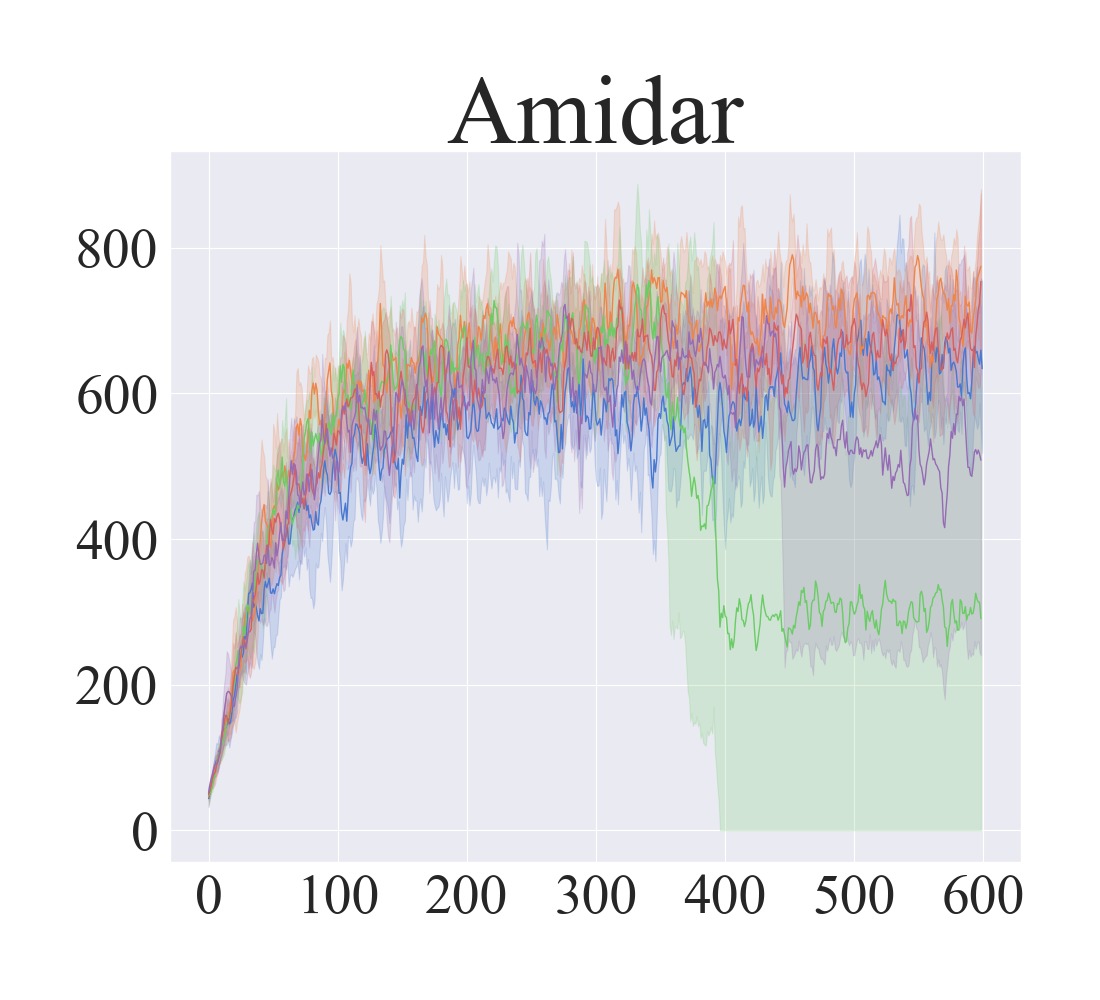}\\
			\end{minipage}%
		}%
		\subfigure{
			\begin{minipage}[t]{0.166\linewidth}
				\centering
				\includegraphics[width=1.05in]{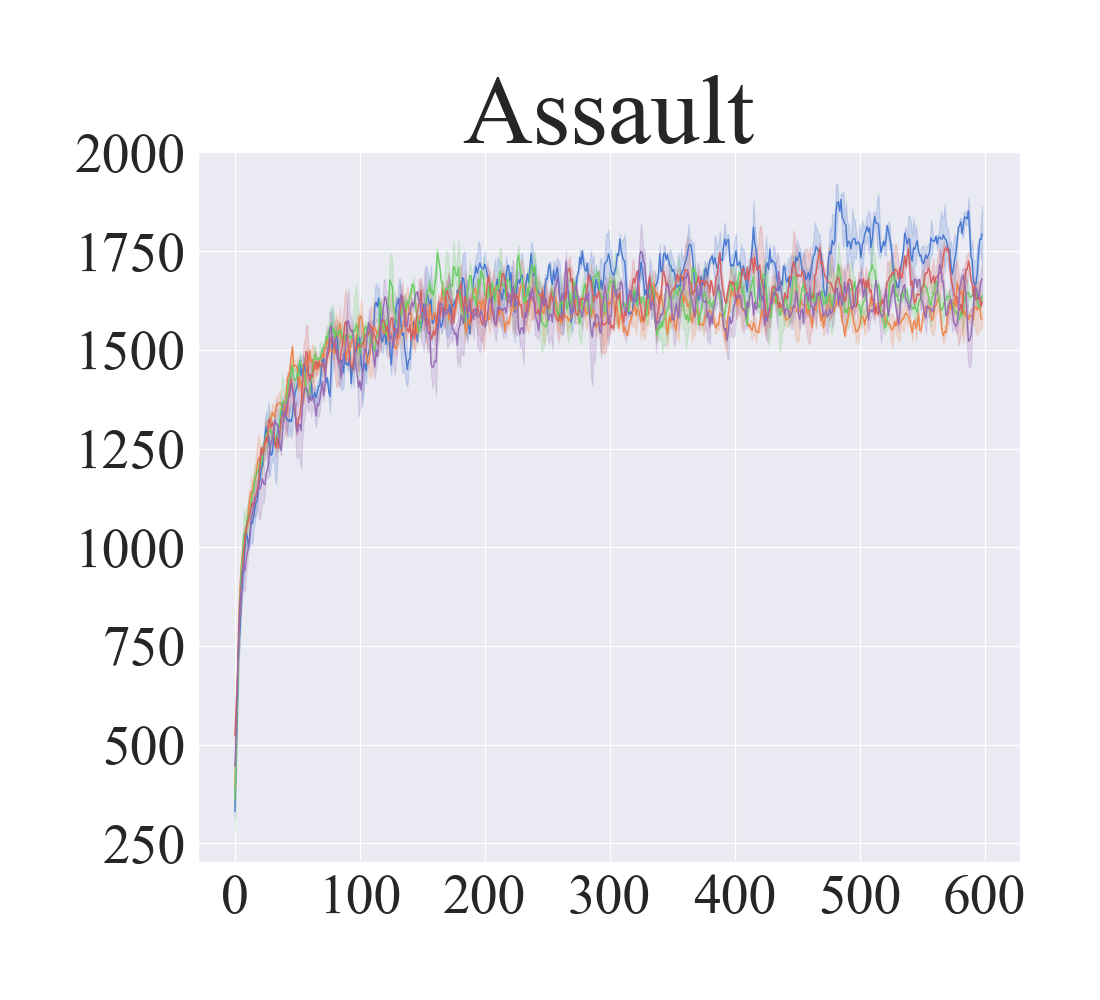}\\
			\end{minipage}%
		}%
		\subfigure{
			\begin{minipage}[t]{0.166\linewidth}
				\centering
				\includegraphics[width=1.05in]{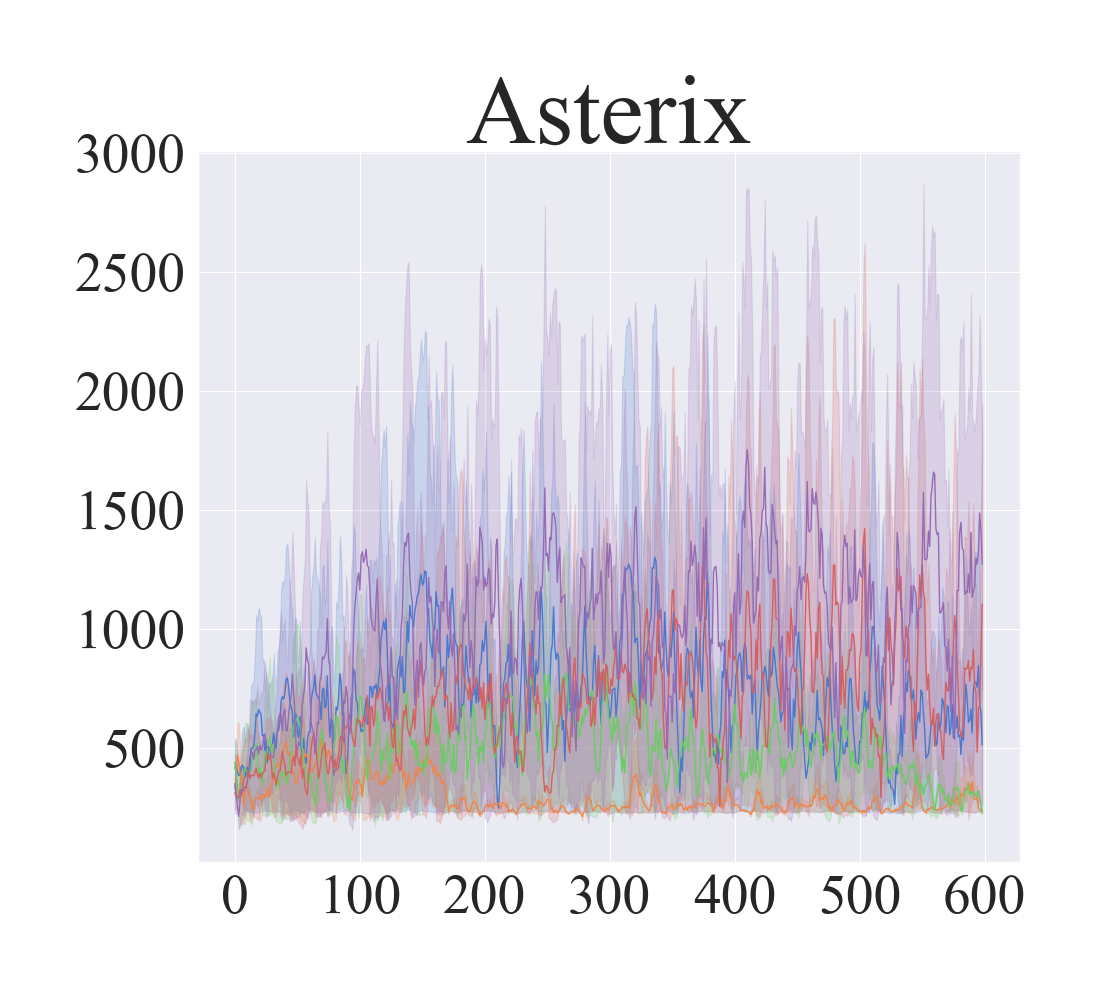}\\
			\end{minipage}%
		}%
		\subfigure{
			\begin{minipage}[t]{0.166\linewidth}
				\centering
				\includegraphics[width=1.05in]{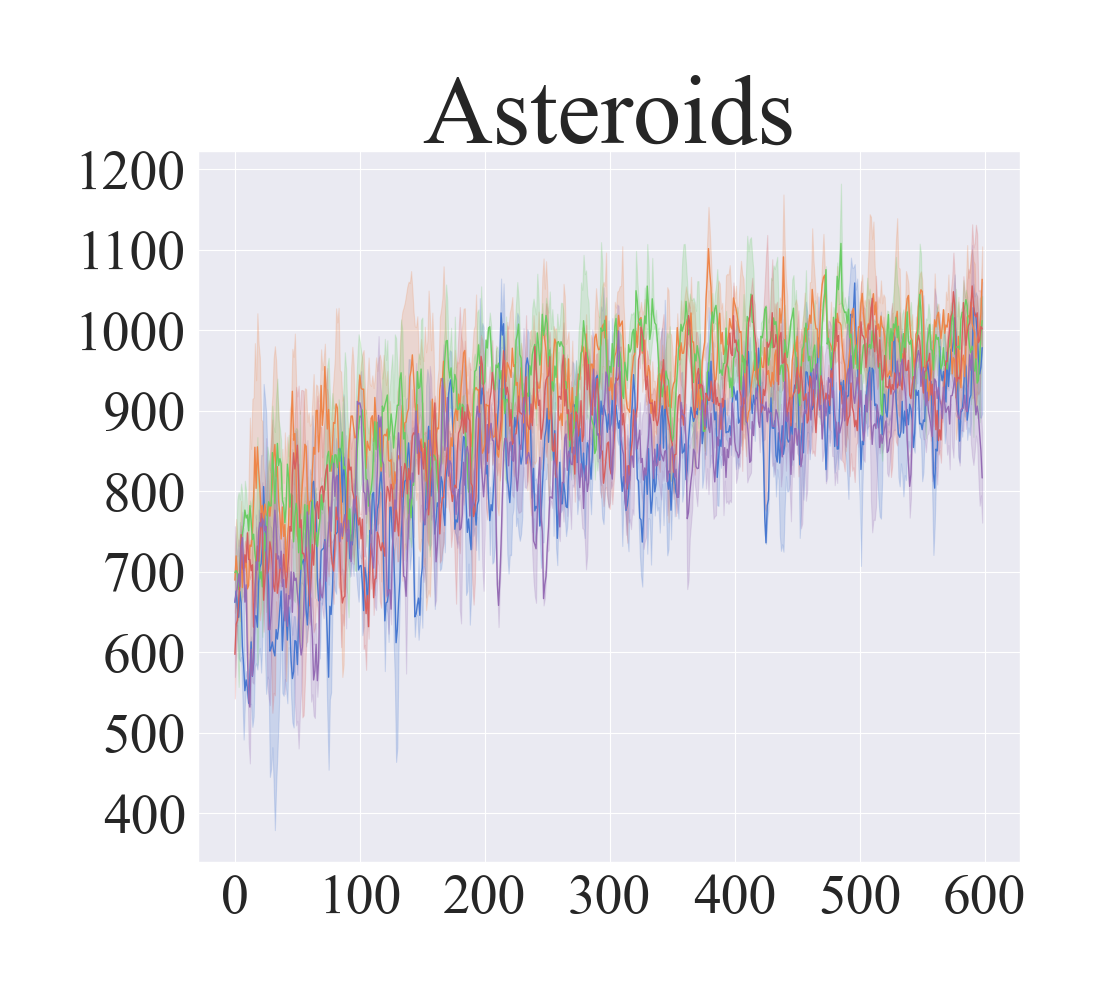}\\\
			\end{minipage}%
		}%
		\vspace{-1.0cm}
		
		\subfigure{
			\begin{minipage}[t]{0.166\linewidth}
				\centering
				\includegraphics[width=1.05in]{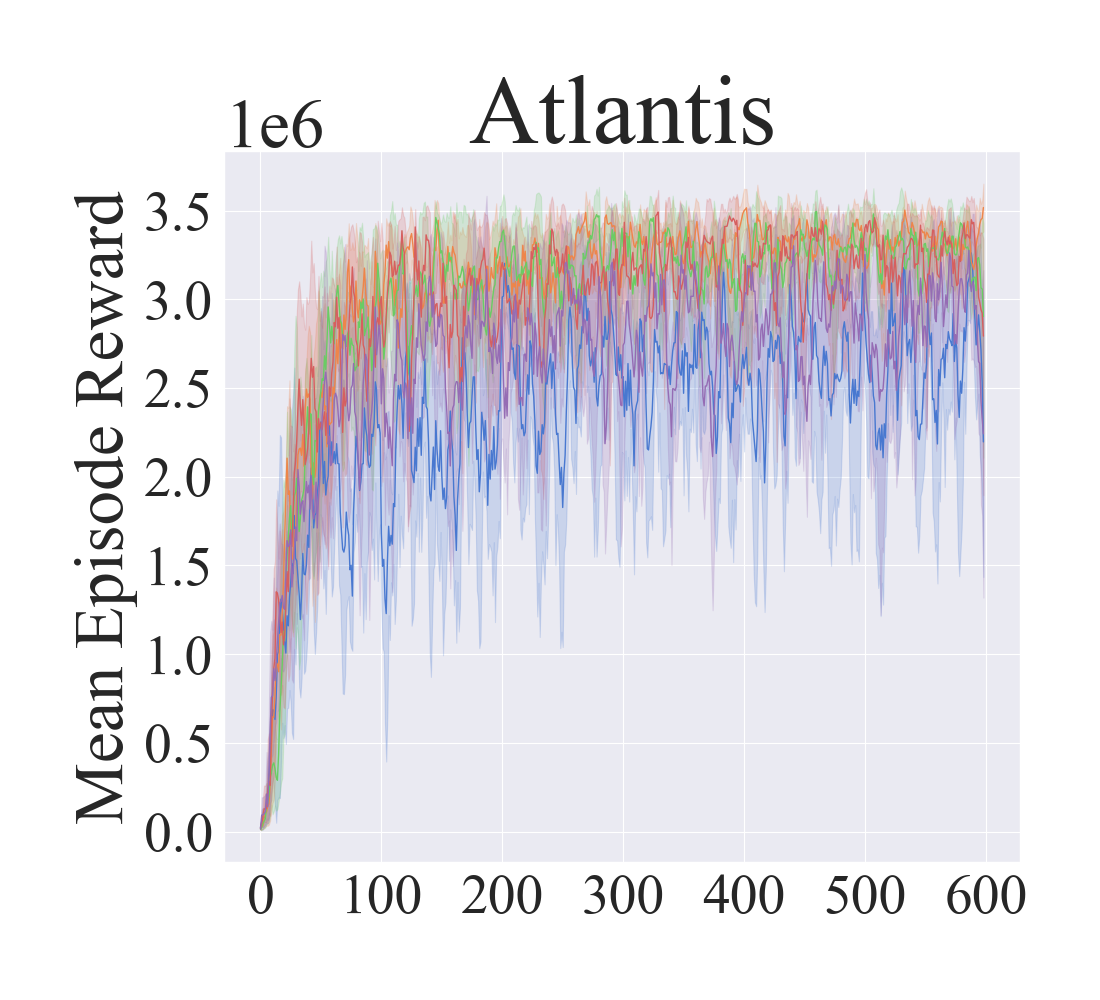}\\
			\end{minipage}%
		}%
		\subfigure{
			\begin{minipage}[t]{0.166\linewidth}
				\centering
				\includegraphics[width=1.05in]{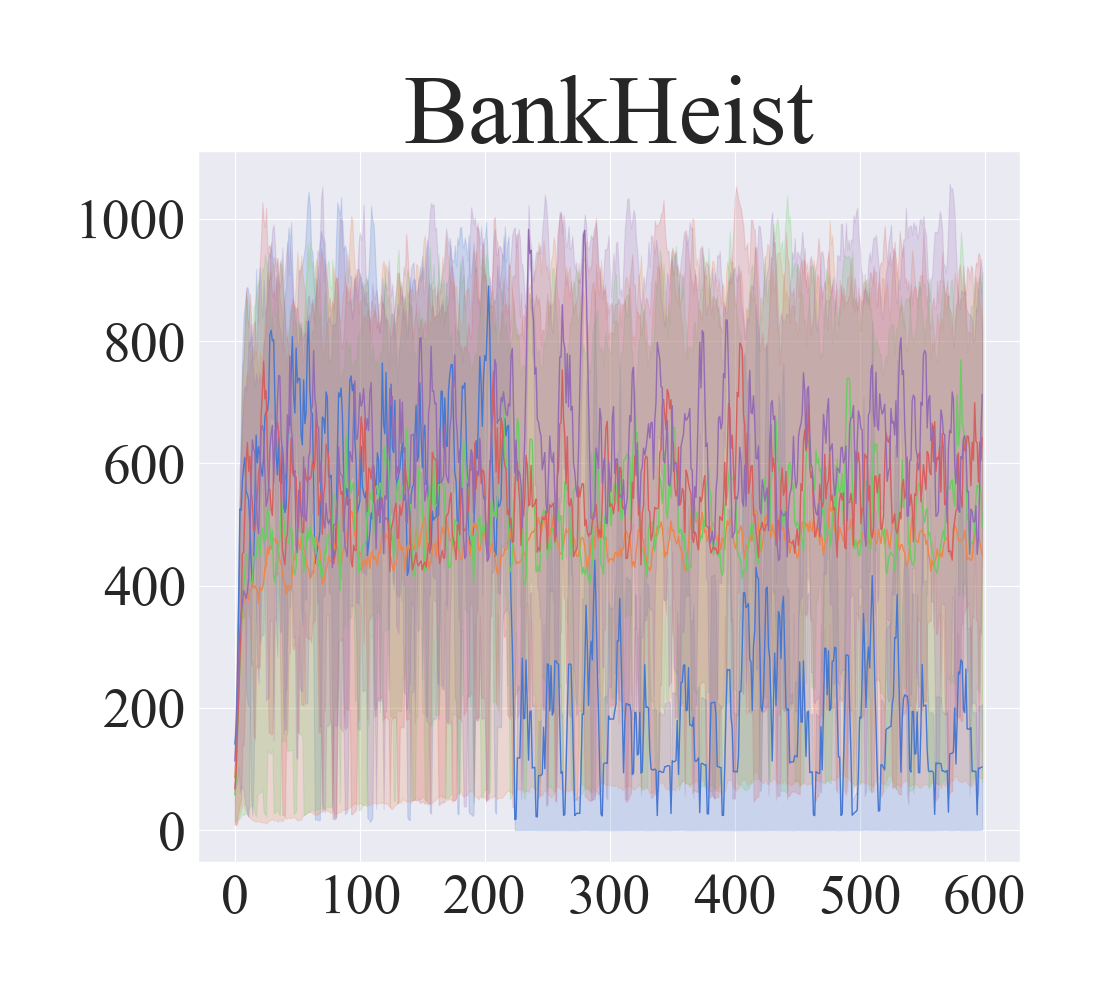}\\
			\end{minipage}%
		}%
		\subfigure{
			\begin{minipage}[t]{0.166\linewidth}
				\centering
				\includegraphics[width=1.05in]{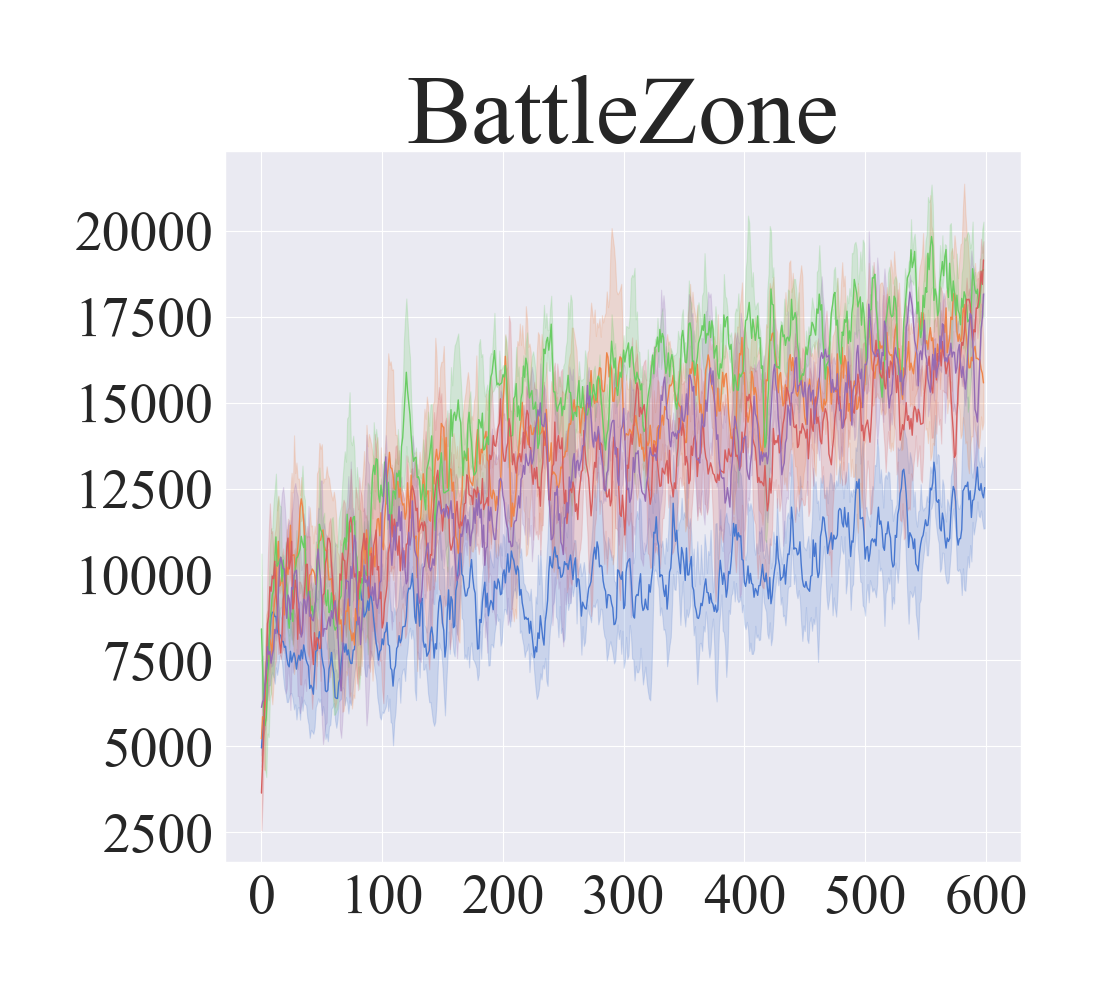}\\
			\end{minipage}%
		}%
		\subfigure{
			\begin{minipage}[t]{0.166\linewidth}
				\centering
				\includegraphics[width=1.05in]{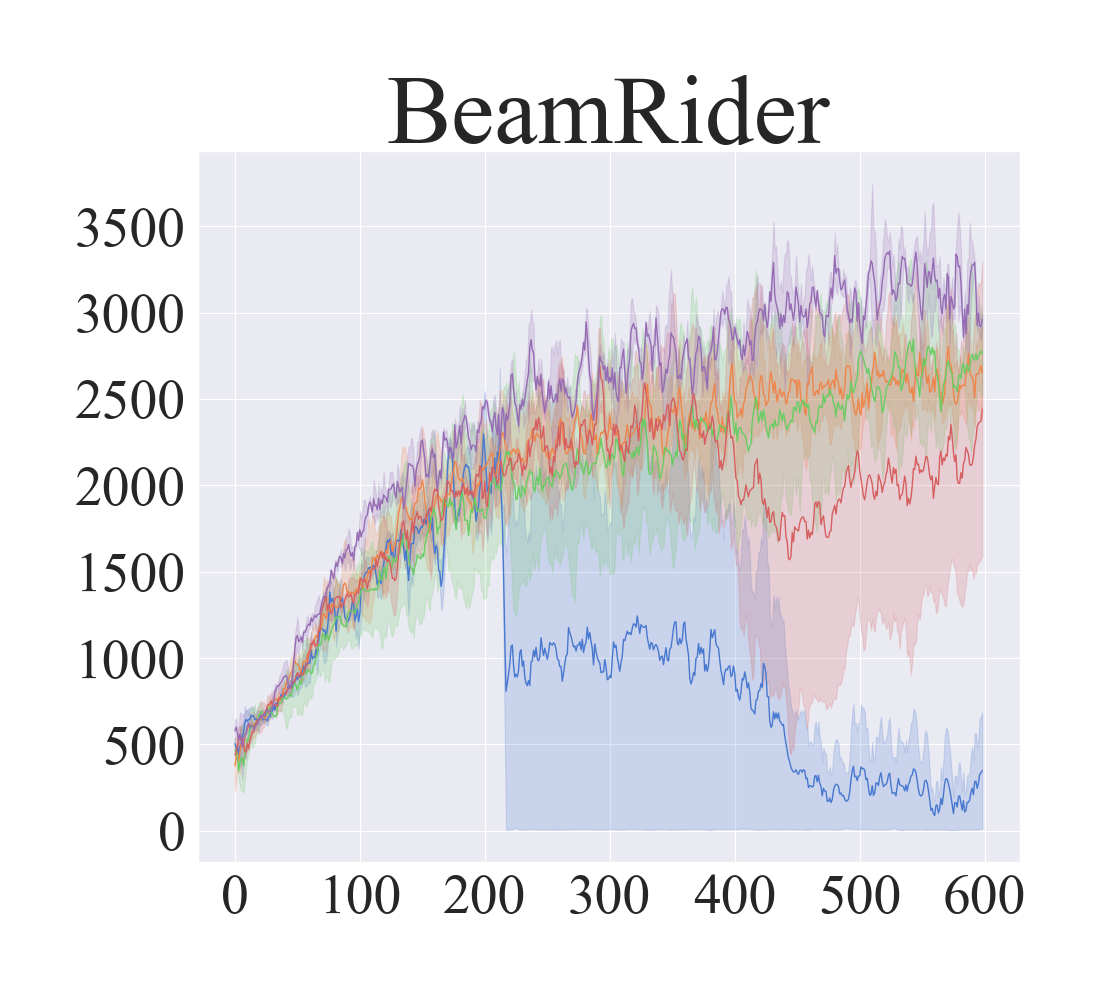}\\
			\end{minipage}%
		}%
		\subfigure{
			\begin{minipage}[t]{0.166\linewidth}
				\centering
				\includegraphics[width=1.05in]{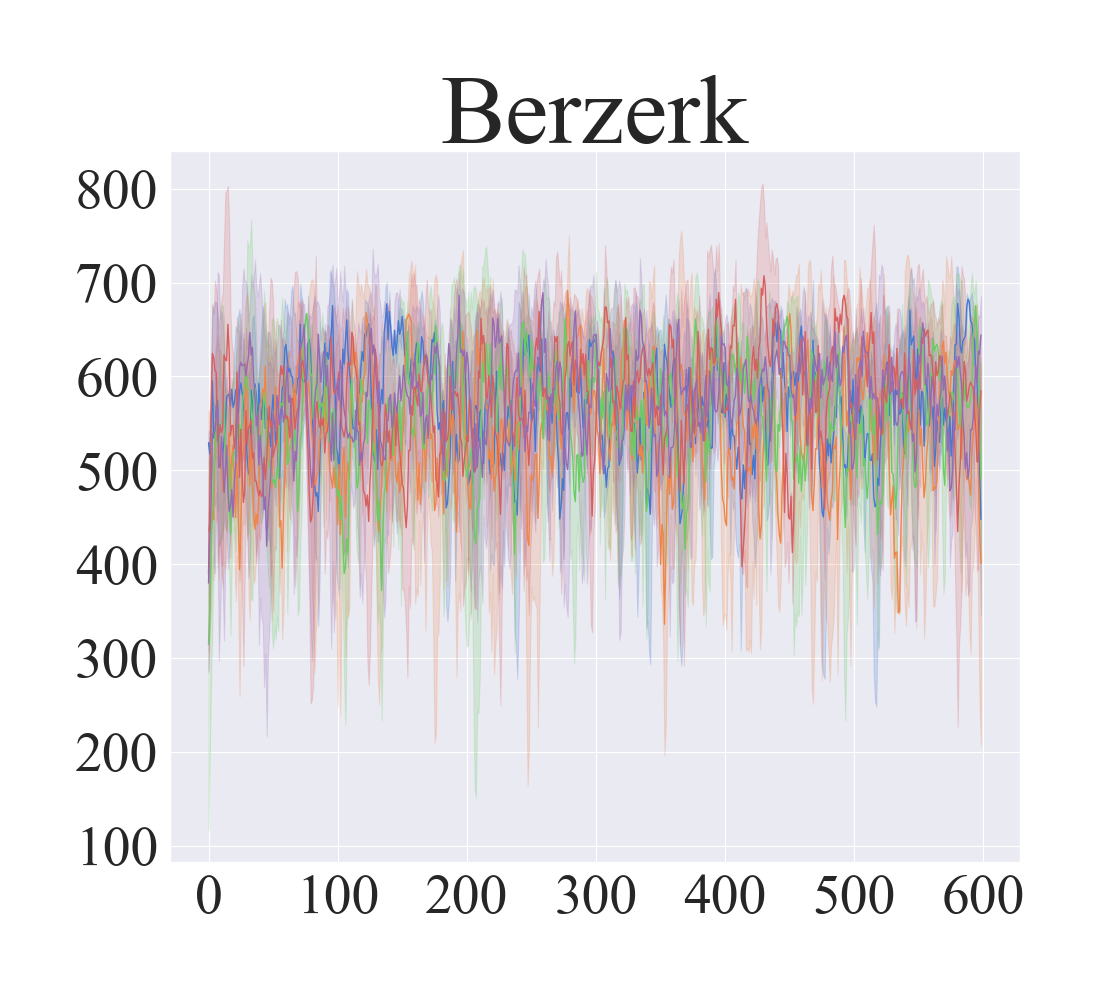}\\
			\end{minipage}%
		}%
		\subfigure{
			\begin{minipage}[t]{0.166\linewidth}
				\centering
				\includegraphics[width=1.05in]{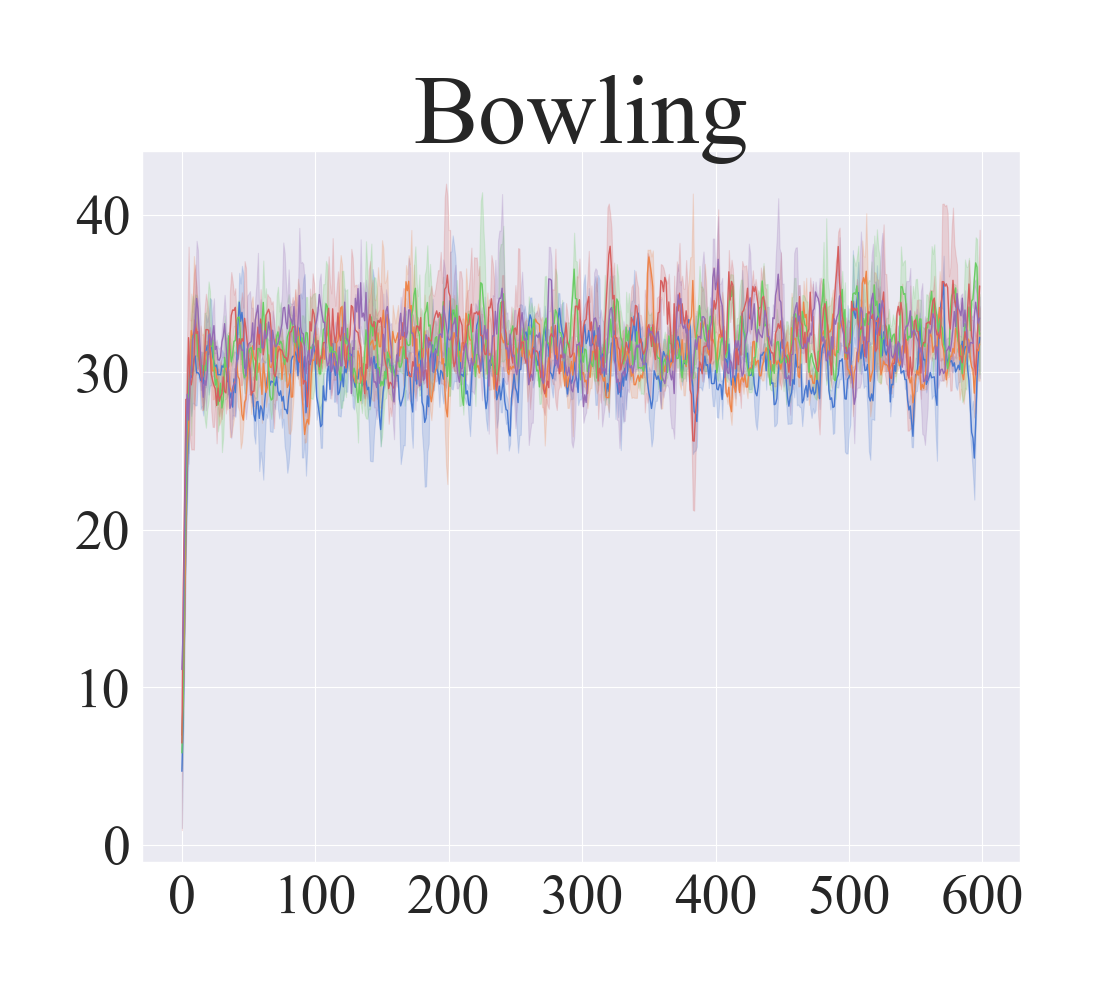}\\
			\end{minipage}%
		}%
		\vspace{-0.6cm}
		
		\subfigure{
			\begin{minipage}[t]{0.166\linewidth}
				\centering
				\includegraphics[width=1.05in]{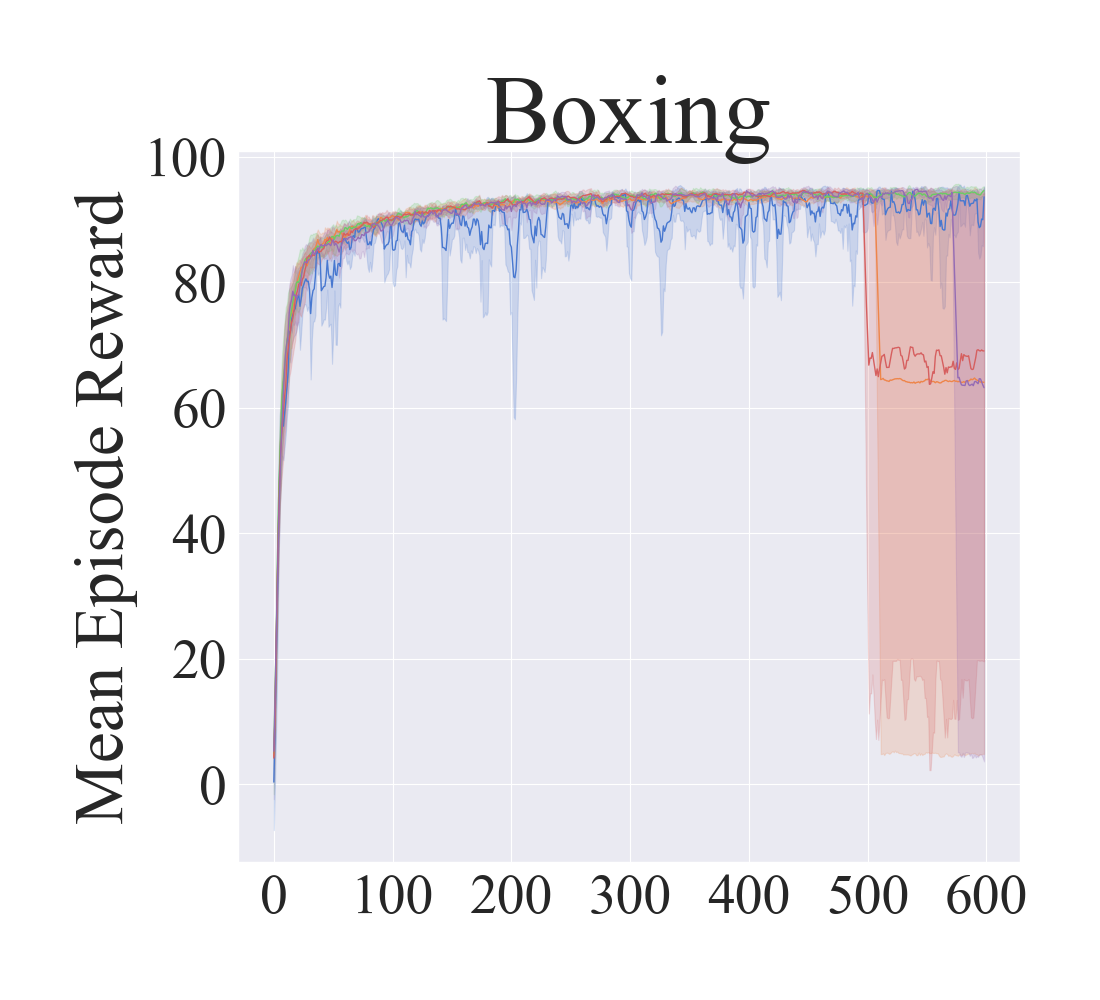}\\
			\end{minipage}%
		}%
		\subfigure{
			\begin{minipage}[t]{0.166\linewidth}
				\centering
				\includegraphics[width=1.05in]{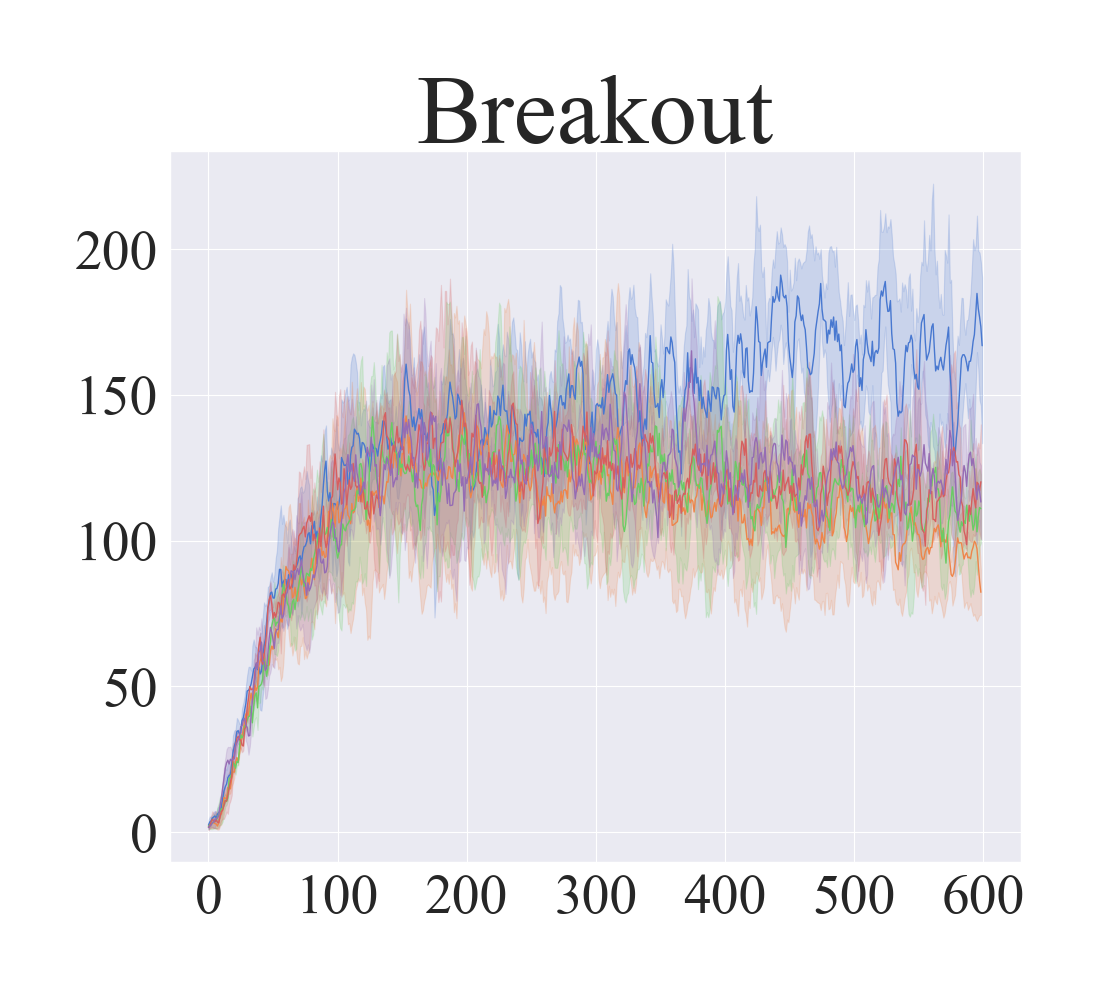}\\
			\end{minipage}%
		}%
		\subfigure{
			\begin{minipage}[t]{0.166\linewidth}
				\centering
				\includegraphics[width=1.05in]{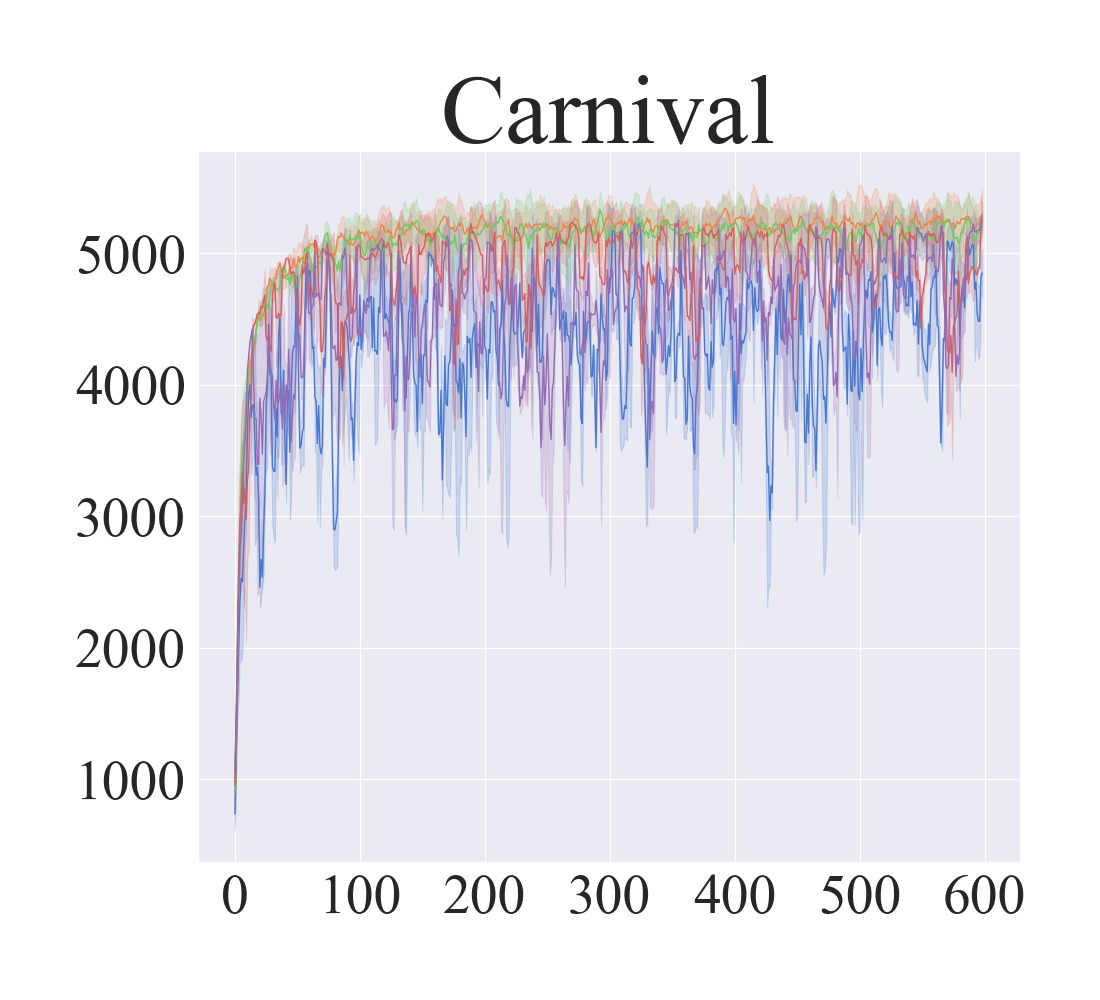}\\
			\end{minipage}%
		}%
		\subfigure{
			\begin{minipage}[t]{0.166\linewidth}
				\centering
				\includegraphics[width=1.05in]{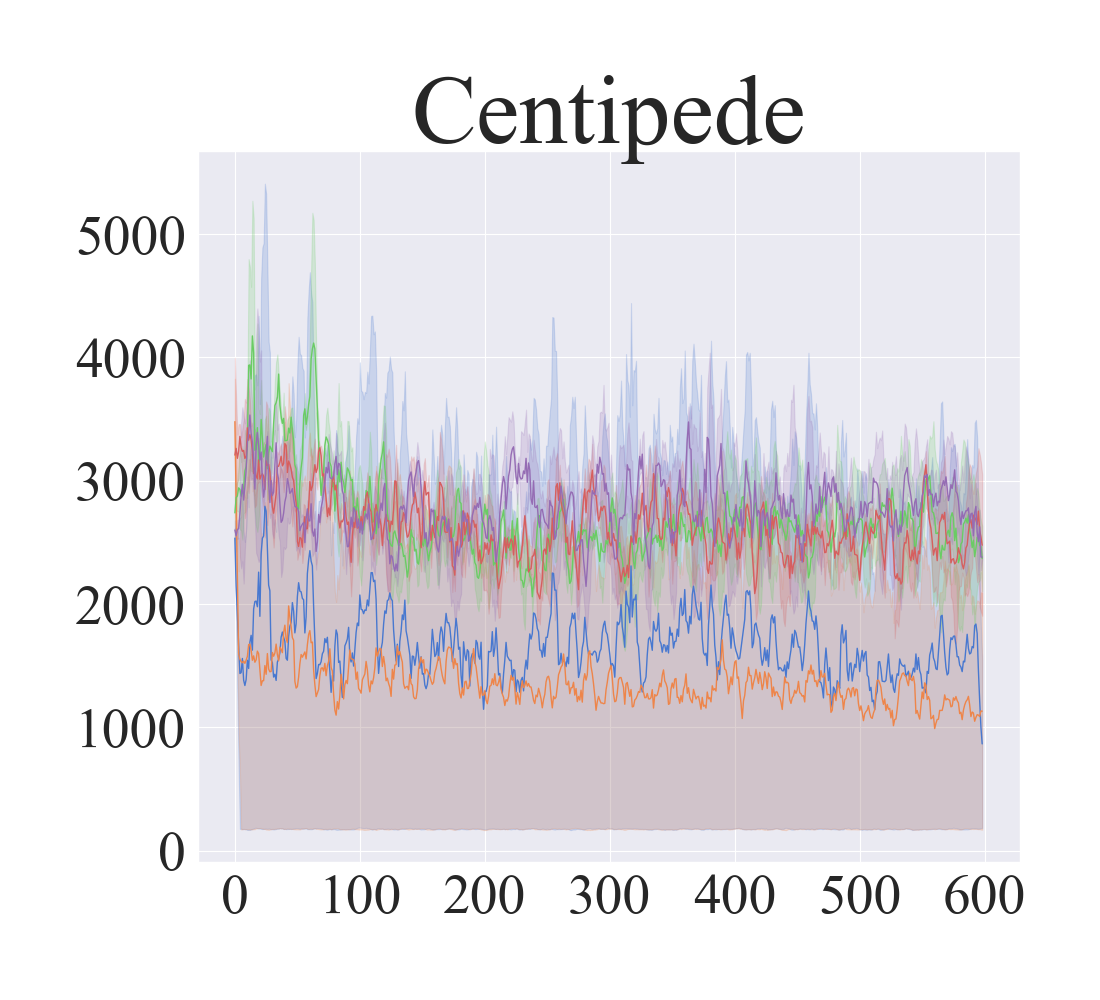}\\
			\end{minipage}%
		}%
		\subfigure{
			\begin{minipage}[t]{0.166\linewidth}
				\centering
				\includegraphics[width=1.05in]{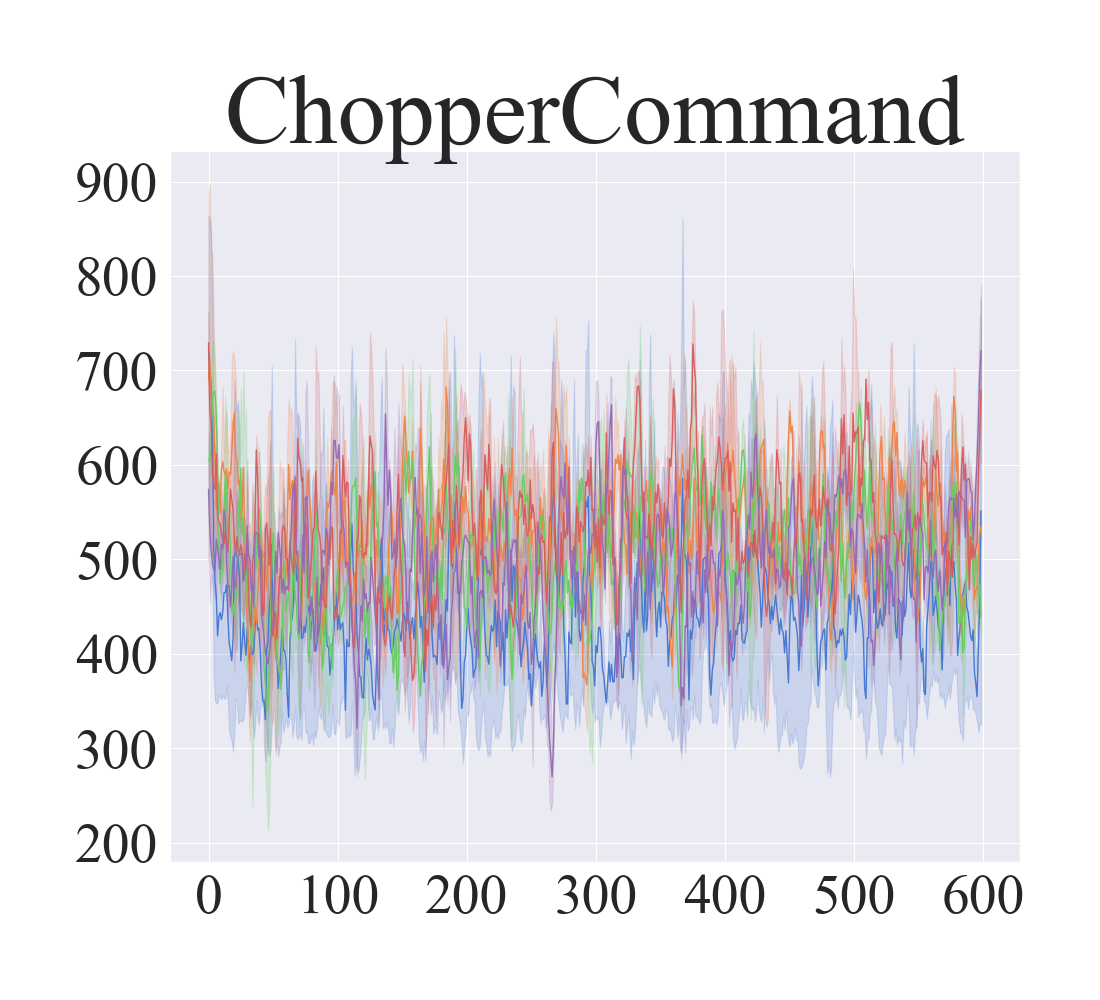}\\
			\end{minipage}%
		}%
		\subfigure{
			\begin{minipage}[t]{0.166\linewidth}
				\centering
				\includegraphics[width=1.05in]{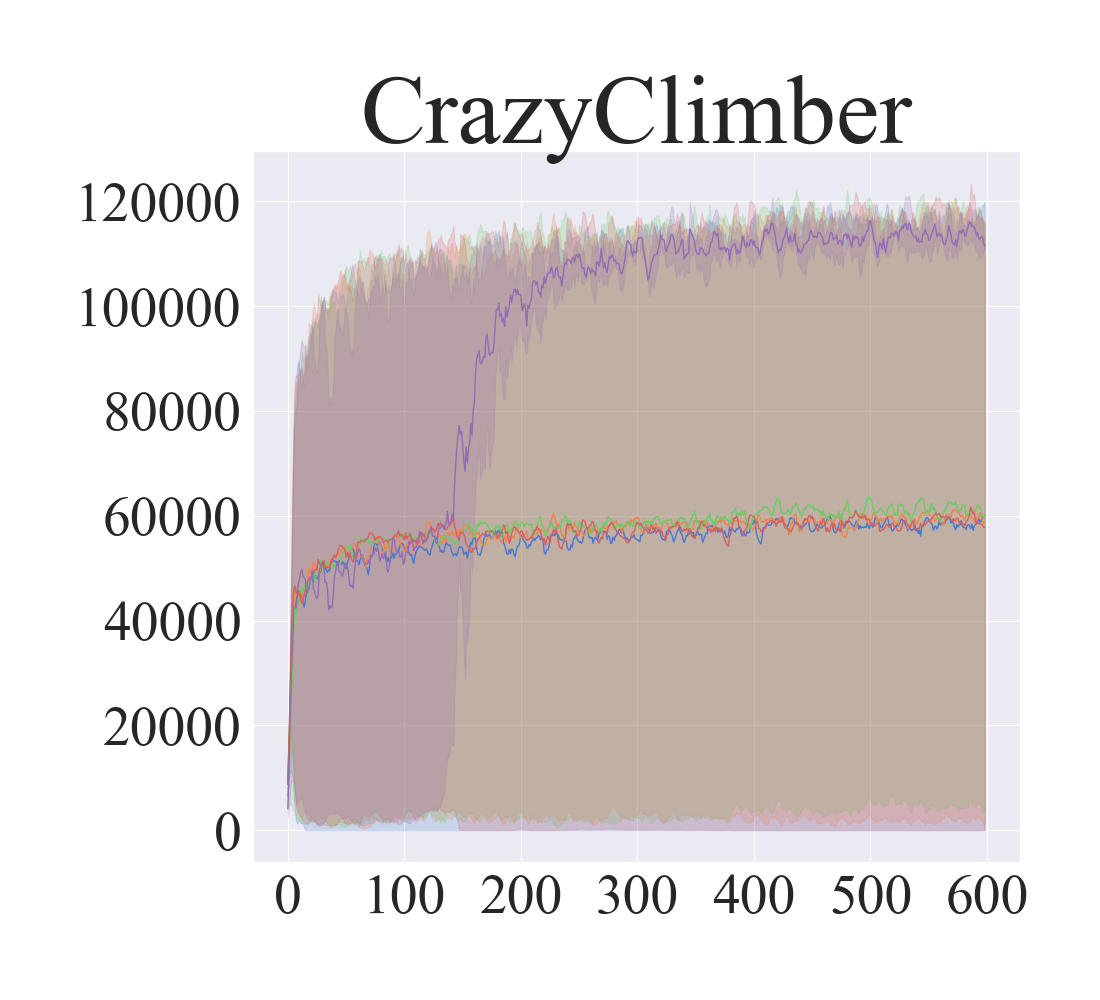}\\
			\end{minipage}%
		}%
		\vspace{-0.6cm}
		
		\subfigure{
			\begin{minipage}[t]{0.166\linewidth}
				\centering
				\includegraphics[width=1.05in]{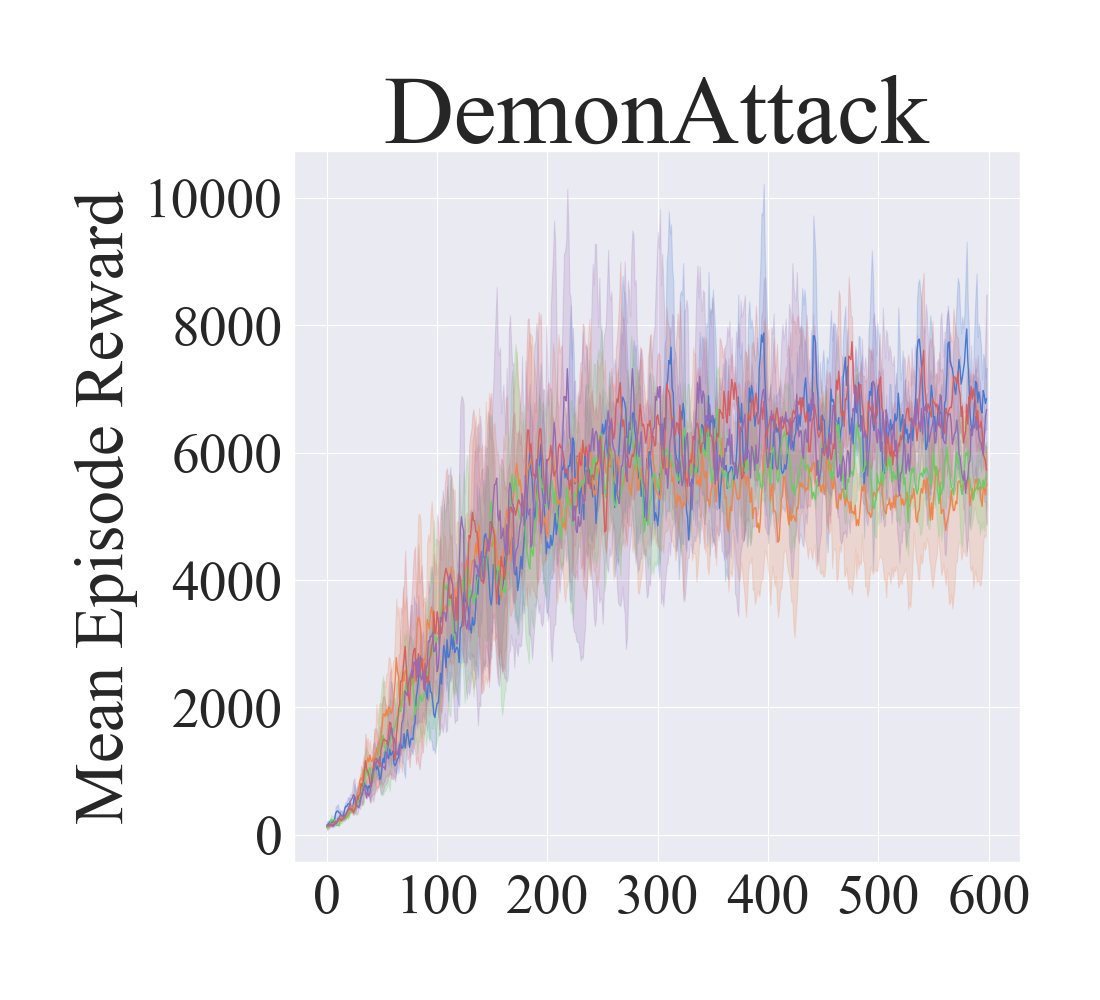}\\
			\end{minipage}%
		}%
		\subfigure{
			\begin{minipage}[t]{0.166\linewidth}
				\centering
				\includegraphics[width=1.05in]{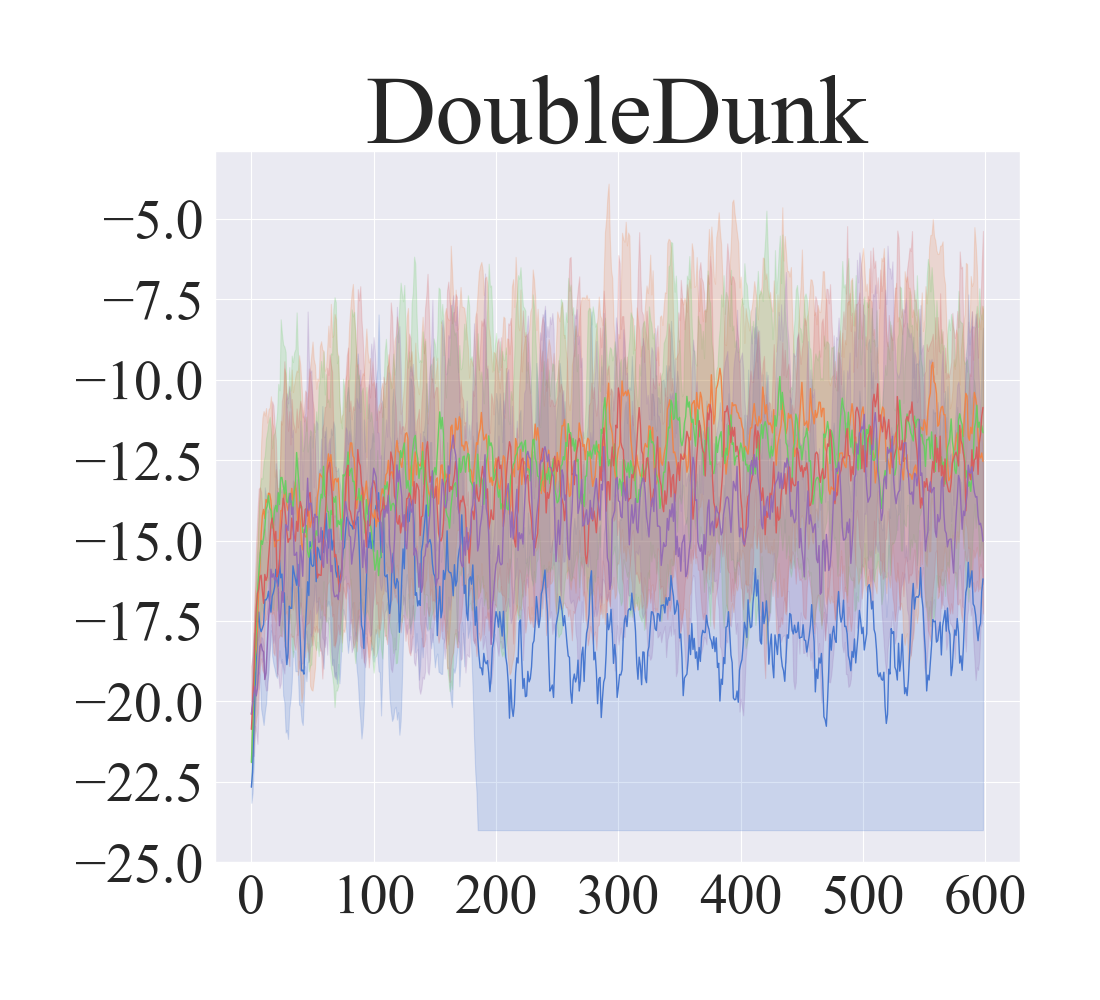}\\
			\end{minipage}%
		}%
		\subfigure{
			\begin{minipage}[t]{0.166\linewidth}
				\centering
				\includegraphics[width=1.05in]{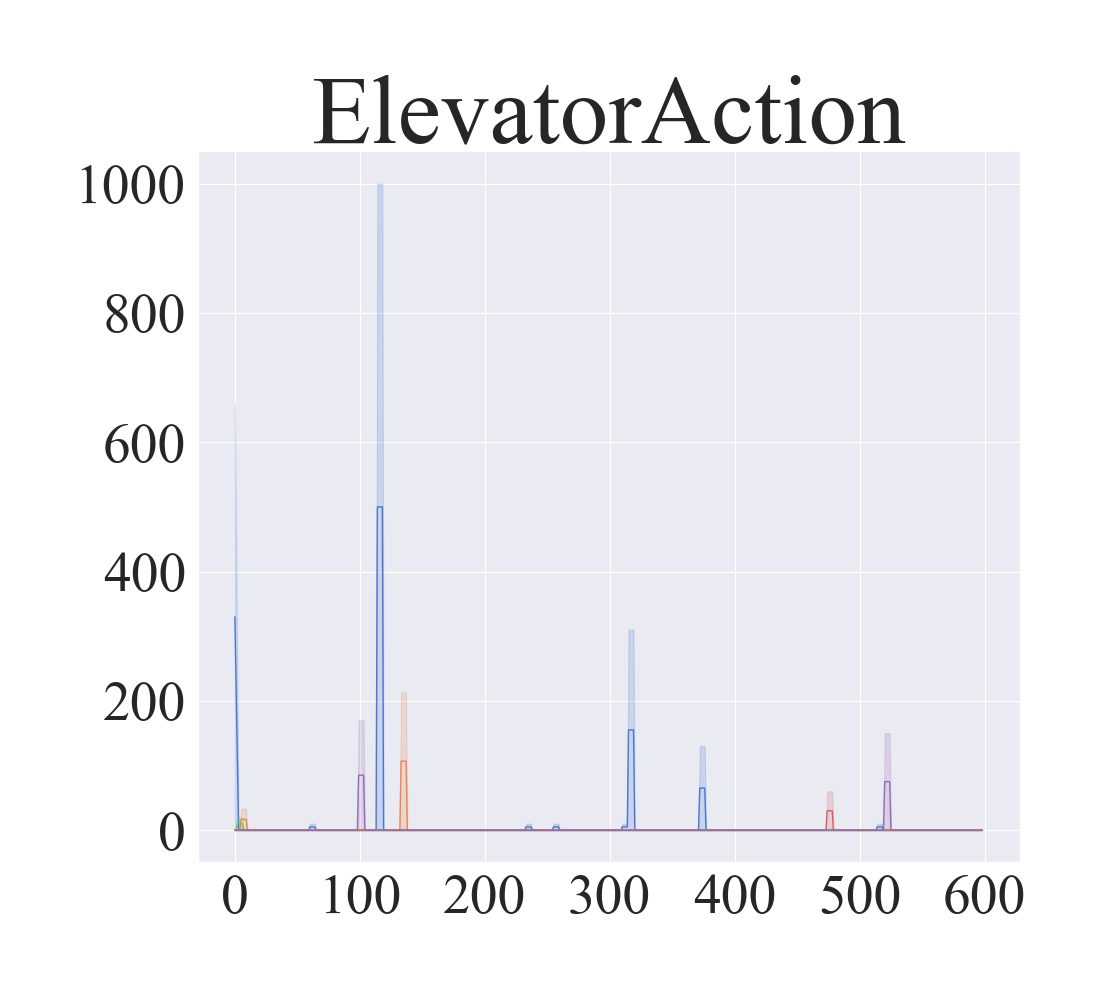}\\
			\end{minipage}%
		}%
		\subfigure{
			\begin{minipage}[t]{0.166\linewidth}
				\centering
				\includegraphics[width=1.05in]{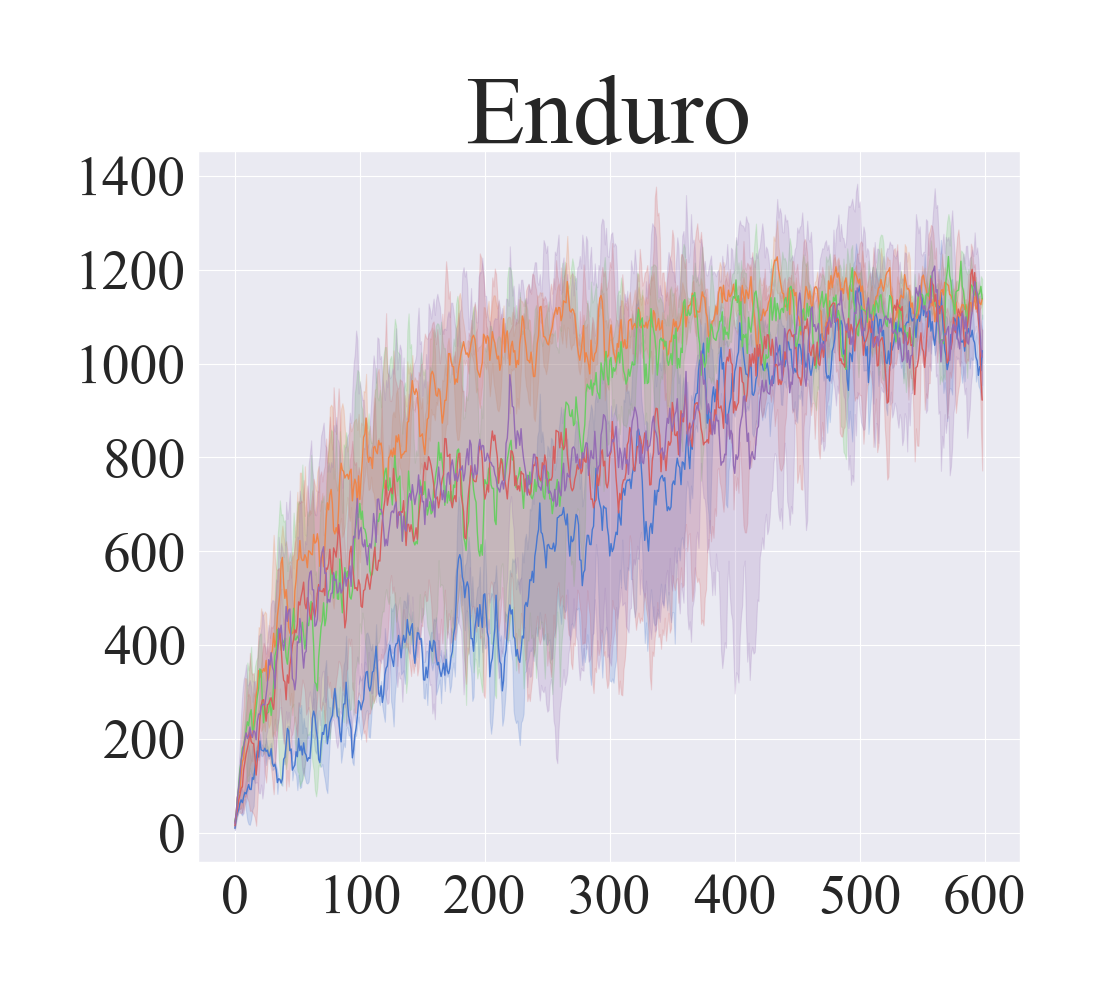}\\
			\end{minipage}%
		}%
		\subfigure{
			\begin{minipage}[t]{0.166\linewidth}
				\centering
				\includegraphics[width=1.05in]{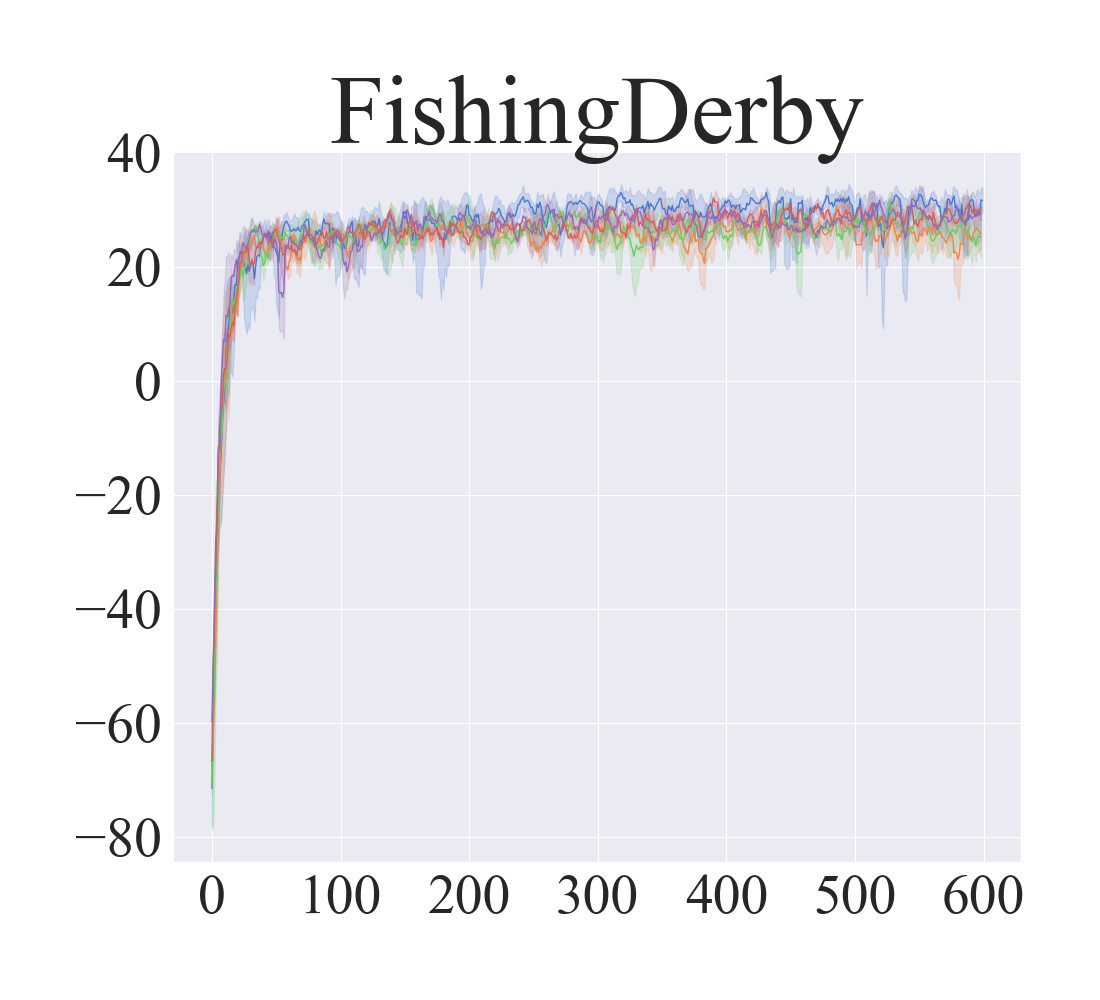}\\
			\end{minipage}%
		}%
		\subfigure{
			\begin{minipage}[t]{0.166\linewidth}
				\centering
				\includegraphics[width=1.05in]{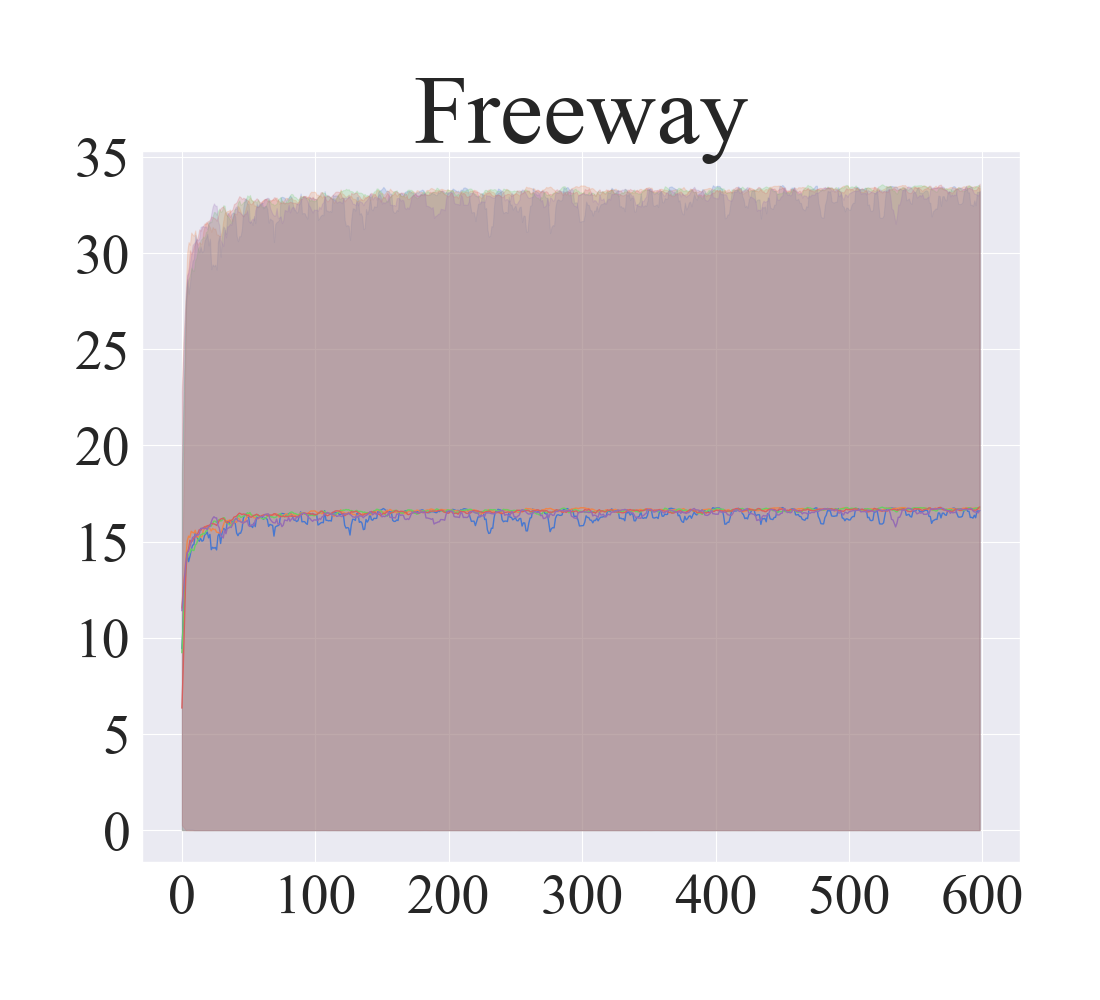}\\
			\end{minipage}%
		}%
		\vspace{-0.6cm}
		
		\subfigure{
			\begin{minipage}[t]{0.166\linewidth}
				\centering
				\includegraphics[width=1.05in]{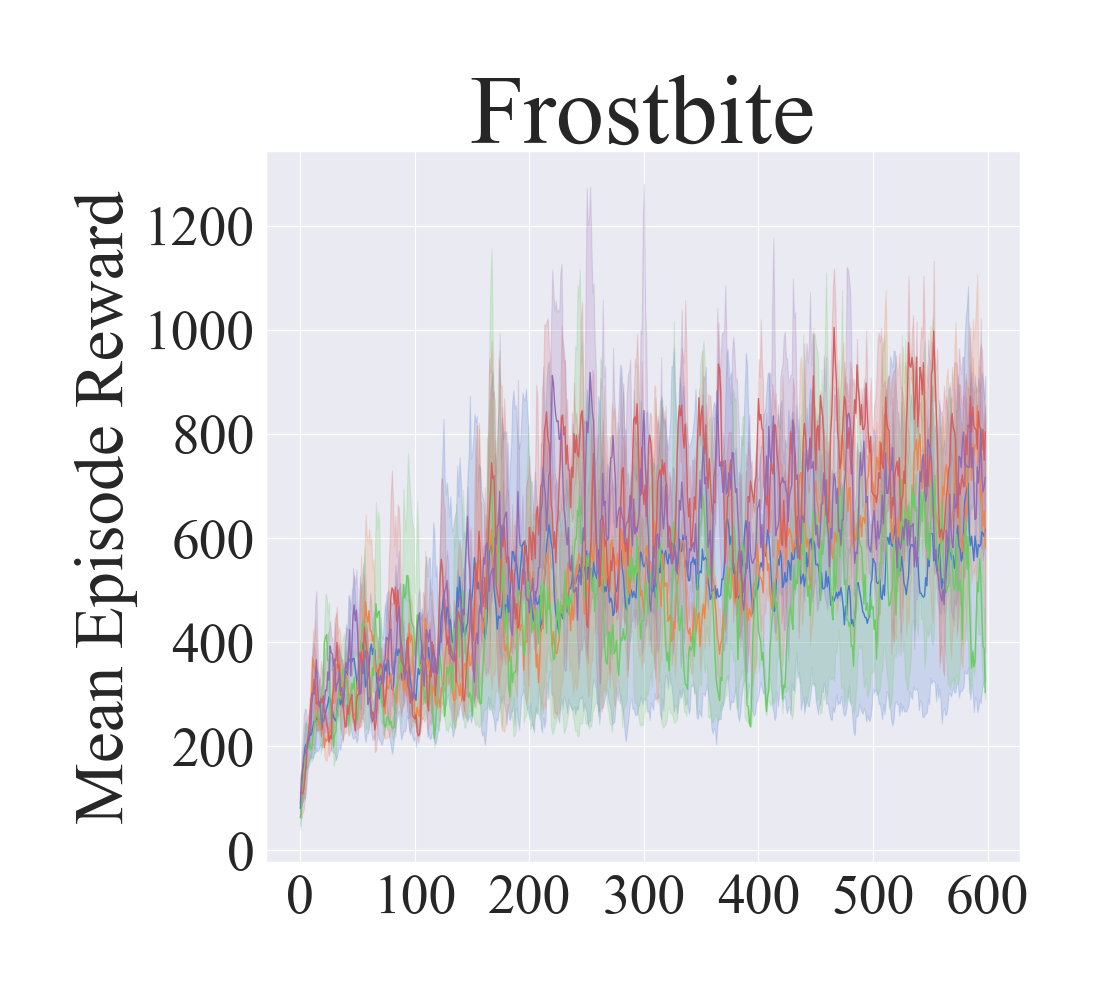}\\
			\end{minipage}%
		}%
		\subfigure{
			\begin{minipage}[t]{0.166\linewidth}
				\centering
				\includegraphics[width=1.05in]{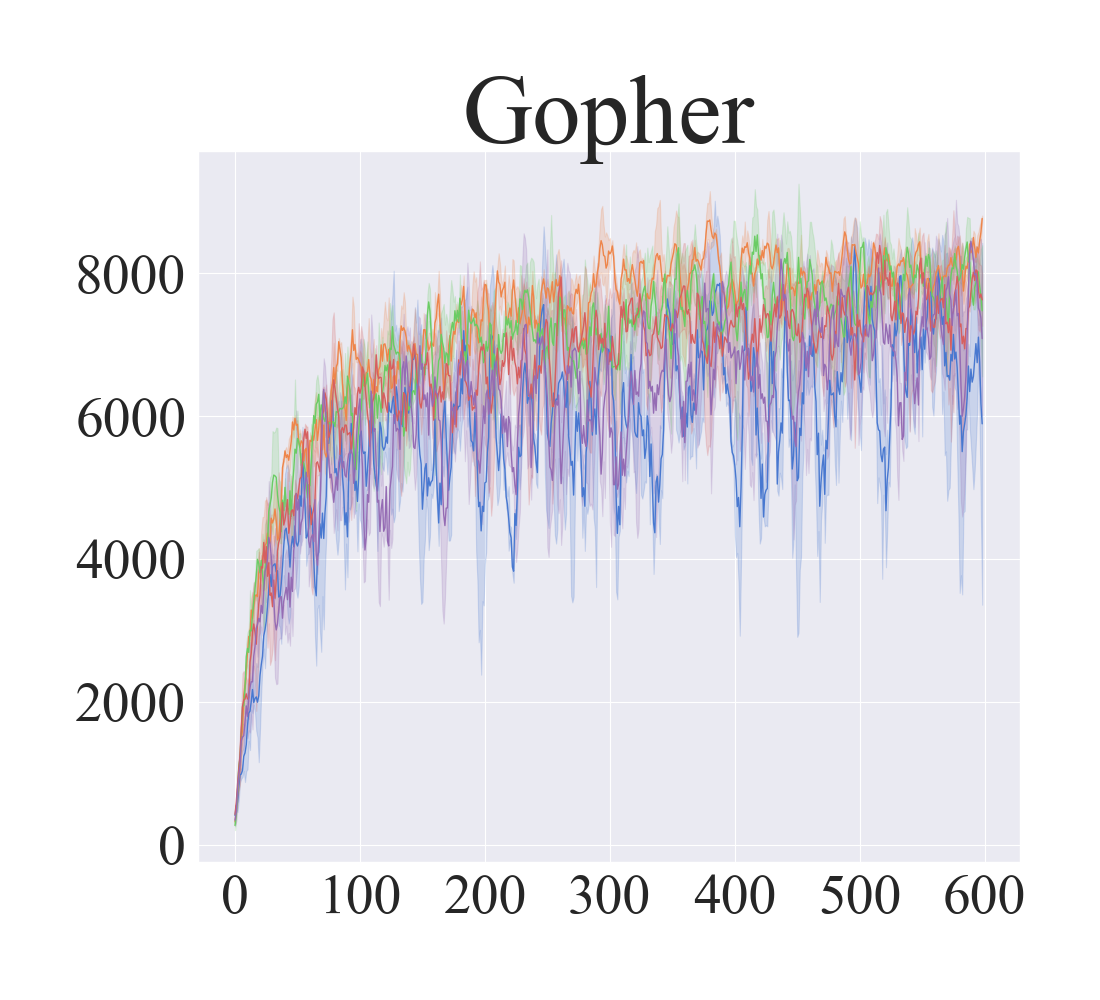}\\
			\end{minipage}%
		}%
		\subfigure{
			\begin{minipage}[t]{0.166\linewidth}
				\centering
				\includegraphics[width=1.05in]{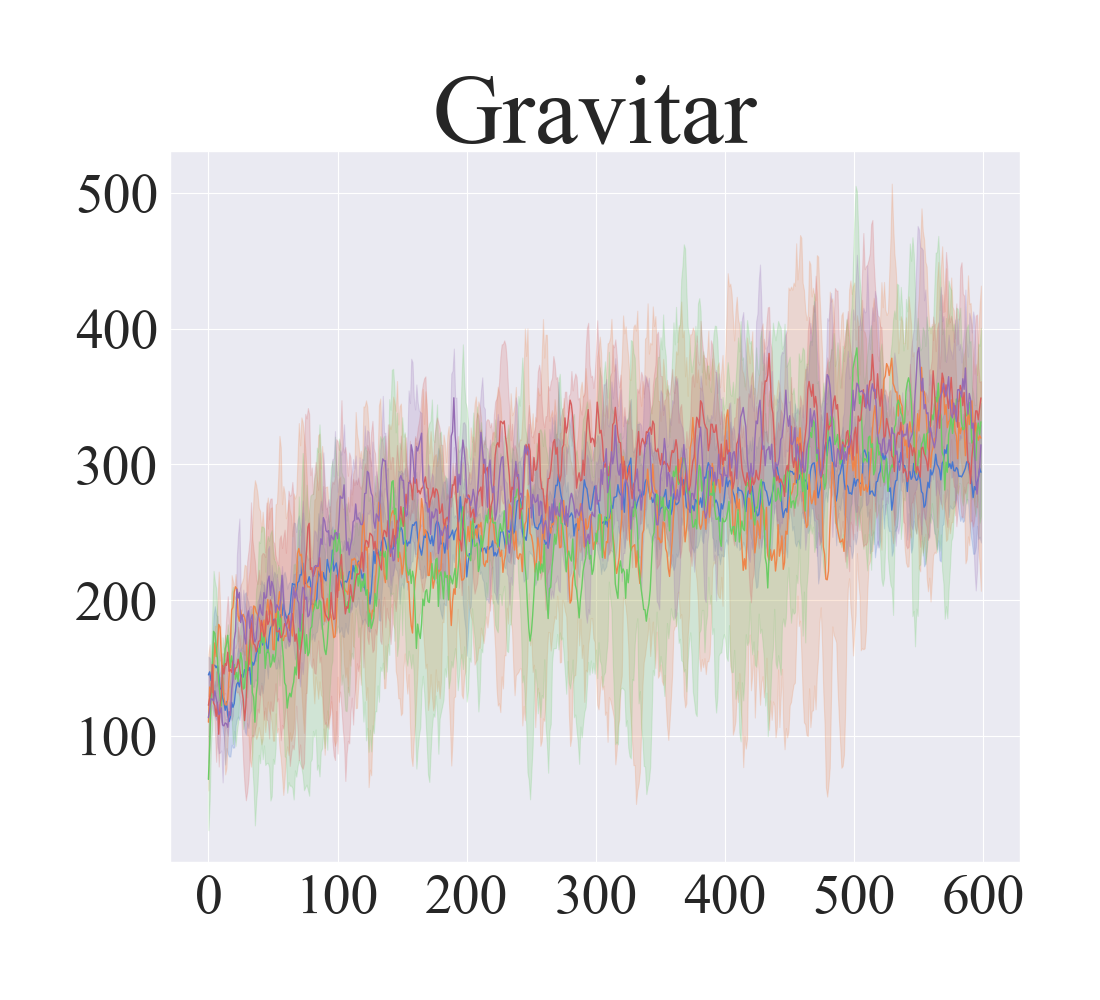}\\
			\end{minipage}%
		}%
		\subfigure{
			\begin{minipage}[t]{0.166\linewidth}
				\centering
				\includegraphics[width=1.05in]{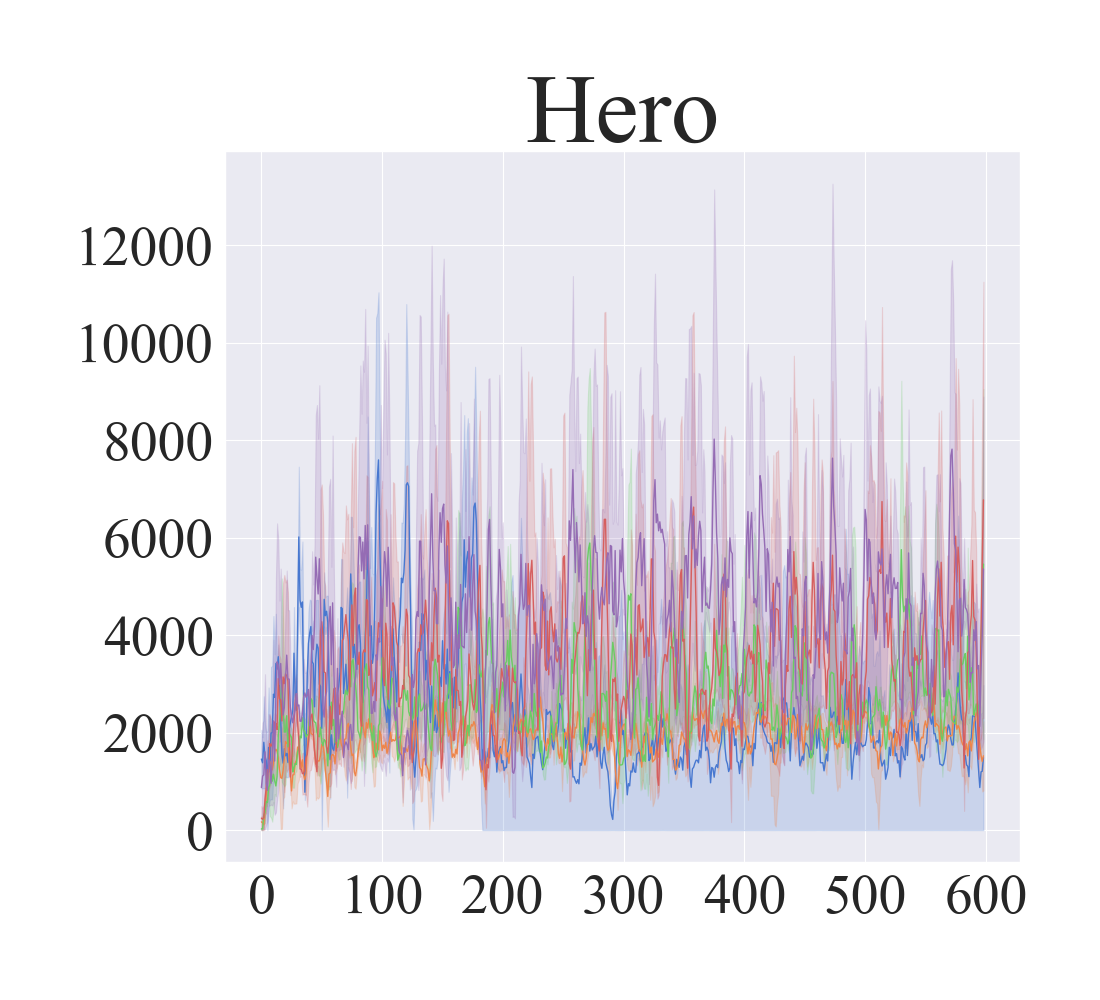}\\
			\end{minipage}%
		}%
		\subfigure{
			\begin{minipage}[t]{0.166\linewidth}
				\centering
				\includegraphics[width=1.05in]{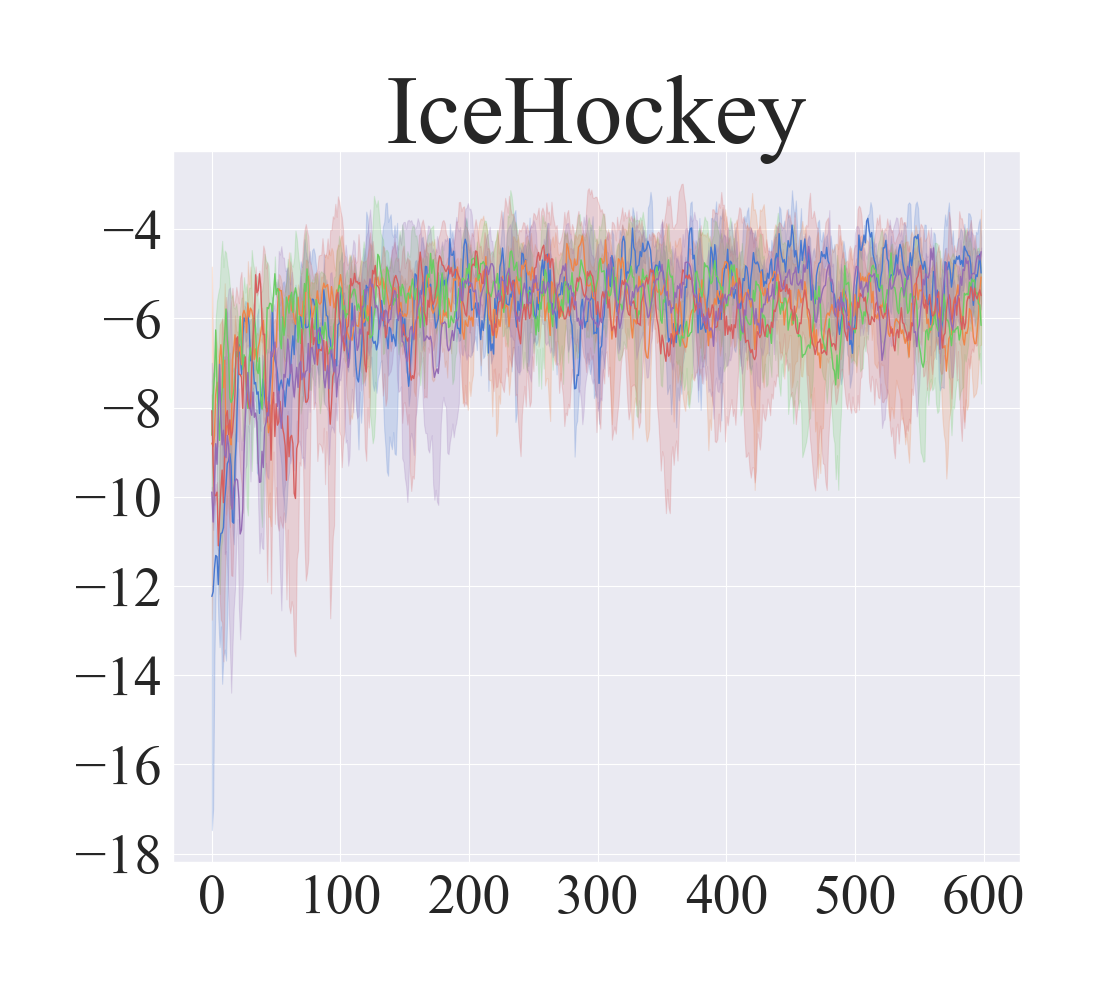}\\
			\end{minipage}%
		}%
		\subfigure{
			\begin{minipage}[t]{0.166\linewidth}
				\centering
				\includegraphics[width=1.05in]{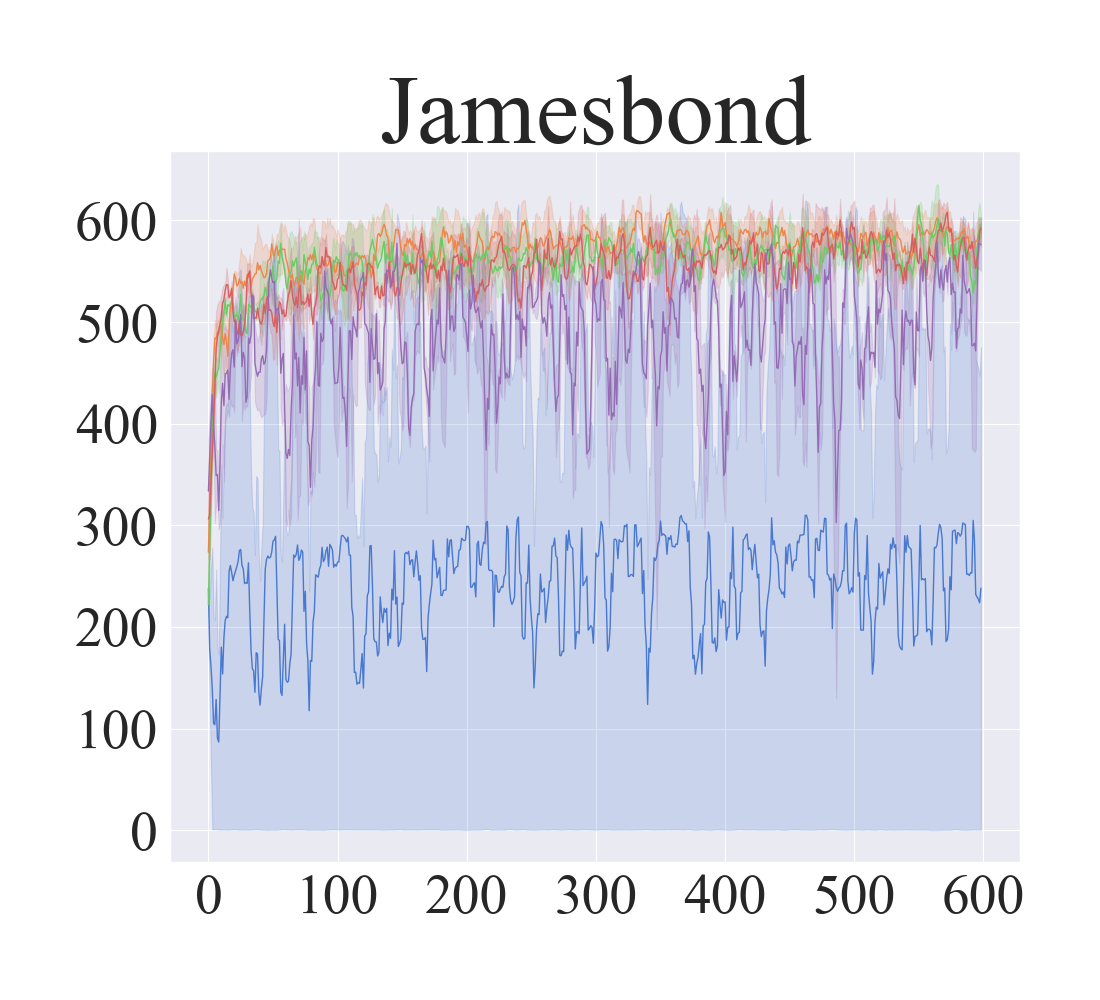}\\
			\end{minipage}%
		}%
		\vspace{-0.6cm}
		
		\subfigure{
			\begin{minipage}[t]{0.166\linewidth}
				\centering
				\includegraphics[width=1.05in]{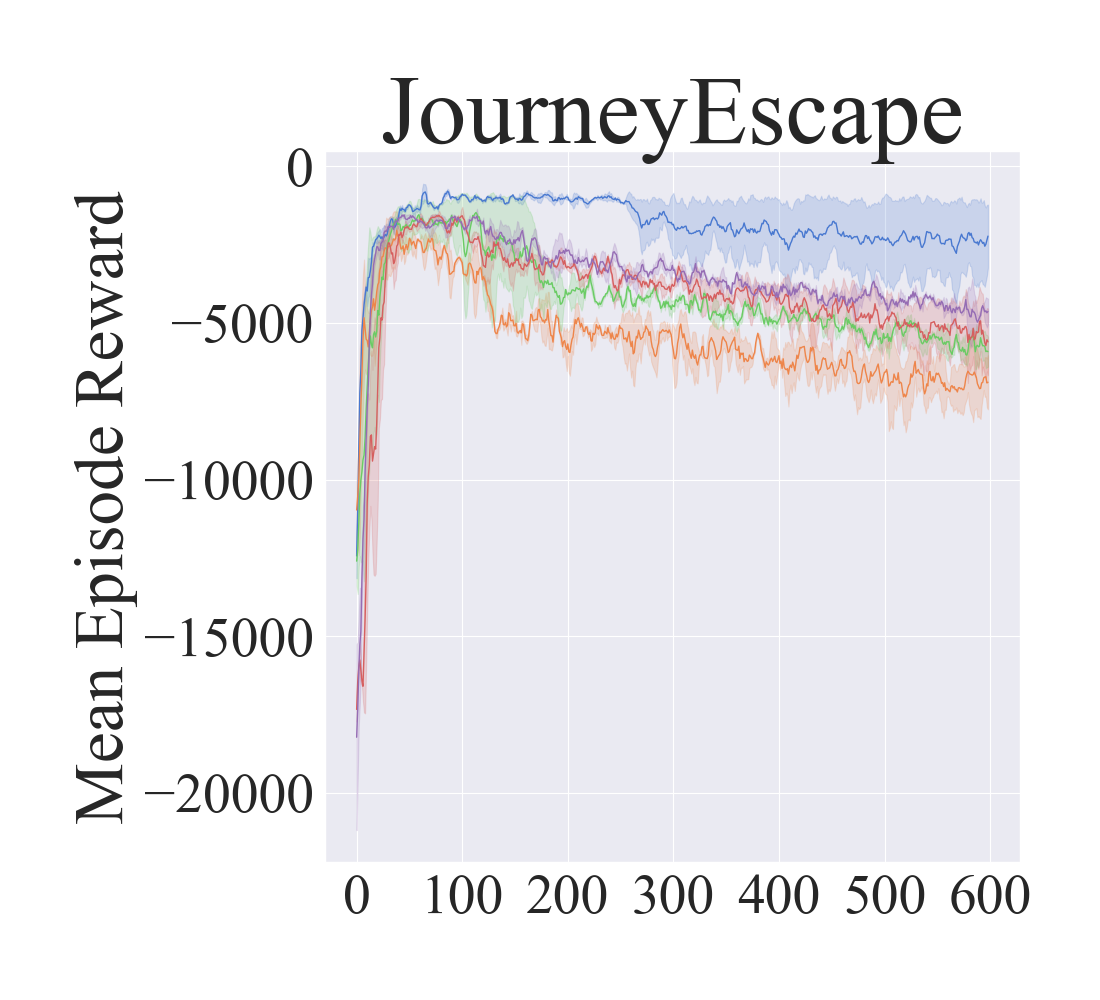}\\
			\end{minipage}%
		}%
		\subfigure{
			\begin{minipage}[t]{0.166\linewidth}
				\centering
				\includegraphics[width=1.05in]{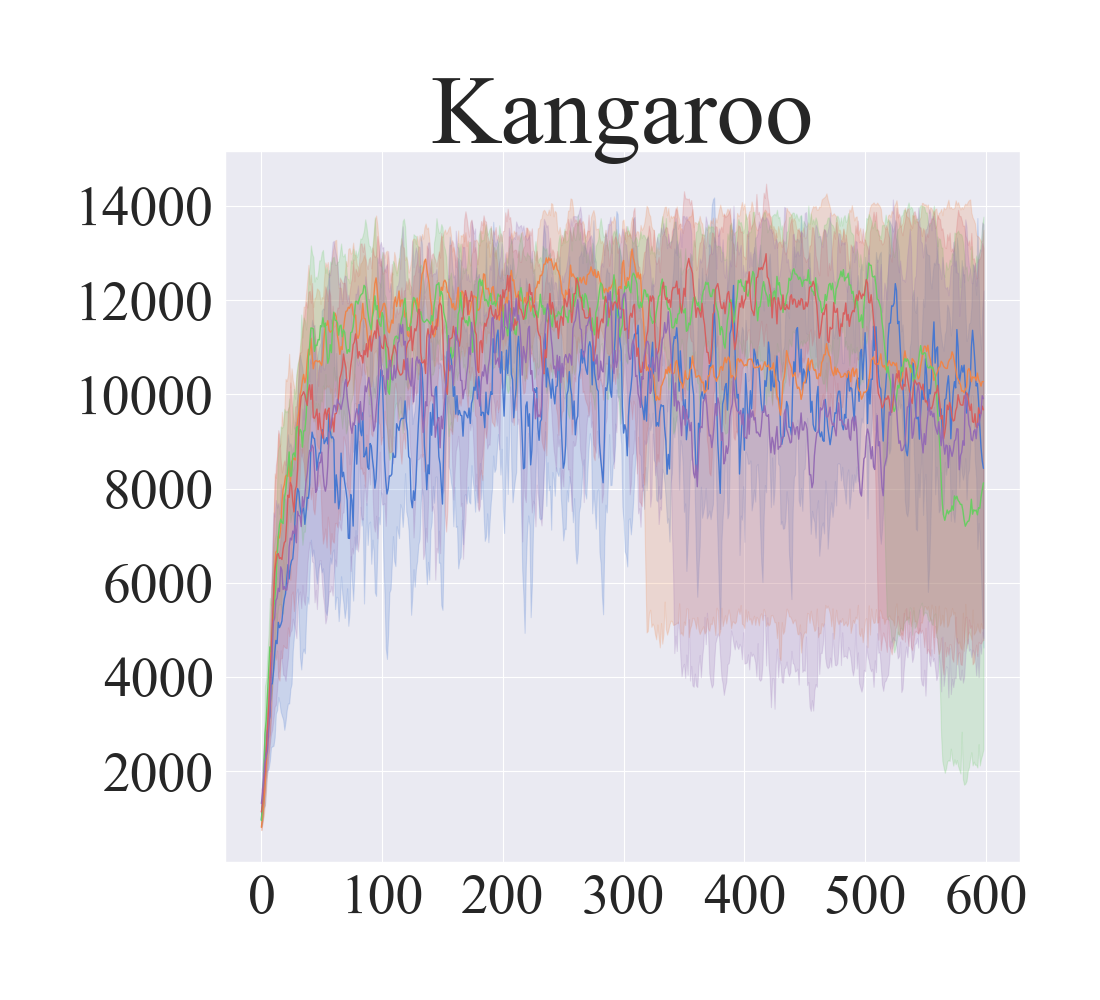}\\
			\end{minipage}%
		}%
		\subfigure{
			\begin{minipage}[t]{0.166\linewidth}
				\centering
				\includegraphics[width=1.05in]{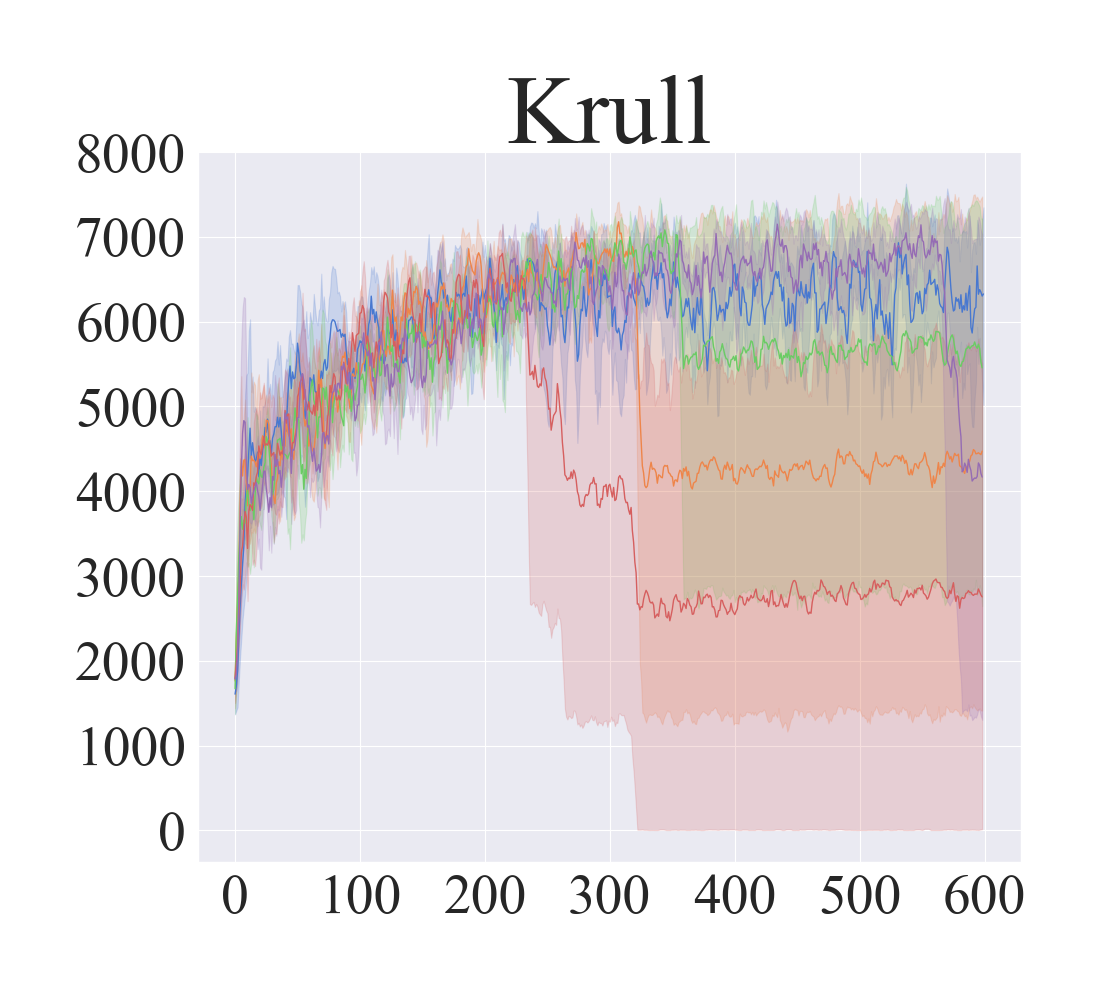}\\
			\end{minipage}%
		}%
		\subfigure{
			\begin{minipage}[t]{0.166\linewidth}
				\centering
				\includegraphics[width=1.05in]{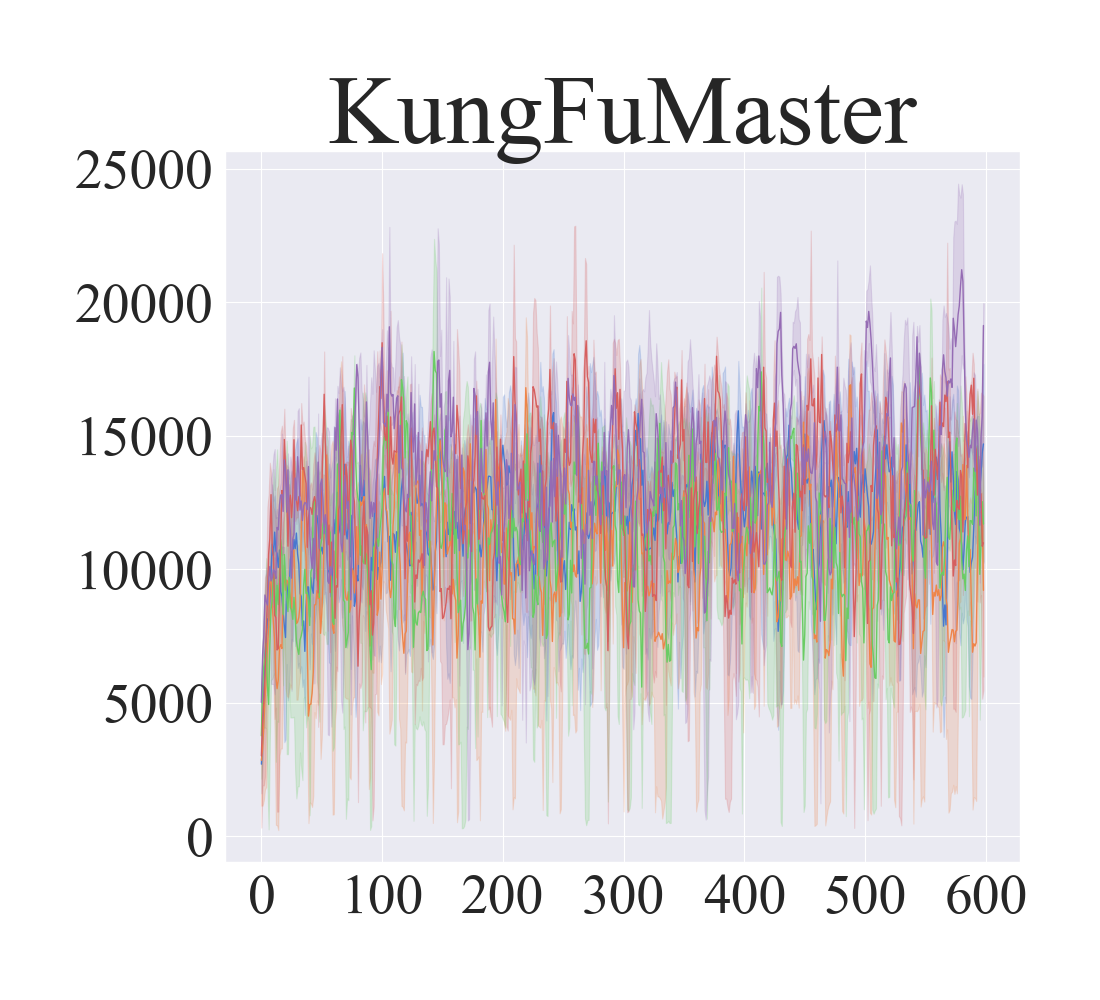}\\
			\end{minipage}%
		}%
		\subfigure{
			\begin{minipage}[t]{0.166\linewidth}
				\centering
				\includegraphics[width=1.05in]{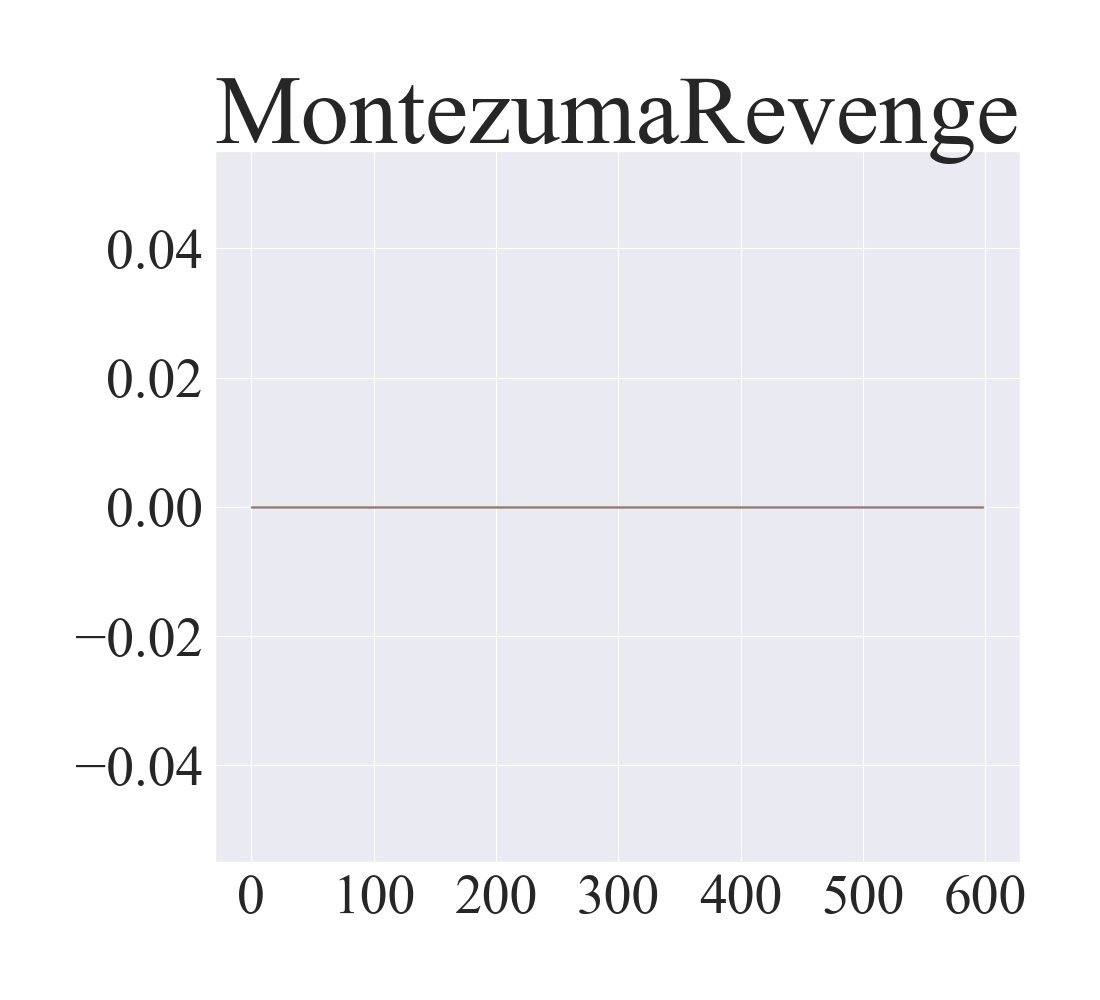}\\
			\end{minipage}%
		}%
		\subfigure{
			\begin{minipage}[t]{0.166\linewidth}
				\centering
				\includegraphics[width=1.05in]{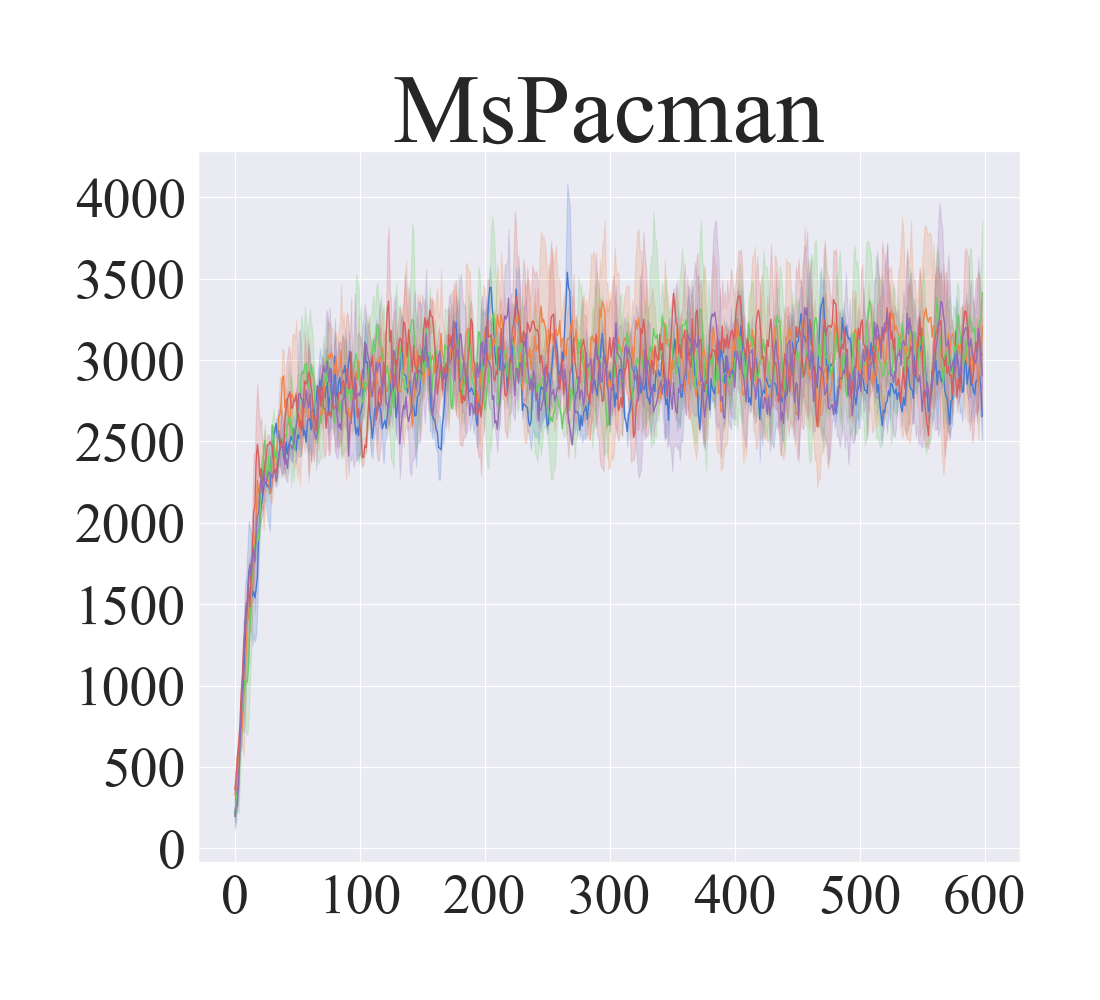}\\
			\end{minipage}%
		}%
		\vspace{-0.6cm}
		
		\subfigure{
			\begin{minipage}[t]{0.166\linewidth}
				\centering
				\includegraphics[width=1.05in]{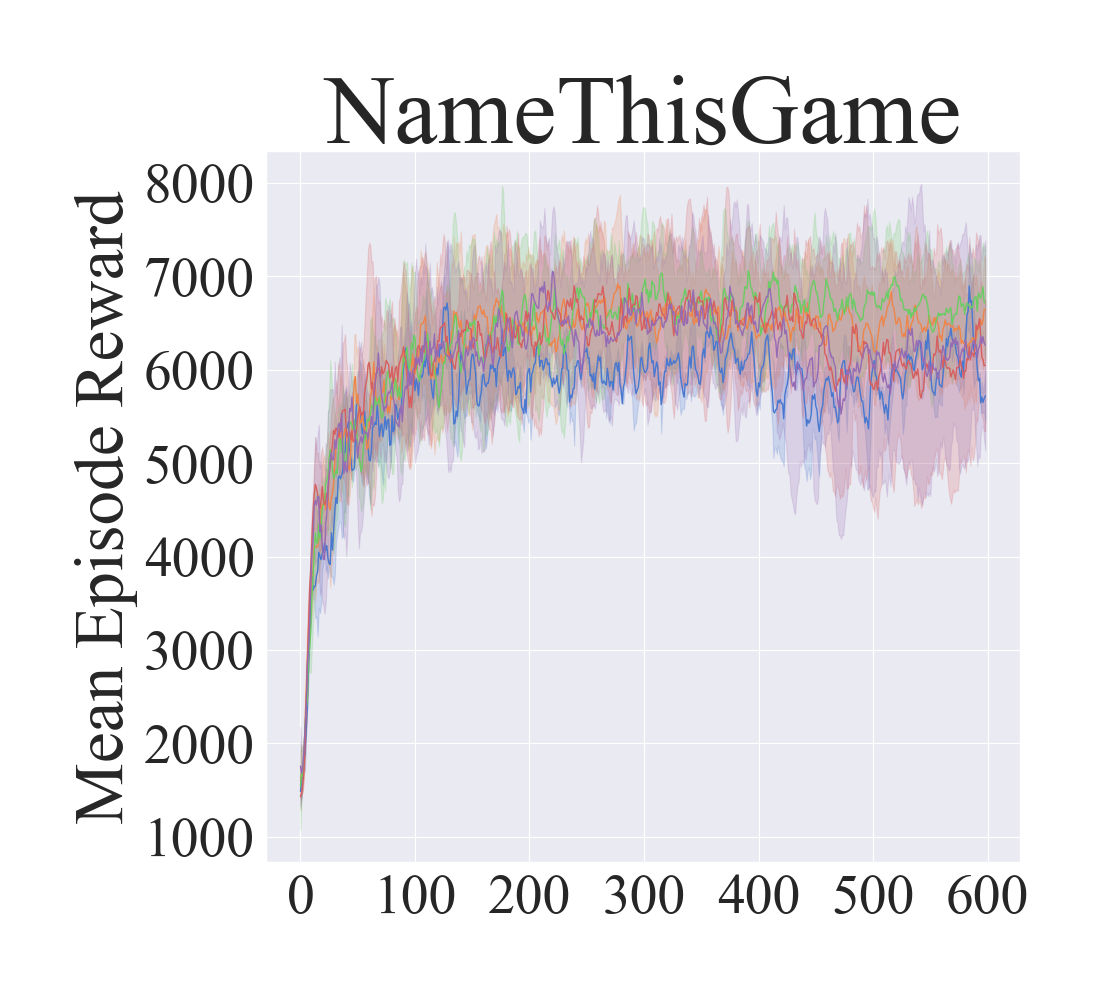}\\
			\end{minipage}%
		}%
		\subfigure{
			\begin{minipage}[t]{0.166\linewidth}
				\centering
				\includegraphics[width=1.05in]{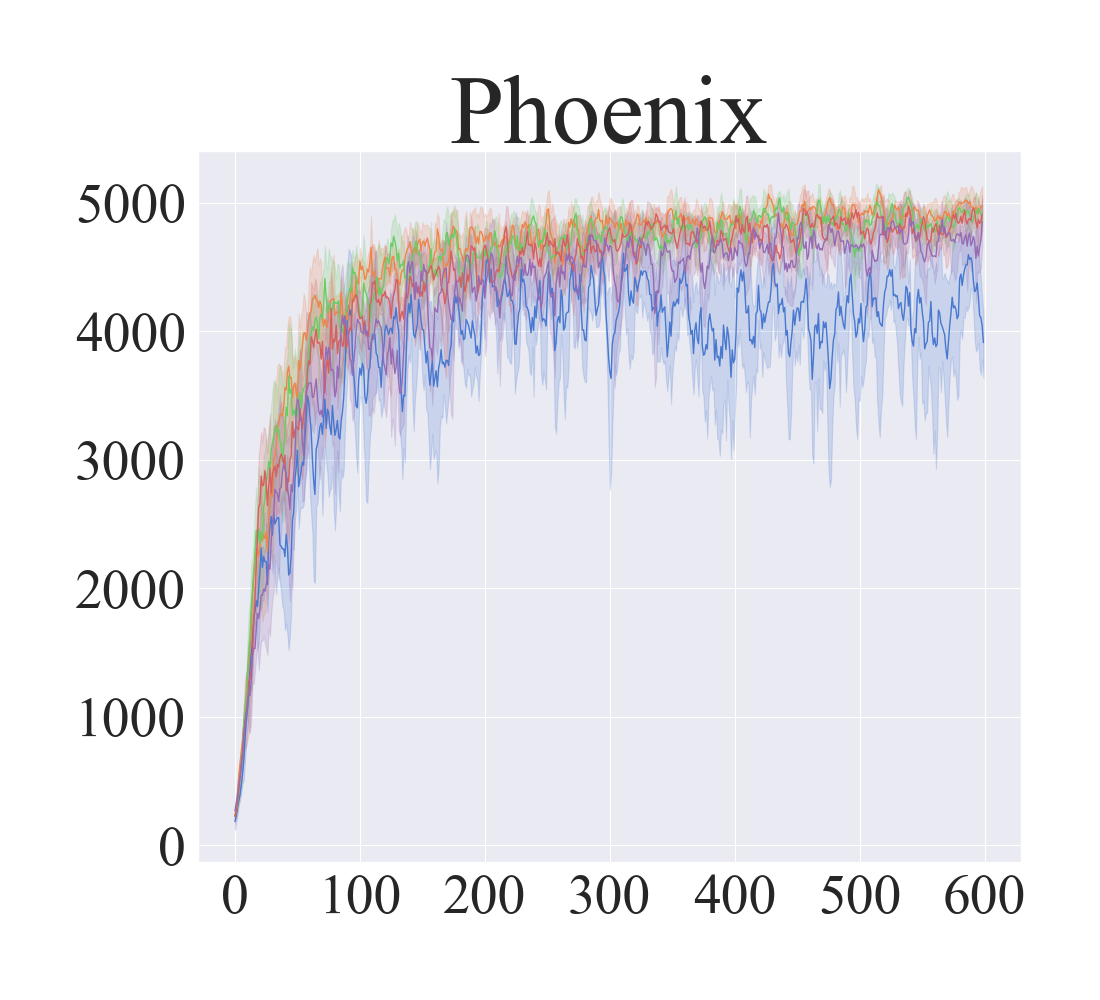}\\
			\end{minipage}%
		}%
		\subfigure{
			\begin{minipage}[t]{0.166\linewidth}
				\centering
				\includegraphics[width=1.05in]{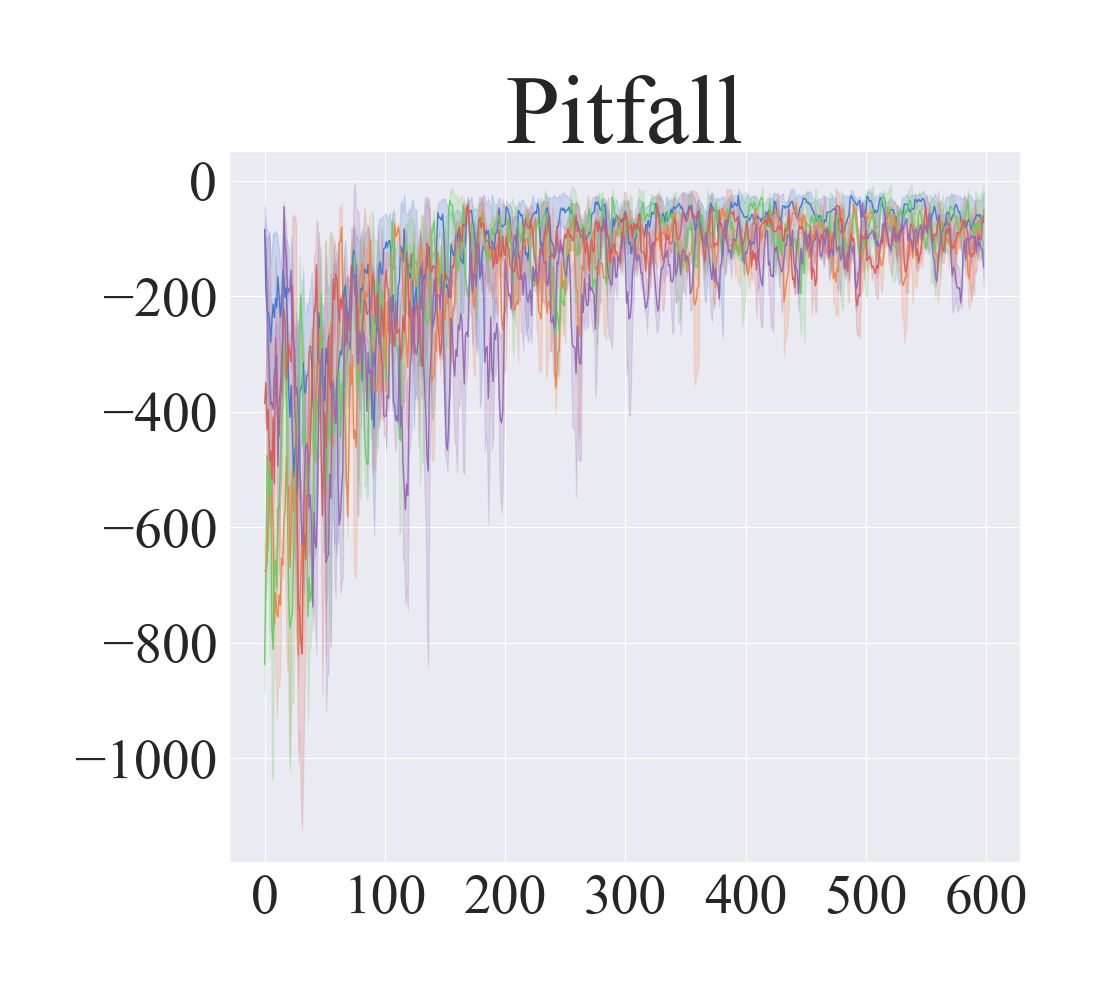}\\
			\end{minipage}%
		}%
		\subfigure{
			\begin{minipage}[t]{0.166\linewidth}
				\centering
				\includegraphics[width=1.05in]{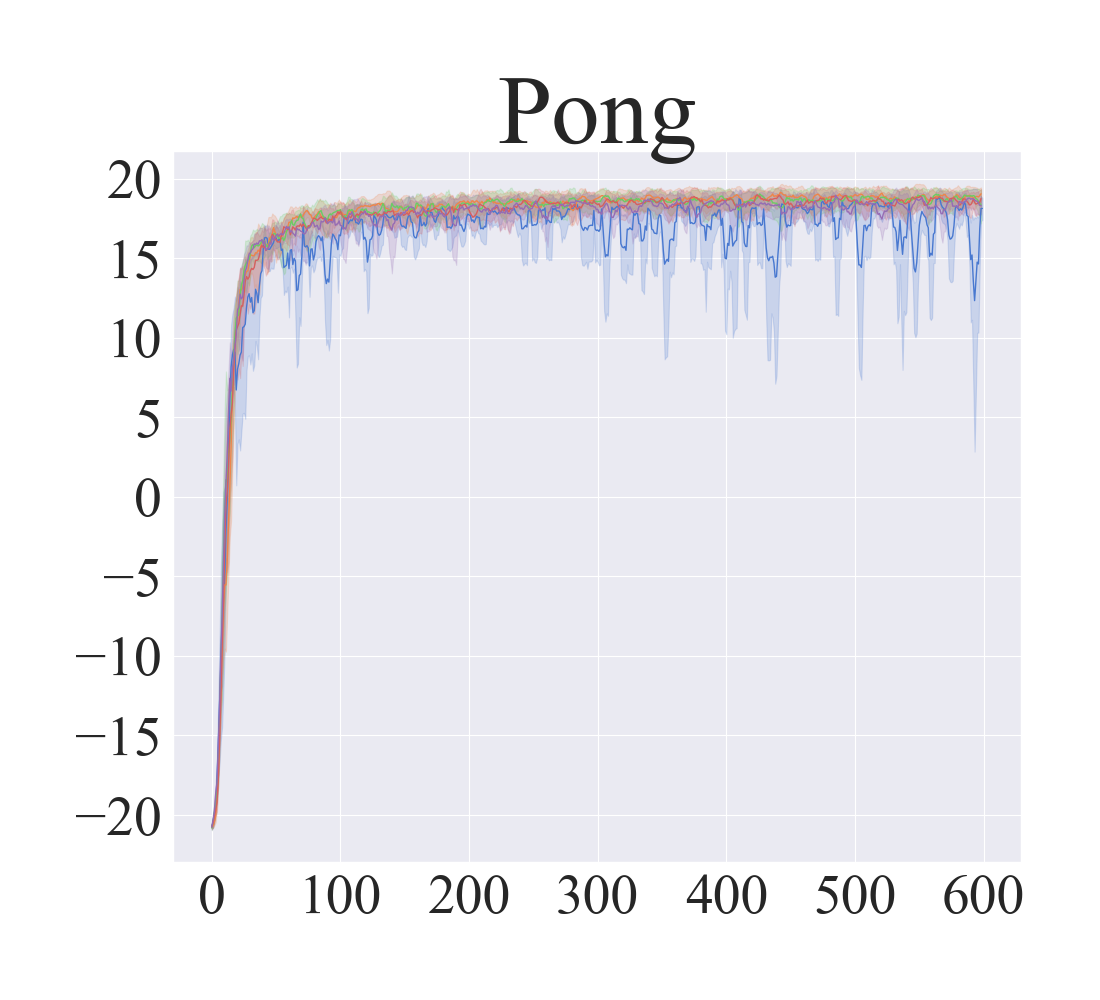}\\
			\end{minipage}%
		}%
		\subfigure{
			\begin{minipage}[t]{0.166\linewidth}
				\centering
				\includegraphics[width=1.05in]{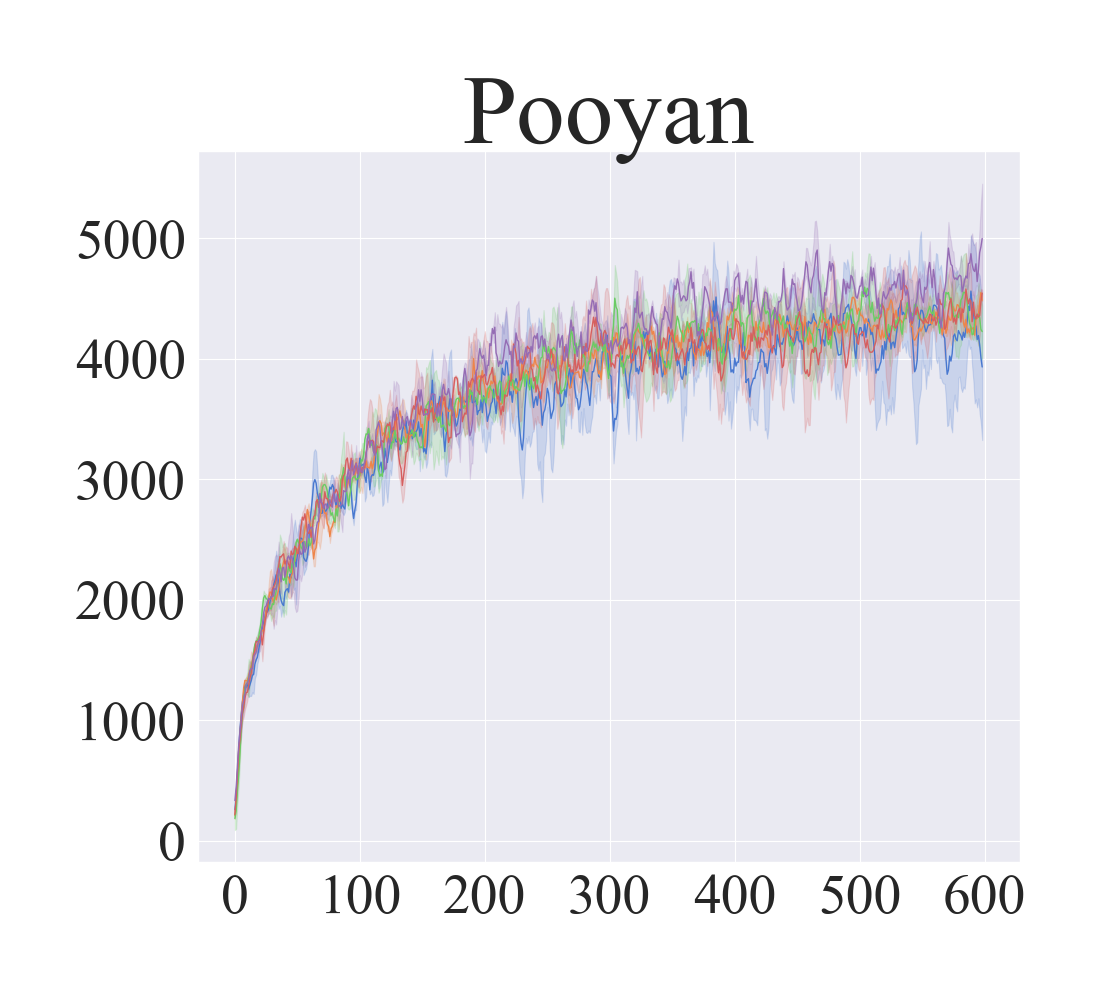}\\
			\end{minipage}%
		}%
		\subfigure{
			\begin{minipage}[t]{0.166\linewidth}
				\centering
				\includegraphics[width=1.05in]{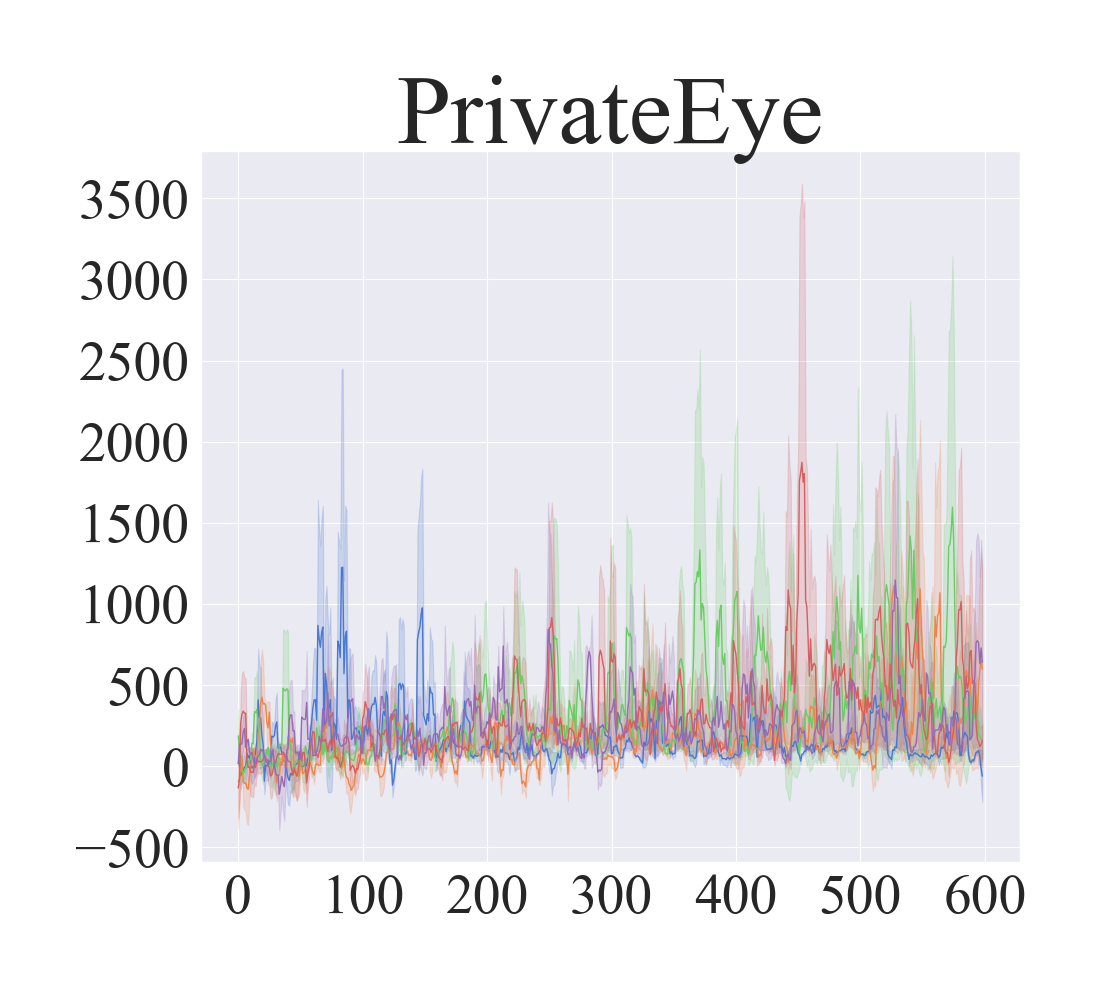}\\
			\end{minipage}%
		}%
		\vspace{-0.6cm}
		
		\subfigure{
			\begin{minipage}[t]{0.166\linewidth}
				\centering
				\includegraphics[width=1.05in]{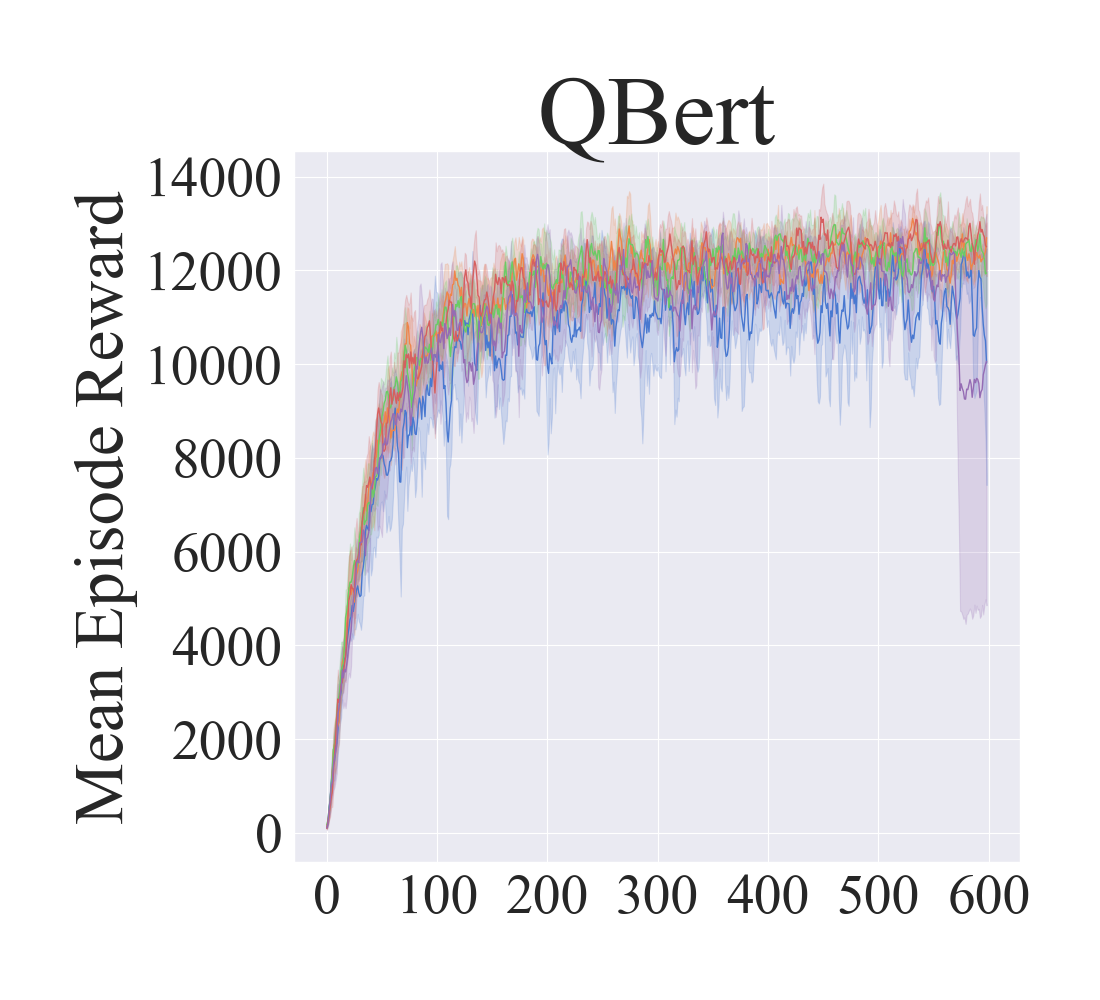}\\
			\end{minipage}%
		}%
		\subfigure{
			\begin{minipage}[t]{0.166\linewidth}
				\centering
				\includegraphics[width=1.05in]{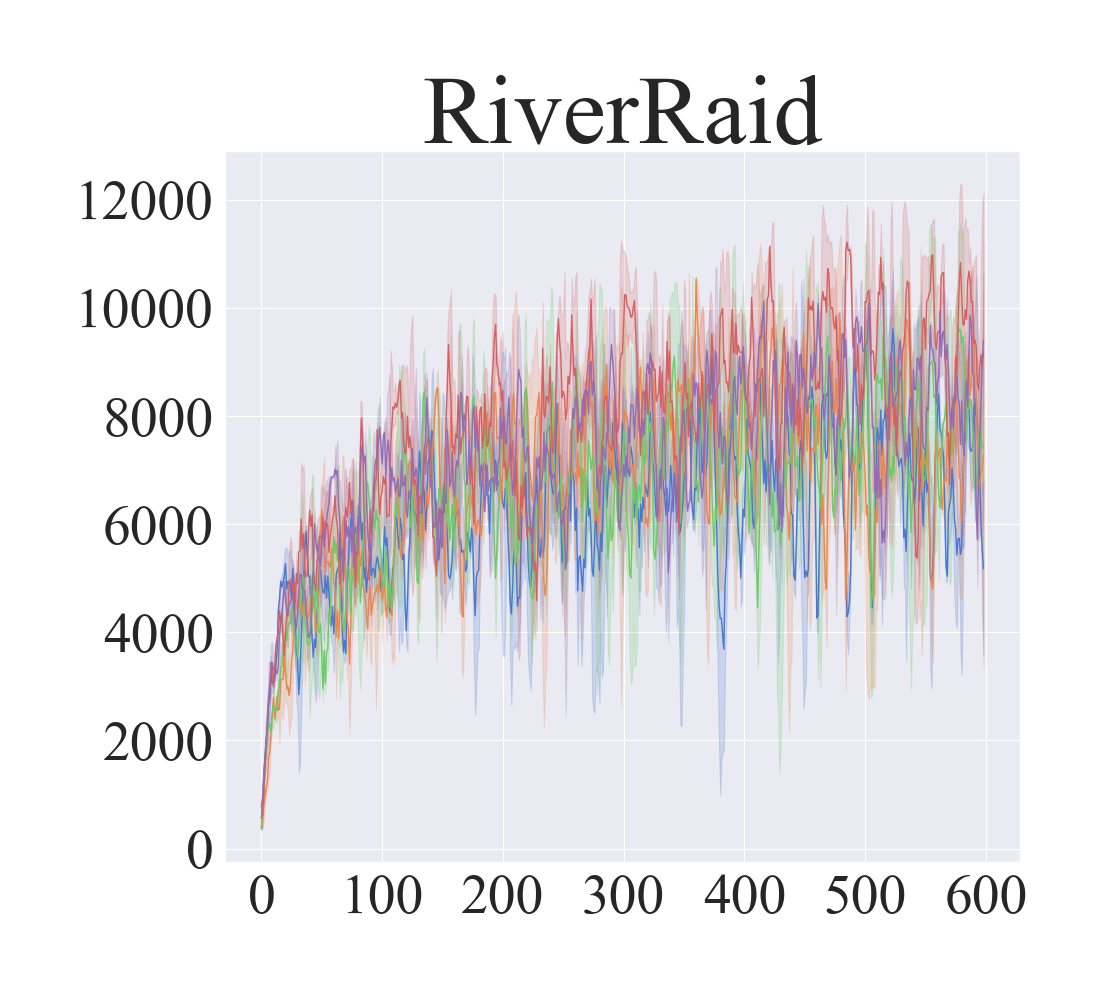}\\
			\end{minipage}%
		}%
		\subfigure{
			\begin{minipage}[t]{0.166\linewidth}
				\centering
				\includegraphics[width=1.05in]{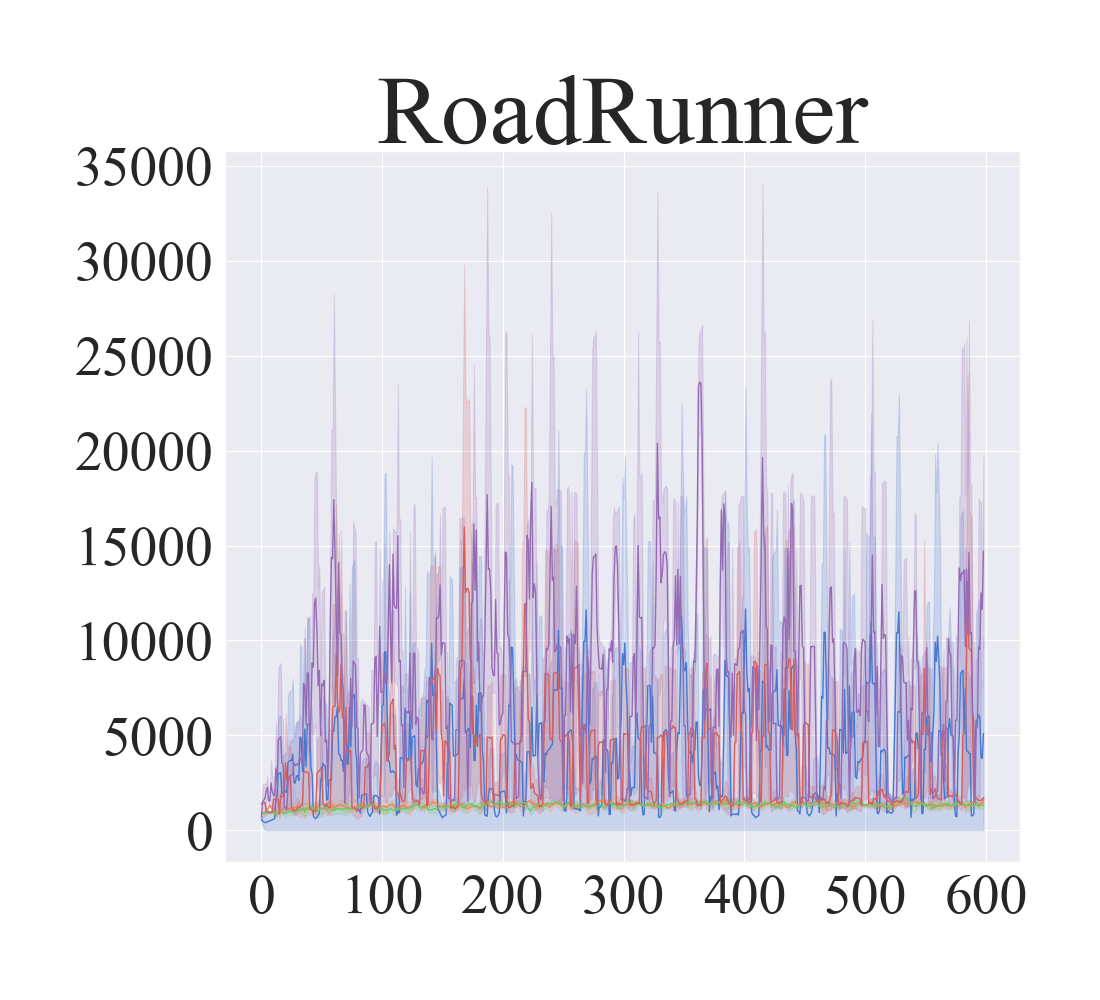}\\
			\end{minipage}%
		}%
		\subfigure{
			\begin{minipage}[t]{0.166\linewidth}
				\centering
				\includegraphics[width=1.05in]{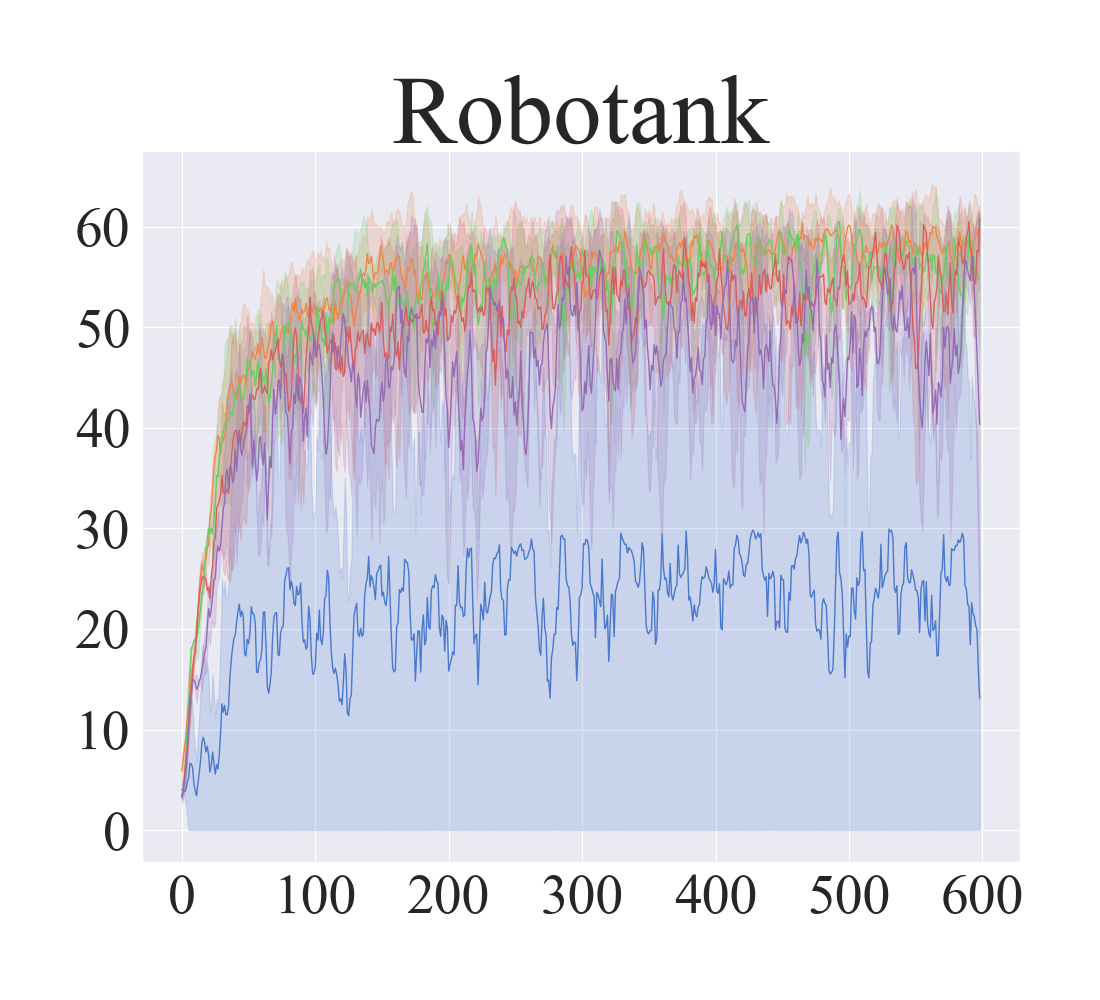}\\
			\end{minipage}%
		}%
		\subfigure{
			\begin{minipage}[t]{0.166\linewidth}
				\centering
				\includegraphics[width=1.05in]{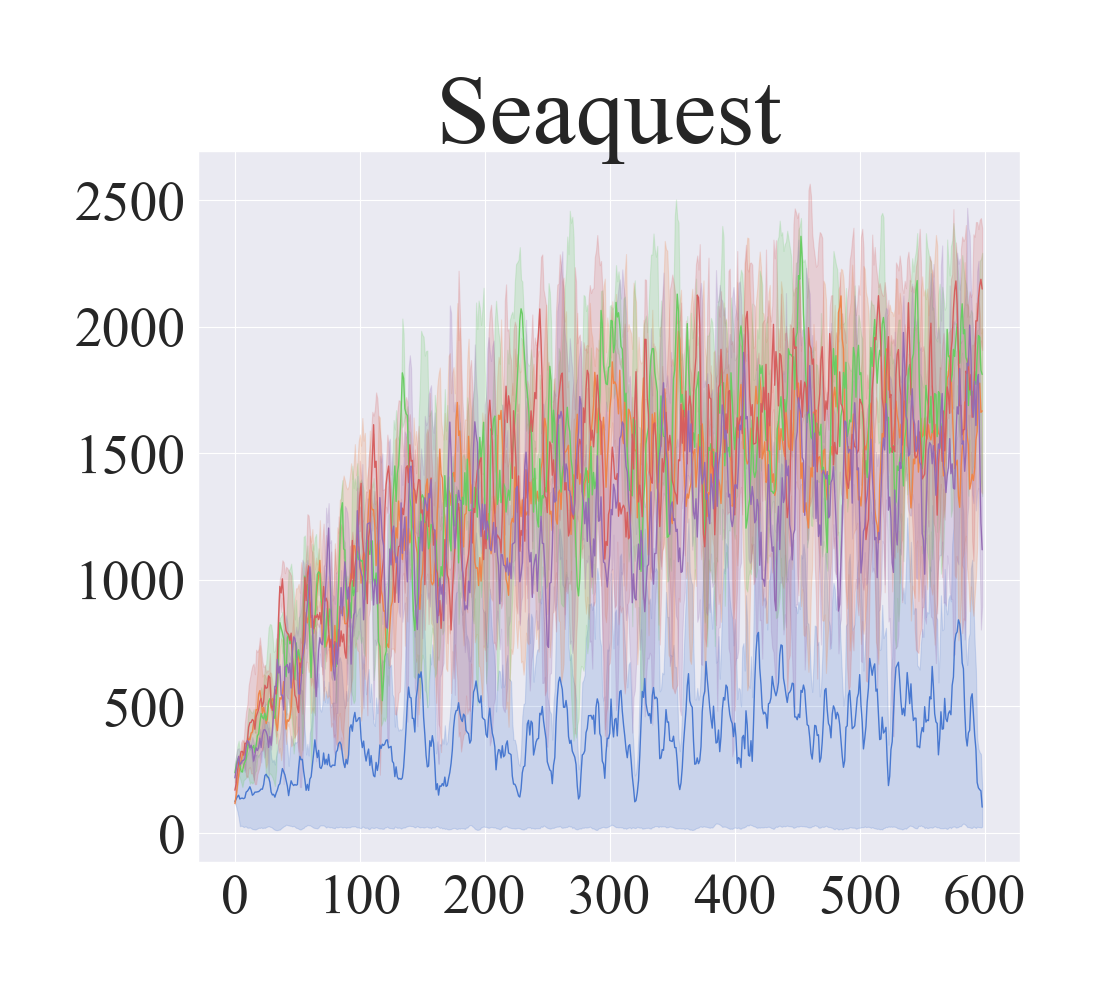}\\
			\end{minipage}%
		}%
		\subfigure{
			\begin{minipage}[t]{0.166\linewidth}
				\centering
				\includegraphics[width=1.05in]{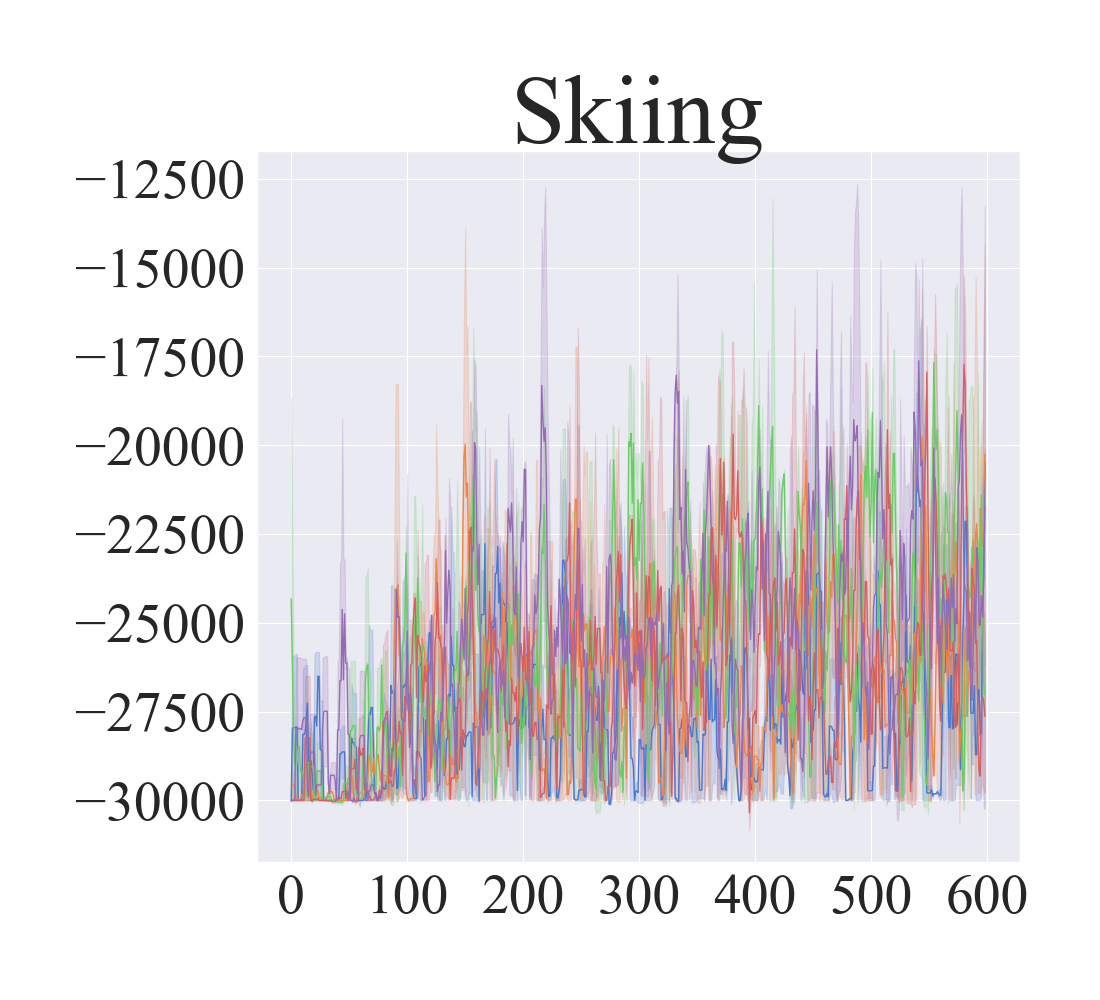}\\
			\end{minipage}%
		}%
		\vspace{-0.6cm}
		
		\subfigure{
			\begin{minipage}[t]{0.166\linewidth}
				\centering
				\includegraphics[width=1.05in]{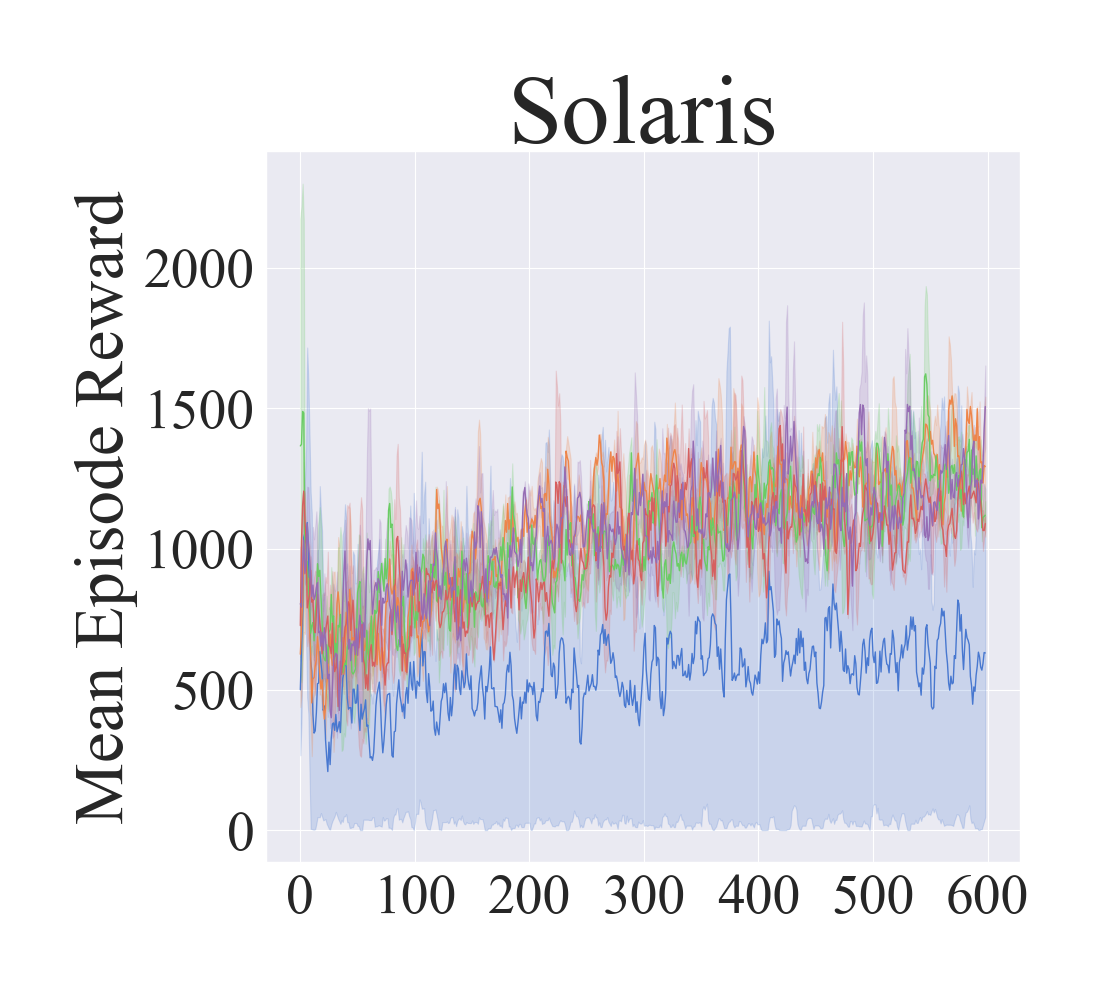}\\
			\end{minipage}%
		}%
		\subfigure{
			\begin{minipage}[t]{0.166\linewidth}
				\centering
				\includegraphics[width=1.05in]{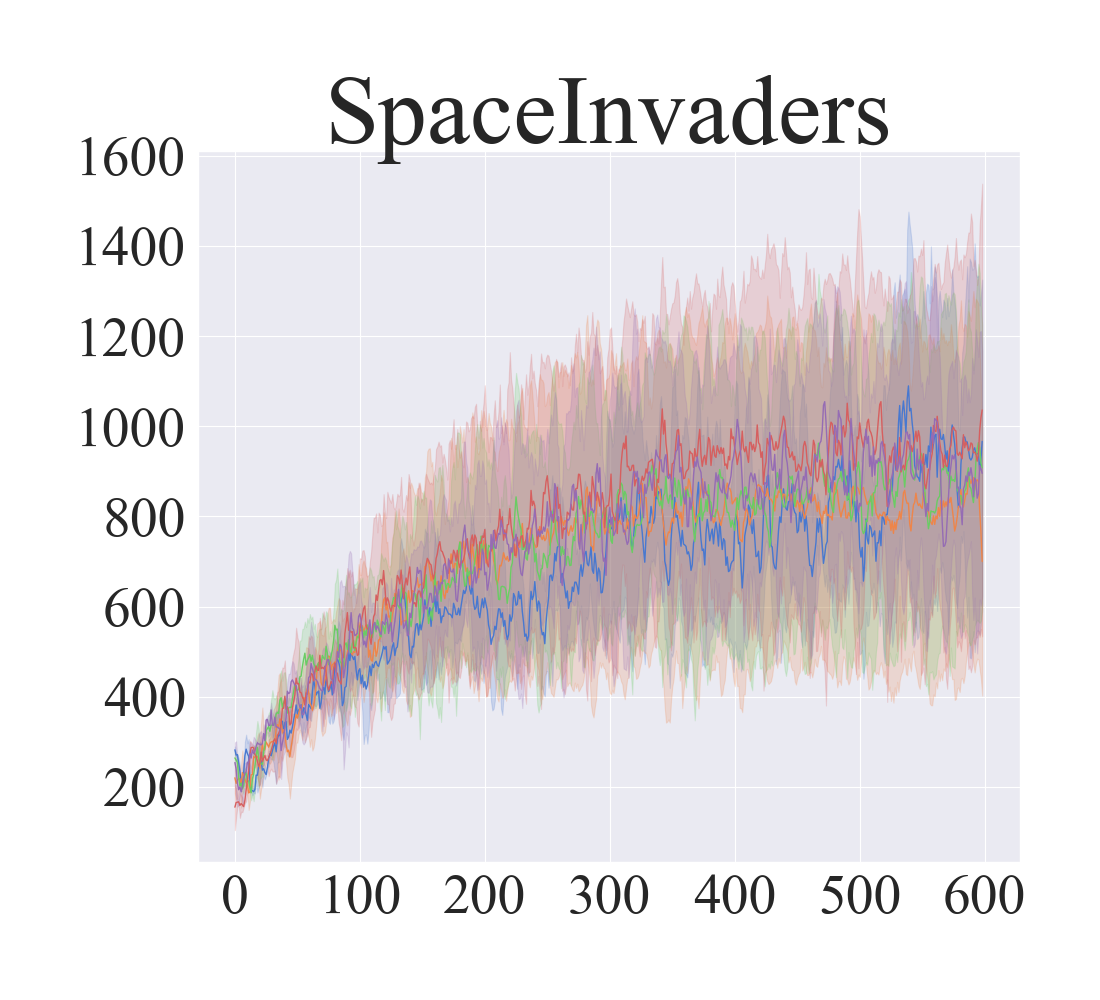}\\
			\end{minipage}%
		}%
		\subfigure{
			\begin{minipage}[t]{0.166\linewidth}
				\centering
				\includegraphics[width=1.05in]{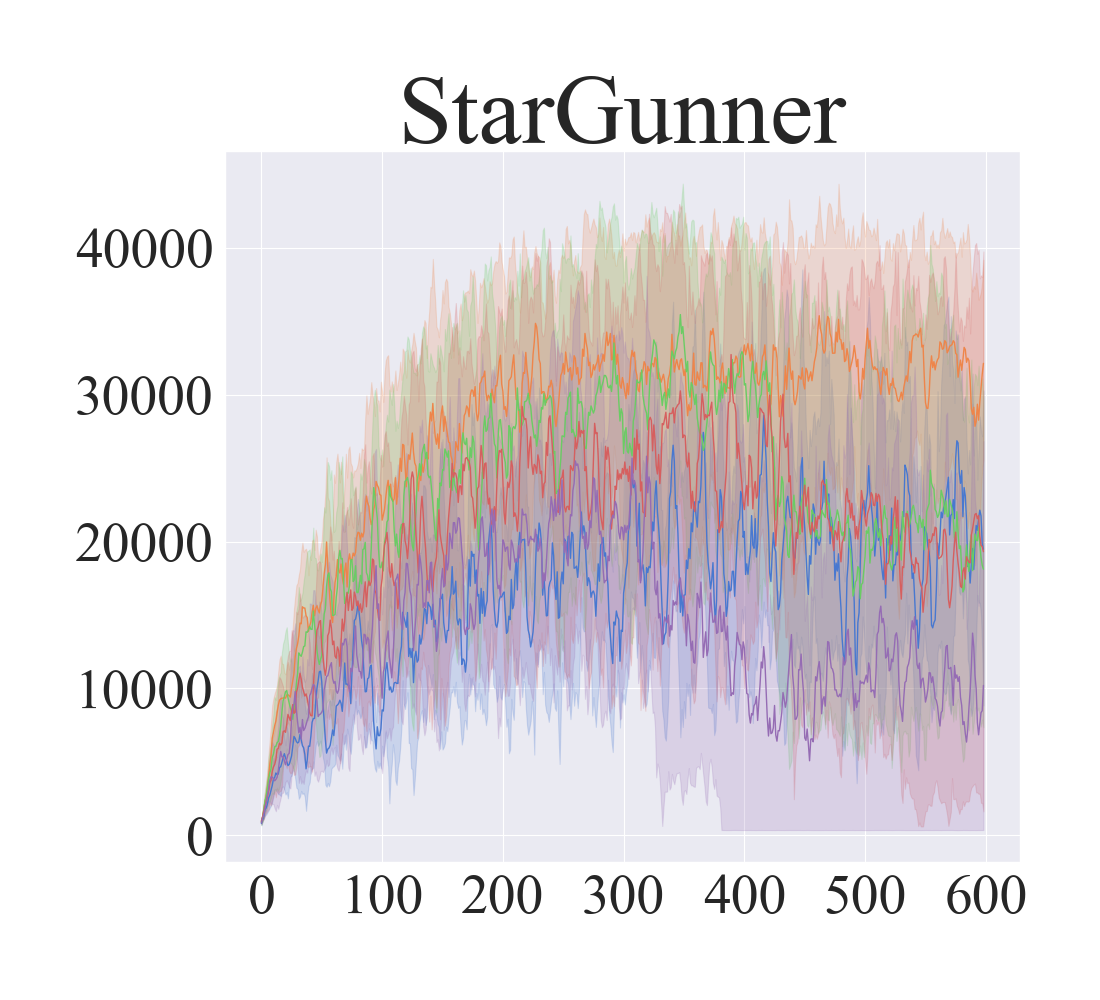}\\
			\end{minipage}%
		}%
		\subfigure{
			\begin{minipage}[t]{0.166\linewidth}
				\centering
				\includegraphics[width=1.05in]{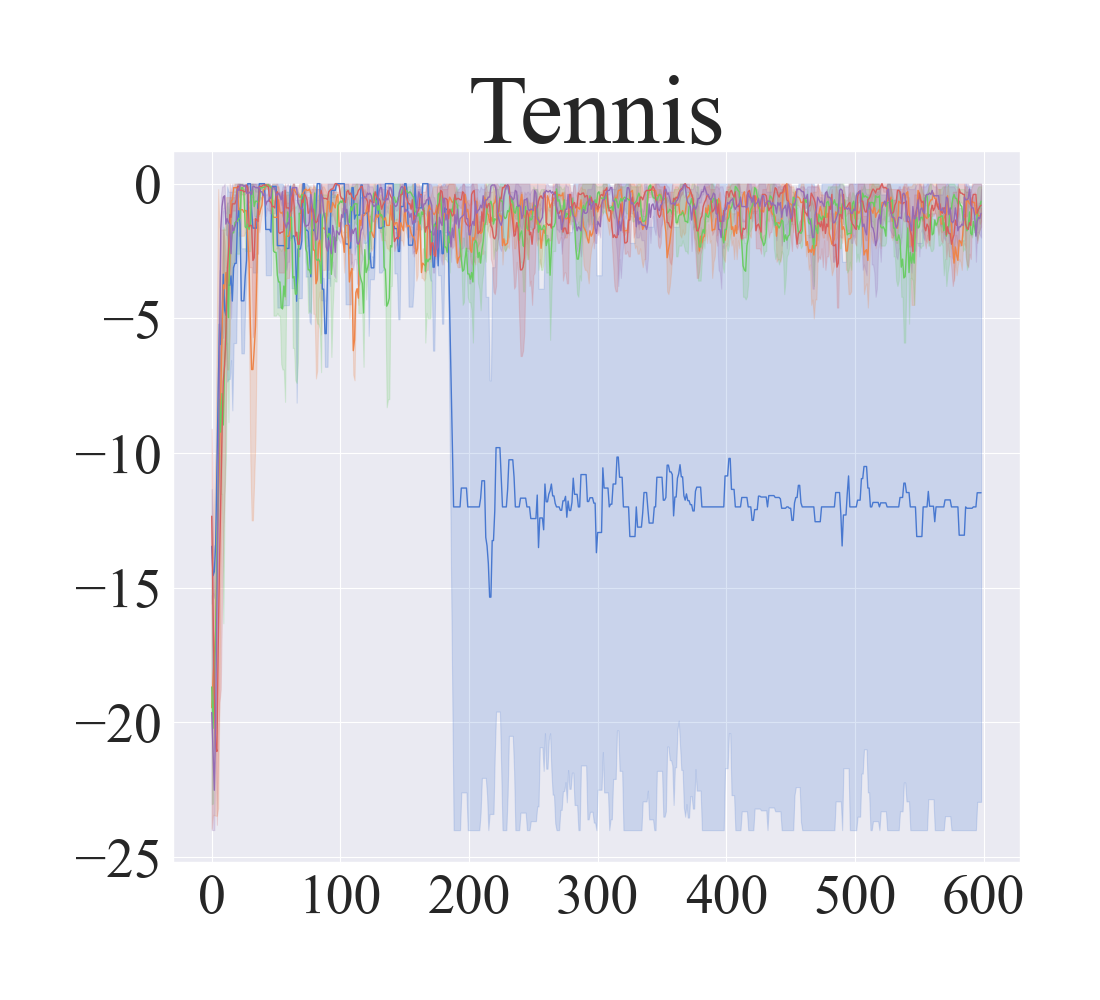}\\
			\end{minipage}%
		}%
		\subfigure{
			\begin{minipage}[t]{0.166\linewidth}
				\centering
				\includegraphics[width=1.05in]{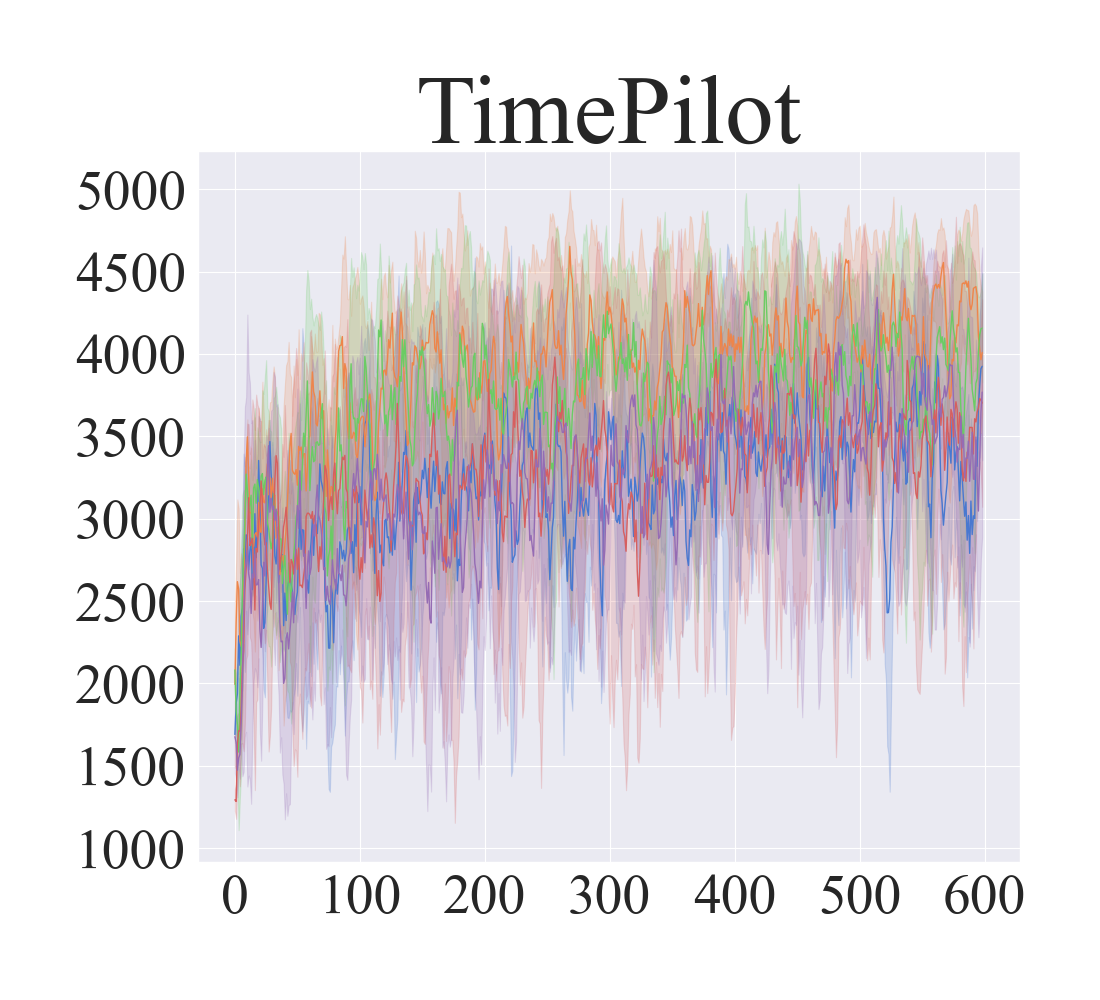}\\
			\end{minipage}%
		}%
		\subfigure{
			\begin{minipage}[t]{0.166\linewidth}
				\centering
				\includegraphics[width=1.05in]{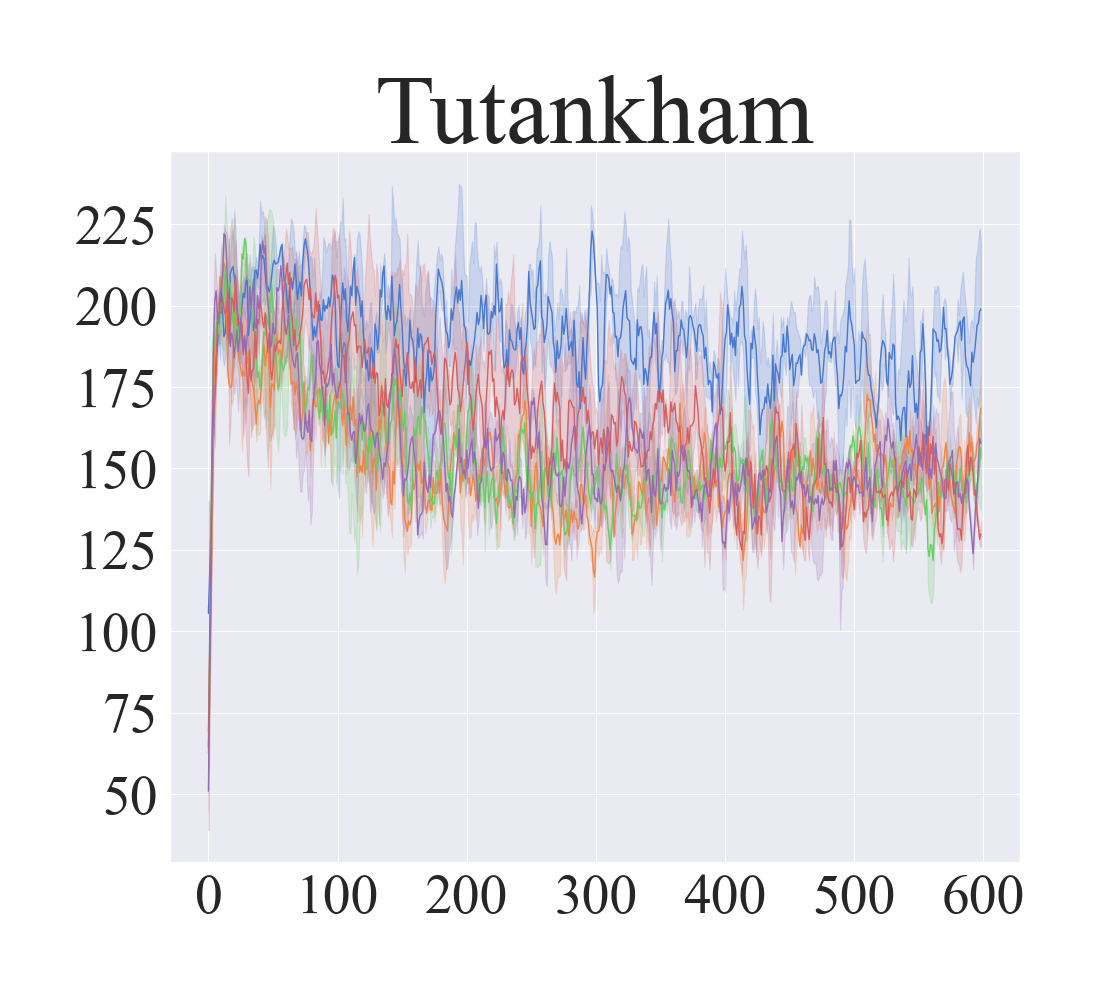}\\
			\end{minipage}%
		}%
		\vspace{-0.6cm}
		
		\subfigure{
			\begin{minipage}[t]{0.166\linewidth}
				\centering
				\includegraphics[width=1.05in]{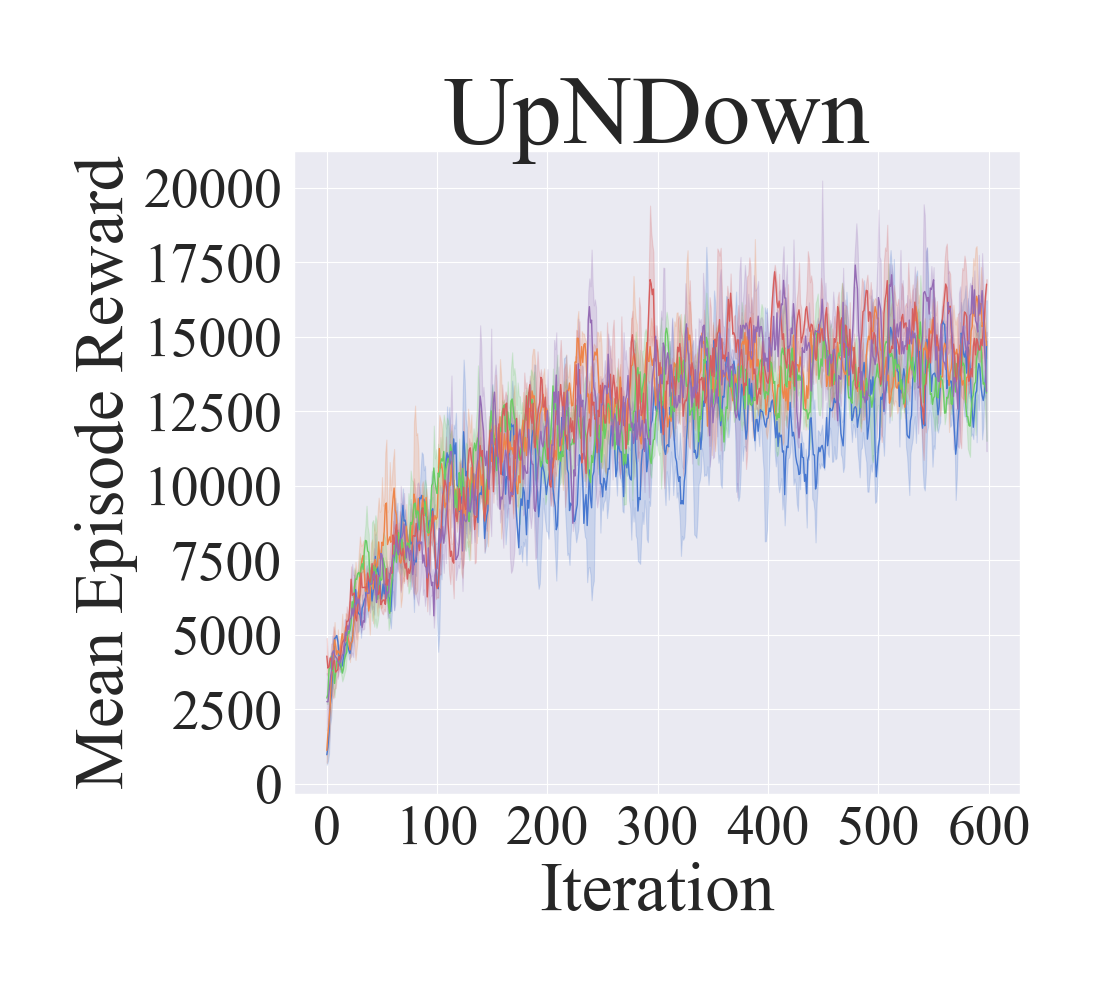}\\
			\end{minipage}%
		}%
		\subfigure{
			\begin{minipage}[t]{0.166\linewidth}
				\centering
				\includegraphics[width=1.05in]{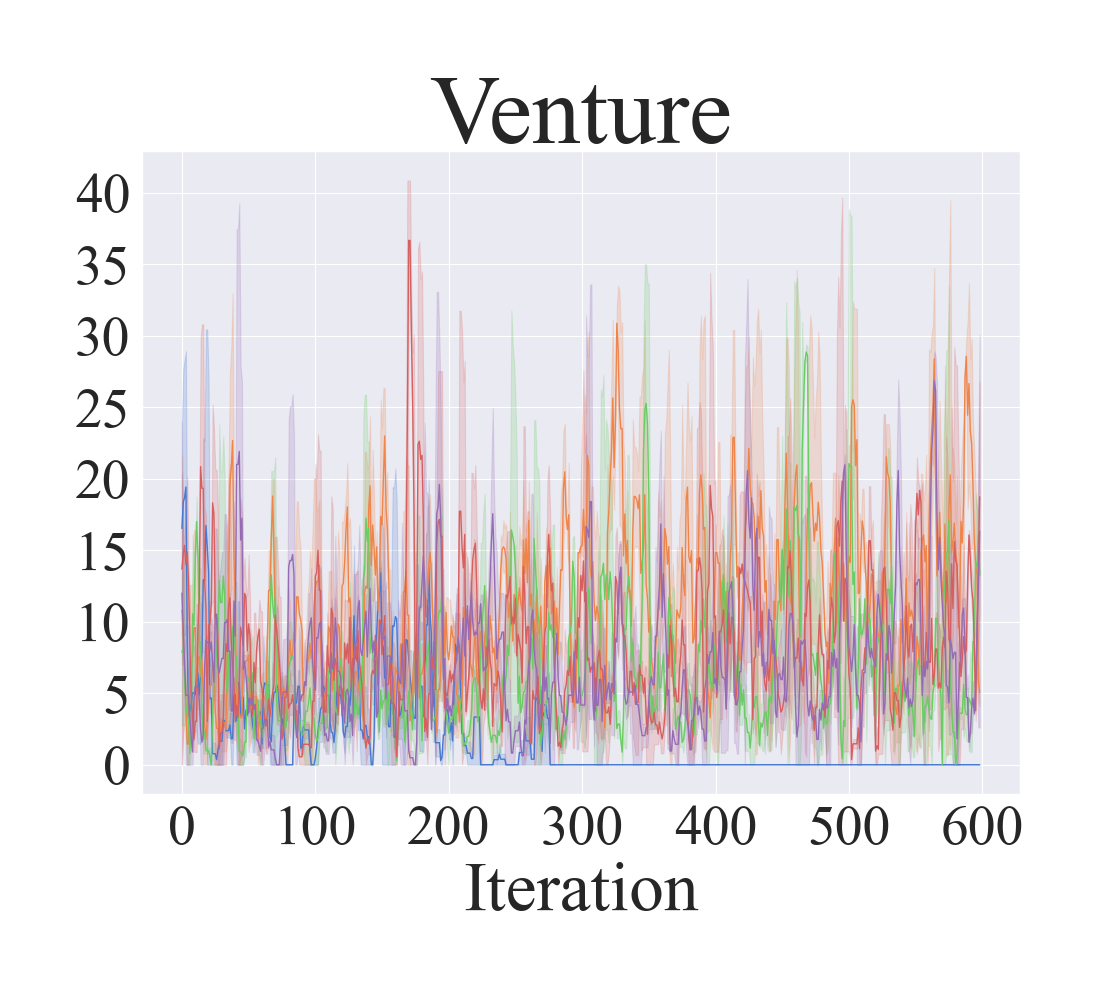}\\
			\end{minipage}%
		}%
		\subfigure{
			\begin{minipage}[t]{0.166\linewidth}
				\centering
				\includegraphics[width=1.05in]{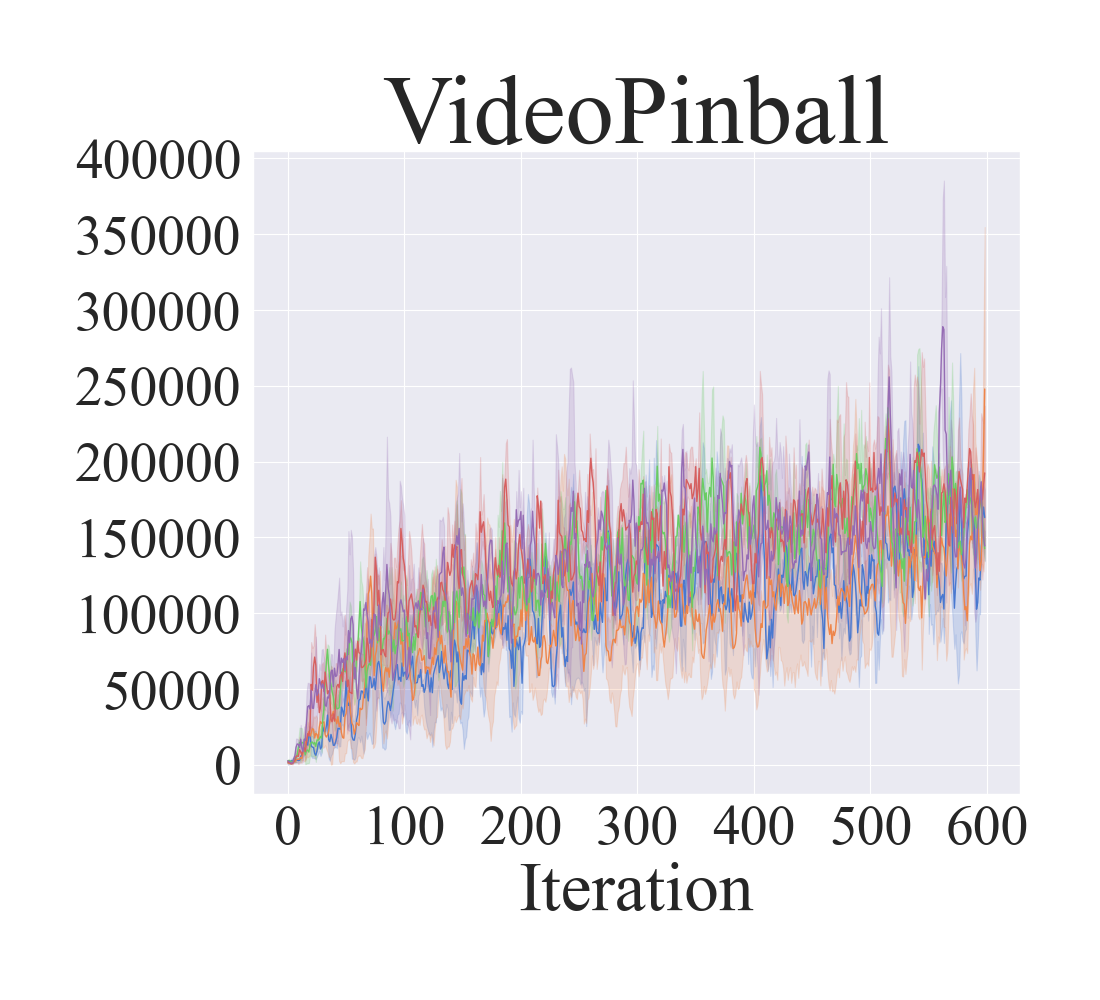}\\
			\end{minipage}%
		}%
		\subfigure{
			\begin{minipage}[t]{0.166\linewidth}
				\centering
				\includegraphics[width=1.05in]{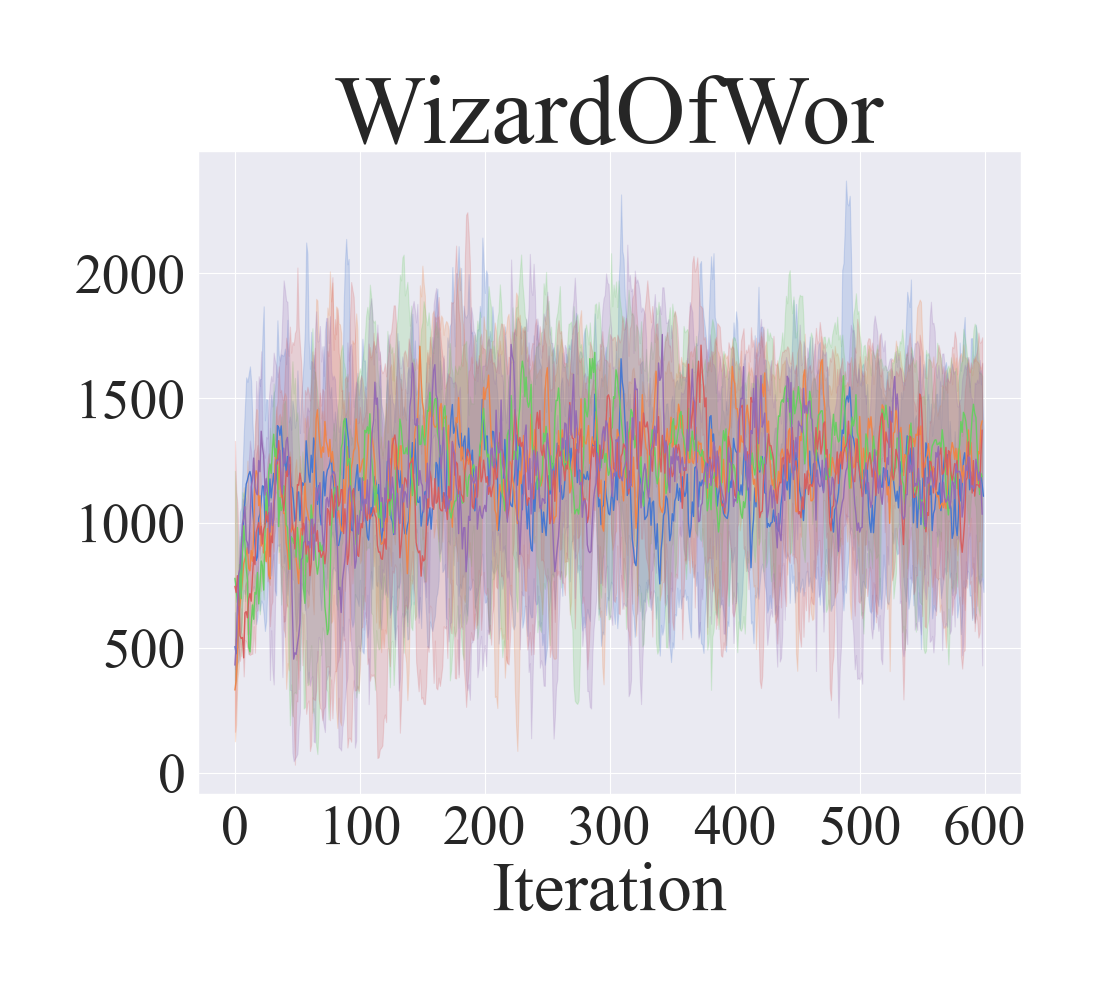}\\
			\end{minipage}%
		}%
		\subfigure{
			\begin{minipage}[t]{0.166\linewidth}
				\centering
				\includegraphics[width=1.05in]{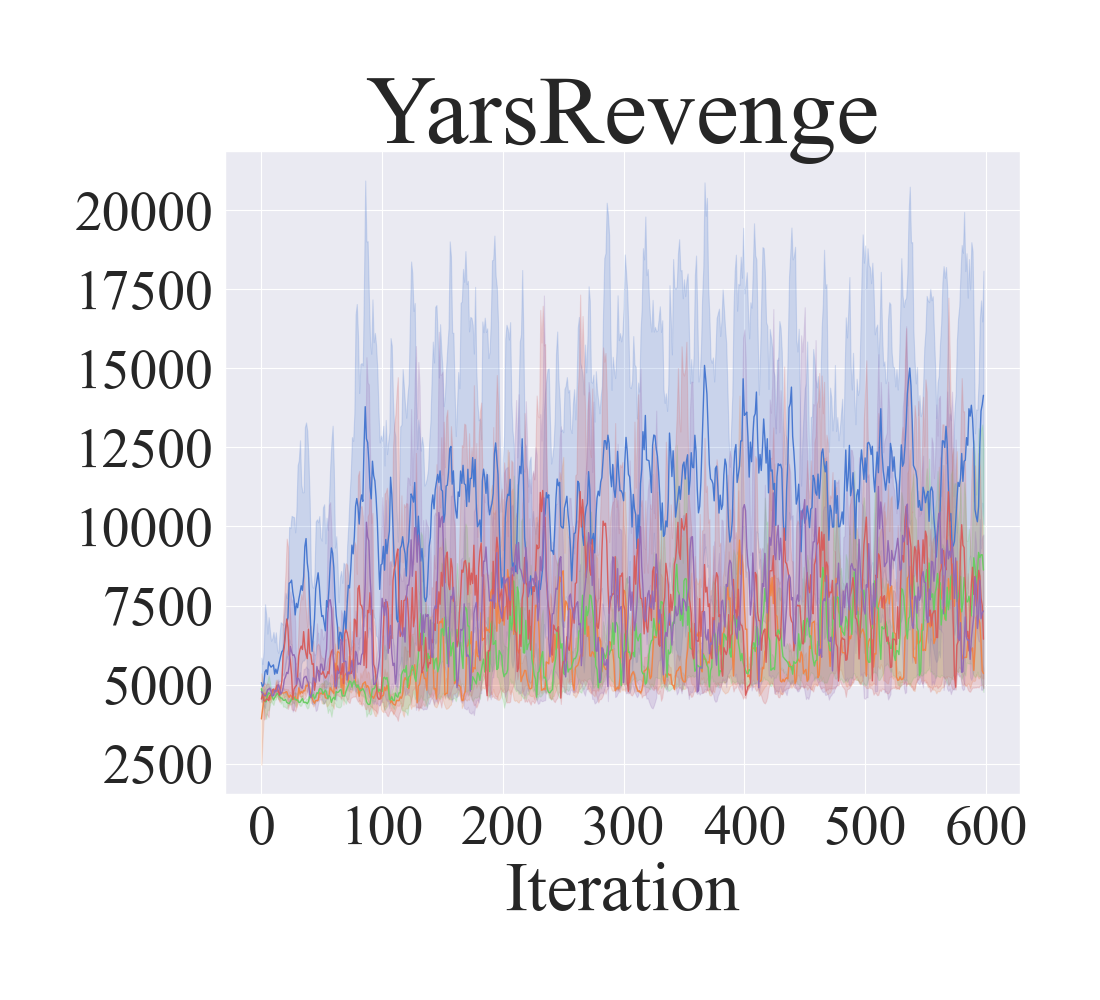}\\
			\end{minipage}%
		}%
		\subfigure{
			\begin{minipage}[t]{0.166\linewidth}
				\centering
				\includegraphics[width=1.05in]{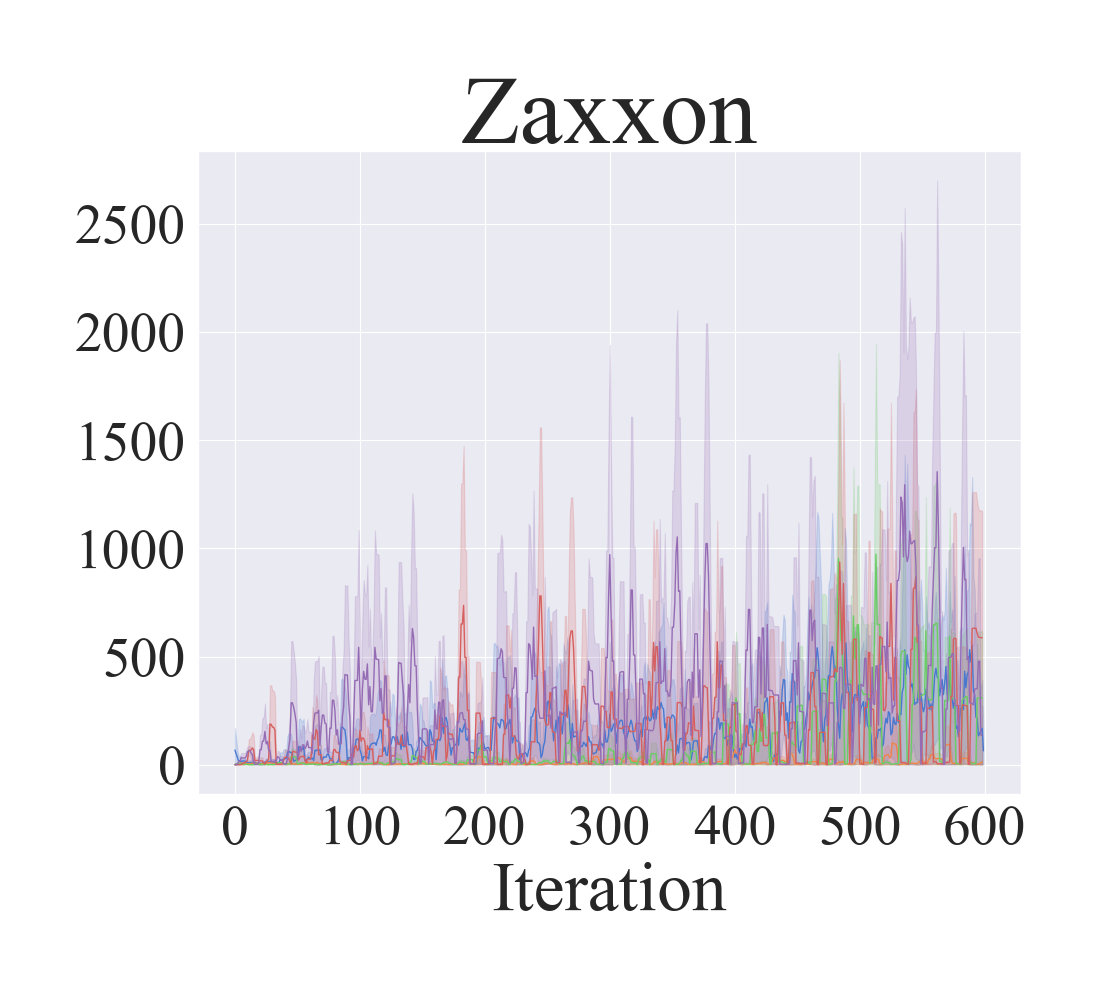}\\
			\end{minipage}%
		}%

		\centering
		\caption{\textbf{Learning curves of all $60$ Atari $2600$ games on medium dataset}}
		\label{fig: Learning curves of all $60$ Atari $2600$ games on medium dataset1}
								
	\end{figure*}

	Please refer Fig. \ref{fig:Learning curves of all $60$ Atari $2600$ games on medium dataset} and Fig. \ref{fig: Learning curves of all $60$ Atari $2600$ games on medium dataset1}.
	
	\subsection{Learning curves of all $60$ Atari $2600$ games on high dataset}
\begin{figure*}[!htb]
		\centering

		\subfigure{
			\begin{minipage}[t]{\linewidth}
				\centering
				\includegraphics[width=4in]{legend.png}\\
			\end{minipage}%
		}%
		
		\subfigure{
			\begin{minipage}[t]{0.166\linewidth}
				\centering
				\includegraphics[width=1.05in]{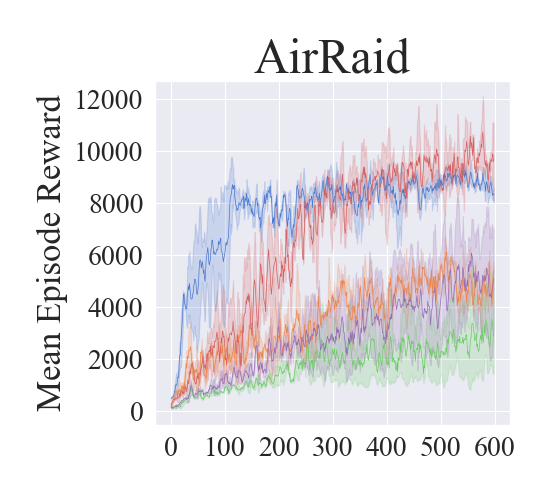}\\
			\end{minipage}%
		}%
		\subfigure{
			\begin{minipage}[t]{0.166\linewidth}
				\centering
				\includegraphics[width=1.05in]{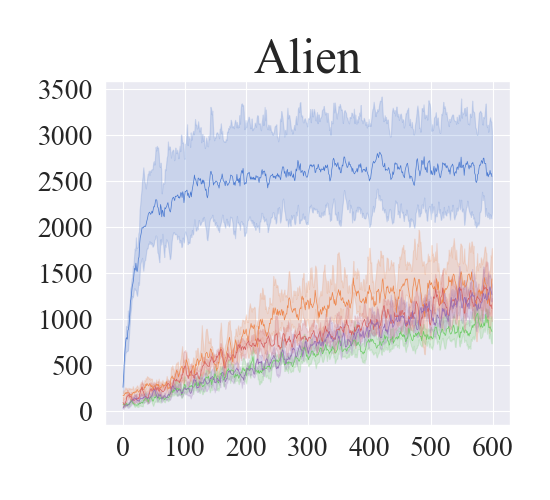}\\
			\end{minipage}%
		}%
		\subfigure{
			\begin{minipage}[t]{0.166\linewidth}
				\centering
				\includegraphics[width=1.05in]{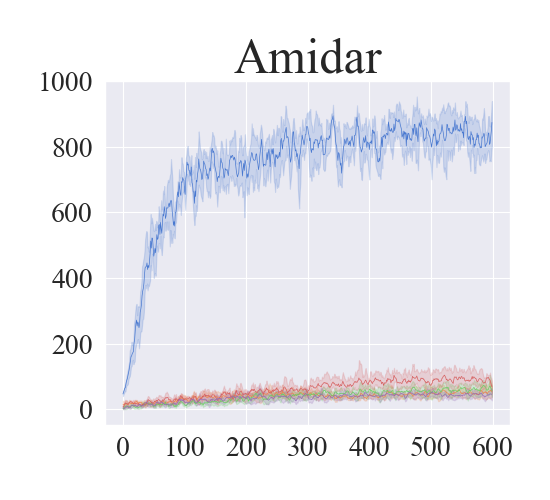}\\
			\end{minipage}%
		}%
		\subfigure{
			\begin{minipage}[t]{0.166\linewidth}
				\centering
				\includegraphics[width=1.05in]{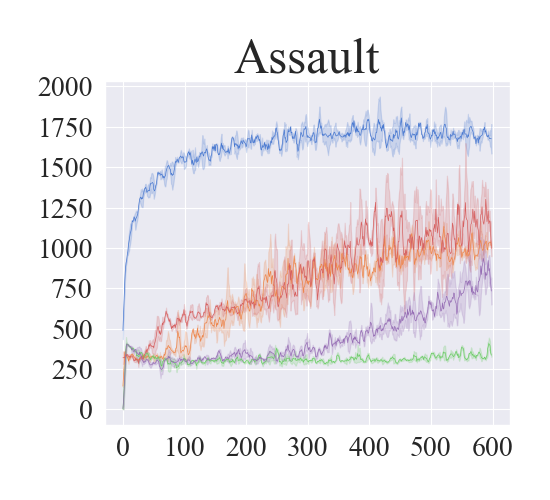}\\
			\end{minipage}%
		}%
		\subfigure{
			\begin{minipage}[t]{0.166\linewidth}
				\centering
				\includegraphics[width=1.05in]{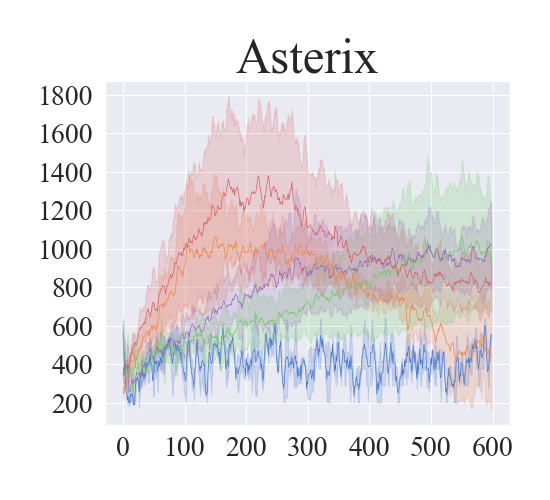}\\
			\end{minipage}%
		}%
		\subfigure{
			\begin{minipage}[t]{0.166\linewidth}
				\centering
				\includegraphics[width=1.05in]{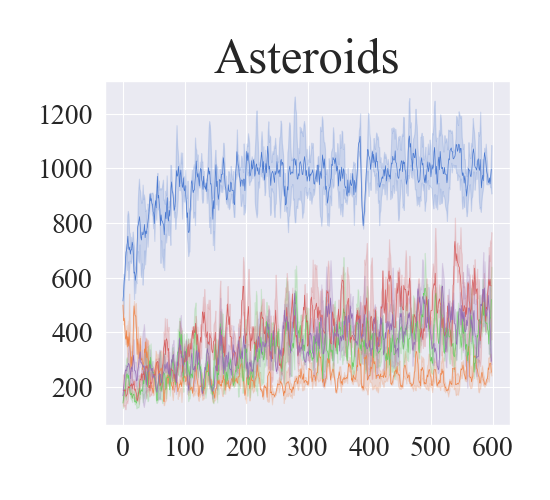}\\\
			\end{minipage}%
		}%
		\vspace{-1.0cm}
		
		\subfigure{
			\begin{minipage}[t]{0.166\linewidth}
				\centering
				\includegraphics[width=1.05in]{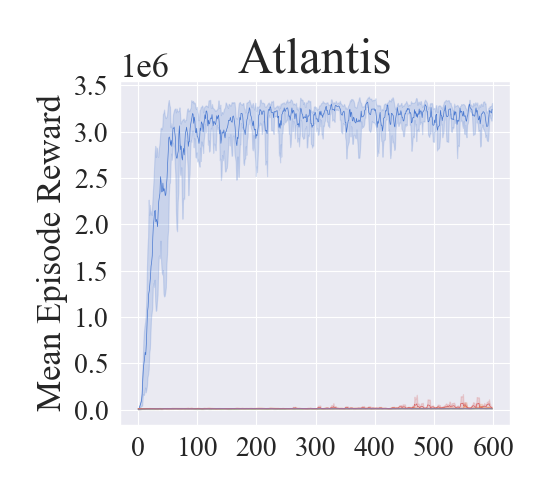}\\
			\end{minipage}%
		}%
		\subfigure{
			\begin{minipage}[t]{0.166\linewidth}
				\centering
				\includegraphics[width=1.05in]{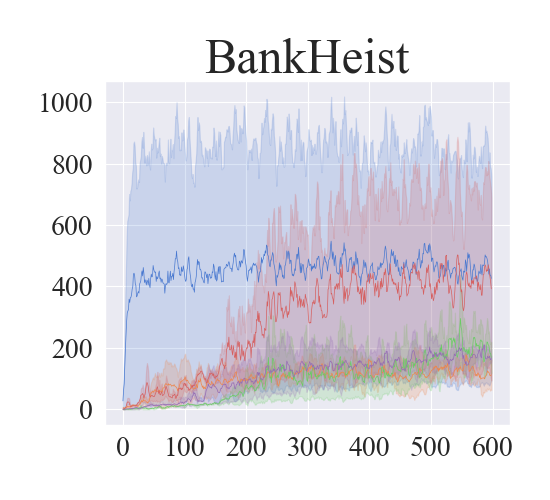}\\
			\end{minipage}%
		}%
		\subfigure{
			\begin{minipage}[t]{0.166\linewidth}
				\centering
				\includegraphics[width=1.05in]{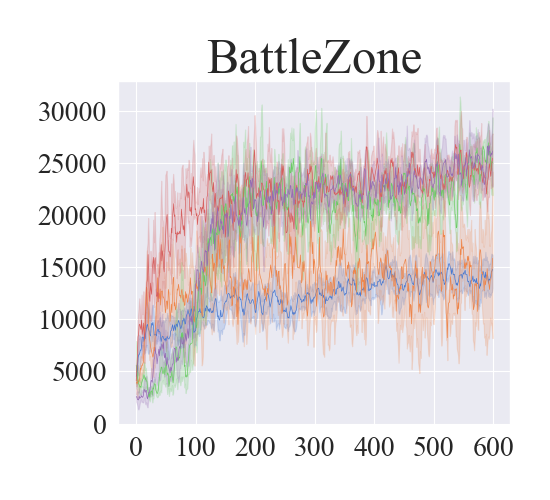}\\
			\end{minipage}%
		}%
		\subfigure{
			\begin{minipage}[t]{0.166\linewidth}
				\centering
				\includegraphics[width=1.05in]{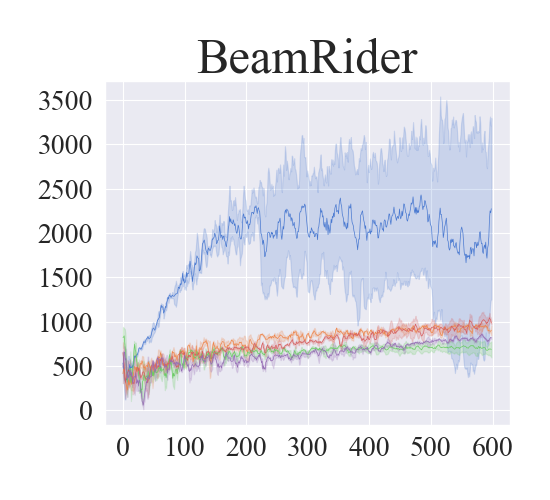}\\
			\end{minipage}%
		}%
		\subfigure{
			\begin{minipage}[t]{0.166\linewidth}
				\centering
				\includegraphics[width=1.05in]{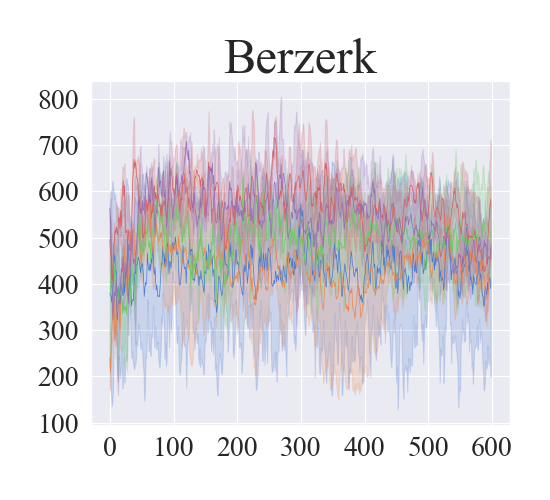}\\
			\end{minipage}%
		}%
		\subfigure{
			\begin{minipage}[t]{0.166\linewidth}
				\centering
				\includegraphics[width=1.05in]{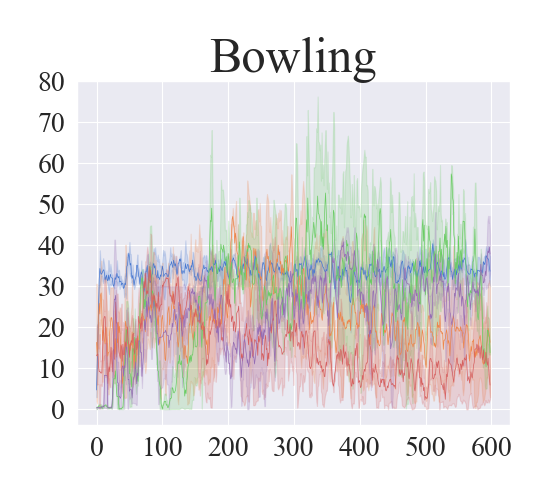}\\
			\end{minipage}%
		}%
		\vspace{-0.6cm}
		
		\subfigure{
			\begin{minipage}[t]{0.166\linewidth}
				\centering
				\includegraphics[width=1.05in]{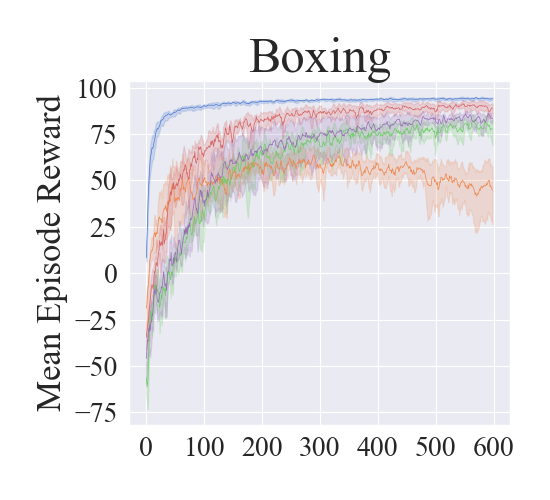}\\
			\end{minipage}%
		}%
		\subfigure{
			\begin{minipage}[t]{0.166\linewidth}
				\centering
				\includegraphics[width=1.05in]{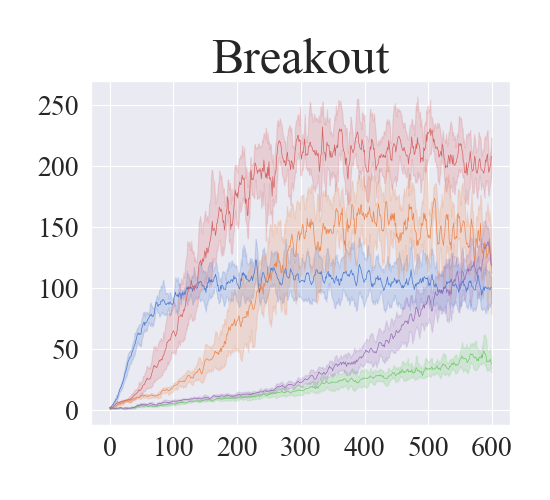}\\
			\end{minipage}%
		}%
		\subfigure{
			\begin{minipage}[t]{0.166\linewidth}
				\centering
				\includegraphics[width=1.05in]{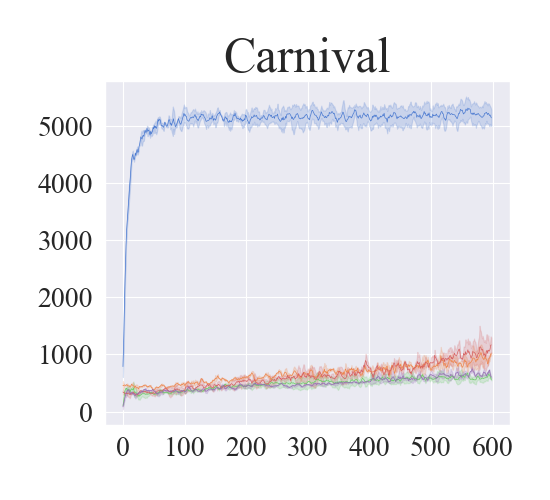}\\
			\end{minipage}%
		}%
		\subfigure{
			\begin{minipage}[t]{0.166\linewidth}
				\centering
				\includegraphics[width=1.05in]{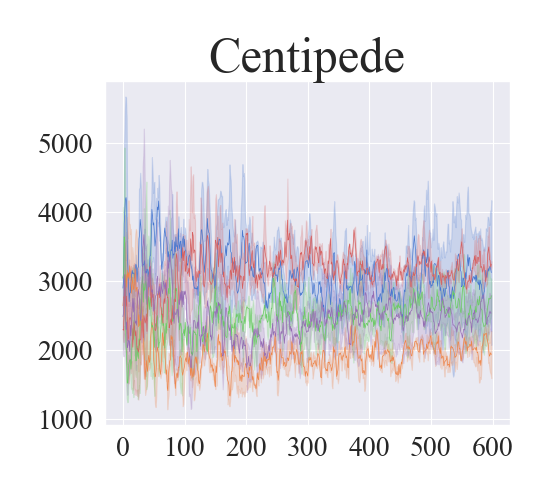}\\
			\end{minipage}%
		}%
		\subfigure{
			\begin{minipage}[t]{0.166\linewidth}
				\centering
				\includegraphics[width=1.05in]{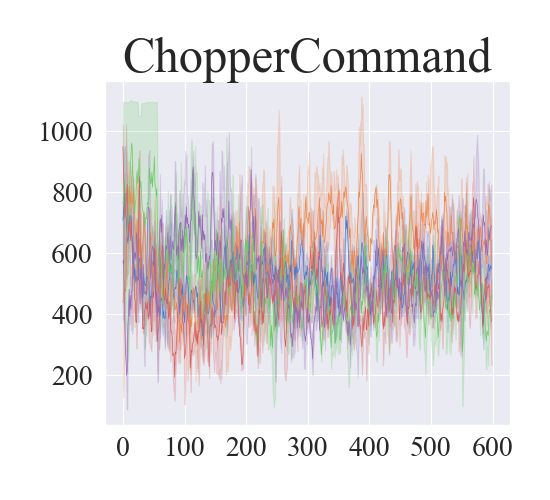}\\
			\end{minipage}%
		}%
		\subfigure{
			\begin{minipage}[t]{0.166\linewidth}
				\centering
				\includegraphics[width=1.05in]{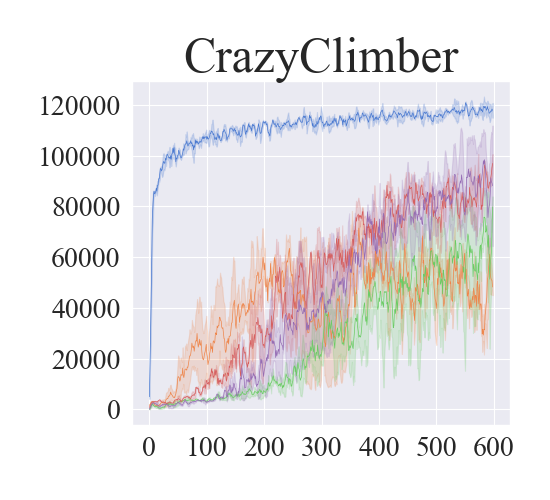}\\
			\end{minipage}%
		}%
		\vspace{-0.6cm}
		
		\subfigure{
			\begin{minipage}[t]{0.166\linewidth}
				\centering
				\includegraphics[width=1.05in]{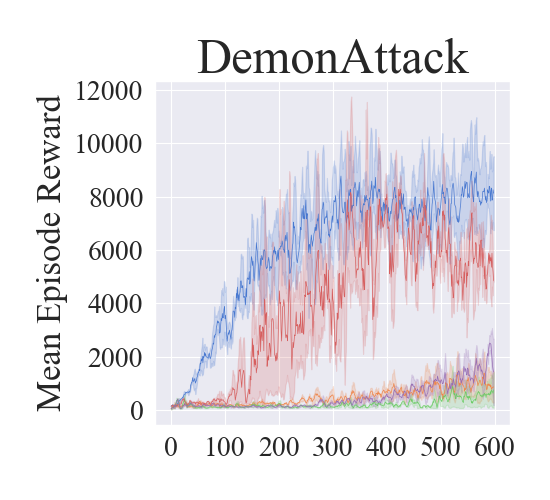}\\
			\end{minipage}%
		}%
		\subfigure{
			\begin{minipage}[t]{0.166\linewidth}
				\centering
				\includegraphics[width=1.05in]{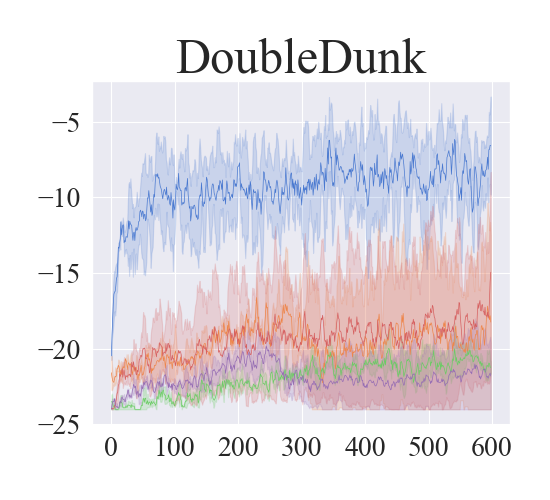}\\
			\end{minipage}%
		}%
		\subfigure{
			\begin{minipage}[t]{0.166\linewidth}
				\centering
				\includegraphics[width=1.05in]{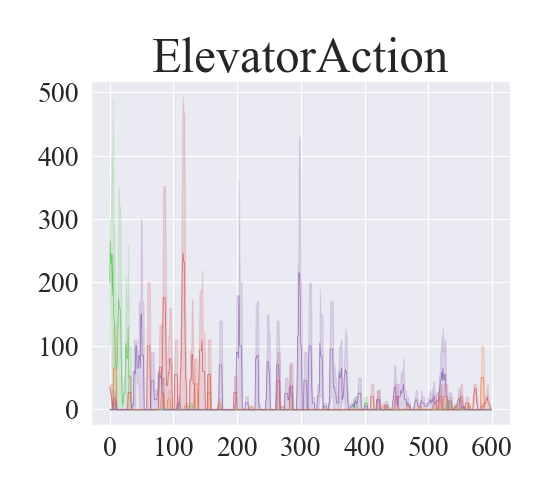}\\
			\end{minipage}%
		}%
		\subfigure{
			\begin{minipage}[t]{0.166\linewidth}
				\centering
				\includegraphics[width=1.05in]{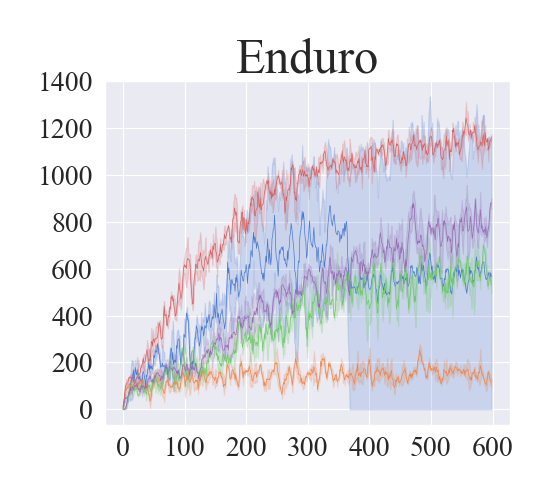}\\
			\end{minipage}%
		}%
		\subfigure{
			\begin{minipage}[t]{0.166\linewidth}
				\centering
				\includegraphics[width=1.05in]{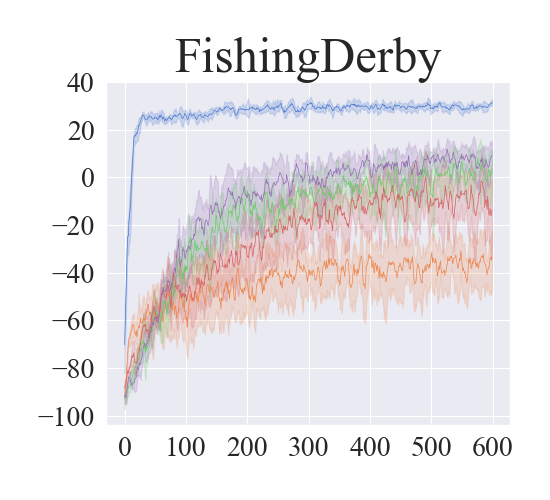}\\
			\end{minipage}%
		}%
		\subfigure{
			\begin{minipage}[t]{0.166\linewidth}
				\centering
				\includegraphics[width=1.05in]{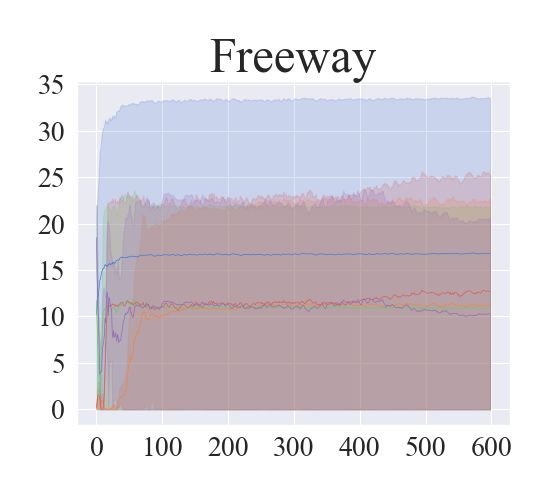}\\
			\end{minipage}%
		}%
		\vspace{-0.6cm}
		
		\subfigure{
			\begin{minipage}[t]{0.166\linewidth}
				\centering
				\includegraphics[width=1.05in]{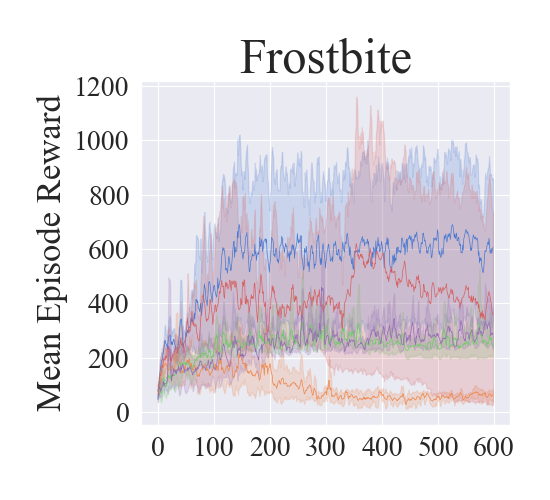}\\
			\end{minipage}%
		}%
		\subfigure{
			\begin{minipage}[t]{0.166\linewidth}
				\centering
				\includegraphics[width=1.05in]{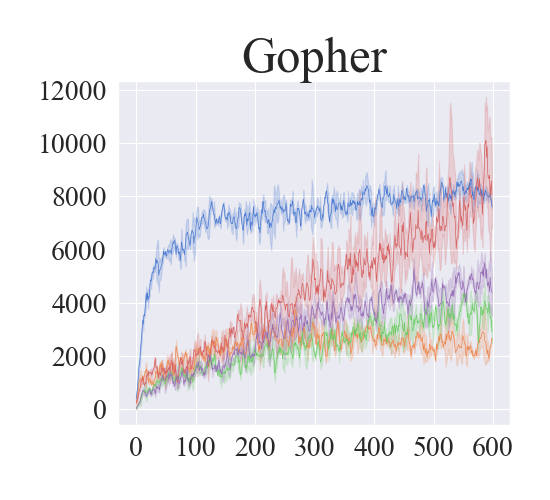}\\
			\end{minipage}%
		}%
		\subfigure{
			\begin{minipage}[t]{0.166\linewidth}
				\centering
				\includegraphics[width=1.05in]{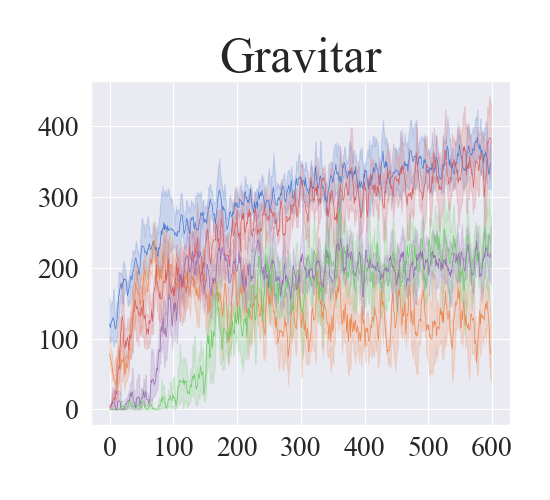}\\
			\end{minipage}%
		}%
		\subfigure{
			\begin{minipage}[t]{0.166\linewidth}
				\centering
				\includegraphics[width=1.05in]{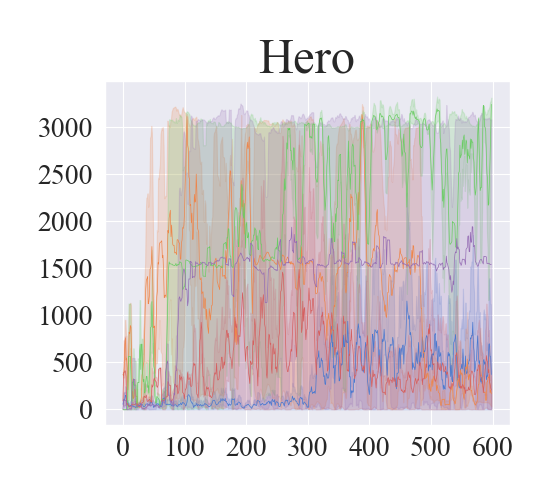}\\
			\end{minipage}%
		}%
		\subfigure{
			\begin{minipage}[t]{0.166\linewidth}
				\centering
				\includegraphics[width=1.05in]{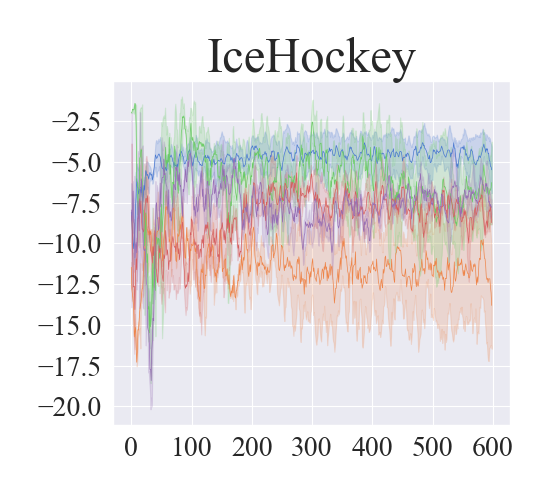}\\
			\end{minipage}%
		}%
		\subfigure{
			\begin{minipage}[t]{0.166\linewidth}
				\centering
				\includegraphics[width=1.05in]{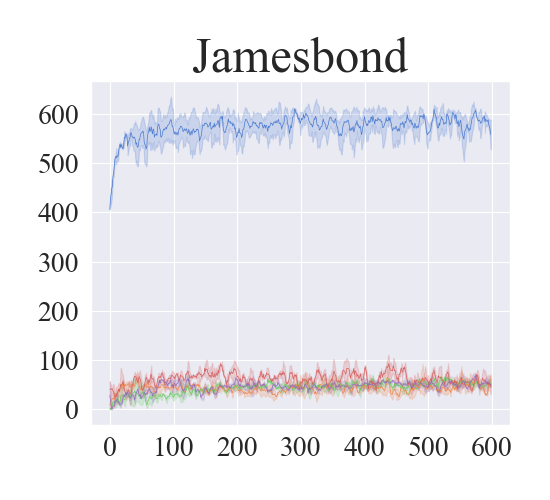}\\
			\end{minipage}%
		}%
		\vspace{-0.6cm}
		
		\subfigure{
			\begin{minipage}[t]{0.166\linewidth}
				\centering
				\includegraphics[width=1.05in]{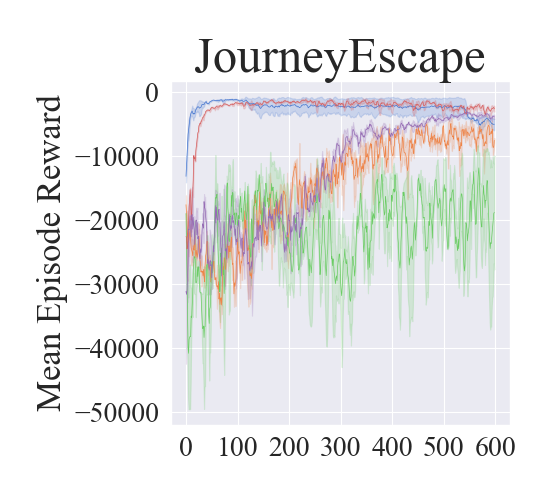}\\
			\end{minipage}%
		}%
		\subfigure{
			\begin{minipage}[t]{0.166\linewidth}
				\centering
				\includegraphics[width=1.05in]{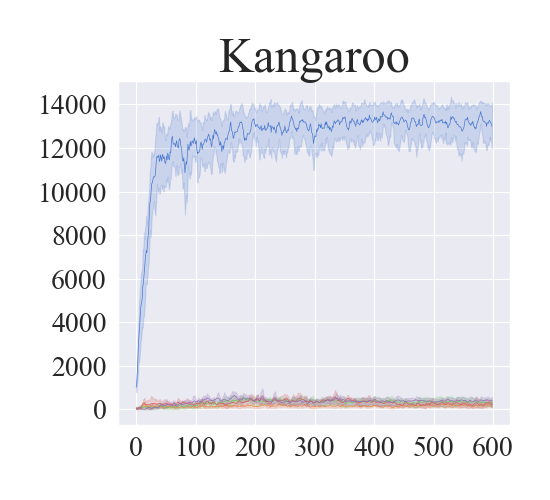}\\
			\end{minipage}%
		}%
		\subfigure{
			\begin{minipage}[t]{0.166\linewidth}
				\centering
				\includegraphics[width=1.05in]{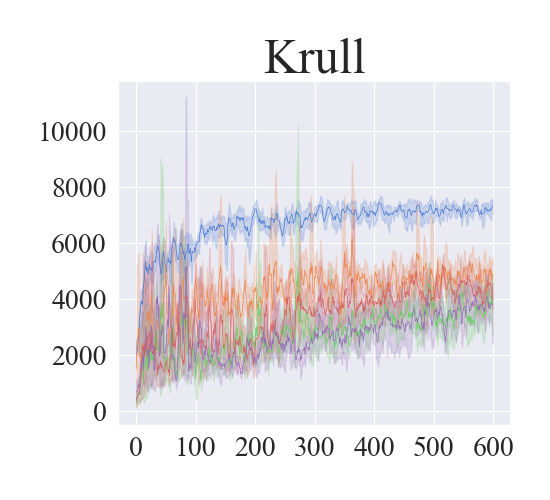}\\
			\end{minipage}%
		}%
		\subfigure{
			\begin{minipage}[t]{0.166\linewidth}
				\centering
				\includegraphics[width=1.05in]{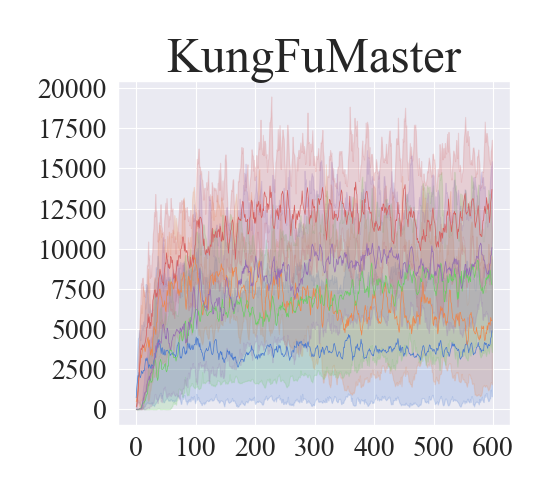}\\
			\end{minipage}%
		}%
		\subfigure{
			\begin{minipage}[t]{0.166\linewidth}
				\centering
				\includegraphics[width=1.05in]{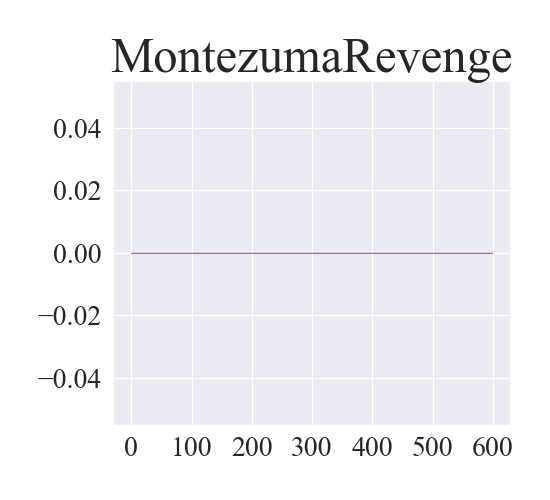}\\
			\end{minipage}%
		}%
		\subfigure{
			\begin{minipage}[t]{0.166\linewidth}
				\centering
				\includegraphics[width=1.05in]{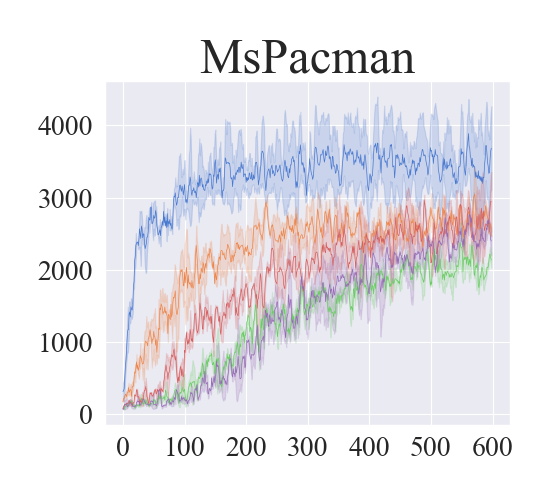}\\
			\end{minipage}%
		}%
		\vspace{-0.6cm}
		
		\subfigure{
			\begin{minipage}[t]{0.166\linewidth}
				\centering
				\includegraphics[width=1.05in]{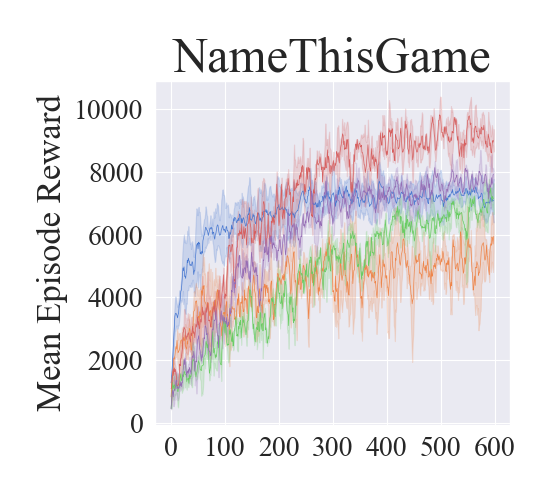}\\
			\end{minipage}%
		}%
		\subfigure{
			\begin{minipage}[t]{0.166\linewidth}
				\centering
				\includegraphics[width=1.05in]{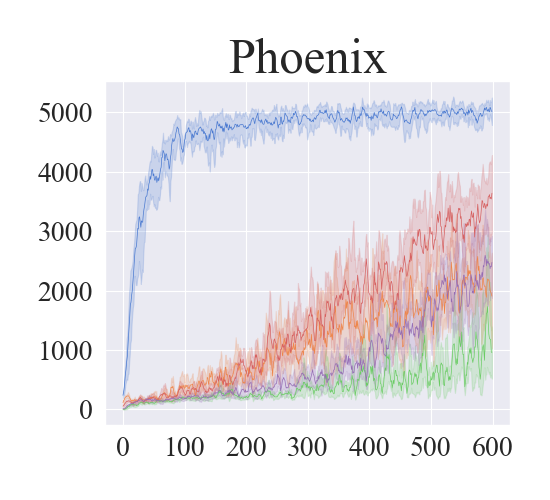}\\
			\end{minipage}%
		}%
		\subfigure{
			\begin{minipage}[t]{0.166\linewidth}
				\centering
				\includegraphics[width=1.05in]{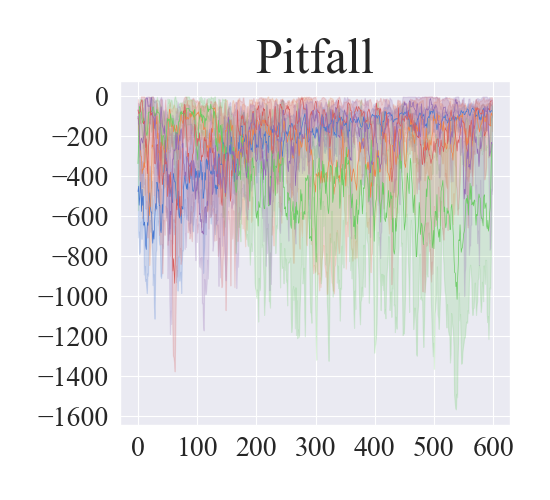}\\
			\end{minipage}%
		}%
		\subfigure{
			\begin{minipage}[t]{0.166\linewidth}
				\centering
				\includegraphics[width=1.05in]{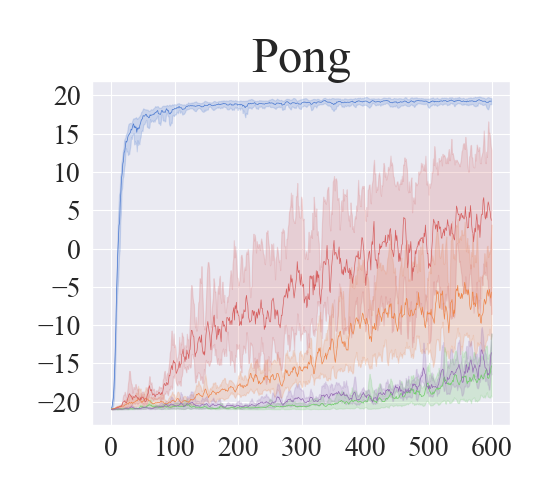}\\
			\end{minipage}%
		}%
		\subfigure{
			\begin{minipage}[t]{0.166\linewidth}
				\centering
				\includegraphics[width=1.05in]{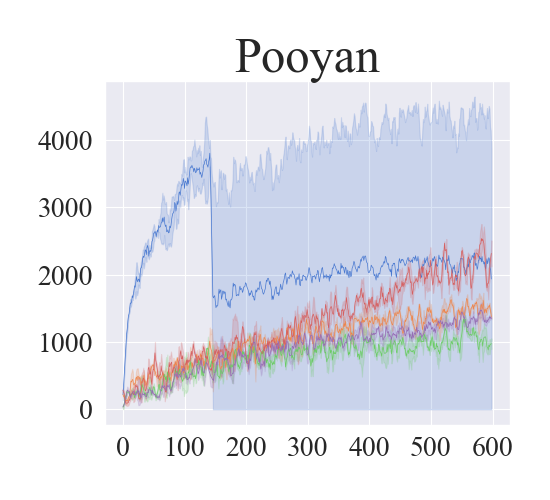}\\
			\end{minipage}%
		}%
		\subfigure{
			\begin{minipage}[t]{0.166\linewidth}
				\centering
				\includegraphics[width=1.05in]{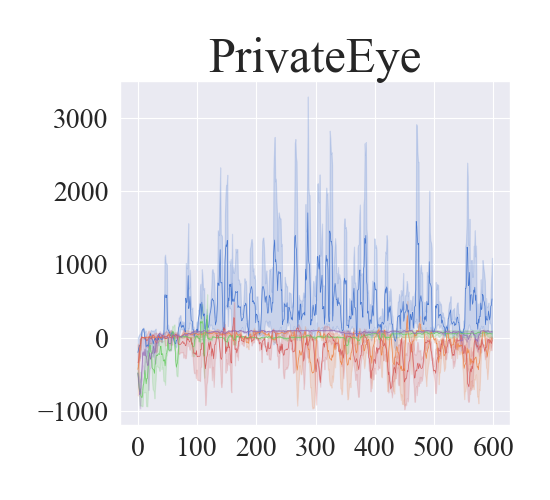}\\
			\end{minipage}%
		}%
		\vspace{-0.6cm}
		
		\subfigure{
			\begin{minipage}[t]{0.166\linewidth}
				\centering
				\includegraphics[width=1.05in]{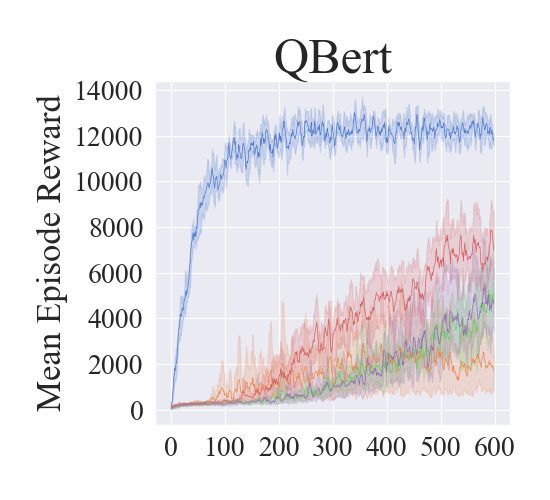}\\
			\end{minipage}%
		}%
		\subfigure{
			\begin{minipage}[t]{0.166\linewidth}
				\centering
				\includegraphics[width=1.05in]{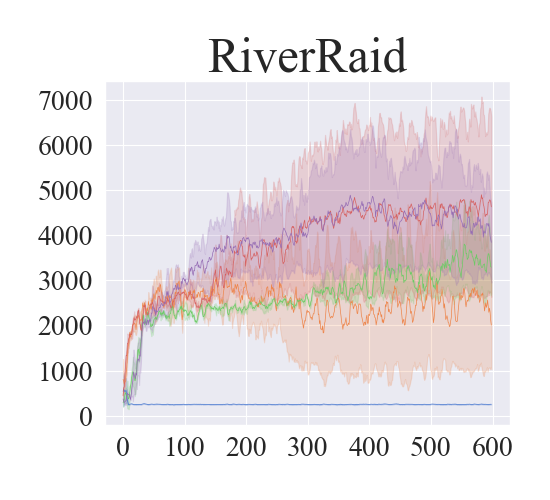}\\
			\end{minipage}%
		}%
		\subfigure{
			\begin{minipage}[t]{0.166\linewidth}
				\centering
				\includegraphics[width=1.05in]{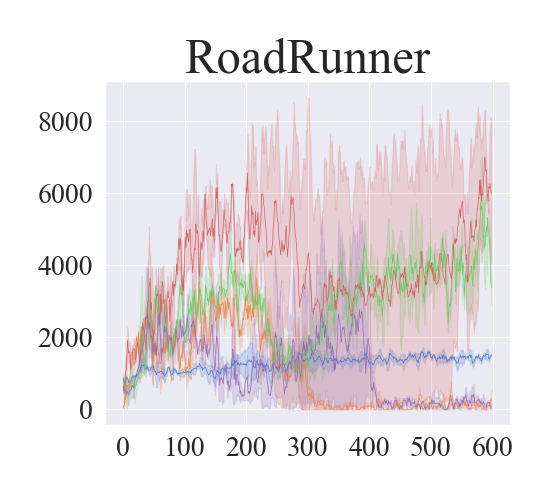}\\
			\end{minipage}%
		}%
		\subfigure{
			\begin{minipage}[t]{0.166\linewidth}
				\centering
				\includegraphics[width=1.05in]{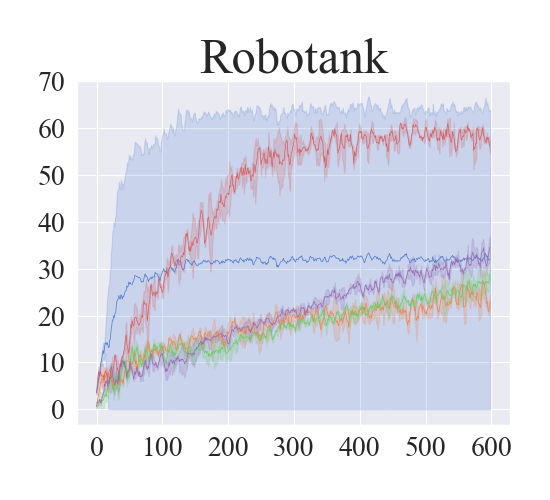}\\
			\end{minipage}%
		}%
		\subfigure{
			\begin{minipage}[t]{0.166\linewidth}
				\centering
				\includegraphics[width=1.05in]{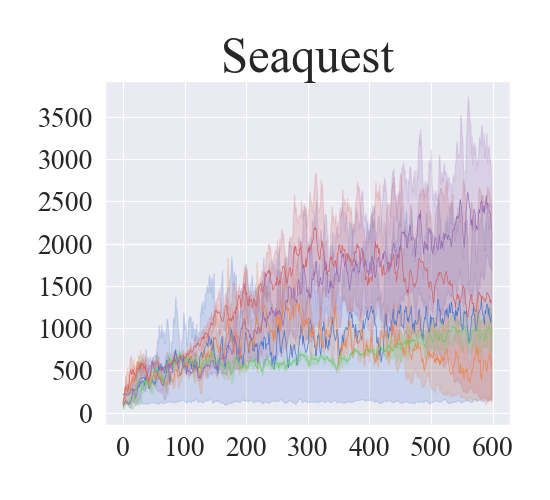}\\
			\end{minipage}%
		}%
		\subfigure{
			\begin{minipage}[t]{0.166\linewidth}
				\centering
				\includegraphics[width=1.05in]{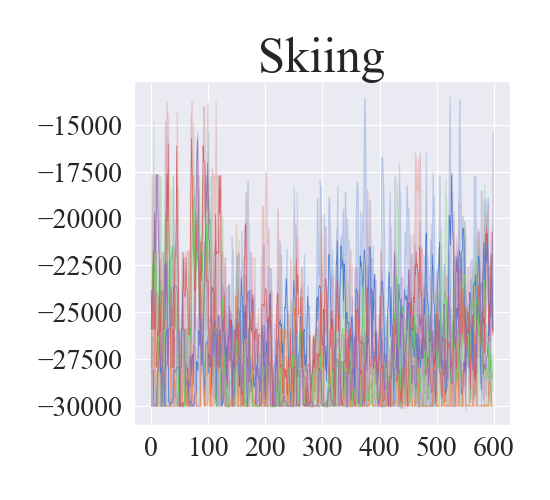}\\
			\end{minipage}%
		}%
		\vspace{-0.6cm}
		
		\subfigure{
			\begin{minipage}[t]{0.166\linewidth}
				\centering
				\includegraphics[width=1.05in]{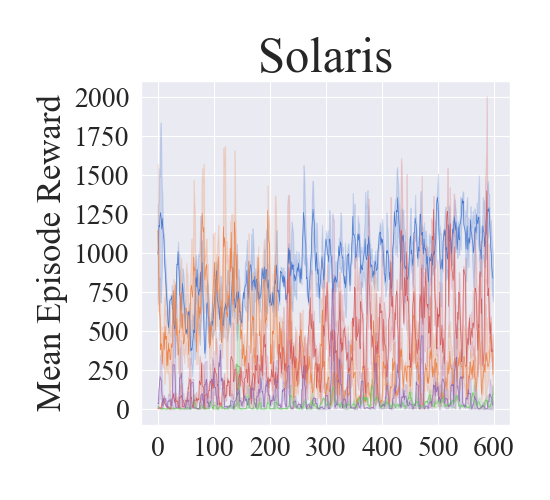}\\
			\end{minipage}%
		}%
		\subfigure{
			\begin{minipage}[t]{0.166\linewidth}
				\centering
				\includegraphics[width=1.05in]{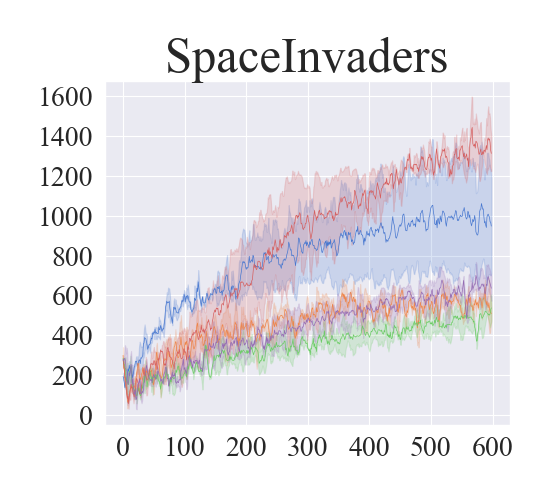}\\
			\end{minipage}%
		}%
		\subfigure{
			\begin{minipage}[t]{0.166\linewidth}
				\centering
				\includegraphics[width=1.05in]{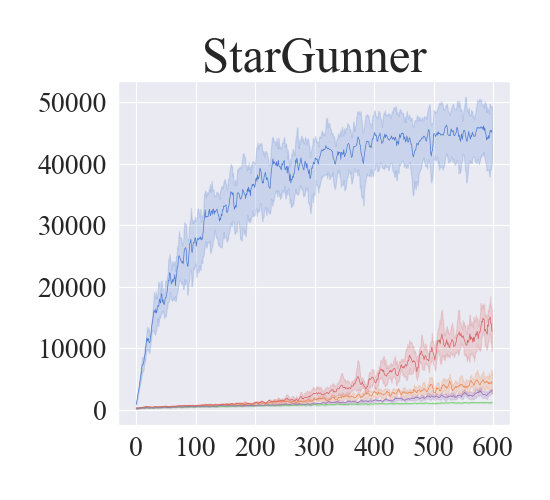}\\
			\end{minipage}%
		}%
		\subfigure{
			\begin{minipage}[t]{0.166\linewidth}
				\centering
				\includegraphics[width=1.05in]{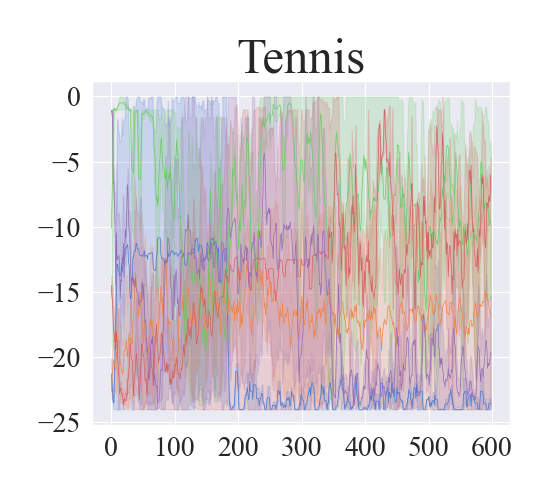}\\
			\end{minipage}%
		}%
		\subfigure{
			\begin{minipage}[t]{0.166\linewidth}
				\centering
				\includegraphics[width=1.05in]{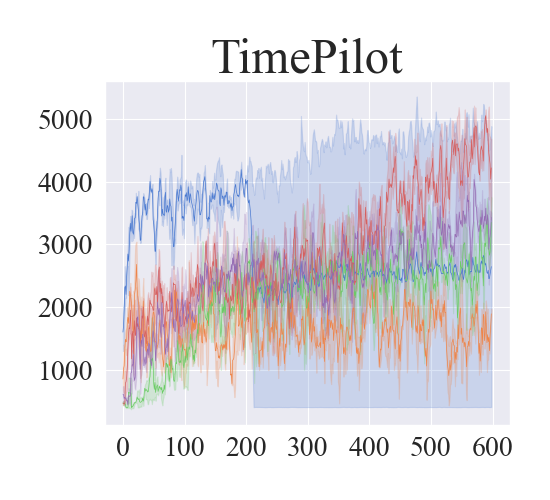}\\
			\end{minipage}%
		}%
		\subfigure{
			\begin{minipage}[t]{0.166\linewidth}
				\centering
				\includegraphics[width=1.05in]{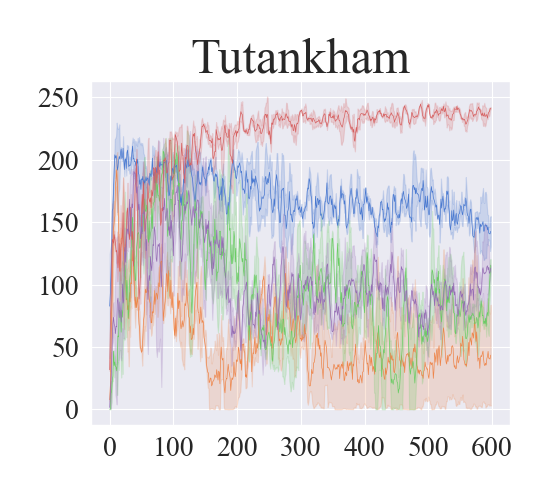}\\
			\end{minipage}%
		}%
		\vspace{-0.6cm}
		
		\subfigure{
			\begin{minipage}[t]{0.166\linewidth}
				\centering
				\includegraphics[width=1.05in]{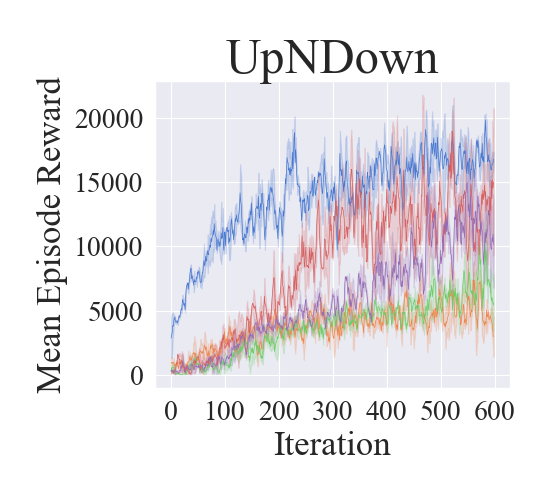}\\
			\end{minipage}%
		}%
		\subfigure{
			\begin{minipage}[t]{0.166\linewidth}
				\centering
				\includegraphics[width=1.05in]{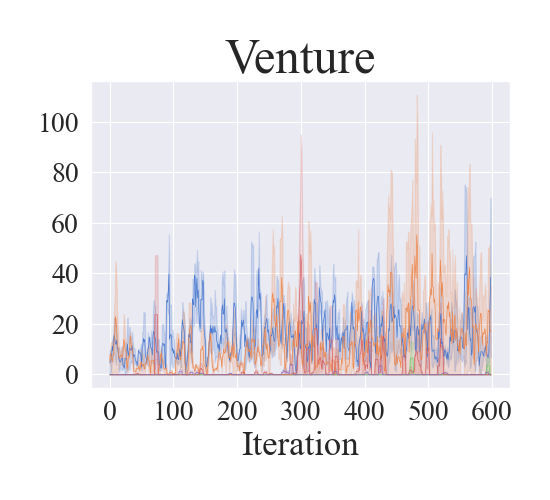}\\
			\end{minipage}%
		}%
		\subfigure{
			\begin{minipage}[t]{0.166\linewidth}
				\centering
				\includegraphics[width=1.05in]{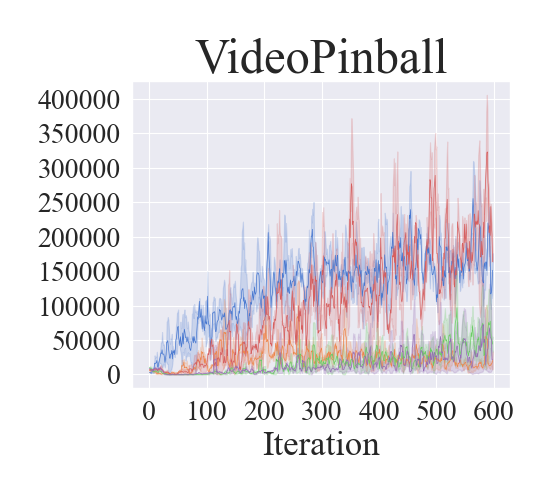}\\
			\end{minipage}%
		}%
		\subfigure{
			\begin{minipage}[t]{0.166\linewidth}
				\centering
				\includegraphics[width=1.05in]{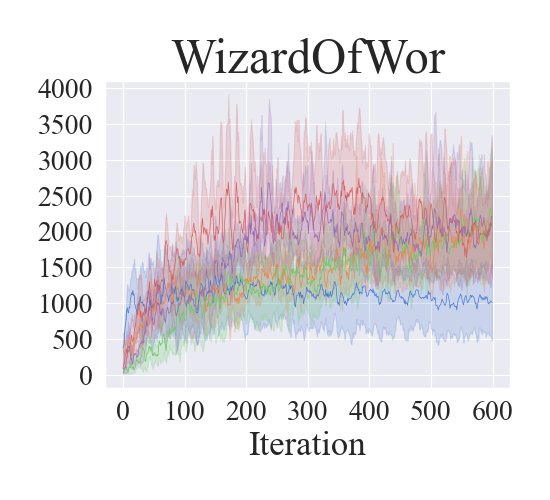}\\
			\end{minipage}%
		}%
		\subfigure{
			\begin{minipage}[t]{0.166\linewidth}
				\centering
				\includegraphics[width=1.05in]{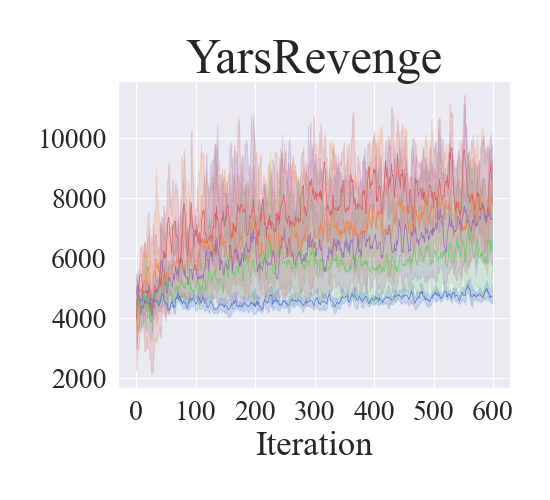}\\
			\end{minipage}%
		}%
		\subfigure{
			\begin{minipage}[t]{0.166\linewidth}
				\centering
				\includegraphics[width=1.05in]{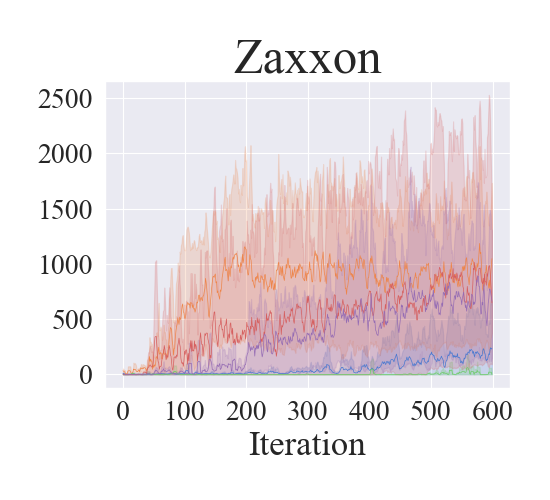}\\
			\end{minipage}%
		}%

		\centering
		\caption{\textbf{Learning curves of all $60$ Atari $2600$ games on high dataset}}
		\label{fig: Learning curves of all $60$ Atari $2600$ games on high dataset}
								
	\end{figure*}

\begin{figure*}[!htb]
		\centering

		\subfigure{
			\begin{minipage}[t]{\linewidth}
				\centering
				\includegraphics[width=4in]{legend2.png}\\
			\end{minipage}%
		}%
		
		\subfigure{
			\begin{minipage}[t]{0.166\linewidth}
				\centering
				\includegraphics[width=1.05in]{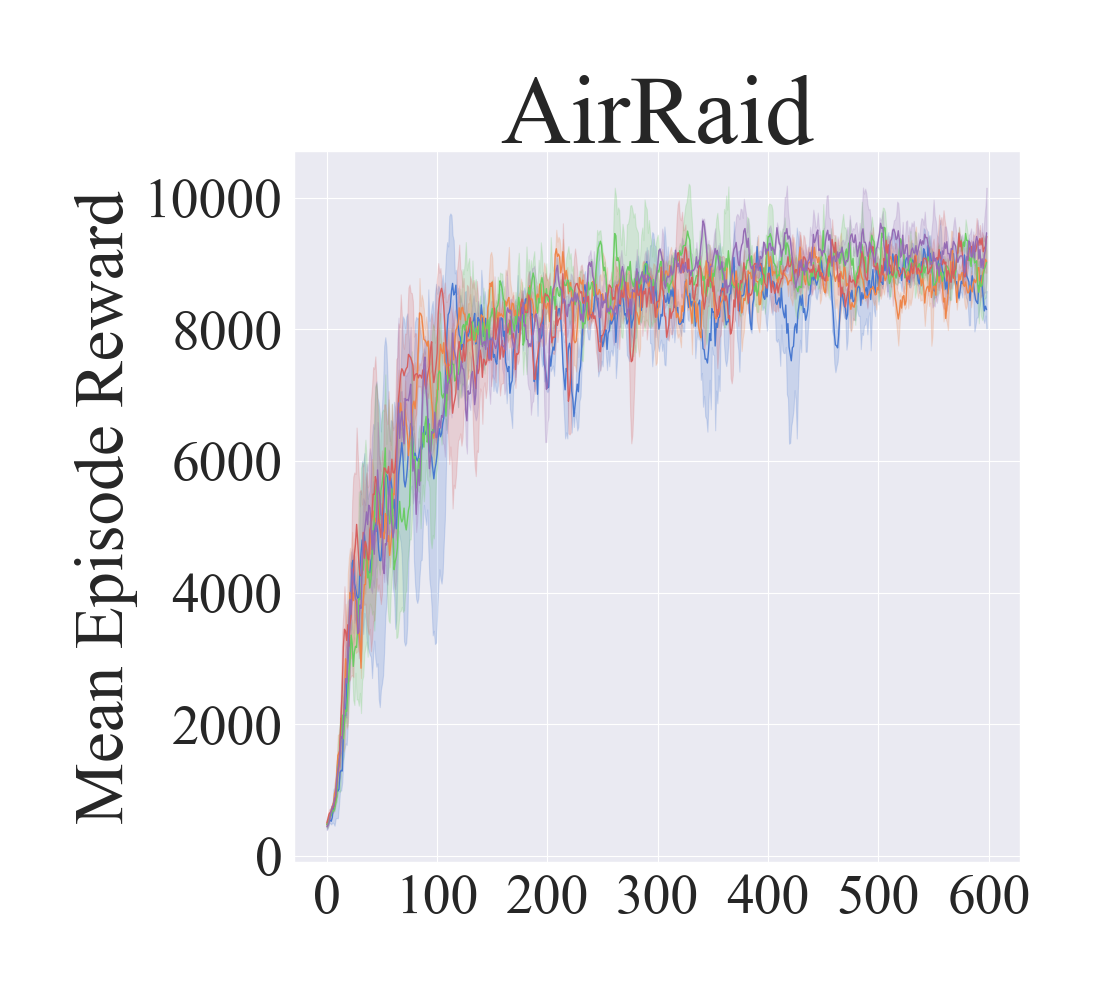}\\
			\end{minipage}%
		}%
		\subfigure{
			\begin{minipage}[t]{0.166\linewidth}
				\centering
				\includegraphics[width=1.05in]{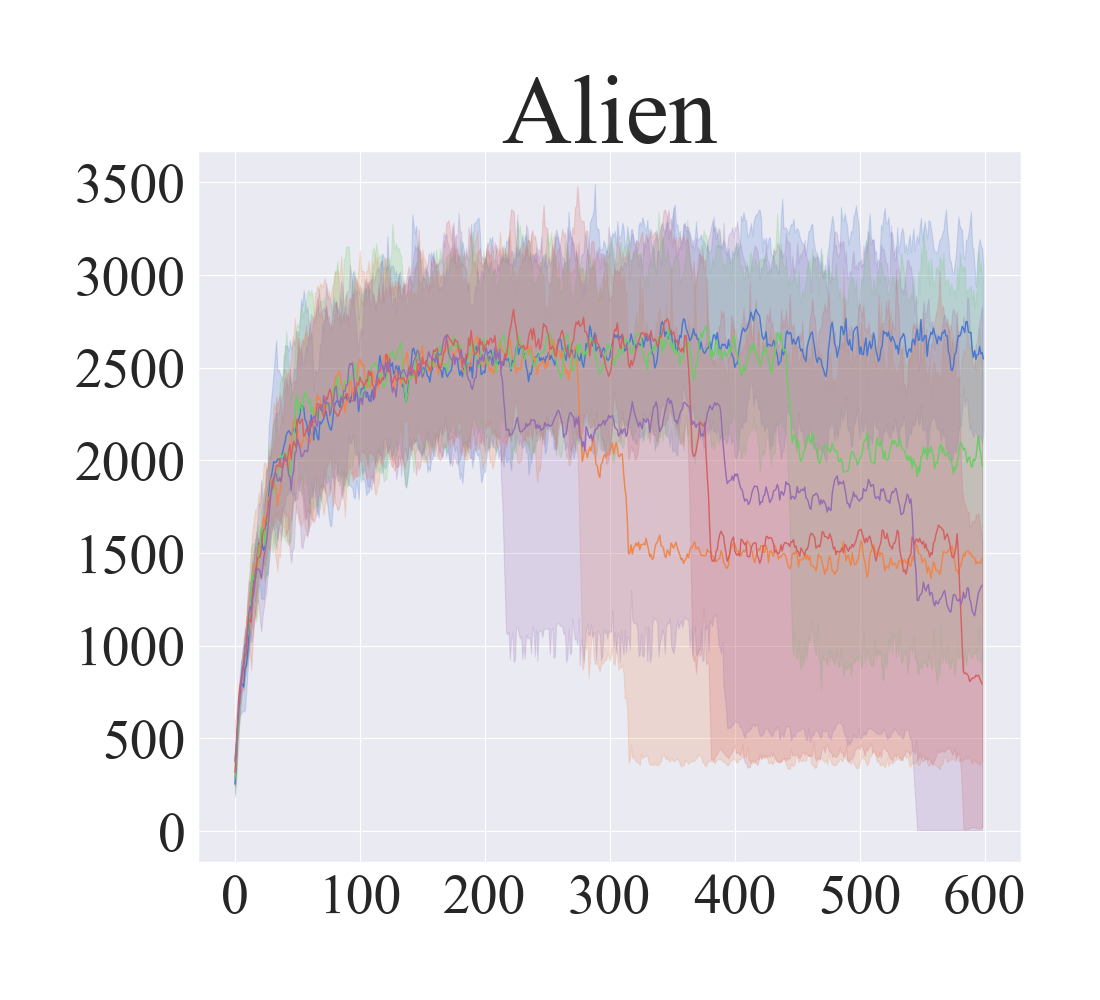}\\
			\end{minipage}%
		}%
		\subfigure{
			\begin{minipage}[t]{0.166\linewidth}
				\centering
				\includegraphics[width=1.05in]{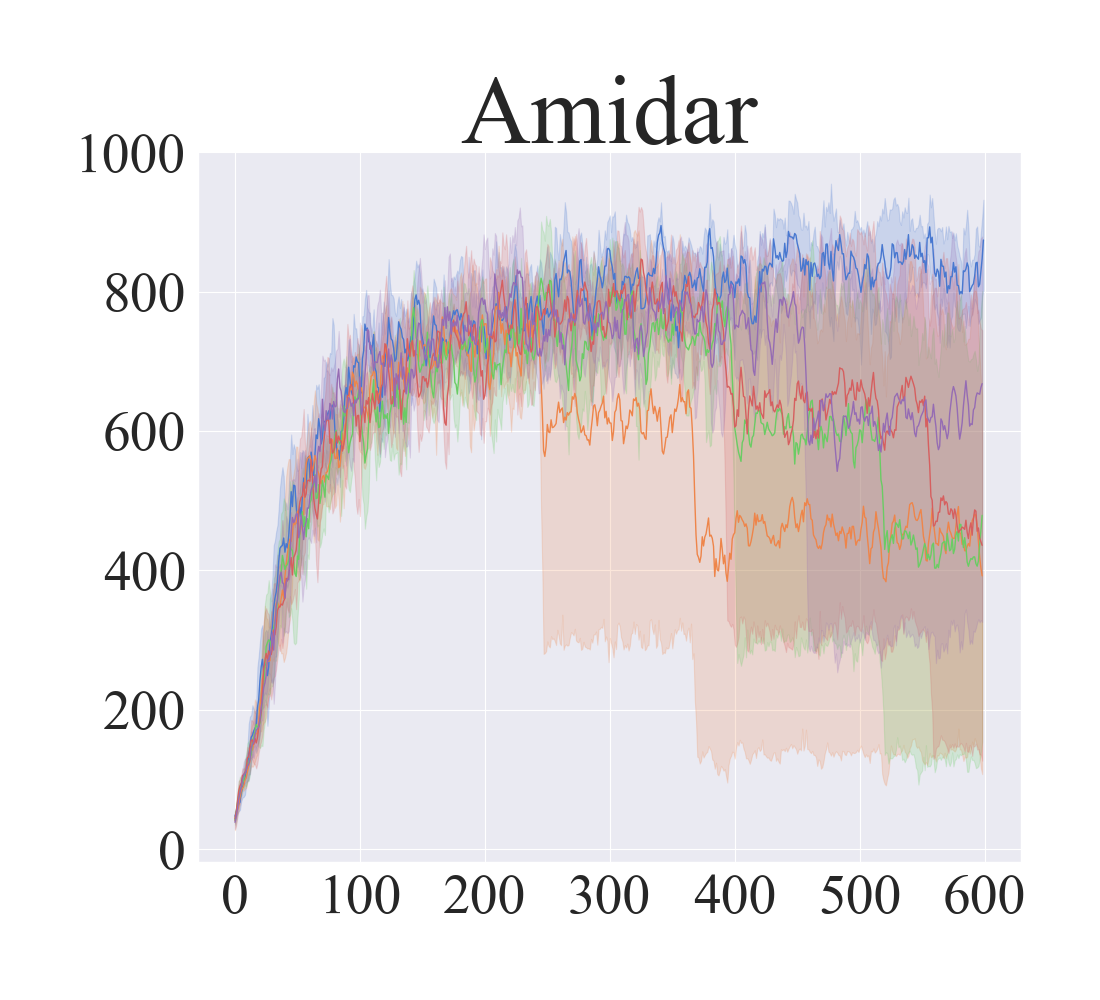}\\
			\end{minipage}%
		}%
		\subfigure{
			\begin{minipage}[t]{0.166\linewidth}
				\centering
				\includegraphics[width=1.05in]{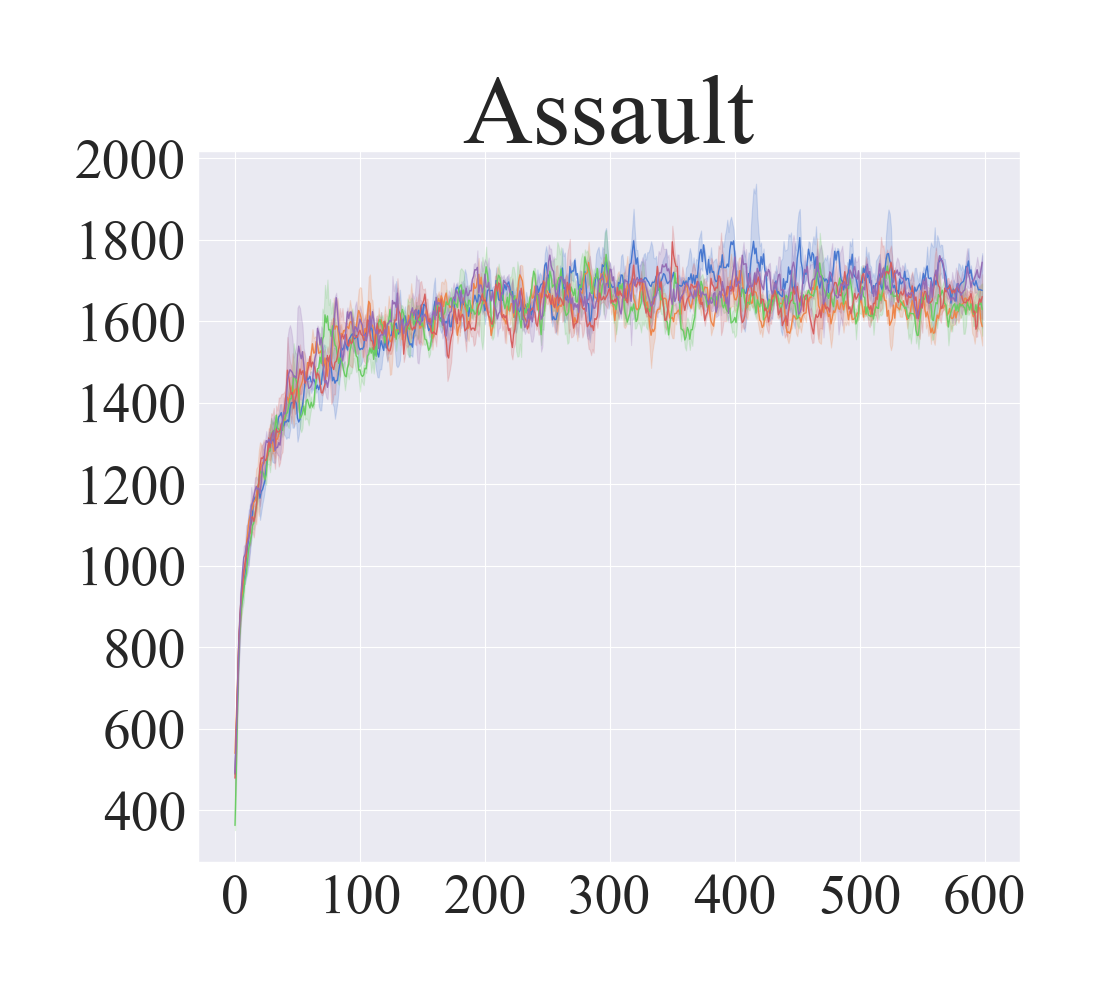}\\
			\end{minipage}%
		}%
		\subfigure{
			\begin{minipage}[t]{0.166\linewidth}
				\centering
				\includegraphics[width=1.05in]{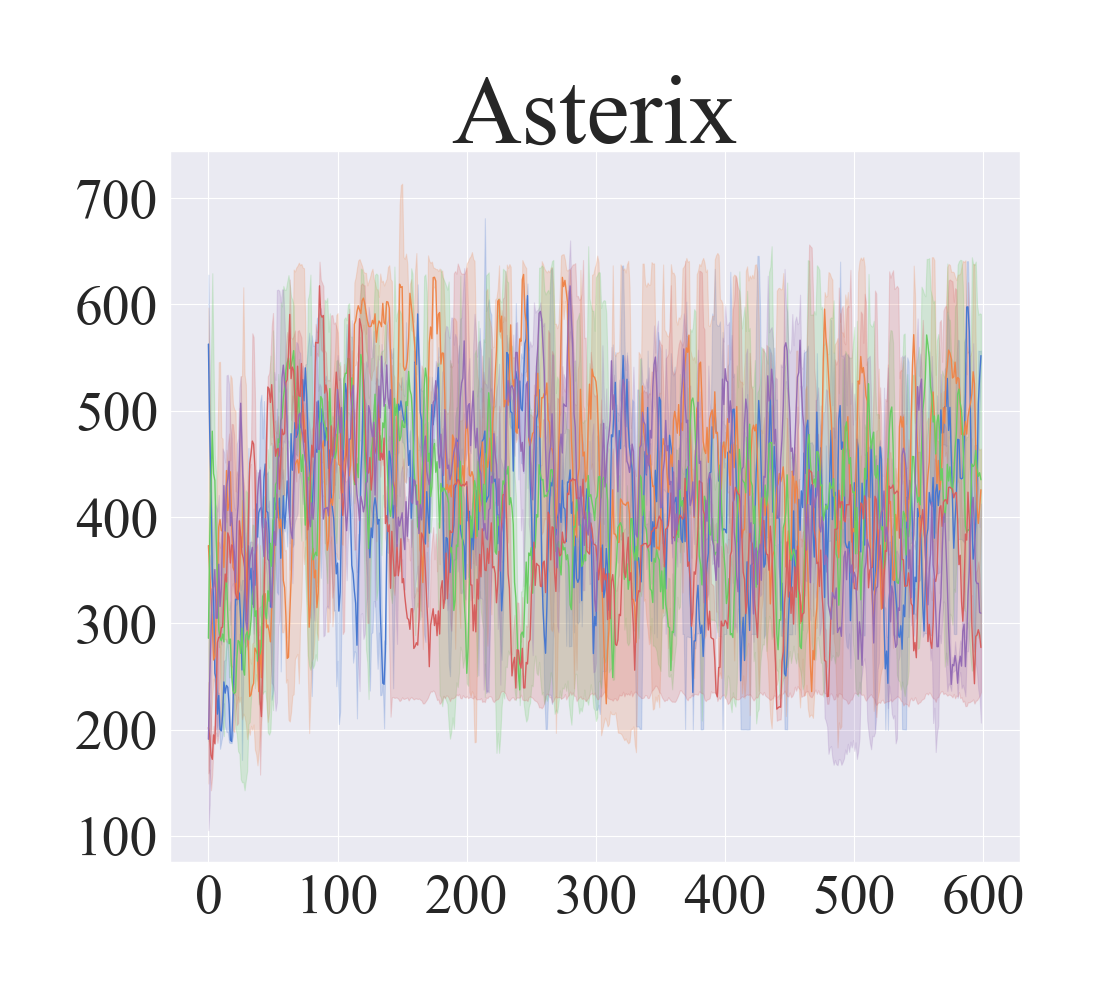}\\
			\end{minipage}%
		}%
		\subfigure{
			\begin{minipage}[t]{0.166\linewidth}
				\centering
				\includegraphics[width=1.05in]{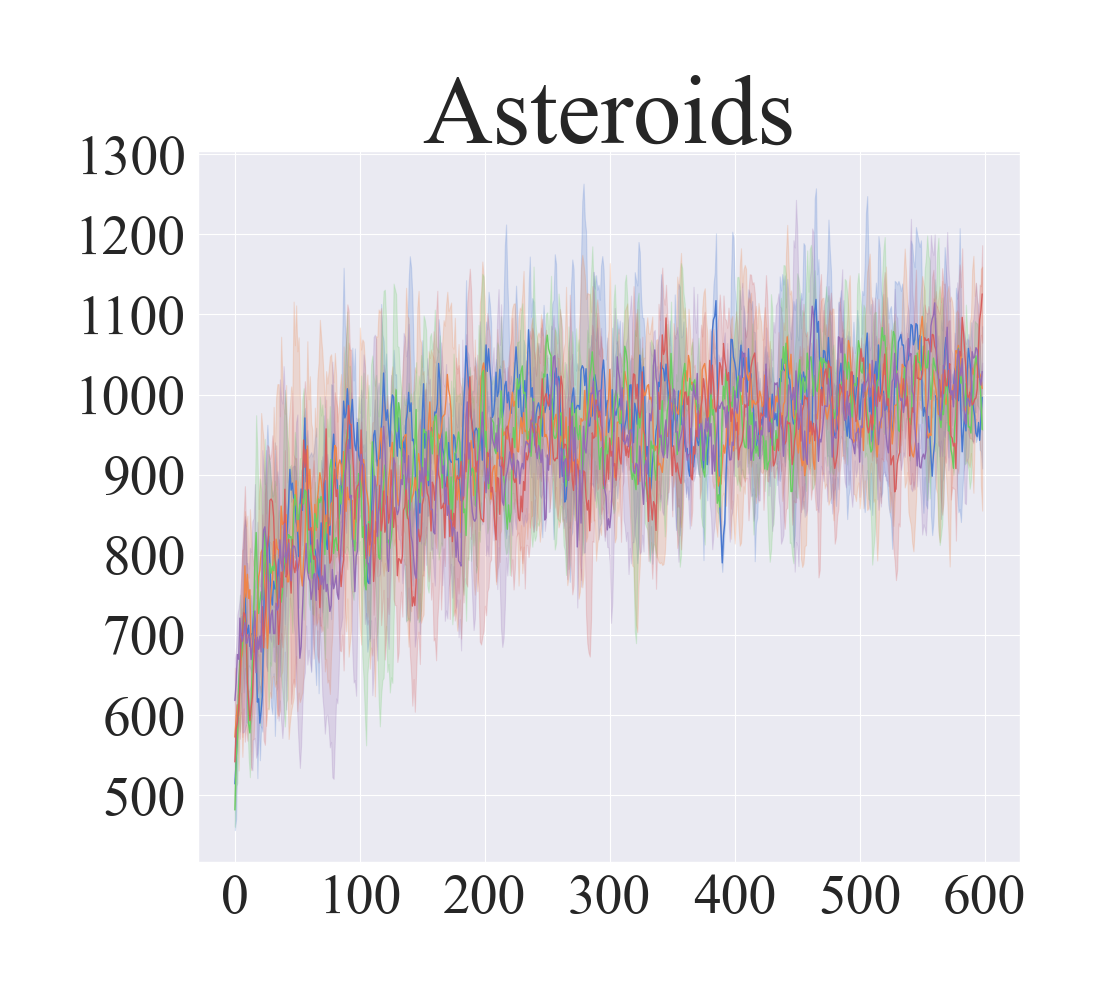}\\\
			\end{minipage}%
		}%
		\vspace{-1.0cm}
		
		\subfigure{
			\begin{minipage}[t]{0.166\linewidth}
				\centering
				\includegraphics[width=1.05in]{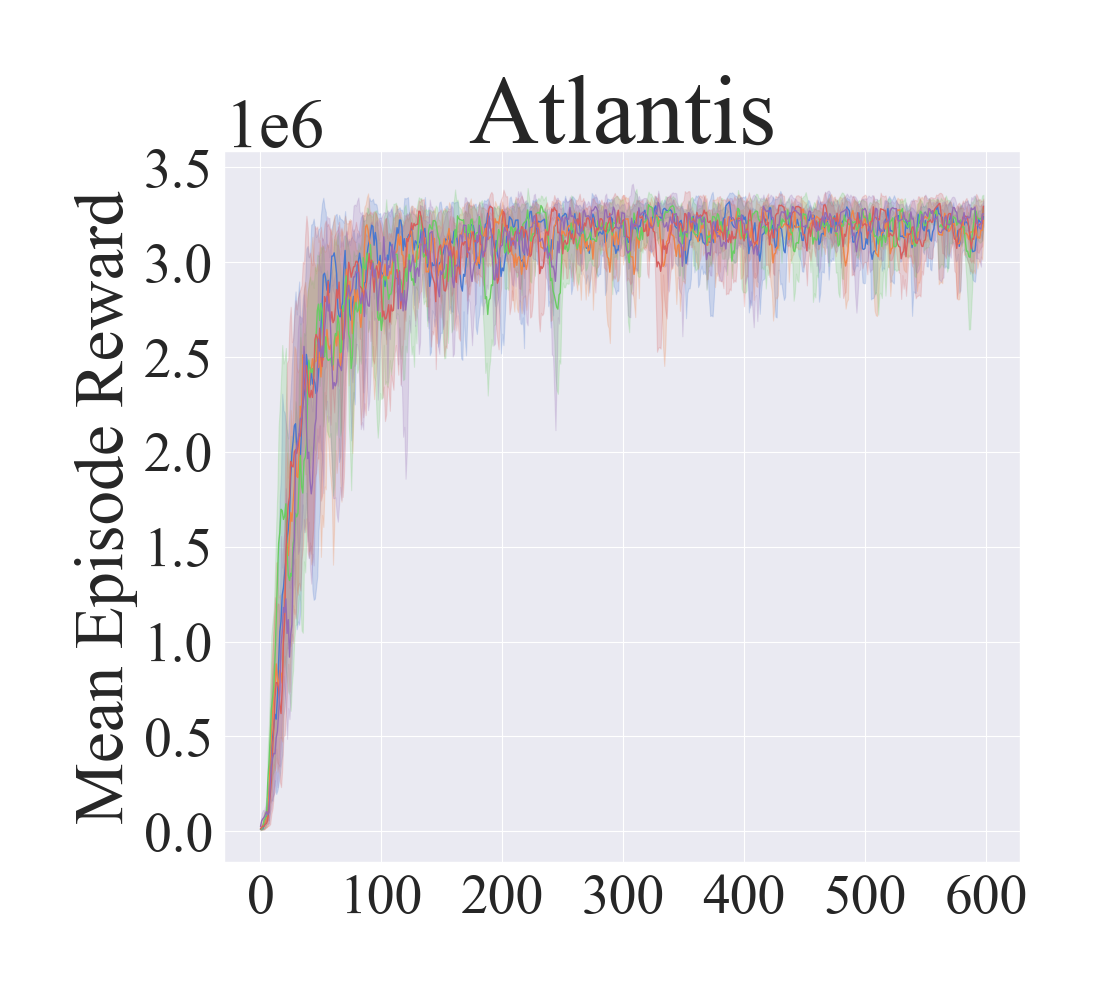}\\
			\end{minipage}%
		}%
		\subfigure{
			\begin{minipage}[t]{0.166\linewidth}
				\centering
				\includegraphics[width=1.05in]{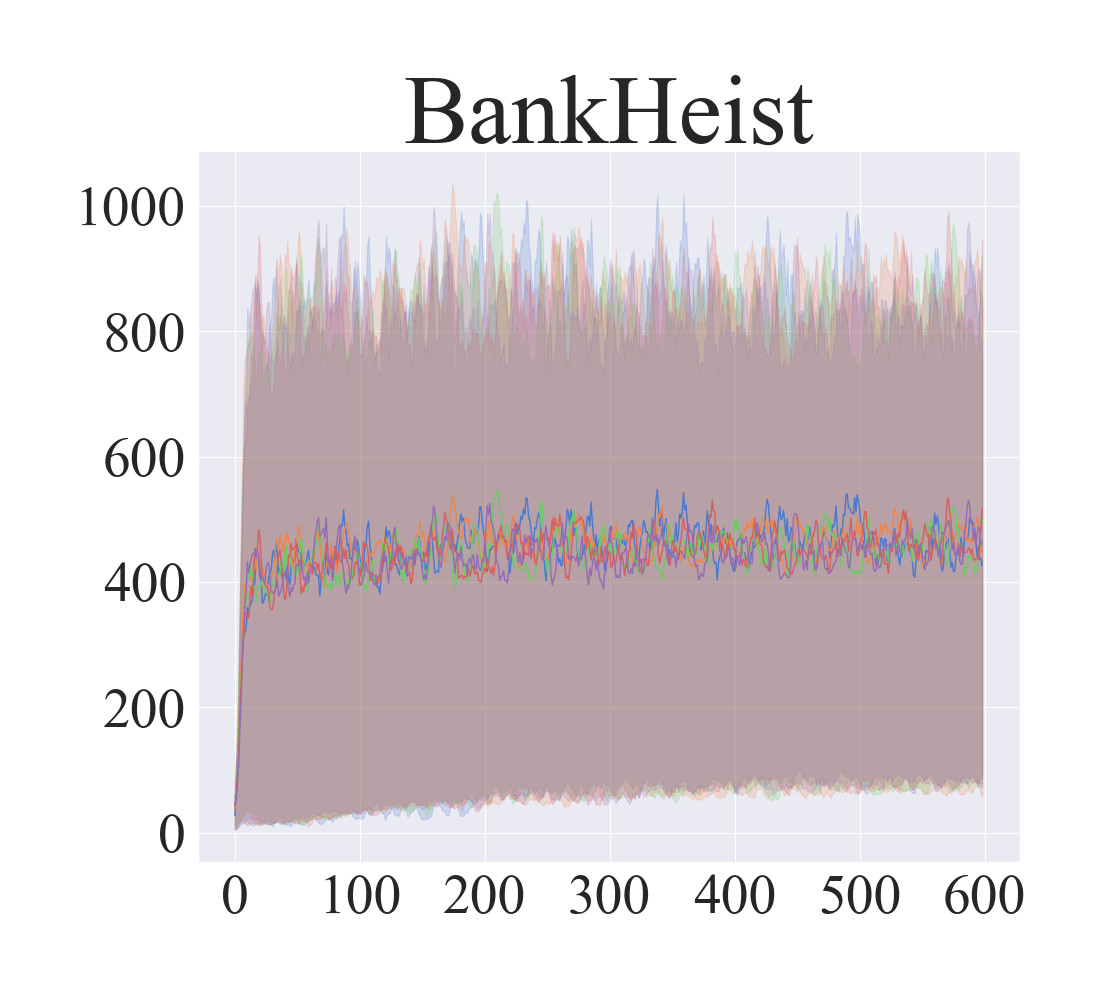}\\
			\end{minipage}%
		}%
		\subfigure{
			\begin{minipage}[t]{0.166\linewidth}
				\centering
				\includegraphics[width=1.05in]{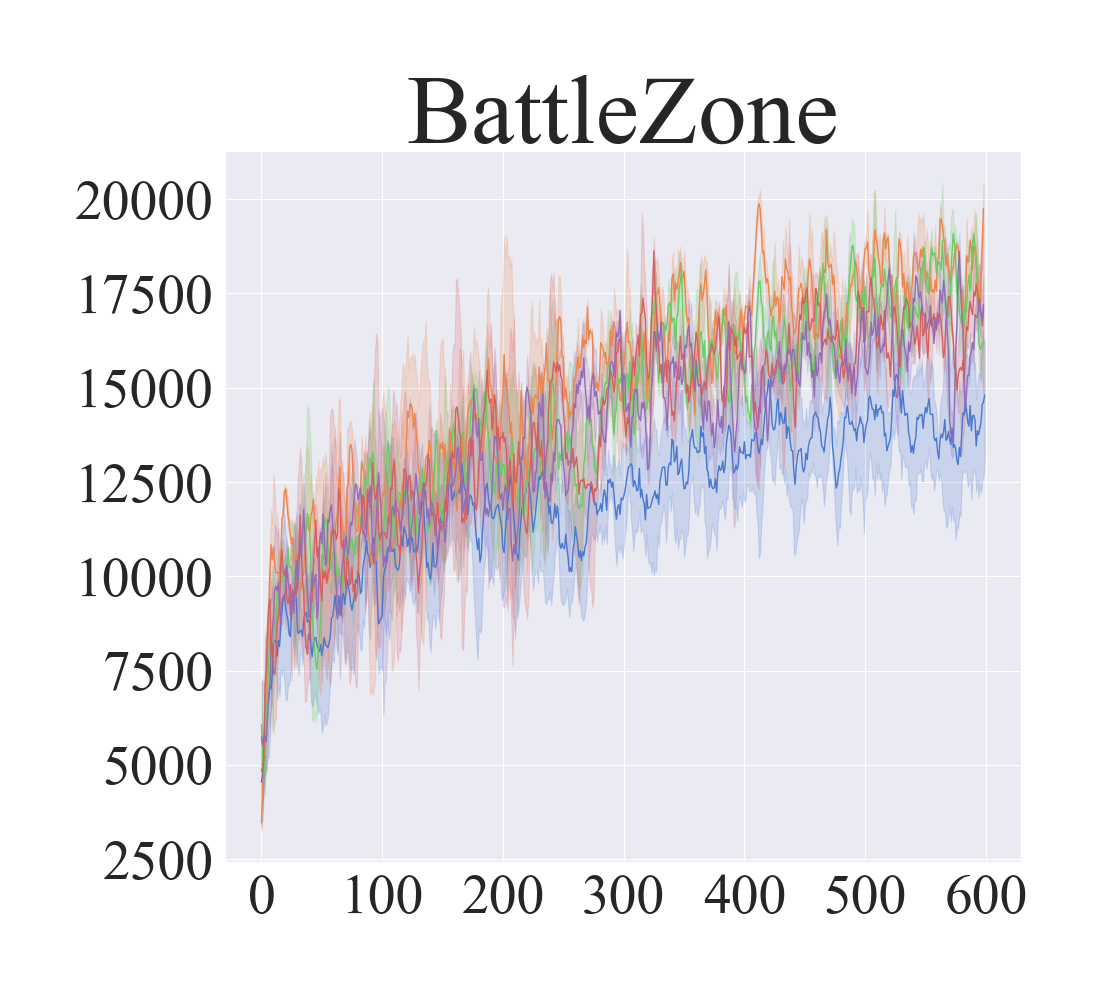}\\
			\end{minipage}%
		}%
		\subfigure{
			\begin{minipage}[t]{0.166\linewidth}
				\centering
				\includegraphics[width=1.05in]{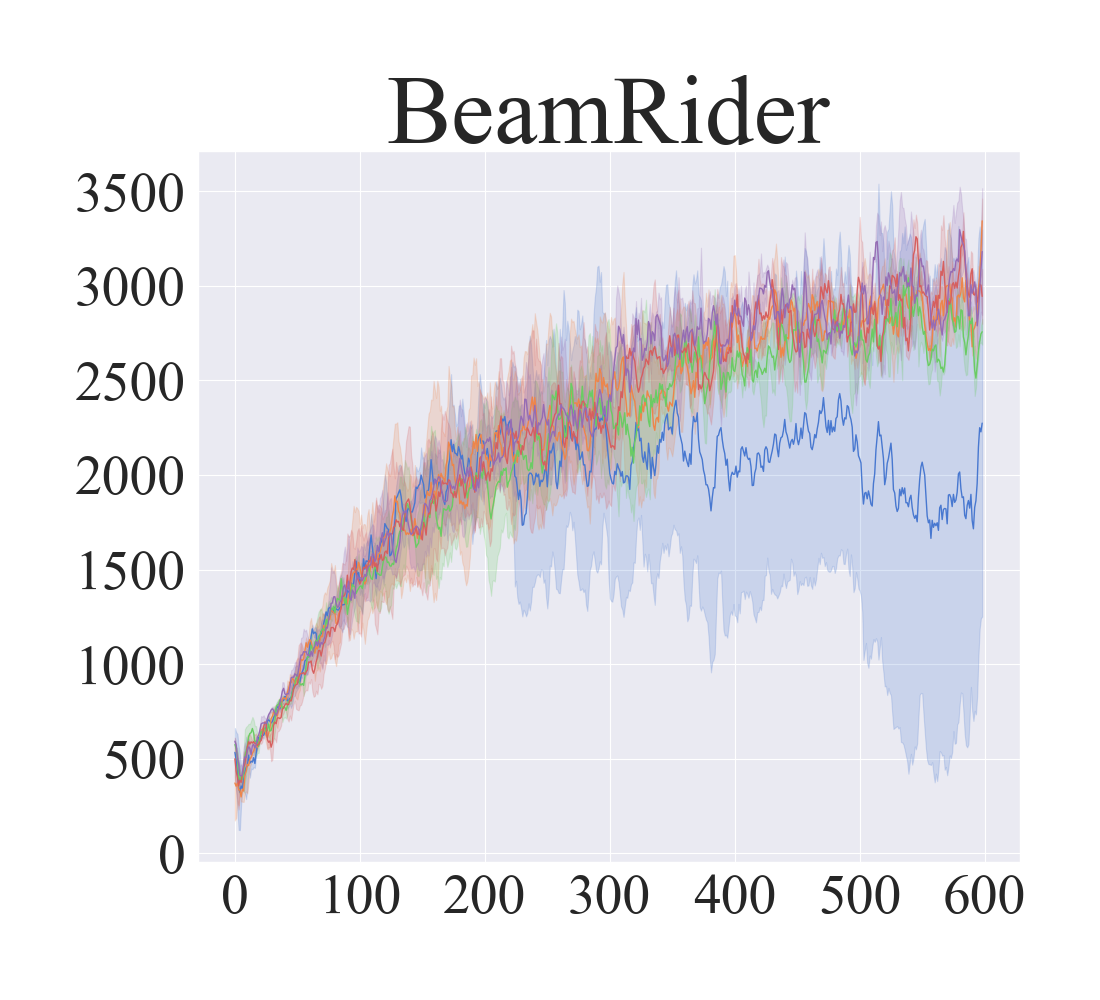}\\
			\end{minipage}%
		}%
		\subfigure{
			\begin{minipage}[t]{0.166\linewidth}
				\centering
				\includegraphics[width=1.05in]{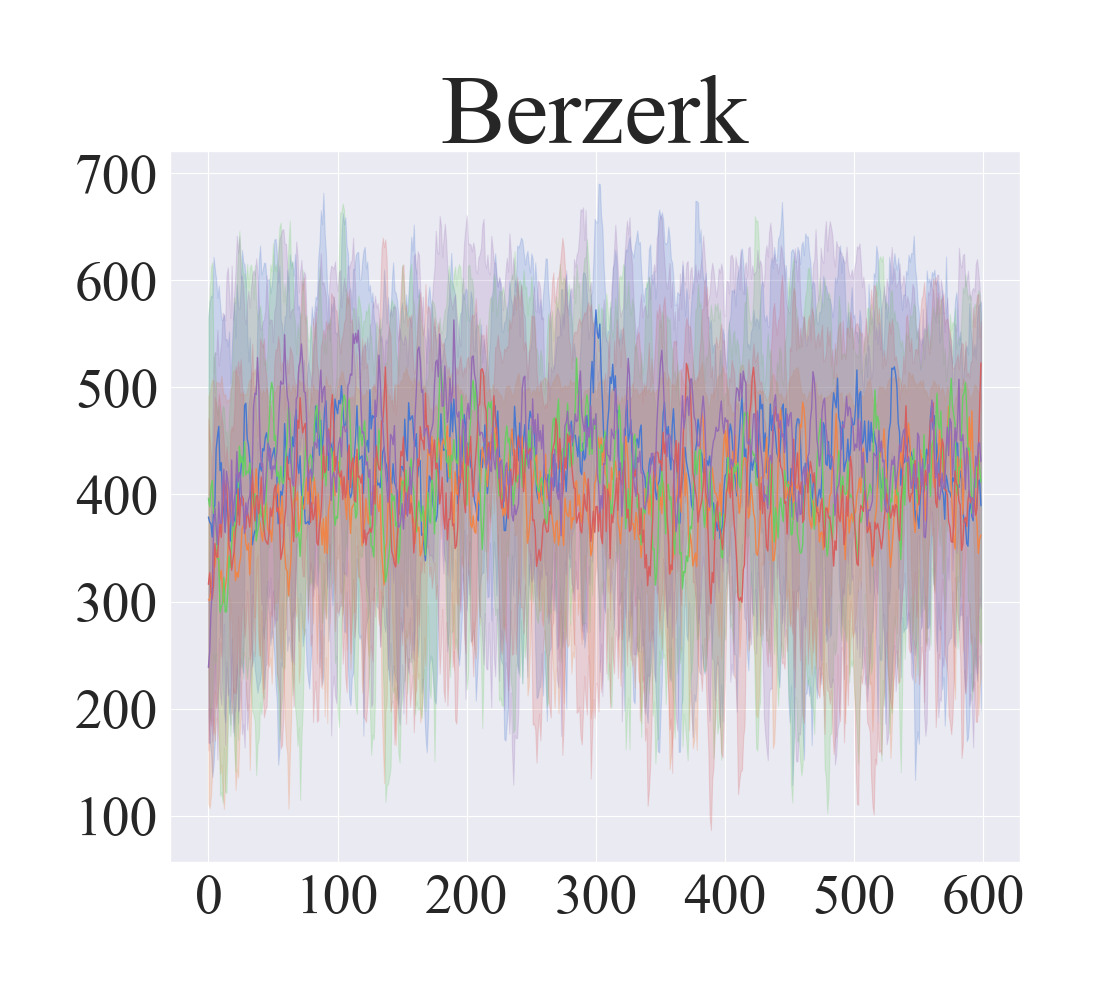}\\
			\end{minipage}%
		}%
		\subfigure{
			\begin{minipage}[t]{0.166\linewidth}
				\centering
				\includegraphics[width=1.05in]{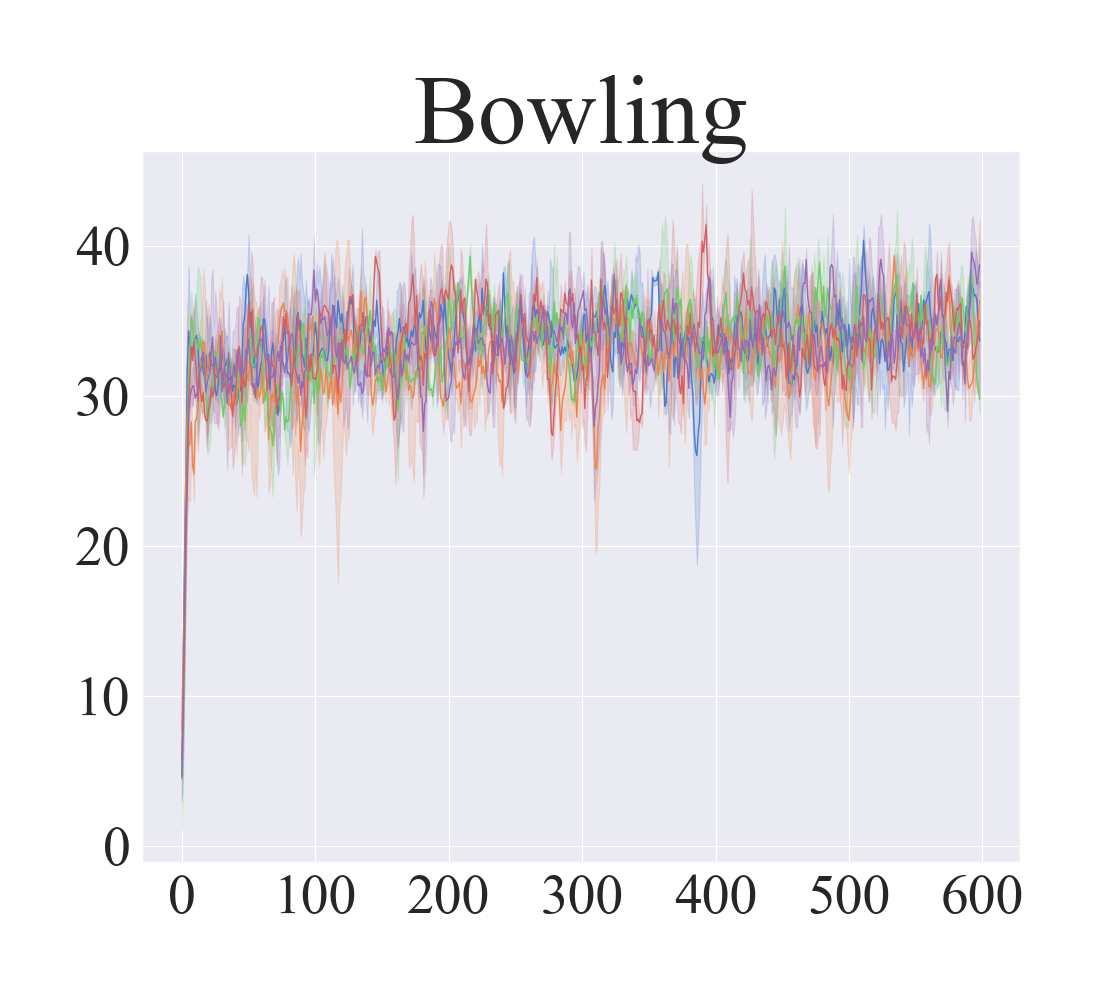}\\
			\end{minipage}%
		}%
		\vspace{-0.6cm}
		
		\subfigure{
			\begin{minipage}[t]{0.166\linewidth}
				\centering
				\includegraphics[width=1.05in]{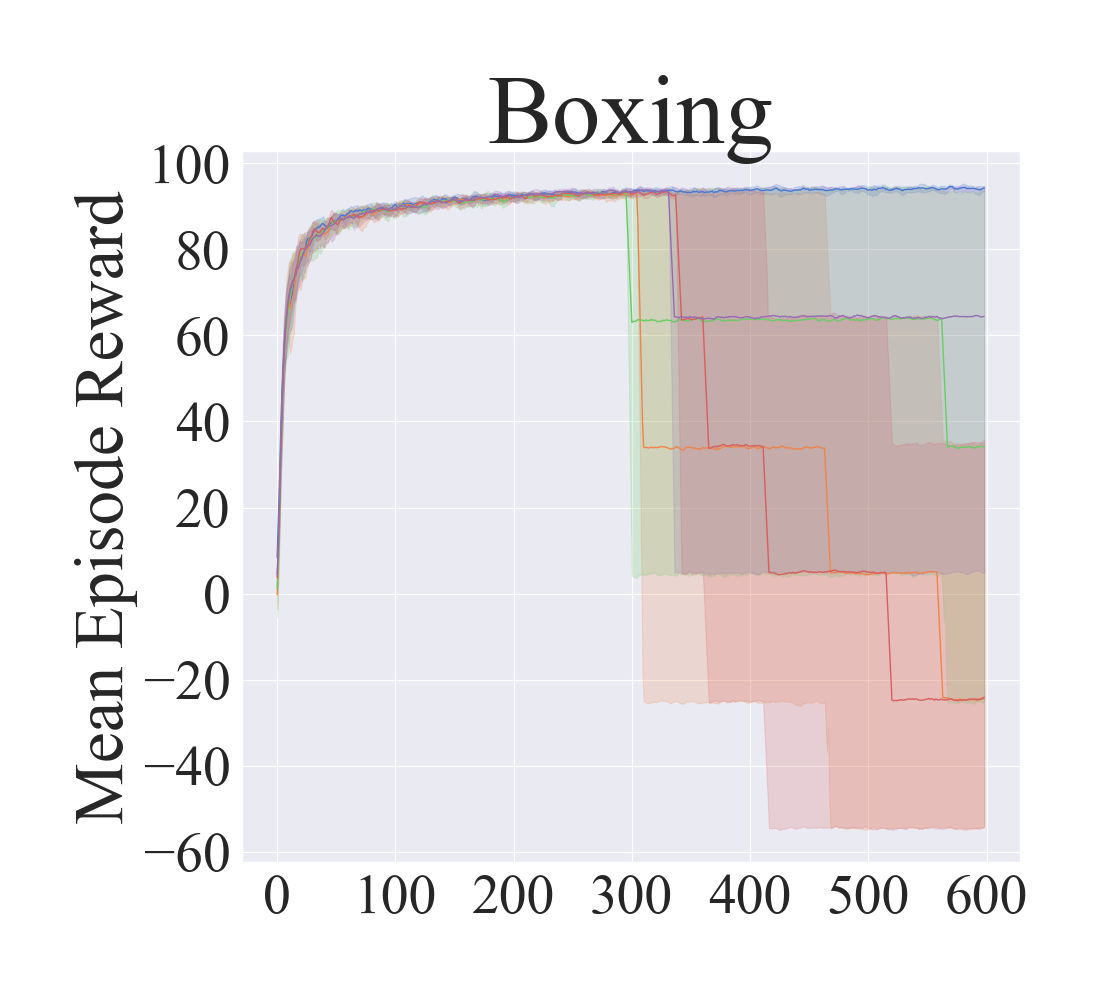}\\
			\end{minipage}%
		}%
		\subfigure{
			\begin{minipage}[t]{0.166\linewidth}
				\centering
				\includegraphics[width=1.05in]{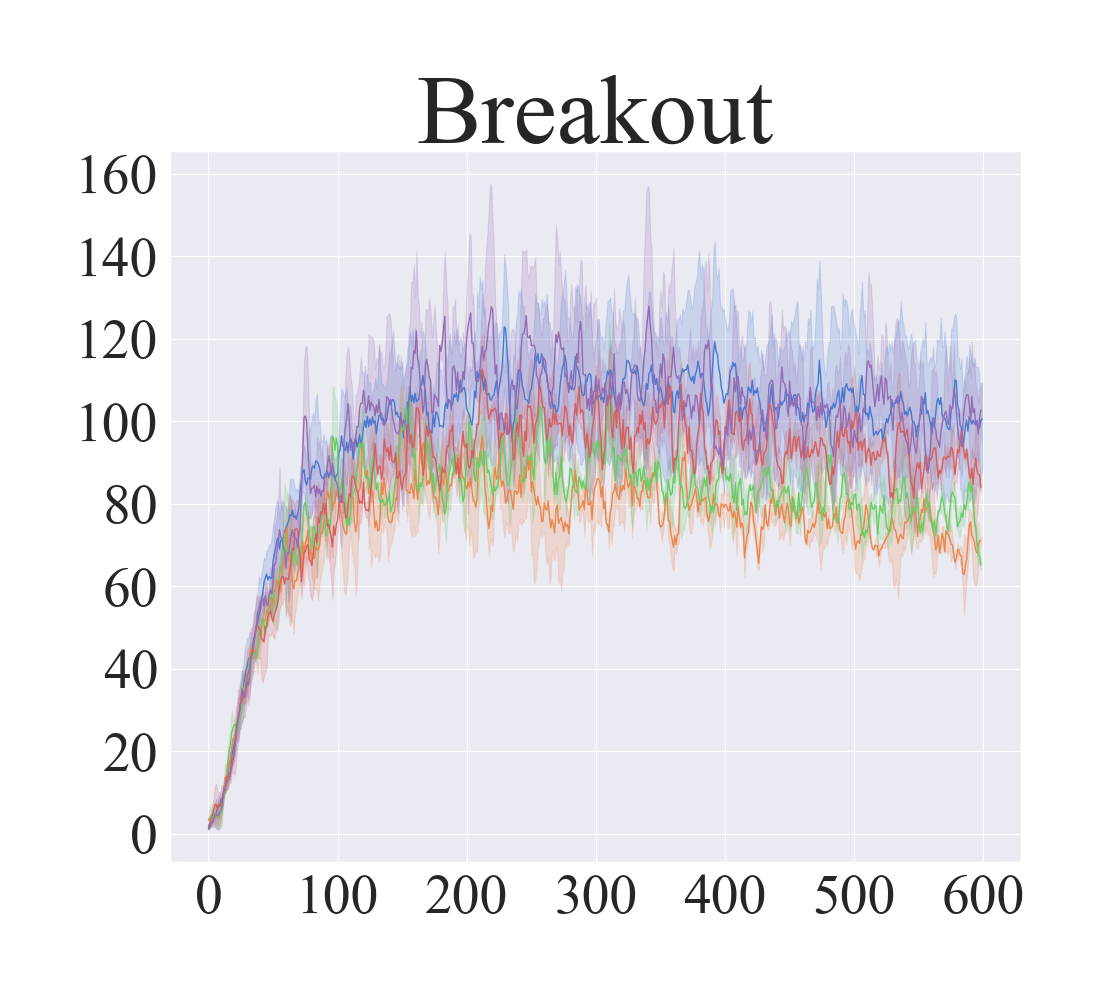}\\
			\end{minipage}%
		}%
		\subfigure{
			\begin{minipage}[t]{0.166\linewidth}
				\centering
				\includegraphics[width=1.05in]{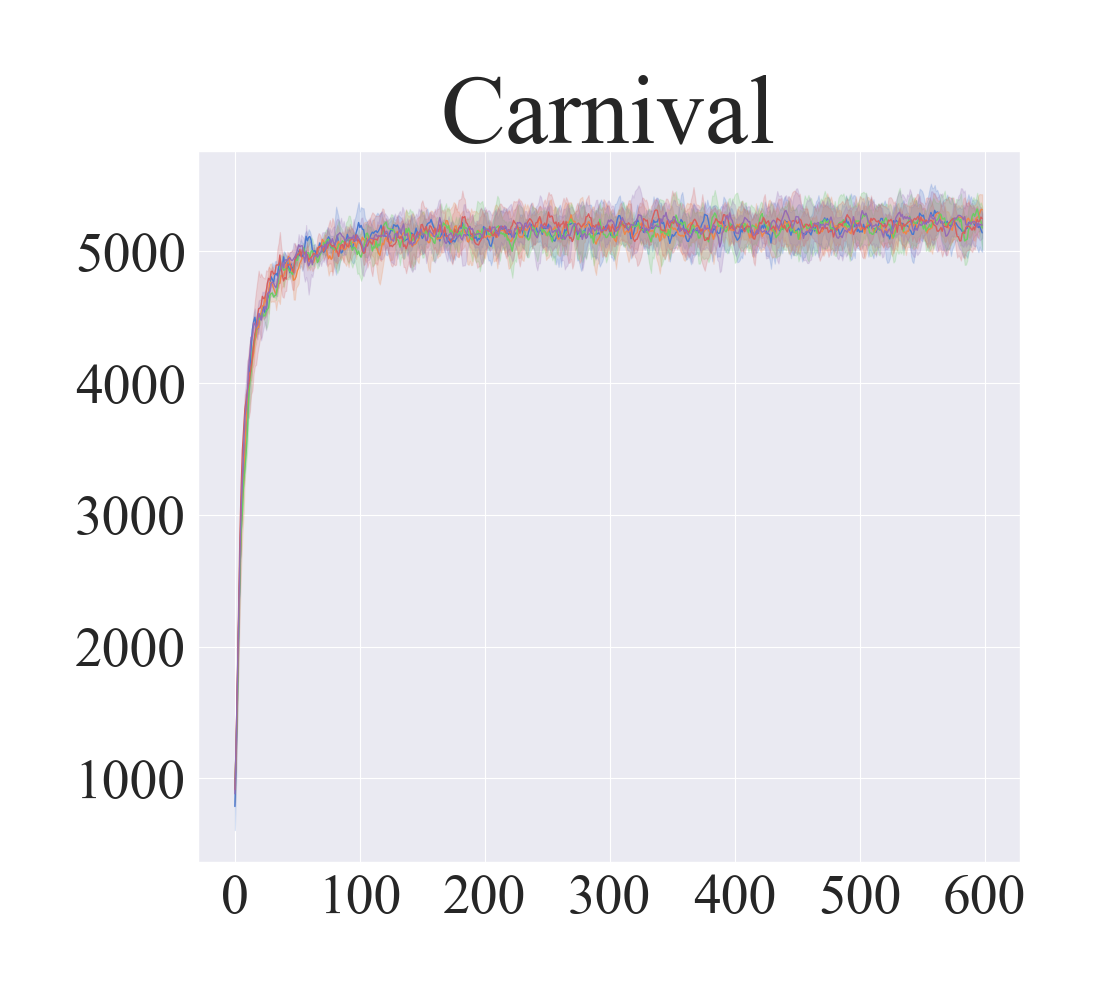}\\
			\end{minipage}%
		}%
		\subfigure{
			\begin{minipage}[t]{0.166\linewidth}
				\centering
				\includegraphics[width=1.05in]{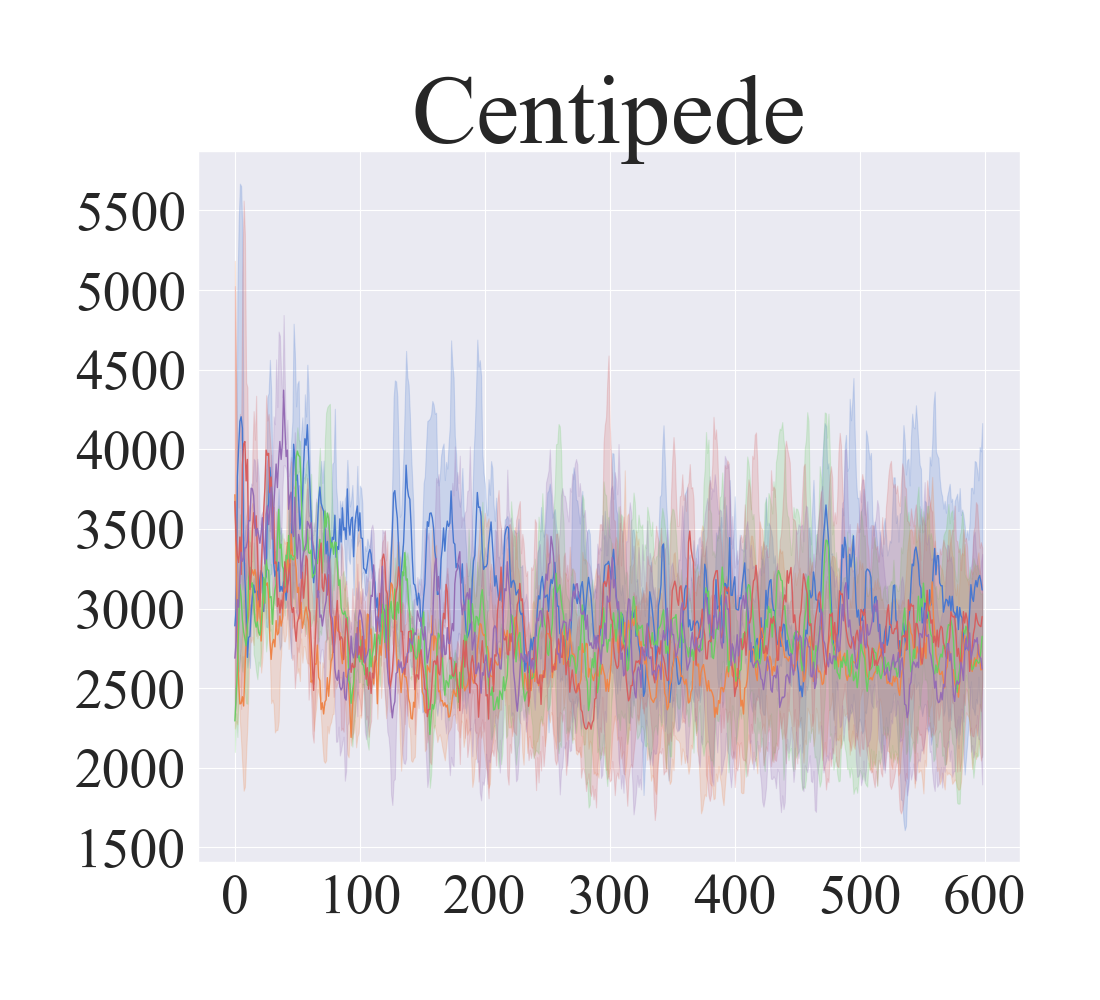}\\
			\end{minipage}%
		}%
		\subfigure{
			\begin{minipage}[t]{0.166\linewidth}
				\centering
				\includegraphics[width=1.05in]{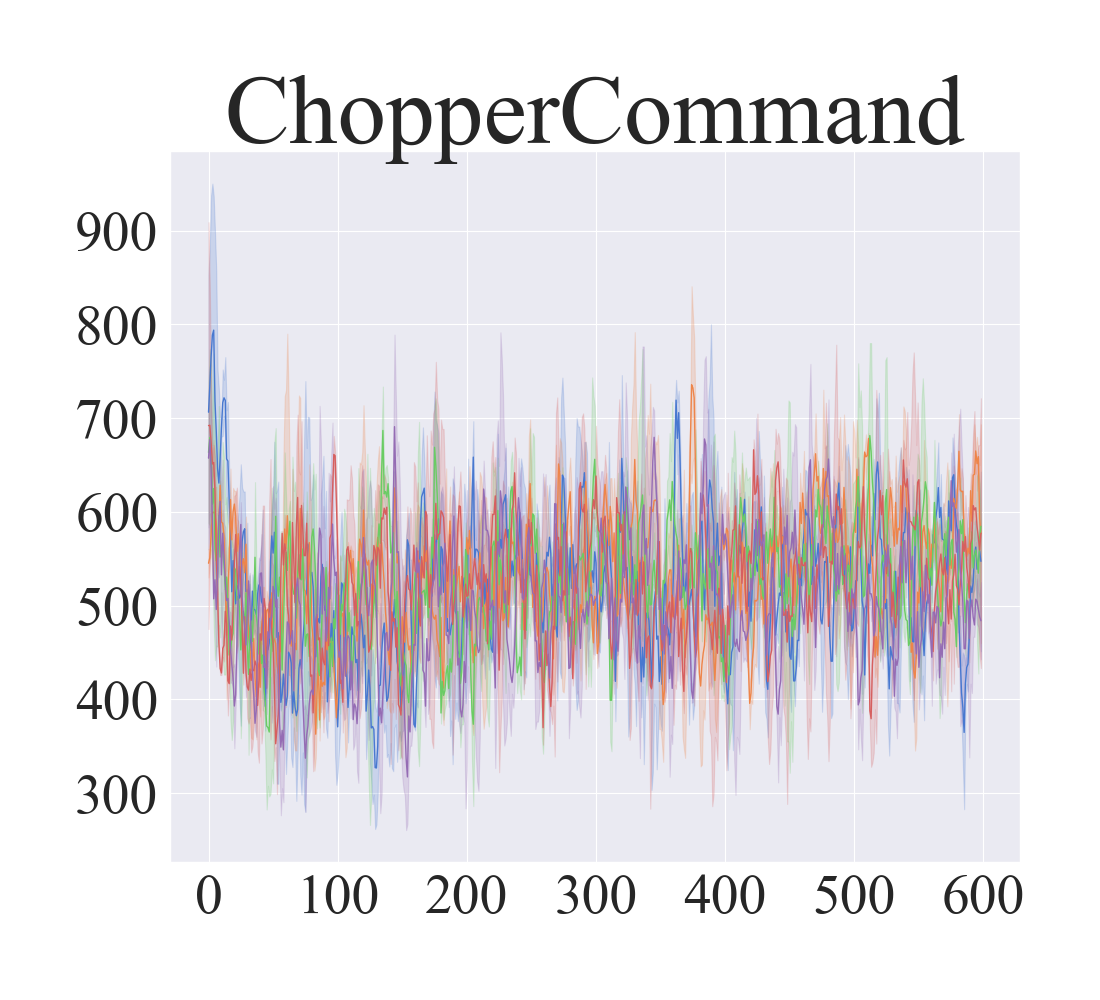}\\
			\end{minipage}%
		}%
		\subfigure{
			\begin{minipage}[t]{0.166\linewidth}
				\centering
				\includegraphics[width=1.05in]{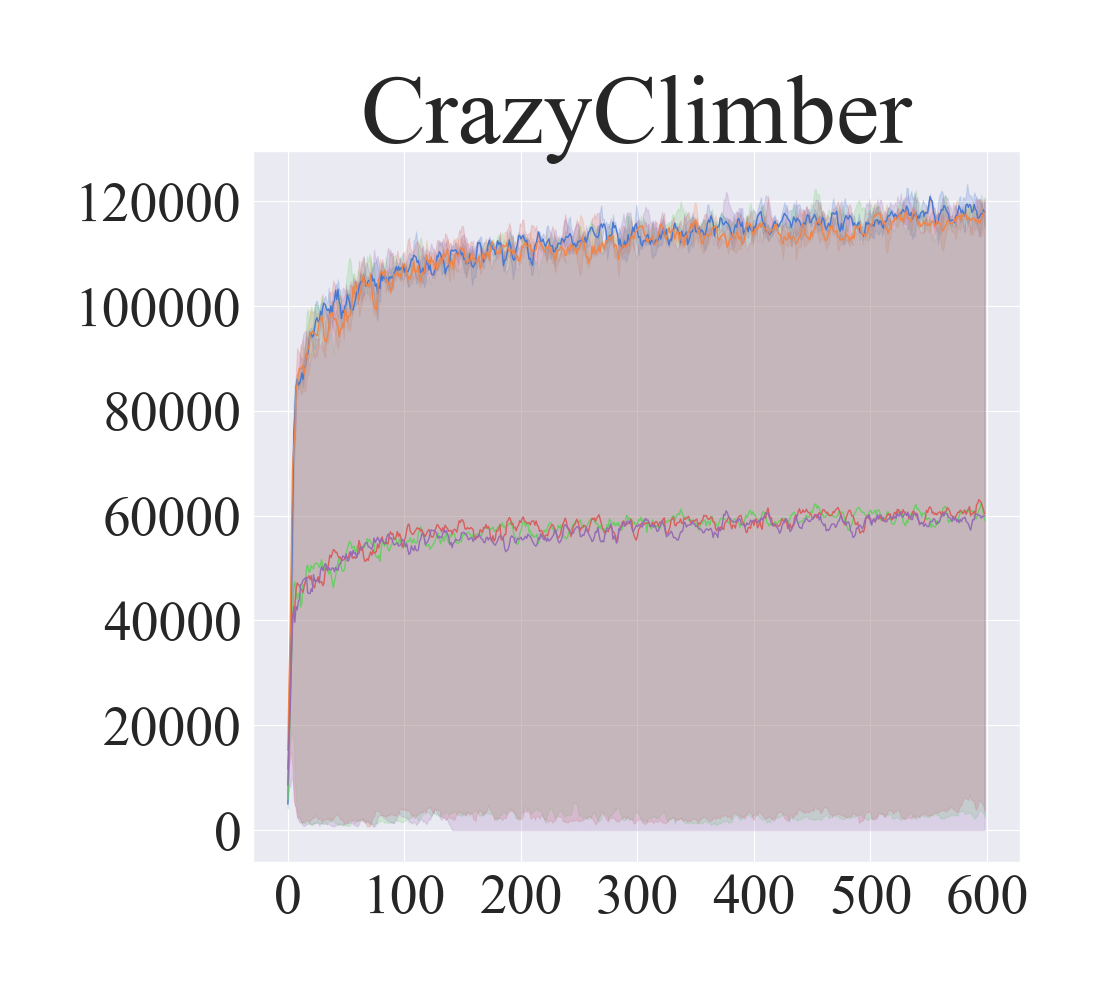}\\
			\end{minipage}%
		}%
		\vspace{-0.6cm}
		
		\subfigure{
			\begin{minipage}[t]{0.166\linewidth}
				\centering
				\includegraphics[width=1.05in]{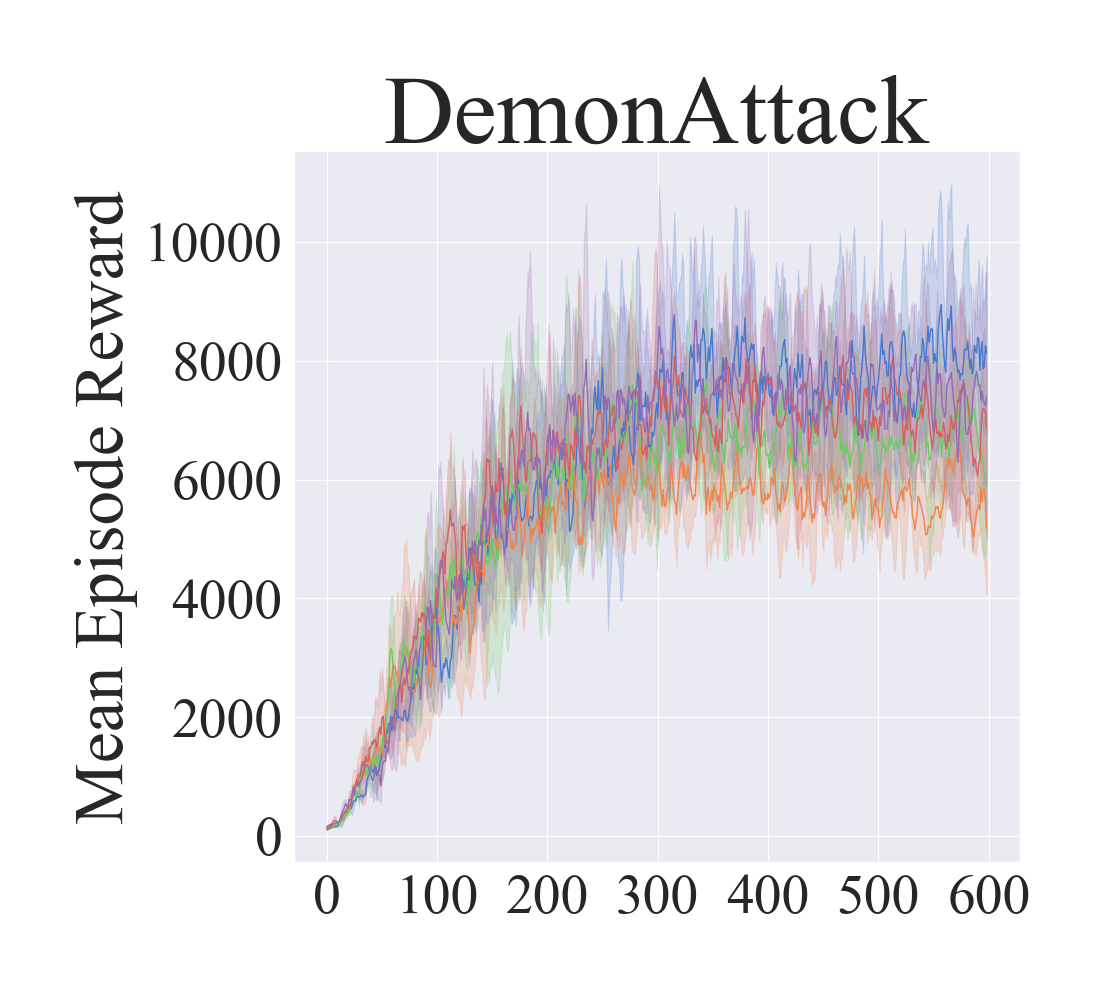}\\
			\end{minipage}%
		}%
		\subfigure{
			\begin{minipage}[t]{0.166\linewidth}
				\centering
				\includegraphics[width=1.05in]{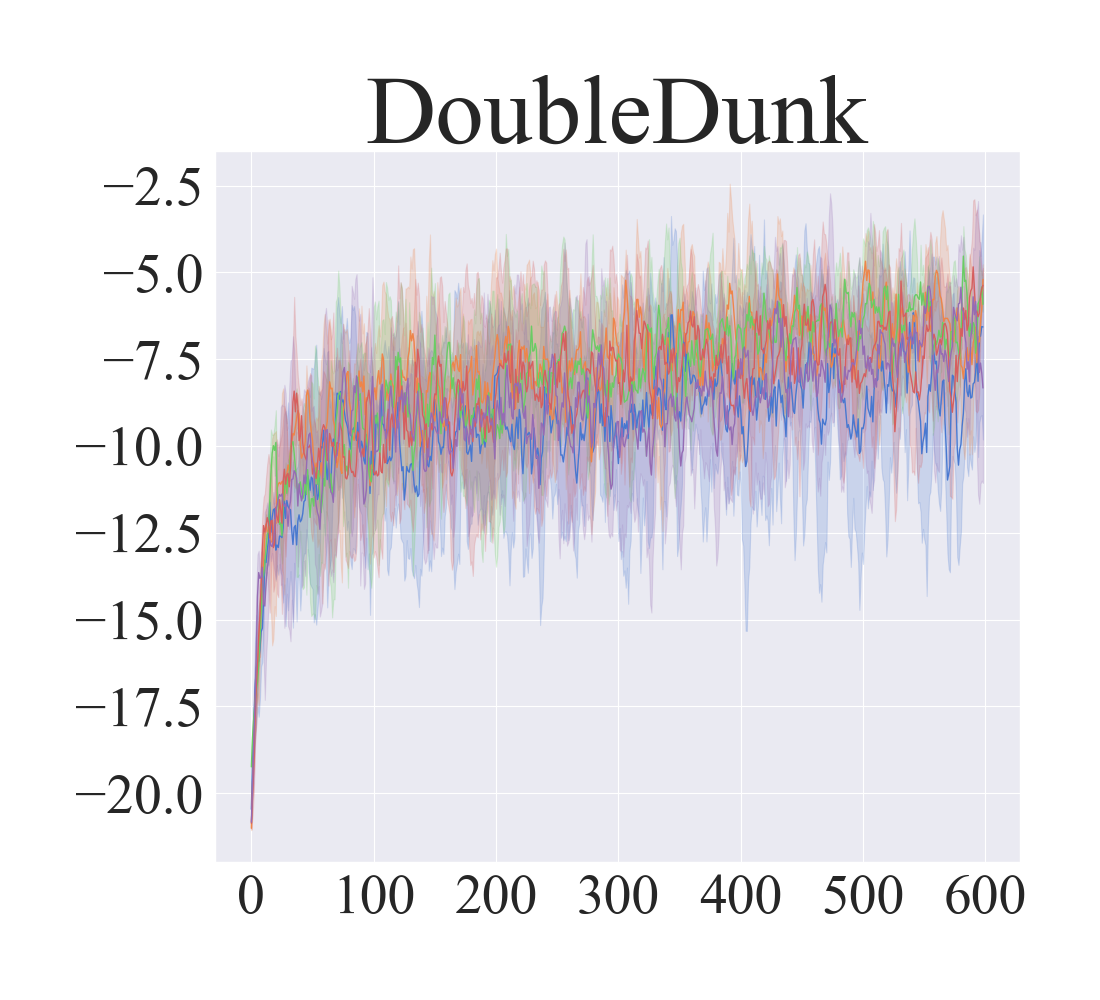}\\
			\end{minipage}%
		}%
		\subfigure{
			\begin{minipage}[t]{0.166\linewidth}
				\centering
				\includegraphics[width=1.05in]{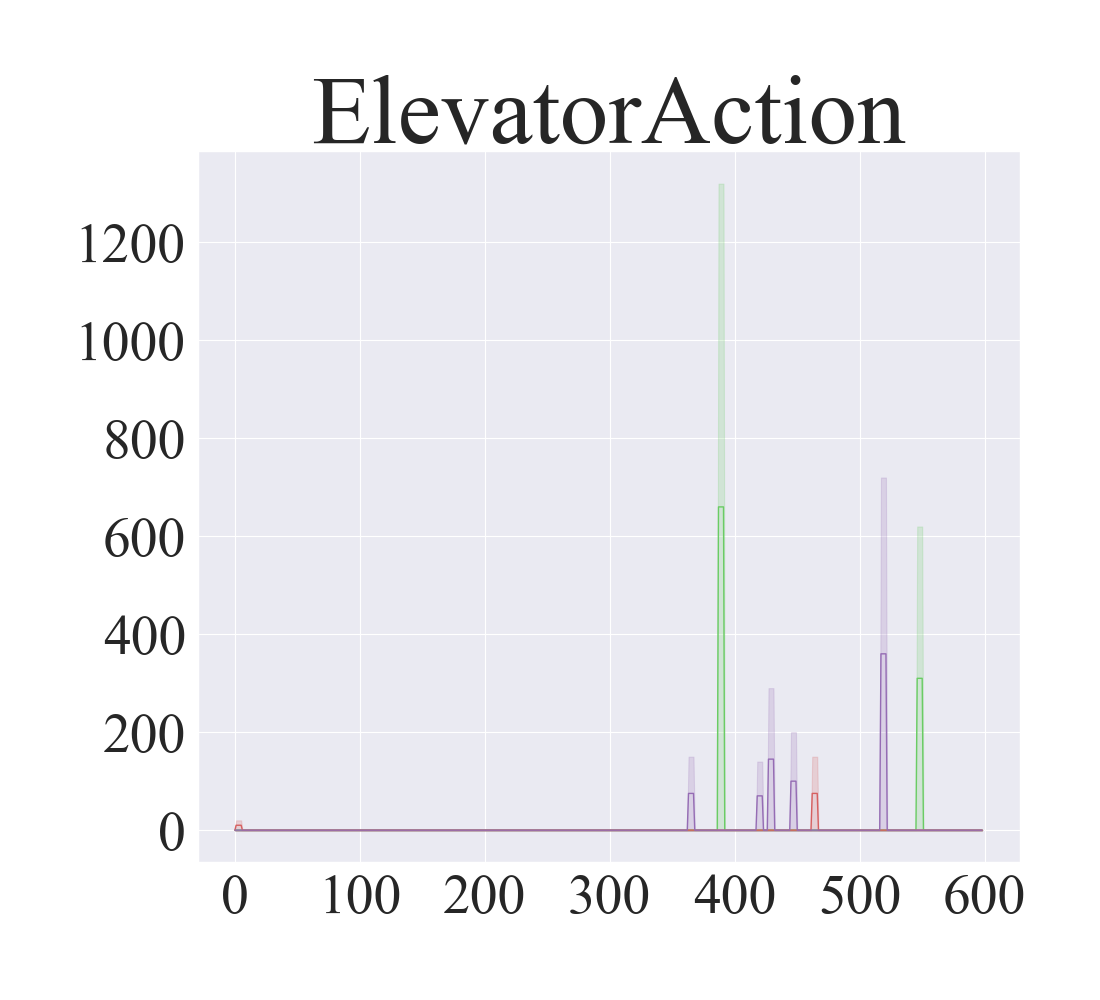}\\
			\end{minipage}%
		}%
		\subfigure{
			\begin{minipage}[t]{0.166\linewidth}
				\centering
				\includegraphics[width=1.05in]{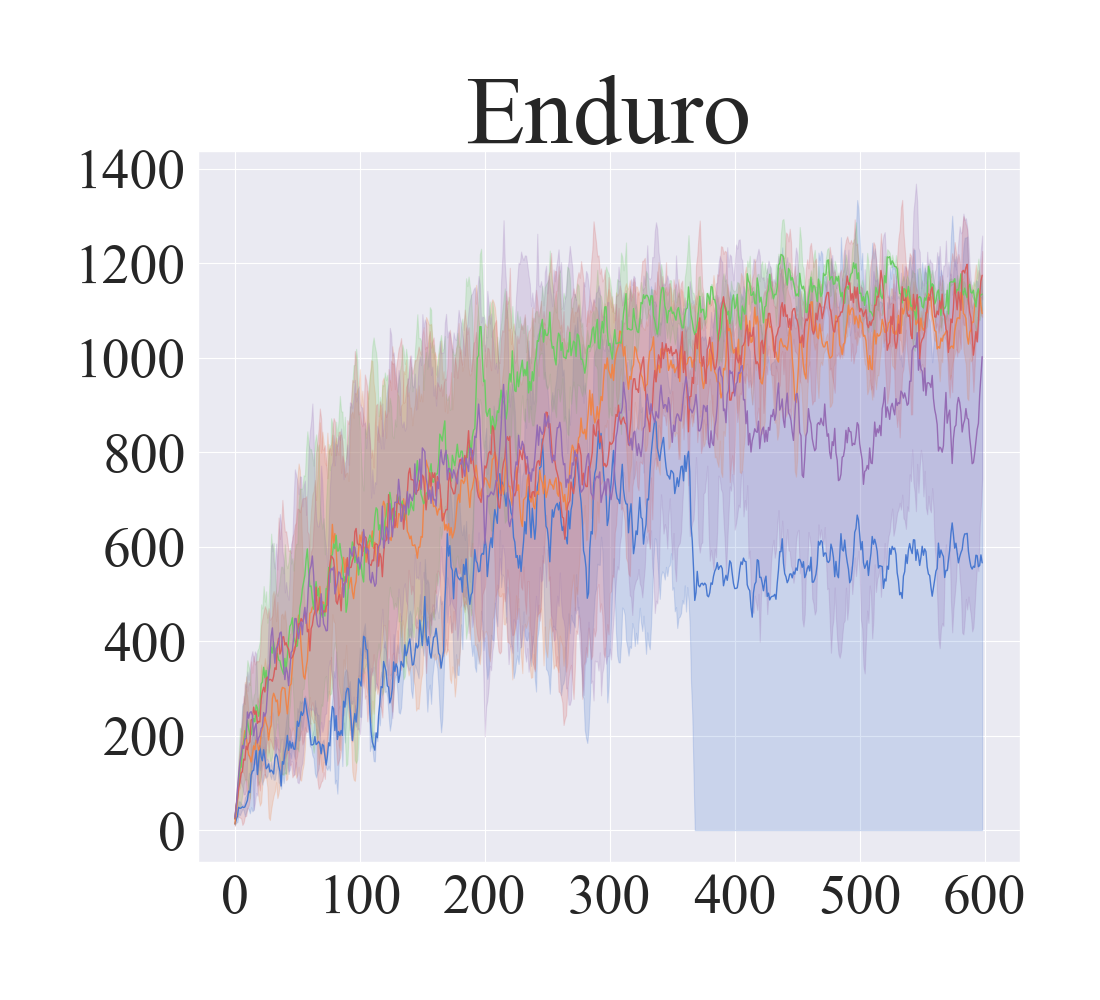}\\
			\end{minipage}%
		}%
		\subfigure{
			\begin{minipage}[t]{0.166\linewidth}
				\centering
				\includegraphics[width=1.05in]{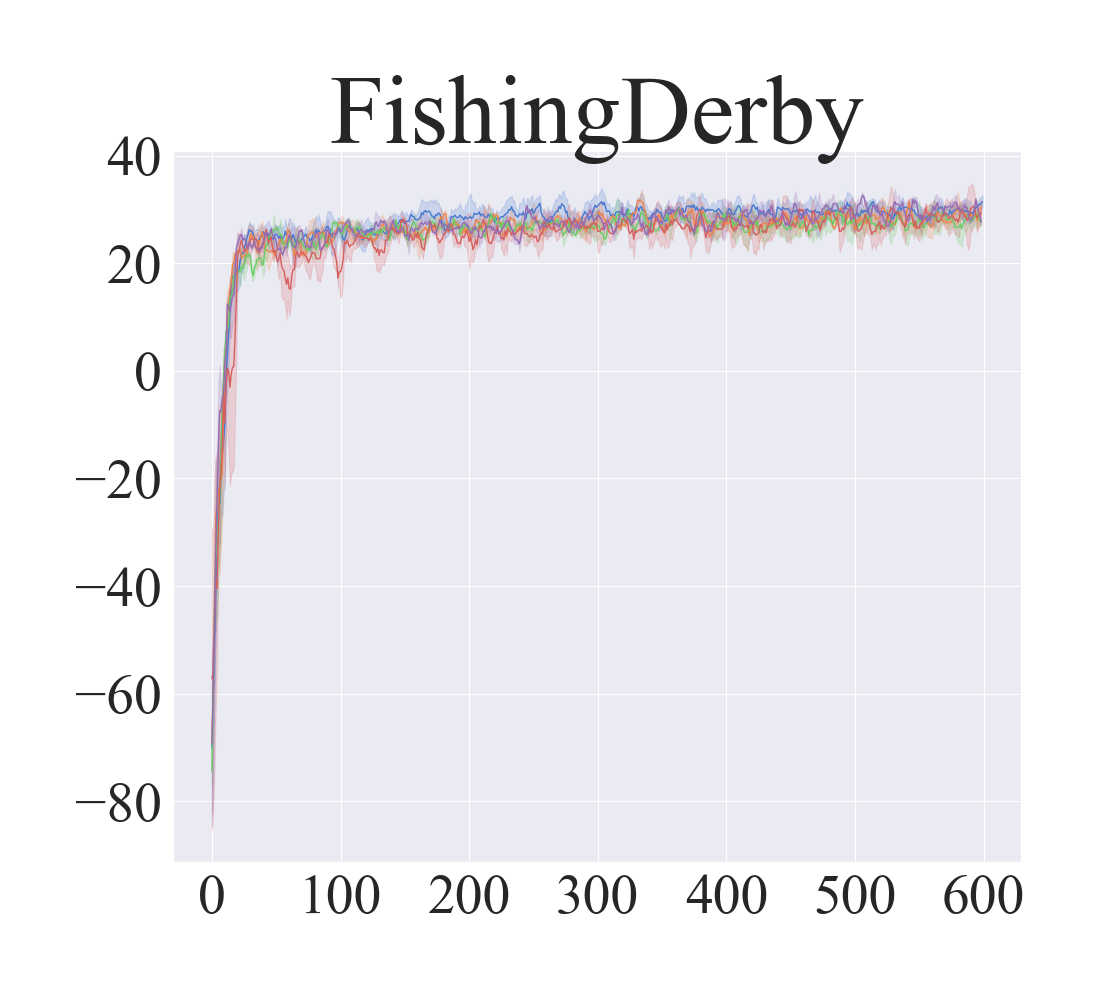}\\
			\end{minipage}%
		}%
		\subfigure{
			\begin{minipage}[t]{0.166\linewidth}
				\centering
				\includegraphics[width=1.05in]{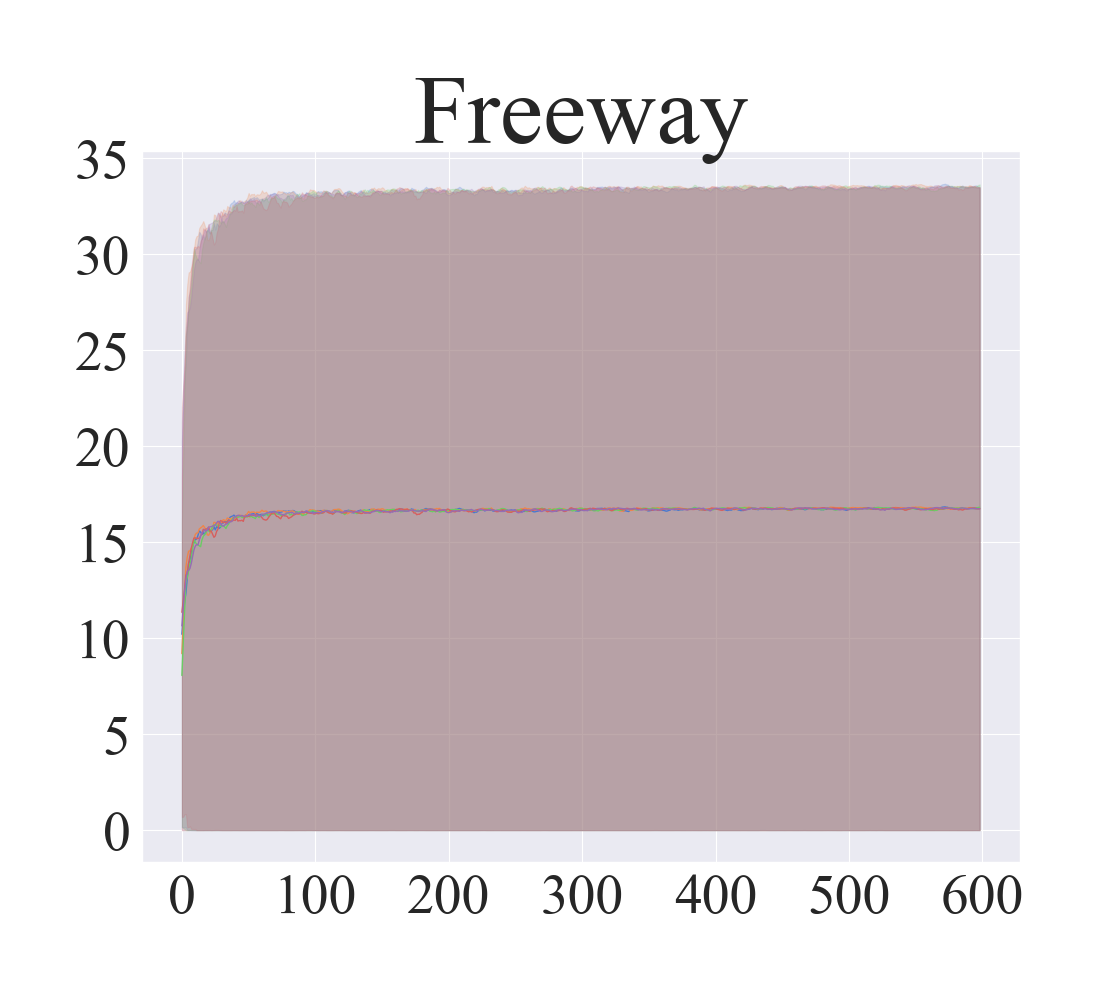}\\
			\end{minipage}%
		}%
		\vspace{-0.6cm}
		
		\subfigure{
			\begin{minipage}[t]{0.166\linewidth}
				\centering
				\includegraphics[width=1.05in]{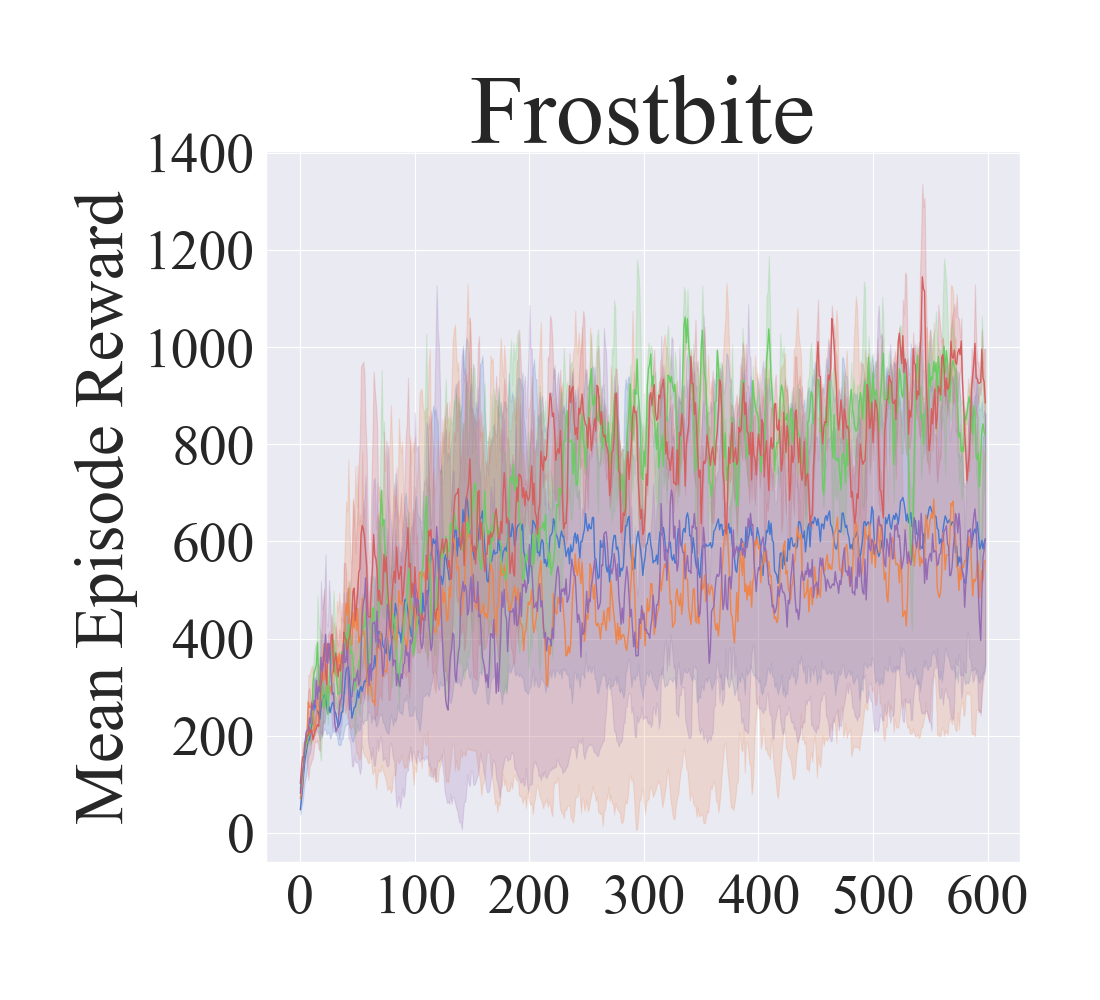}\\
			\end{minipage}%
		}%
		\subfigure{
			\begin{minipage}[t]{0.166\linewidth}
				\centering
				\includegraphics[width=1.05in]{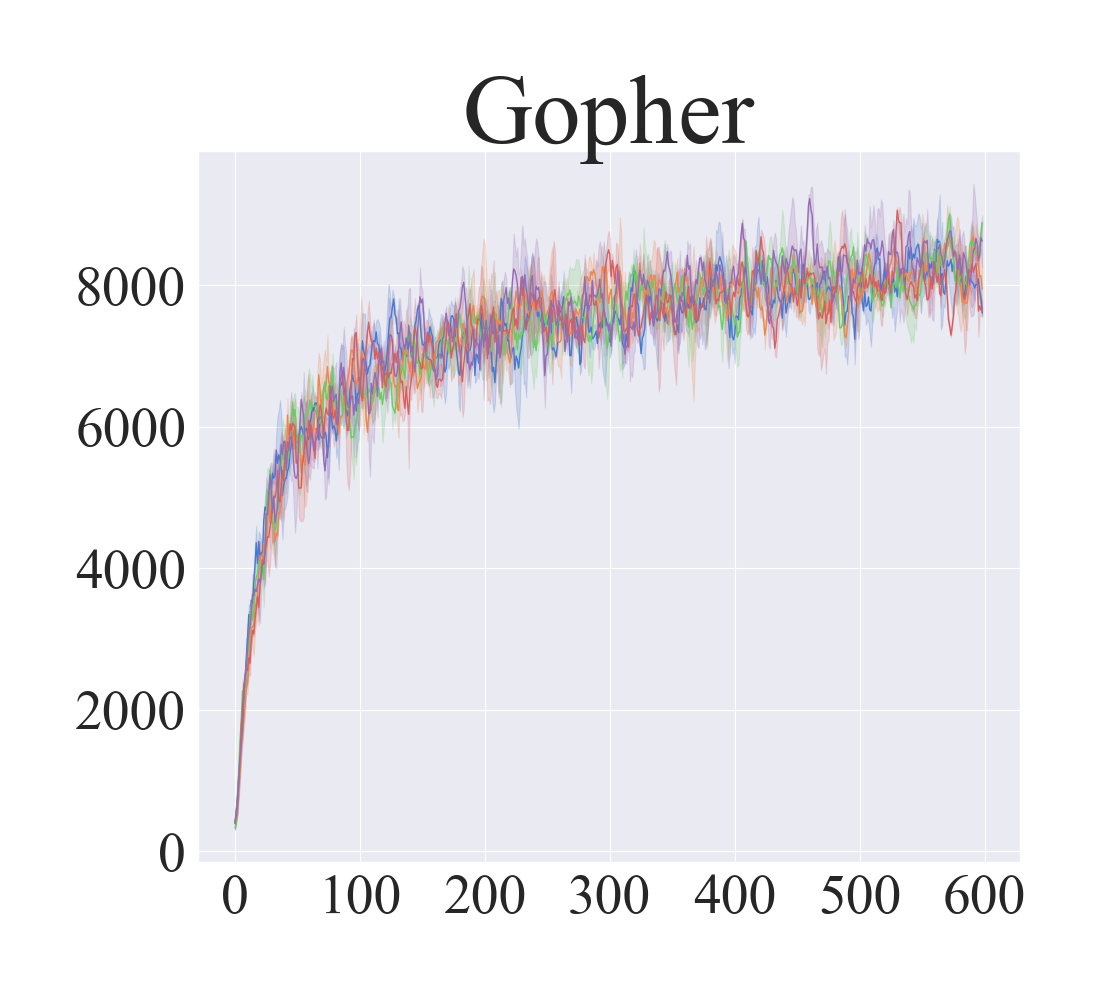}\\
			\end{minipage}%
		}%
		\subfigure{
			\begin{minipage}[t]{0.166\linewidth}
				\centering
				\includegraphics[width=1.05in]{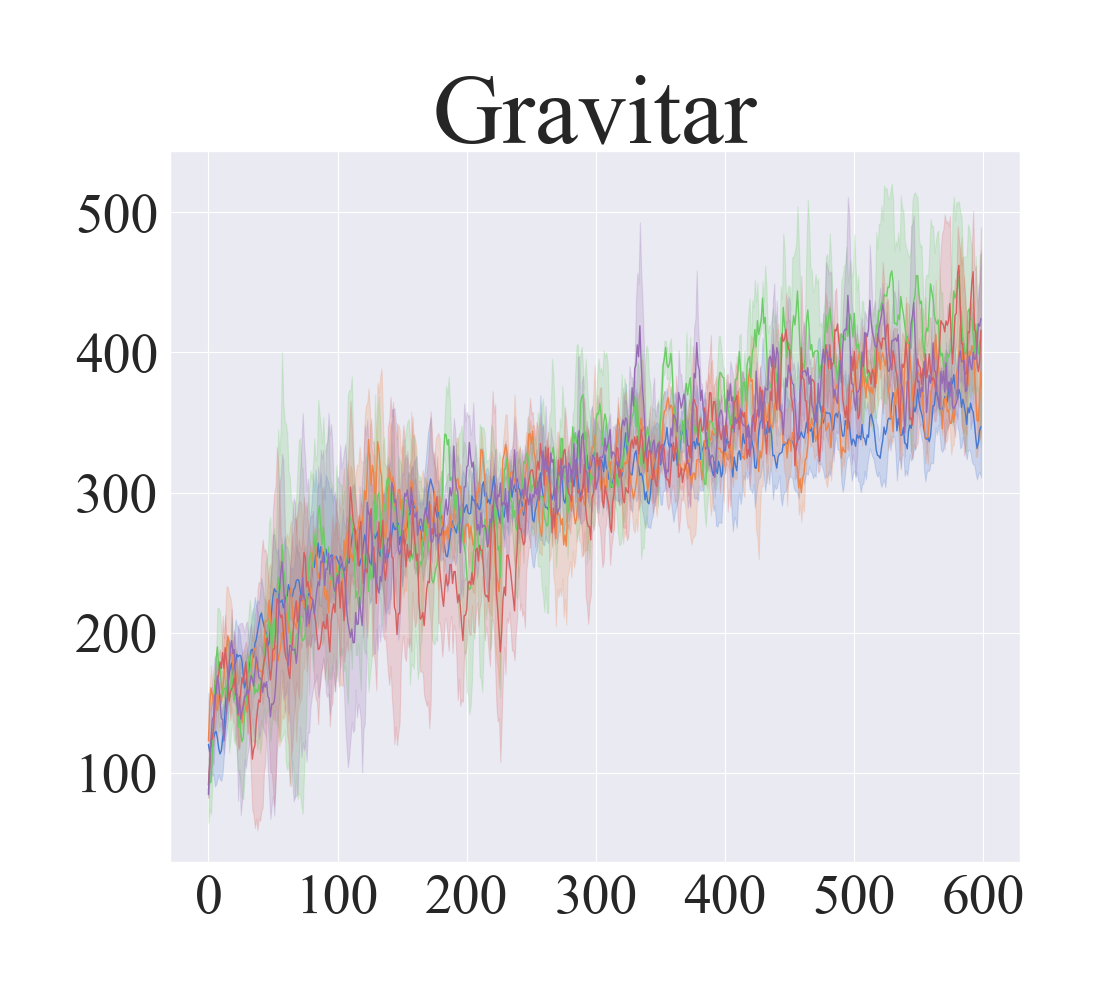}\\
			\end{minipage}%
		}%
		\subfigure{
			\begin{minipage}[t]{0.166\linewidth}
				\centering
				\includegraphics[width=1.05in]{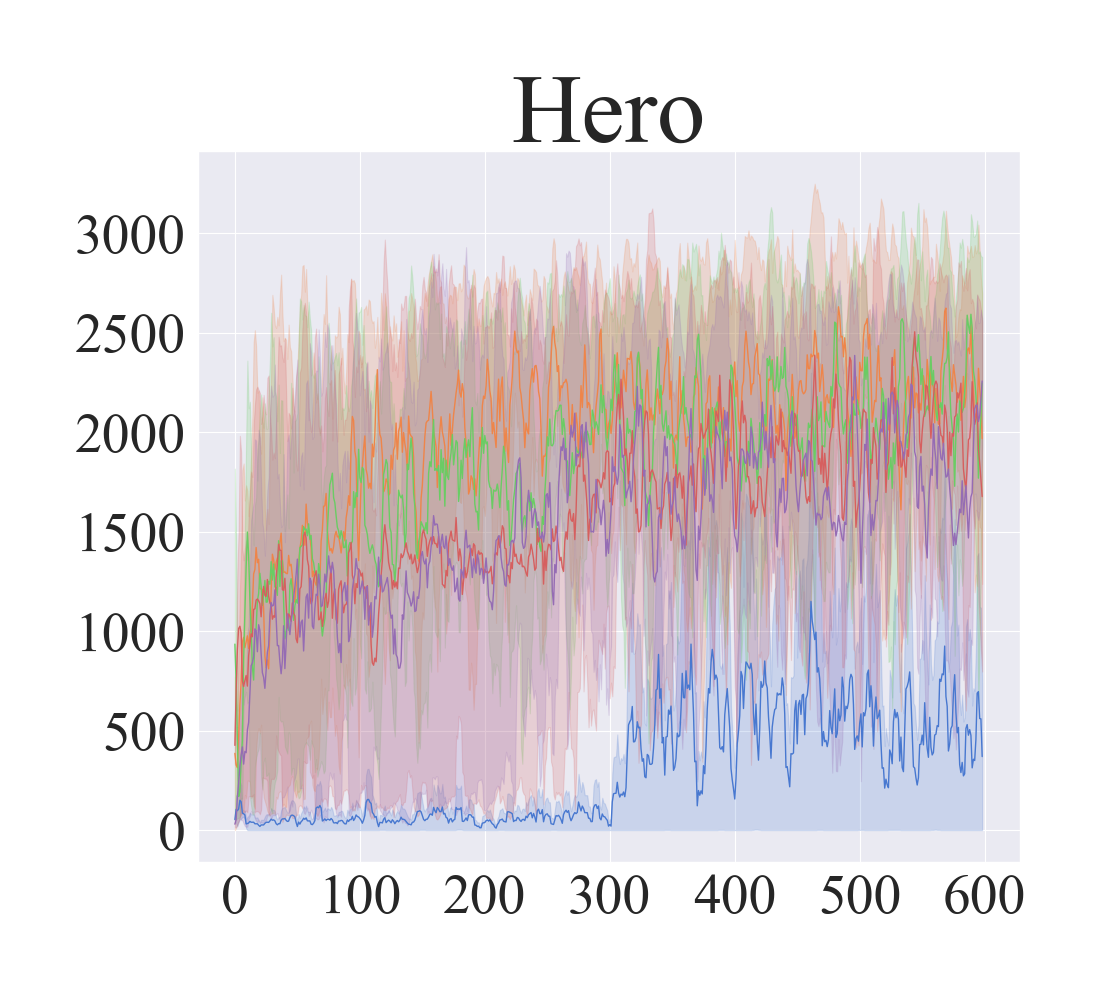}\\
			\end{minipage}%
		}%
		\subfigure{
			\begin{minipage}[t]{0.166\linewidth}
				\centering
				\includegraphics[width=1.05in]{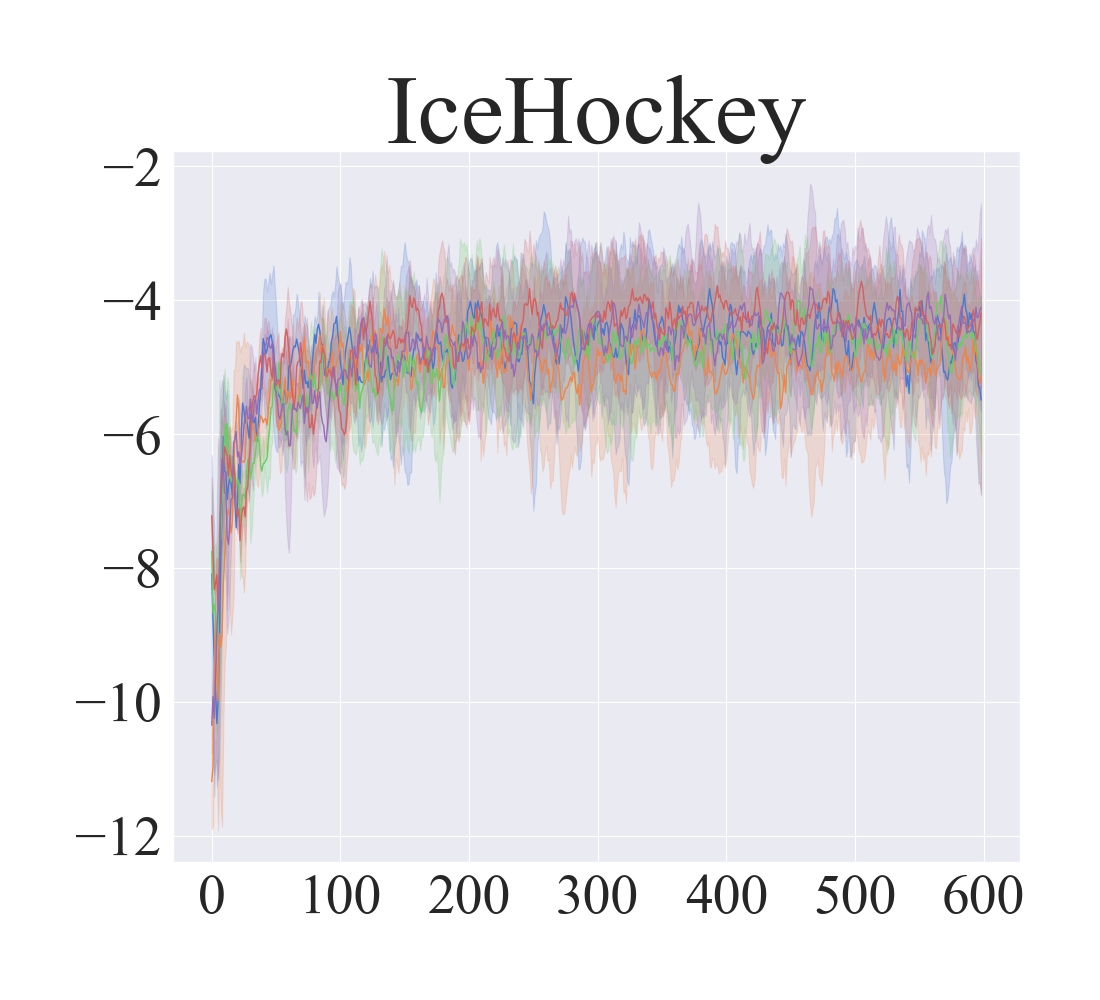}\\
			\end{minipage}%
		}%
		\subfigure{
			\begin{minipage}[t]{0.166\linewidth}
				\centering
				\includegraphics[width=1.05in]{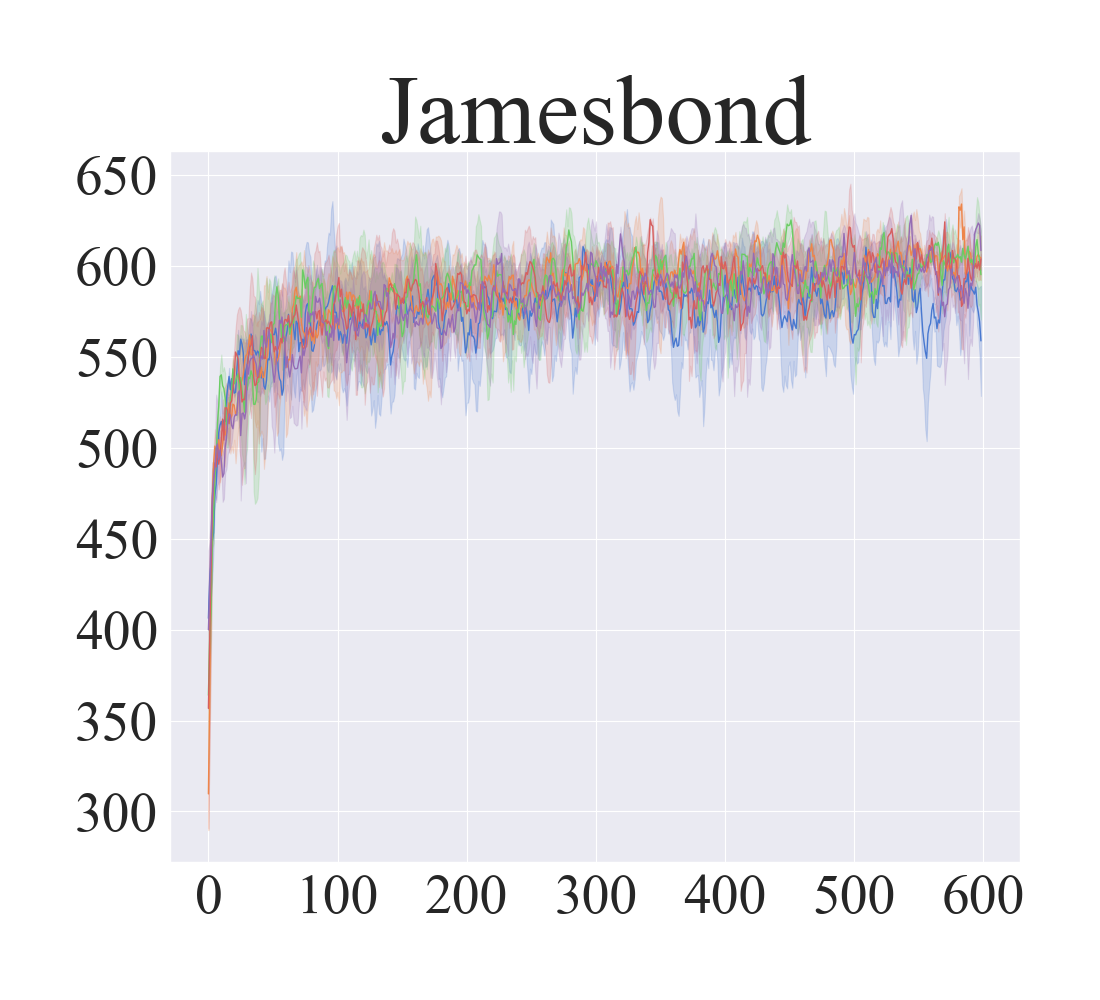}\\
			\end{minipage}%
		}%
		\vspace{-0.6cm}
		
		\subfigure{
			\begin{minipage}[t]{0.166\linewidth}
				\centering
				\includegraphics[width=1.05in]{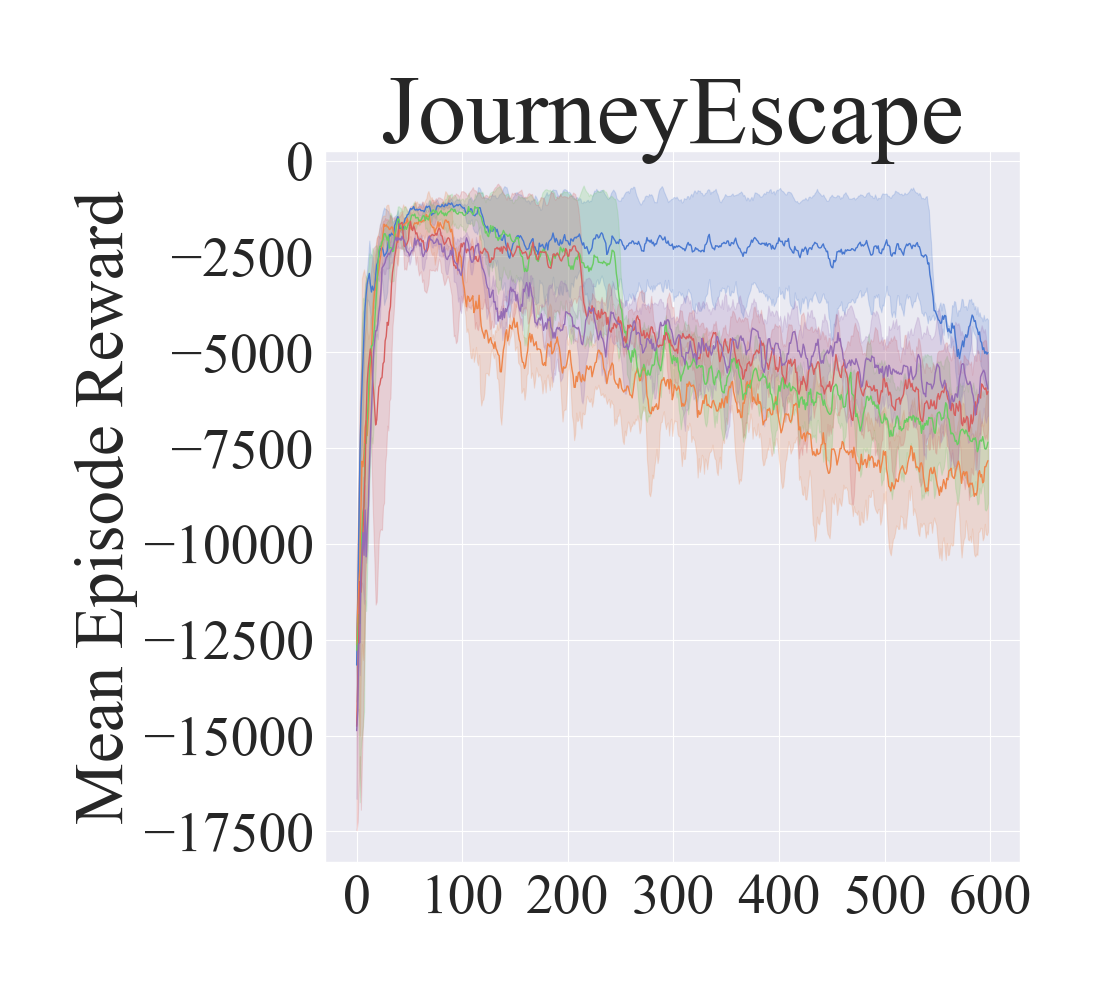}\\
			\end{minipage}%
		}%
		\subfigure{
			\begin{minipage}[t]{0.166\linewidth}
				\centering
				\includegraphics[width=1.05in]{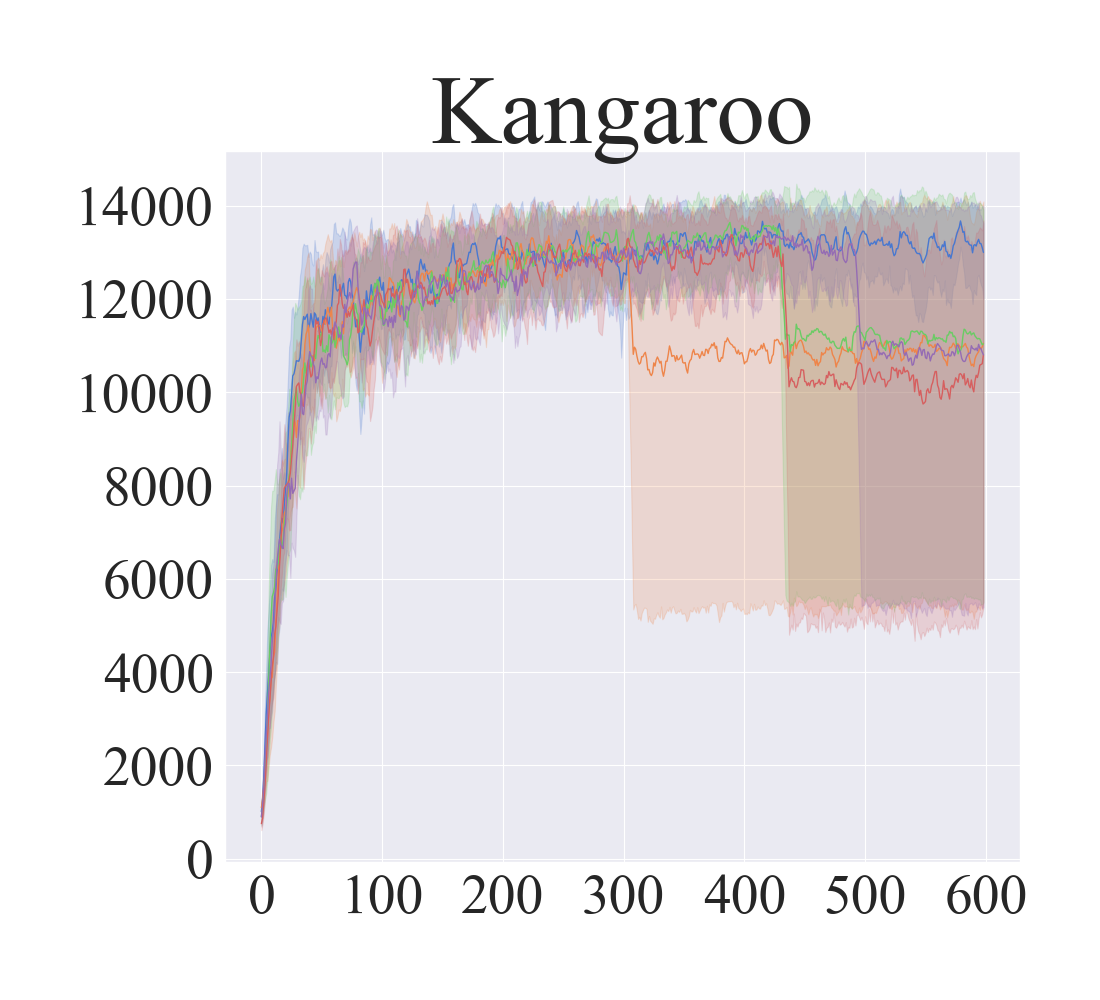}\\
			\end{minipage}%
		}%
		\subfigure{
			\begin{minipage}[t]{0.166\linewidth}
				\centering
				\includegraphics[width=1.05in]{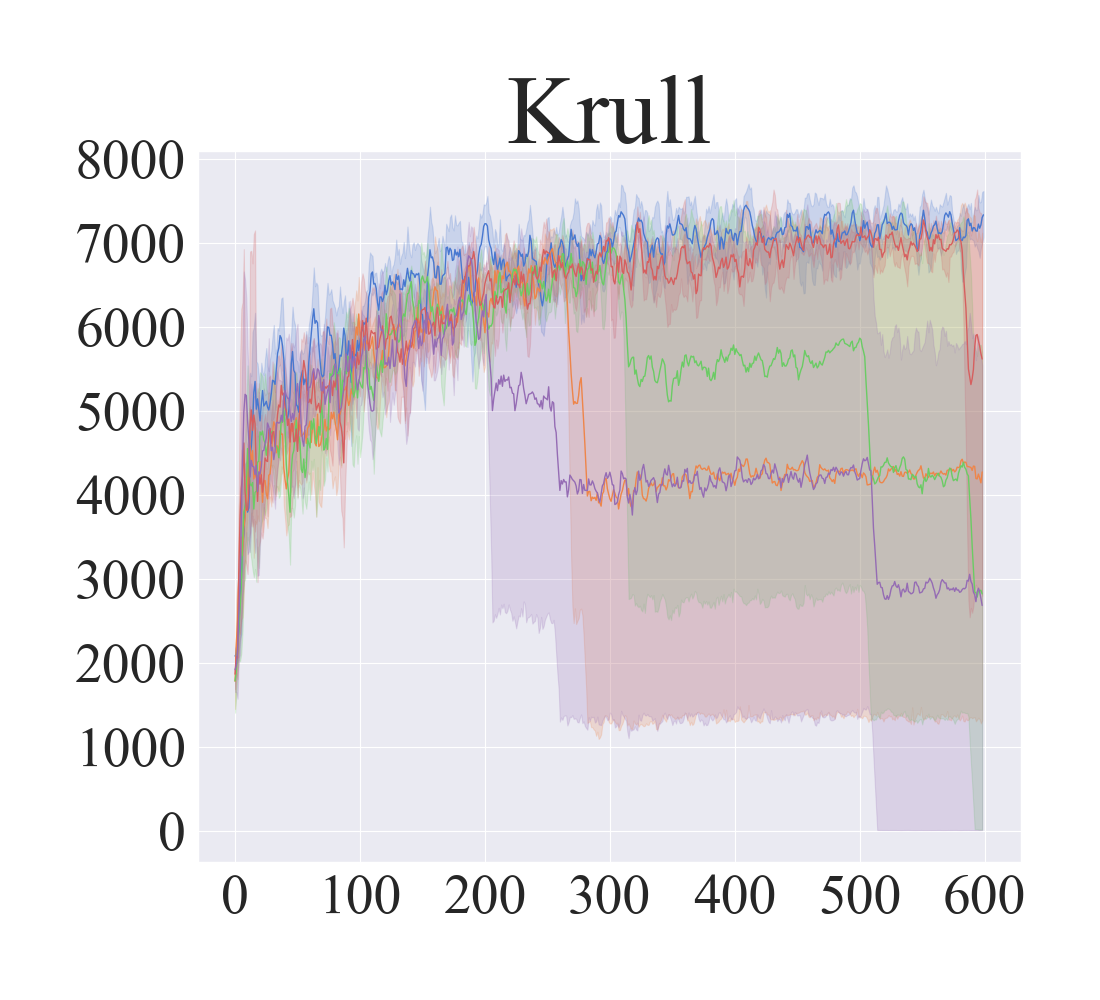}\\
			\end{minipage}%
		}%
		\subfigure{
			\begin{minipage}[t]{0.166\linewidth}
				\centering
				\includegraphics[width=1.05in]{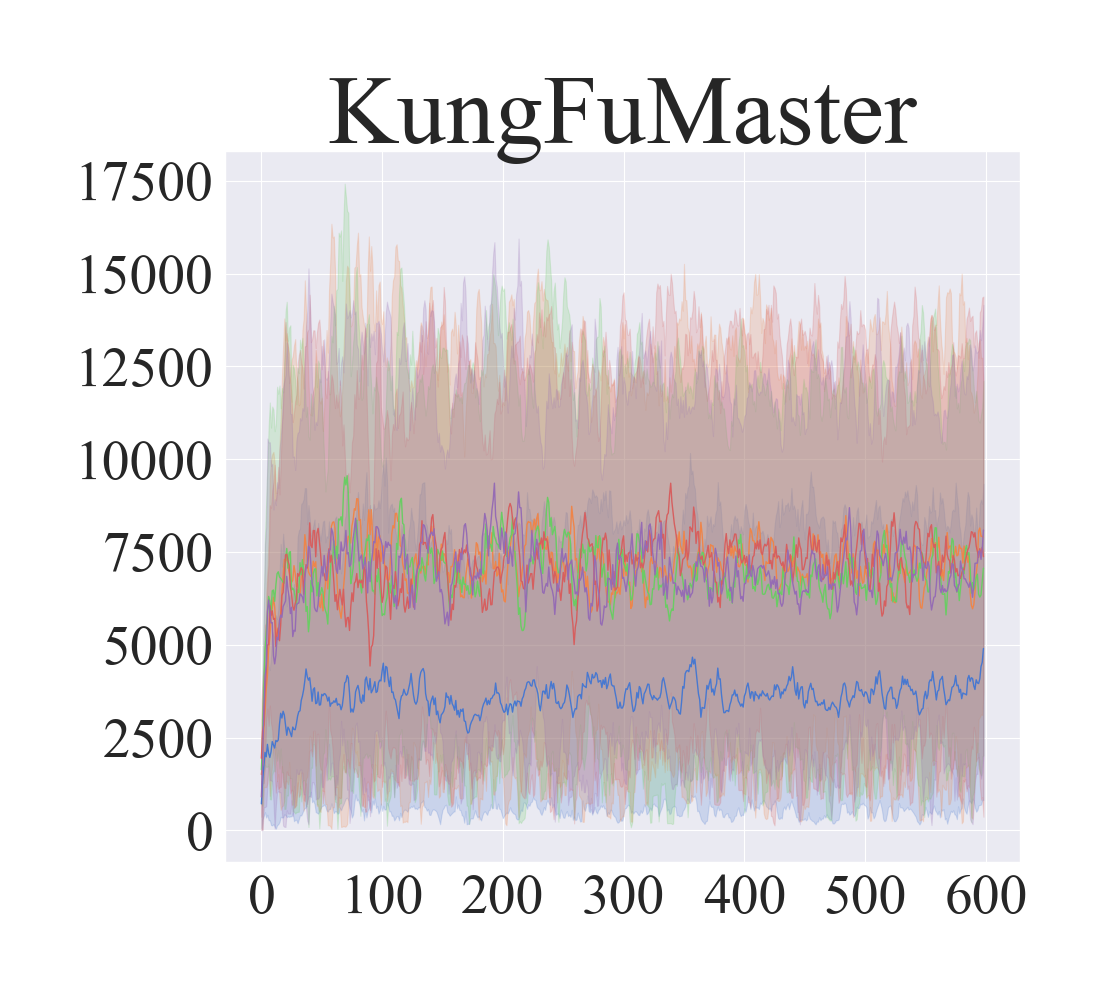}\\
			\end{minipage}%
		}%
		\subfigure{
			\begin{minipage}[t]{0.166\linewidth}
				\centering
				\includegraphics[width=1.05in]{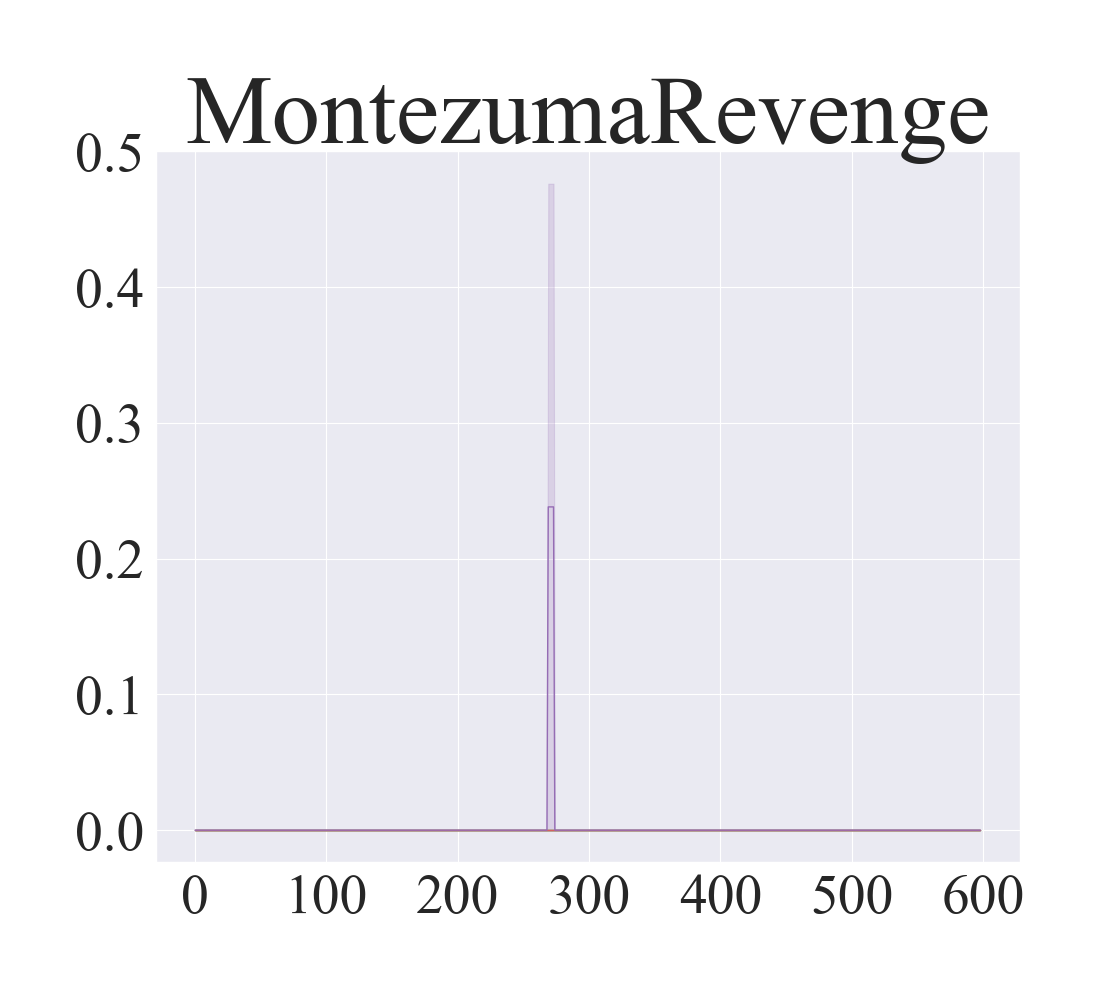}\\
			\end{minipage}%
		}%
		\subfigure{
			\begin{minipage}[t]{0.166\linewidth}
				\centering
				\includegraphics[width=1.05in]{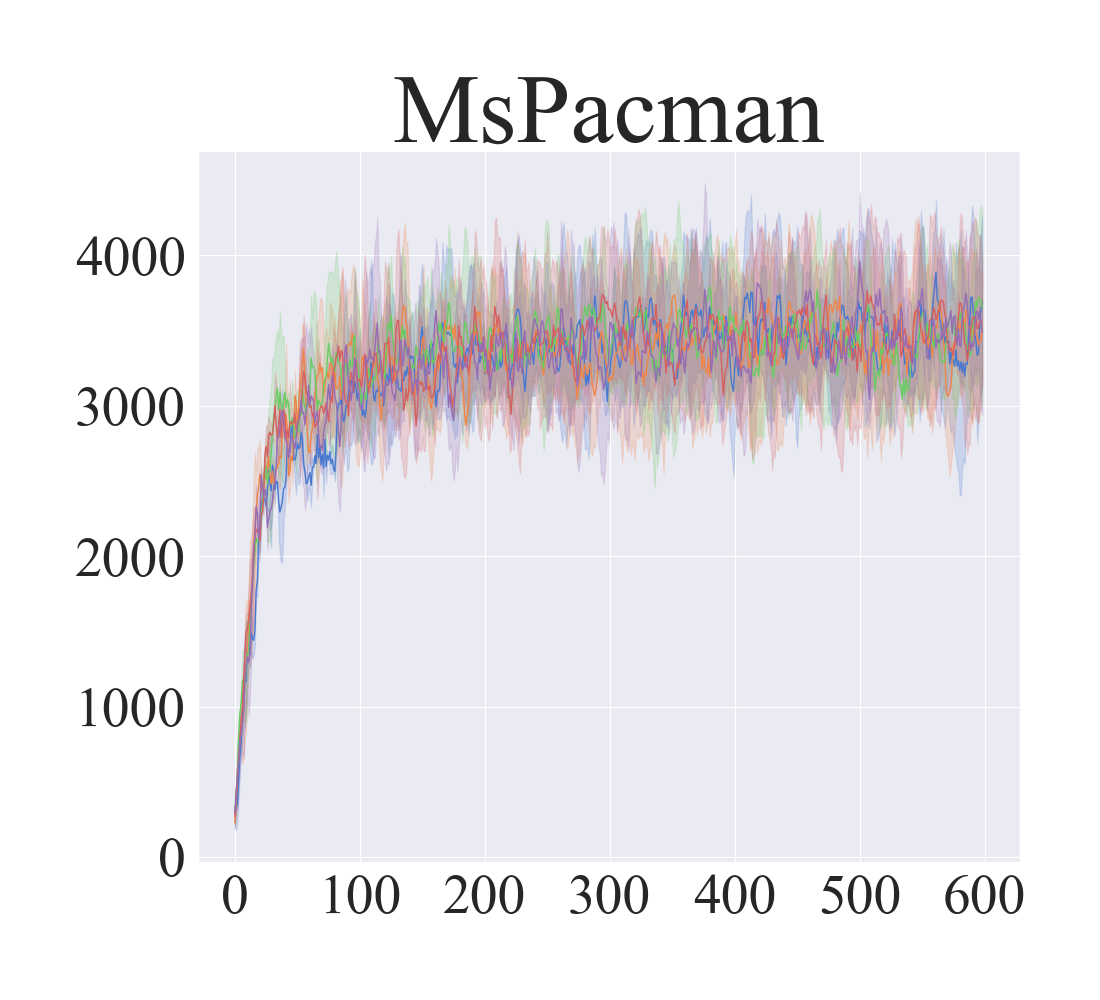}\\
			\end{minipage}%
		}%
		\vspace{-0.6cm}
		
		\subfigure{
			\begin{minipage}[t]{0.166\linewidth}
				\centering
				\includegraphics[width=1.05in]{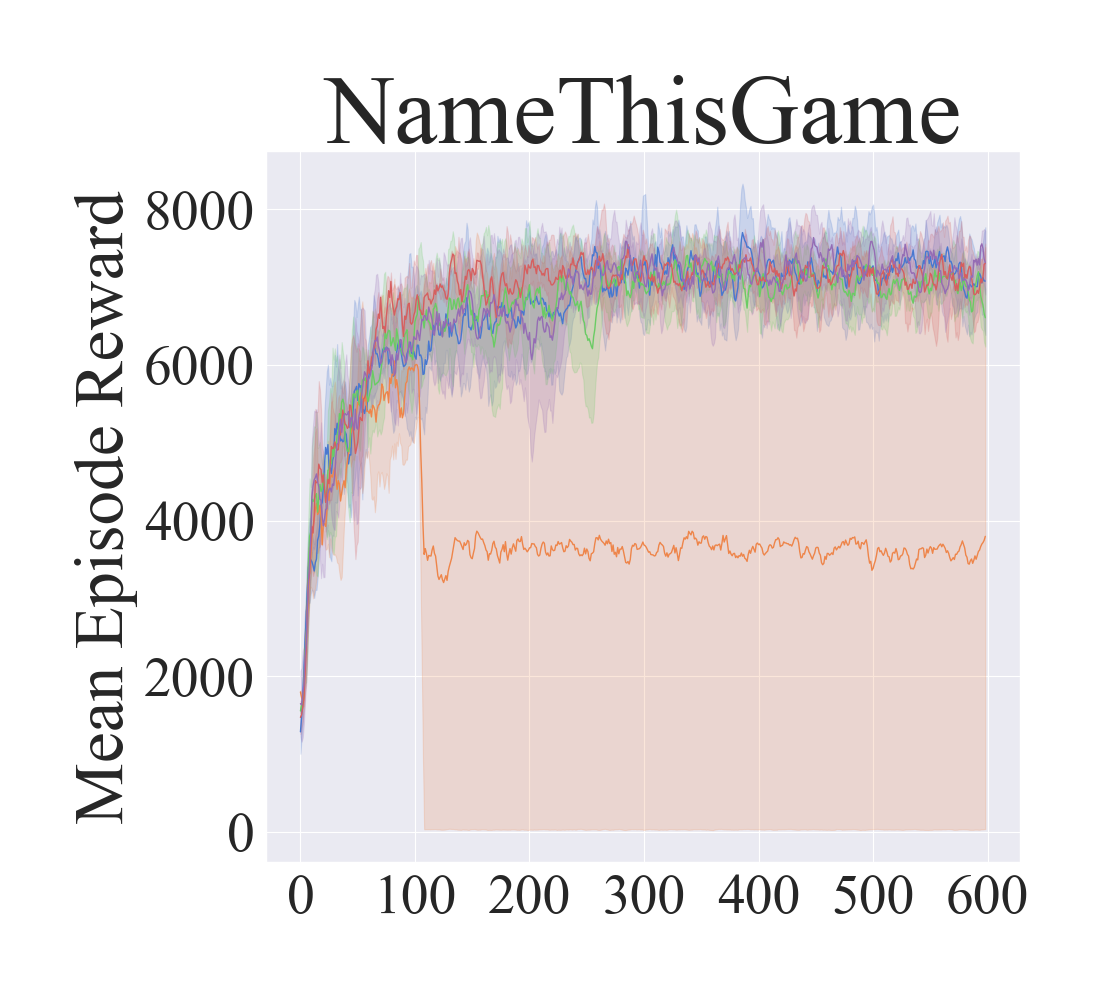}\\
			\end{minipage}%
		}%
		\subfigure{
			\begin{minipage}[t]{0.166\linewidth}
				\centering
				\includegraphics[width=1.05in]{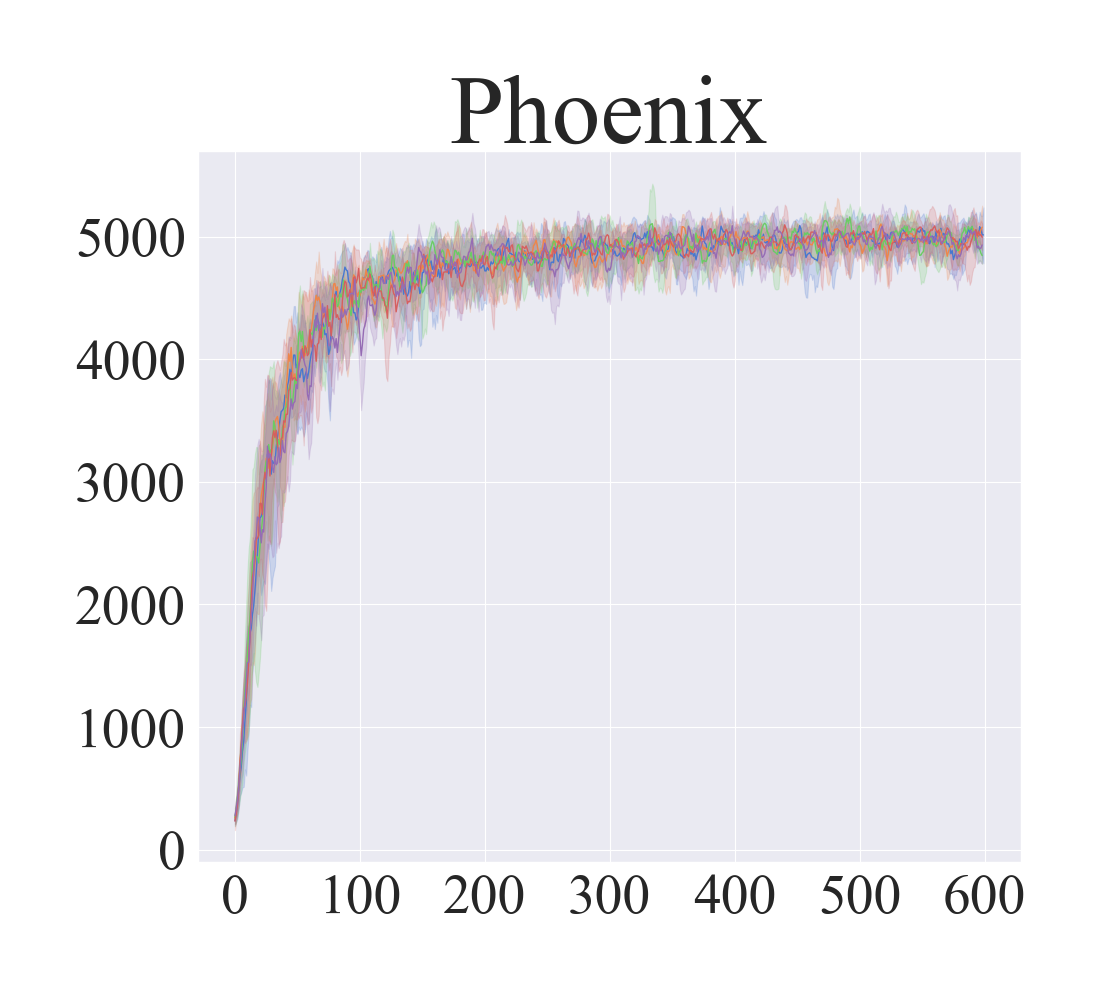}\\
			\end{minipage}%
		}%
		\subfigure{
			\begin{minipage}[t]{0.166\linewidth}
				\centering
				\includegraphics[width=1.05in]{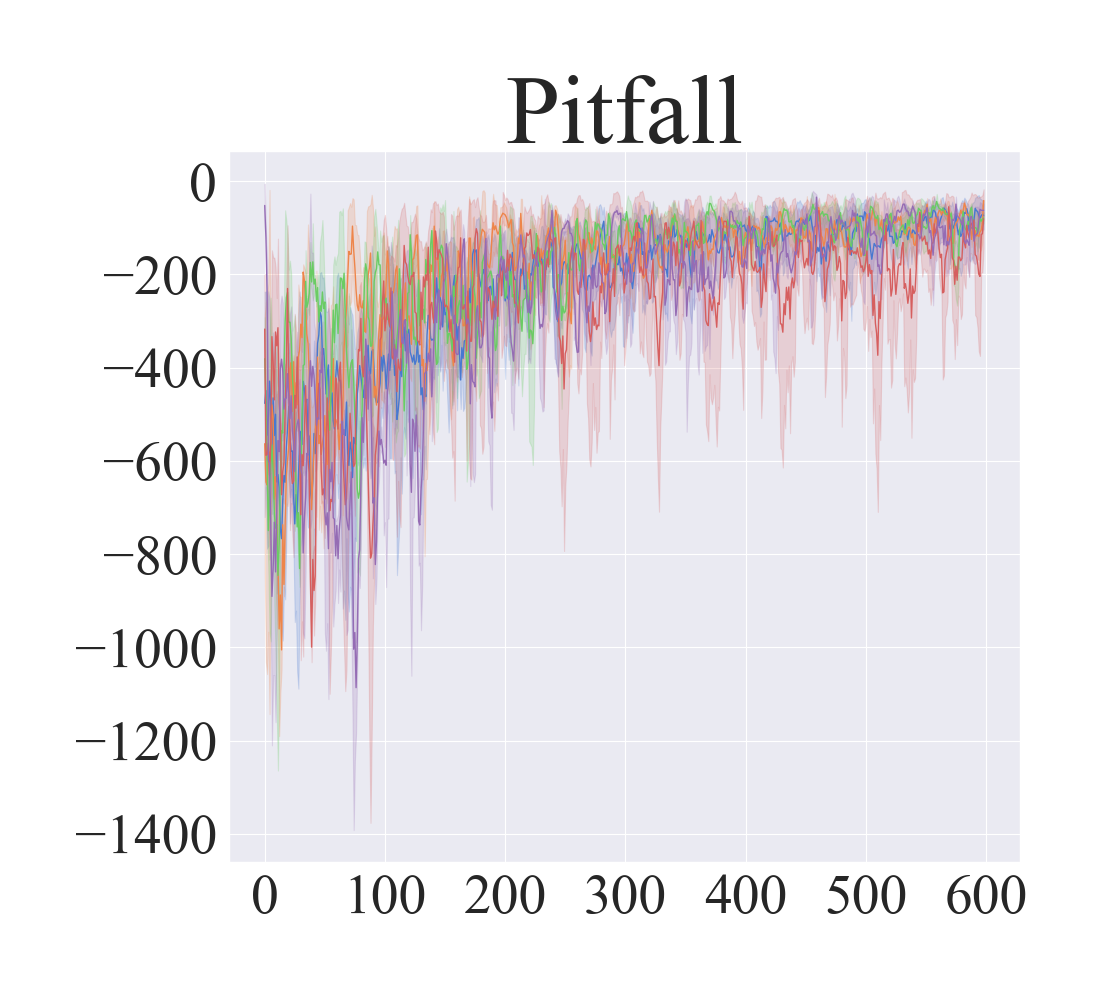}\\
			\end{minipage}%
		}%
		\subfigure{
			\begin{minipage}[t]{0.166\linewidth}
				\centering
				\includegraphics[width=1.05in]{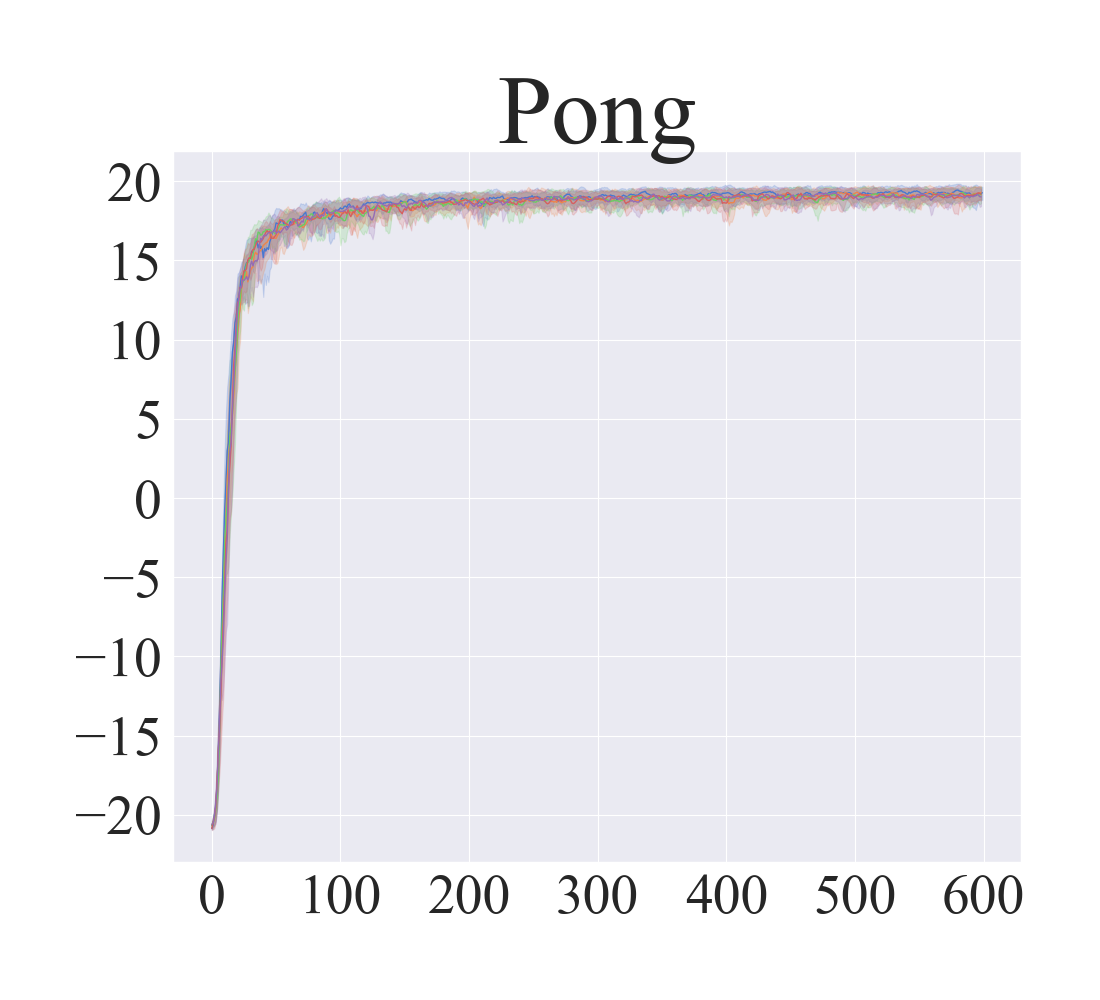}\\
			\end{minipage}%
		}%
		\subfigure{
			\begin{minipage}[t]{0.166\linewidth}
				\centering
				\includegraphics[width=1.05in]{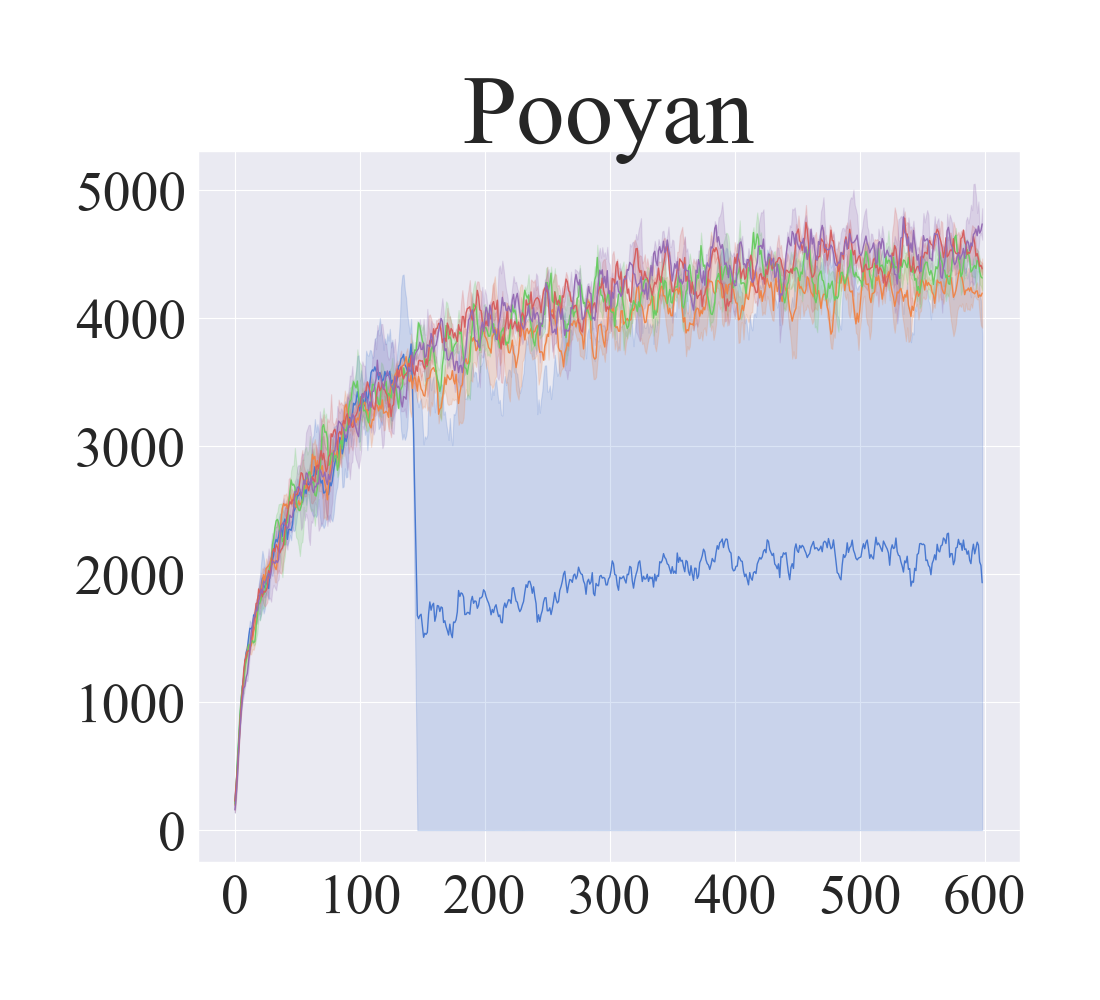}\\
			\end{minipage}%
		}%
		\subfigure{
			\begin{minipage}[t]{0.166\linewidth}
				\centering
				\includegraphics[width=1.05in]{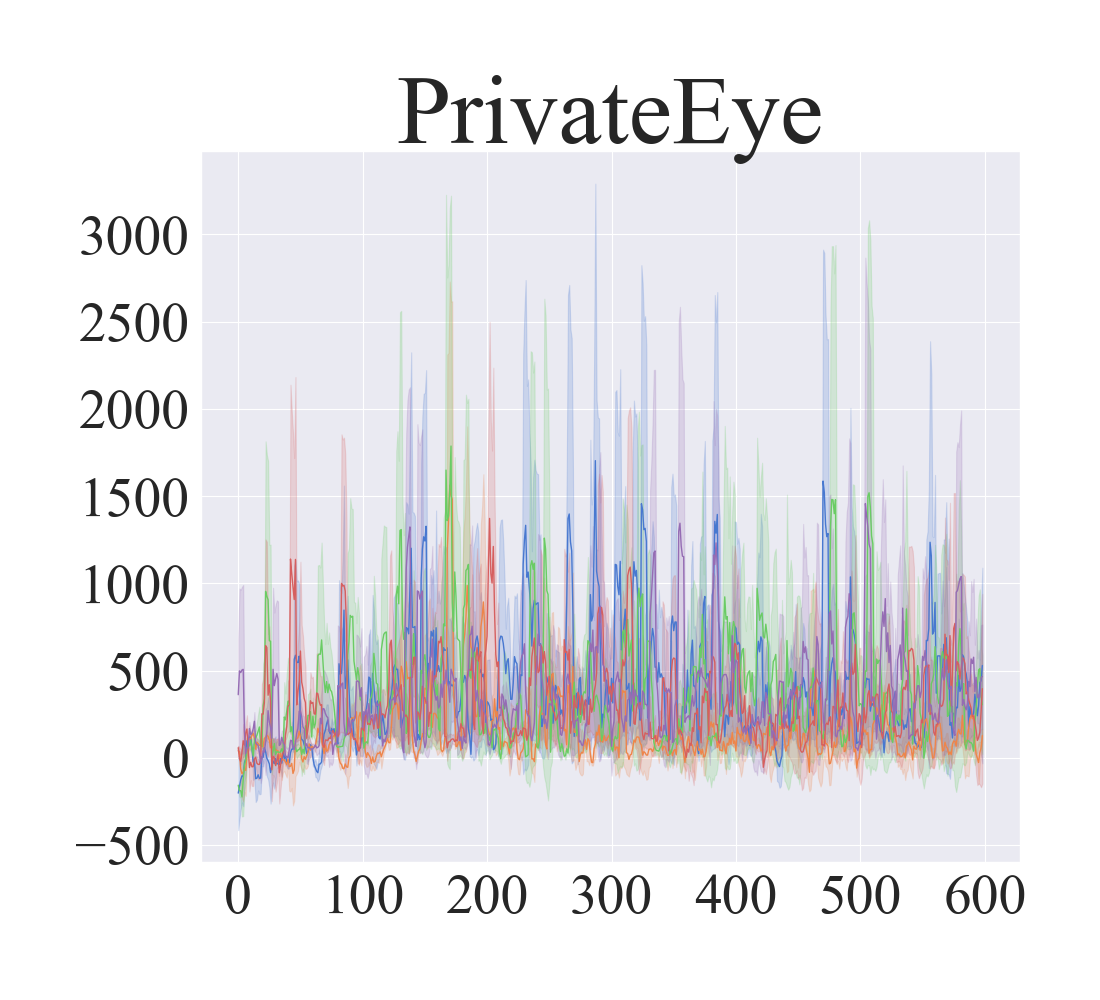}\\
			\end{minipage}%
		}%
		\vspace{-0.6cm}
		
		\subfigure{
			\begin{minipage}[t]{0.166\linewidth}
				\centering
				\includegraphics[width=1.05in]{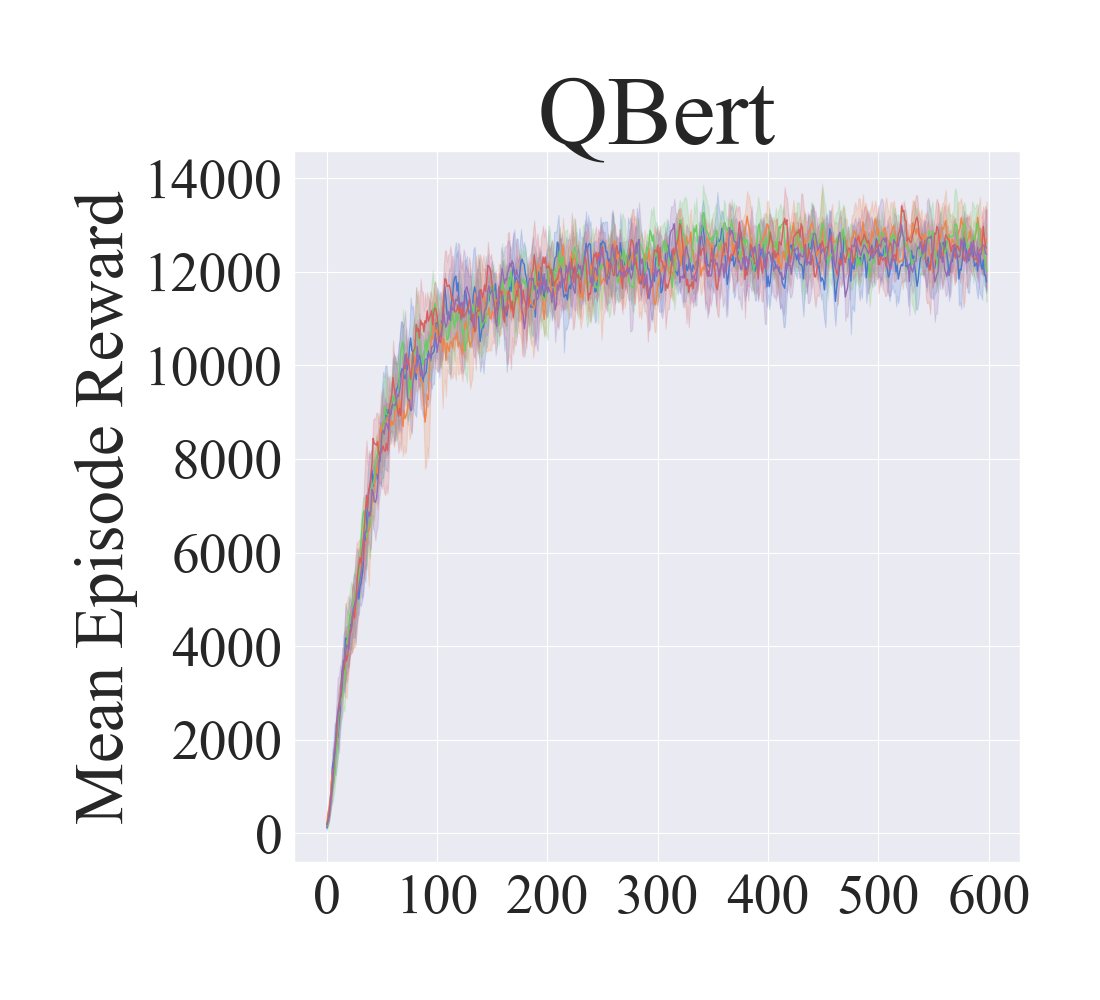}\\
			\end{minipage}%
		}%
		\subfigure{
			\begin{minipage}[t]{0.166\linewidth}
				\centering
				\includegraphics[width=1.05in]{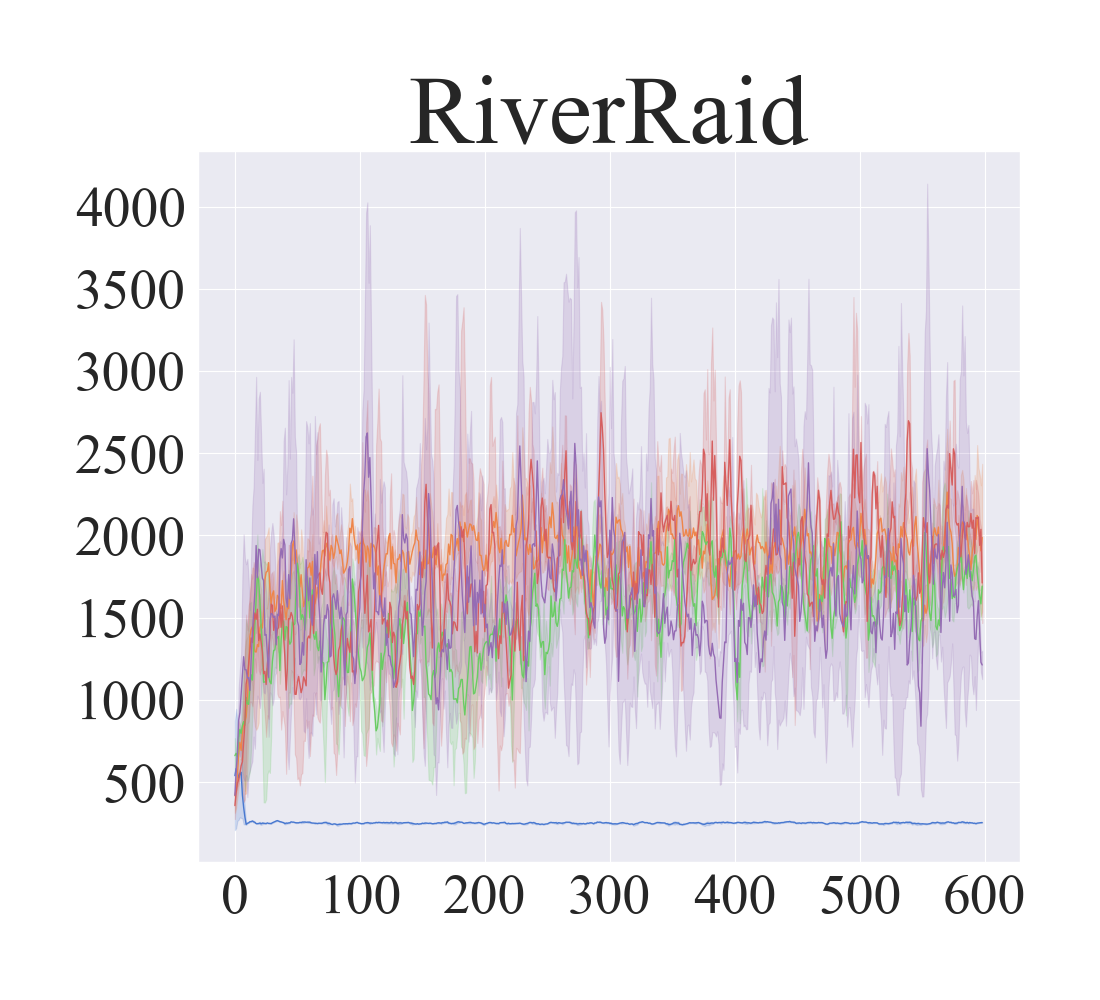}\\
			\end{minipage}%
		}%
		\subfigure{
			\begin{minipage}[t]{0.166\linewidth}
				\centering
				\includegraphics[width=1.05in]{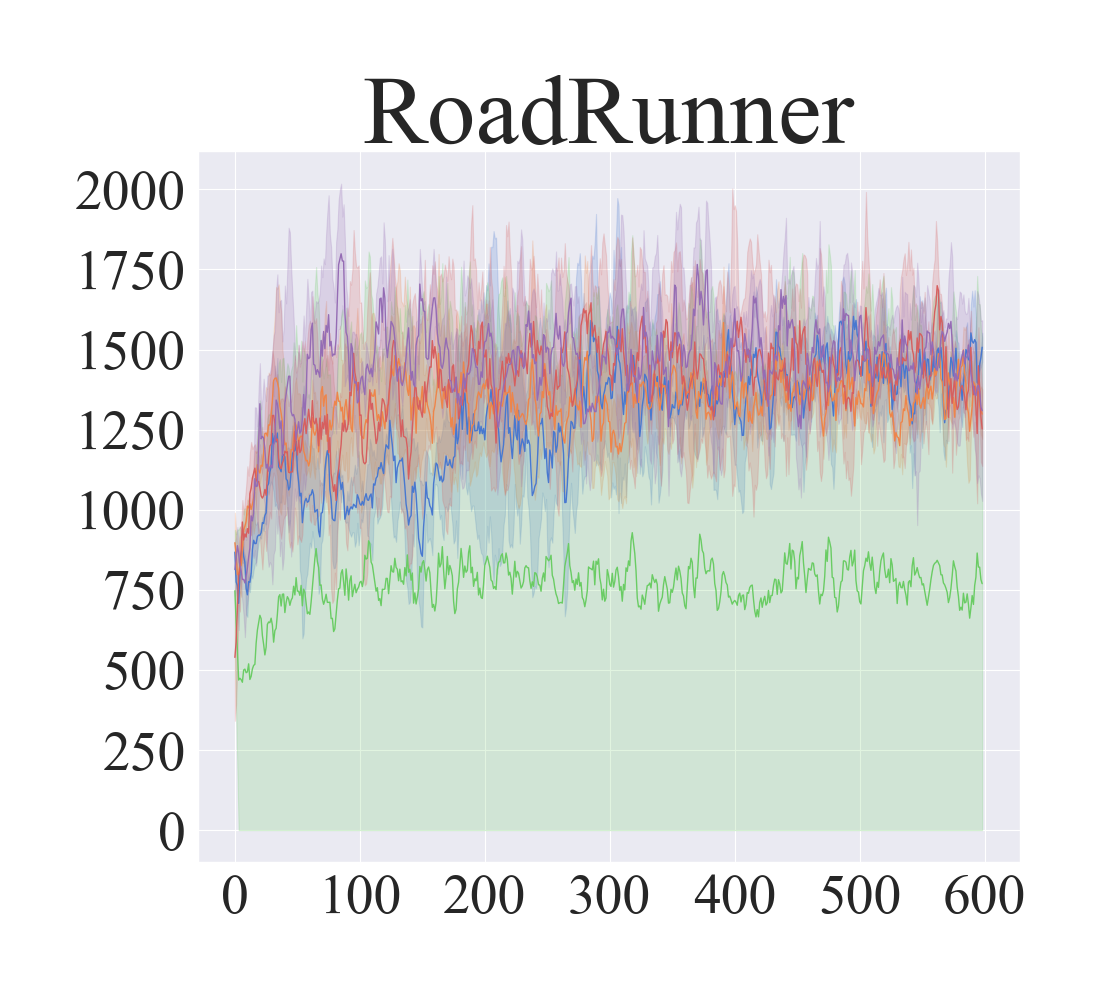}\\
			\end{minipage}%
		}%
		\subfigure{
			\begin{minipage}[t]{0.166\linewidth}
				\centering
				\includegraphics[width=1.05in]{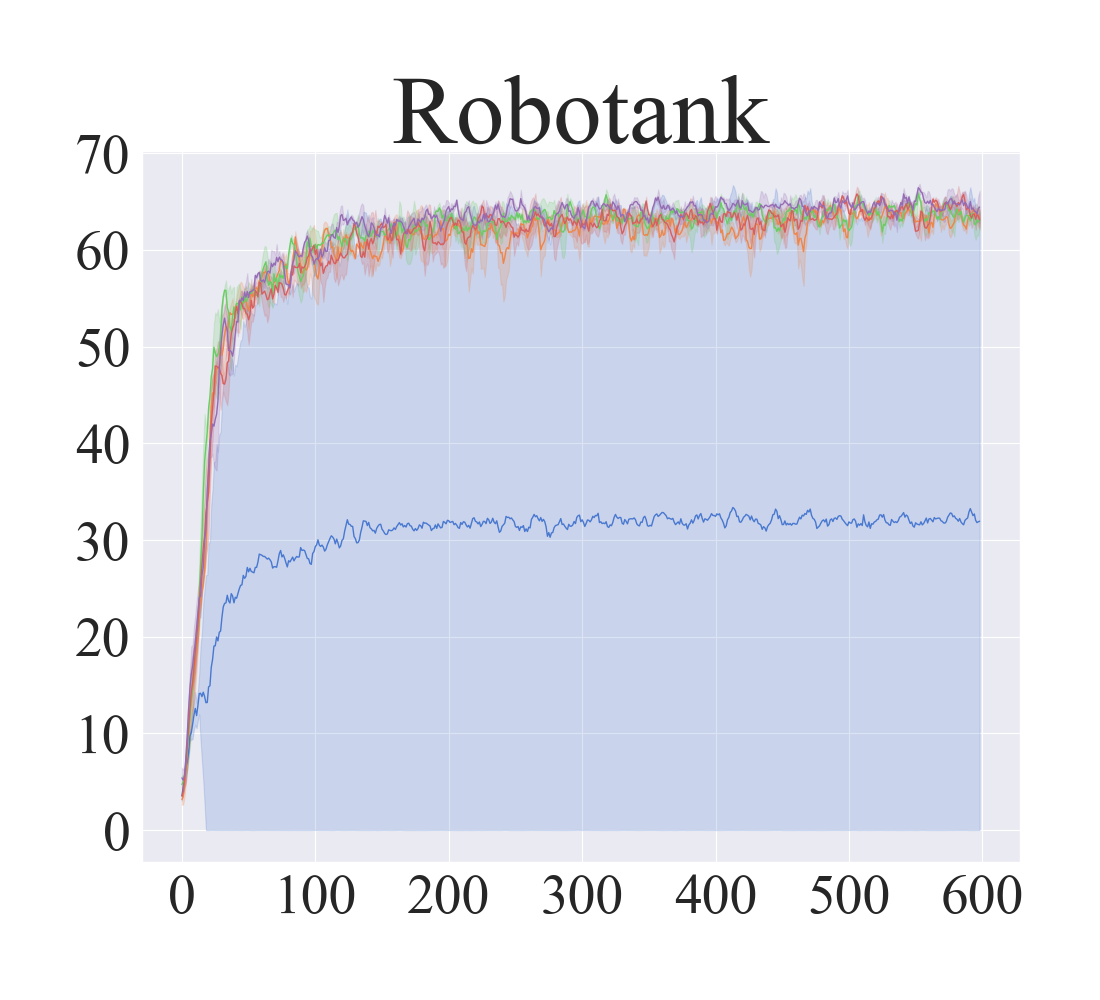}\\
			\end{minipage}%
		}%
		\subfigure{
			\begin{minipage}[t]{0.166\linewidth}
				\centering
				\includegraphics[width=1.05in]{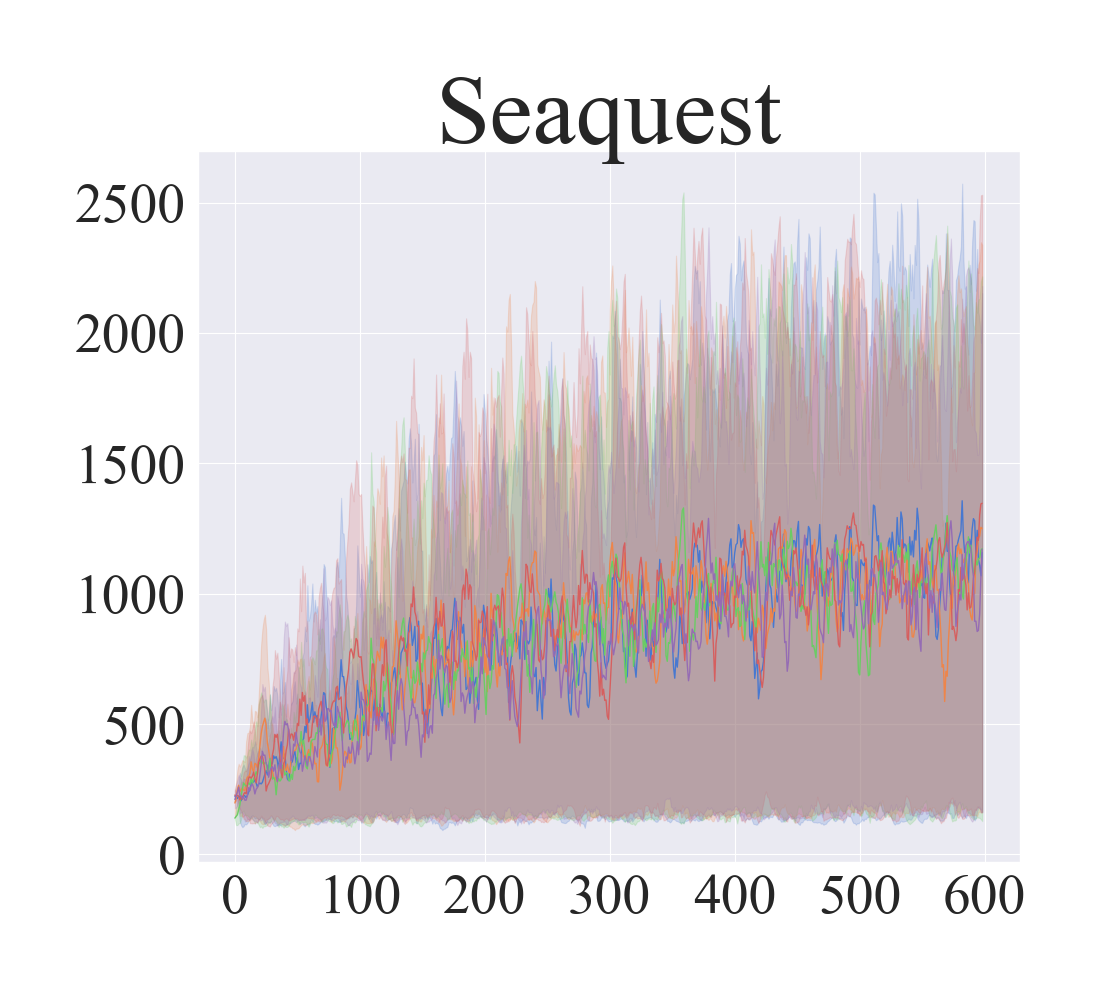}\\
			\end{minipage}%
		}%
		\subfigure{
			\begin{minipage}[t]{0.166\linewidth}
				\centering
				\includegraphics[width=1.05in]{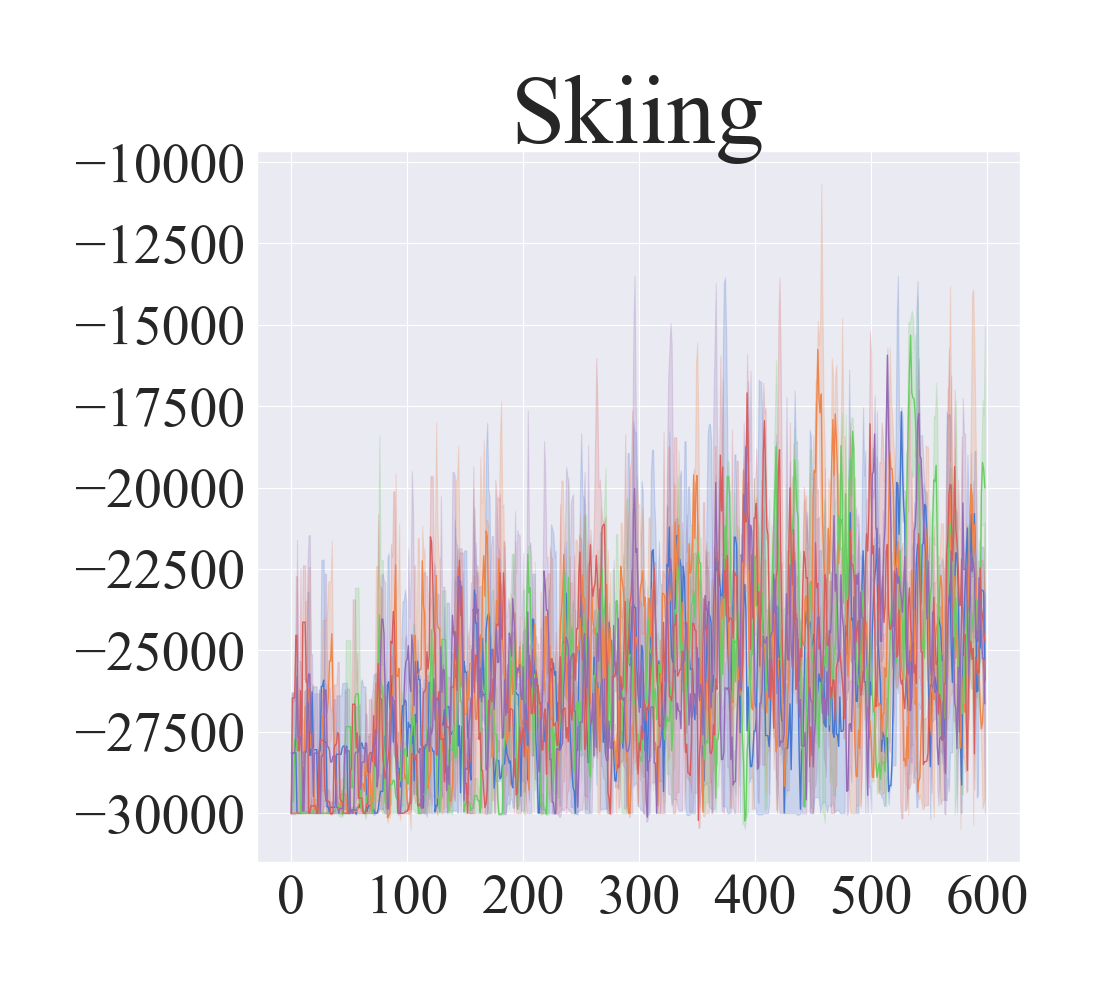}\\
			\end{minipage}%
		}%
		\vspace{-0.6cm}
		
		\subfigure{
			\begin{minipage}[t]{0.166\linewidth}
				\centering
				\includegraphics[width=1.05in]{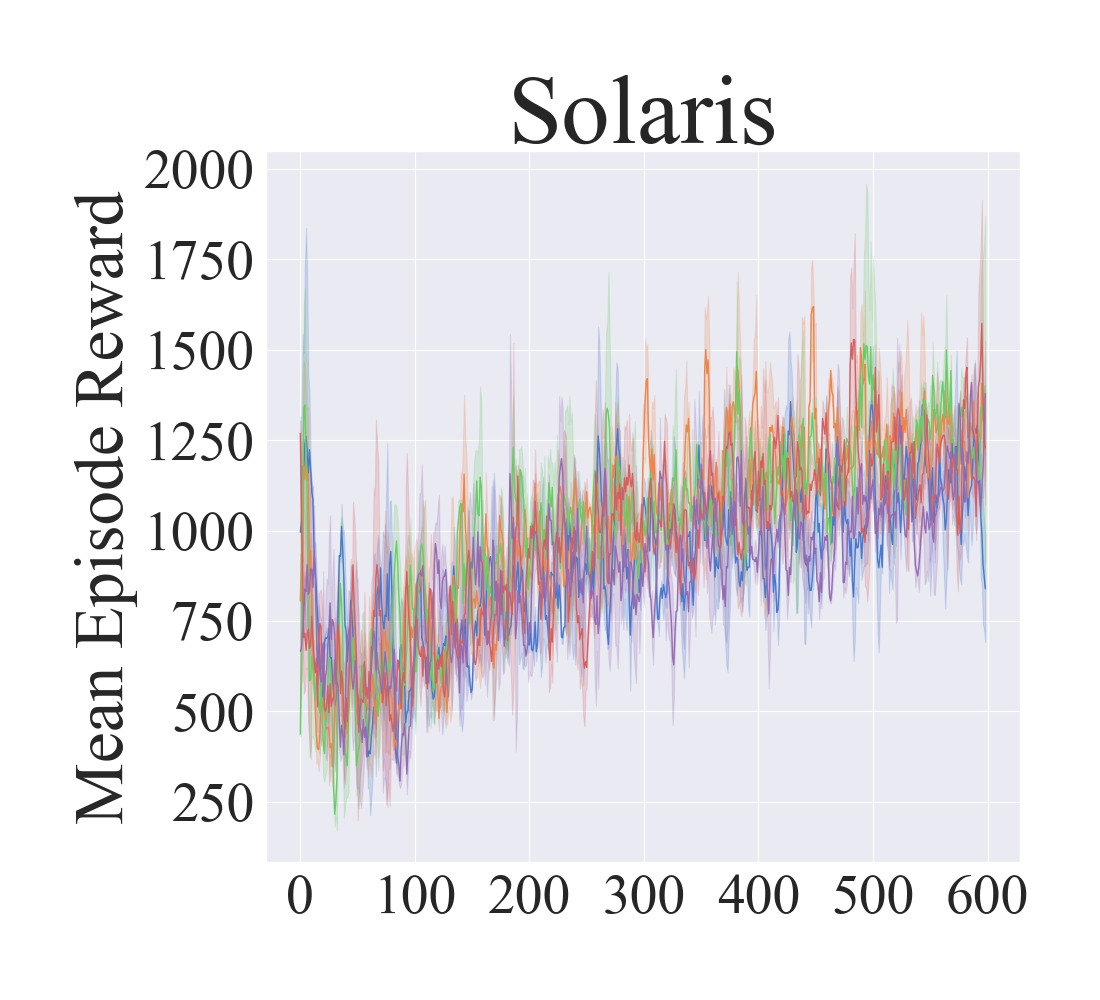}\\
			\end{minipage}%
		}%
		\subfigure{
			\begin{minipage}[t]{0.166\linewidth}
				\centering
				\includegraphics[width=1.05in]{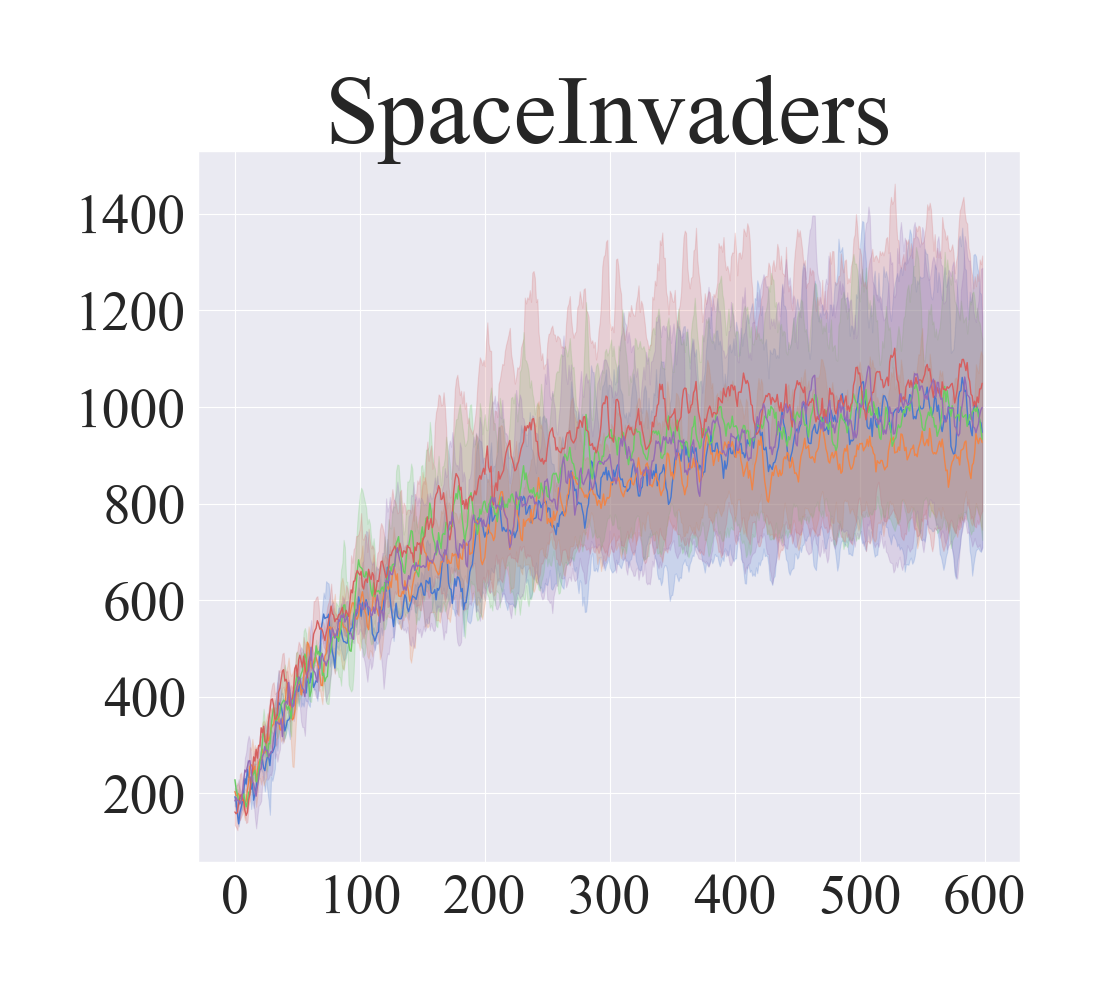}\\
			\end{minipage}%
		}%
		\subfigure{
			\begin{minipage}[t]{0.166\linewidth}
				\centering
				\includegraphics[width=1.05in]{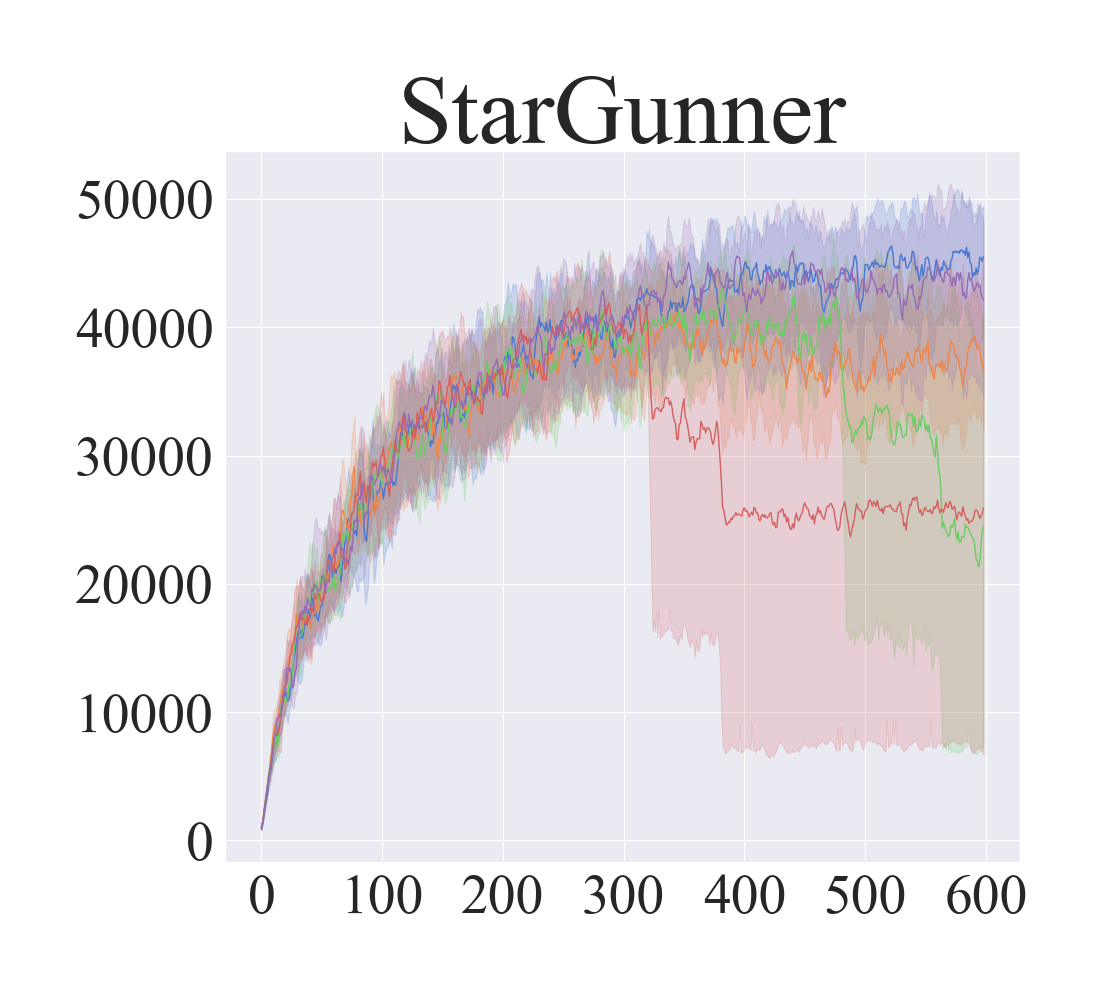}\\
			\end{minipage}%
		}%
		\subfigure{
			\begin{minipage}[t]{0.166\linewidth}
				\centering
				\includegraphics[width=1.05in]{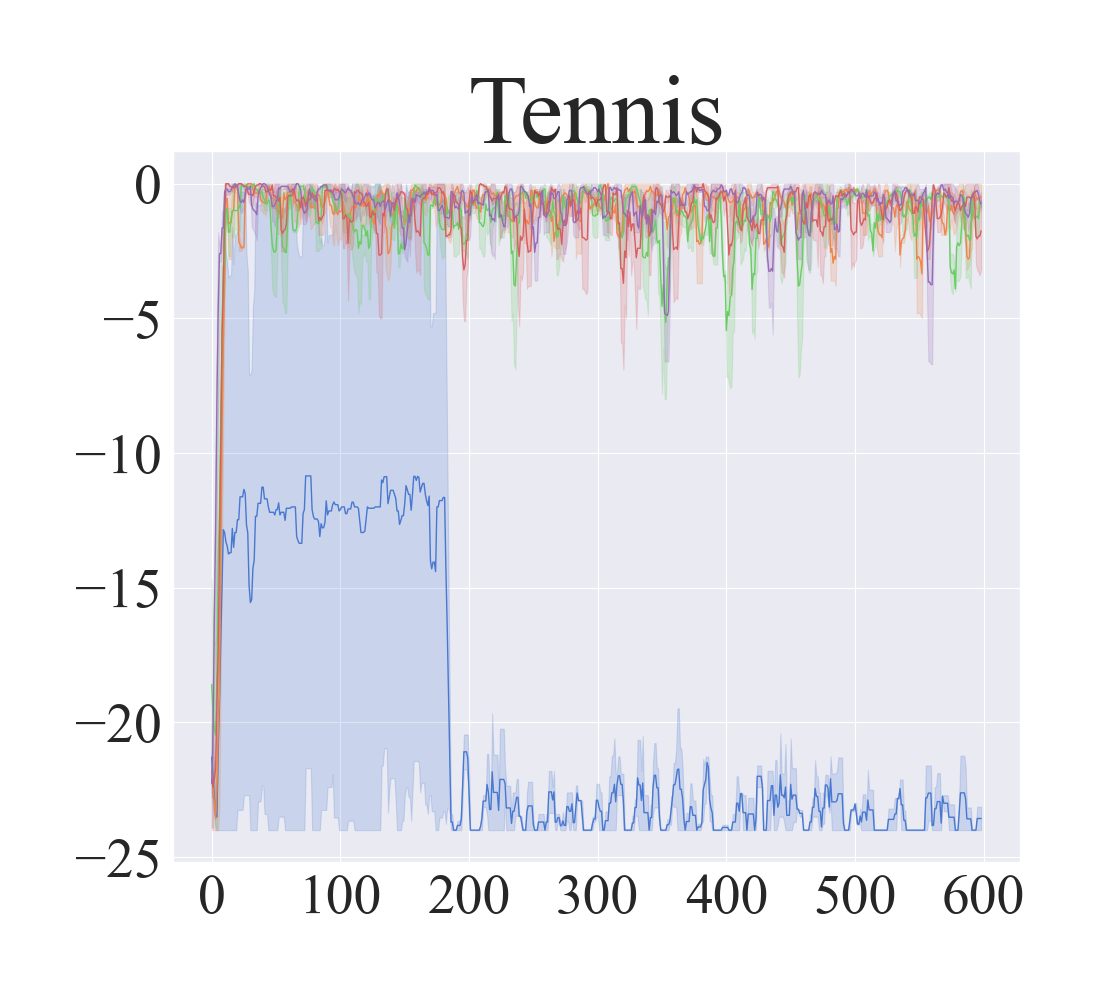}\\
			\end{minipage}%
		}%
		\subfigure{
			\begin{minipage}[t]{0.166\linewidth}
				\centering
				\includegraphics[width=1.05in]{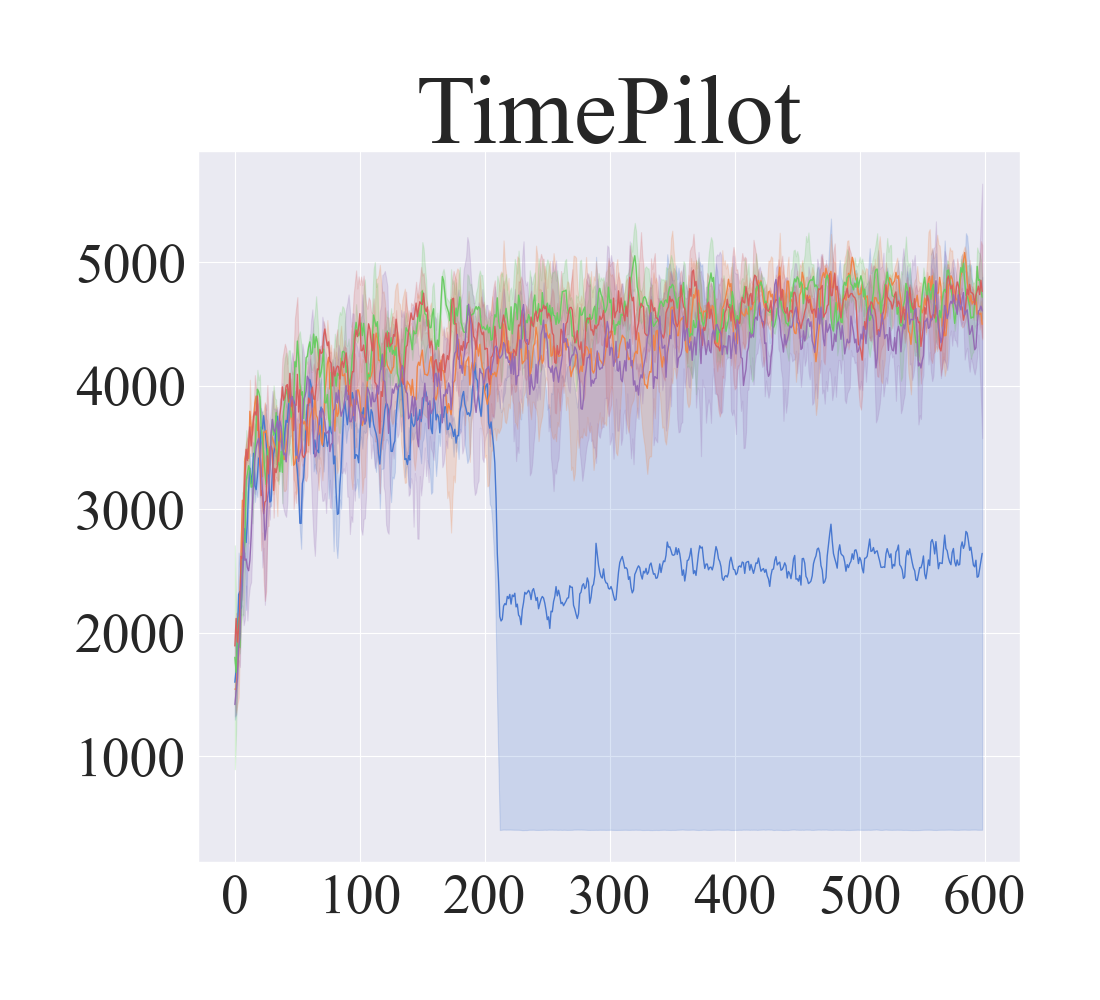}\\
			\end{minipage}%
		}%
		\subfigure{
			\begin{minipage}[t]{0.166\linewidth}
				\centering
				\includegraphics[width=1.05in]{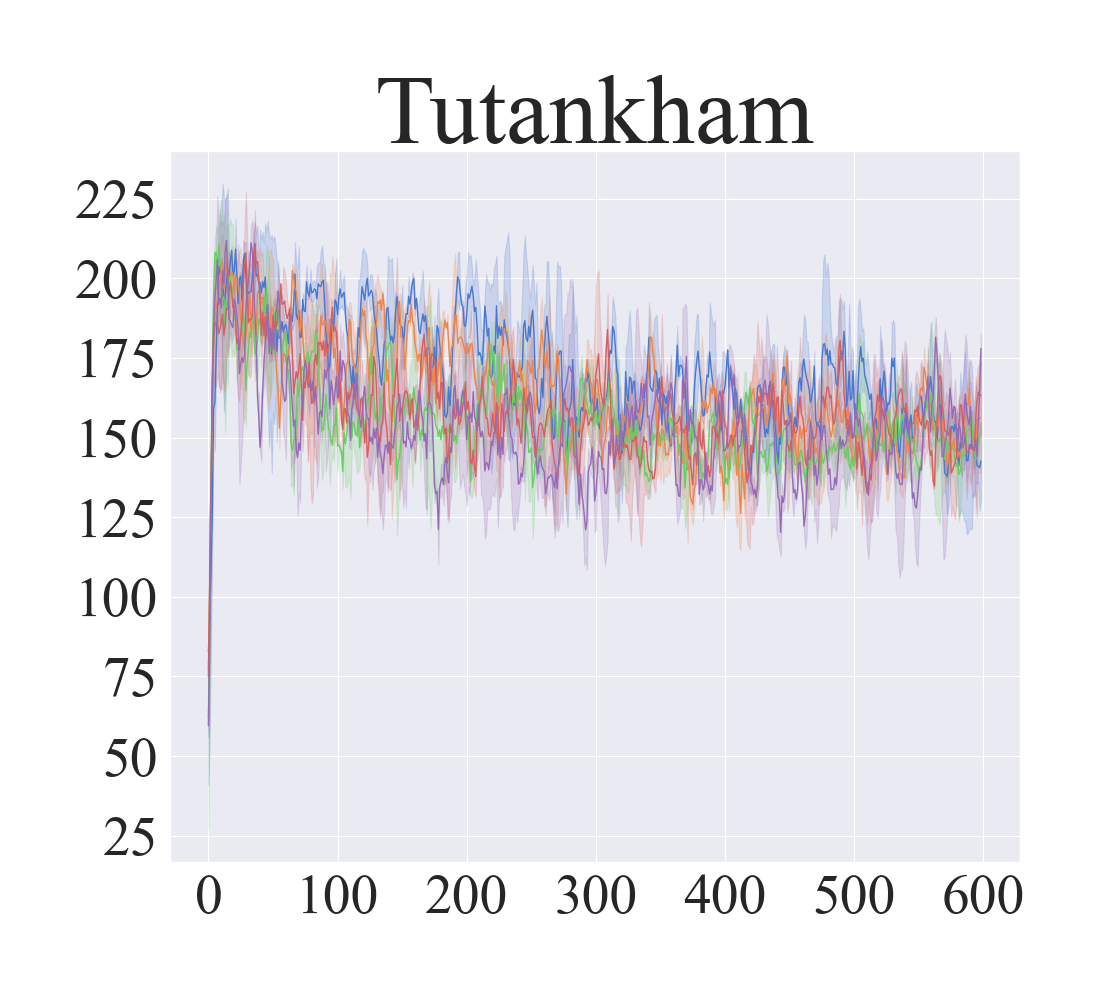}\\
			\end{minipage}%
		}%
		\vspace{-0.6cm}
		
		\subfigure{
			\begin{minipage}[t]{0.166\linewidth}
				\centering
				\includegraphics[width=1.05in]{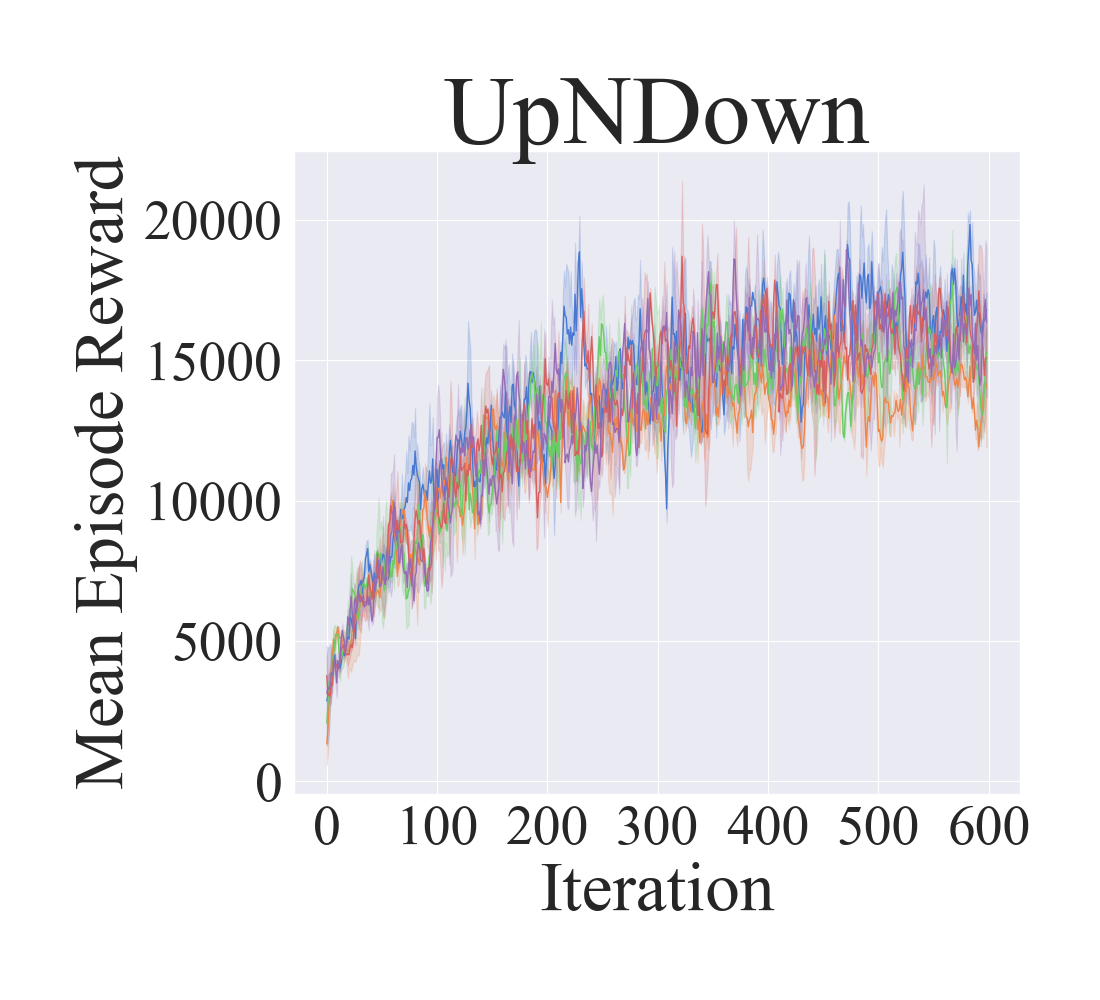}\\
			\end{minipage}%
		}%
		\subfigure{
			\begin{minipage}[t]{0.166\linewidth}
				\centering
				\includegraphics[width=1.05in]{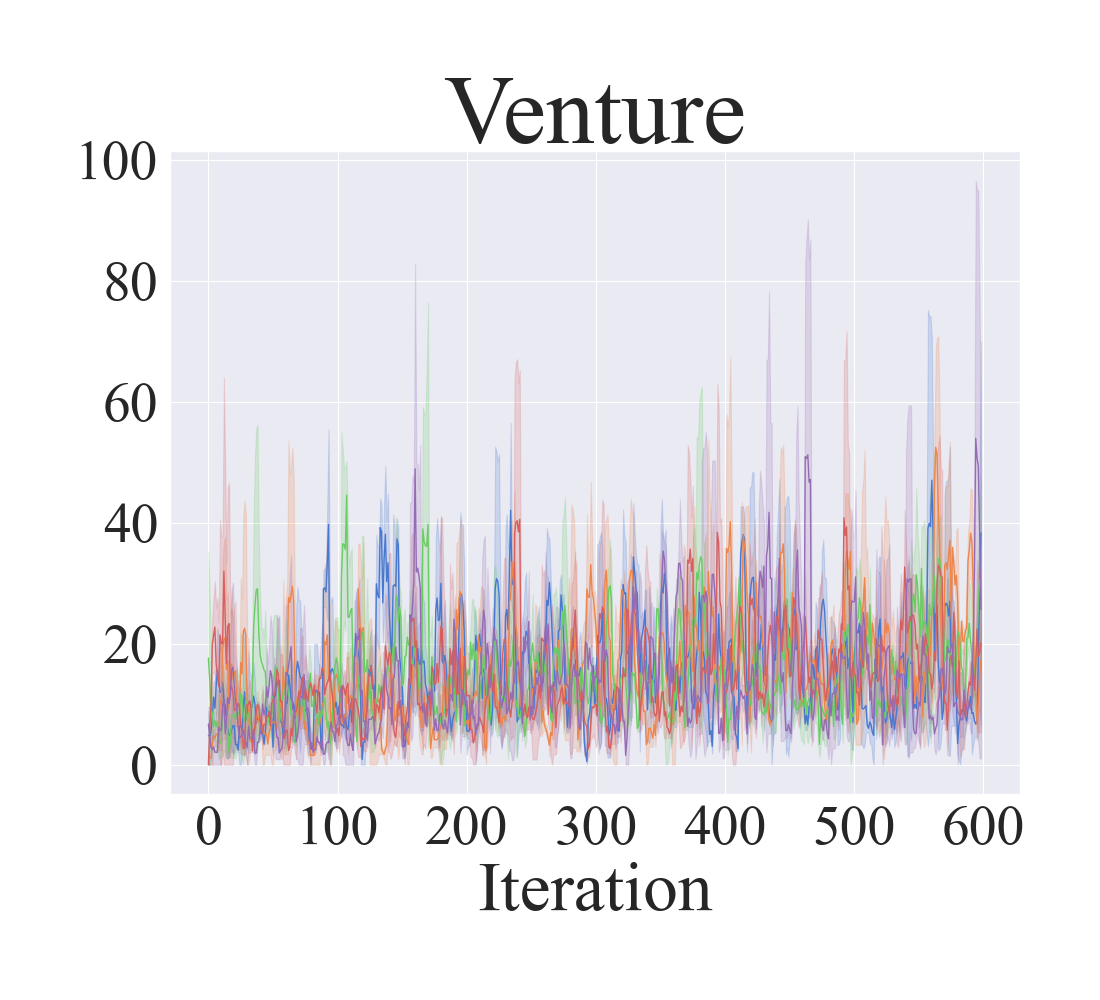}\\
			\end{minipage}%
		}%
		\subfigure{
			\begin{minipage}[t]{0.166\linewidth}
				\centering
				\includegraphics[width=1.05in]{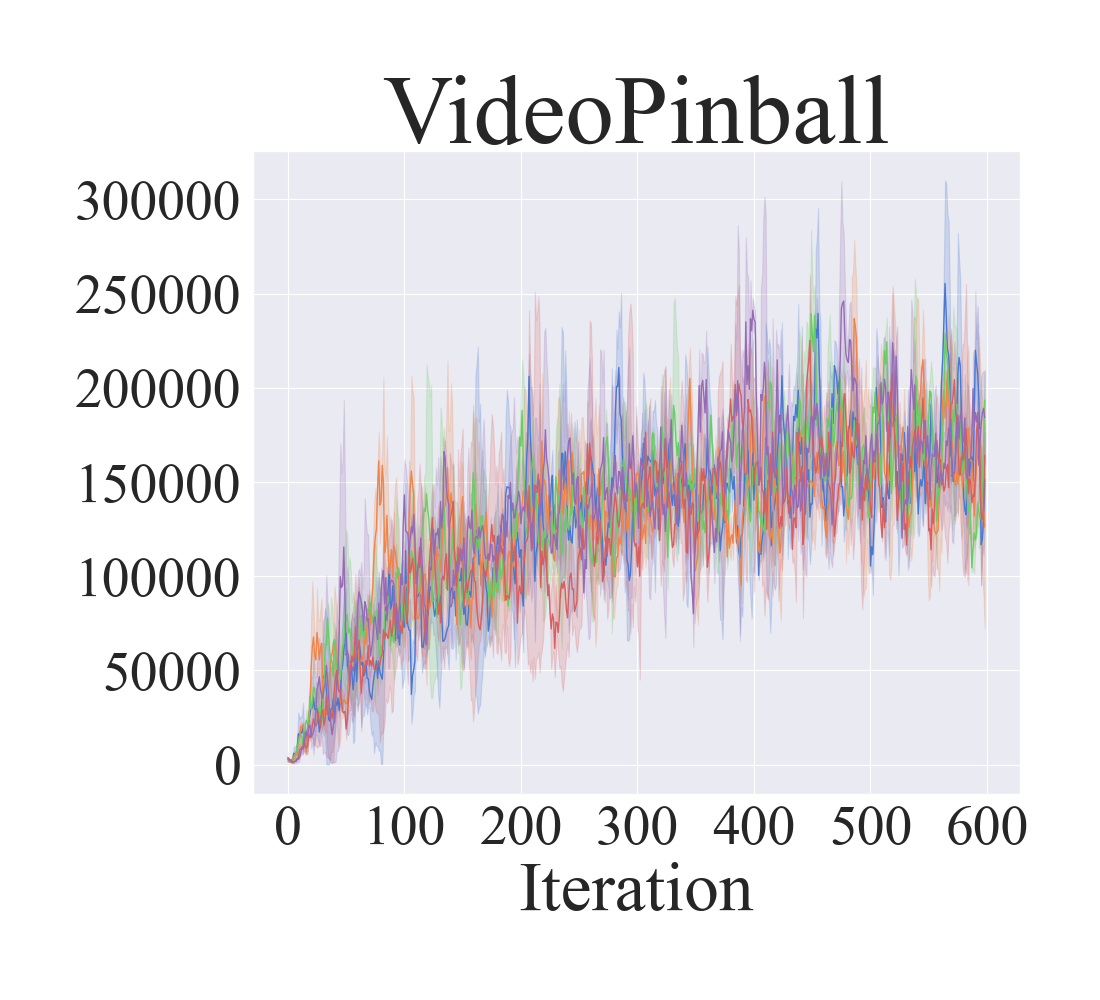}\\
			\end{minipage}%
		}%
		\subfigure{
			\begin{minipage}[t]{0.166\linewidth}
				\centering
				\includegraphics[width=1.05in]{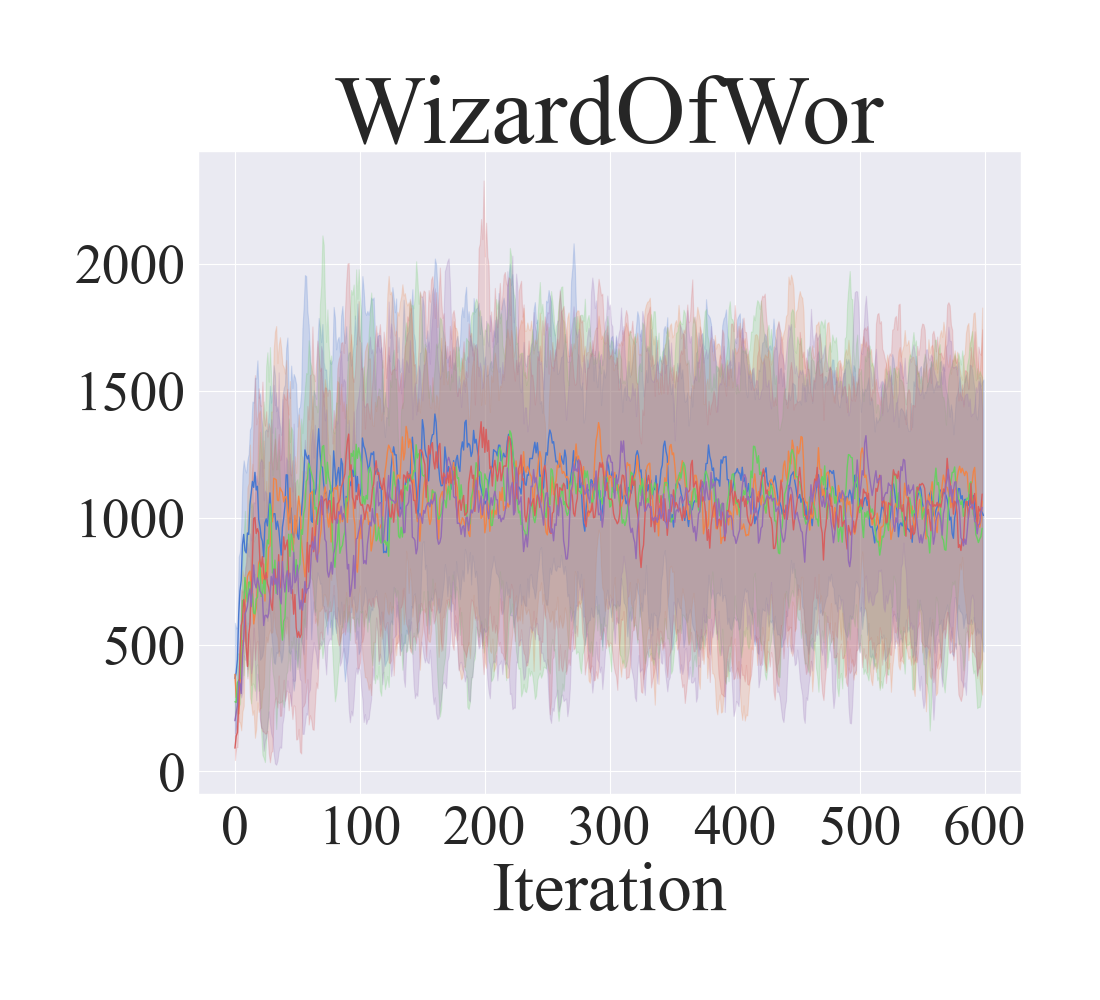}\\
			\end{minipage}%
		}%
		\subfigure{
			\begin{minipage}[t]{0.166\linewidth}
				\centering
				\includegraphics[width=1.05in]{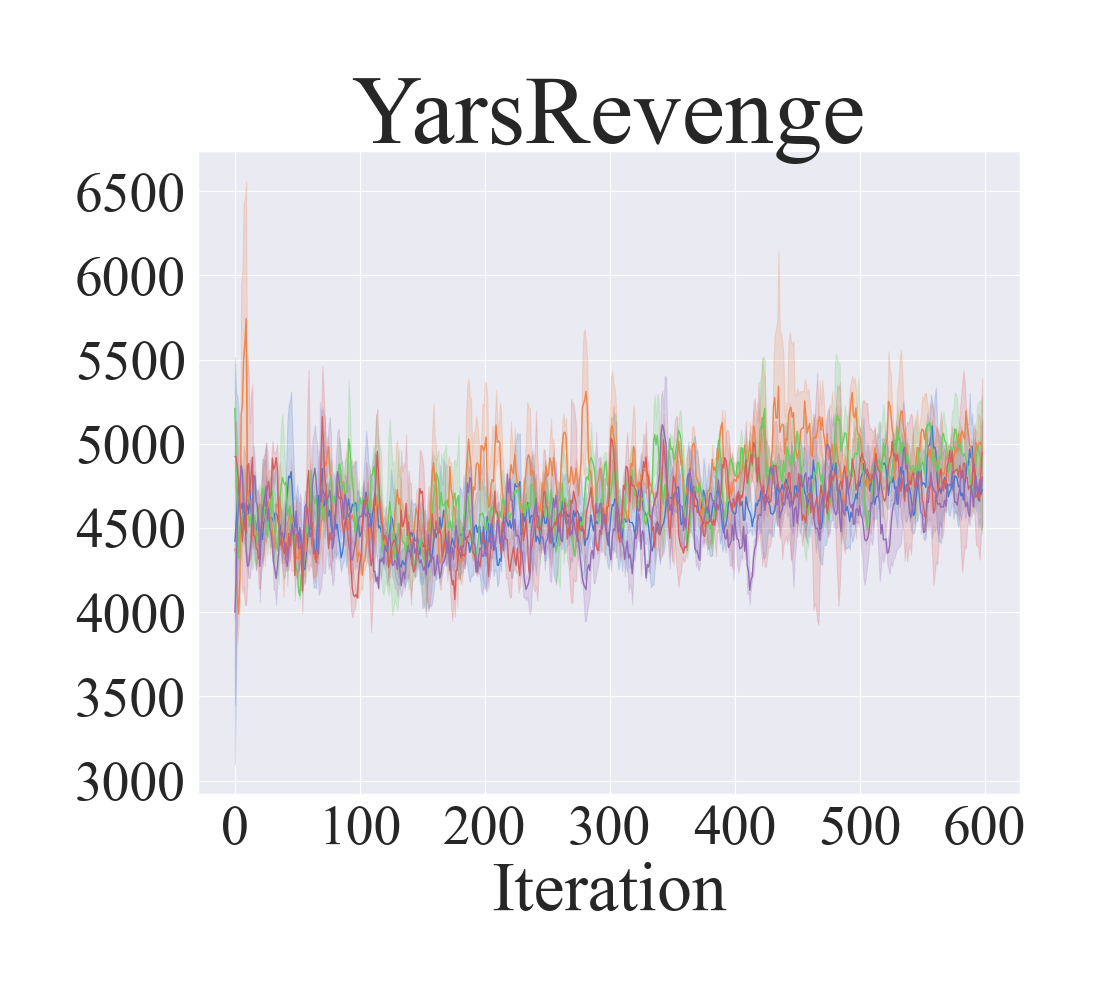}\\
			\end{minipage}%
		}%
		\subfigure{
			\begin{minipage}[t]{0.166\linewidth}
				\centering
				\includegraphics[width=1.05in]{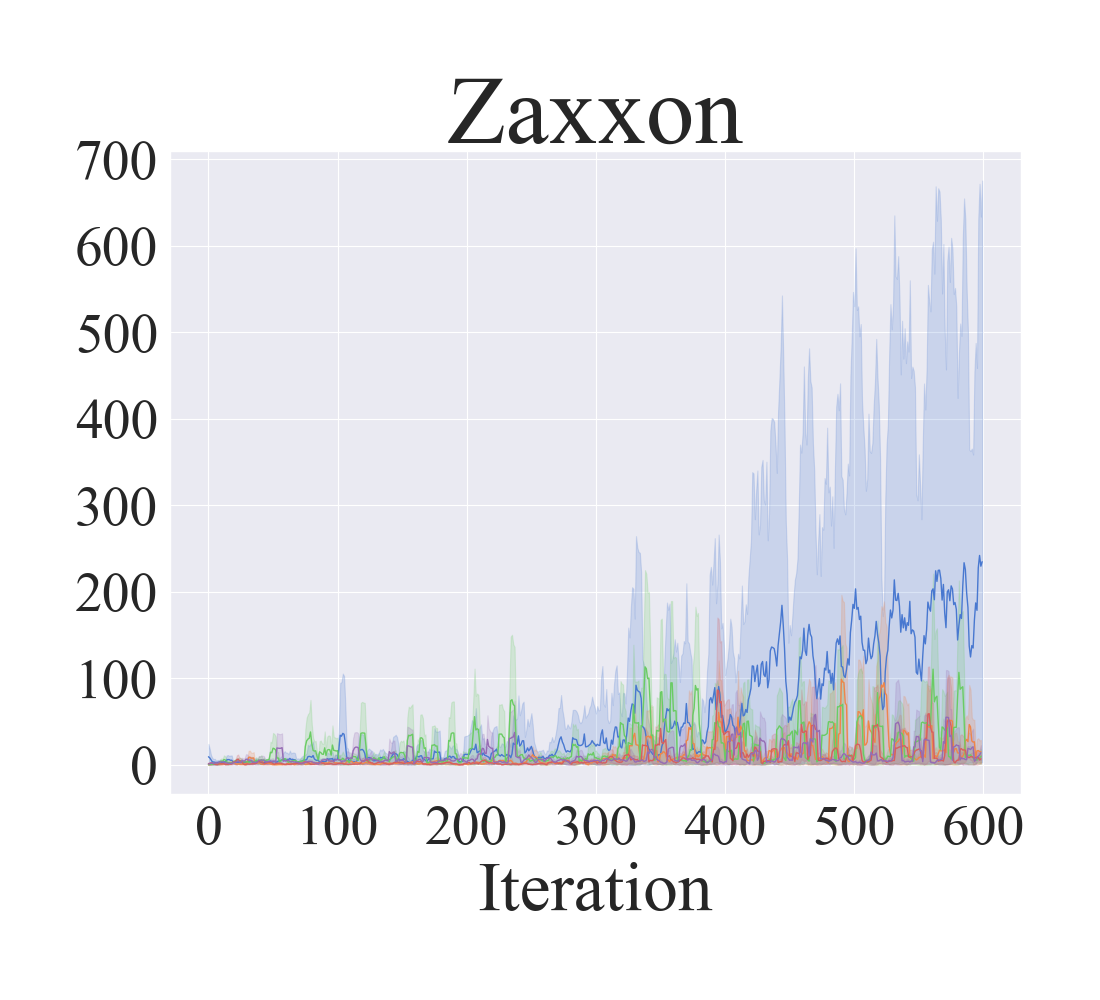}\\
			\end{minipage}%
		}%

		\centering
		\caption{\textbf{Learning curves of all $60$ Atari $2600$ games on high dataset}}
		\label{fig: Learning curves of all $60$ Atari $2600$ games on high dataset1}
								
	\end{figure*}

	Please refer Fig. \ref{fig: Learning curves of all $60$ Atari $2600$ games on high dataset} and Fig. \ref{fig: Learning curves of all $60$ Atari $2600$ games on high dataset1}.

	\subsection{Comparison on different datasets}
\begin{figure*}[!htb]
		\centering
		\subfigure{
			\begin{minipage}[t]{0.333\linewidth}
				\centering
				\includegraphics[width=2.3in]{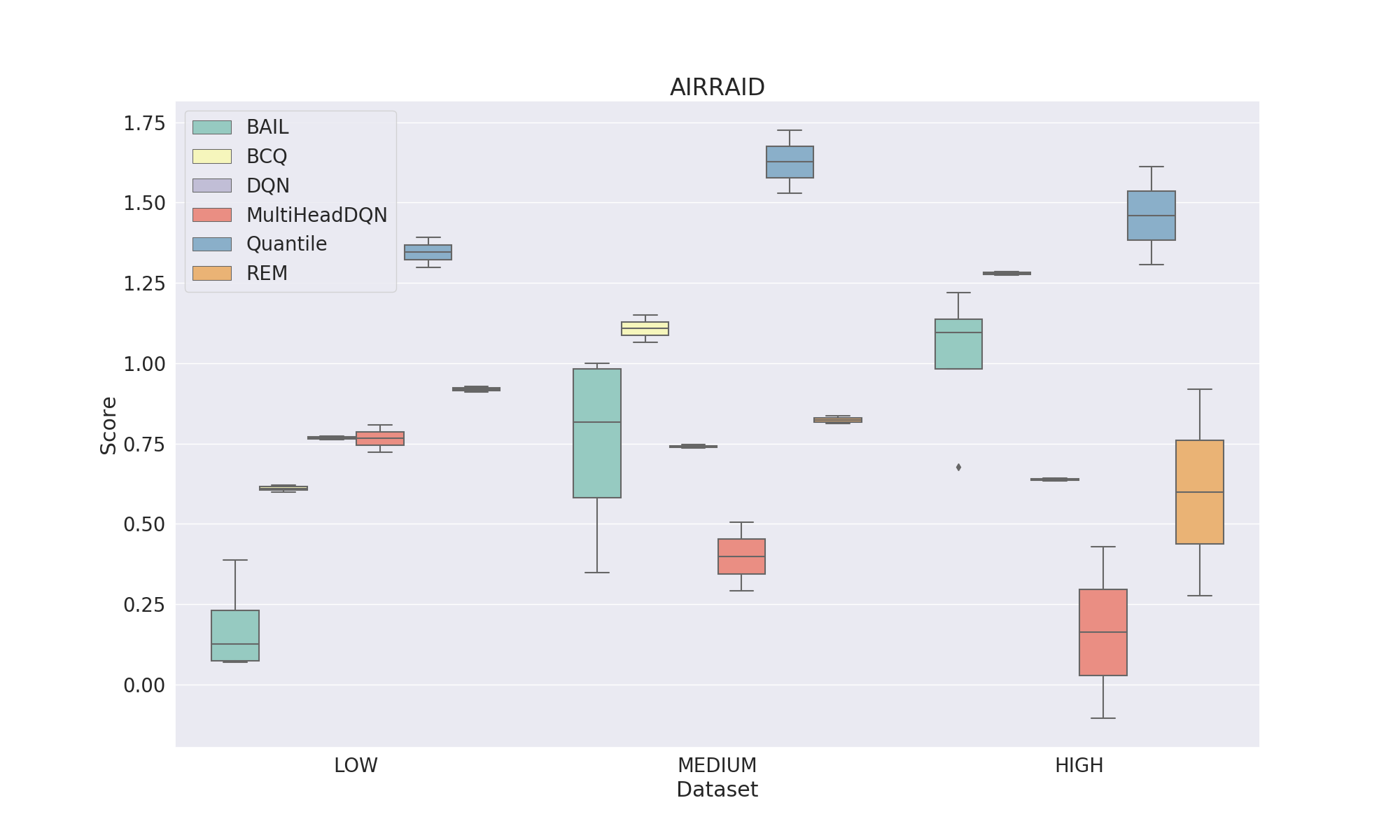}\\
				\vspace{0.01cm}
			\end{minipage}%
		}%
		\subfigure{
			\begin{minipage}[t]{0.333\linewidth}
				\centering
				\includegraphics[width=2.3in]{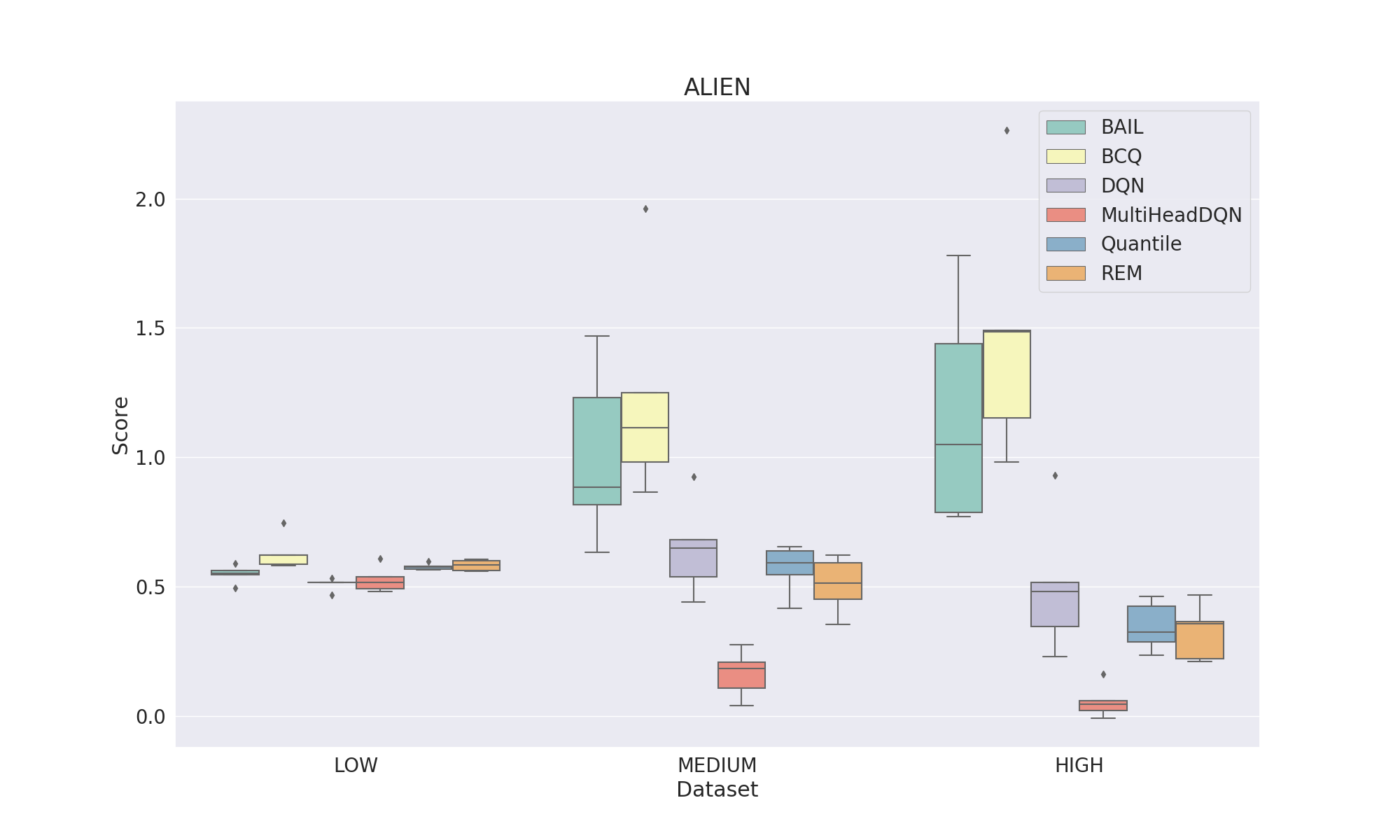}\\
				\vspace{0.01cm}
			\end{minipage}%
		}%
		\subfigure{
			\begin{minipage}[t]{0.333\linewidth}
				\centering
				\includegraphics[width=2.3in]{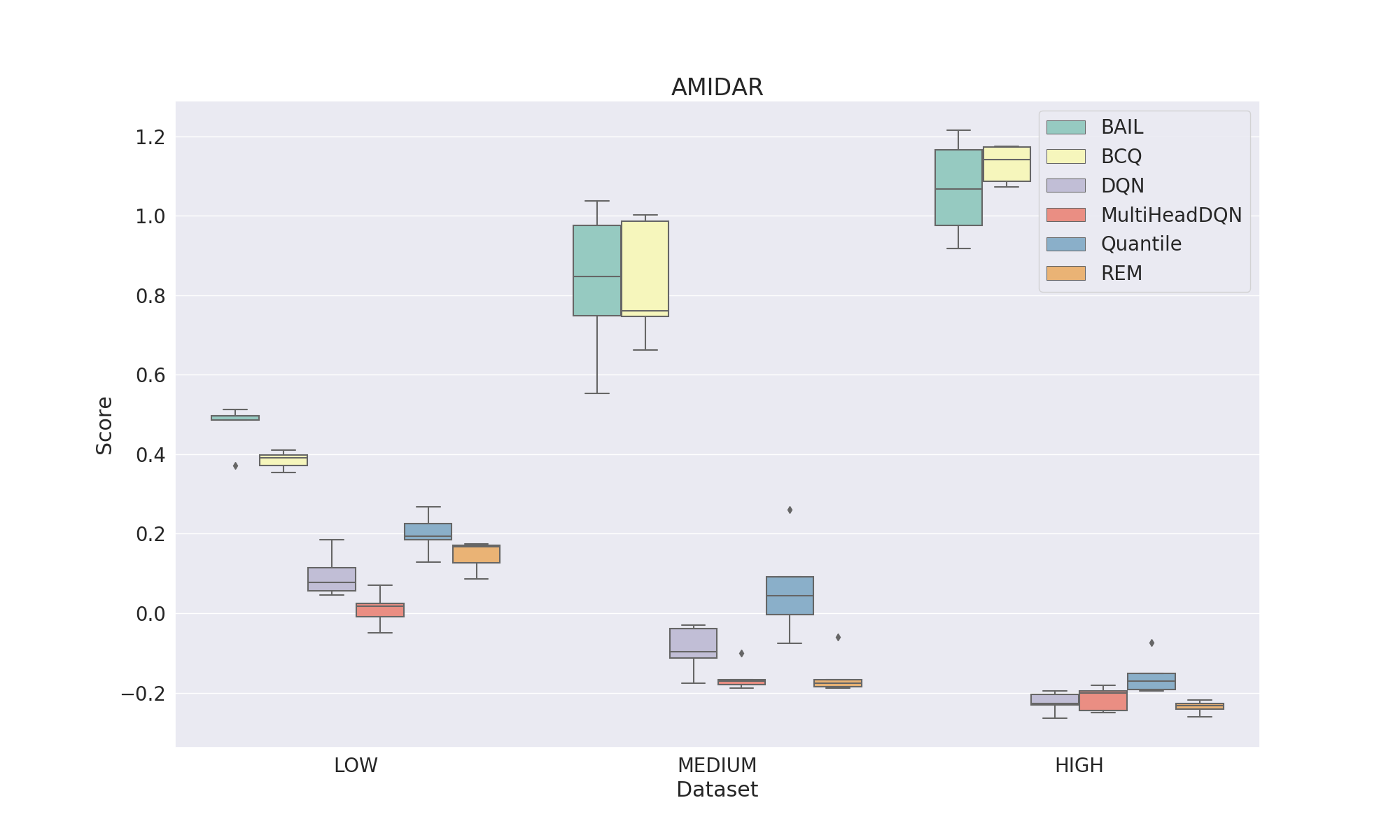}\\
				\vspace{0.01cm}
			\end{minipage}%
		}%

		\subfigure{
			\begin{minipage}[t]{0.333\linewidth}
				\centering
				\includegraphics[width=2.3in]{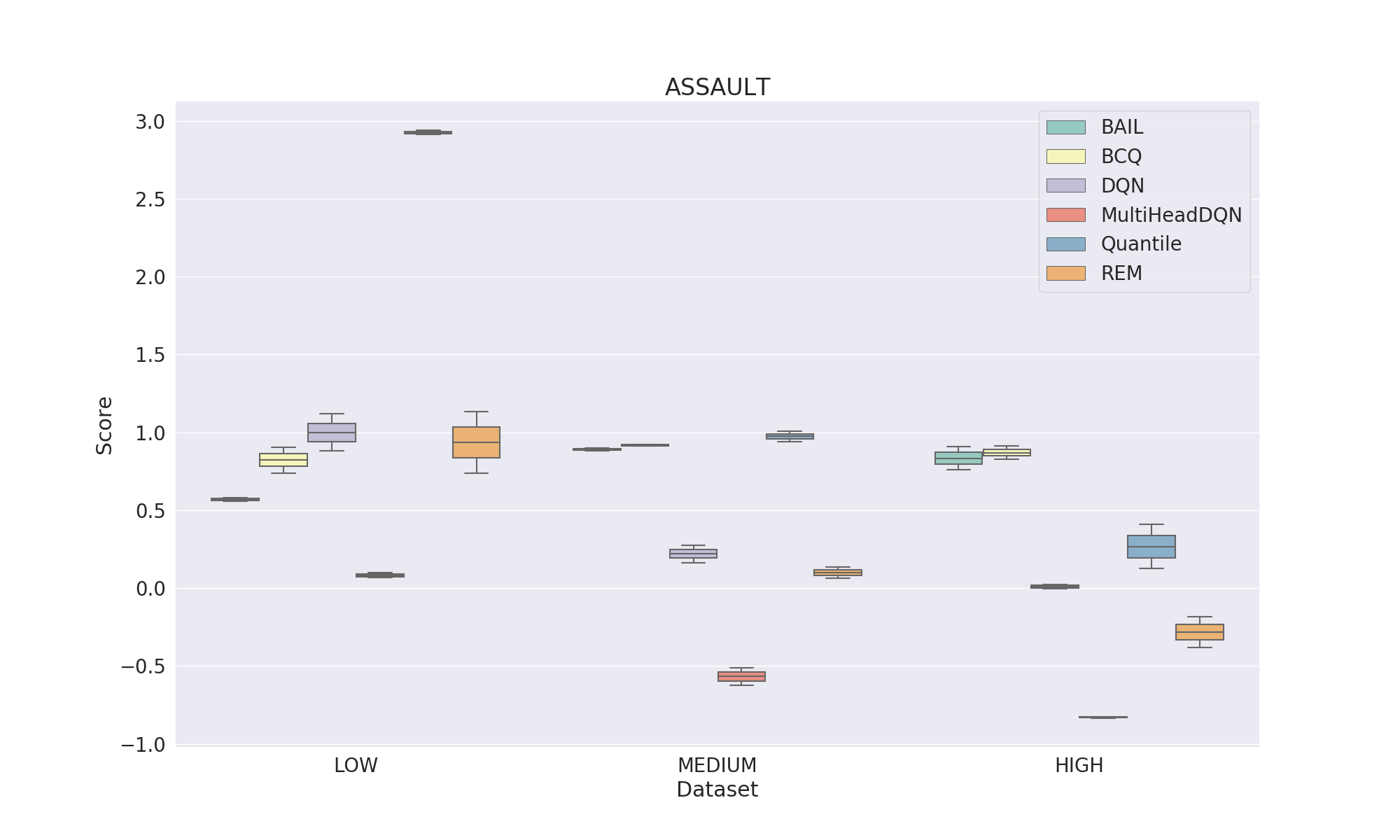}\\
				\vspace{0.01cm}
			\end{minipage}%
		}%
		\subfigure{
			\begin{minipage}[t]{0.333\linewidth}
				\centering
				\includegraphics[width=2.3in]{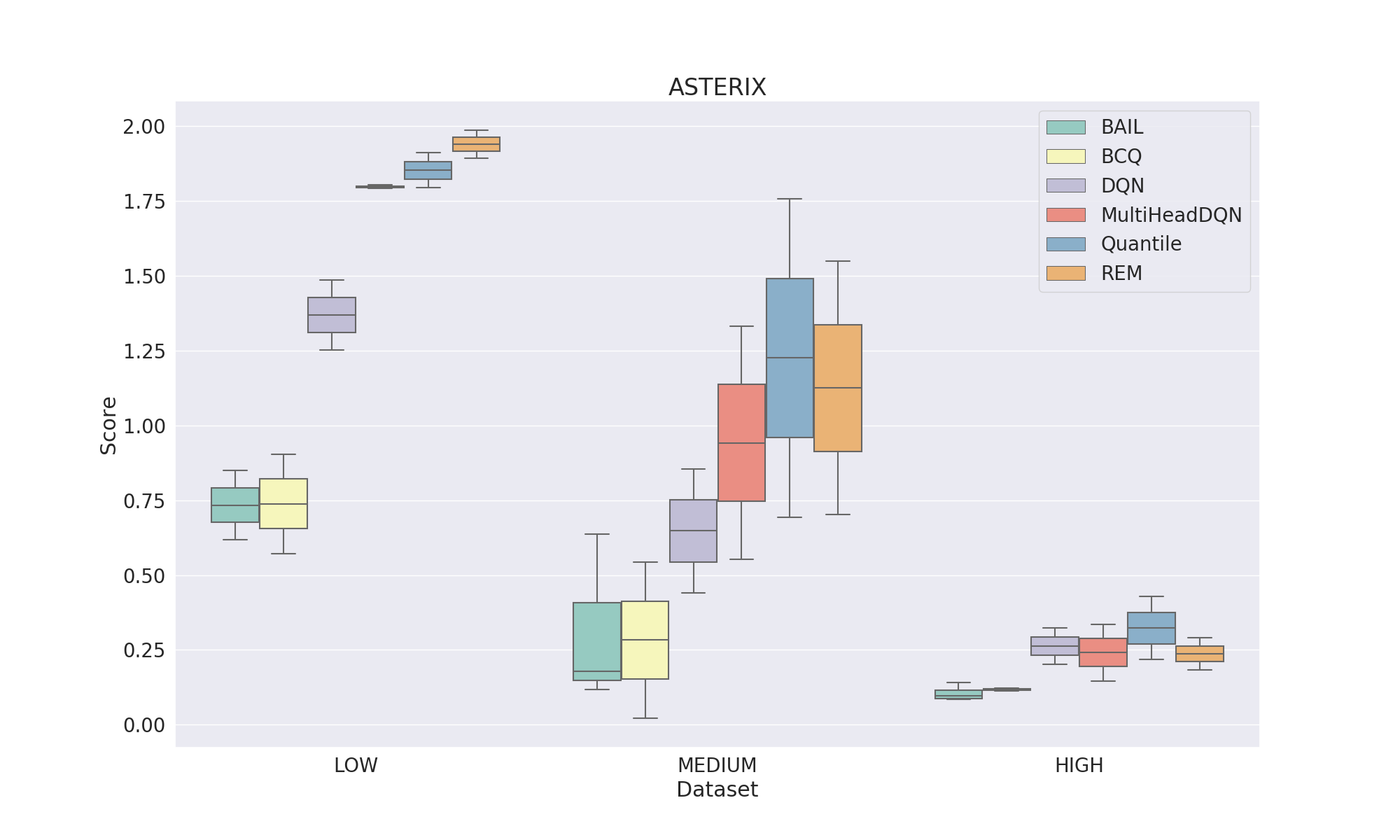}\\
				\vspace{0.01cm}
			\end{minipage}%
		}%
		\subfigure{
			\begin{minipage}[t]{0.333\linewidth}
				\centering
				\includegraphics[width=2.3in]{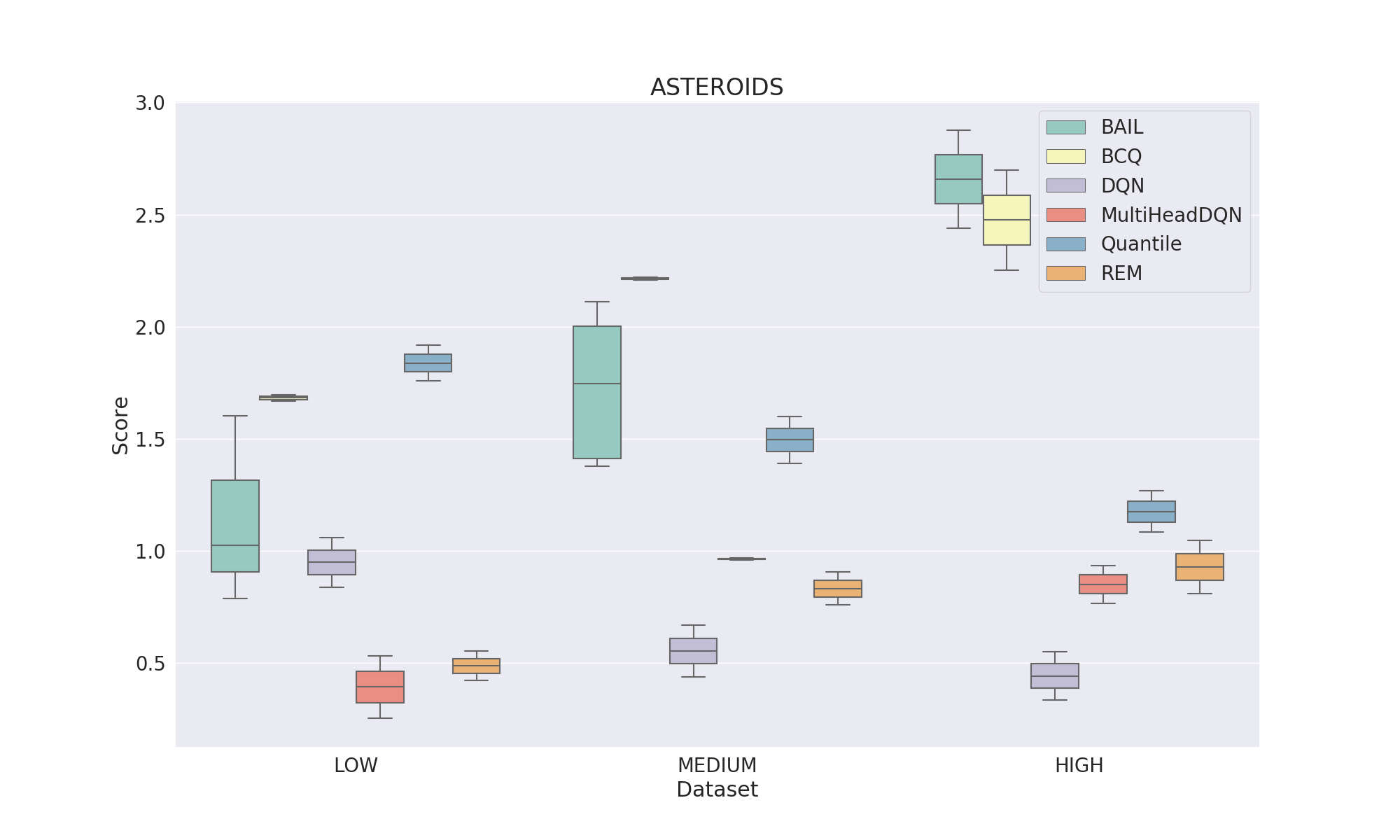}\\
				\vspace{0.01cm}
			\end{minipage}%
		}%
								
		\subfigure{
			\begin{minipage}[t]{0.333\linewidth}
				\centering
				\includegraphics[width=2.3in]{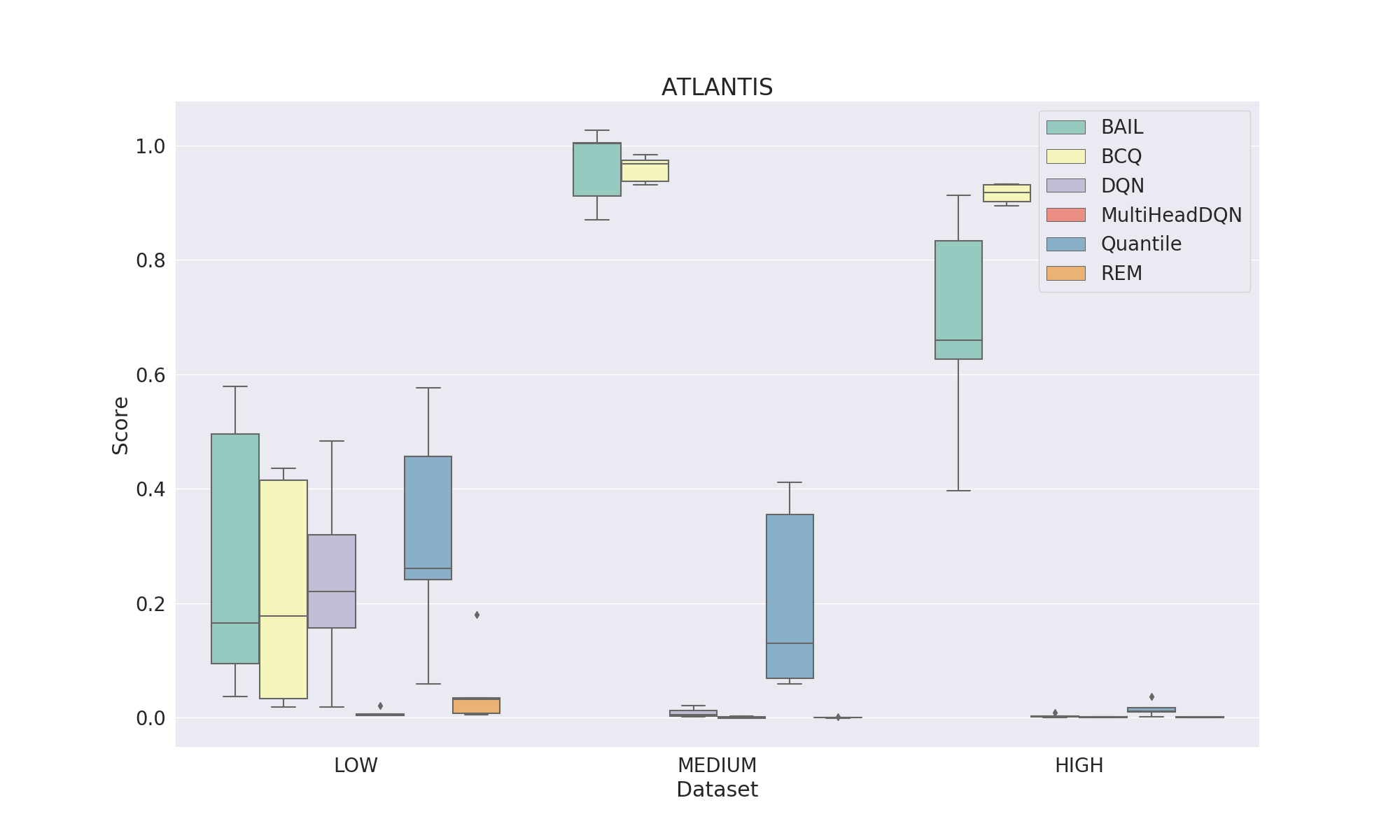}\\
				\vspace{0.01cm}
			\end{minipage}%
		}%
		\subfigure{
			\begin{minipage}[t]{0.333\linewidth}
				\centering
				\includegraphics[width=2.3in]{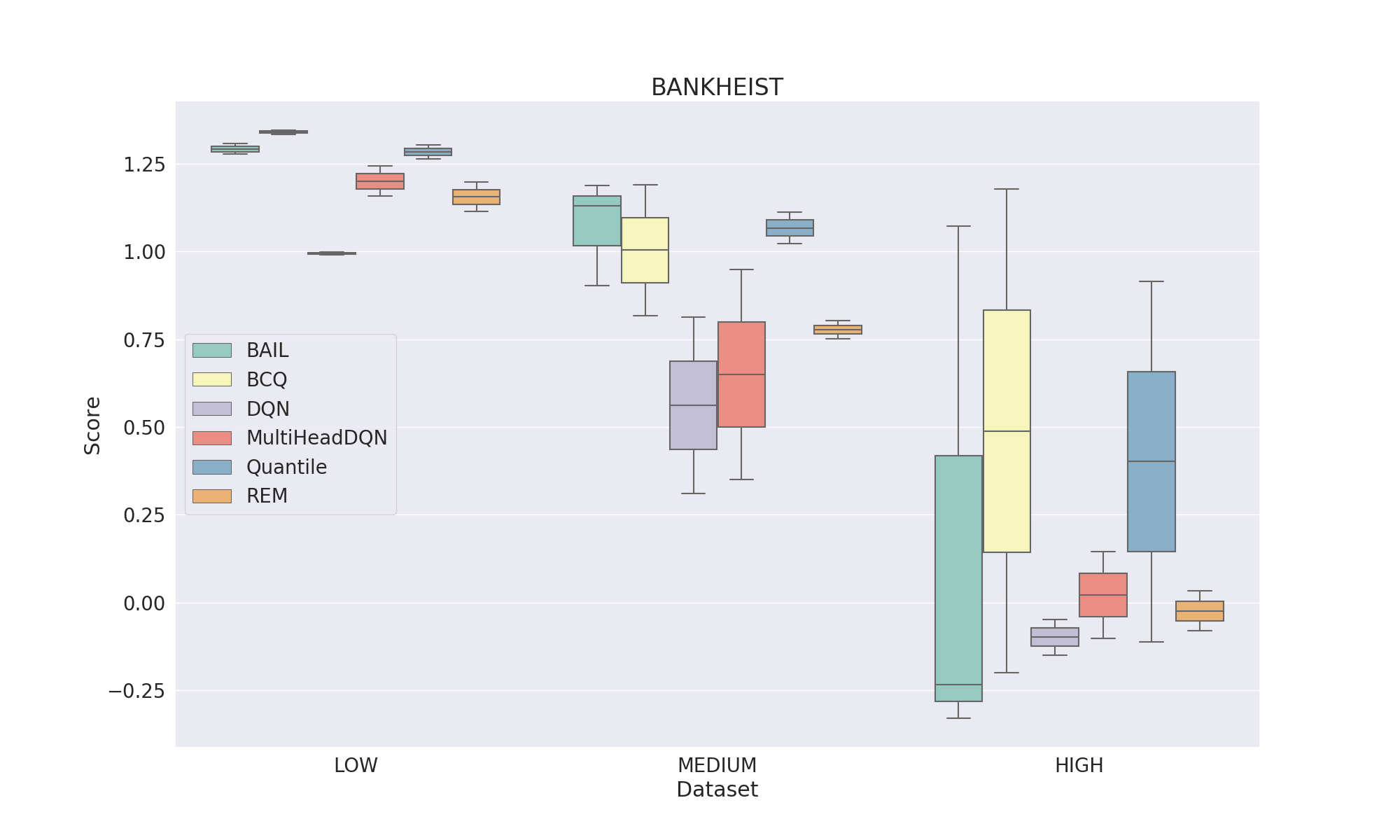}\\
				\vspace{0.01cm}
			\end{minipage}%
		}%
		\subfigure{
			\begin{minipage}[t]{0.333\linewidth}
				\centering
				\includegraphics[width=2.3in]{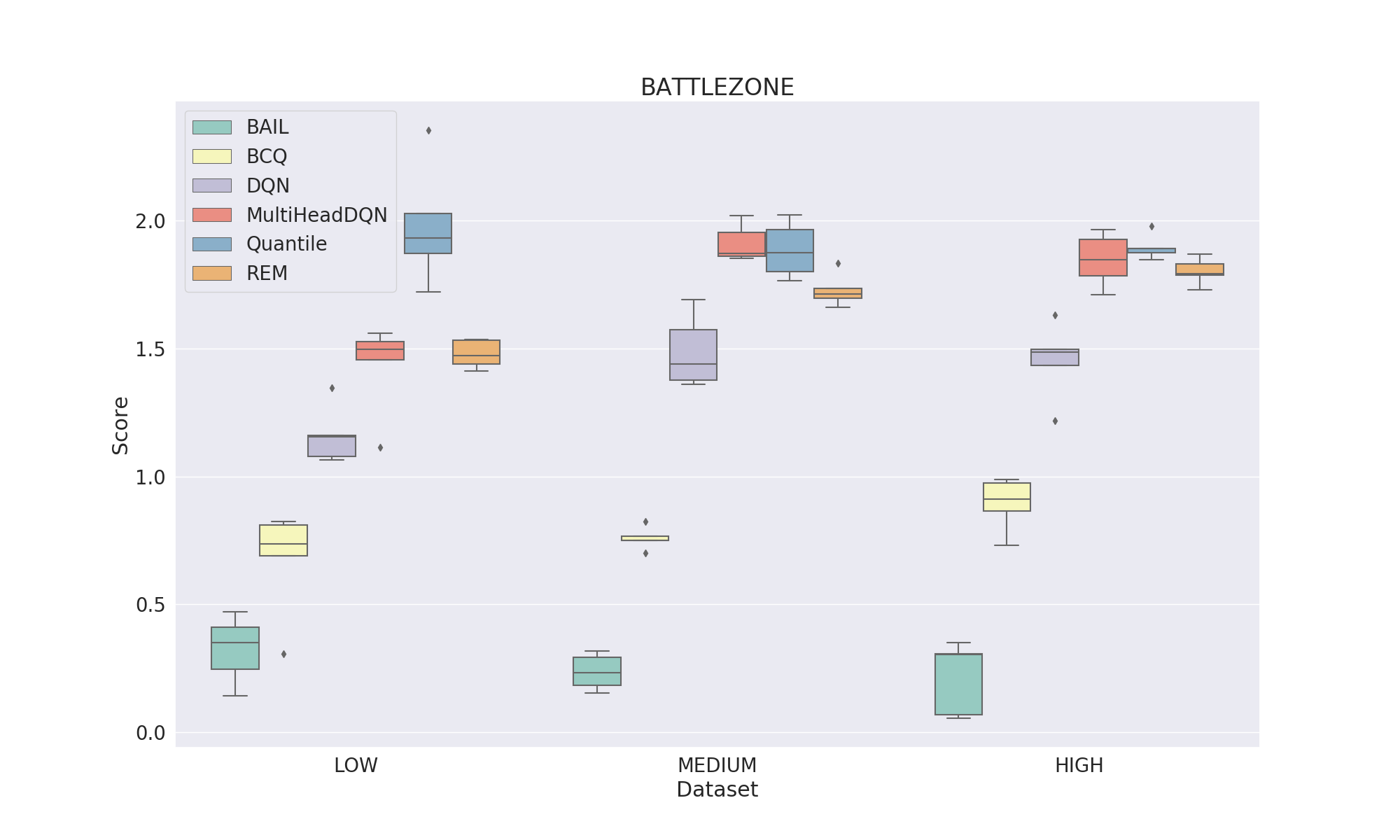}\\
				\vspace{0.01cm}
			\end{minipage}%
		}%

		\subfigure{
			\begin{minipage}[t]{0.333\linewidth}
				\centering
				\includegraphics[width=2.3in]{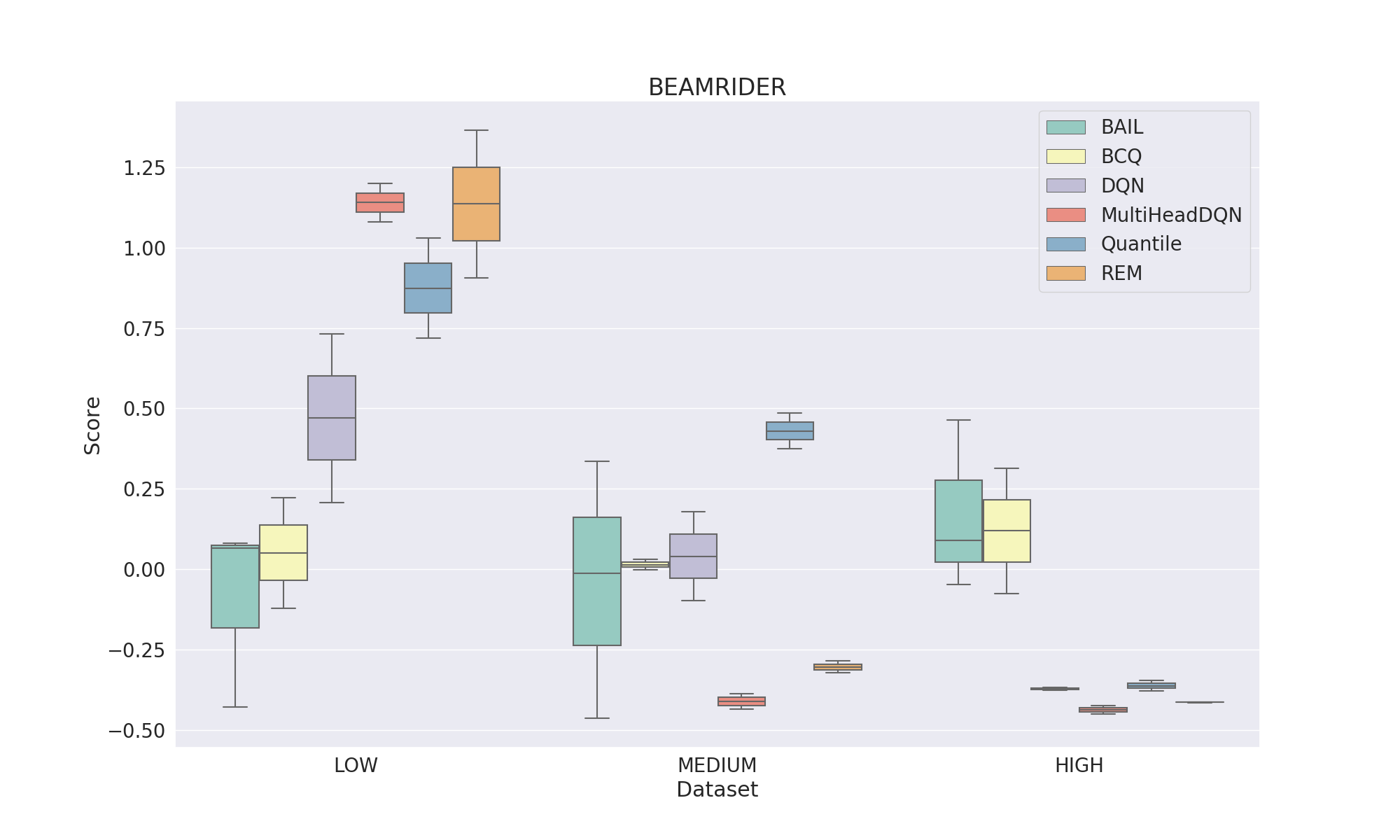}\\
				\vspace{0.01cm}
			\end{minipage}%
		}%
		\subfigure{
			\begin{minipage}[t]{0.333\linewidth}
				\centering
				\includegraphics[width=2.3in]{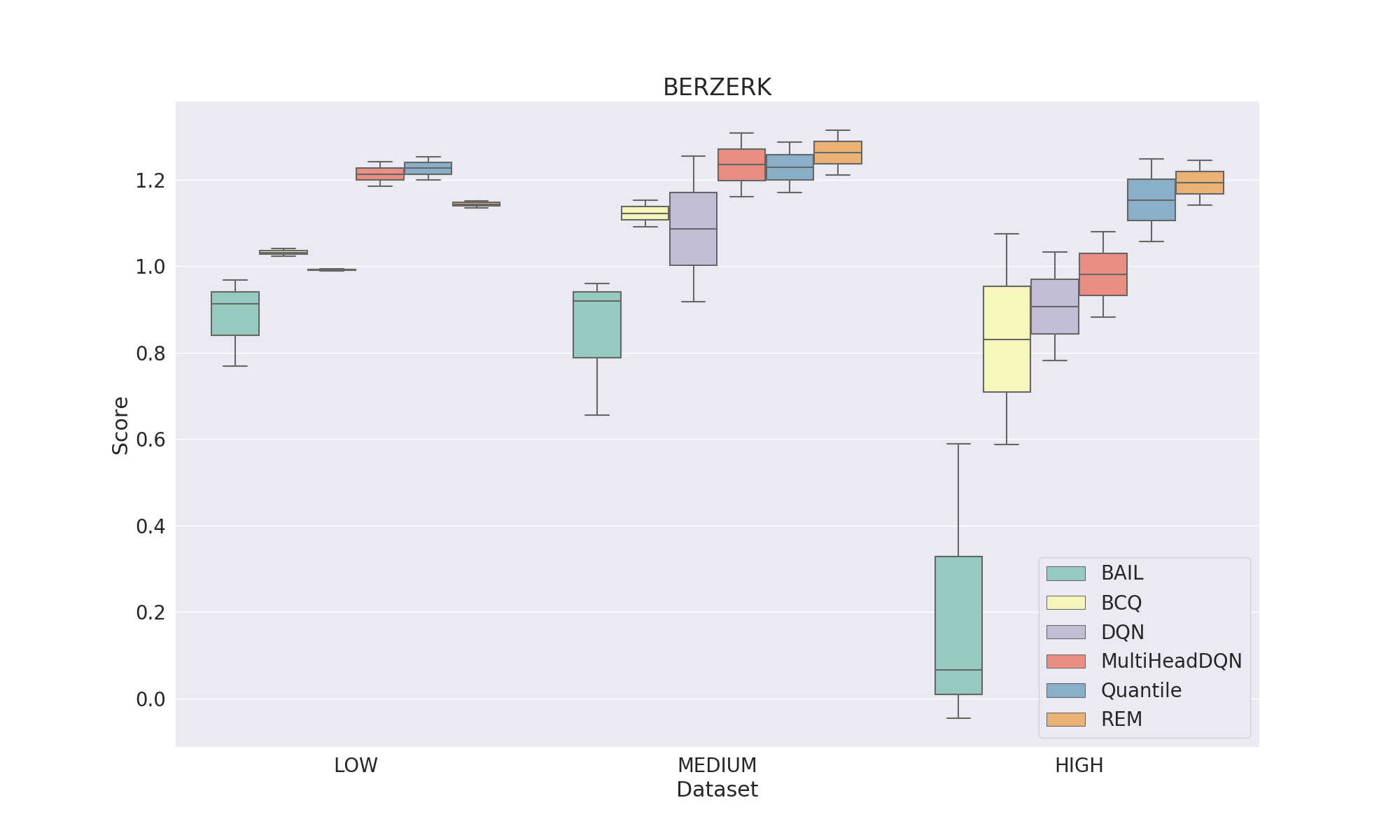}\\
				\vspace{0.01cm}
			\end{minipage}%
		}%
		\subfigure{
			\begin{minipage}[t]{0.333\linewidth}
				\centering
				\includegraphics[width=2.3in]{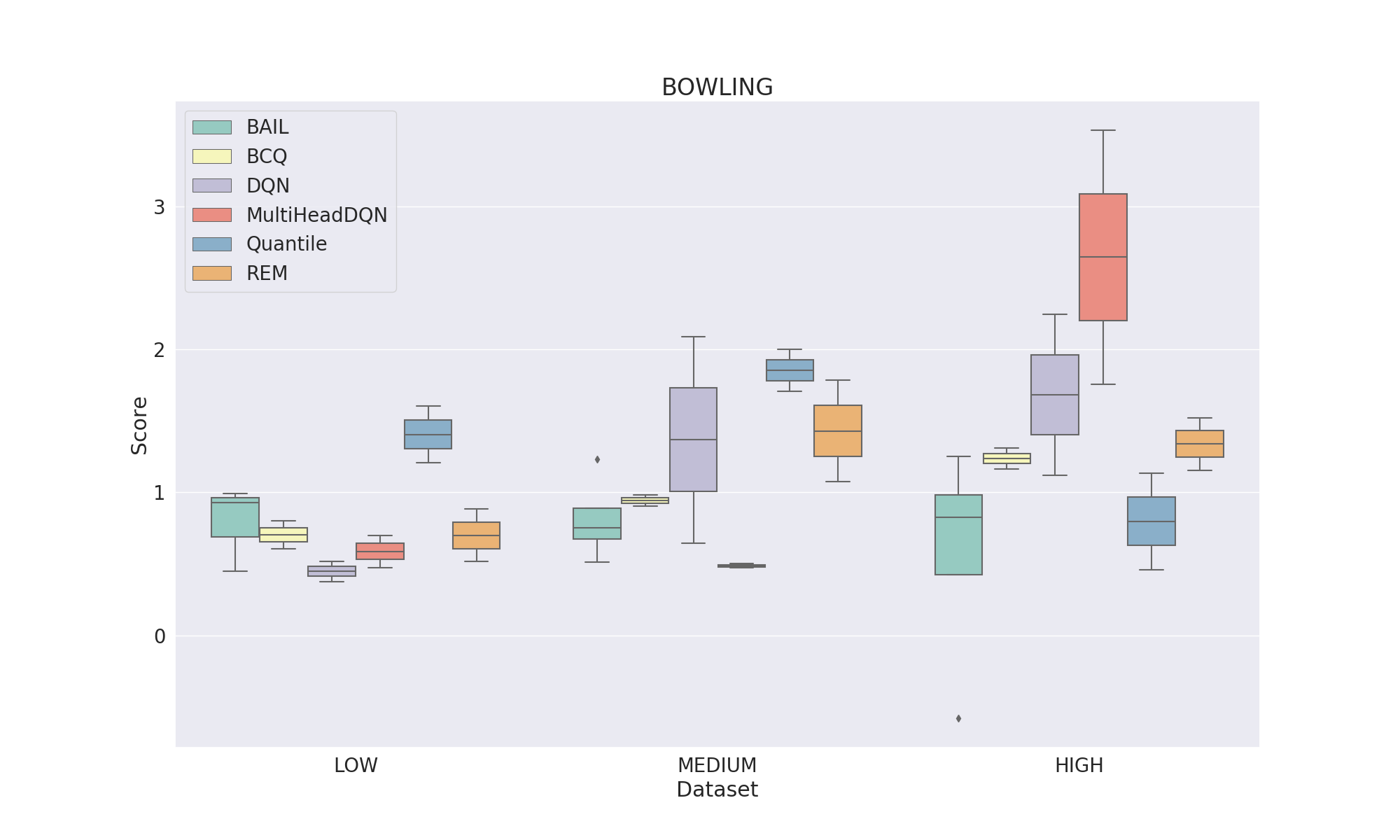}\\
				\vspace{0.01cm}
			\end{minipage}%
		}%

		\subfigure{
			\begin{minipage}[t]{0.333\linewidth}
				\centering
				\includegraphics[width=2.3in]{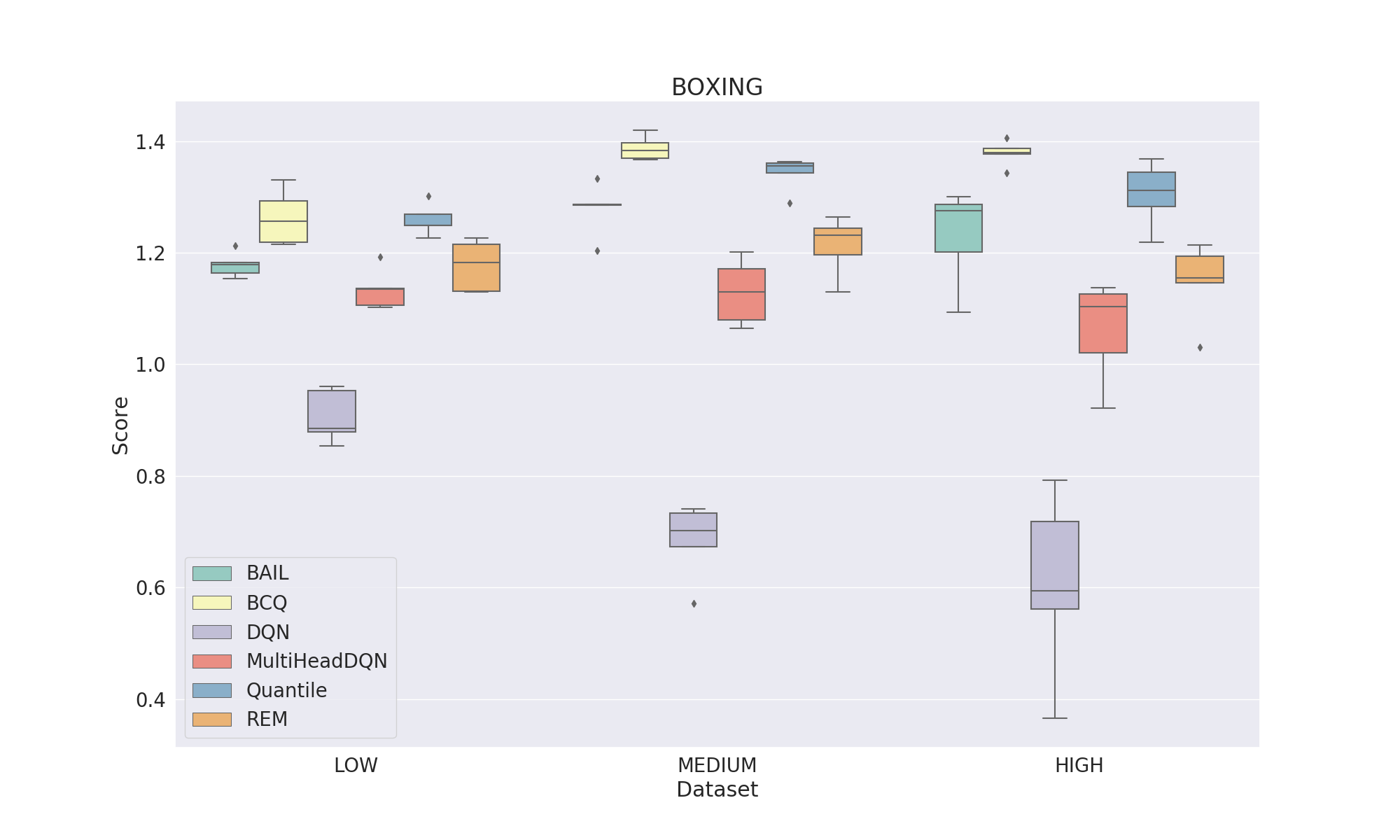}\\
				\vspace{0.01cm}
			\end{minipage}%
		}%
		\subfigure{
			\begin{minipage}[t]{0.333\linewidth}
				\centering
				\includegraphics[width=2.3in]{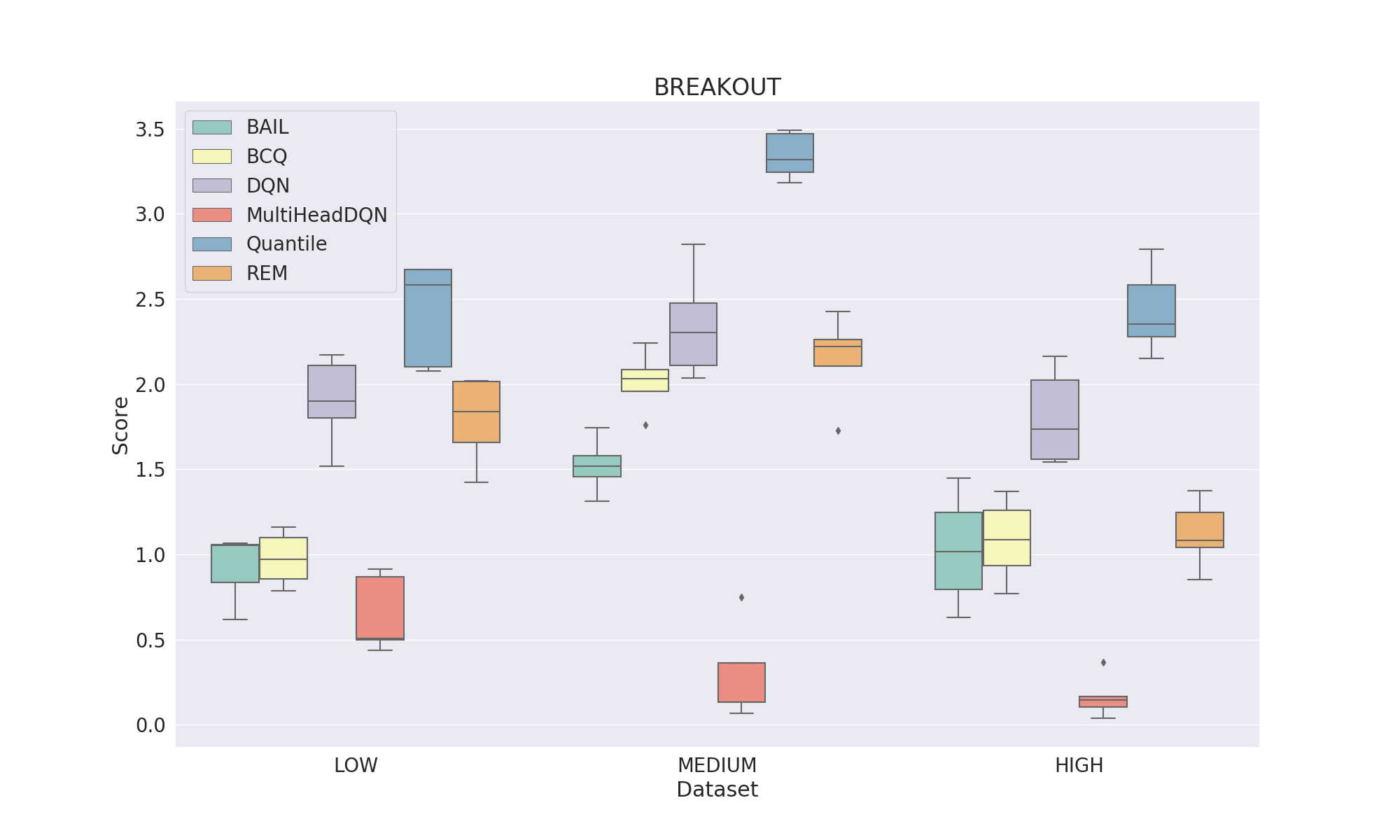}\\
				\vspace{0.01cm}
			\end{minipage}%
		}%
		\subfigure{
			\begin{minipage}[t]{0.333\linewidth}
				\centering
				\includegraphics[width=2.3in]{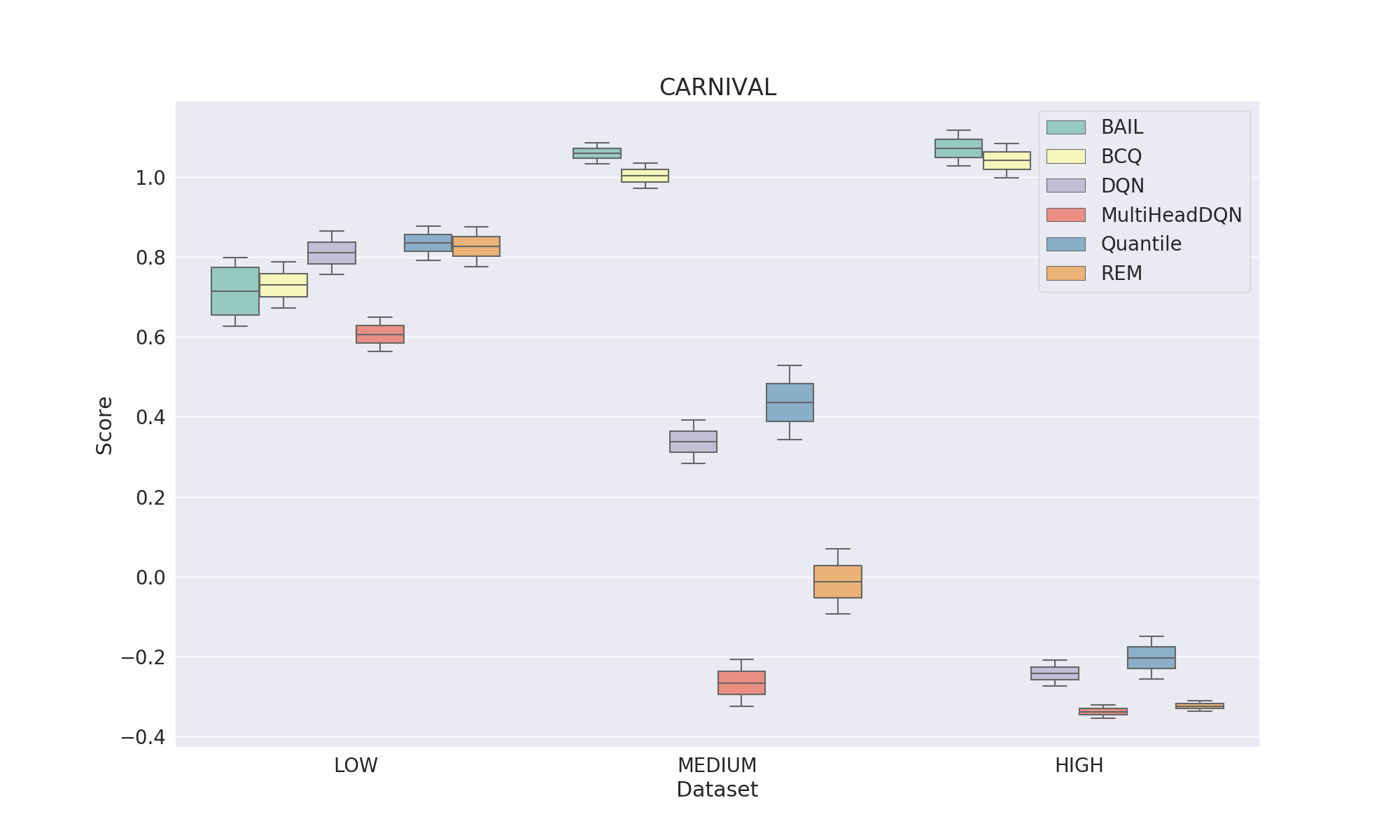}\\
				\vspace{0.01cm}
			\end{minipage}%
		}%

		\subfigure{
			\begin{minipage}[t]{0.333\linewidth}
				\centering
				\includegraphics[width=2.3in]{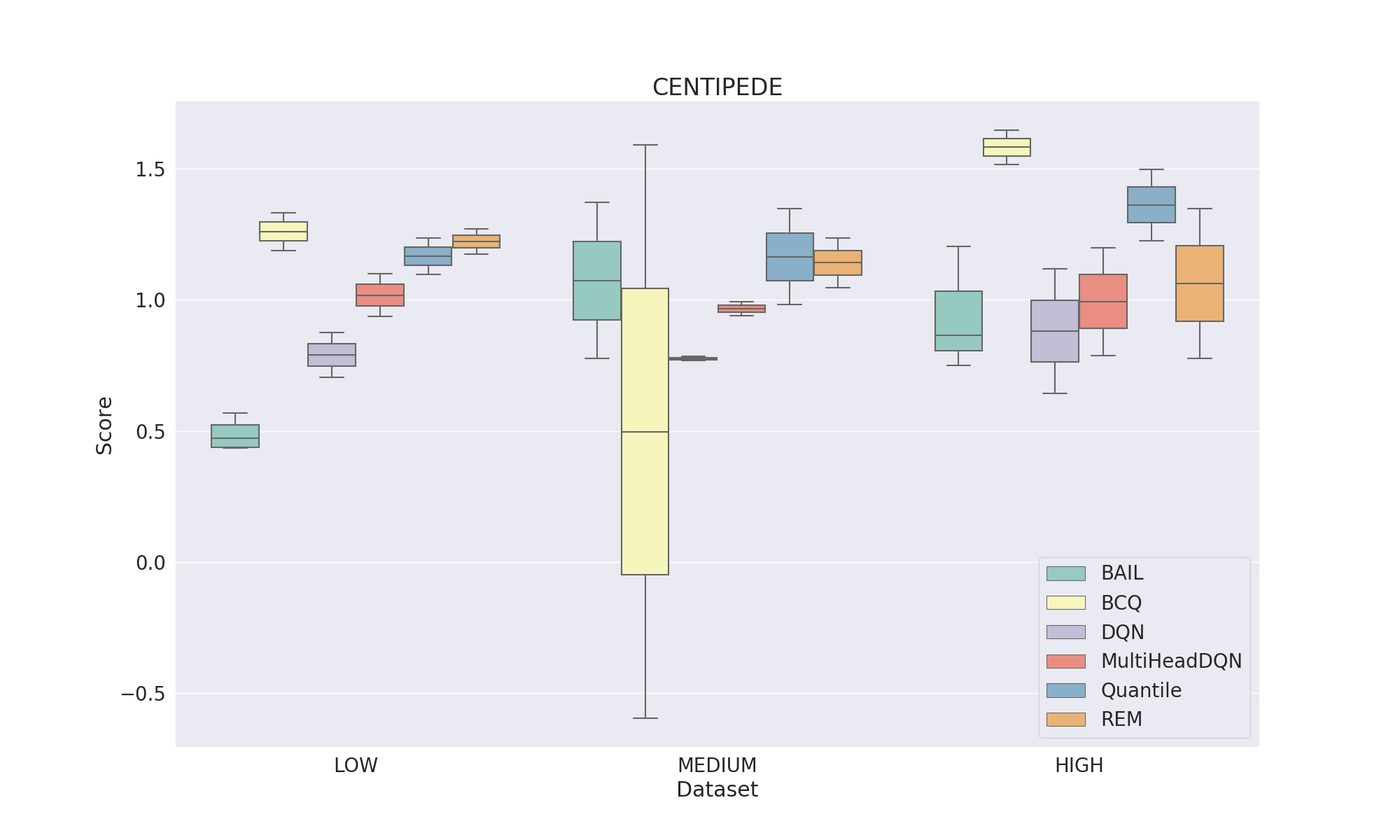}\\
				\vspace{0.01cm}
			\end{minipage}%
		}%
		\subfigure{
			\begin{minipage}[t]{0.333\linewidth}
				\centering
				\includegraphics[width=2.3in]{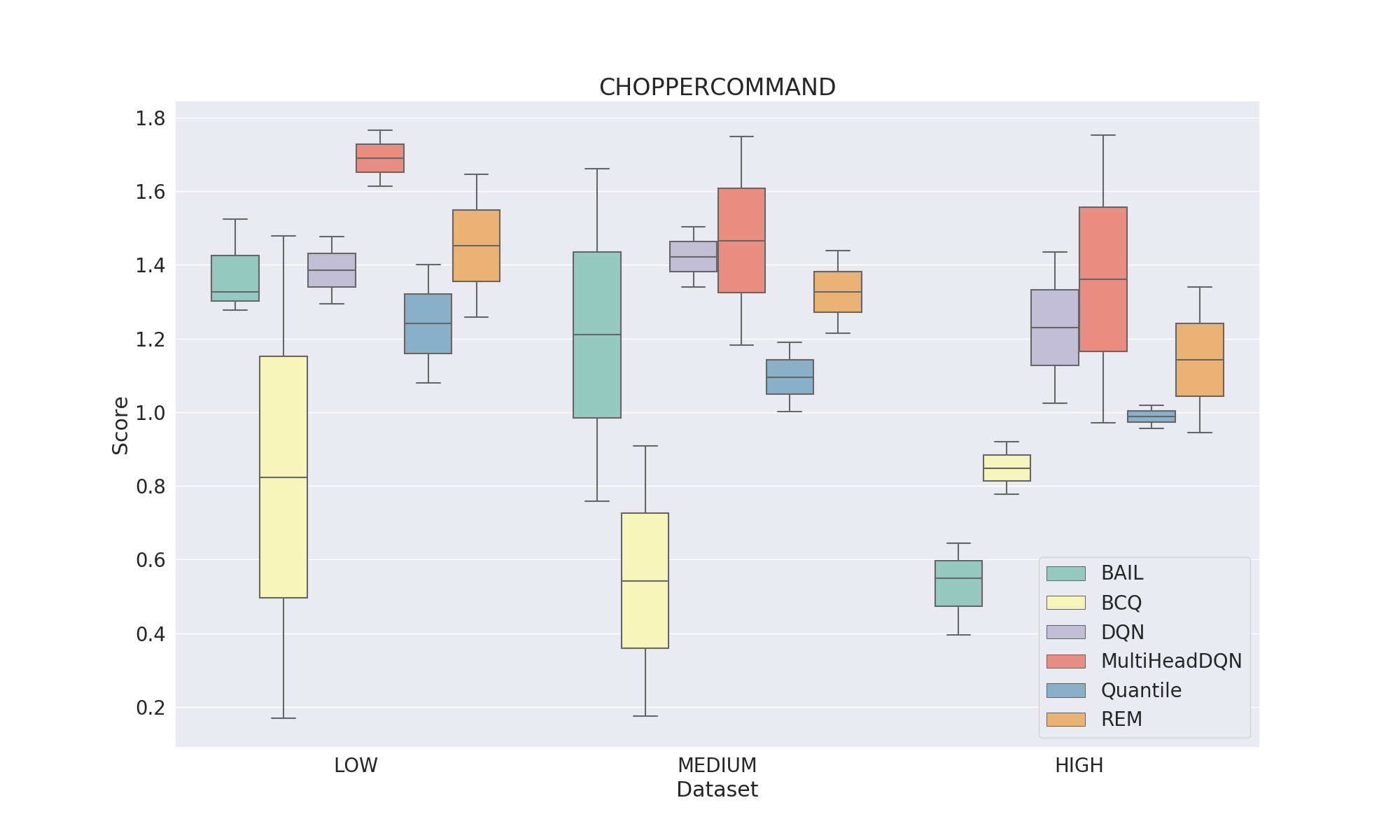}\\
				\vspace{0.01cm}
			\end{minipage}%
		}%
		\subfigure{
			\begin{minipage}[t]{0.333\linewidth}
				\centering
				\includegraphics[width=2.3in]{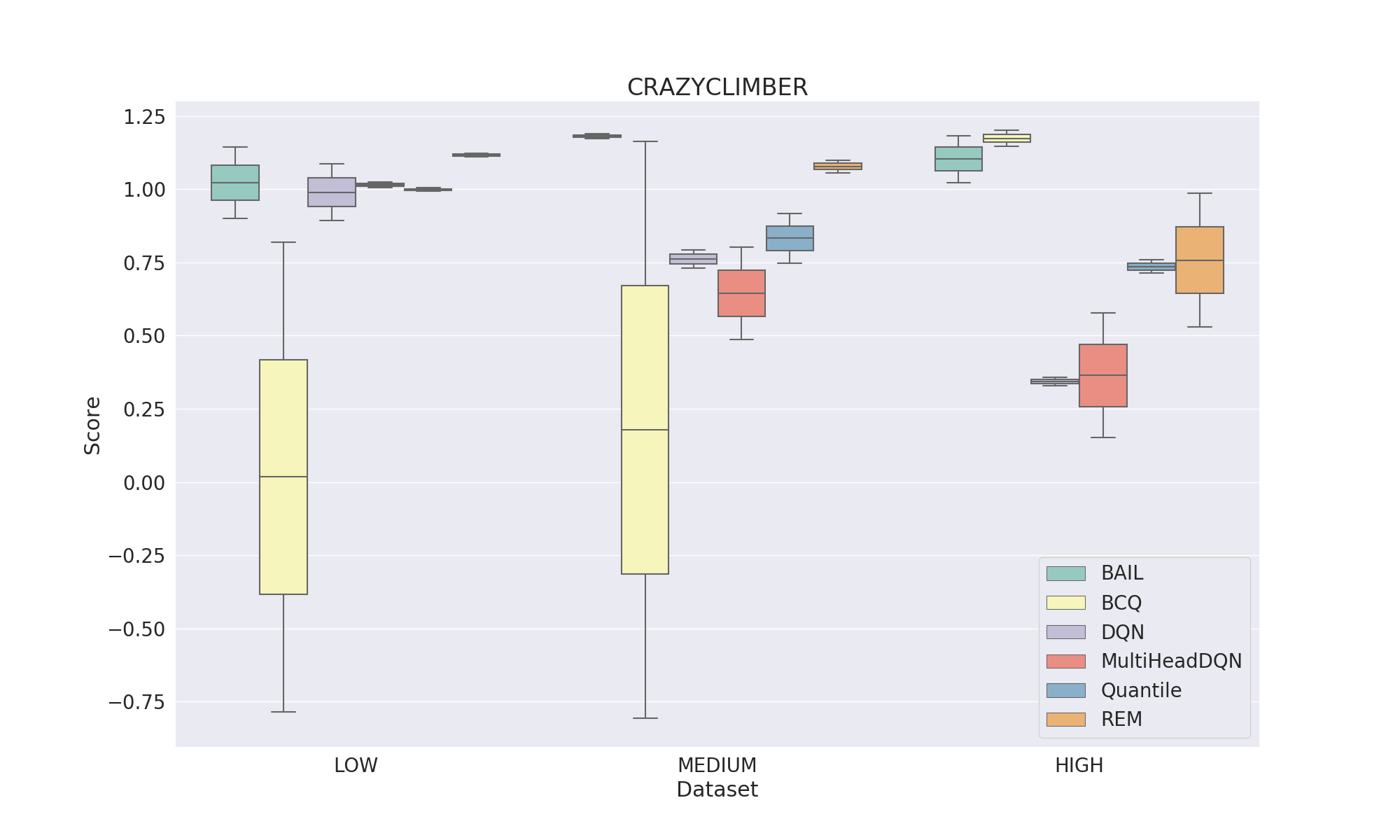}\\
				\vspace{0.01cm}
			\end{minipage}%
		}%

		\centering
		\caption{\textbf{Comparison between baselines on different datasets from Game Alien to Game CrazyClimber}}
		\label{fig: Comparison between baselines on different datasets from Game Alien to Game CrazyClimber}
								
	\end{figure*}

	Please refer Fig. \ref{fig: Comparison between baselines on different datasets from Game Alien to Game CrazyClimber} - Fig. \ref{fig: Comparison between TR-BCQ and the best baselines on different datasets from Game UpNDown to Game Zaxxon}
	
	\begin{figure*}[!htb]
		\centering

		\subfigure{
			\begin{minipage}[t]{0.333\linewidth}
				\centering
				\includegraphics[width=2.3in]{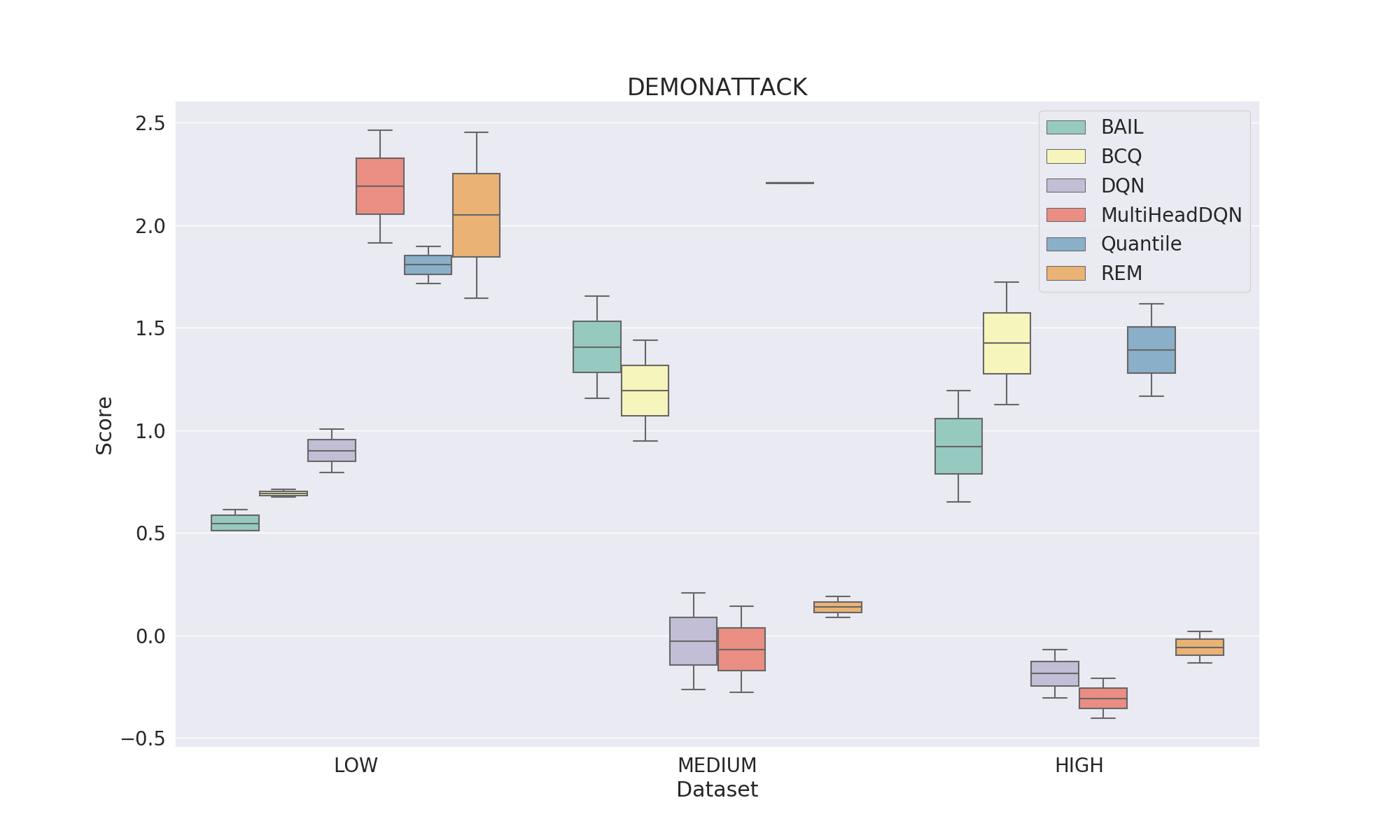}\\
				\vspace{0.01cm}
			\end{minipage}%
		}%
		\subfigure{
			\begin{minipage}[t]{0.333\linewidth}
				\centering
				\includegraphics[width=2.3in]{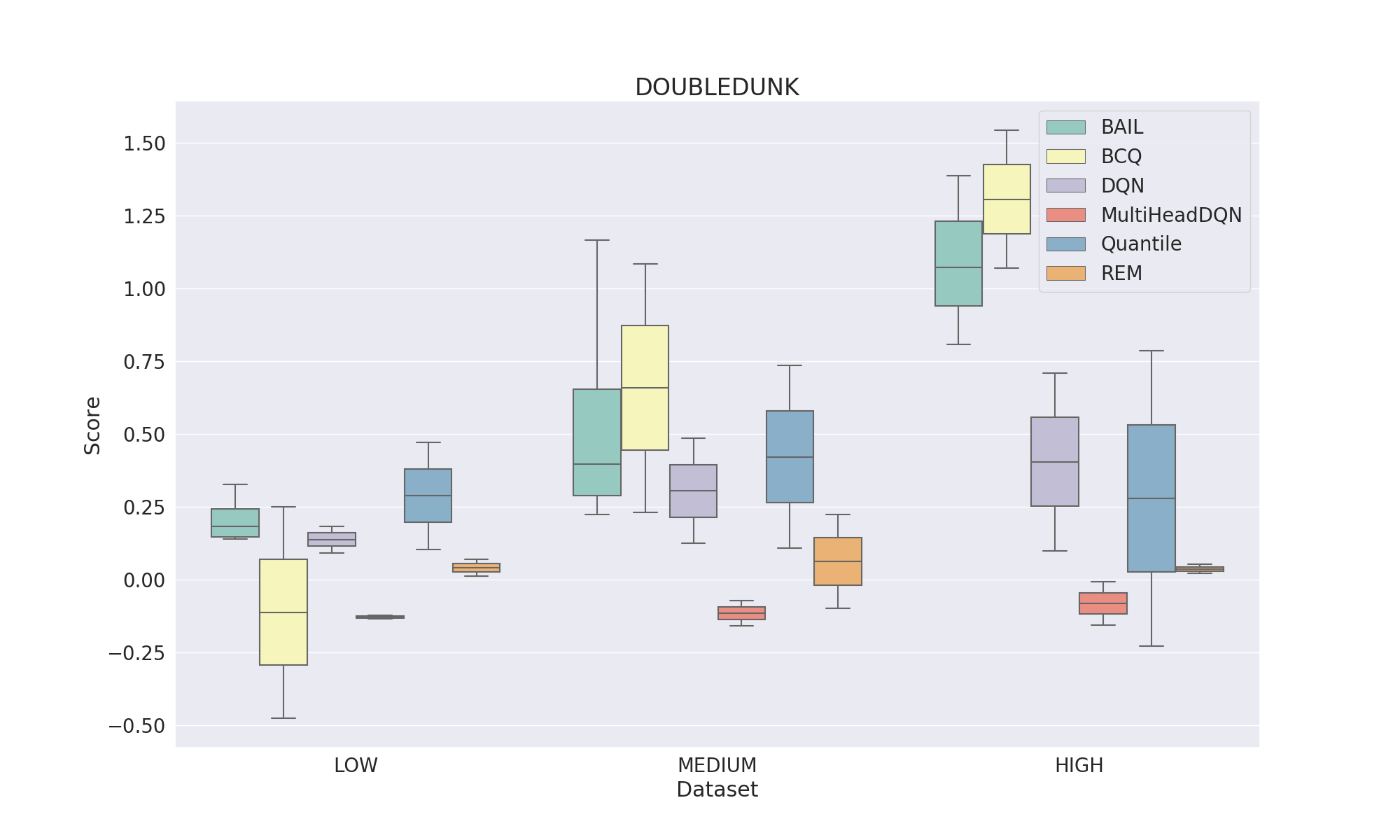}\\
				\vspace{0.01cm}
			\end{minipage}%
		}%
		\subfigure{
			\begin{minipage}[t]{0.333\linewidth}
				\centering
				\includegraphics[width=2.3in]{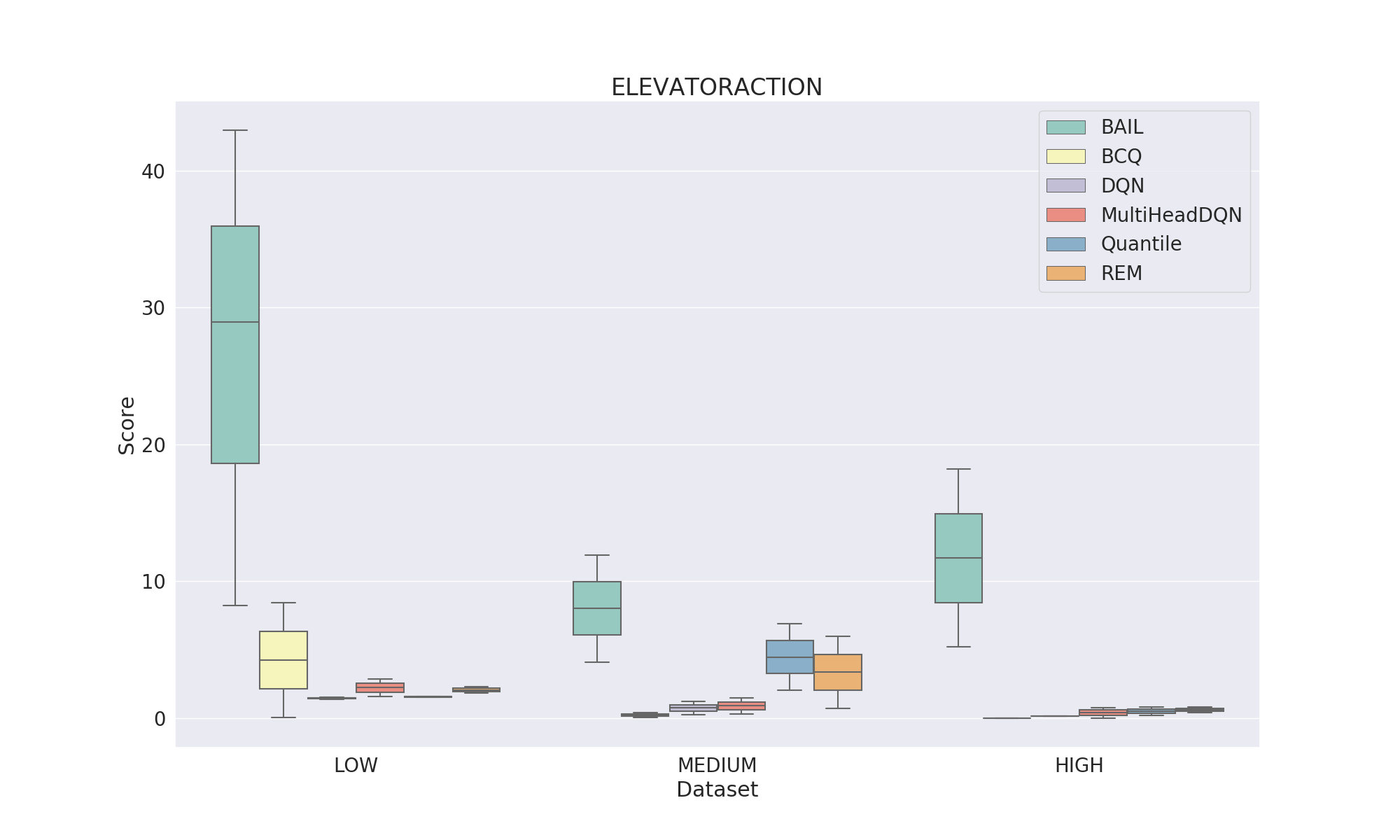}\\
				\vspace{0.01cm}
			\end{minipage}%
		}%

		\subfigure{
			\begin{minipage}[t]{0.333\linewidth}
				\centering
				\includegraphics[width=2.3in]{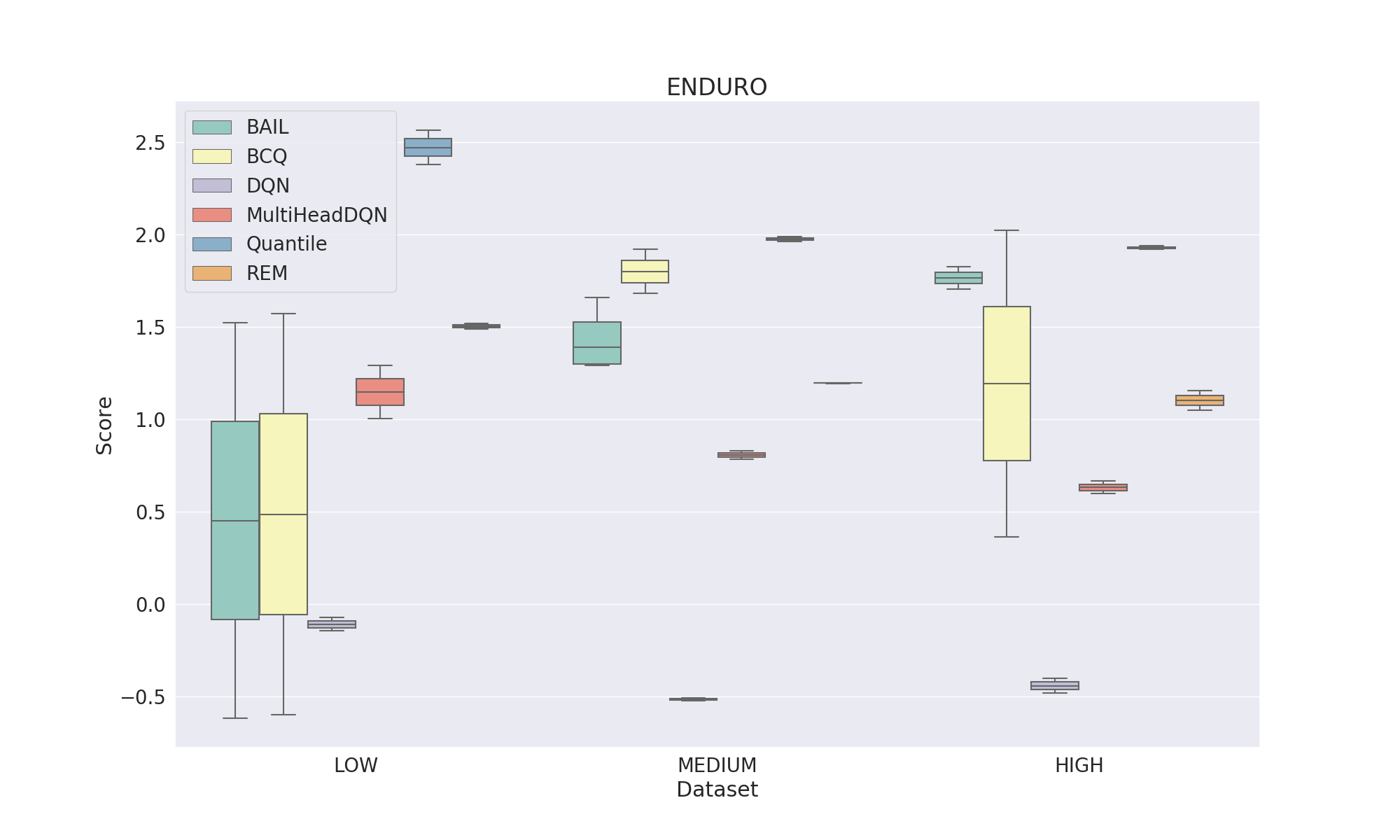}\\
				\vspace{0.01cm}
			\end{minipage}%
		}%
		\subfigure{
			\begin{minipage}[t]{0.333\linewidth}
				\centering
				\includegraphics[width=2.3in]{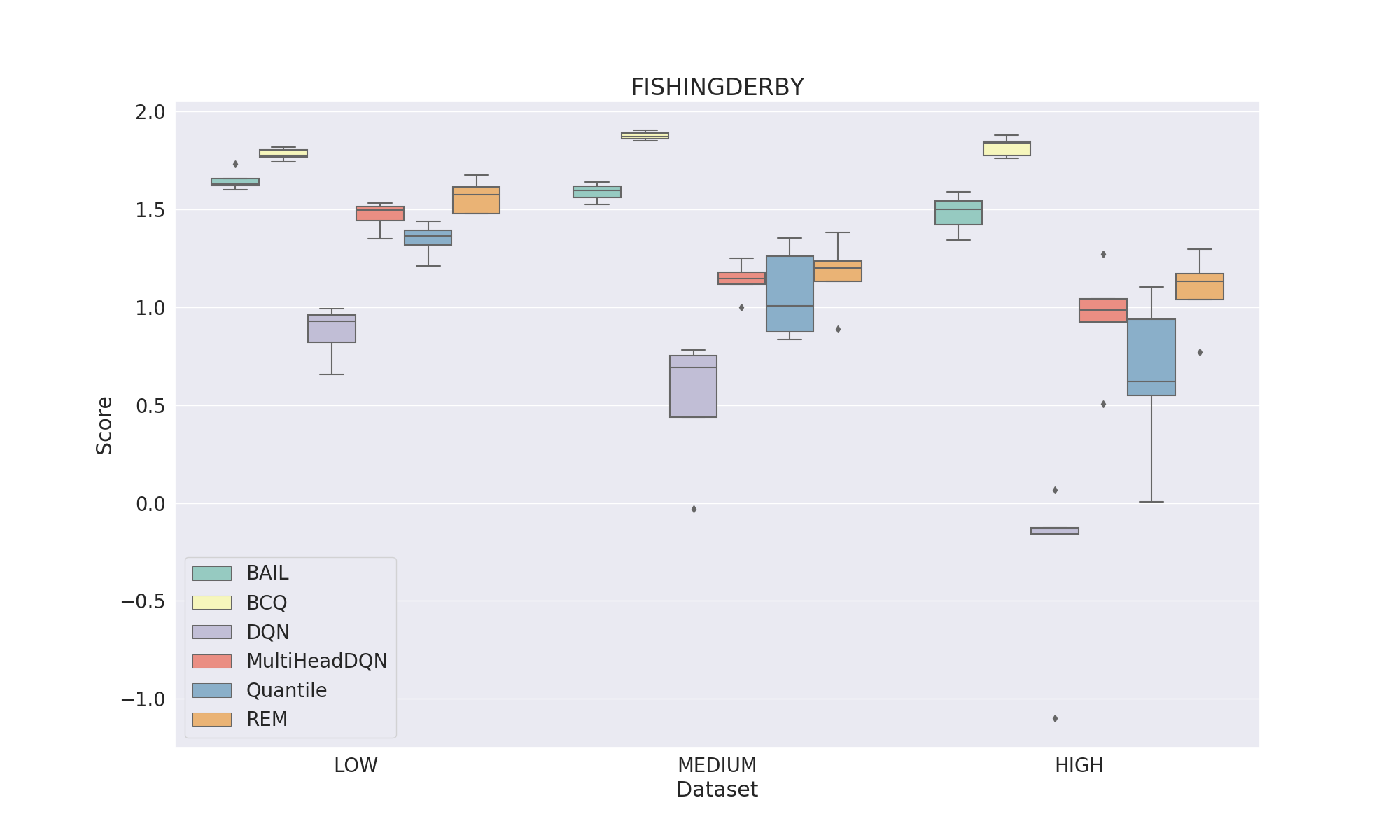}\\
				\vspace{0.01cm}
			\end{minipage}%
		}%
		\subfigure{
			\begin{minipage}[t]{0.333\linewidth}
				\centering
				\includegraphics[width=2.3in]{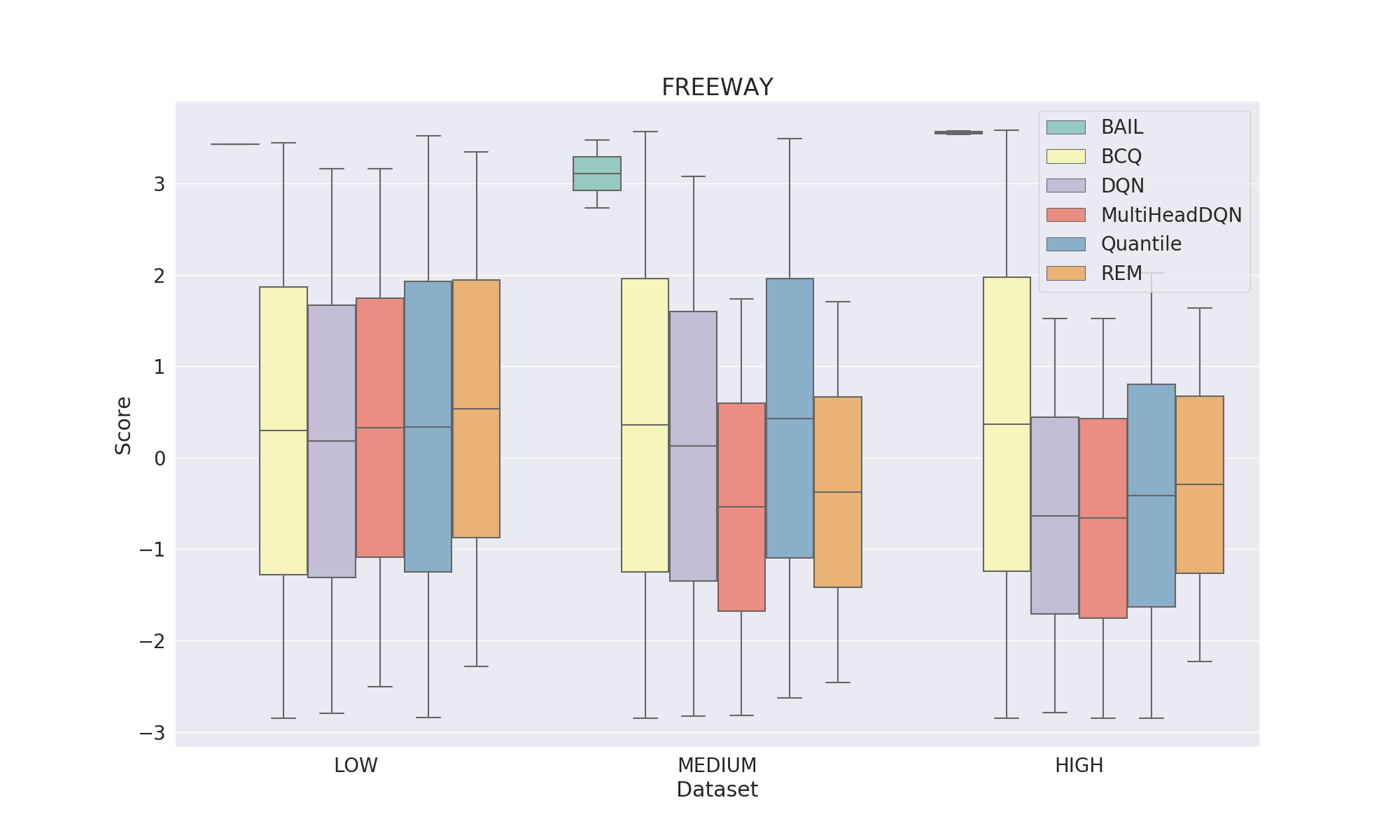}\\
				\vspace{0.01cm}
			\end{minipage}%
		}%

		\subfigure{
			\begin{minipage}[t]{0.333\linewidth}
				\centering
				\includegraphics[width=2.3in]{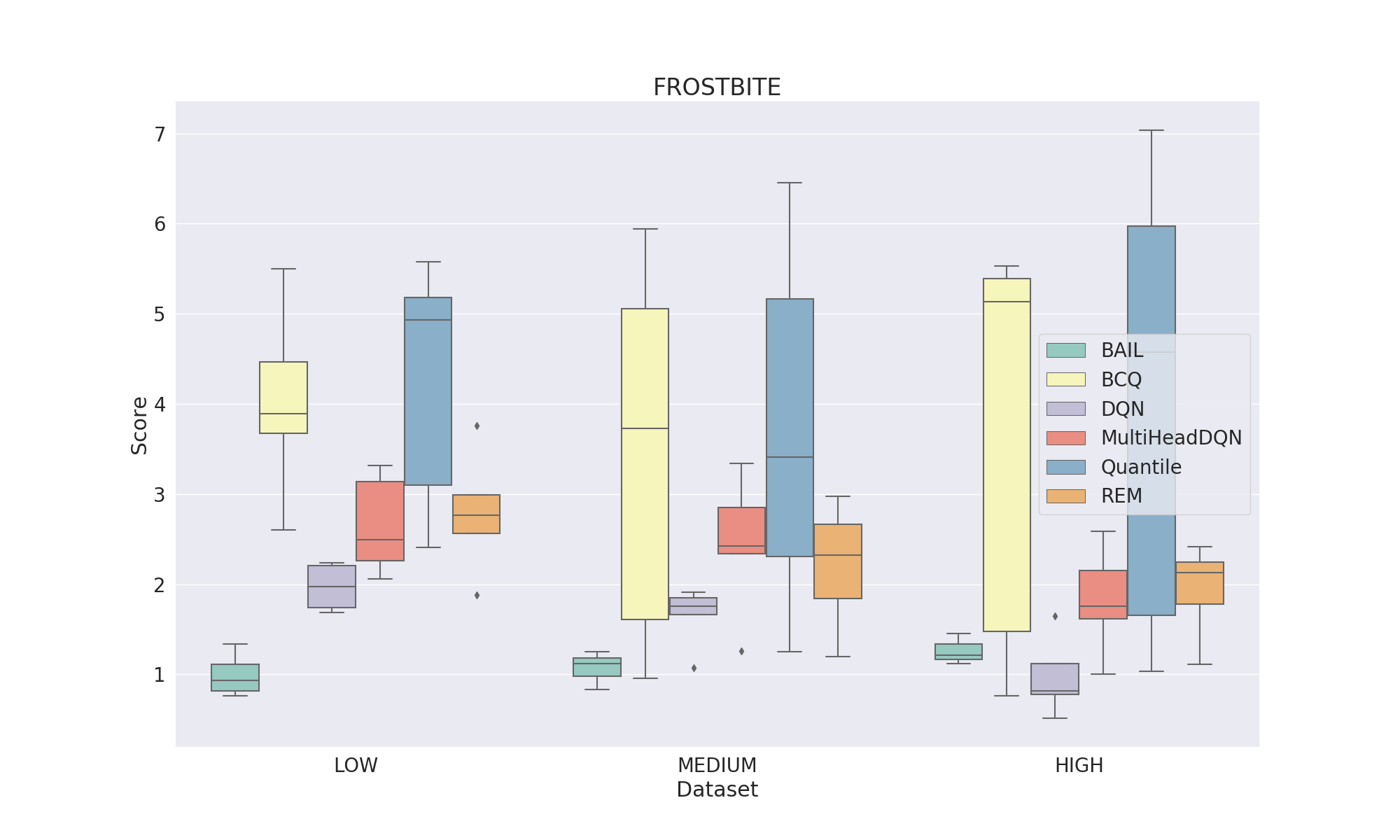}\\
				\vspace{0.01cm}
			\end{minipage}%
		}%
		\subfigure{
			\begin{minipage}[t]{0.333\linewidth}
				\centering
				\includegraphics[width=2.3in]{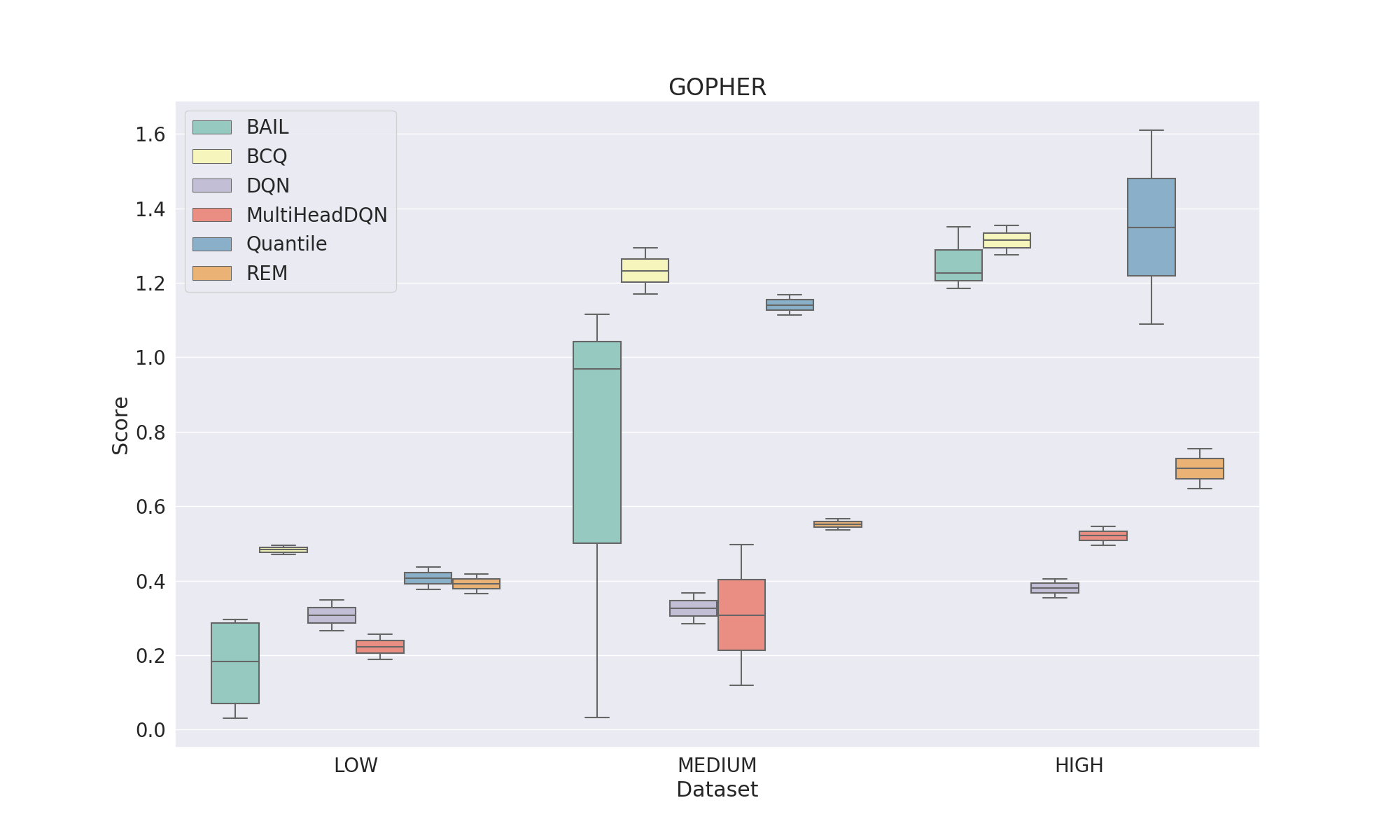}\\
				\vspace{0.01cm}
			\end{minipage}%
		}%
		\subfigure{
			\begin{minipage}[t]{0.333\linewidth}
				\centering
				\includegraphics[width=2.3in]{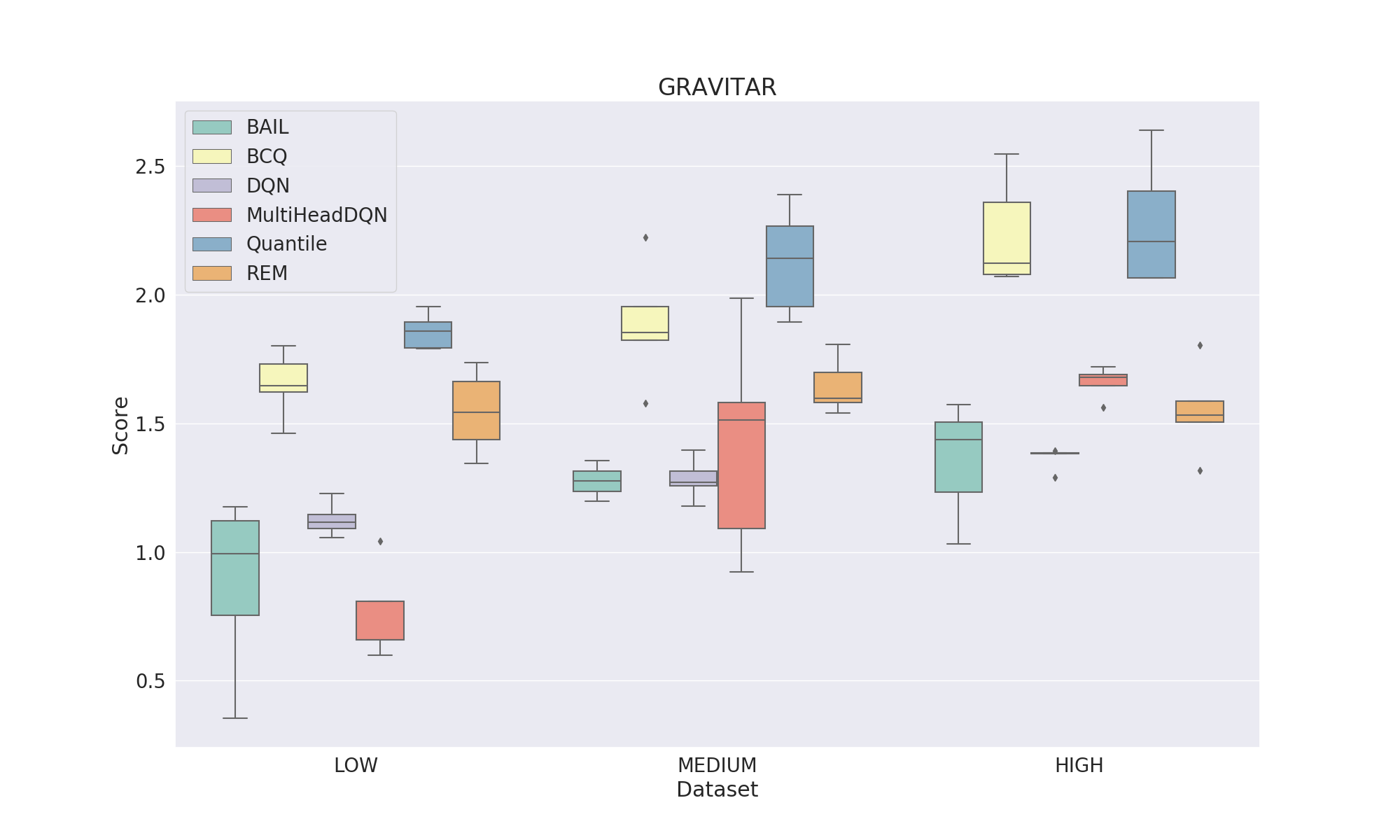}\\
				\vspace{0.01cm}
			\end{minipage}%
		}%

		\subfigure{
			\begin{minipage}[t]{0.333\linewidth}
				\centering
				\includegraphics[width=2.3in]{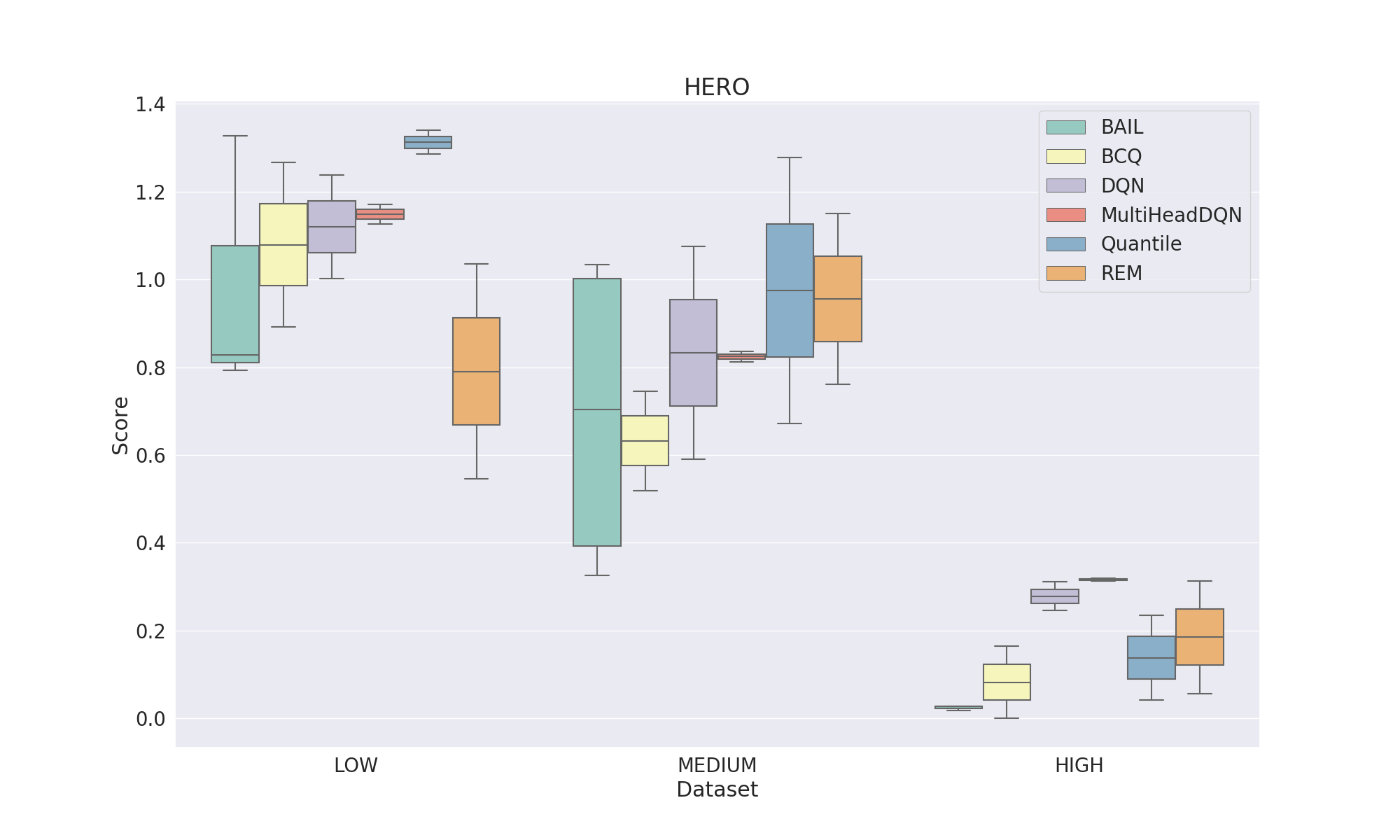}\\
				\vspace{0.01cm}
			\end{minipage}%
		}%
		\subfigure{
			\begin{minipage}[t]{0.333\linewidth}
				\centering
				\includegraphics[width=2.3in]{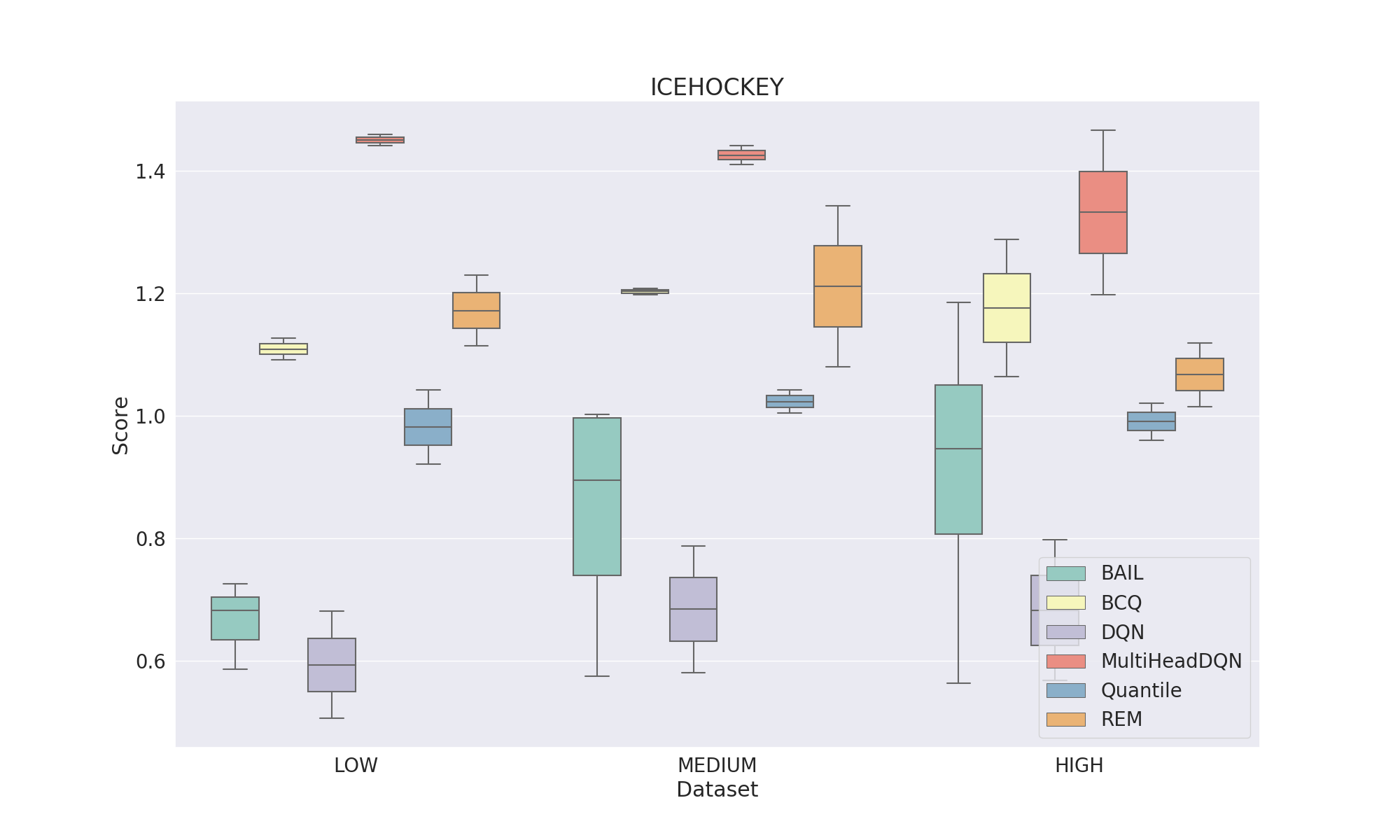}\\
				\vspace{0.01cm}
			\end{minipage}%
		}%
		\subfigure{
			\begin{minipage}[t]{0.333\linewidth}
				\centering
				\includegraphics[width=2.3in]{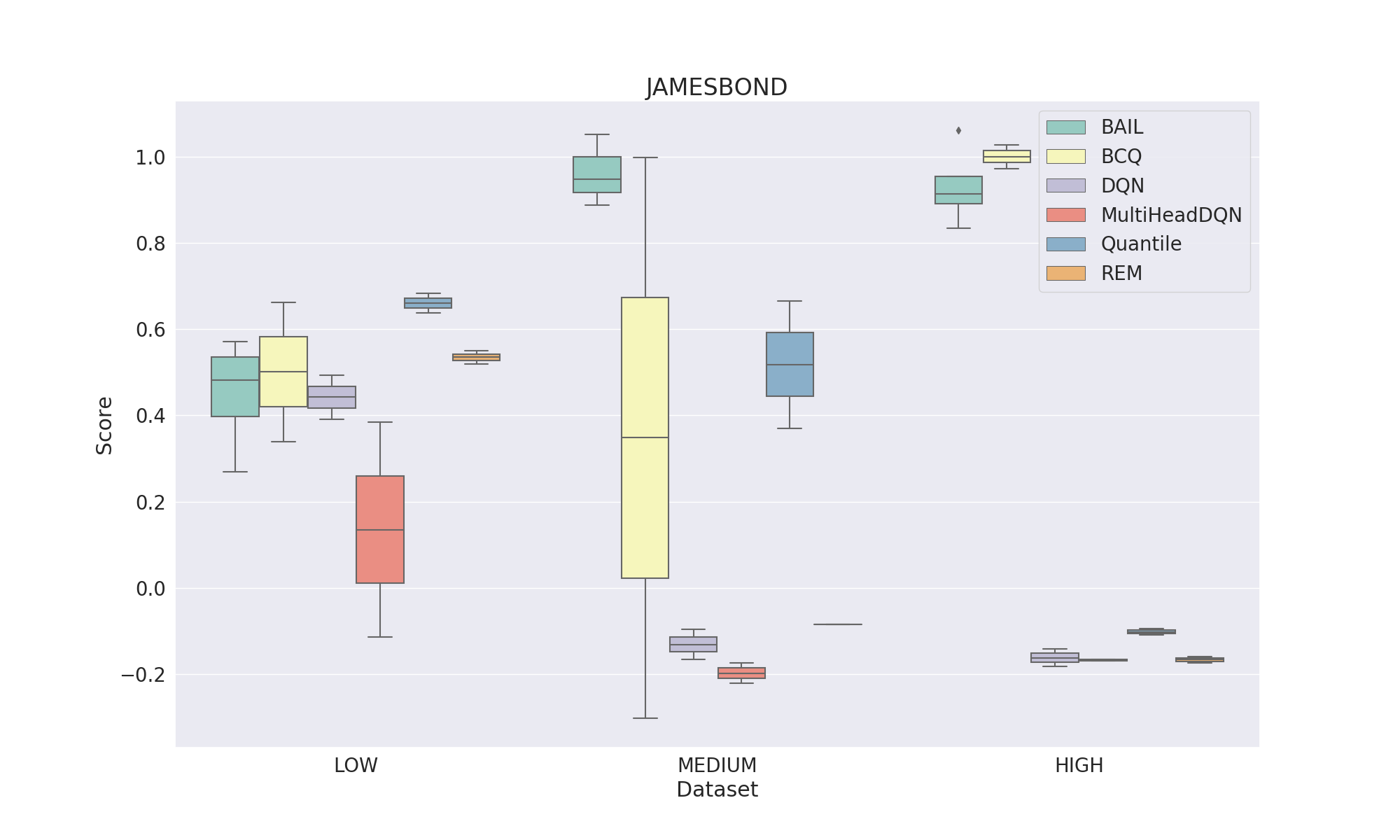}\\
				\vspace{0.01cm}
			\end{minipage}%
		}%

		\subfigure{
			\begin{minipage}[t]{0.333\linewidth}
				\centering
				\includegraphics[width=2.3in]{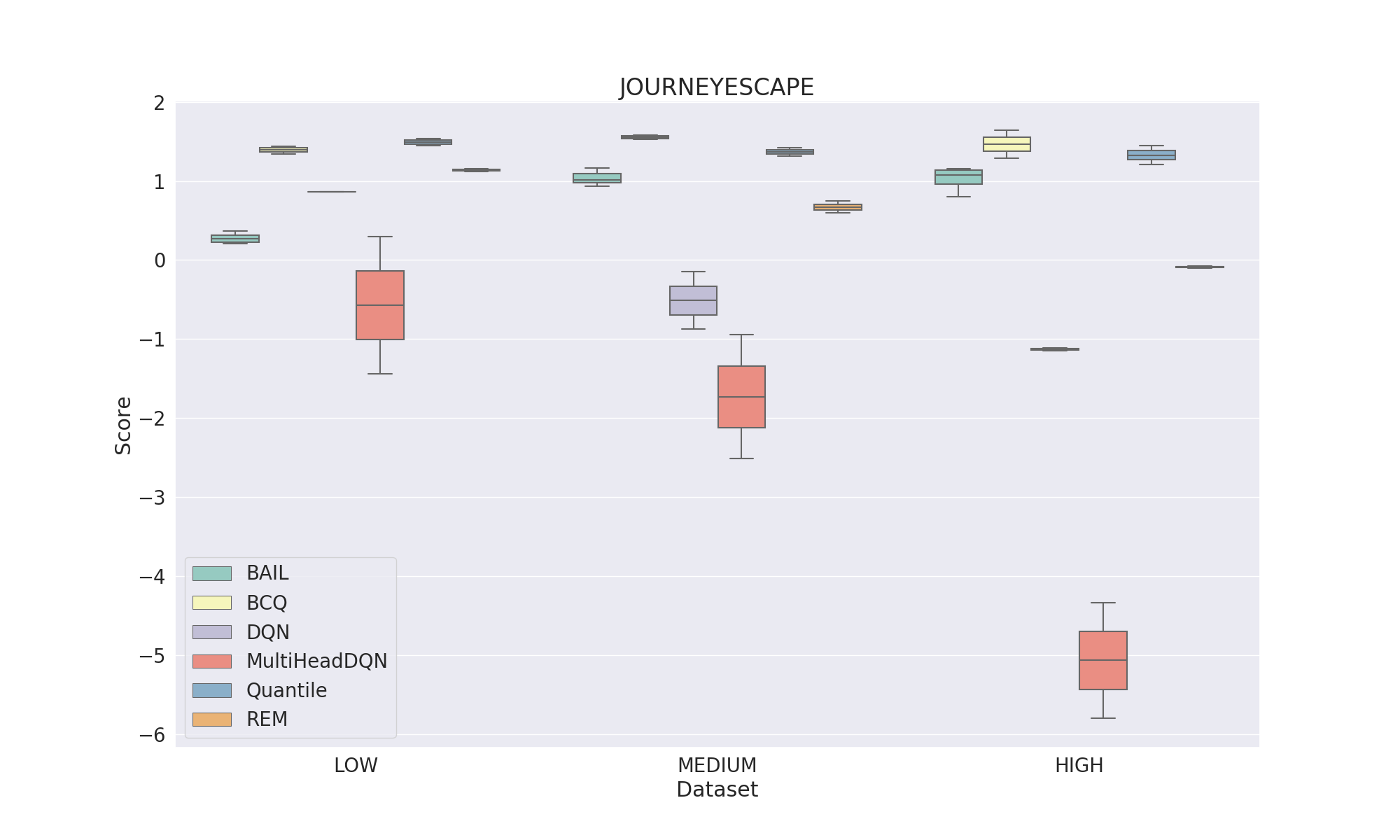}\\
				\vspace{0.01cm}
			\end{minipage}%
		}%
		\subfigure{
			\begin{minipage}[t]{0.333\linewidth}
				\centering
				\includegraphics[width=2.3in]{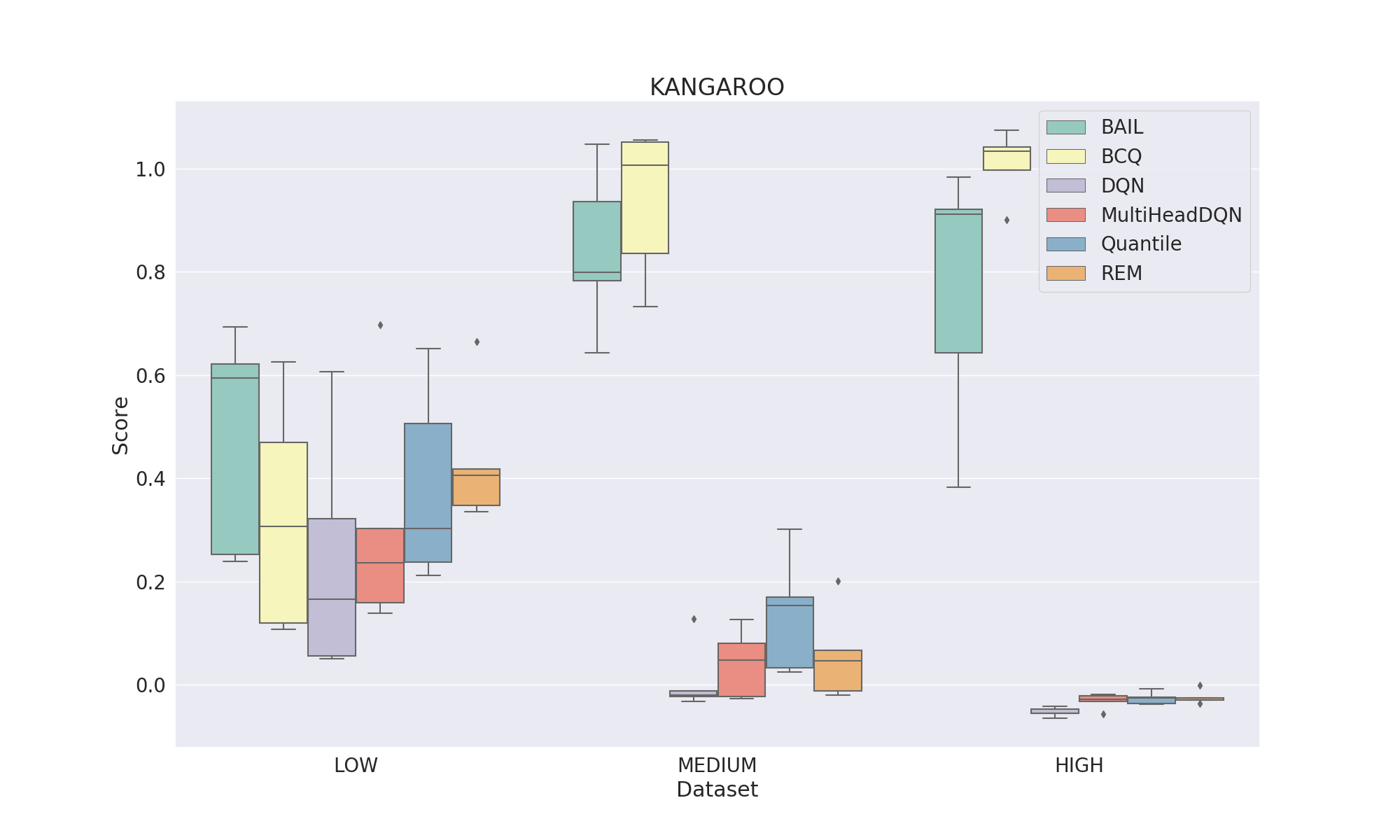}\\
				\vspace{0.01cm}
			\end{minipage}%
		}%
		\subfigure{
			\begin{minipage}[t]{0.333\linewidth}
				\centering
				\includegraphics[width=2.3in]{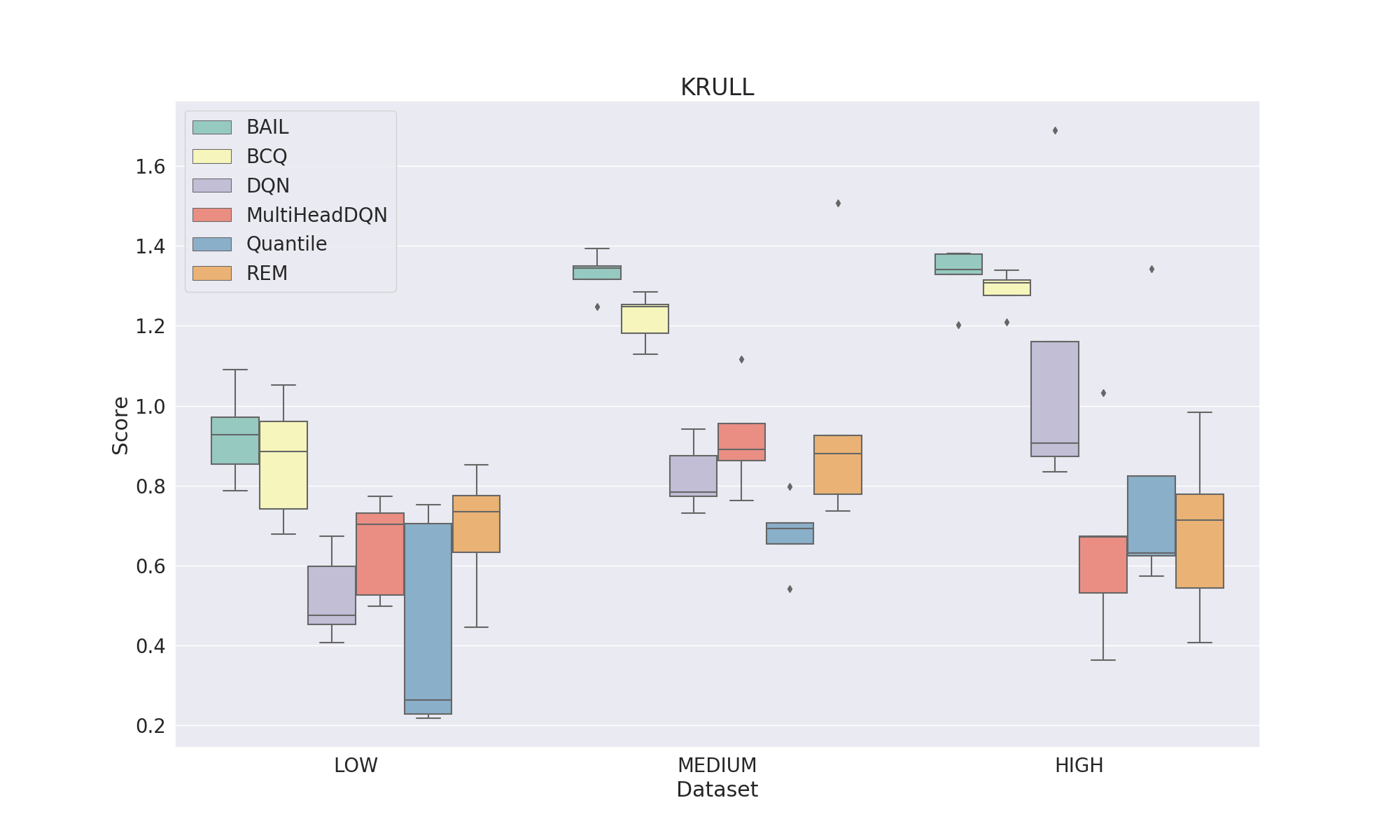}\\
				\vspace{0.01cm}
			\end{minipage}%
		}%

		\subfigure{
			\begin{minipage}[t]{0.333\linewidth}
				\centering
				\includegraphics[width=2.3in]{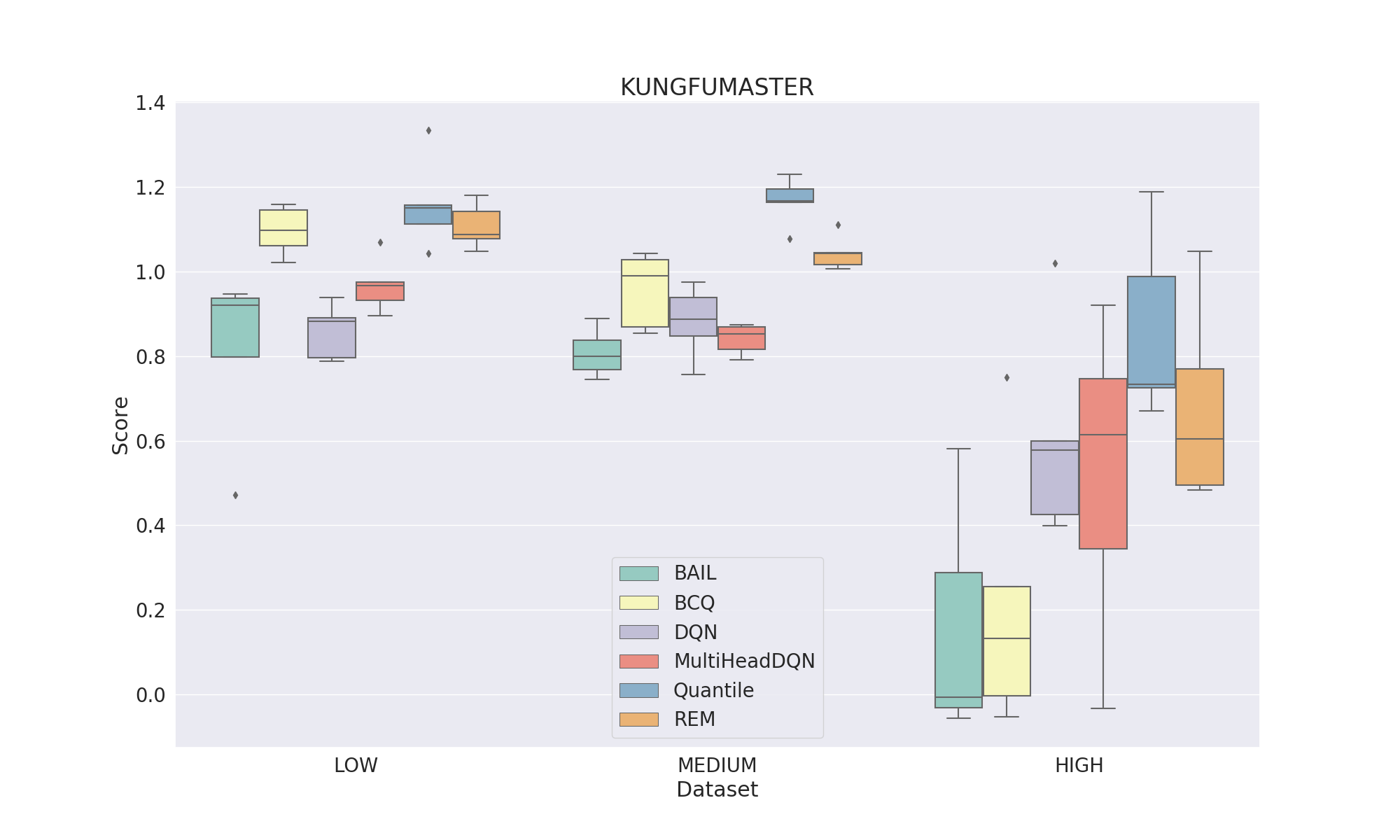}\\
				\vspace{0.01cm}
			\end{minipage}%
		}%
		\subfigure{
			\begin{minipage}[t]{0.333\linewidth}
				\centering
				\includegraphics[width=2.3in]{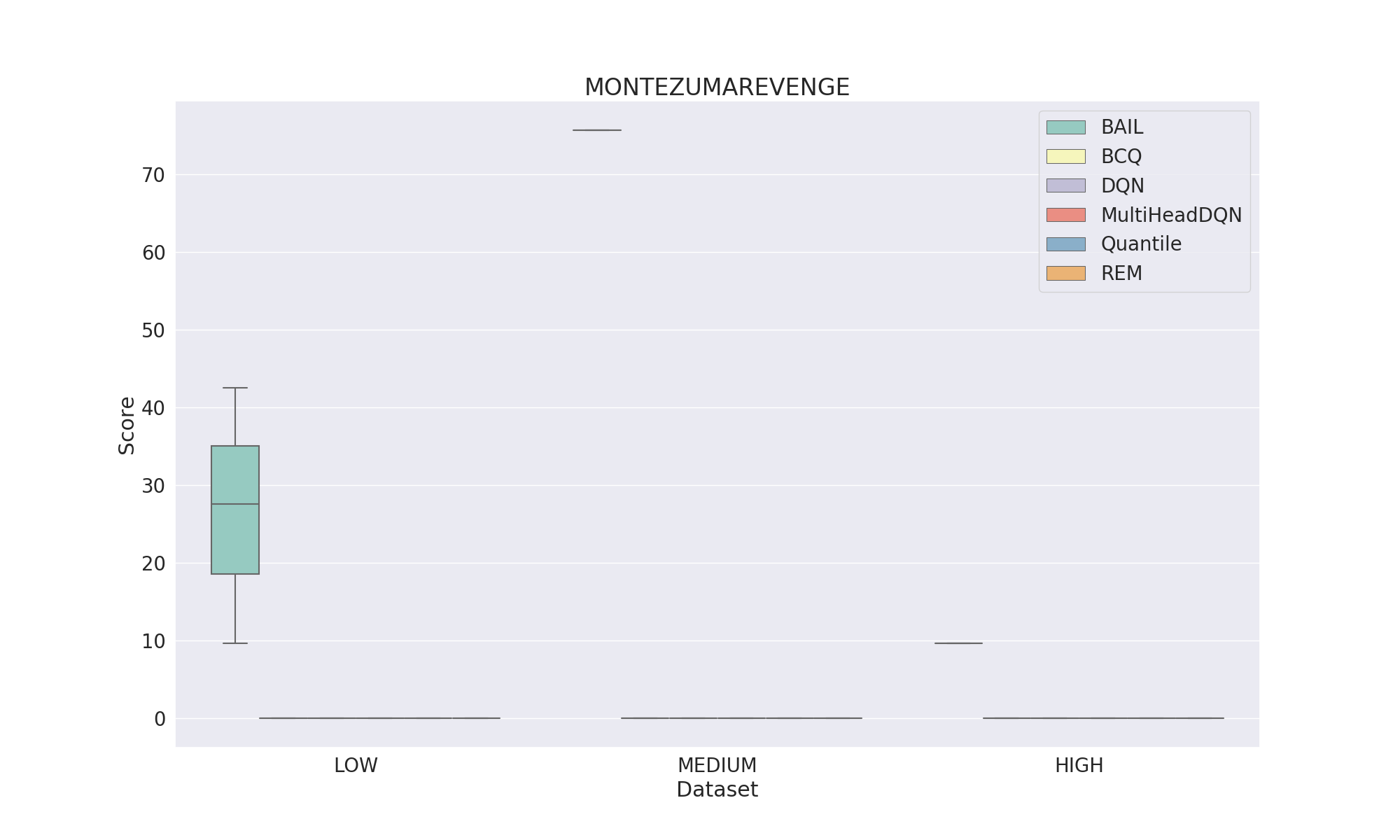}\\
				\vspace{0.01cm}
			\end{minipage}%
		}%
		\subfigure{
			\begin{minipage}[t]{0.333\linewidth}
				\centering
				\includegraphics[width=2.3in]{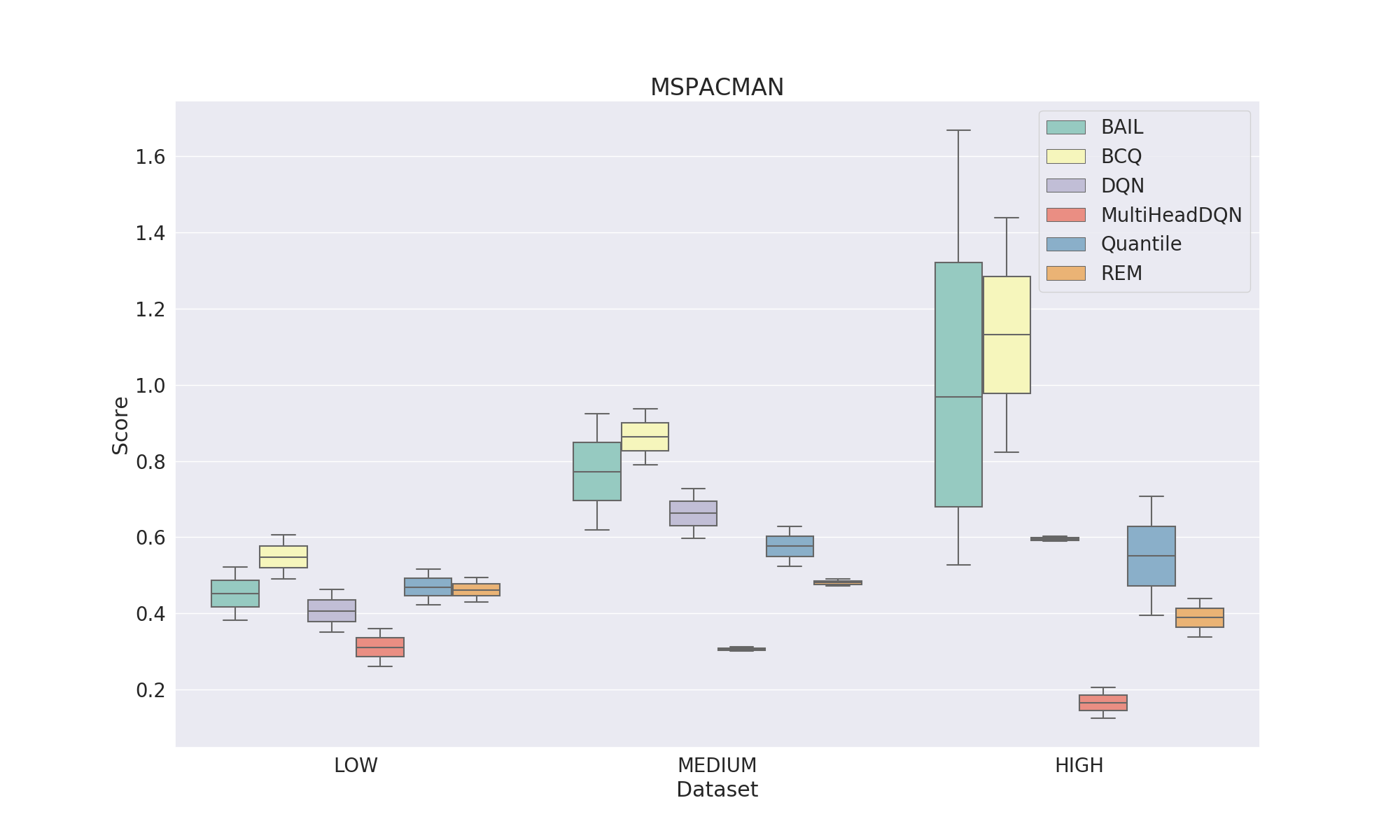}\\
				\vspace{0.01cm}
			\end{minipage}%
		}%

		\centering
		\caption{\textbf{Comparison between baselines on different datasets from Game DemonAttack to Game MsPacman}}
		\label{fig: Comparison between baselines on different datasets from Game DemonAttack to Game MsPacman}
								
	\end{figure*}

	\begin{figure*}[!htb]
		\centering
		
		\subfigure{
			\begin{minipage}[t]{0.333\linewidth}
				\centering
				\includegraphics[width=2.3in]{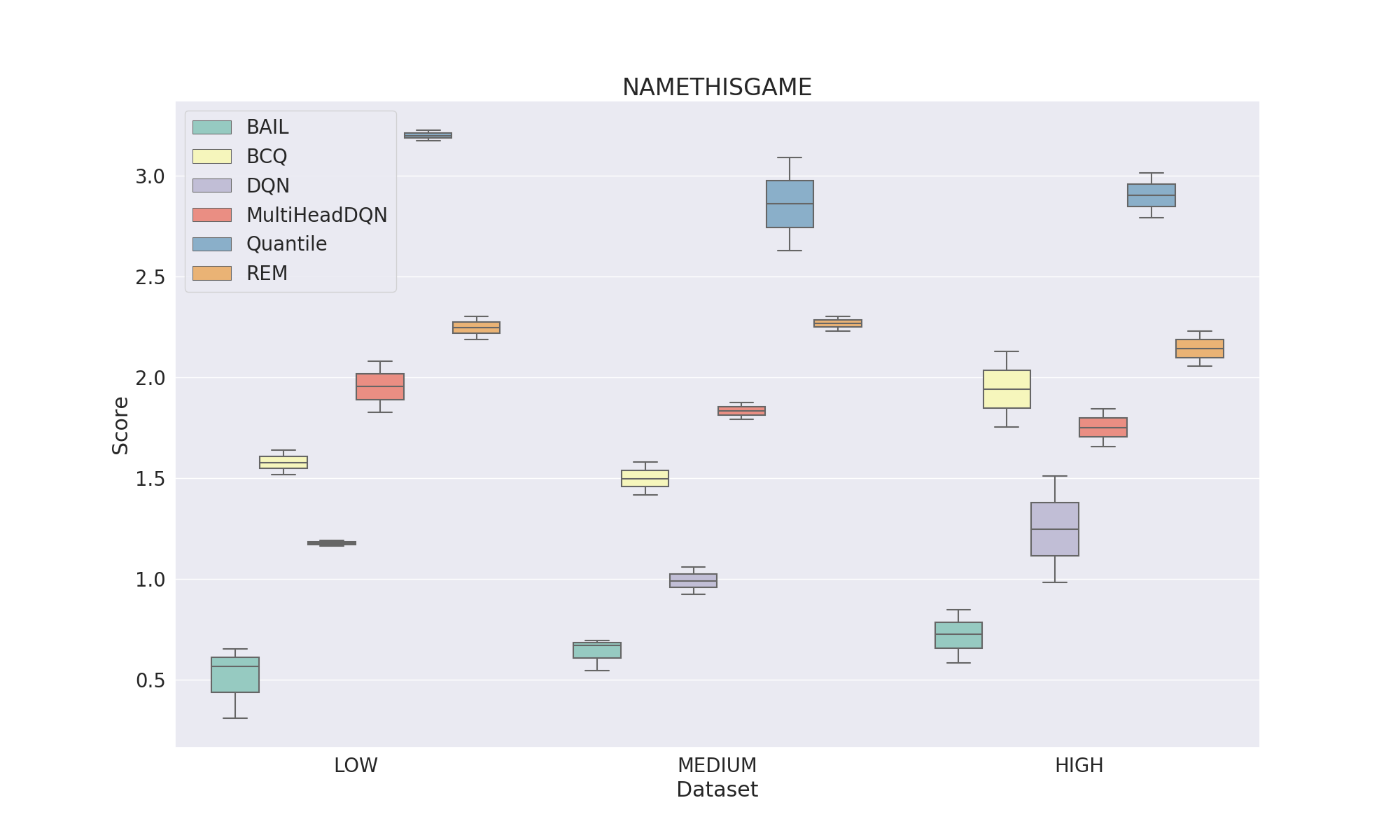}\\
				\vspace{0.01cm}
			\end{minipage}%
		}%
		\subfigure{
			\begin{minipage}[t]{0.333\linewidth}
				\centering
				\includegraphics[width=2.3in]{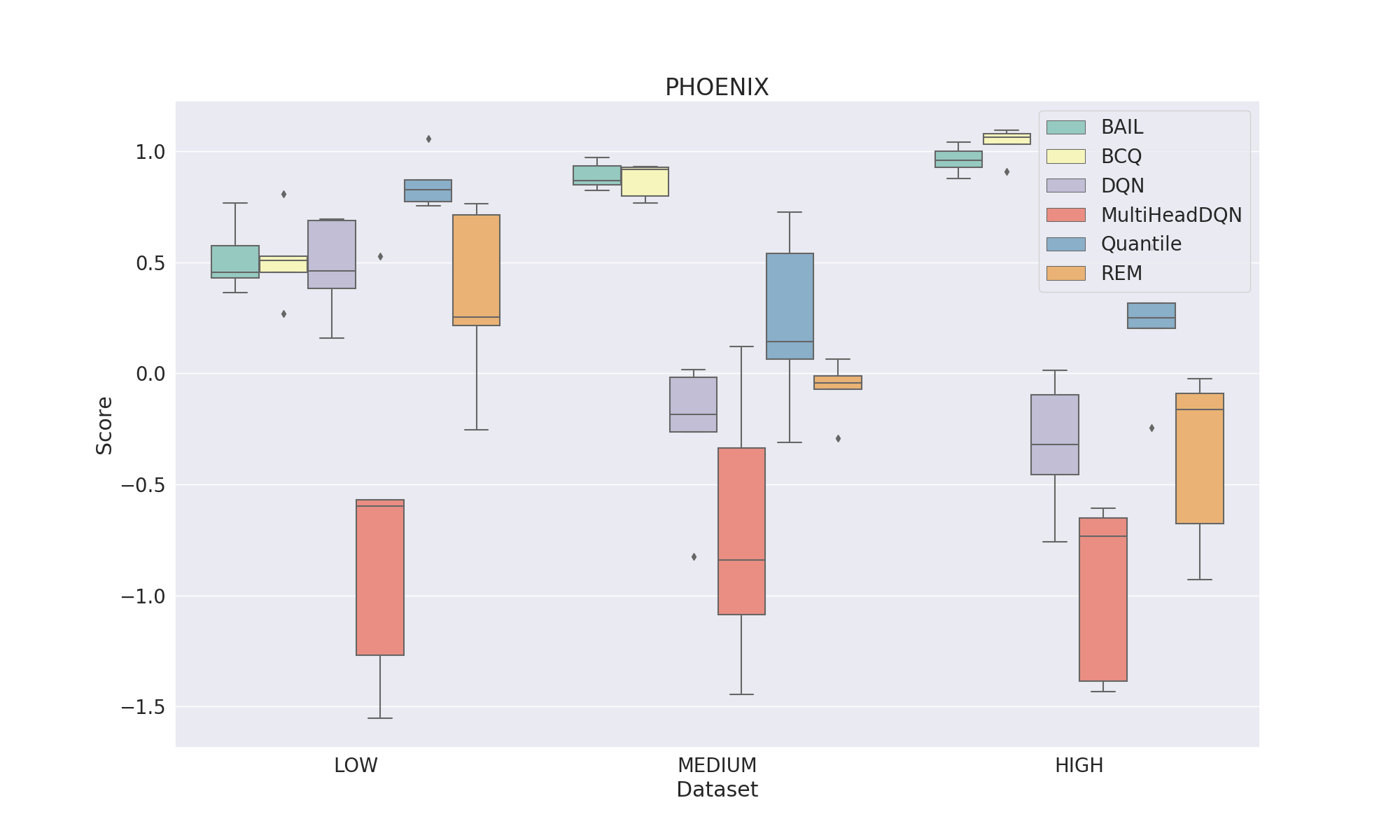}\\
				\vspace{0.01cm}
			\end{minipage}%
		}%
		\subfigure{
			\begin{minipage}[t]{0.333\linewidth}
				\centering
				\includegraphics[width=2.3in]{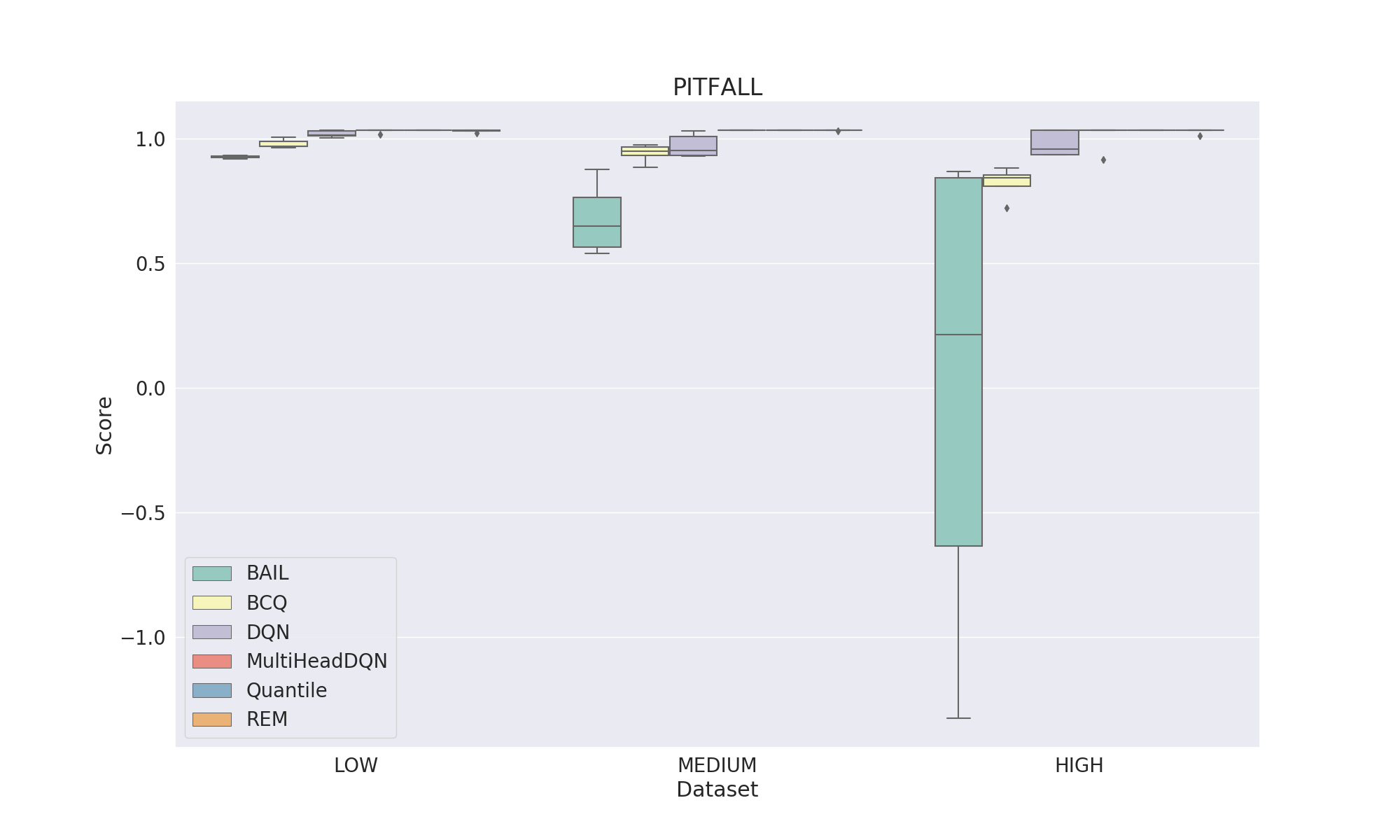}\\
				\vspace{0.01cm}
			\end{minipage}%
		}%

		\subfigure{
			\begin{minipage}[t]{0.333\linewidth}
				\centering
				\includegraphics[width=2.3in]{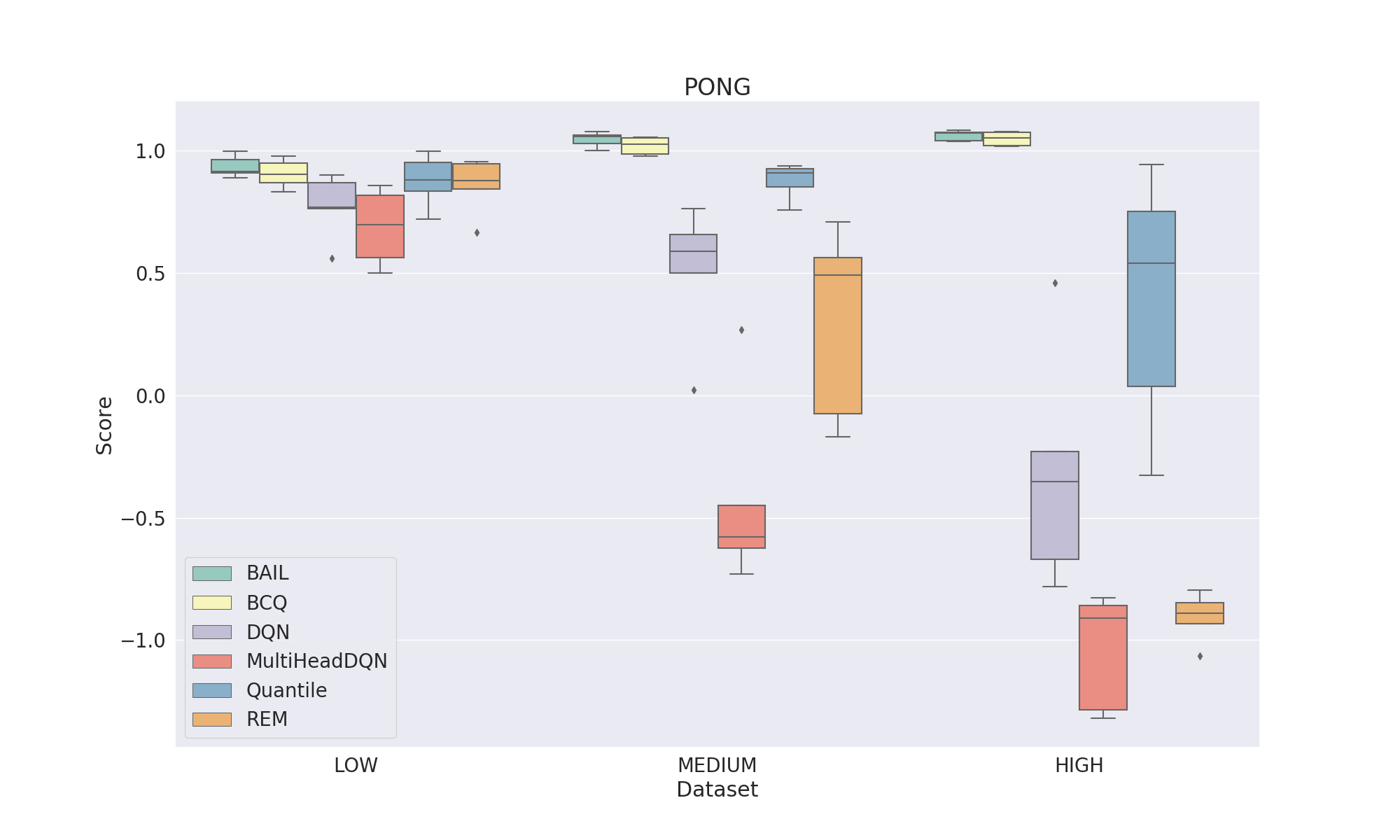}\\
				\vspace{0.01cm}
			\end{minipage}%
		}%
		\subfigure{
			\begin{minipage}[t]{0.333\linewidth}
				\centering
				\includegraphics[width=2.3in]{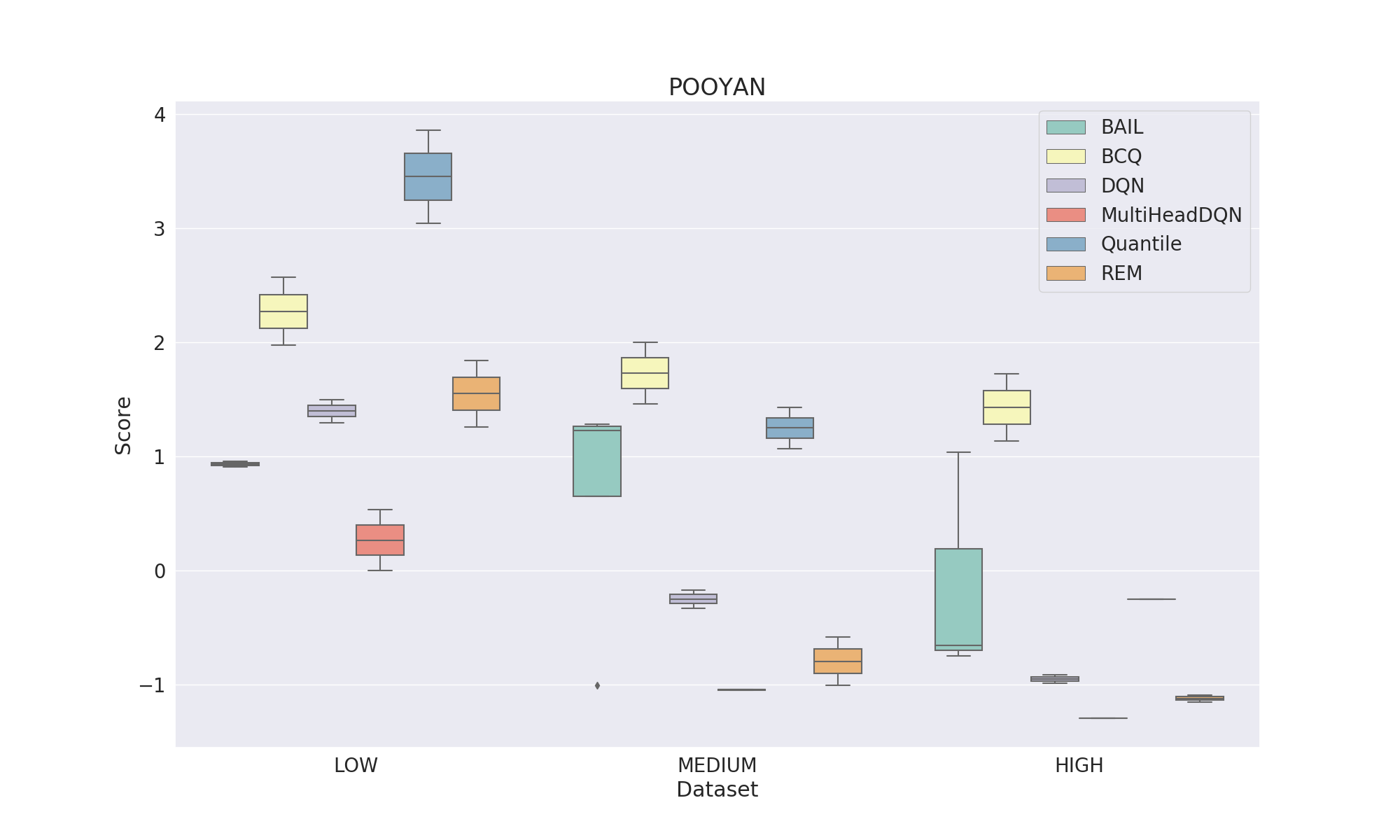}\\
				\vspace{0.01cm}
			\end{minipage}%
		}%
		\subfigure{
			\begin{minipage}[t]{0.333\linewidth}
				\centering
				\includegraphics[width=2.3in]{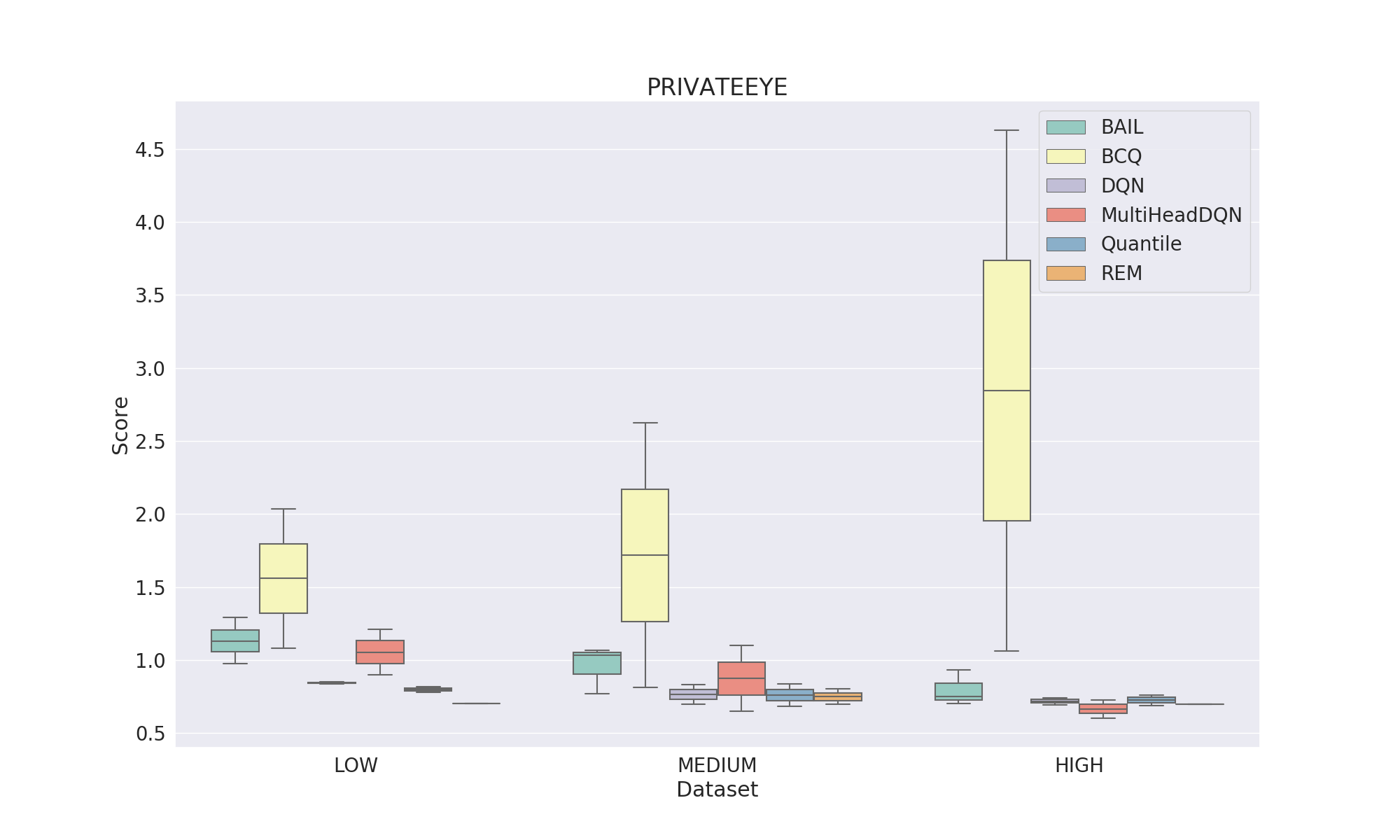}\\
				\vspace{0.01cm}
			\end{minipage}%
		}%

		\subfigure{
			\begin{minipage}[t]{0.333\linewidth}
				\centering
				\includegraphics[width=2.3in]{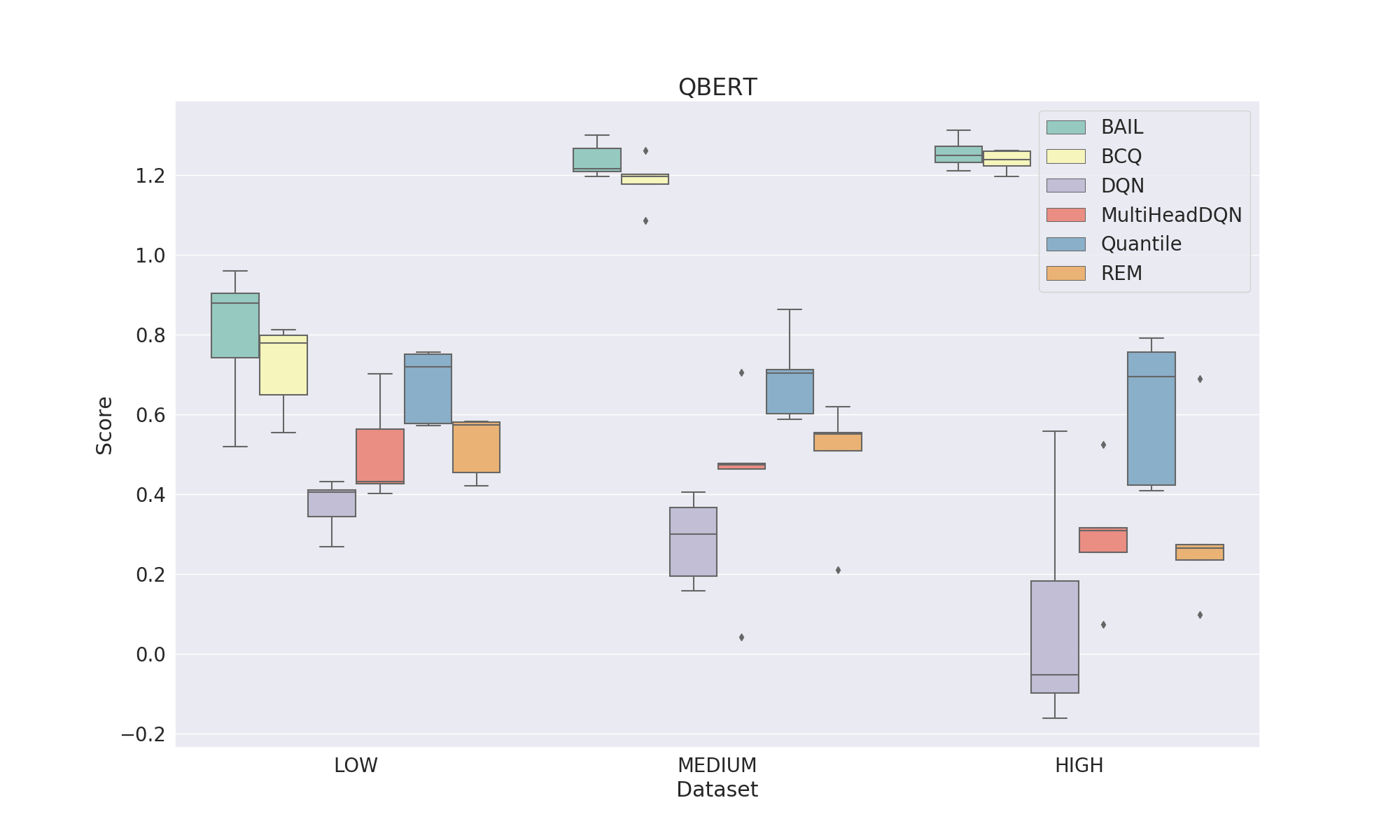}\\
				\vspace{0.01cm}
			\end{minipage}%
		}%
		\subfigure{
			\begin{minipage}[t]{0.333\linewidth}
				\centering
				\includegraphics[width=2.3in]{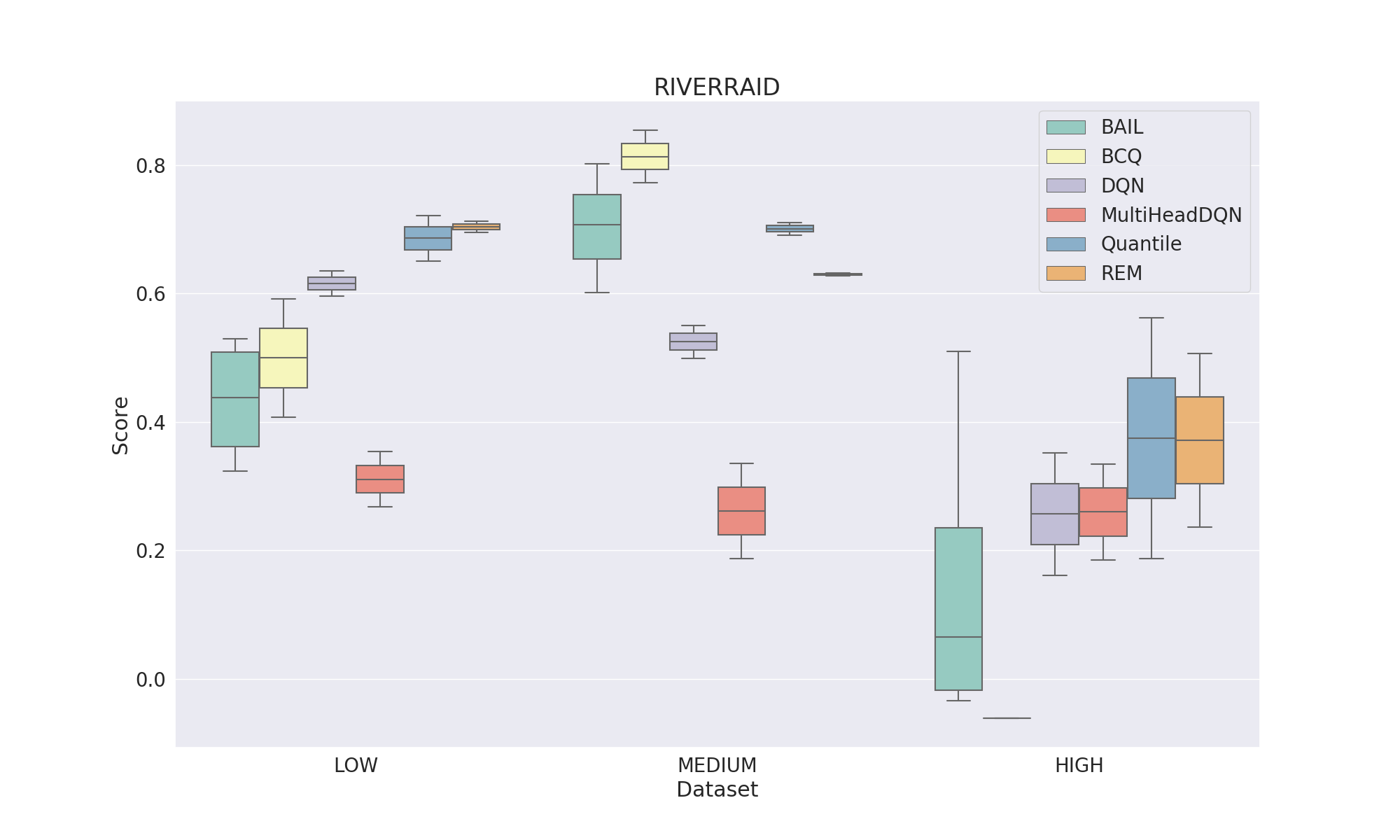}\\
				\vspace{0.01cm}
			\end{minipage}%
		}%
		\subfigure{
			\begin{minipage}[t]{0.333\linewidth}
				\centering
				\includegraphics[width=2.3in]{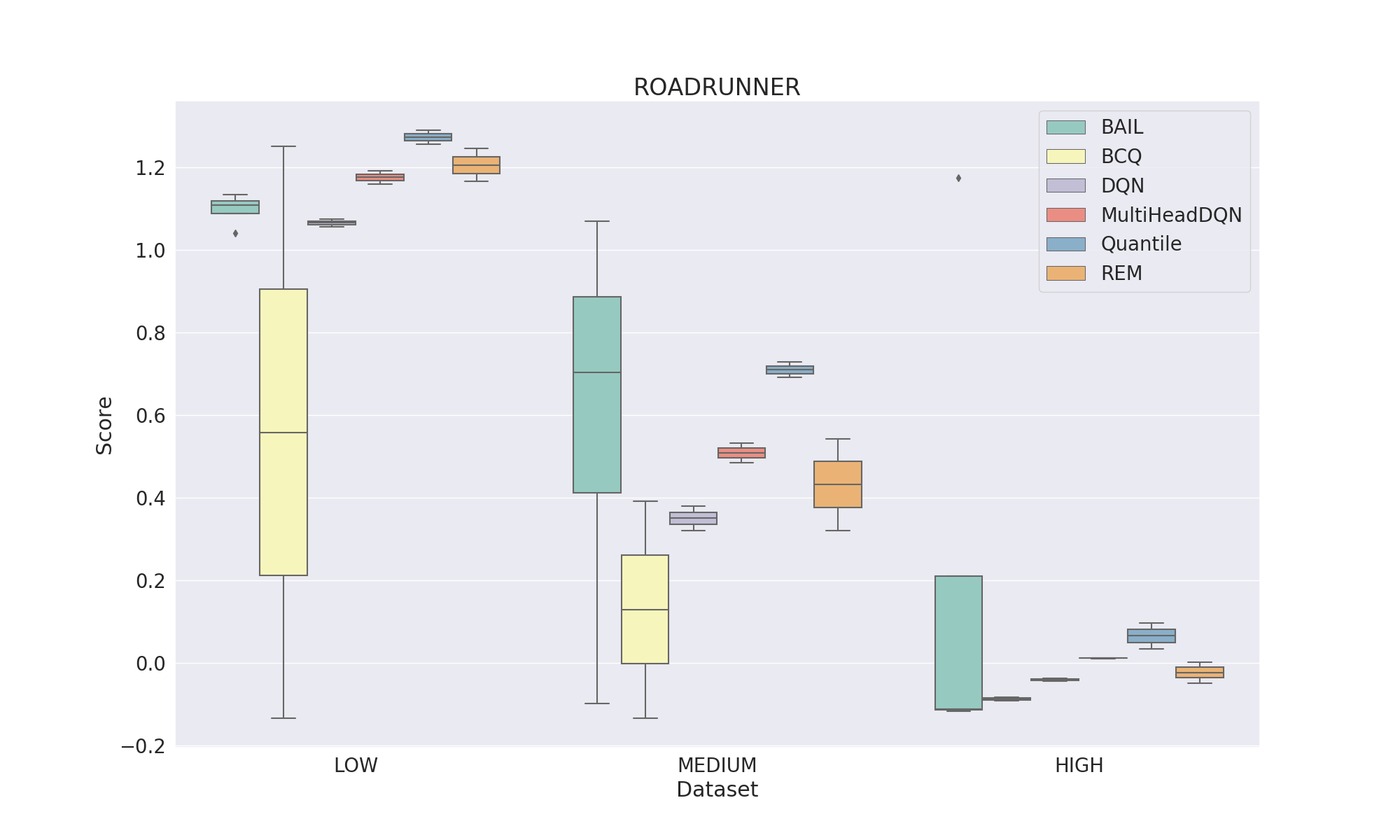}\\
				\vspace{0.01cm}
			\end{minipage}%
		}%

		\subfigure{
			\begin{minipage}[t]{0.333\linewidth}
				\centering
				\includegraphics[width=2.3in]{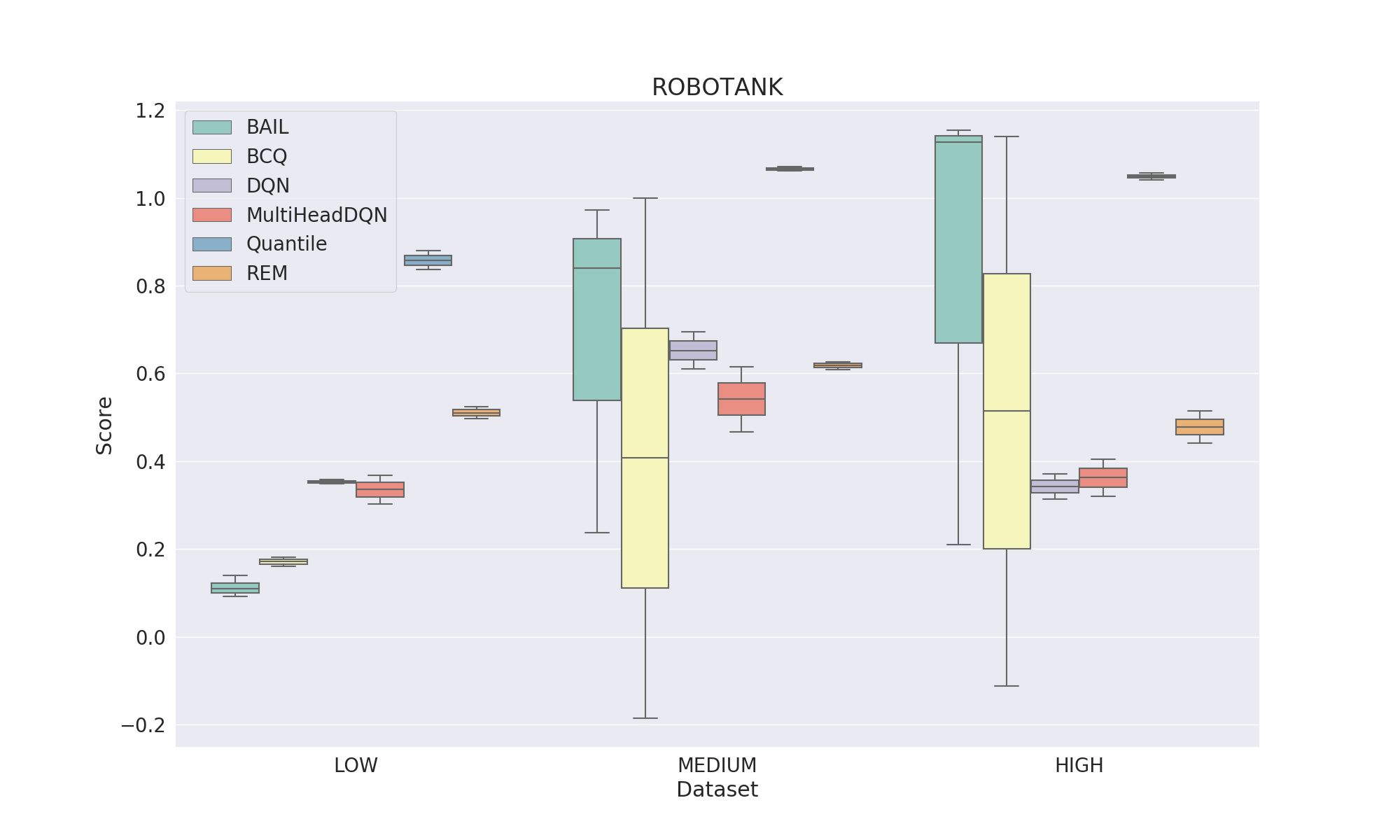}\\
				\vspace{0.01cm}
			\end{minipage}%
		}%
		\subfigure{
			\begin{minipage}[t]{0.333\linewidth}
				\centering
				\includegraphics[width=2.3in]{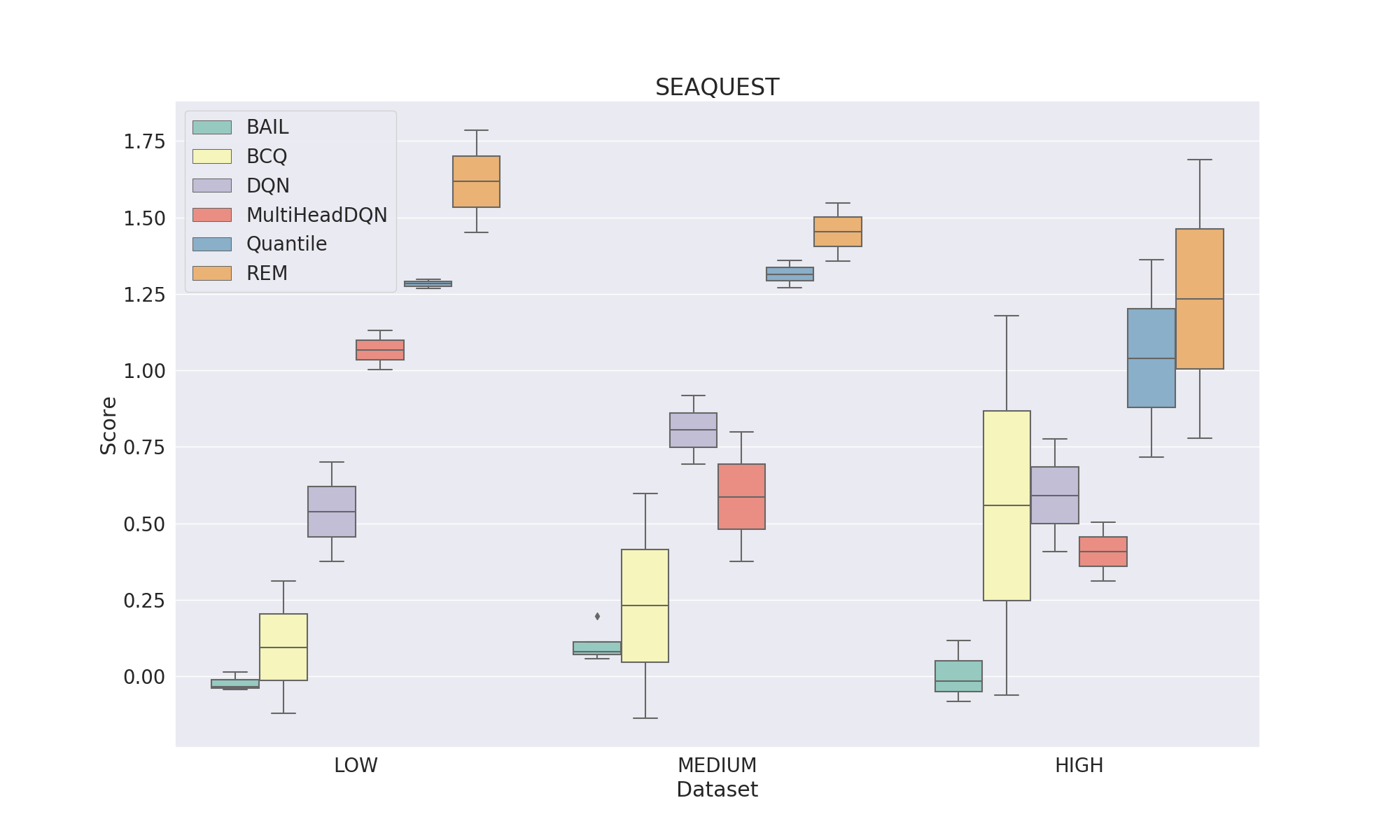}\\
				\vspace{0.01cm}
			\end{minipage}%
		}%
		\subfigure{
			\begin{minipage}[t]{0.333\linewidth}
				\centering
				\includegraphics[width=2.3in]{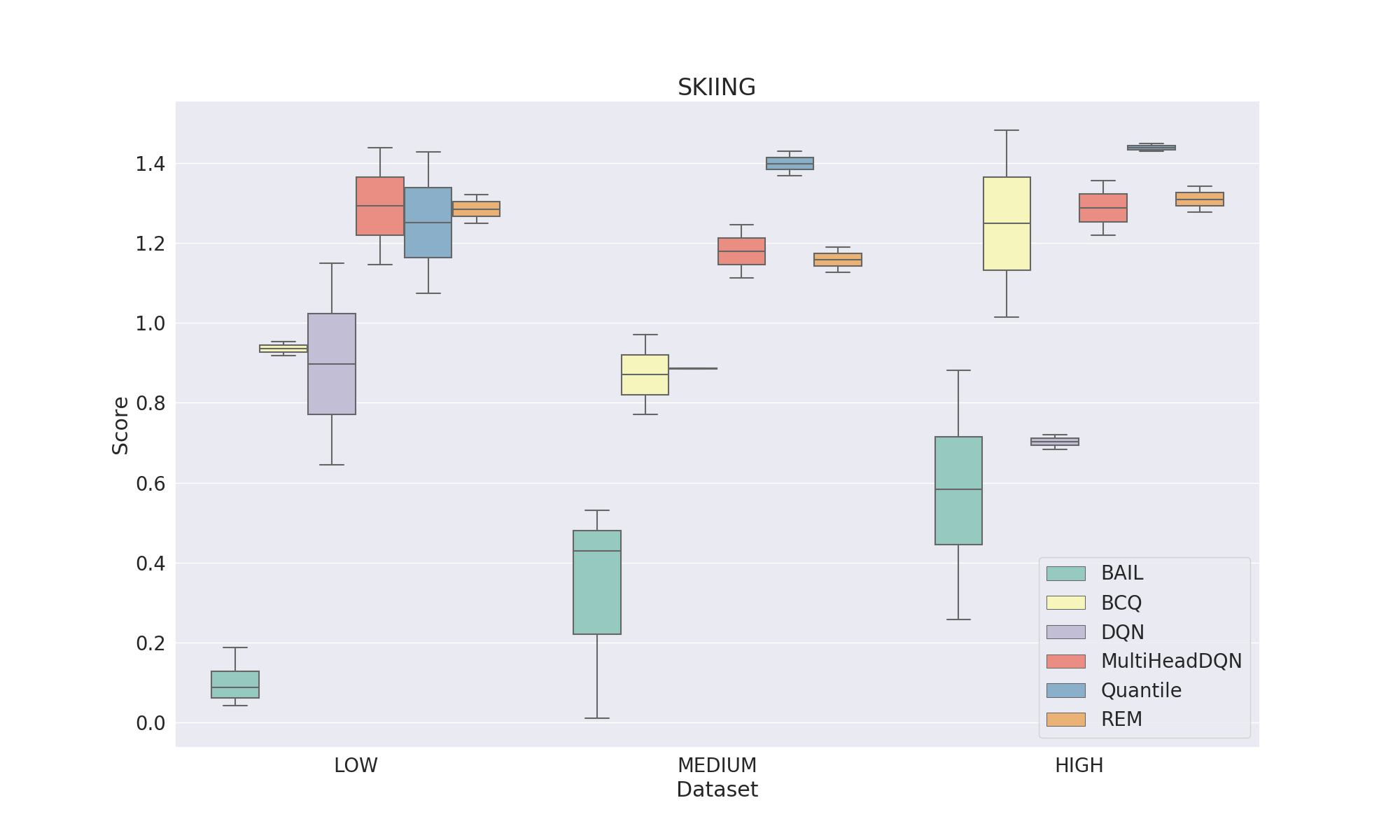}\\
				\vspace{0.01cm}
			\end{minipage}%
		}%

		\subfigure{
			\begin{minipage}[t]{0.333\linewidth}
				\centering
				\includegraphics[width=2.3in]{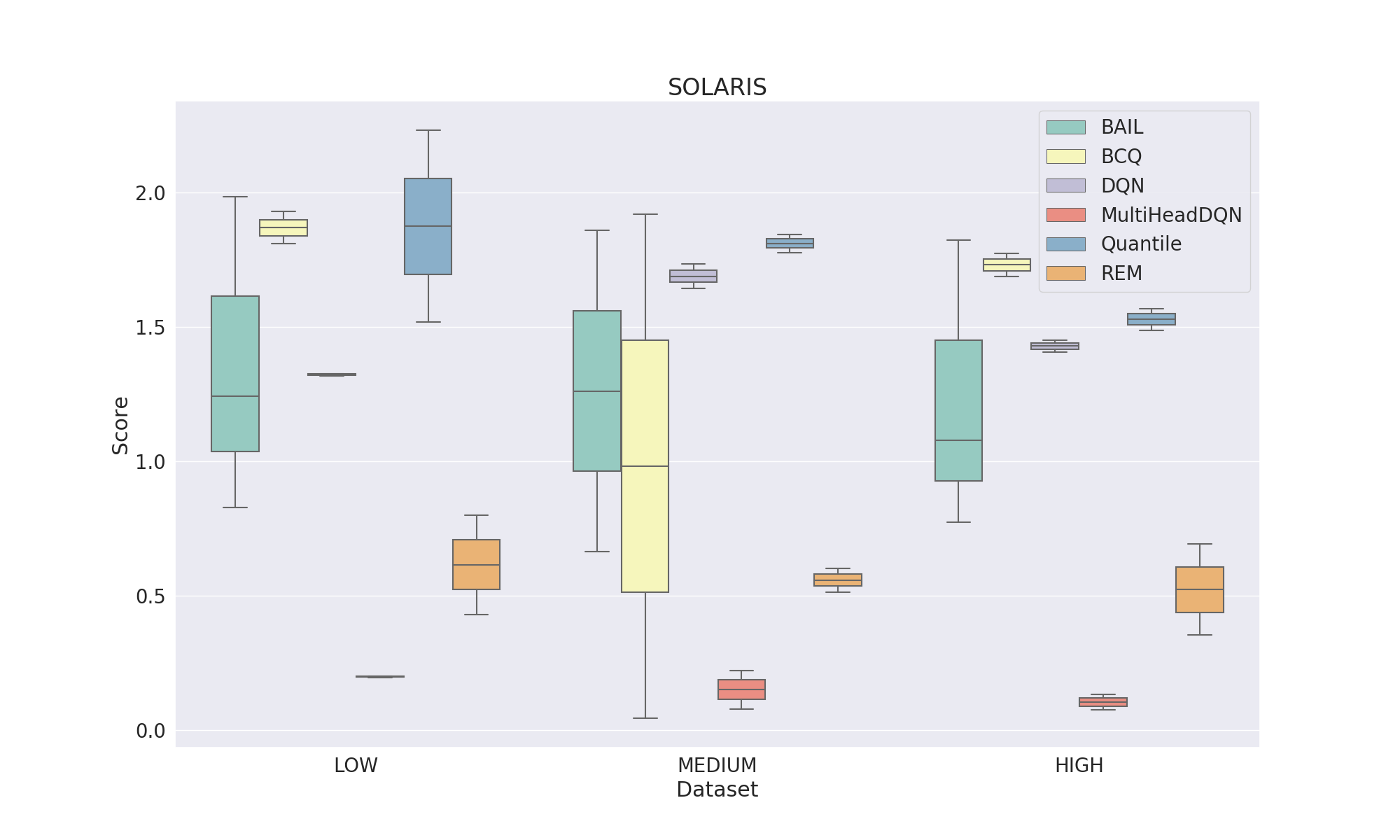}\\
				\vspace{0.01cm}
			\end{minipage}%
		}%
		\subfigure{
			\begin{minipage}[t]{0.333\linewidth}
				\centering
				\includegraphics[width=2.3in]{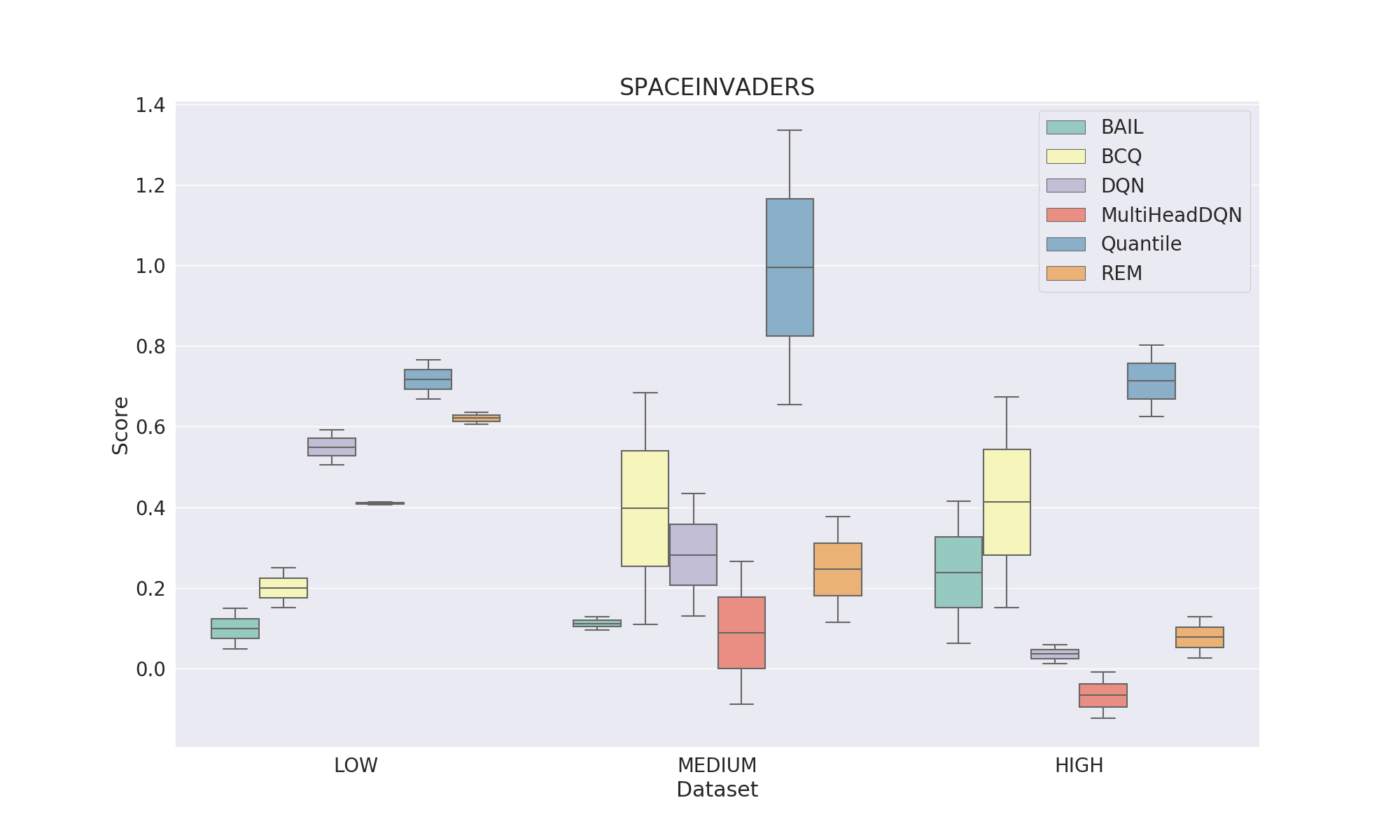}\\
				\vspace{0.01cm}
			\end{minipage}%
		}%
		\subfigure{
			\begin{minipage}[t]{0.333\linewidth}
				\centering
				\includegraphics[width=2.3in]{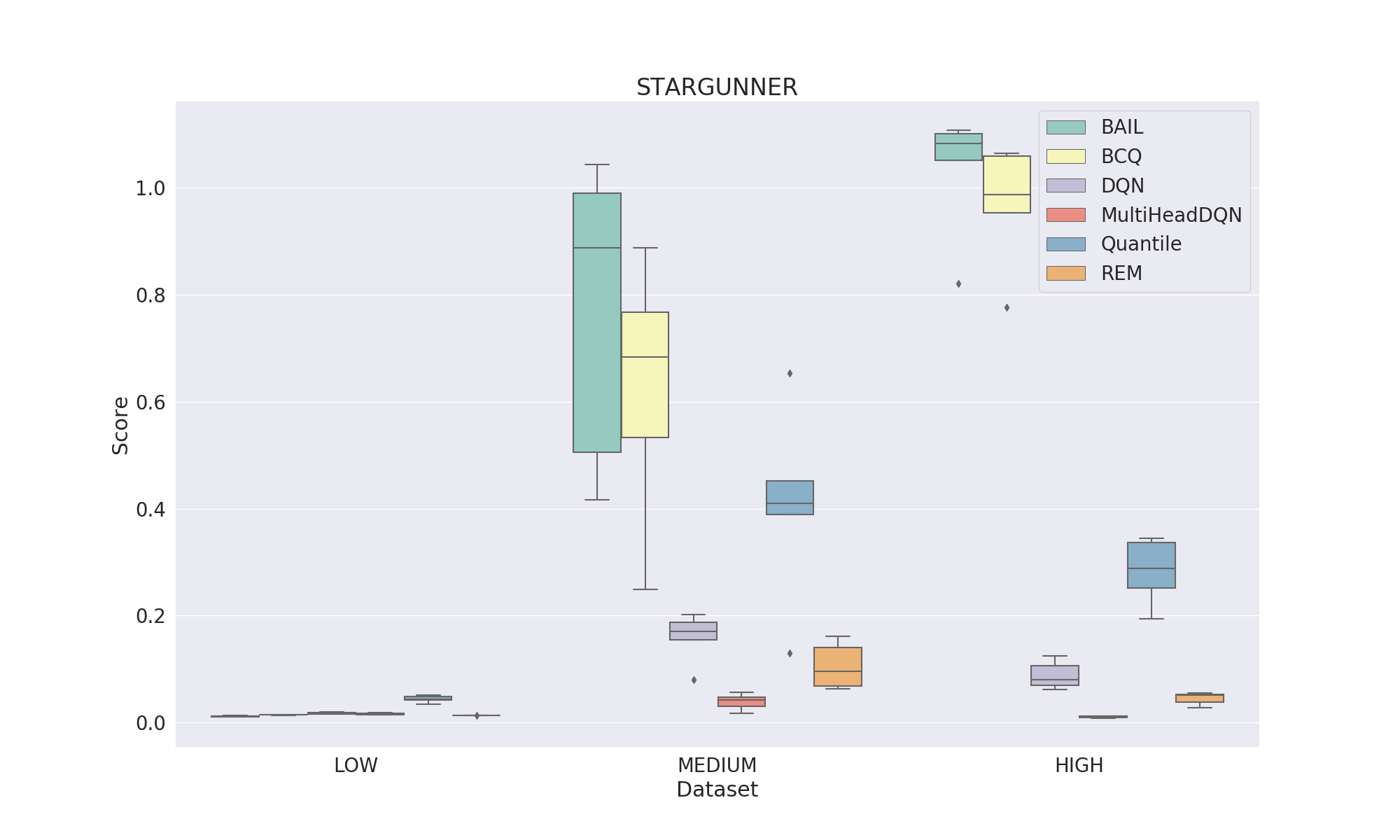}\\
				\vspace{0.01cm}
			\end{minipage}%
		}%

		\subfigure{
			\begin{minipage}[t]{0.333\linewidth}
				\centering
				\includegraphics[width=2.3in]{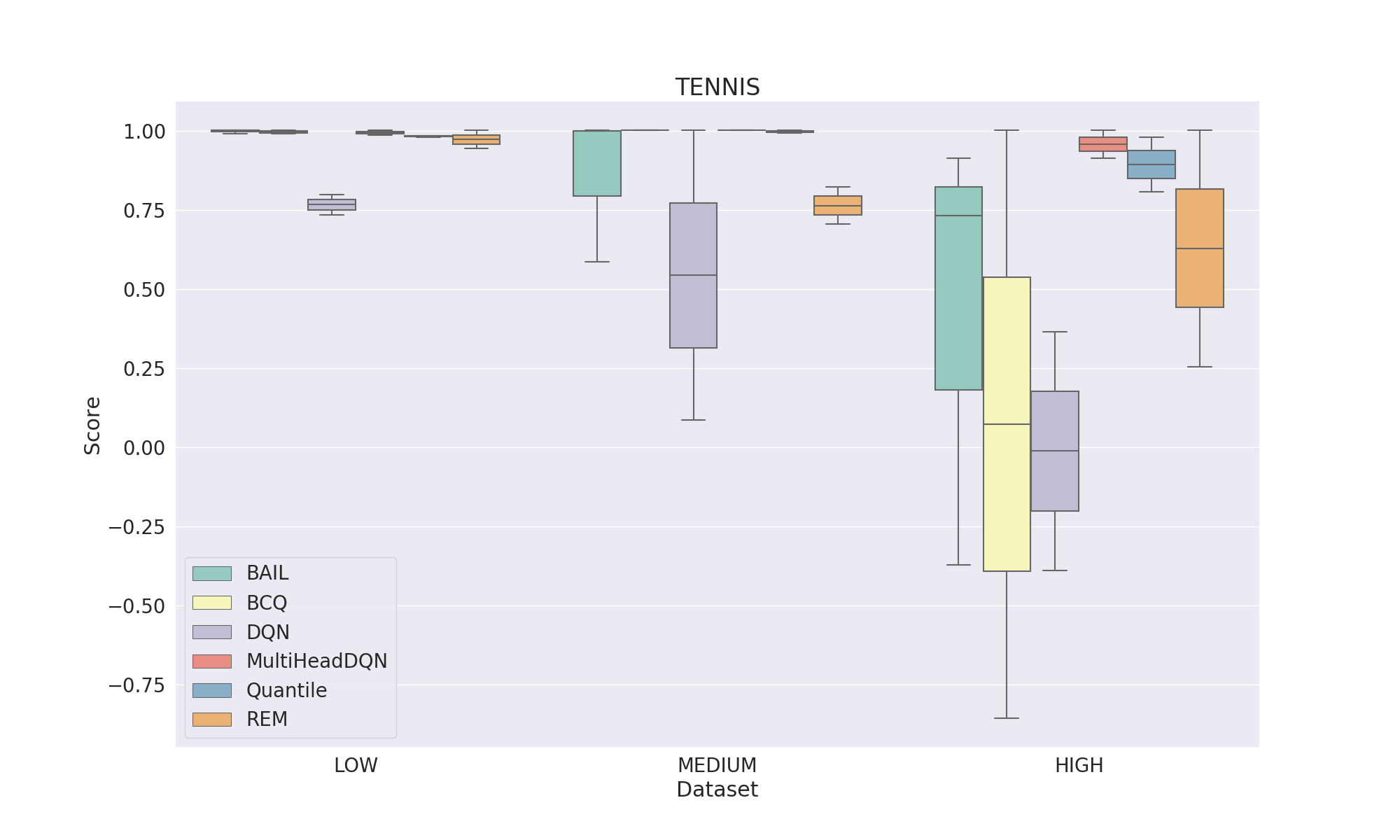}\\
				\vspace{0.01cm}
			\end{minipage}%
		}%
		\subfigure{
			\begin{minipage}[t]{0.333\linewidth}
				\centering
				\includegraphics[width=2.3in]{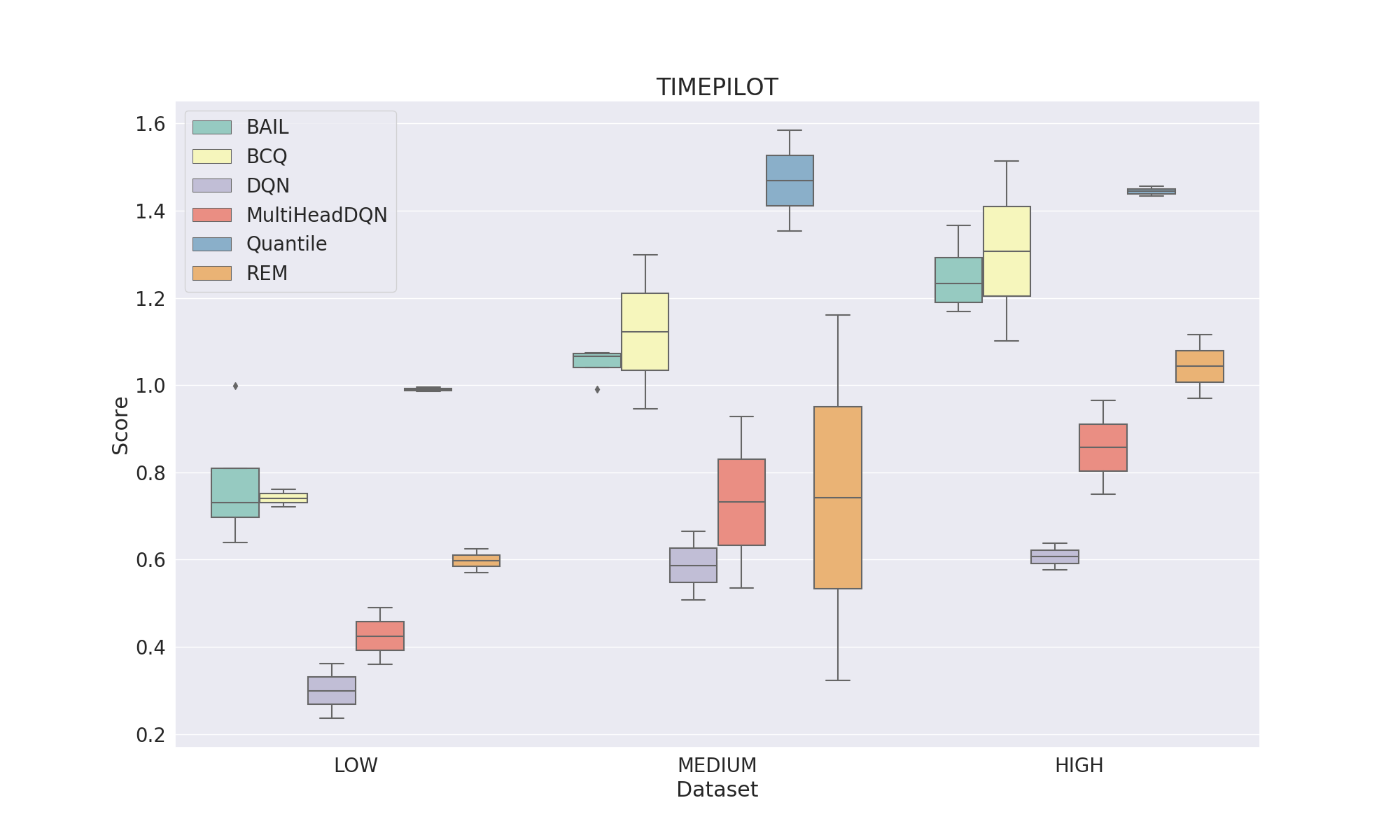}\\
				\vspace{0.01cm}
			\end{minipage}%
		}%
		\subfigure{
			\begin{minipage}[t]{0.333\linewidth}
				\centering
				\includegraphics[width=2.3in]{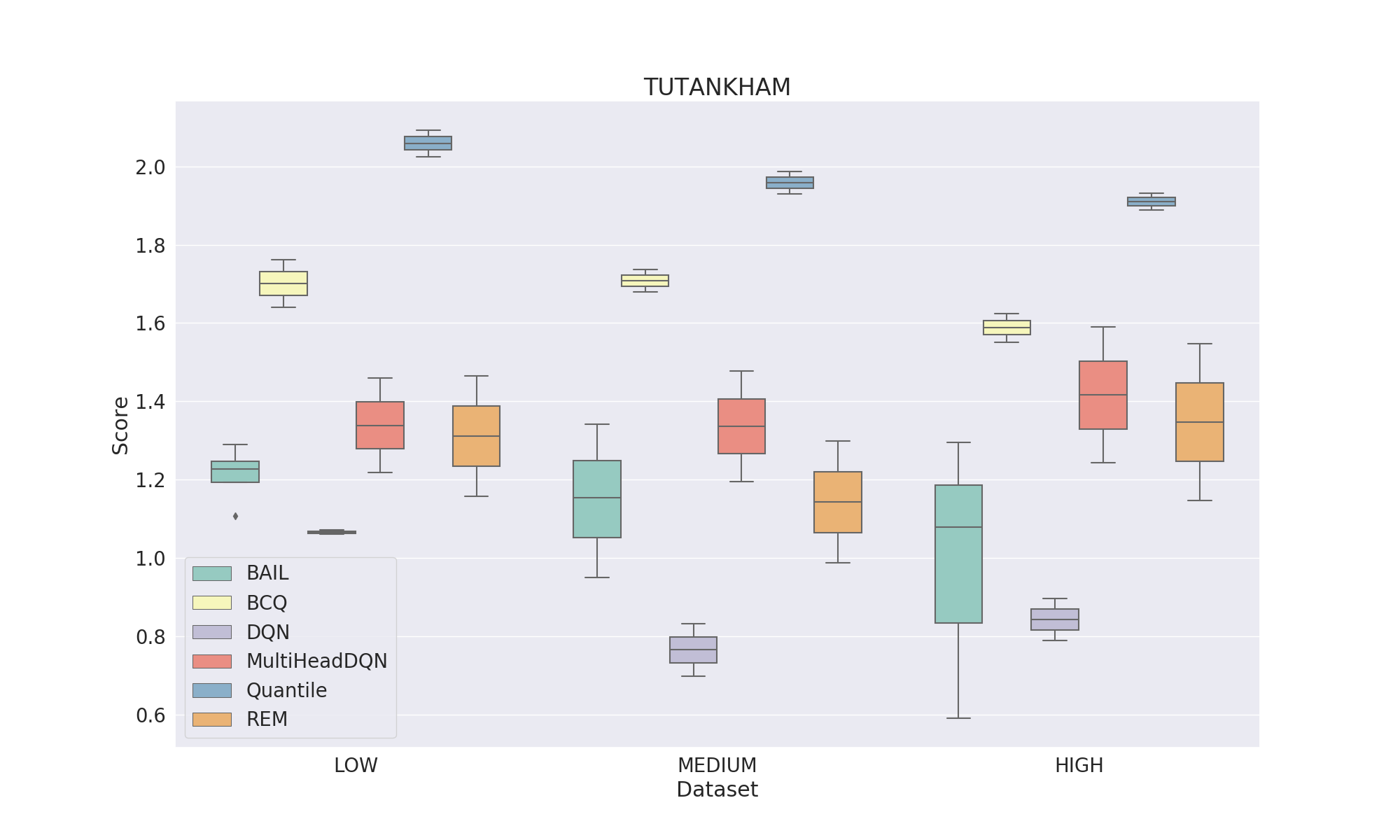}\\
				\vspace{0.01cm}
			\end{minipage}%
		}%

		\centering
		\caption{\textbf{Comparison between baselines on different datasets from Game NameThisGame to Game Tutankham}}
		\label{fig: Comparison between baselines on different datasets from Game NameThisGame to Game Tutankham}
								
	\end{figure*}

	\begin{figure*}[htb]
		\centering

		\vspace{-8cm}

		\subfigure{
			\begin{minipage}[t]{0.333\linewidth}
				\centering
				\includegraphics[width=2.3in]{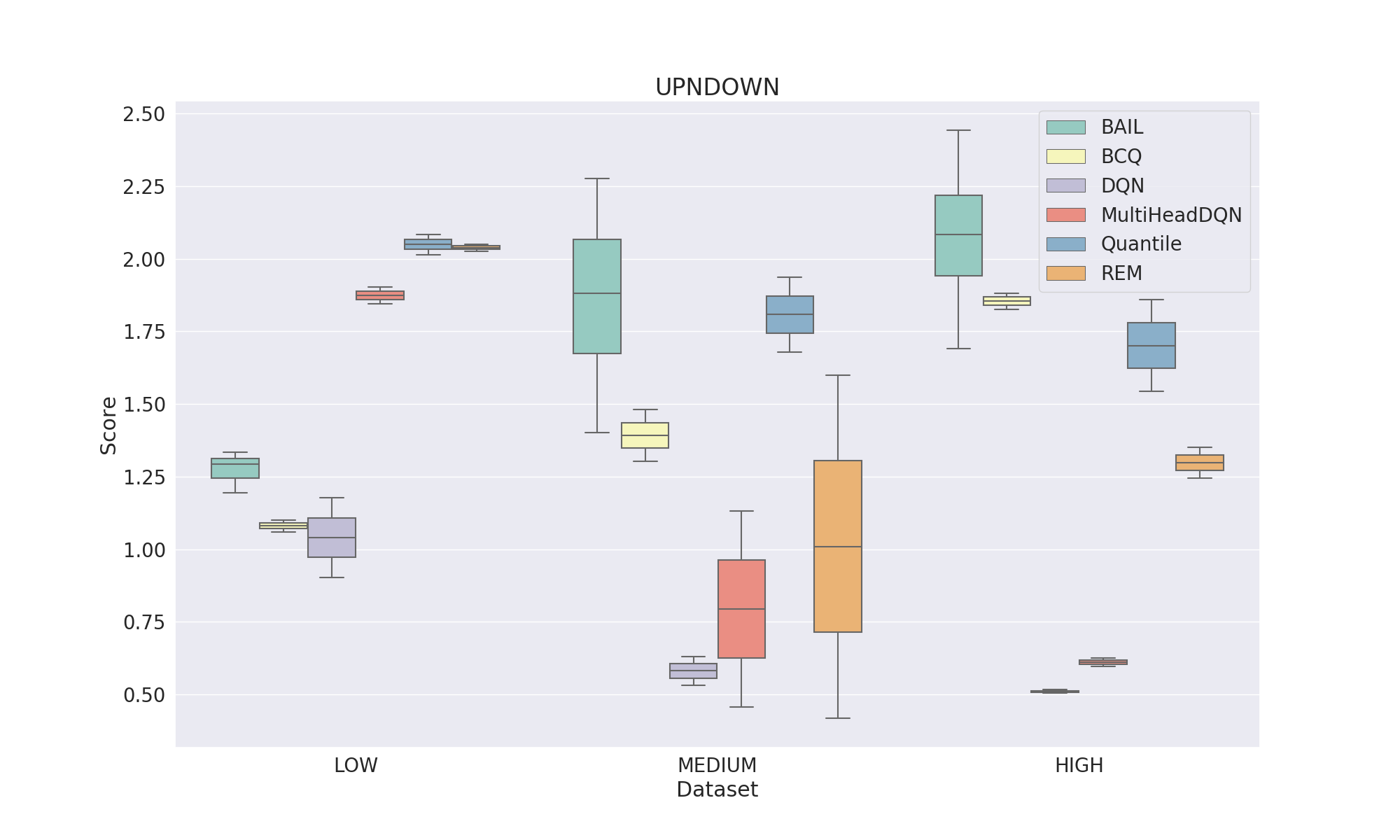}\\
				\vspace{0.01cm}
			\end{minipage}%
		}%
		\subfigure{
			\begin{minipage}[t]{0.333\linewidth}
				\centering
				\includegraphics[width=2.3in]{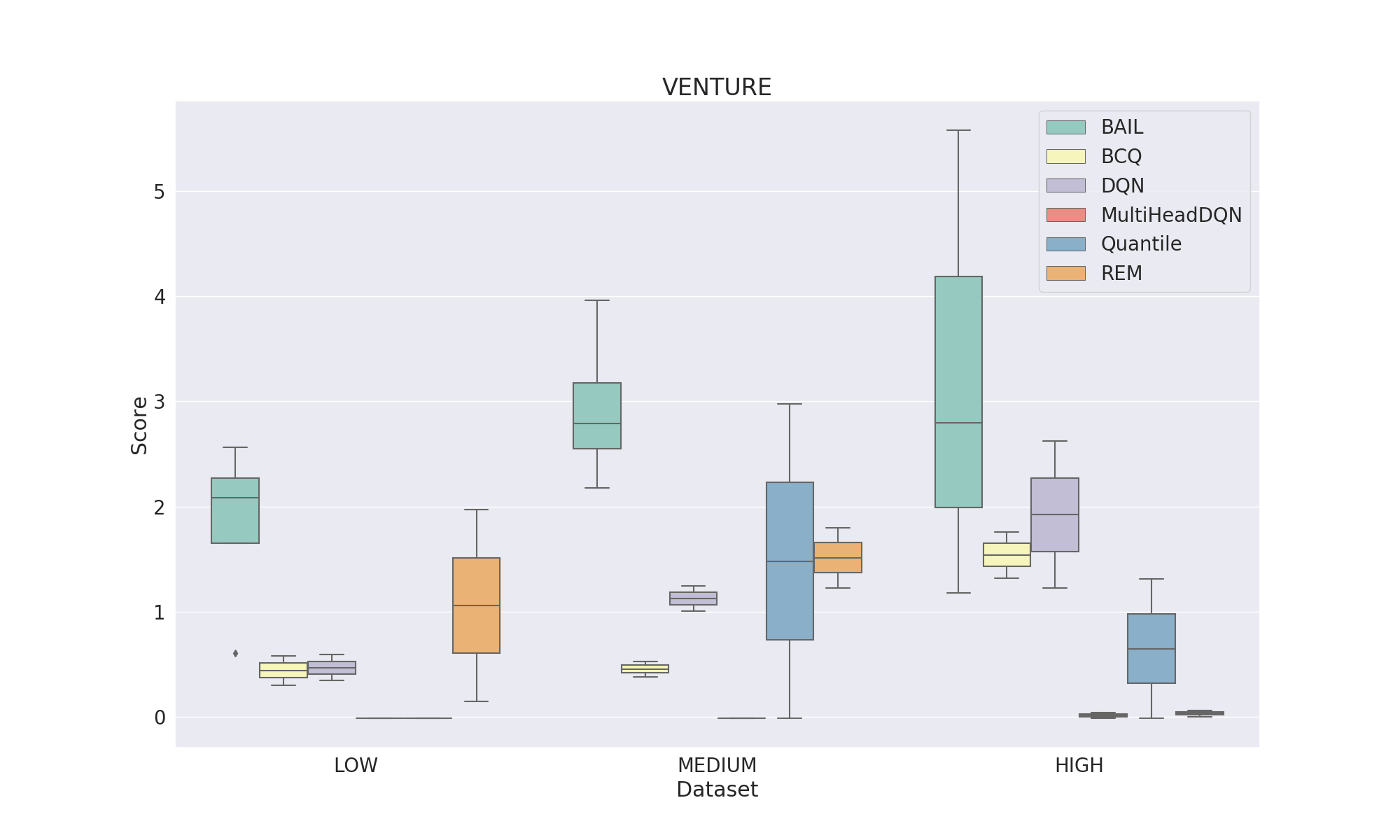}\\
				\vspace{0.01cm}
			\end{minipage}%
		}%
		\subfigure{
			\begin{minipage}[t]{0.333\linewidth}
				\centering
				\includegraphics[width=2.3in]{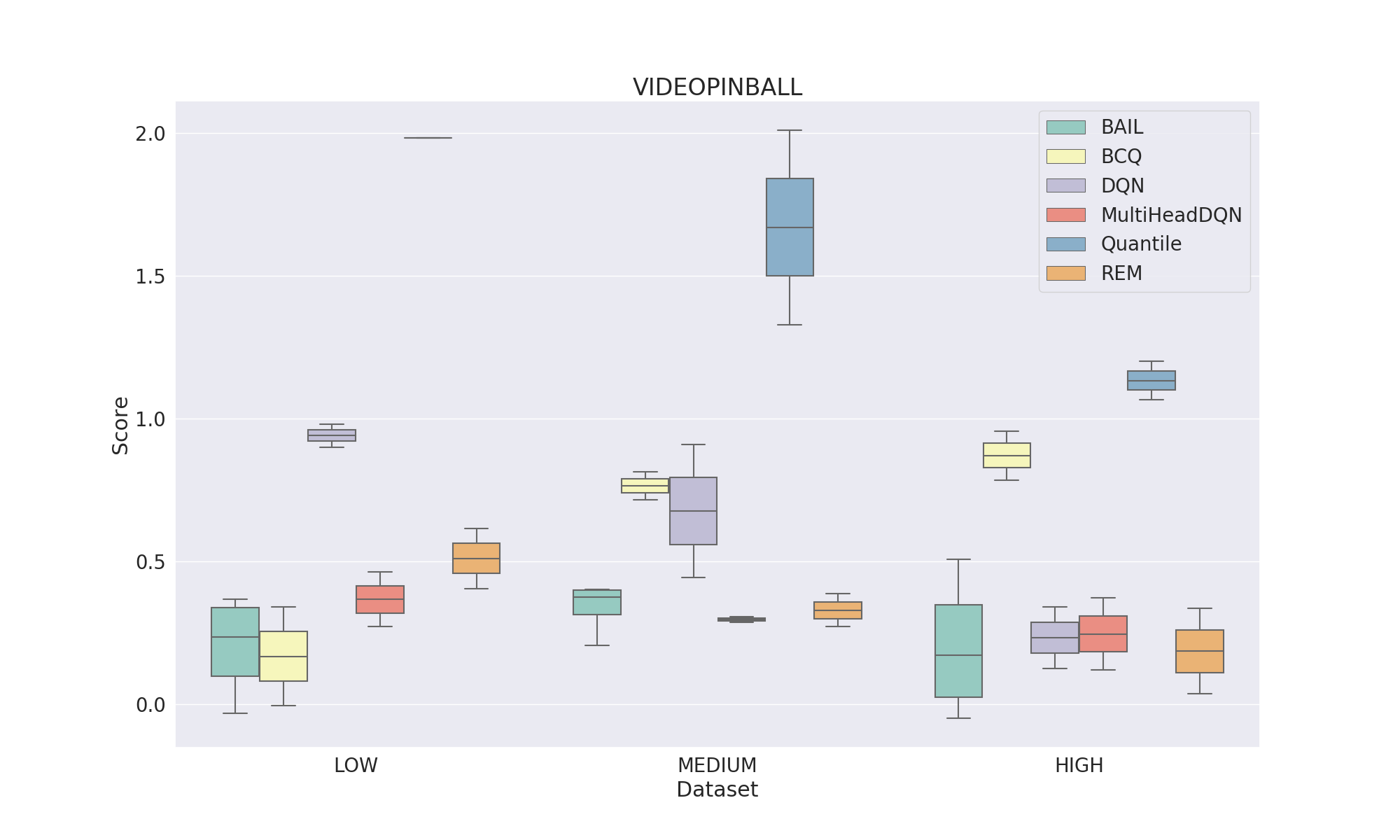}\\
				\vspace{0.01cm}
			\end{minipage}%
		}%

		\subfigure{
			\begin{minipage}[t]{0.333\linewidth}
				\centering
				\includegraphics[width=2.3in]{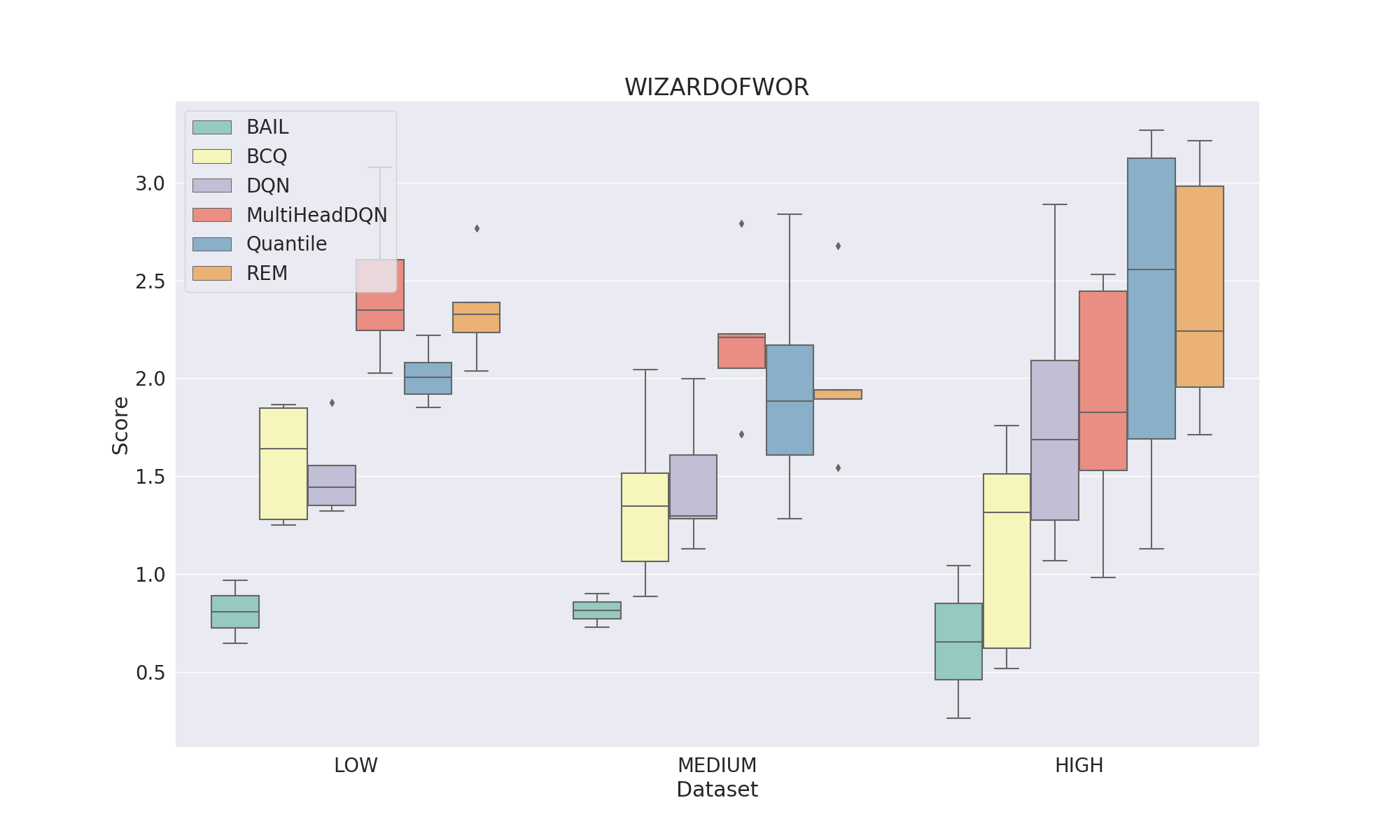}\\
				\vspace{0.01cm}
			\end{minipage}%
		}%
		\subfigure{
			\begin{minipage}[t]{0.333\linewidth}
				\centering
				\includegraphics[width=2.3in]{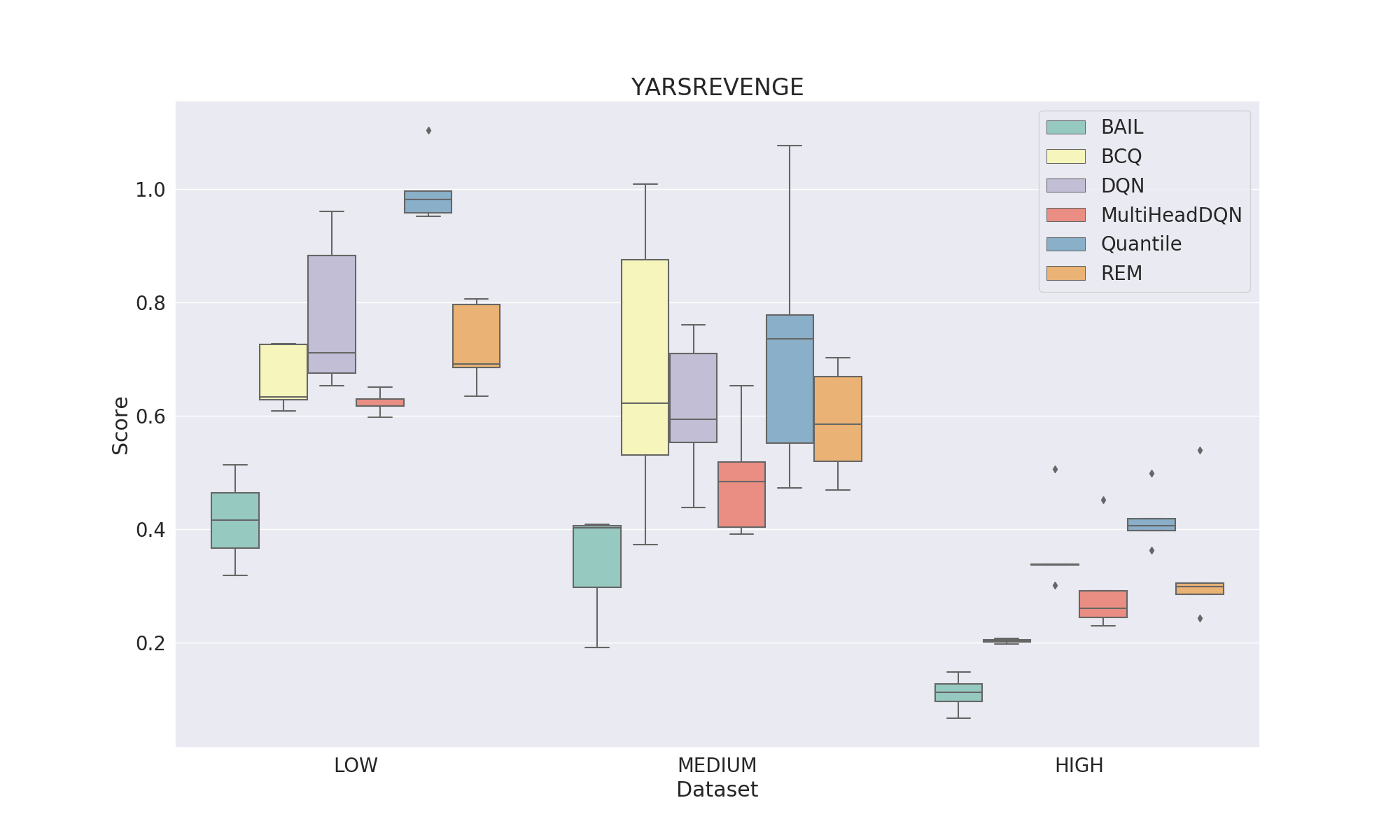}\\
				\vspace{0.01cm}
			\end{minipage}%
		}%
		\subfigure{
			\begin{minipage}[t]{0.333\linewidth}
				\centering
				\includegraphics[width=2.3in]{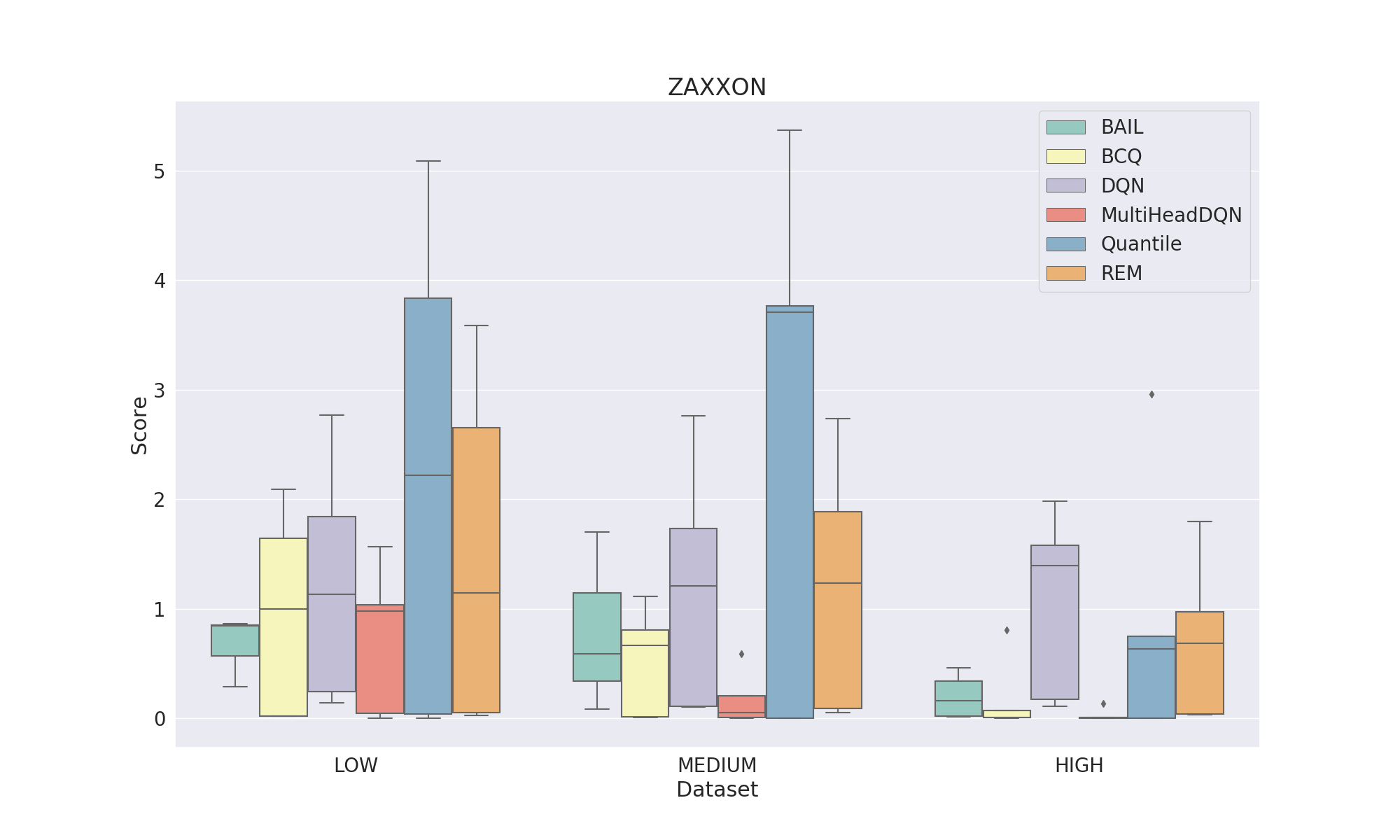}\\
				\vspace{0.01cm}
			\end{minipage}%
		}%

		\centering
		\caption{\textbf{Comparison between baselines on different datasets from Game UpNDown to Game Zaxxon}}
		\label{fig: Comparison between baselines on different datasets from Game UpNDown to Game Zaxxon}
								
	\end{figure*}

	\begin{figure*}[!htb]
		\centering
		\subfigure{
			\begin{minipage}[t]{0.333\linewidth}
				\centering
				\includegraphics[width=2.3in]{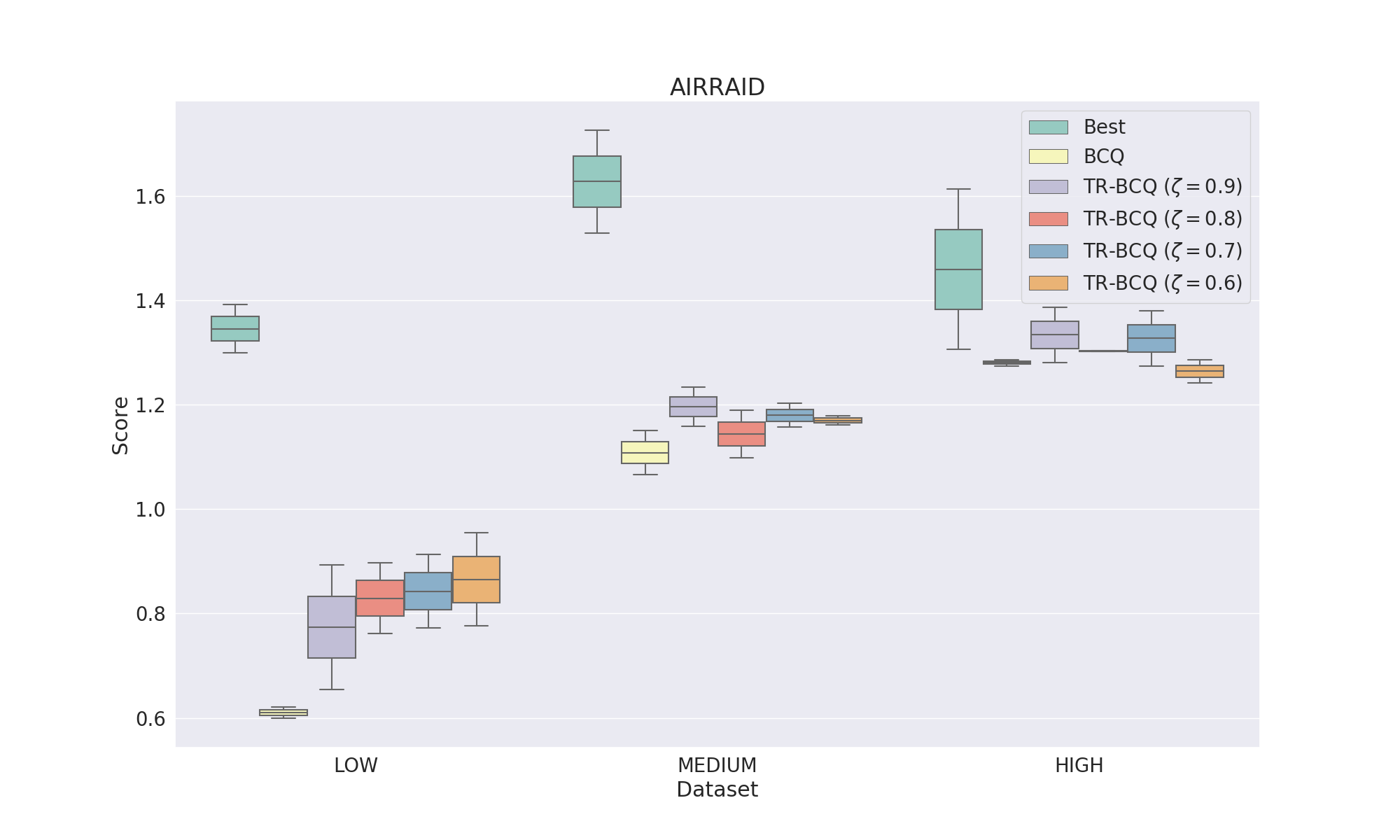}\\
				\vspace{0.01cm}
			\end{minipage}%
		}%
		\subfigure{
			\begin{minipage}[t]{0.333\linewidth}
				\centering
				\includegraphics[width=2.3in]{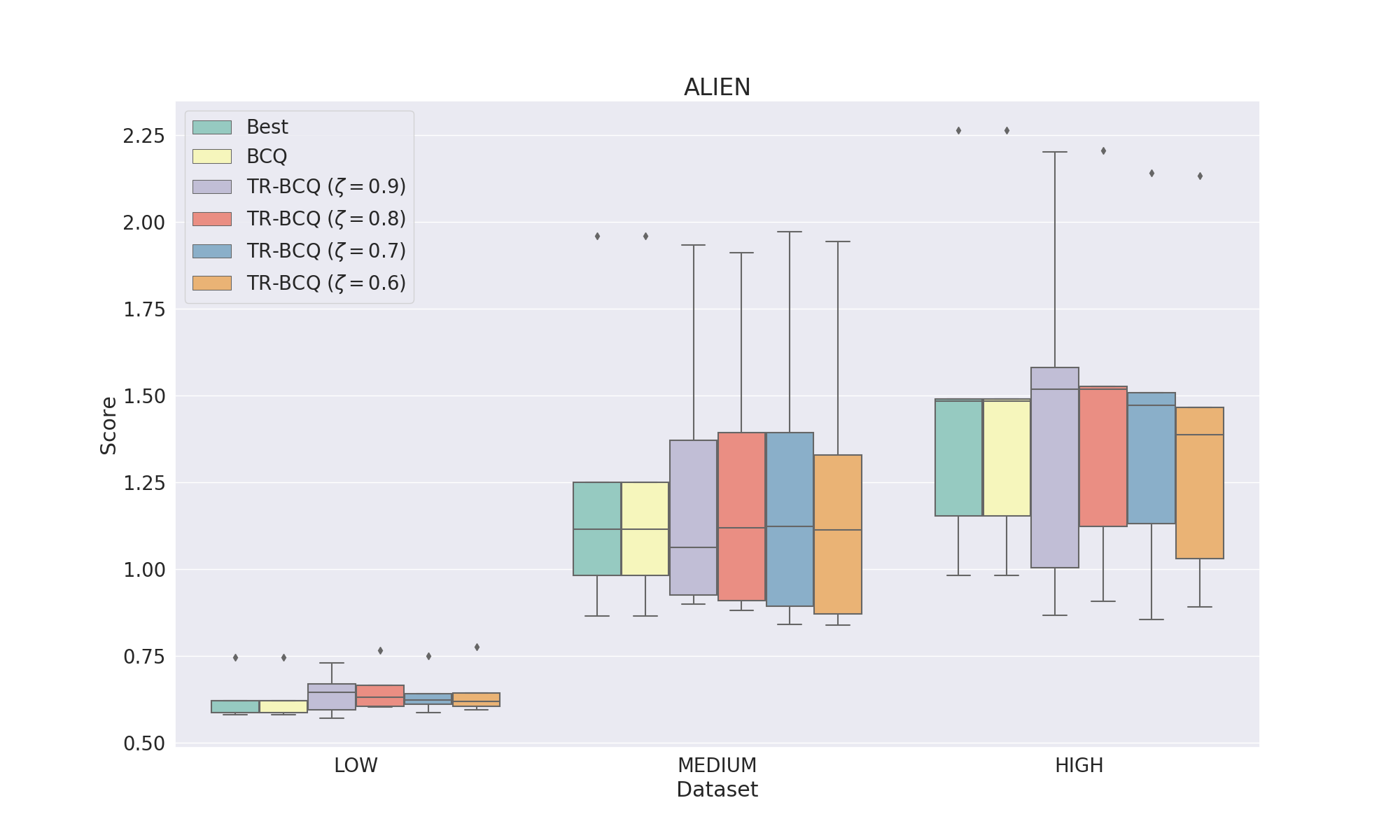}\\
				\vspace{0.01cm}
			\end{minipage}%
		}%
		\subfigure{
			\begin{minipage}[t]{0.333\linewidth}
				\centering
				\includegraphics[width=2.3in]{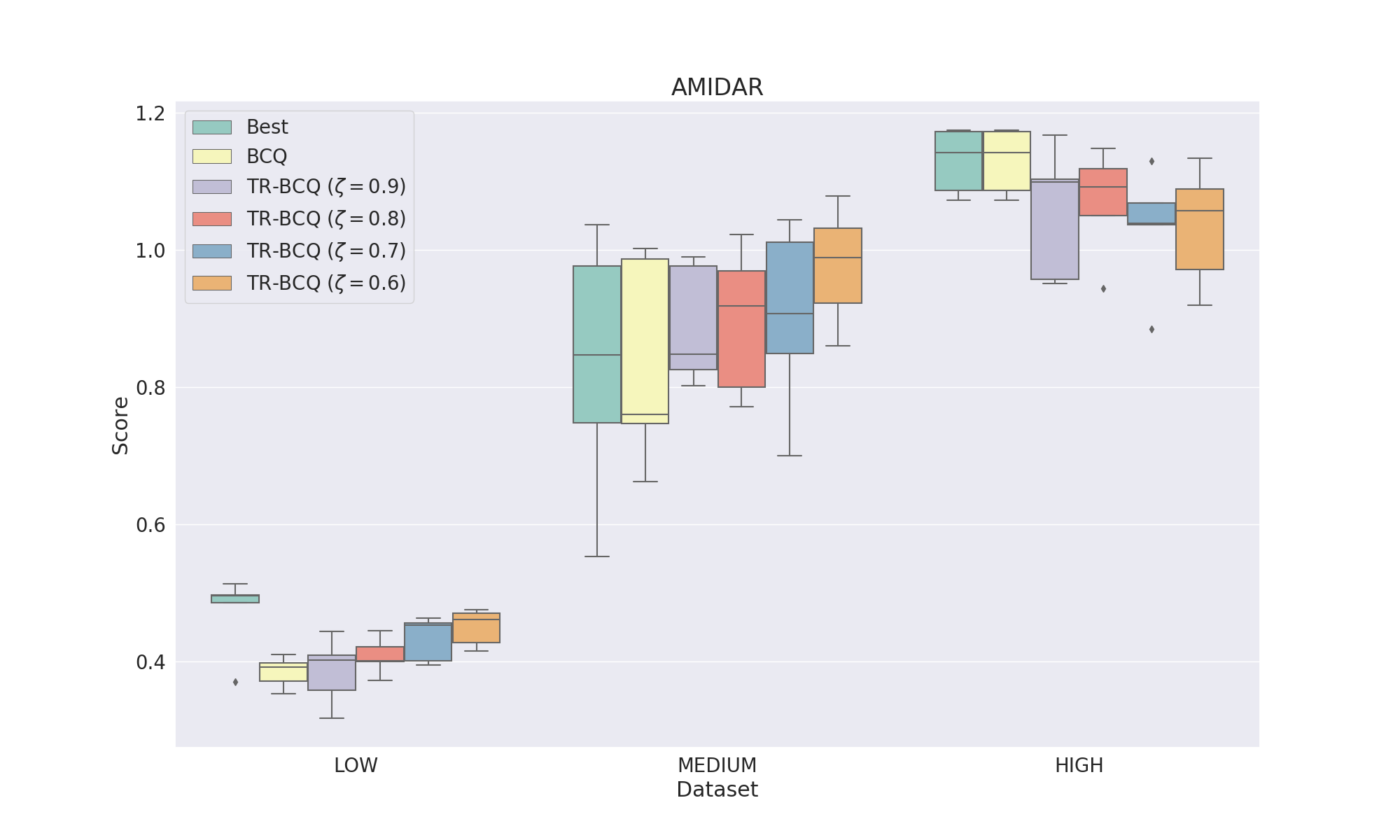}\\
				\vspace{0.01cm}
			\end{minipage}%
		}%

		\subfigure{
			\begin{minipage}[t]{0.333\linewidth}
				\centering
				\includegraphics[width=2.3in]{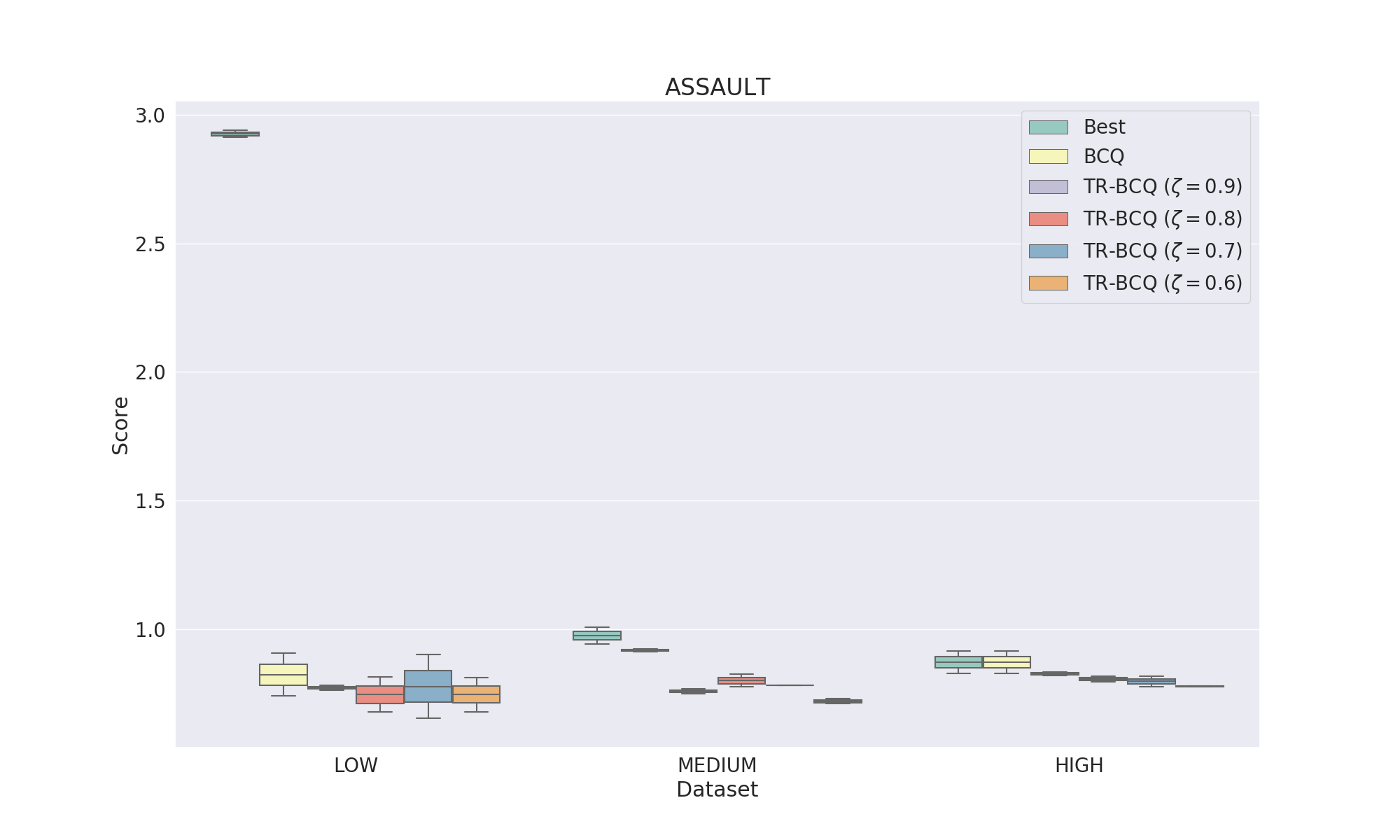}\\
				\vspace{0.01cm}
			\end{minipage}%
		}%
		\subfigure{
			\begin{minipage}[t]{0.333\linewidth}
				\centering
				\includegraphics[width=2.3in]{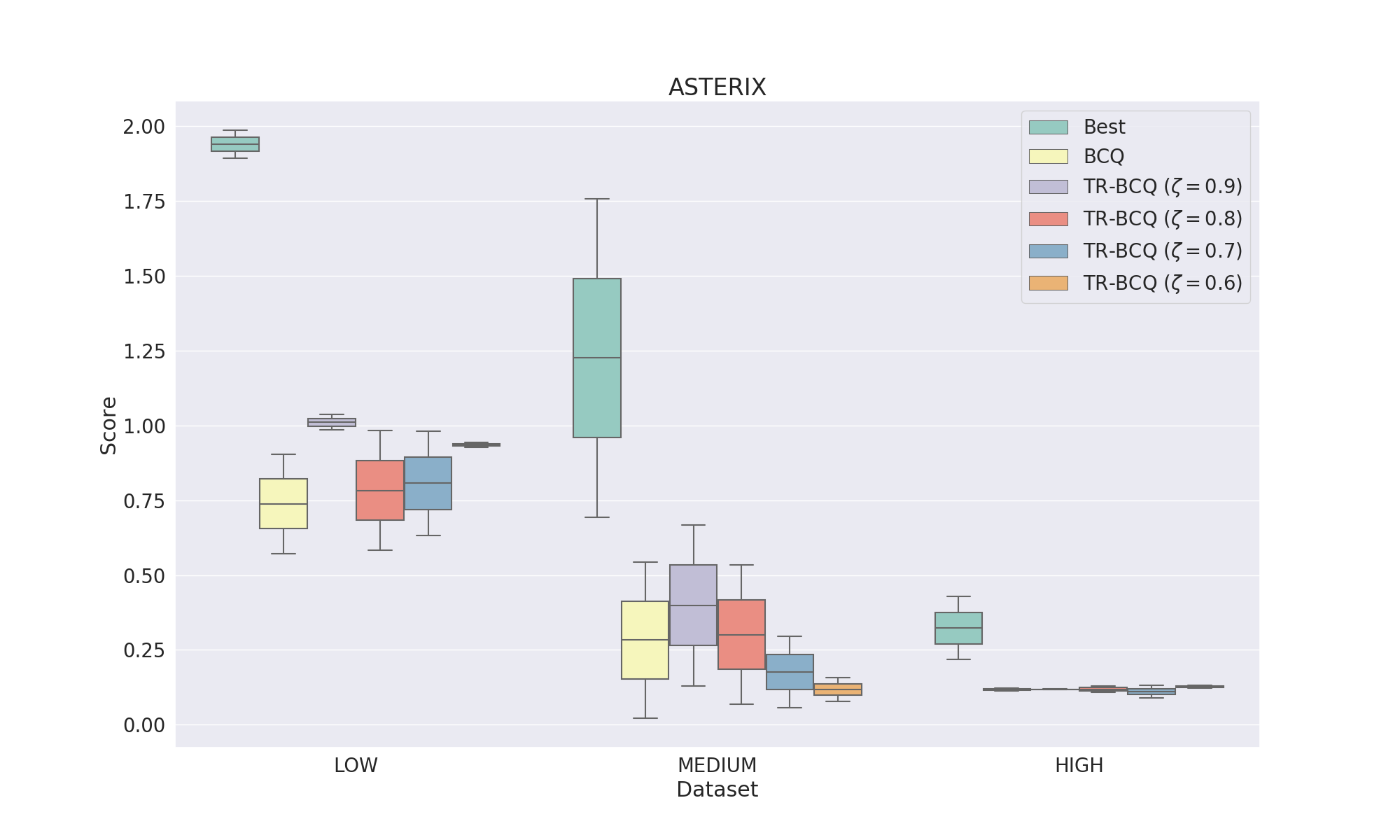}\\
				\vspace{0.01cm}
			\end{minipage}%
		}%
		\subfigure{
			\begin{minipage}[t]{0.333\linewidth}
				\centering
				\includegraphics[width=2.3in]{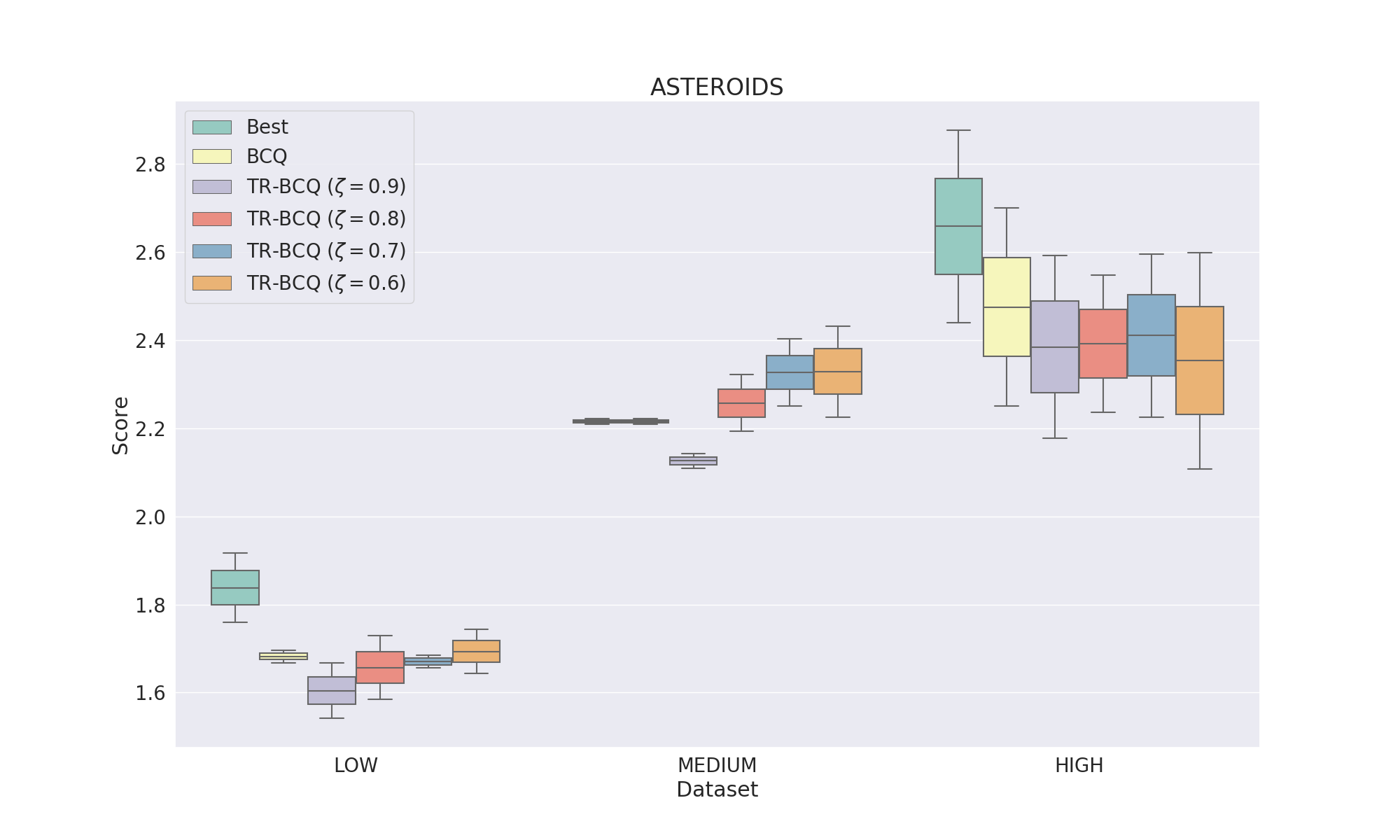}\\
				\vspace{0.01cm}
			\end{minipage}%
		}%
								
		\subfigure{
			\begin{minipage}[t]{0.333\linewidth}
				\centering
				\includegraphics[width=2.3in]{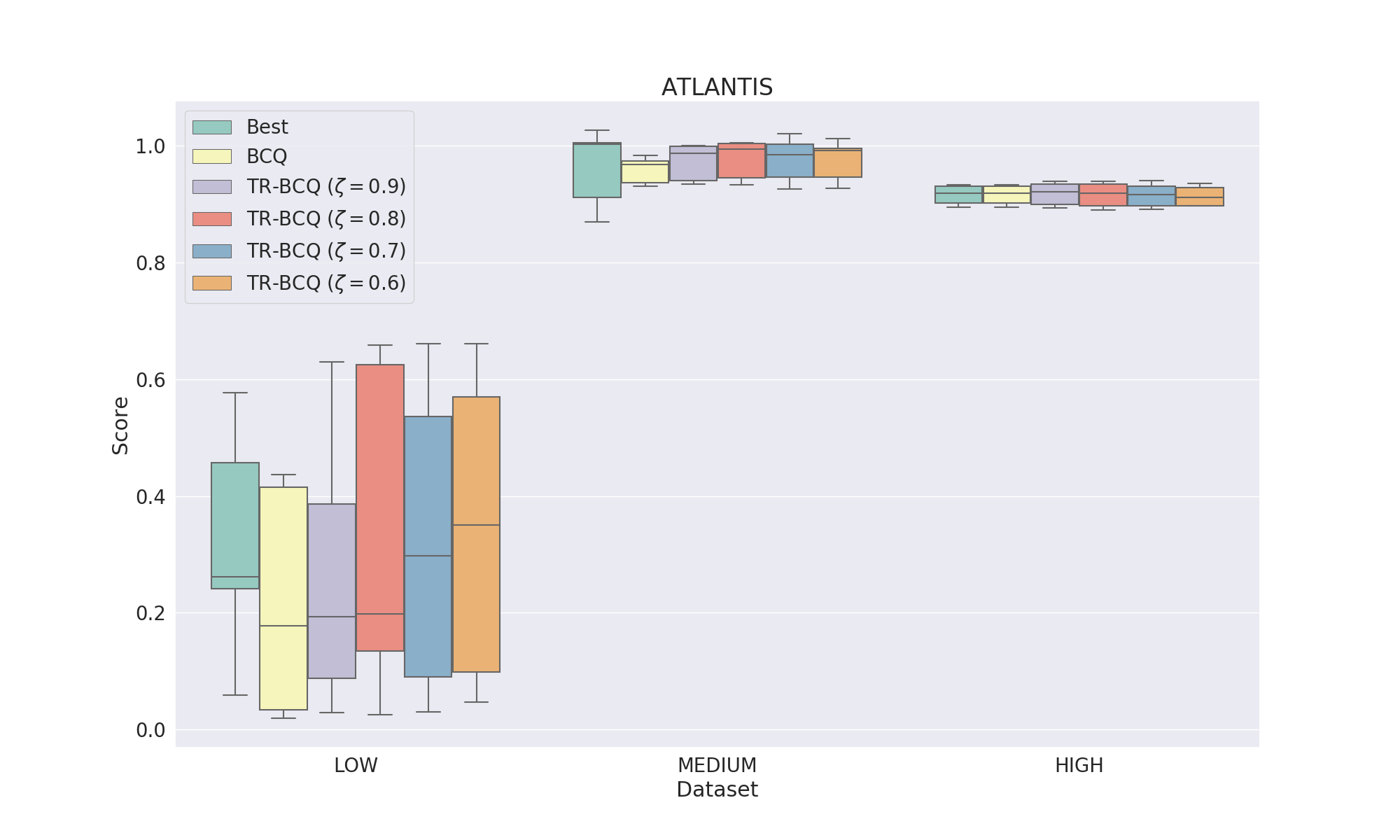}\\
				\vspace{0.01cm}
			\end{minipage}%
		}%
		\subfigure{
			\begin{minipage}[t]{0.333\linewidth}
				\centering
				\includegraphics[width=2.3in]{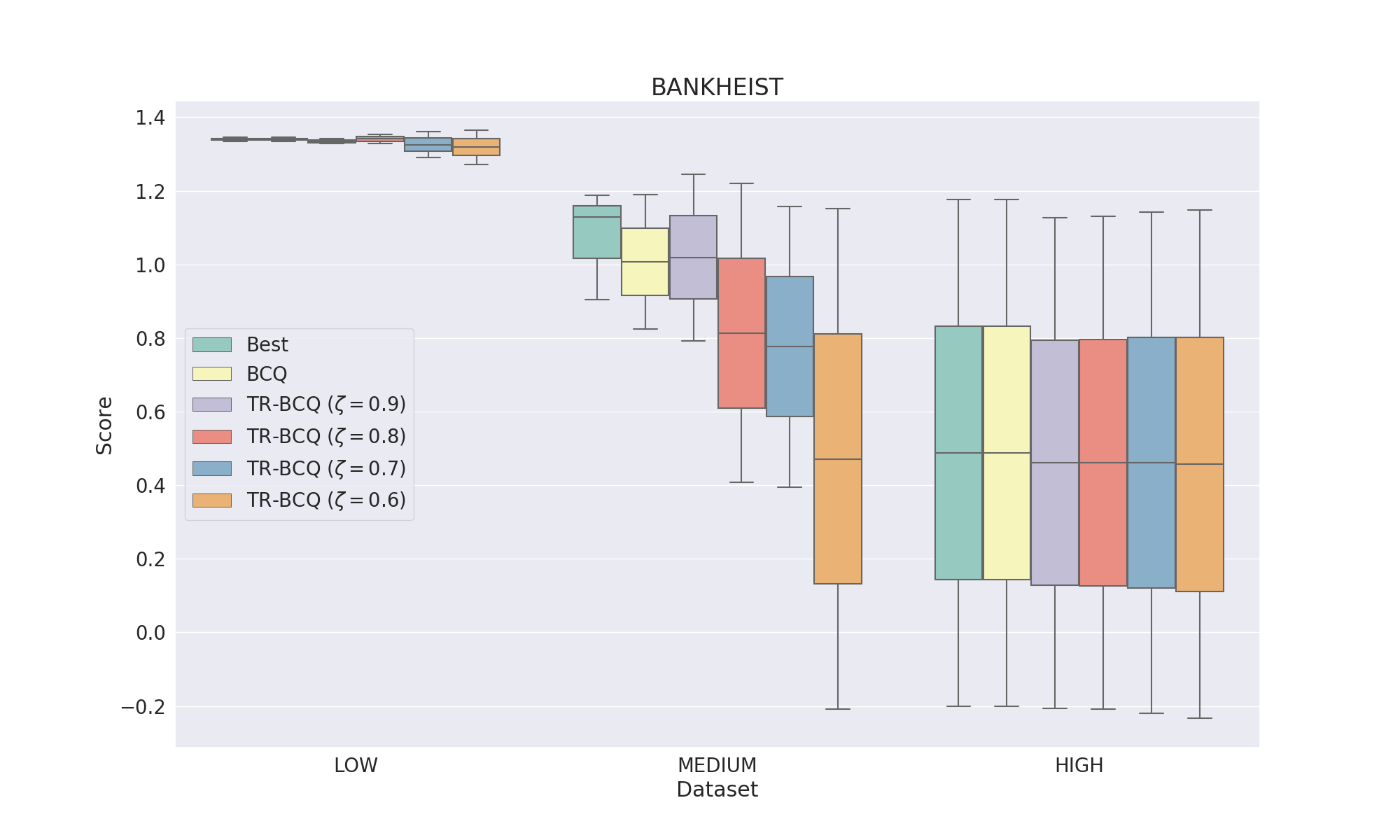}\\
				\vspace{0.01cm}
			\end{minipage}%
		}%
		\subfigure{
			\begin{minipage}[t]{0.333\linewidth}
				\centering
				\includegraphics[width=2.3in]{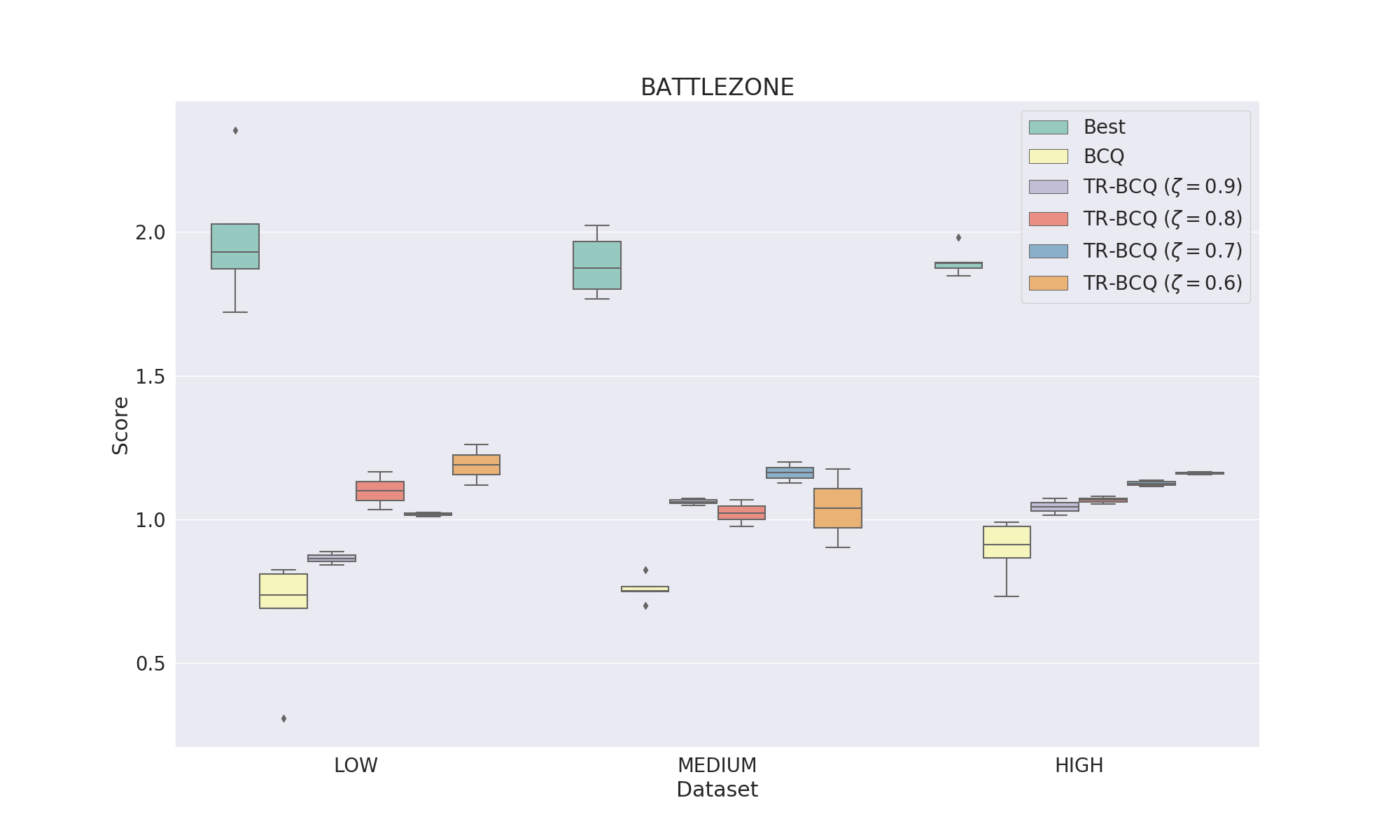}\\
				\vspace{0.01cm}
			\end{minipage}%
		}%

		\subfigure{
			\begin{minipage}[t]{0.333\linewidth}
				\centering
				\includegraphics[width=2.3in]{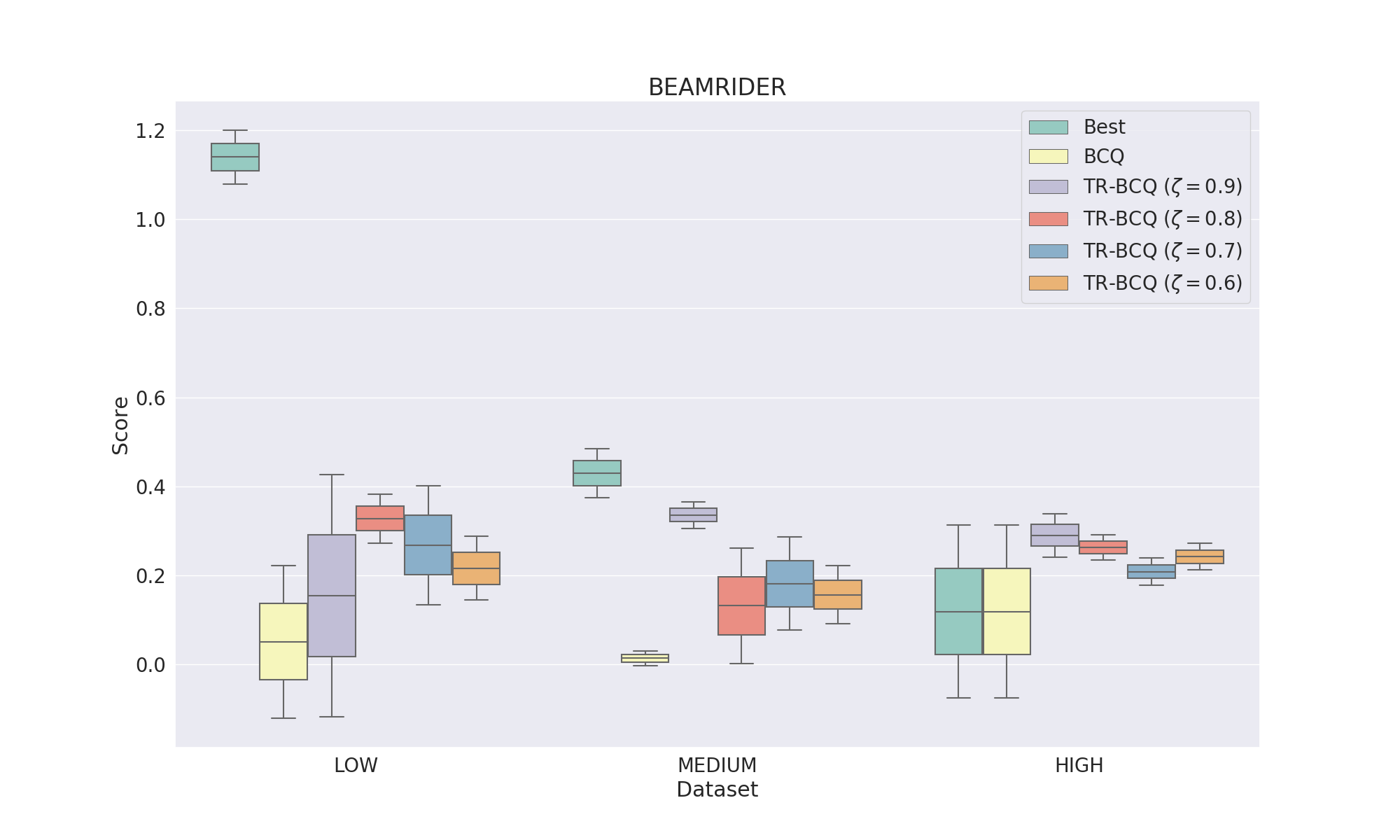}\\
				\vspace{0.01cm}
			\end{minipage}%
		}%
		\subfigure{
			\begin{minipage}[t]{0.333\linewidth}
				\centering
				\includegraphics[width=2.3in]{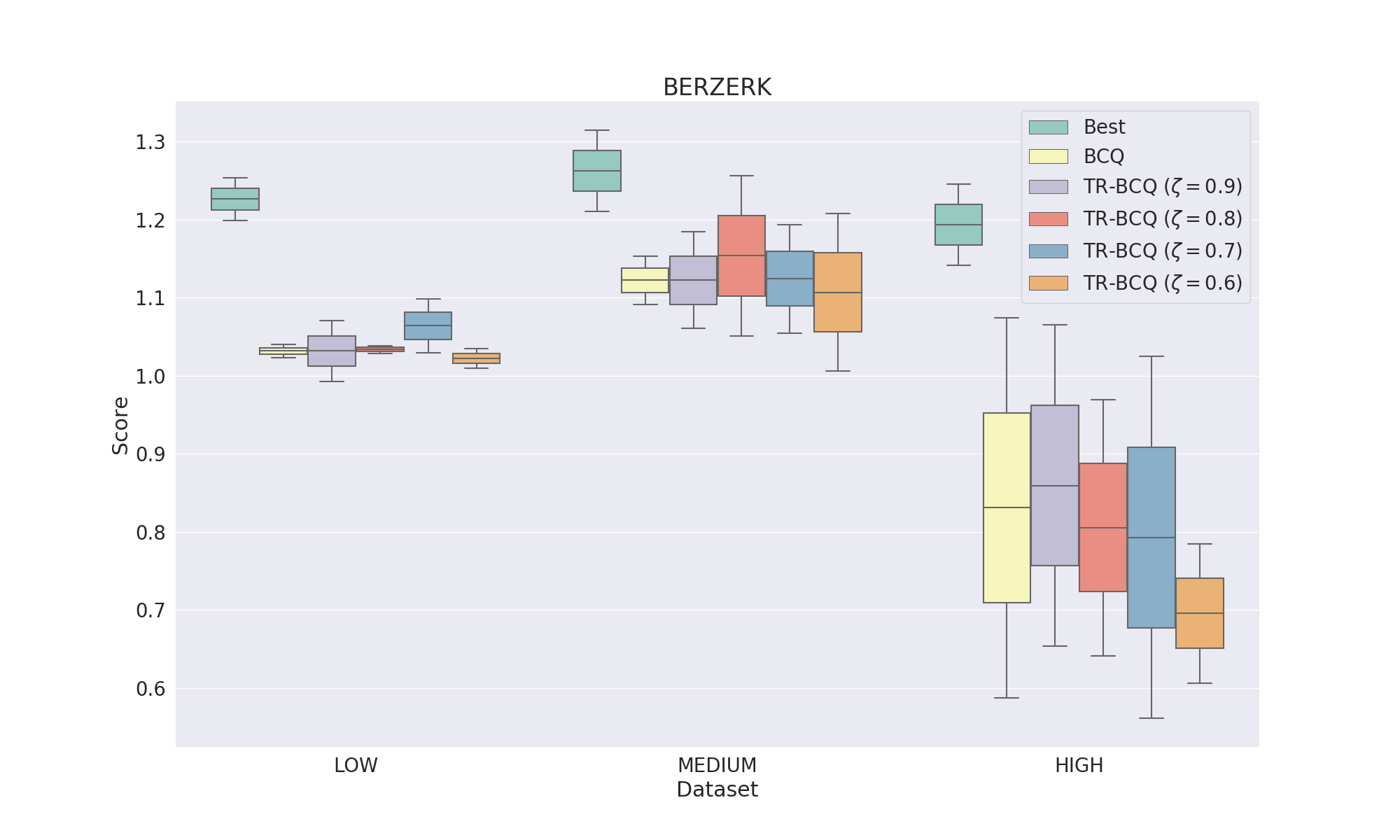}\\
				\vspace{0.01cm}
			\end{minipage}%
		}%
		\subfigure{
			\begin{minipage}[t]{0.333\linewidth}
				\centering
				\includegraphics[width=2.3in]{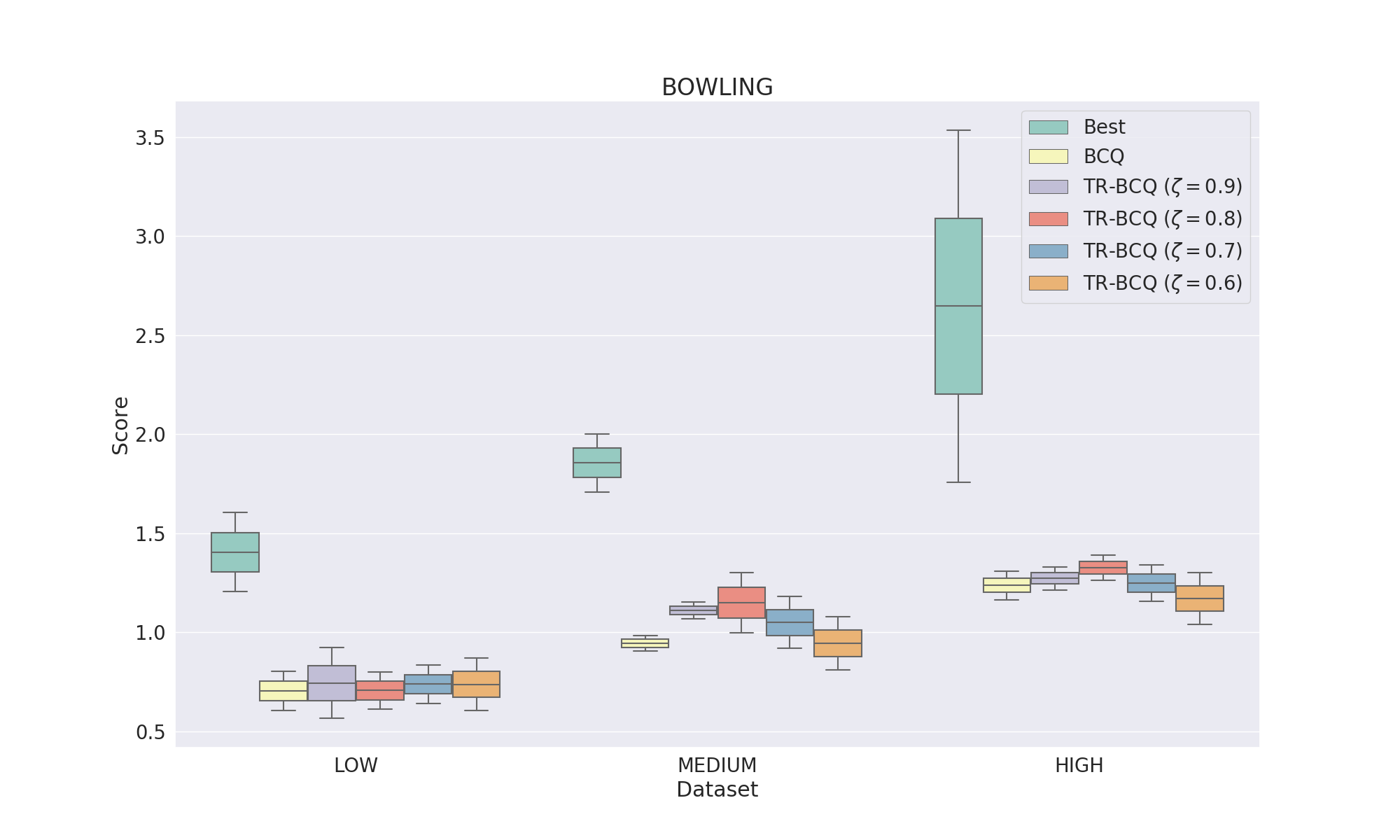}\\
				\vspace{0.01cm}
			\end{minipage}%
		}%

		\subfigure{
			\begin{minipage}[t]{0.333\linewidth}
				\centering
				\includegraphics[width=2.3in]{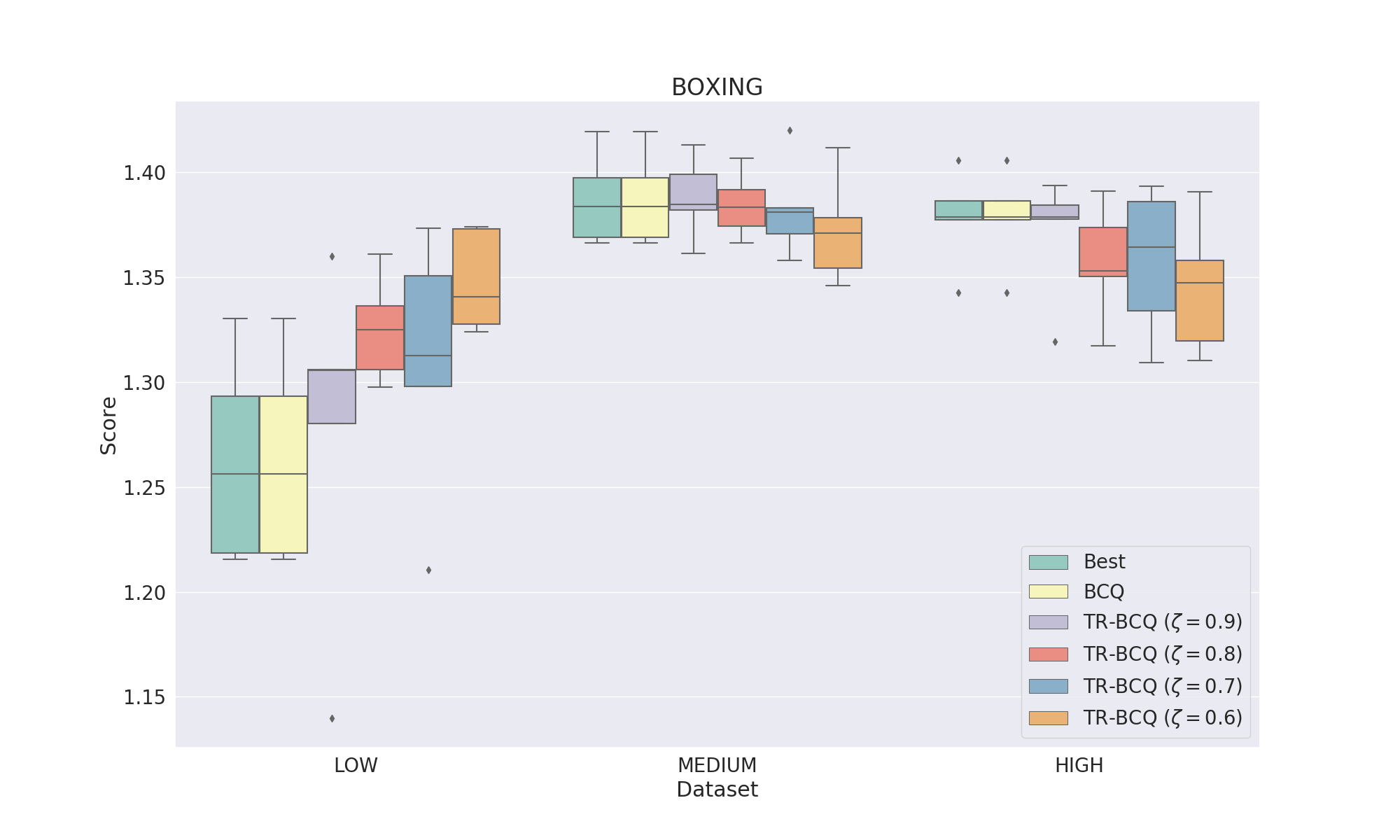}\\
				\vspace{0.01cm}
			\end{minipage}%
		}%
		\subfigure{
			\begin{minipage}[t]{0.333\linewidth}
				\centering
				\includegraphics[width=2.3in]{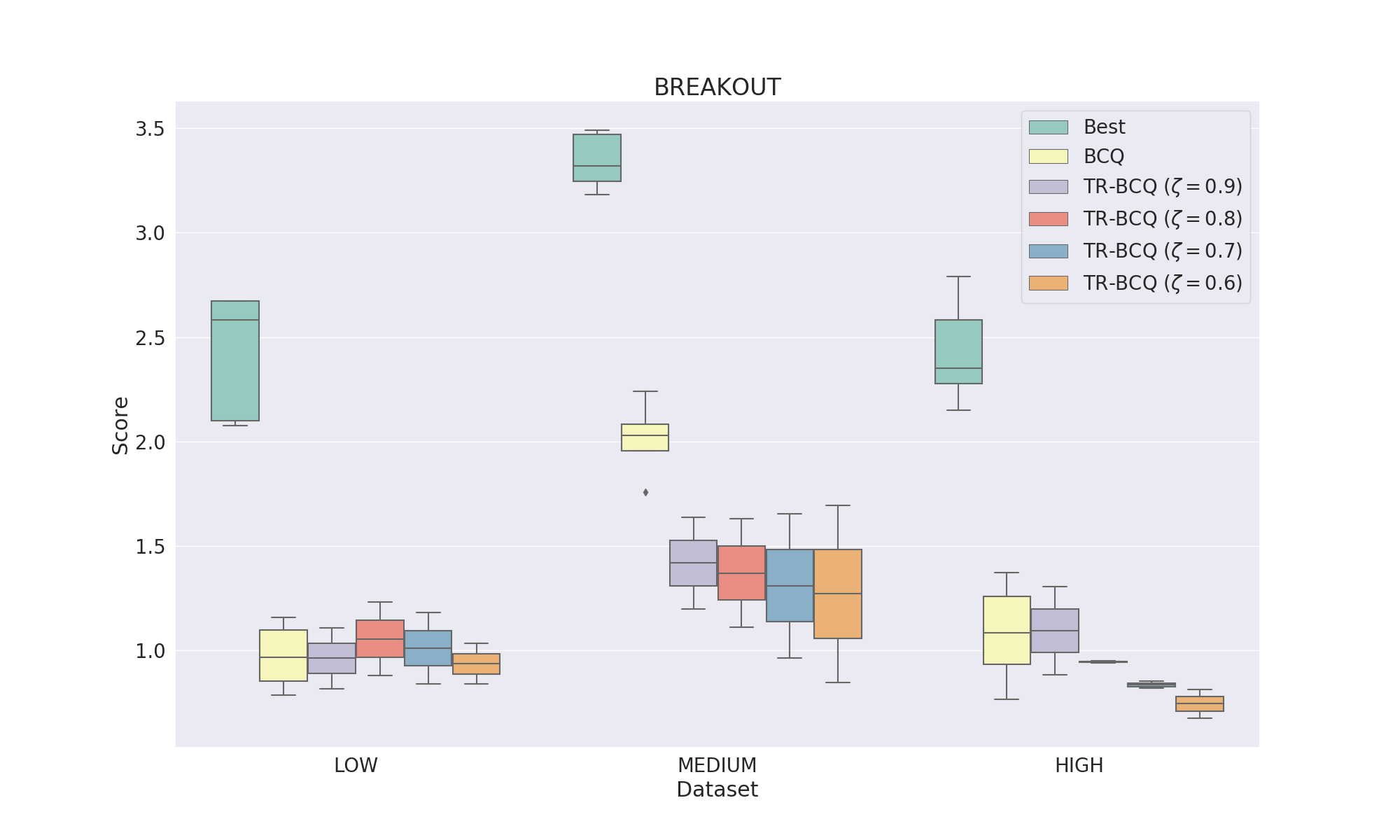}\\
				\vspace{0.01cm}
			\end{minipage}%
		}%
		\subfigure{
			\begin{minipage}[t]{0.333\linewidth}
				\centering
				\includegraphics[width=2.3in]{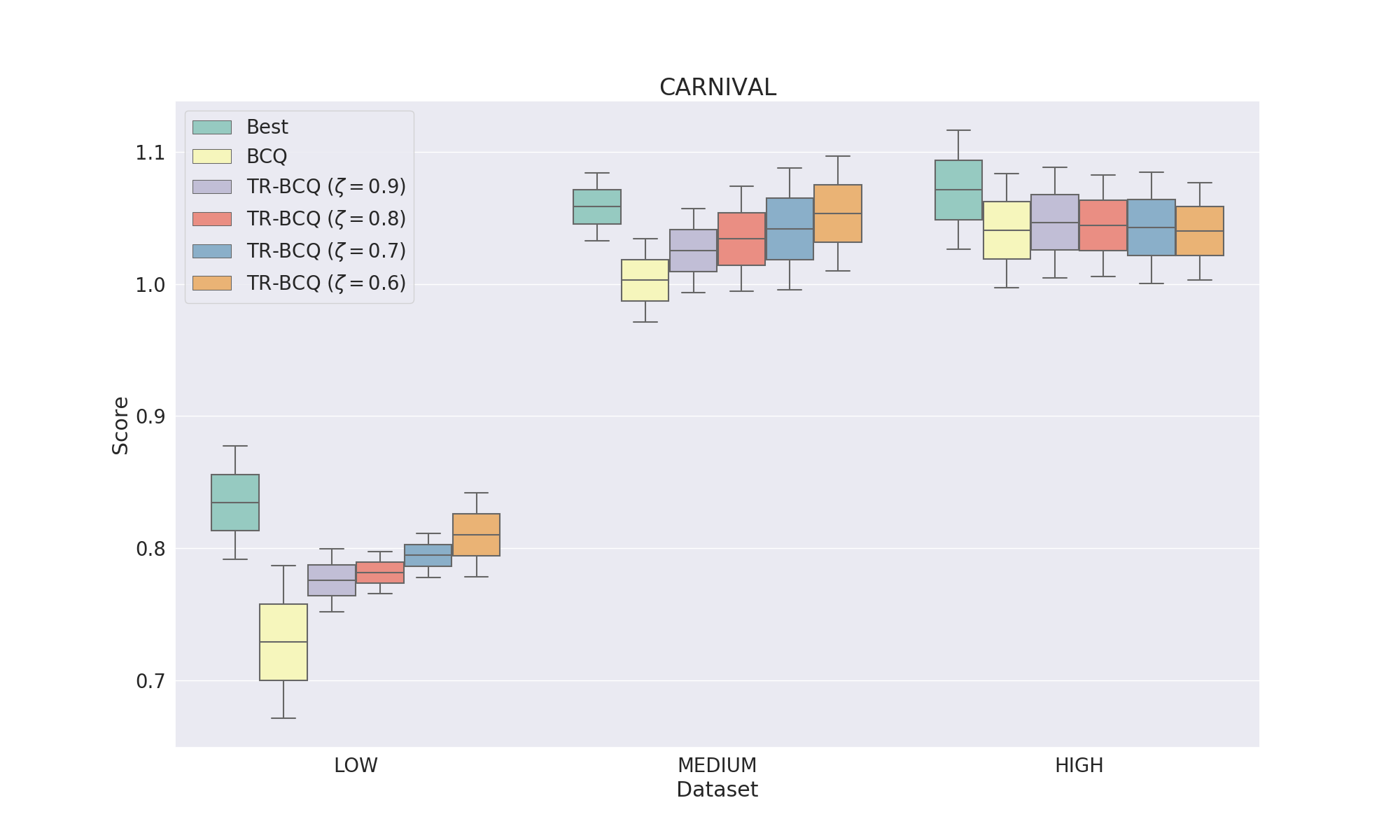}\\
				\vspace{0.01cm}
			\end{minipage}%
		}%

		\subfigure{
			\begin{minipage}[t]{0.333\linewidth}
				\centering
				\includegraphics[width=2.3in]{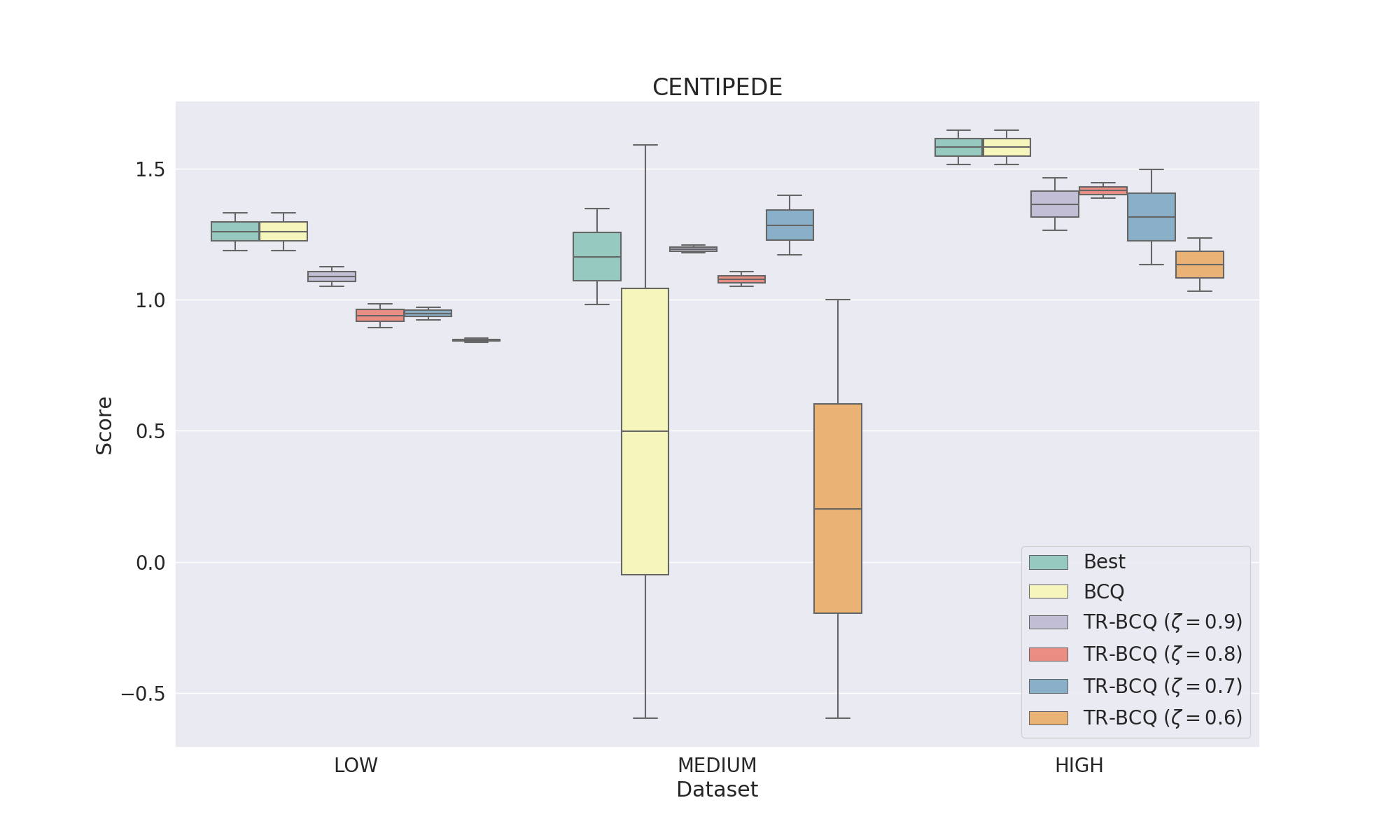}\\
				\vspace{0.01cm}
			\end{minipage}%
		}%
		\subfigure{
			\begin{minipage}[t]{0.333\linewidth}
				\centering
				\includegraphics[width=2.3in]{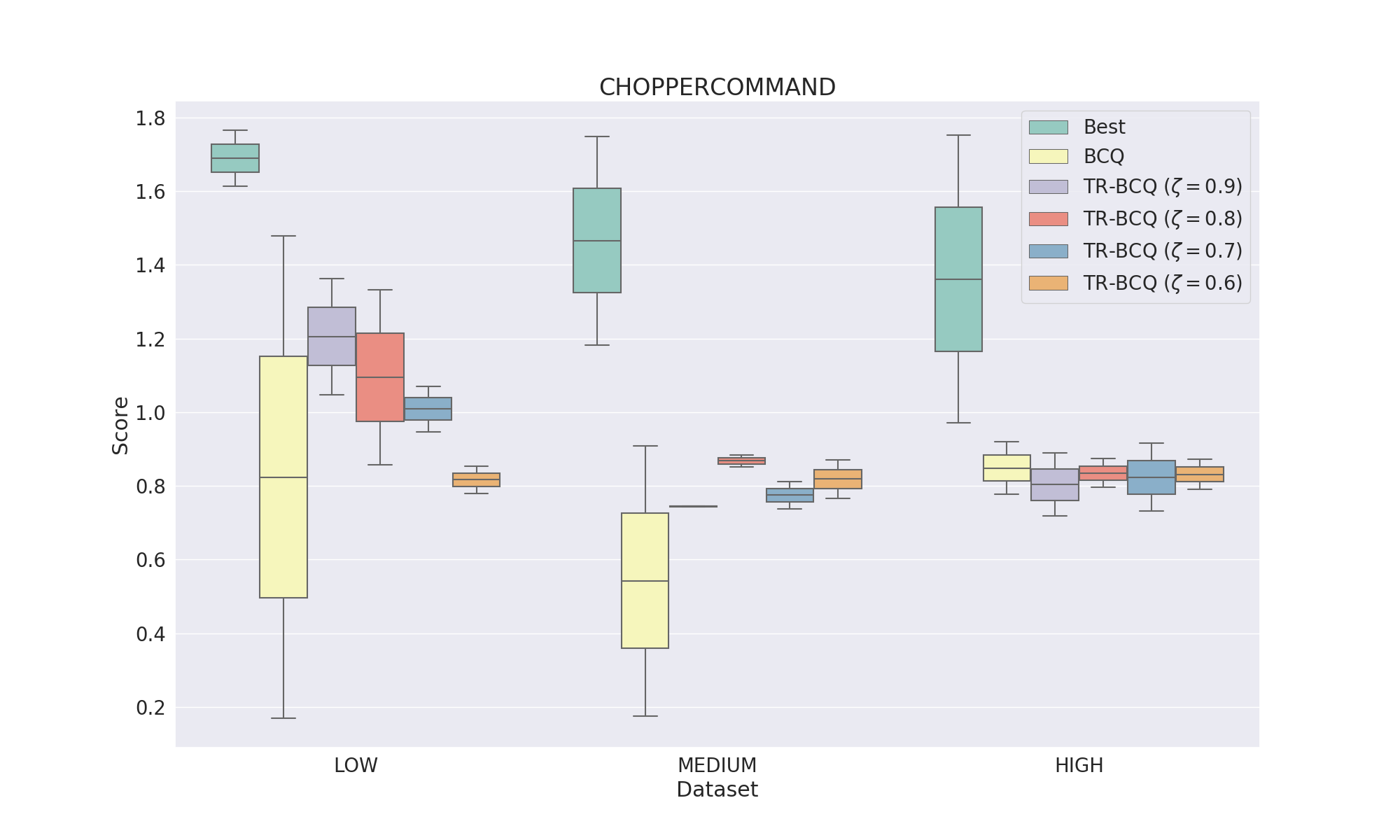}\\
				\vspace{0.01cm}
			\end{minipage}%
		}%
		\subfigure{
			\begin{minipage}[t]{0.333\linewidth}
				\centering
				\includegraphics[width=2.3in]{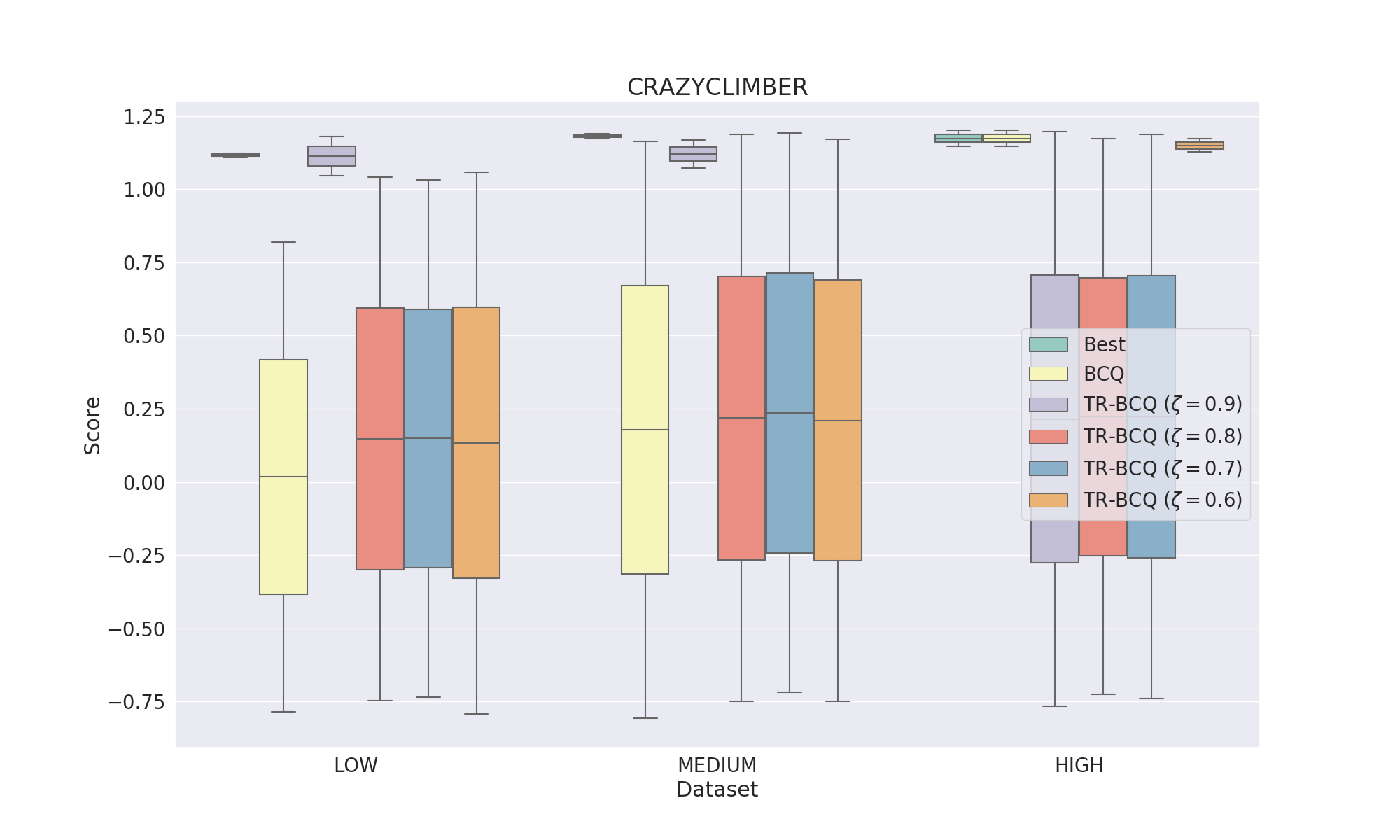}\\
				\vspace{0.01cm}
			\end{minipage}%
		}%

		\centering
		\caption{\textbf{Comparison between  TR-BCQ and the best baselines on different datasets from Game Alien to Game CrazyClimber}}
		\label{fig: Comparison between  TR-BCQ and best baselines on different datasets from Game Alien to Game CrazyClimber}
								
	\end{figure*}

	\begin{figure*}[!htb]
		\centering

		\subfigure{
			\begin{minipage}[t]{0.333\linewidth}
				\centering
				\includegraphics[width=2.3in]{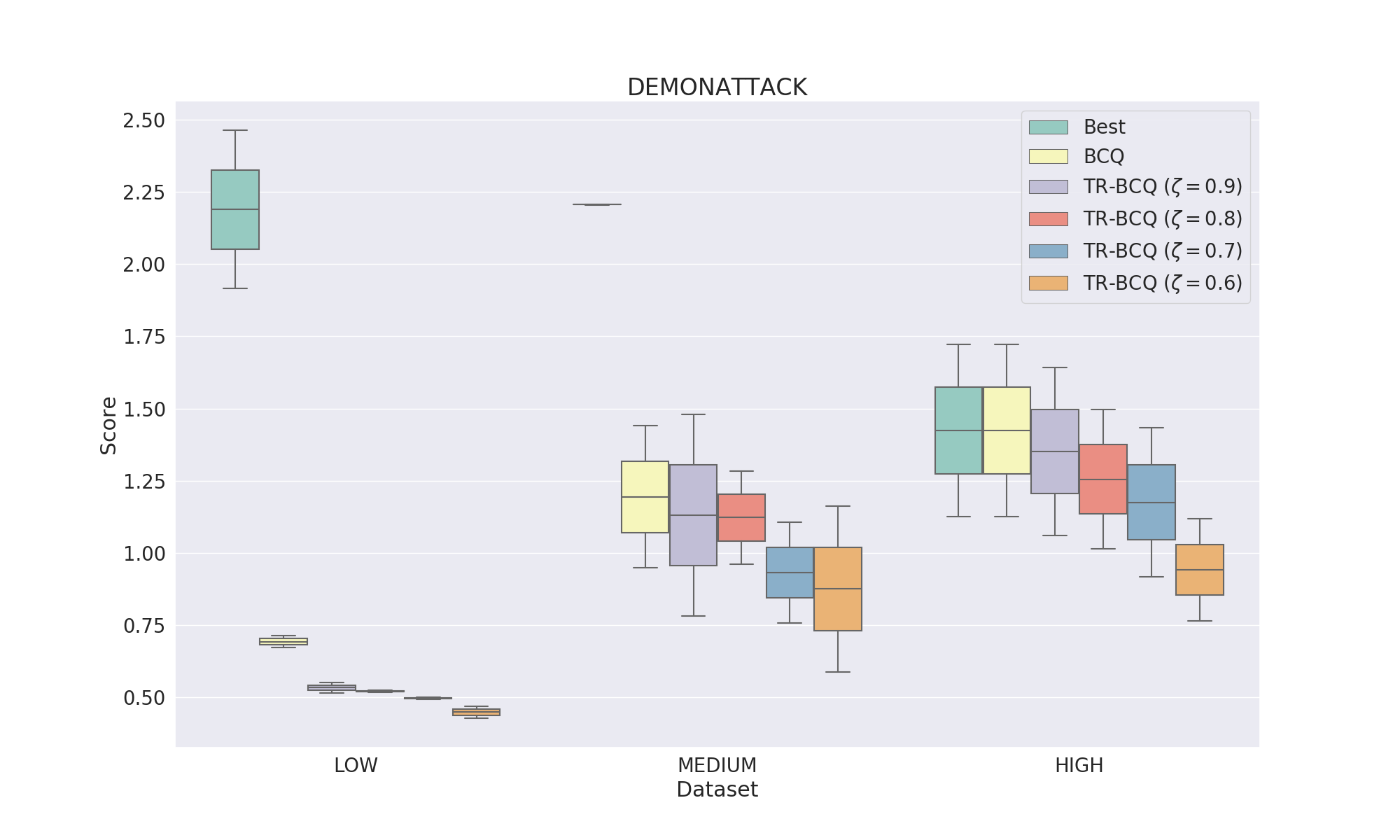}\\
				\vspace{0.01cm}
			\end{minipage}%
		}%
		\subfigure{
			\begin{minipage}[t]{0.333\linewidth}
				\centering
				\includegraphics[width=2.3in]{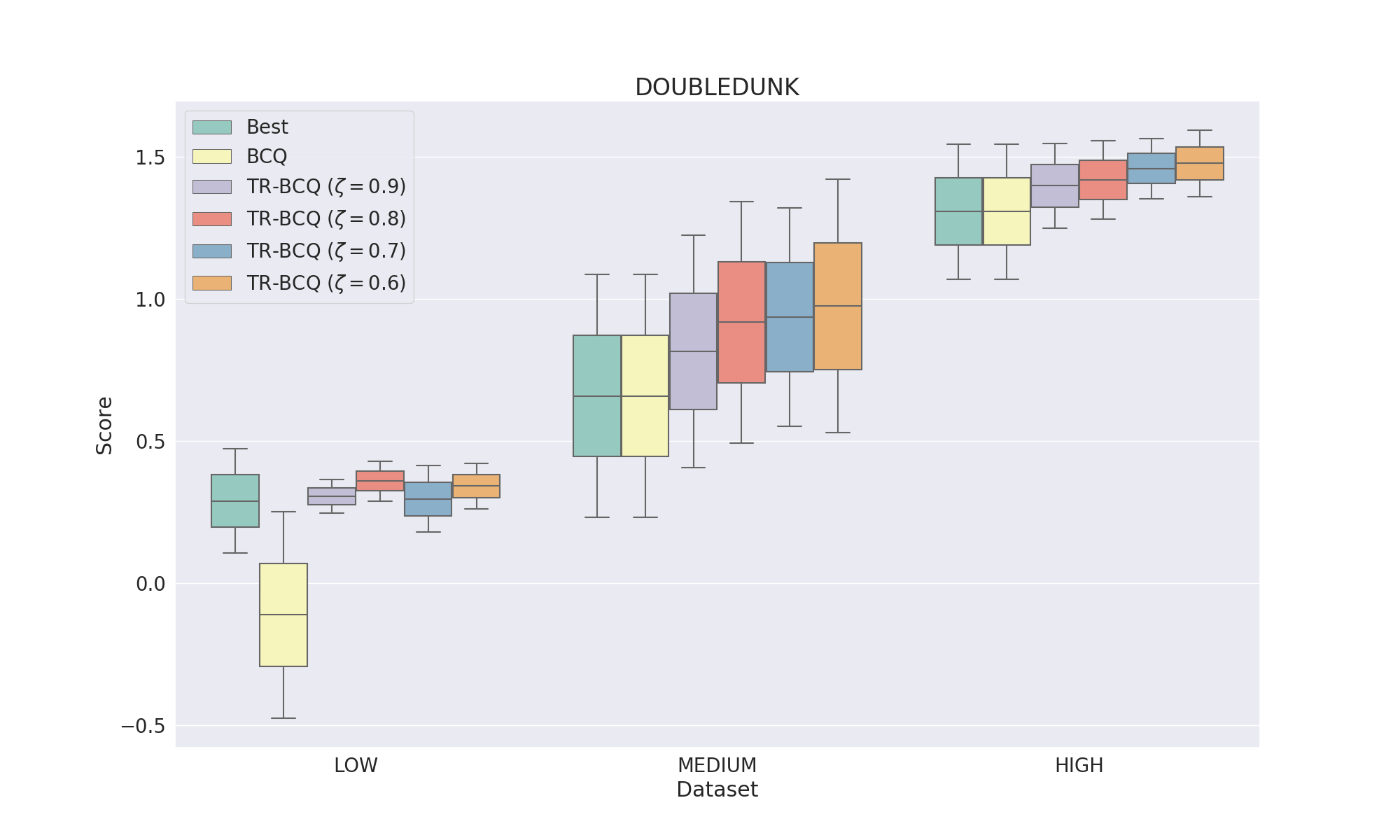}\\
				\vspace{0.01cm}
			\end{minipage}%
		}%
		\subfigure{
			\begin{minipage}[t]{0.333\linewidth}
				\centering
				\includegraphics[width=2.3in]{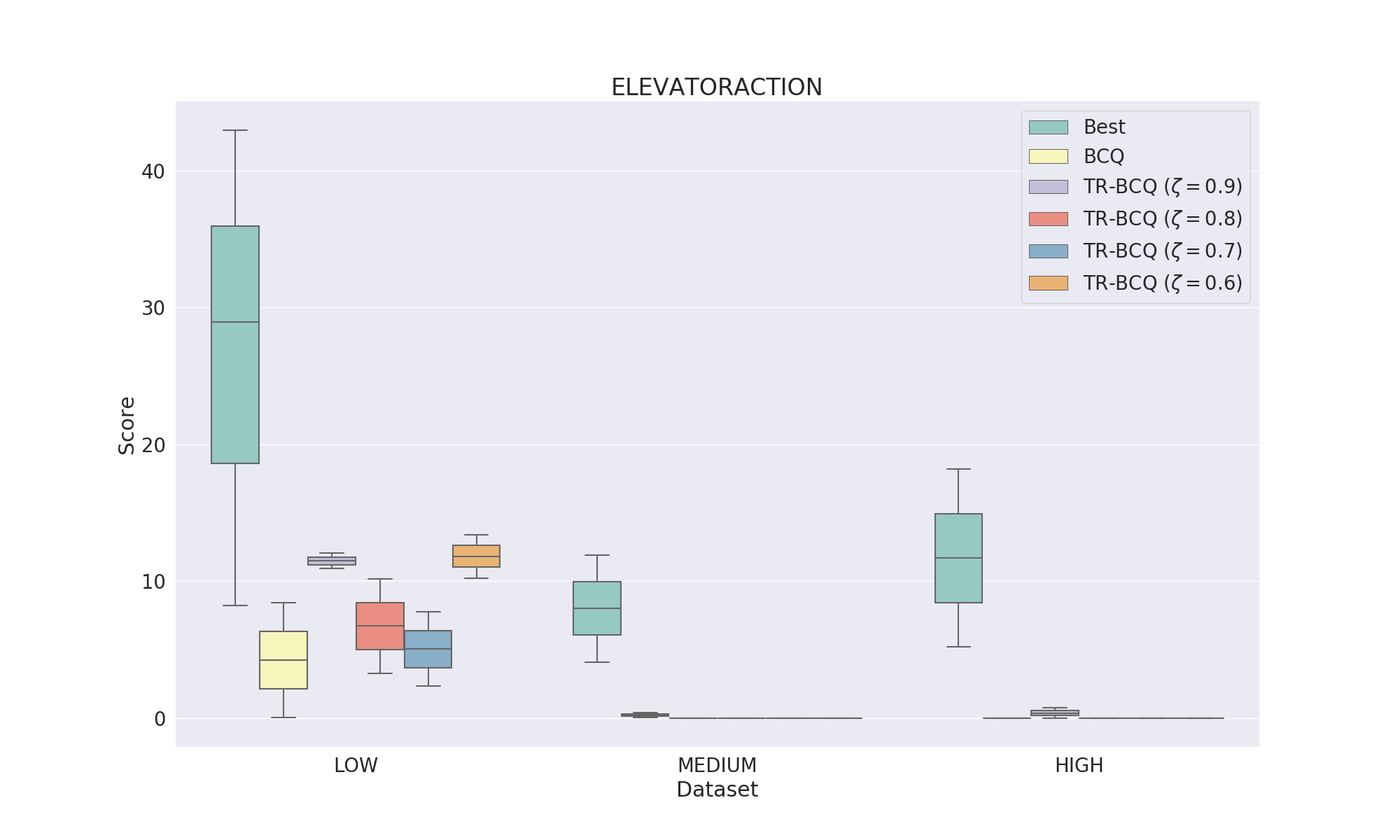}\\
				\vspace{0.01cm}
			\end{minipage}%
		}%

		\subfigure{
			\begin{minipage}[t]{0.333\linewidth}
				\centering
				\includegraphics[width=2.3in]{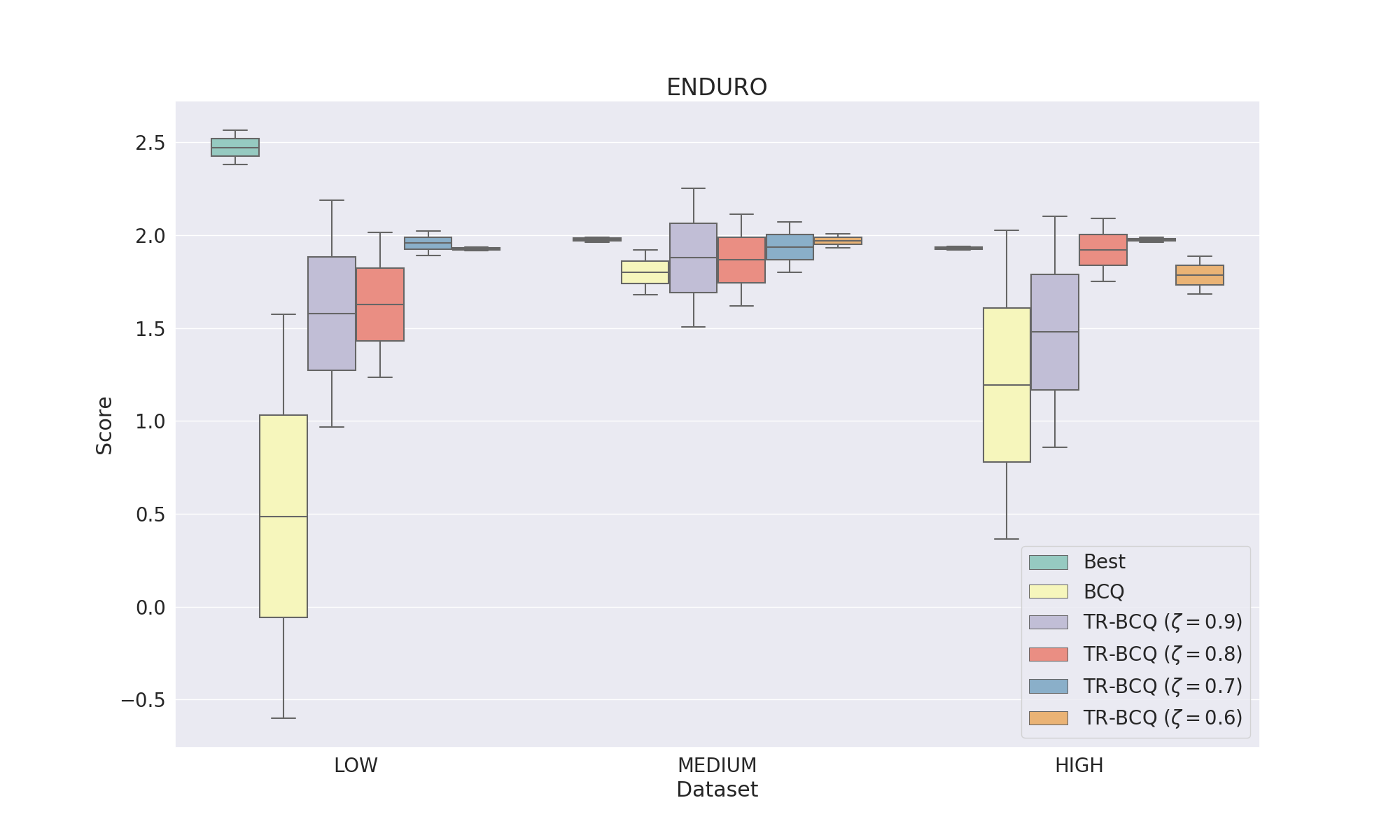}\\
				\vspace{0.01cm}
			\end{minipage}%
		}%
		\subfigure{
			\begin{minipage}[t]{0.333\linewidth}
				\centering
				\includegraphics[width=2.3in]{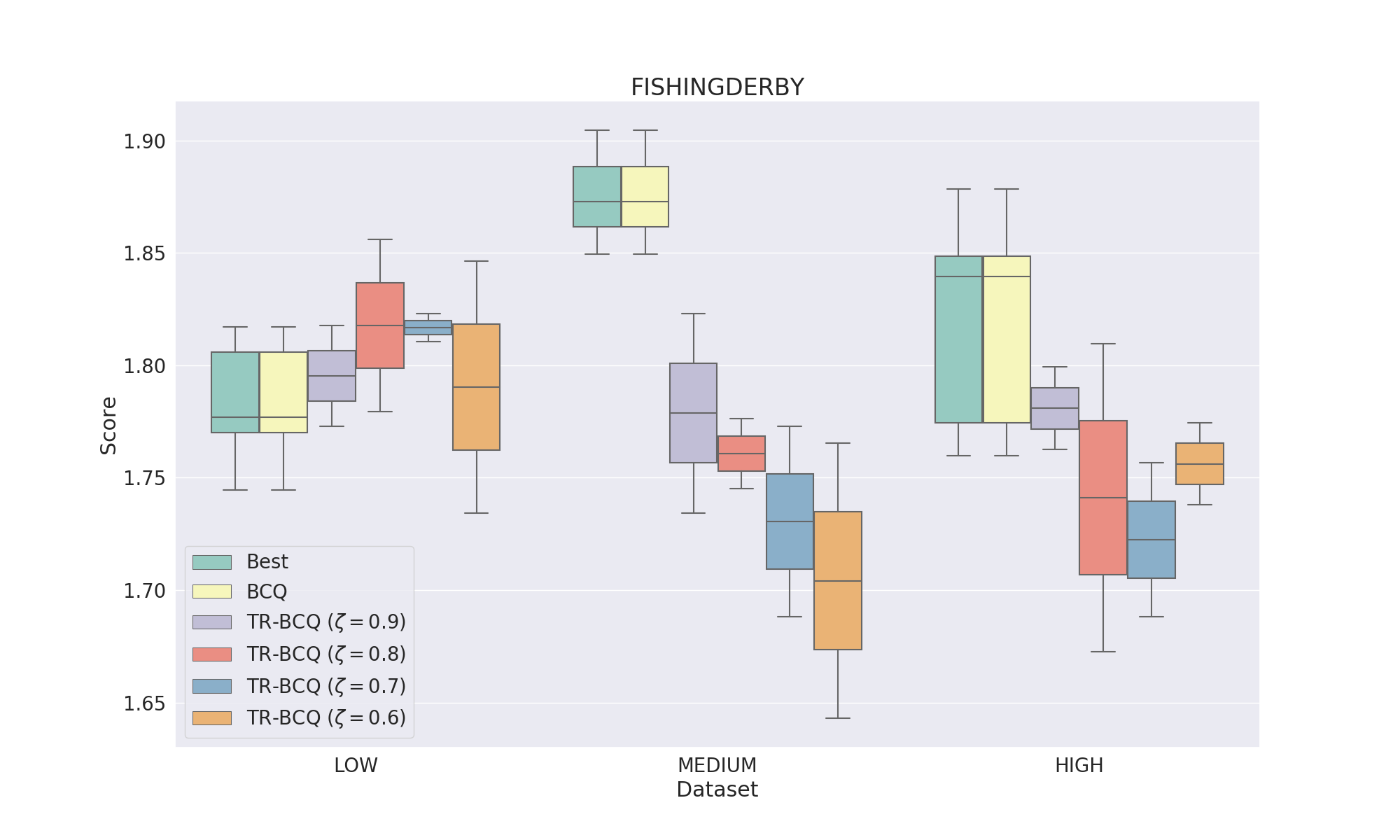}\\
				\vspace{0.01cm}
			\end{minipage}%
		}%
		\subfigure{
			\begin{minipage}[t]{0.333\linewidth}
				\centering
				\includegraphics[width=2.3in]{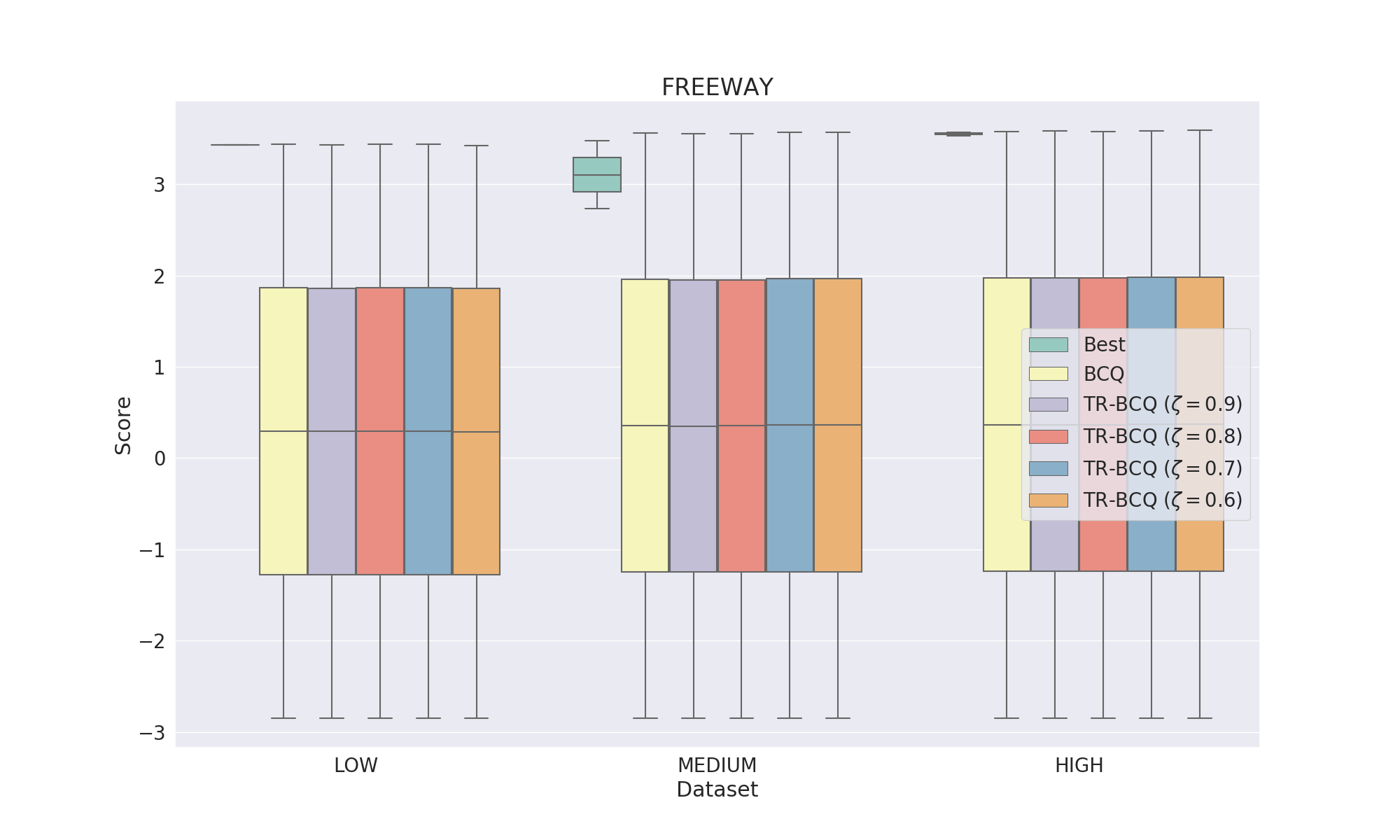}\\
				\vspace{0.01cm}
			\end{minipage}%
		}%

		\subfigure{
			\begin{minipage}[t]{0.333\linewidth}
				\centering
				\includegraphics[width=2.3in]{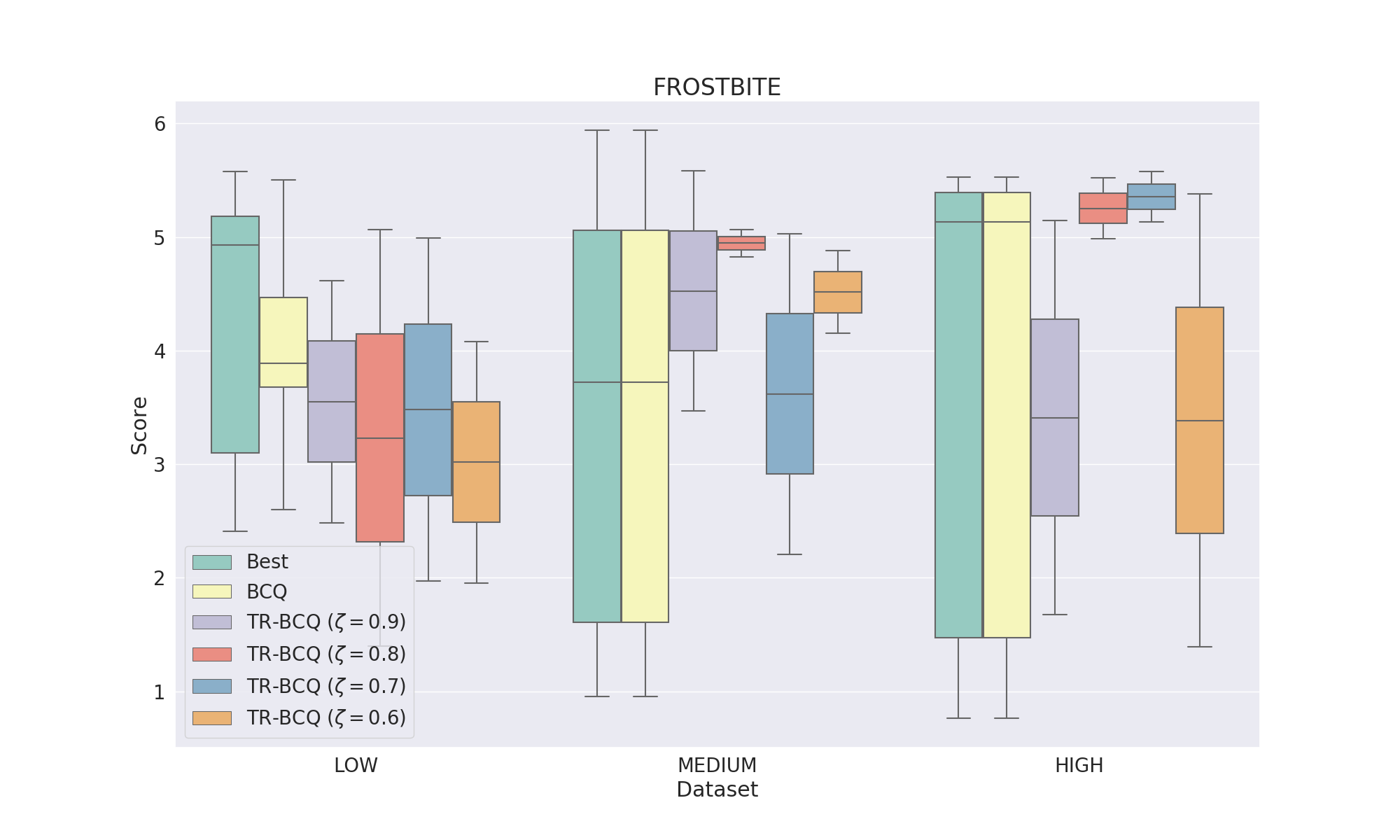}\\
				\vspace{0.01cm}
			\end{minipage}%
		}%
		\subfigure{
			\begin{minipage}[t]{0.333\linewidth}
				\centering
				\includegraphics[width=2.3in]{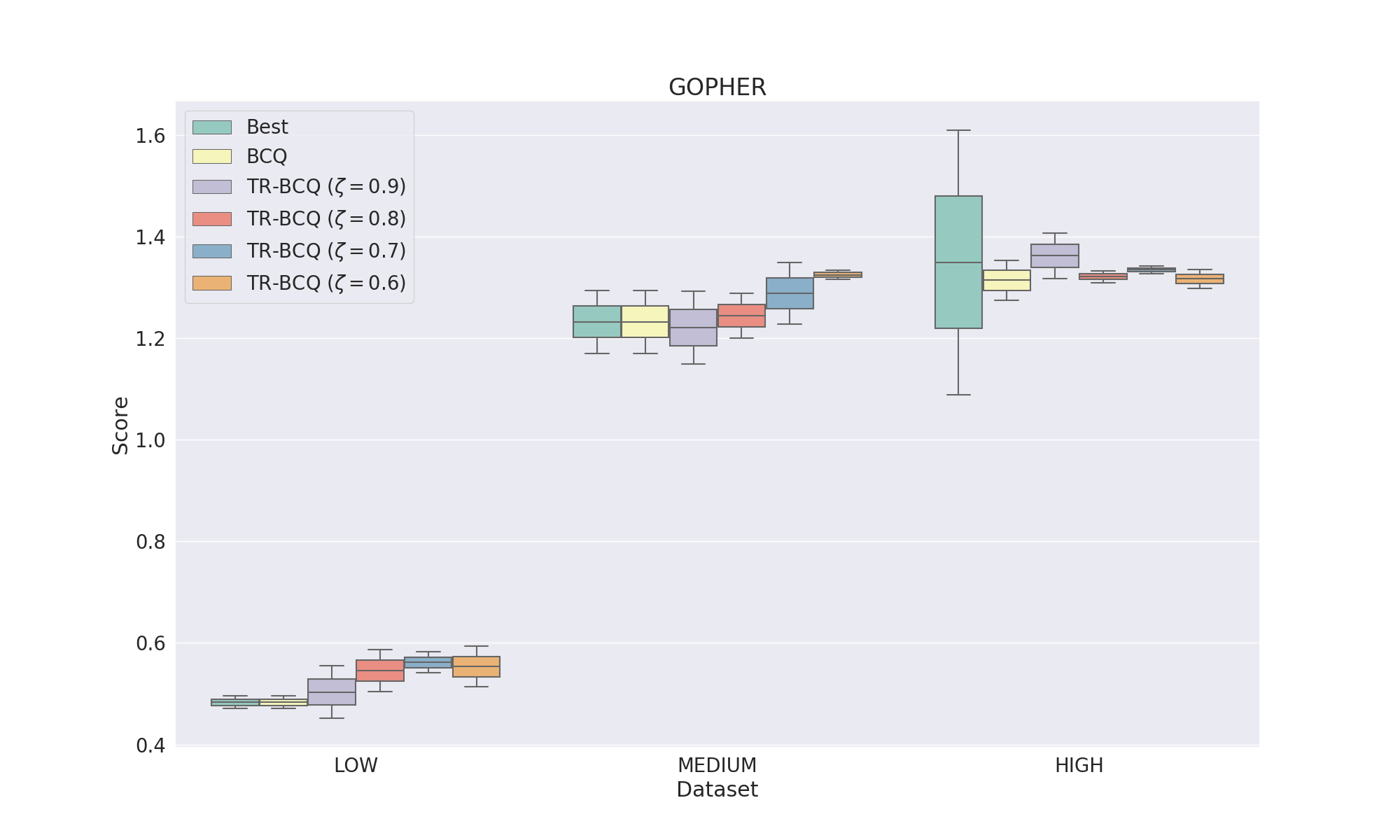}\\
				\vspace{0.01cm}
			\end{minipage}%
		}%
		\subfigure{
			\begin{minipage}[t]{0.333\linewidth}
				\centering
				\includegraphics[width=2.3in]{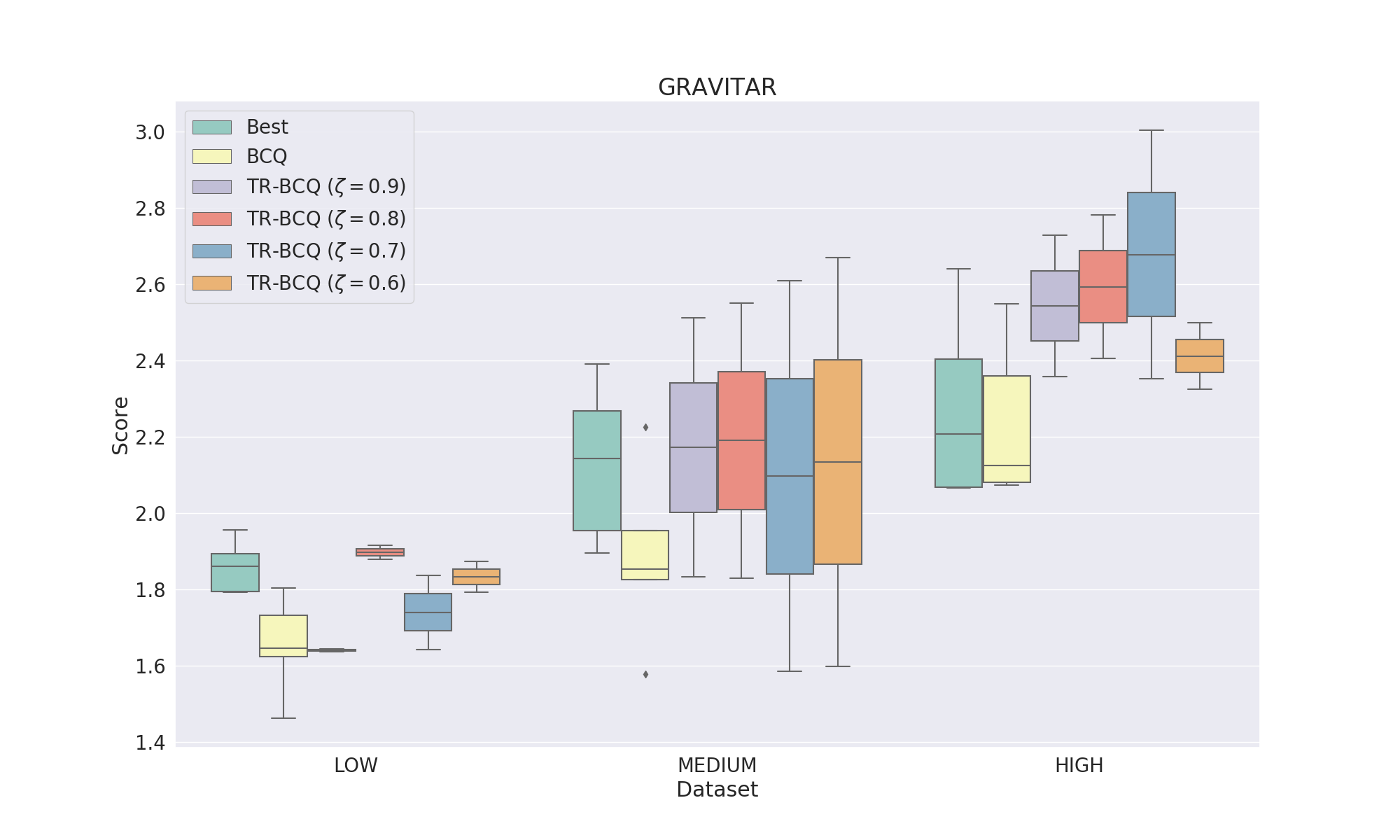}\\
				\vspace{0.01cm}
			\end{minipage}%
		}%

		\subfigure{
			\begin{minipage}[t]{0.333\linewidth}
				\centering
				\includegraphics[width=2.3in]{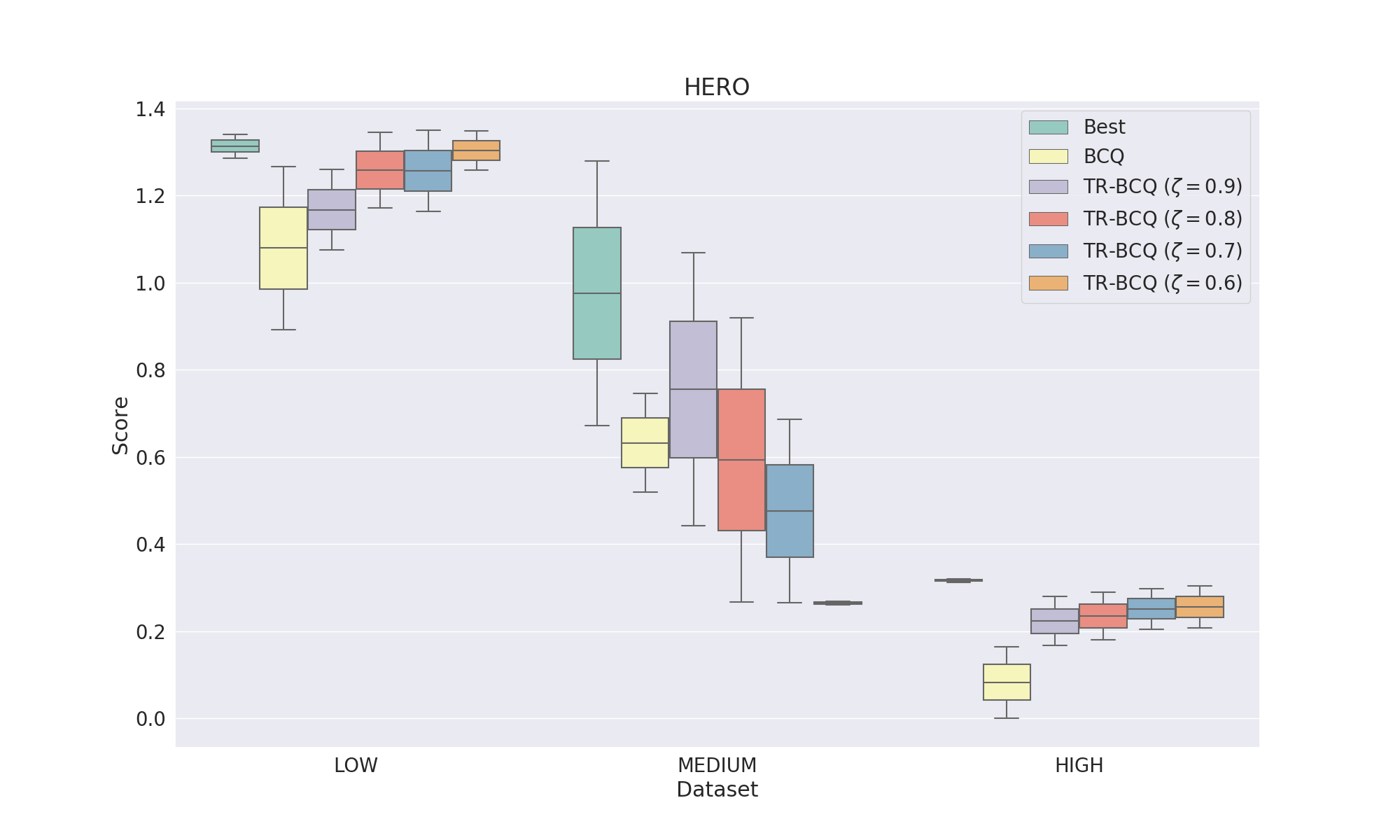}\\
				\vspace{0.01cm}
			\end{minipage}%
		}%
		\subfigure{
			\begin{minipage}[t]{0.333\linewidth}
				\centering
				\includegraphics[width=2.3in]{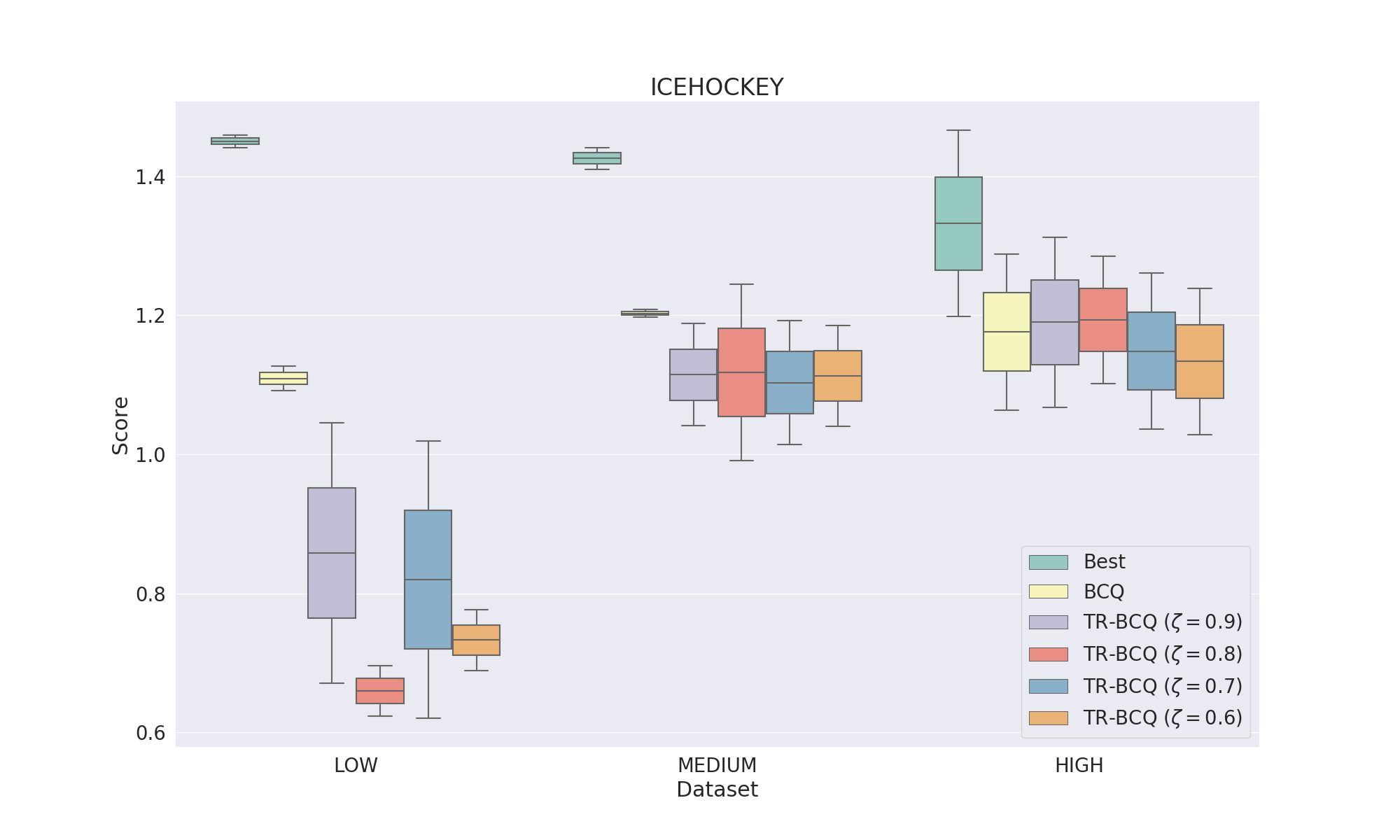}\\
				\vspace{0.01cm}
			\end{minipage}%
		}%
		\subfigure{
			\begin{minipage}[t]{0.333\linewidth}
				\centering
				\includegraphics[width=2.3in]{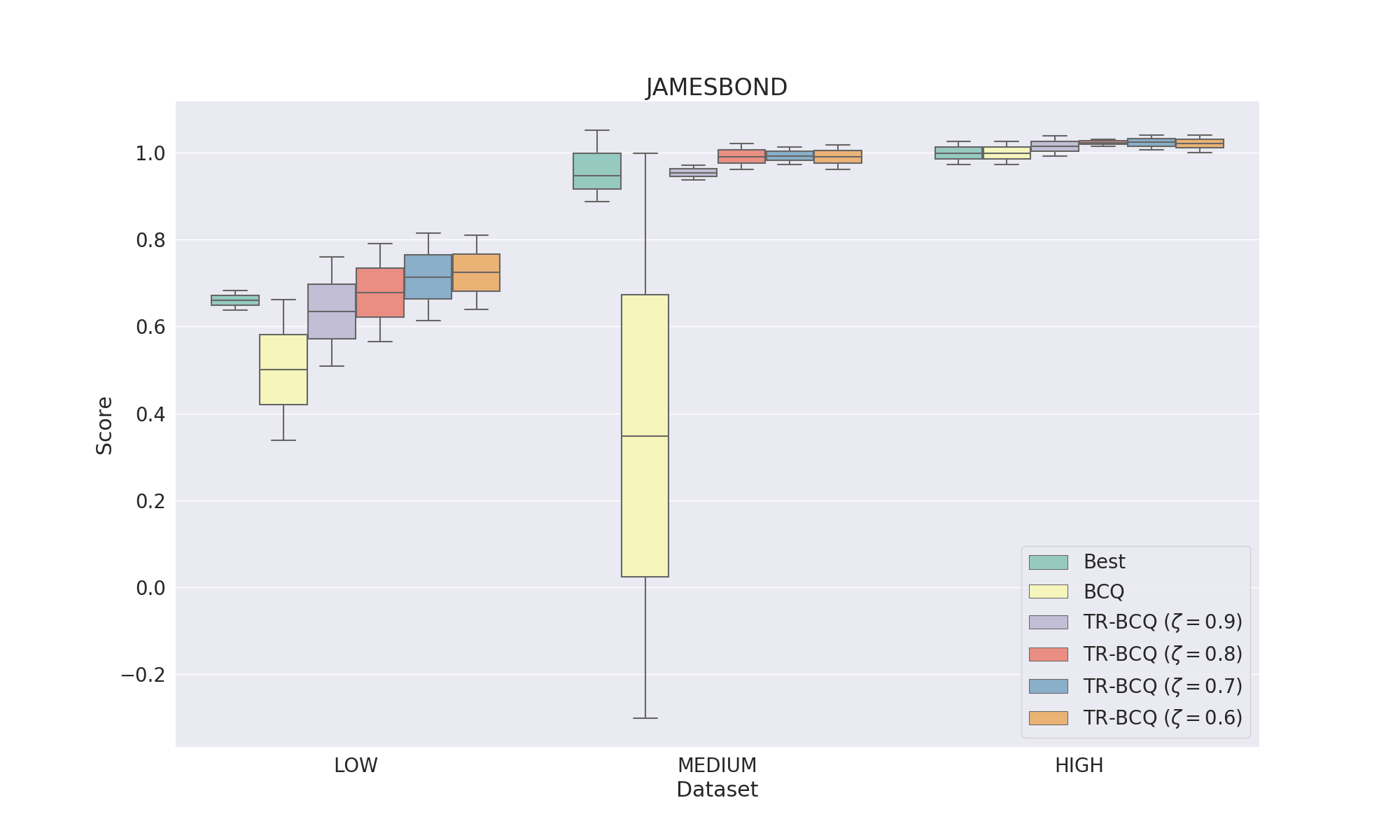}\\
				\vspace{0.01cm}
			\end{minipage}%
		}%

		\subfigure{
			\begin{minipage}[t]{0.333\linewidth}
				\centering
				\includegraphics[width=2.3in]{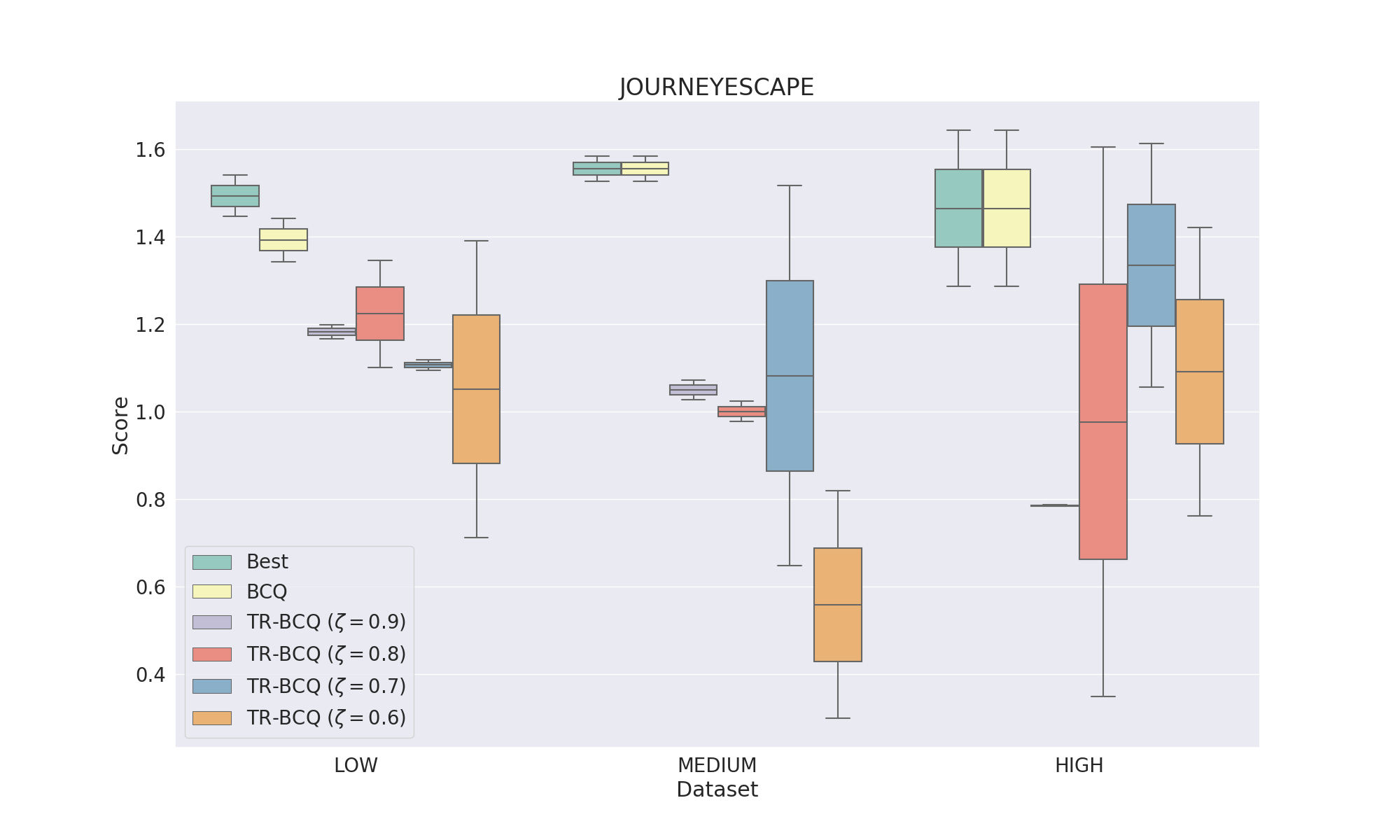}\\
				\vspace{0.01cm}
			\end{minipage}%
		}%
		\subfigure{
			\begin{minipage}[t]{0.333\linewidth}
				\centering
				\includegraphics[width=2.3in]{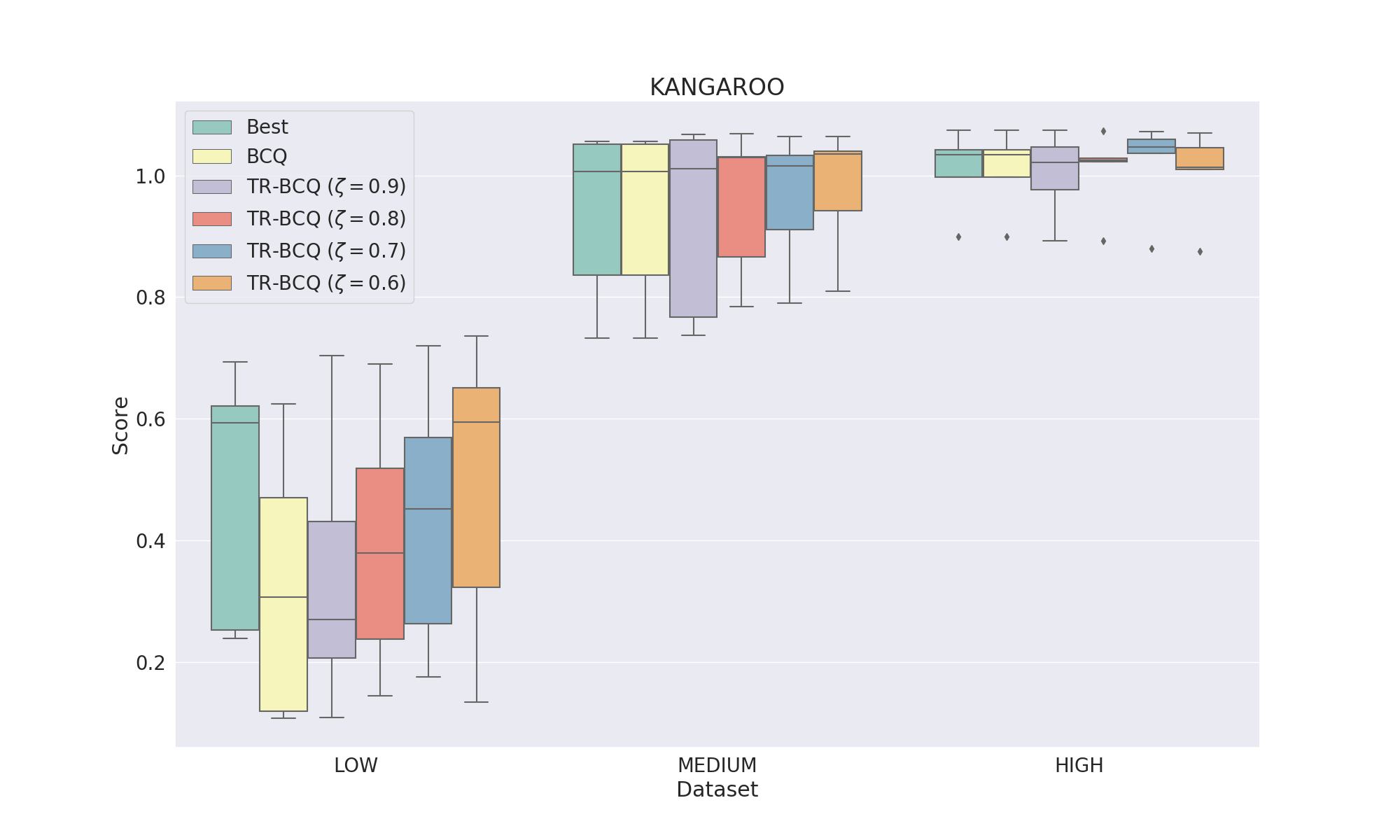}\\
				\vspace{0.01cm}
			\end{minipage}%
		}%
		\subfigure{
			\begin{minipage}[t]{0.333\linewidth}
				\centering
				\includegraphics[width=2.3in]{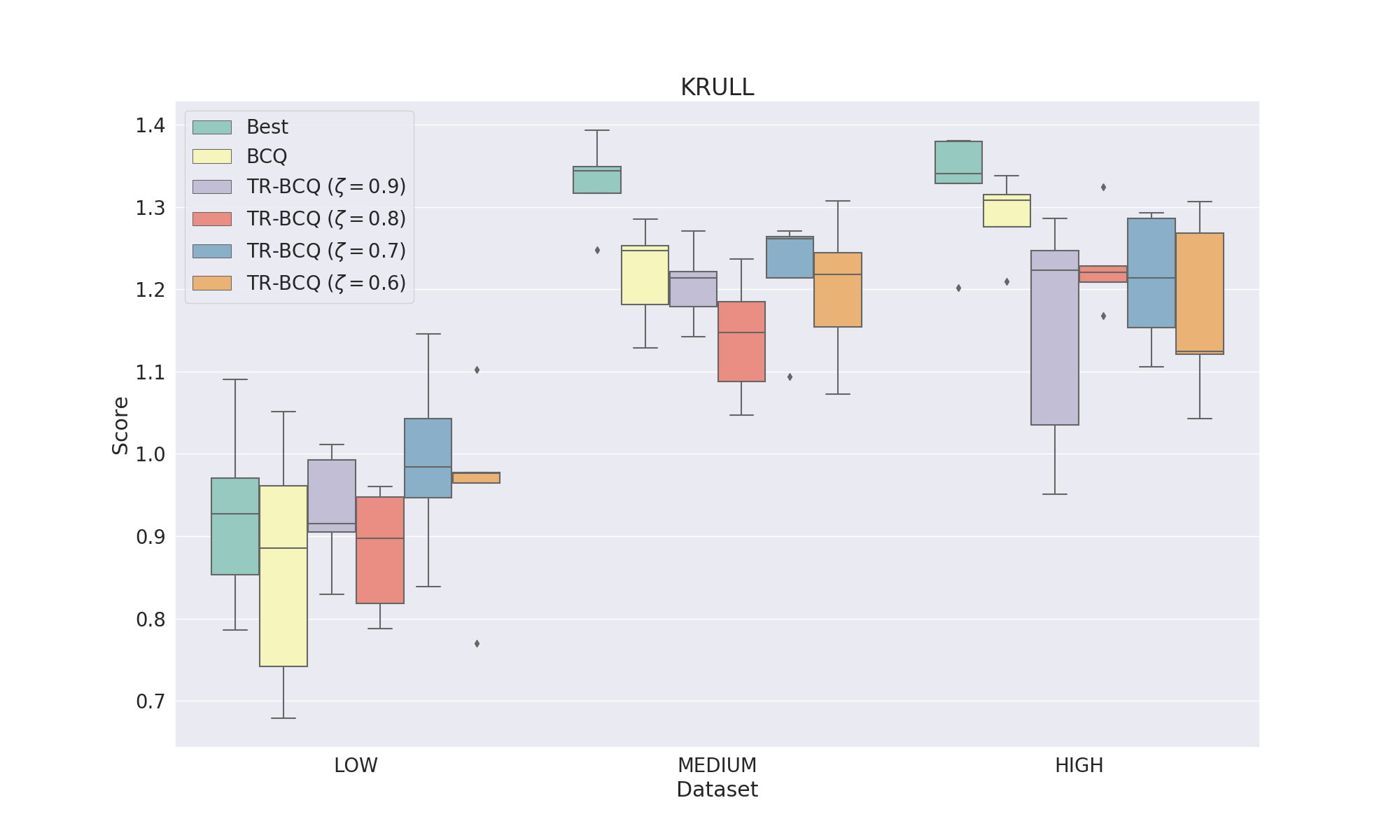}\\
				\vspace{0.01cm}
			\end{minipage}%
		}%

		\subfigure{
			\begin{minipage}[t]{0.333\linewidth}
				\centering
				\includegraphics[width=2.3in]{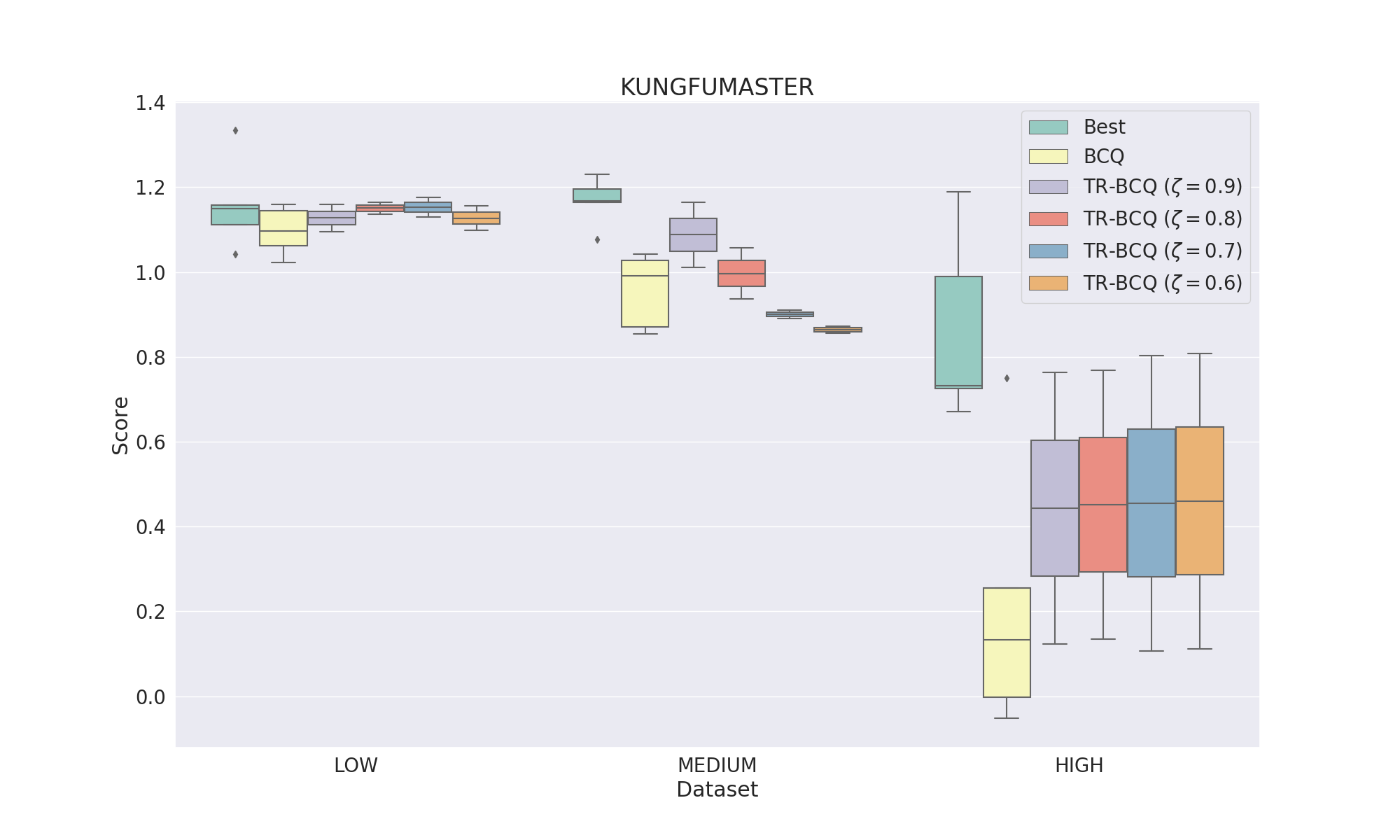}\\
				\vspace{0.01cm}
			\end{minipage}%
		}%
		\subfigure{
			\begin{minipage}[t]{0.333\linewidth}
				\centering
				\includegraphics[width=2.3in]{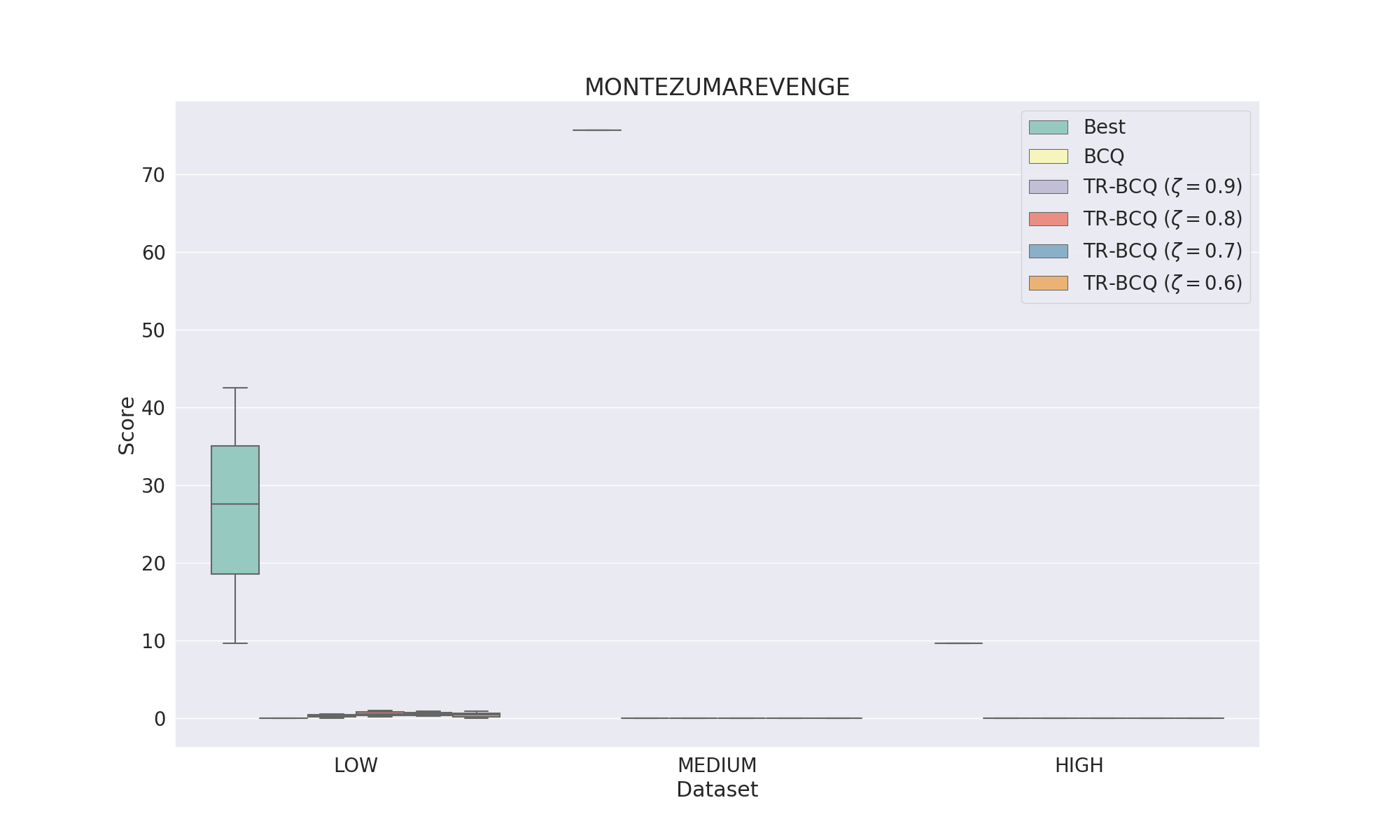}\\
				\vspace{0.01cm}
			\end{minipage}%
		}%
		\subfigure{
			\begin{minipage}[t]{0.333\linewidth}
				\centering
				\includegraphics[width=2.3in]{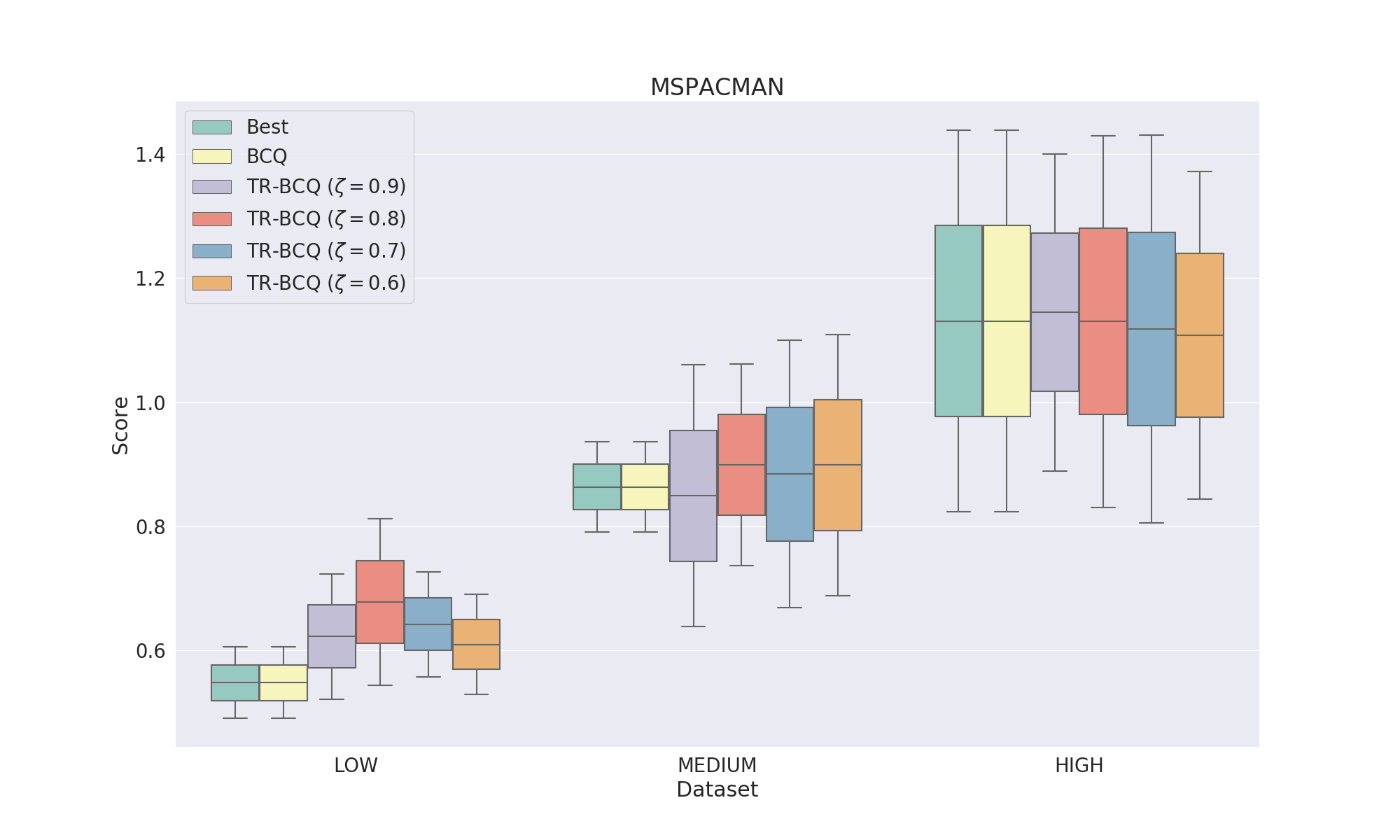}\\
				\vspace{0.01cm}
			\end{minipage}%
		}%

		\centering
		\caption{\textbf{Comparison between  TR-BCQ and the best baselines on different datasets from Game DemonAttack to Game MsPacman}}
		\label{fig: Comparison between  TR-BCQ and best baselines on different datasets from Game DemonAttack to Game MsPacman}
								
	\end{figure*}

	\begin{figure*}[!htb]
		\centering
		
		\subfigure{
			\begin{minipage}[t]{0.333\linewidth}
				\centering
				\includegraphics[width=2.3in]{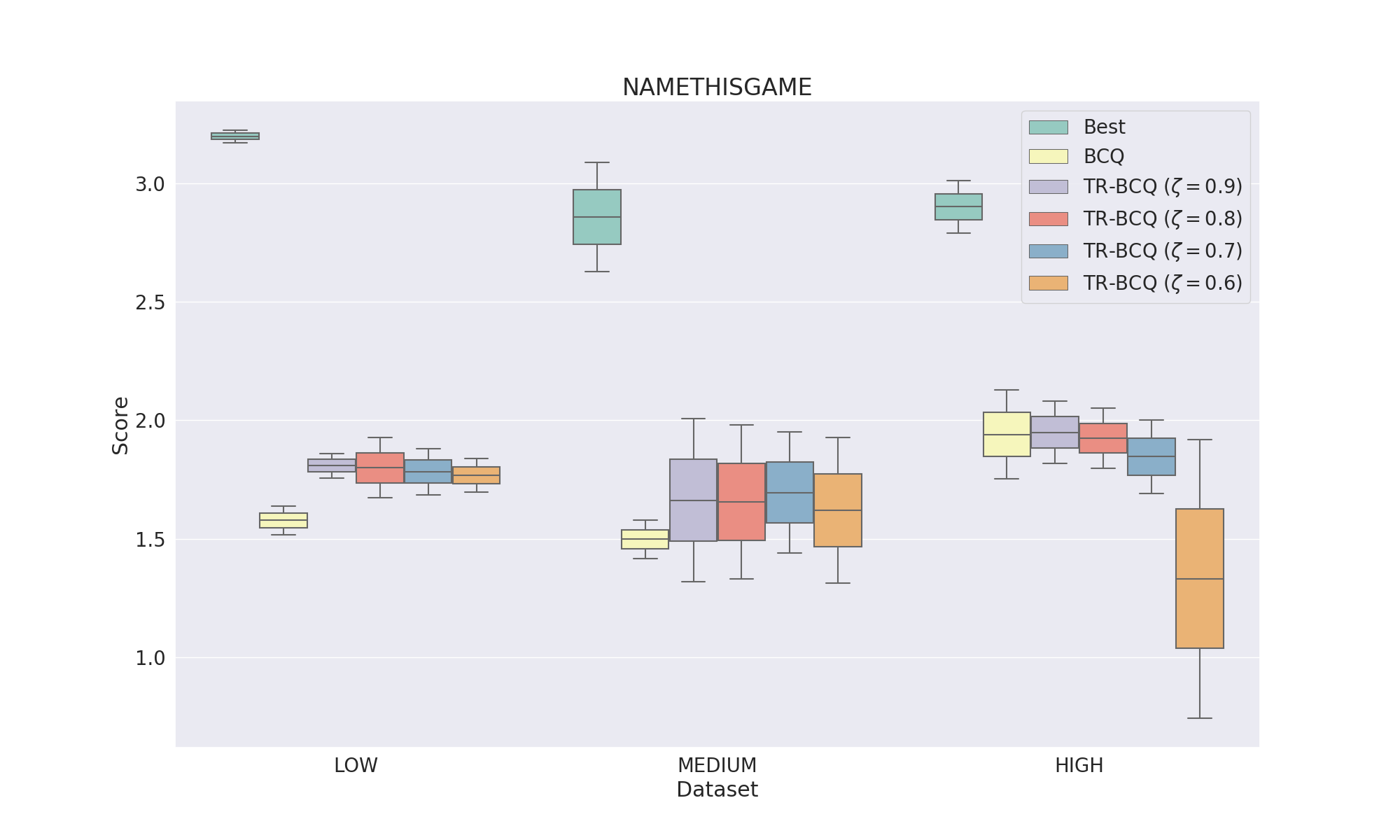}\\
				\vspace{0.01cm}
			\end{minipage}%
		}%
		\subfigure{
			\begin{minipage}[t]{0.333\linewidth}
				\centering
				\includegraphics[width=2.3in]{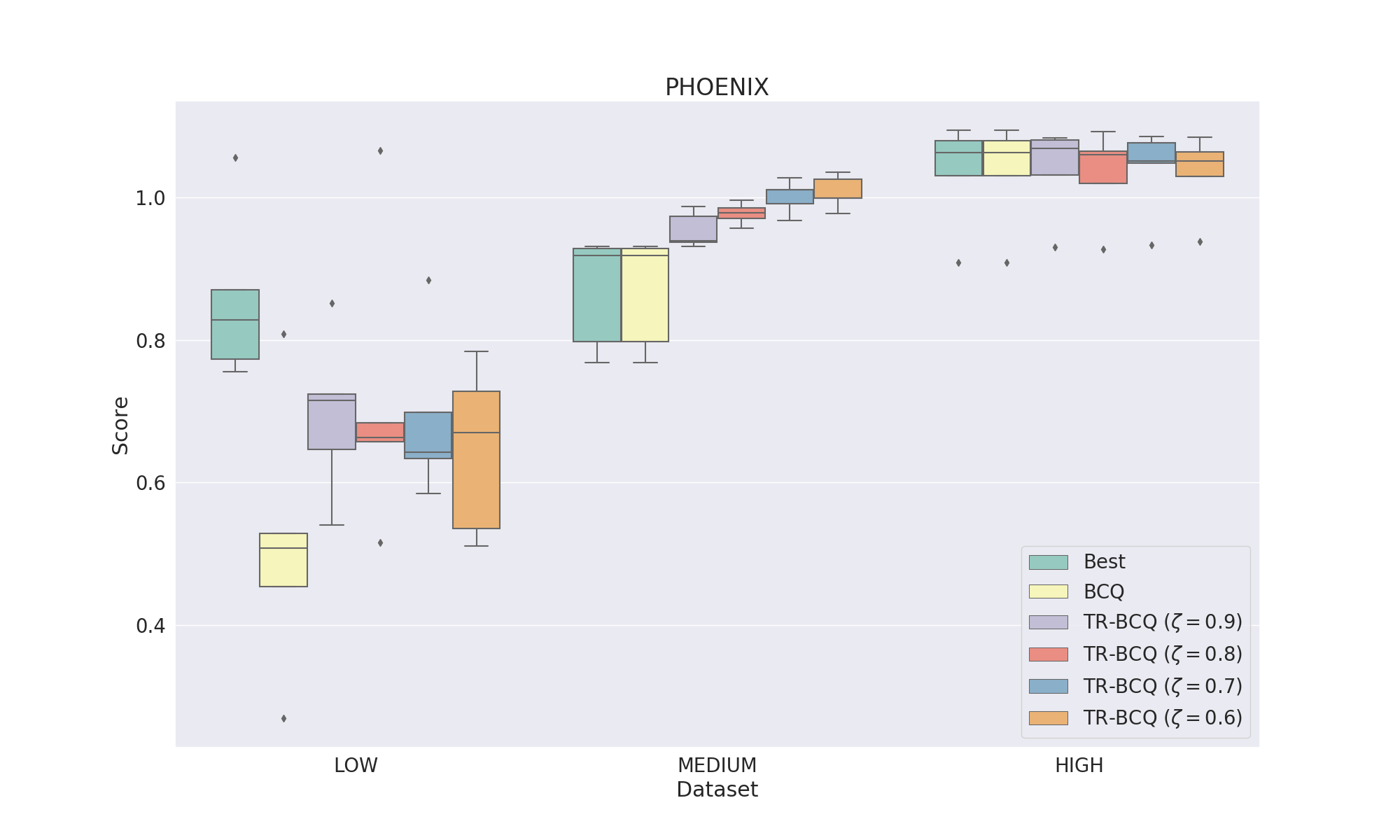}\\
				\vspace{0.01cm}
			\end{minipage}%
		}%
		\subfigure{
			\begin{minipage}[t]{0.333\linewidth}
				\centering
				\includegraphics[width=2.3in]{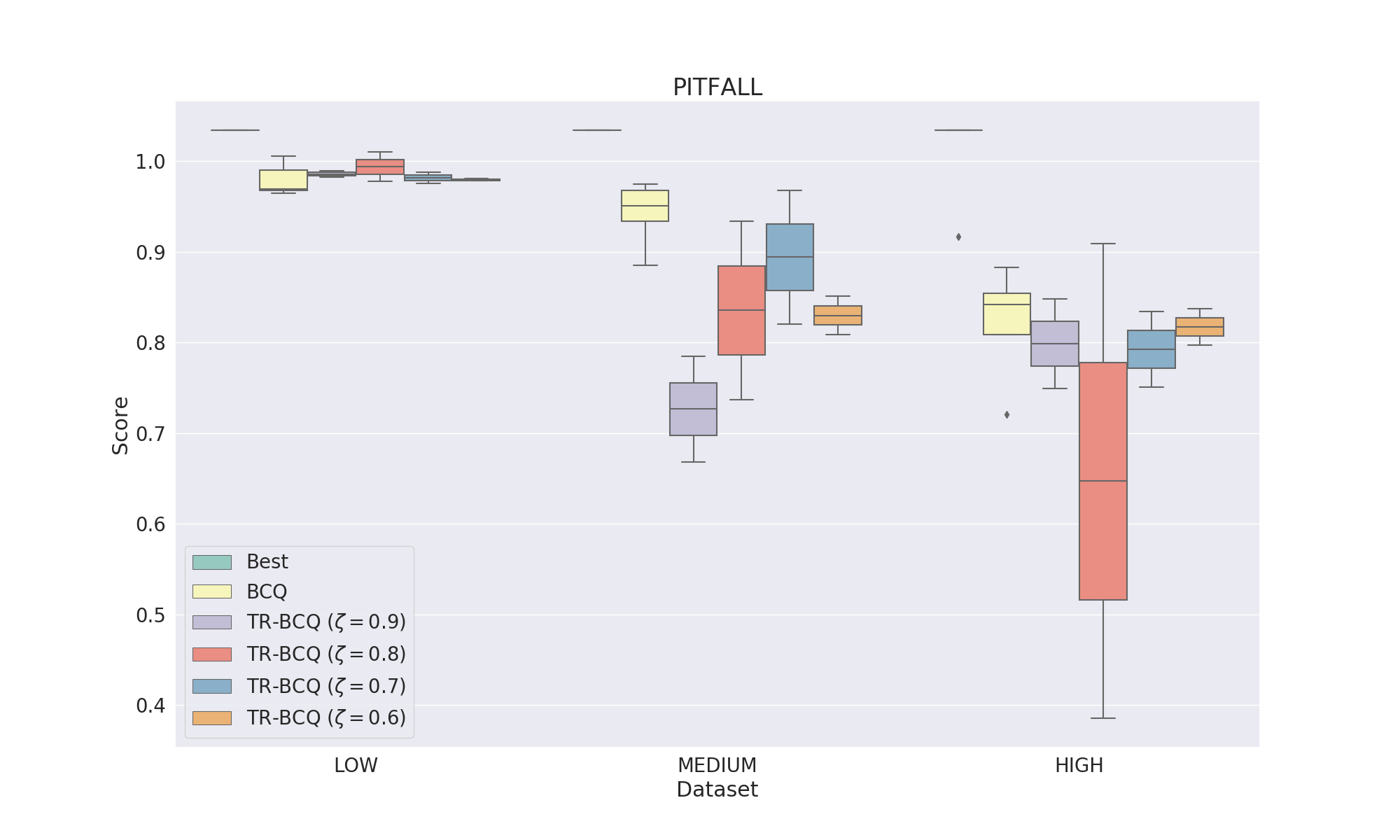}\\
				\vspace{0.01cm}
			\end{minipage}%
		}%

		\subfigure{
			\begin{minipage}[t]{0.333\linewidth}
				\centering
				\includegraphics[width=2.3in]{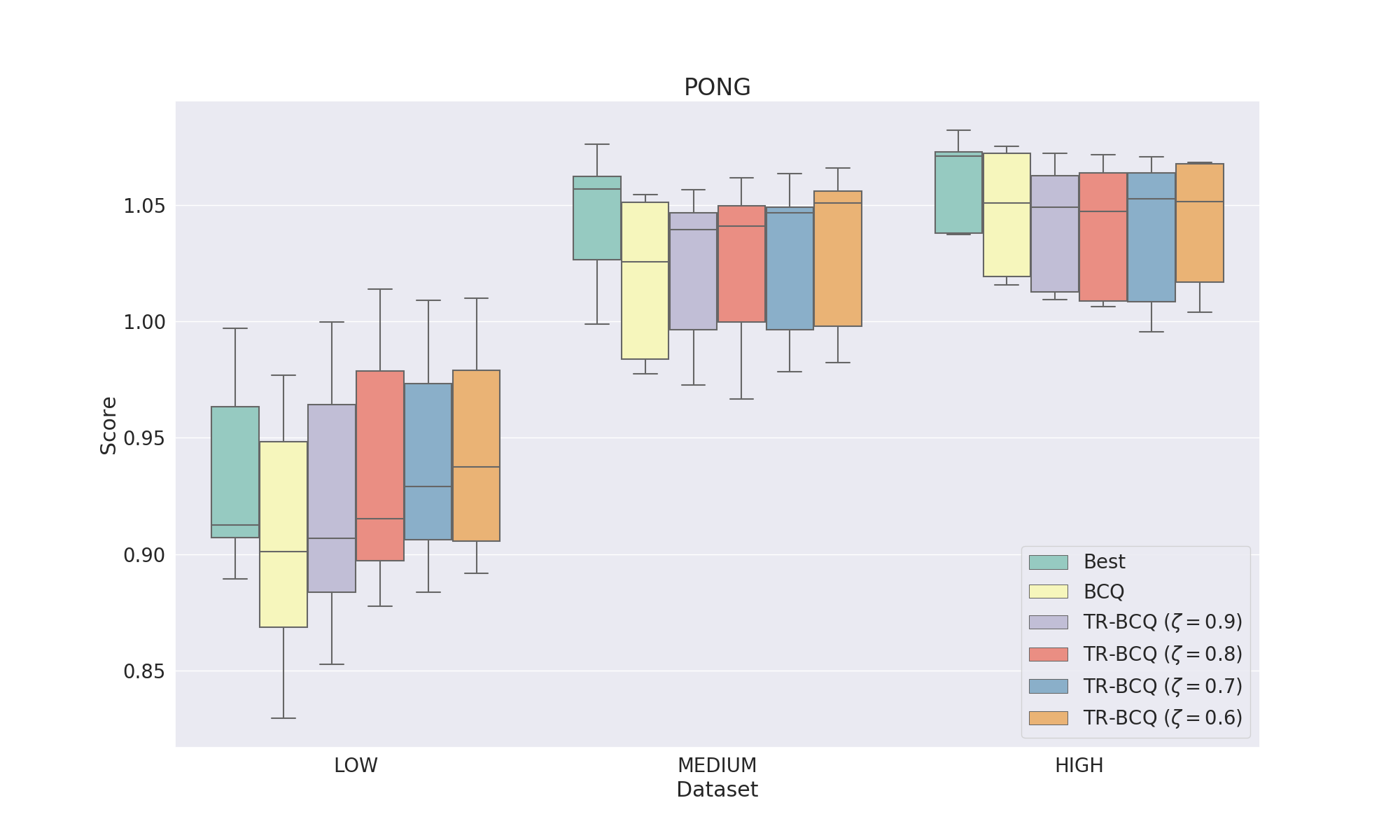}\\
				\vspace{0.01cm}
			\end{minipage}%
		}%
		\subfigure{
			\begin{minipage}[t]{0.333\linewidth}
				\centering
				\includegraphics[width=2.3in]{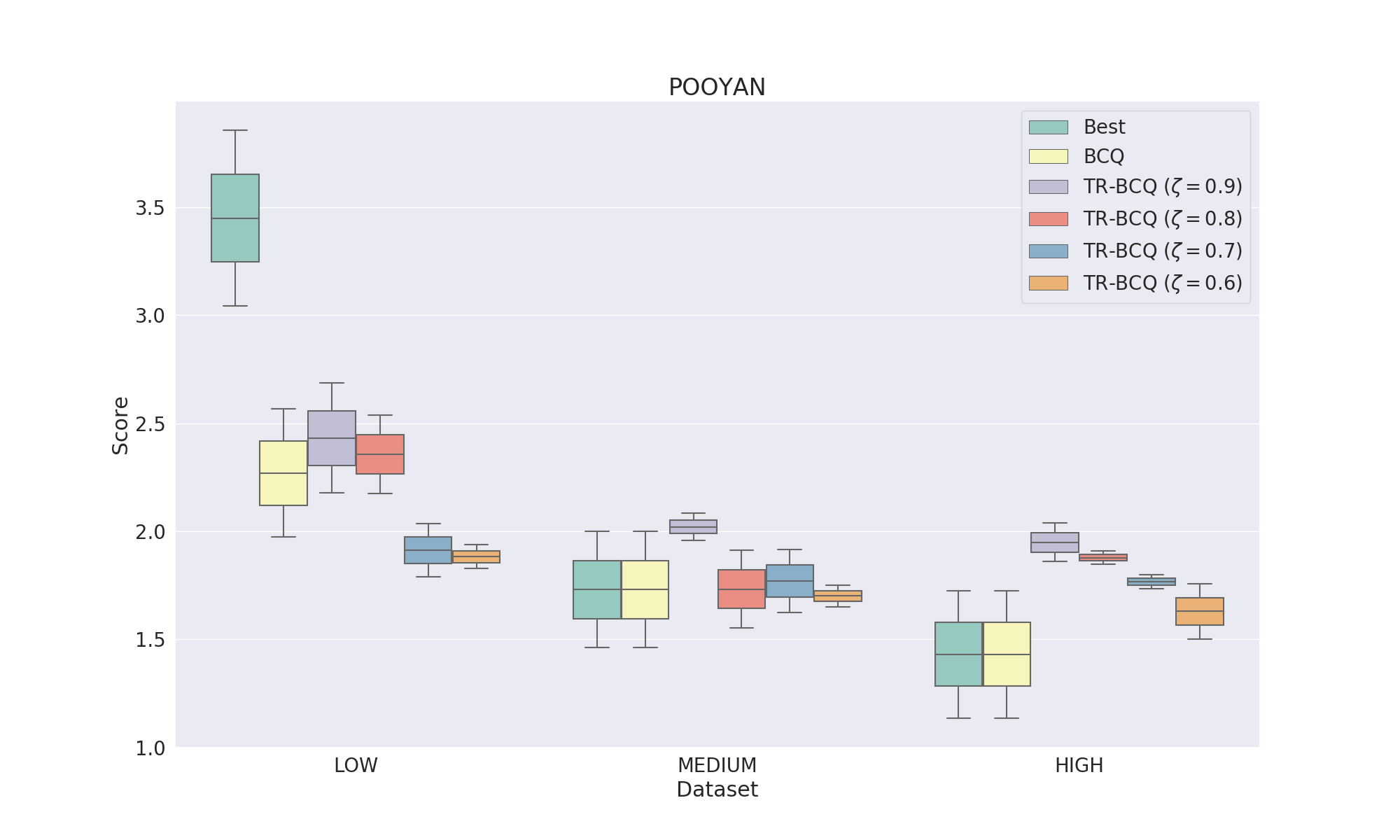}\\
				\vspace{0.01cm}
			\end{minipage}%
		}%
		\subfigure{
			\begin{minipage}[t]{0.333\linewidth}
				\centering
				\includegraphics[width=2.3in]{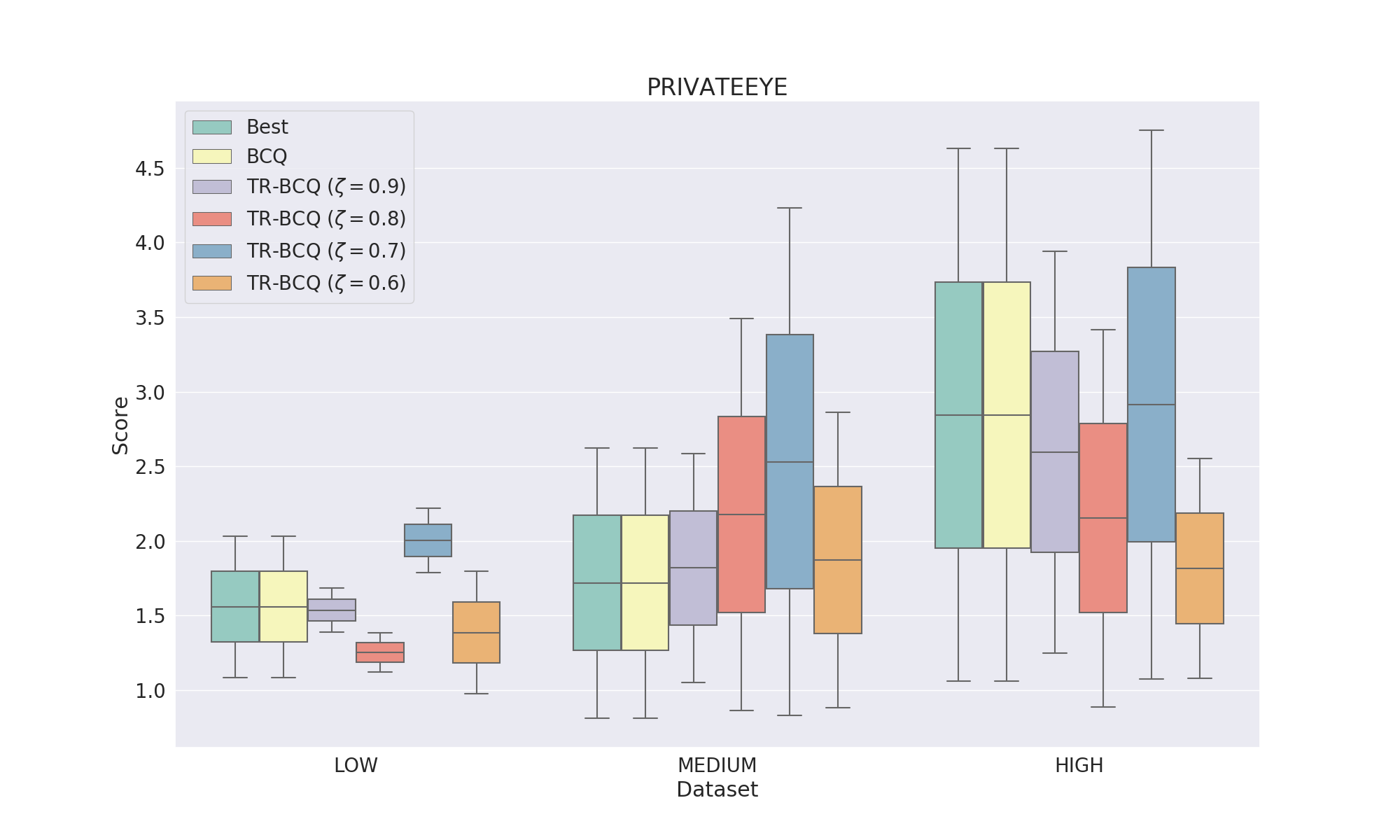}\\
				\vspace{0.01cm}
			\end{minipage}%
		}%

		\subfigure{
			\begin{minipage}[t]{0.333\linewidth}
				\centering
				\includegraphics[width=2.3in]{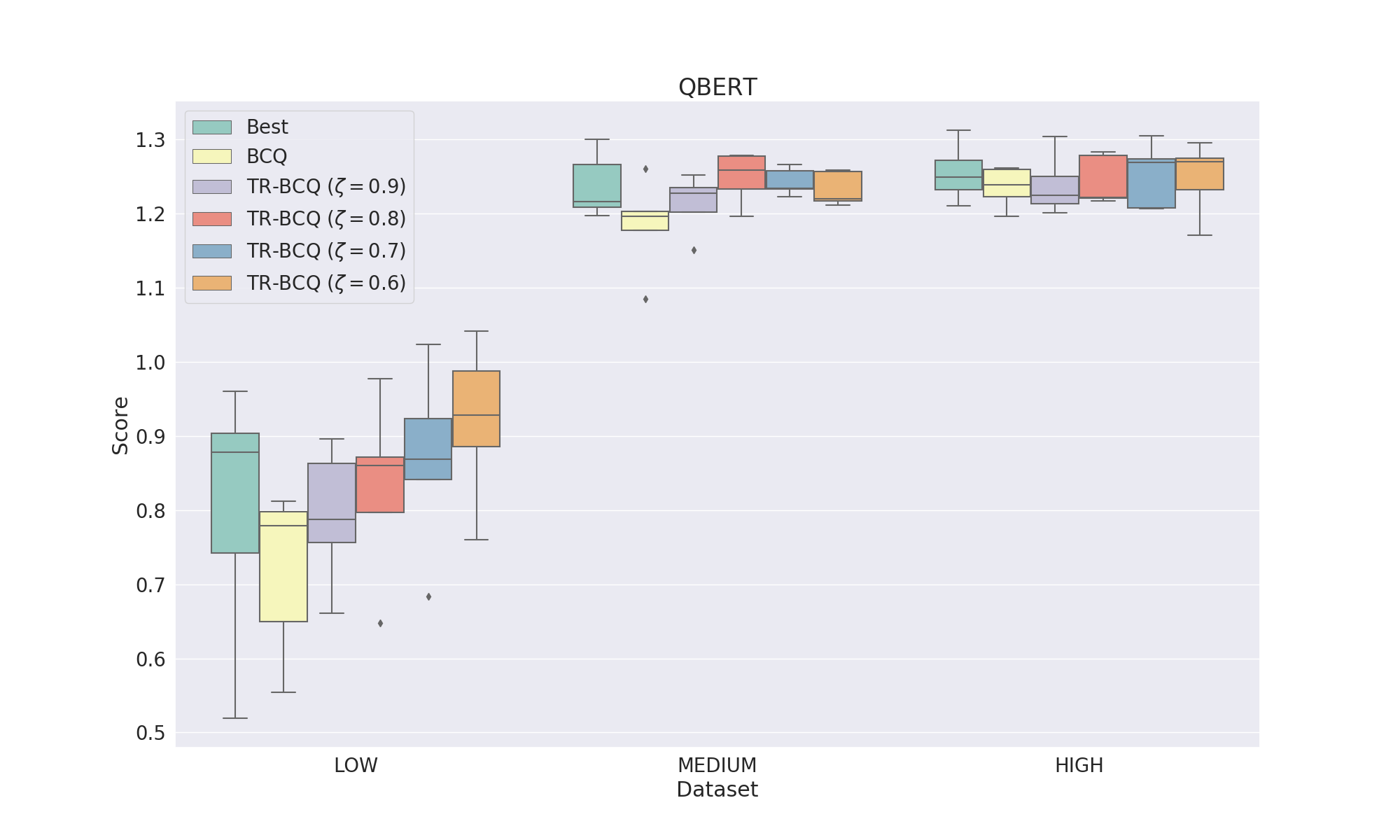}\\
				\vspace{0.01cm}
			\end{minipage}%
		}%
		\subfigure{
			\begin{minipage}[t]{0.333\linewidth}
				\centering
				\includegraphics[width=2.3in]{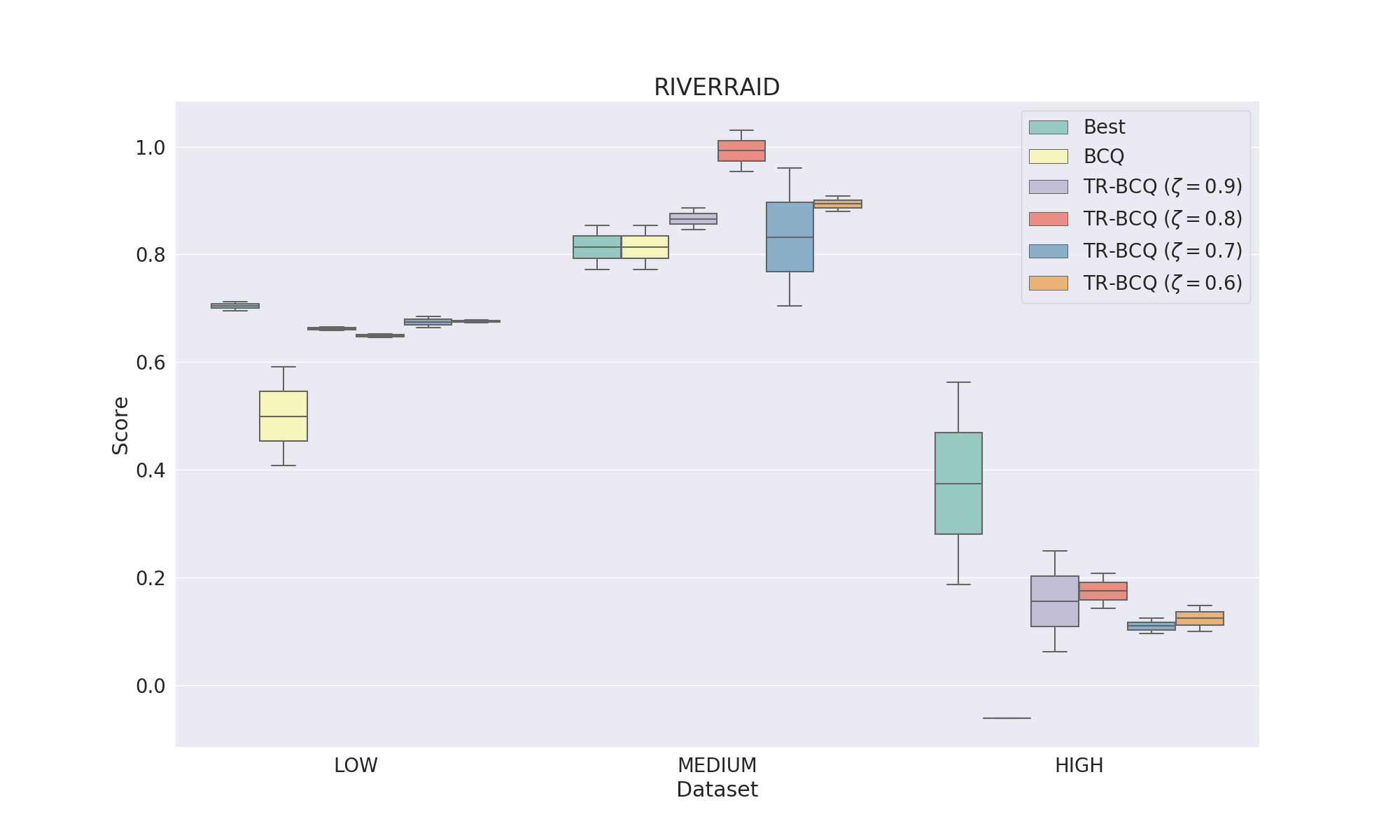}\\
				\vspace{0.01cm}
			\end{minipage}%
		}%
		\subfigure{
			\begin{minipage}[t]{0.333\linewidth}
				\centering
				\includegraphics[width=2.3in]{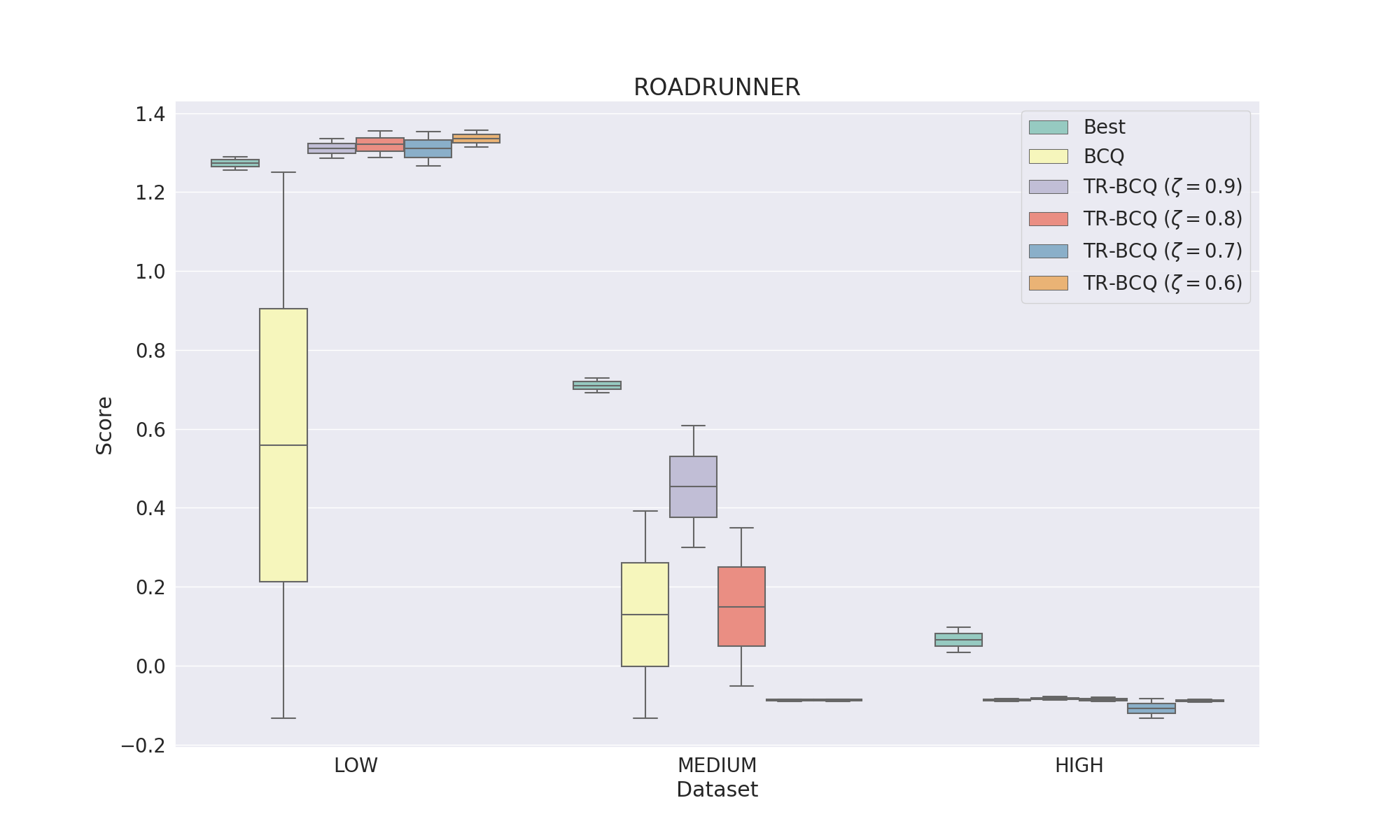}\\
				\vspace{0.01cm}
			\end{minipage}%
		}%

		\subfigure{
			\begin{minipage}[t]{0.333\linewidth}
				\centering
				\includegraphics[width=2.3in]{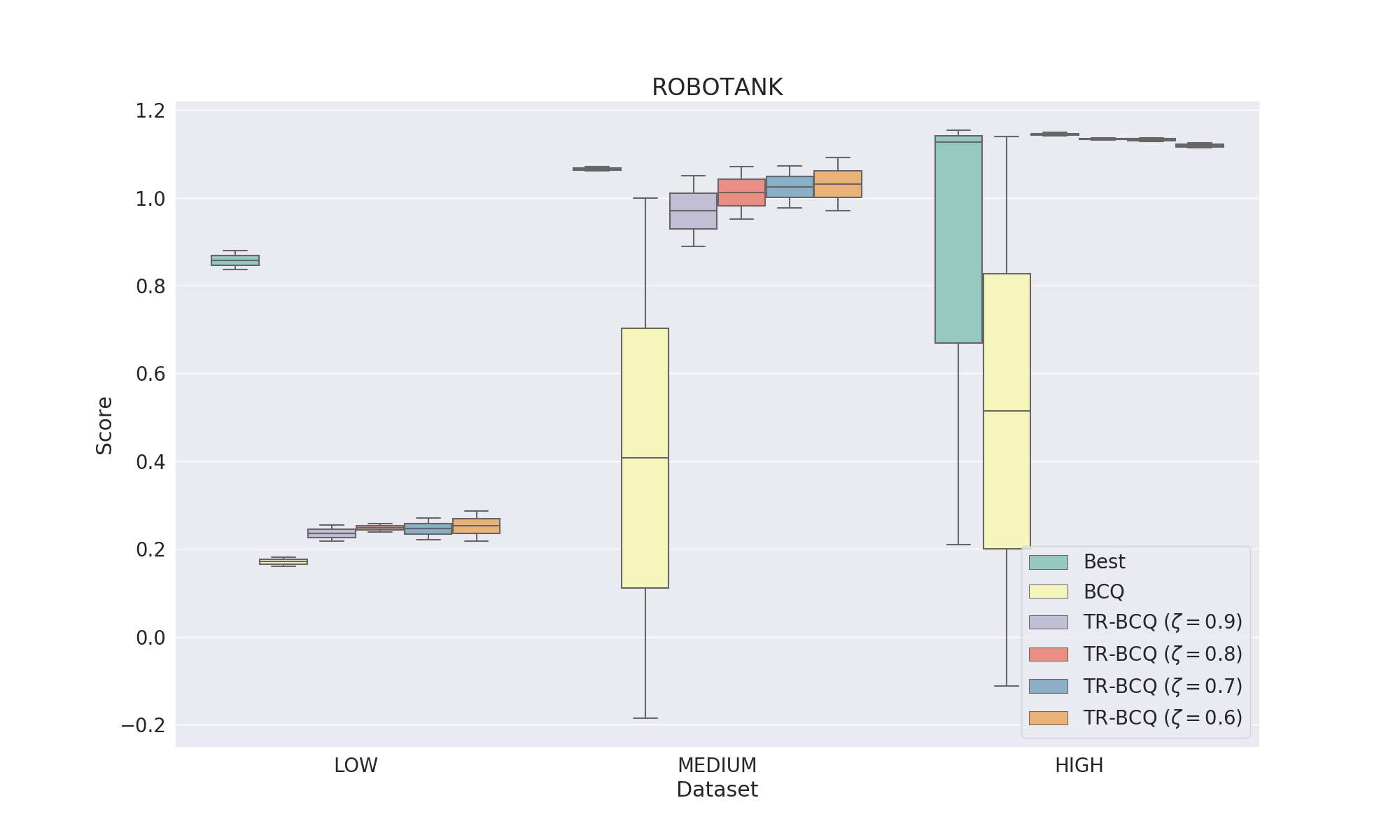}\\
				\vspace{0.01cm}
			\end{minipage}%
		}%
		\subfigure{
			\begin{minipage}[t]{0.333\linewidth}
				\centering
				\includegraphics[width=2.3in]{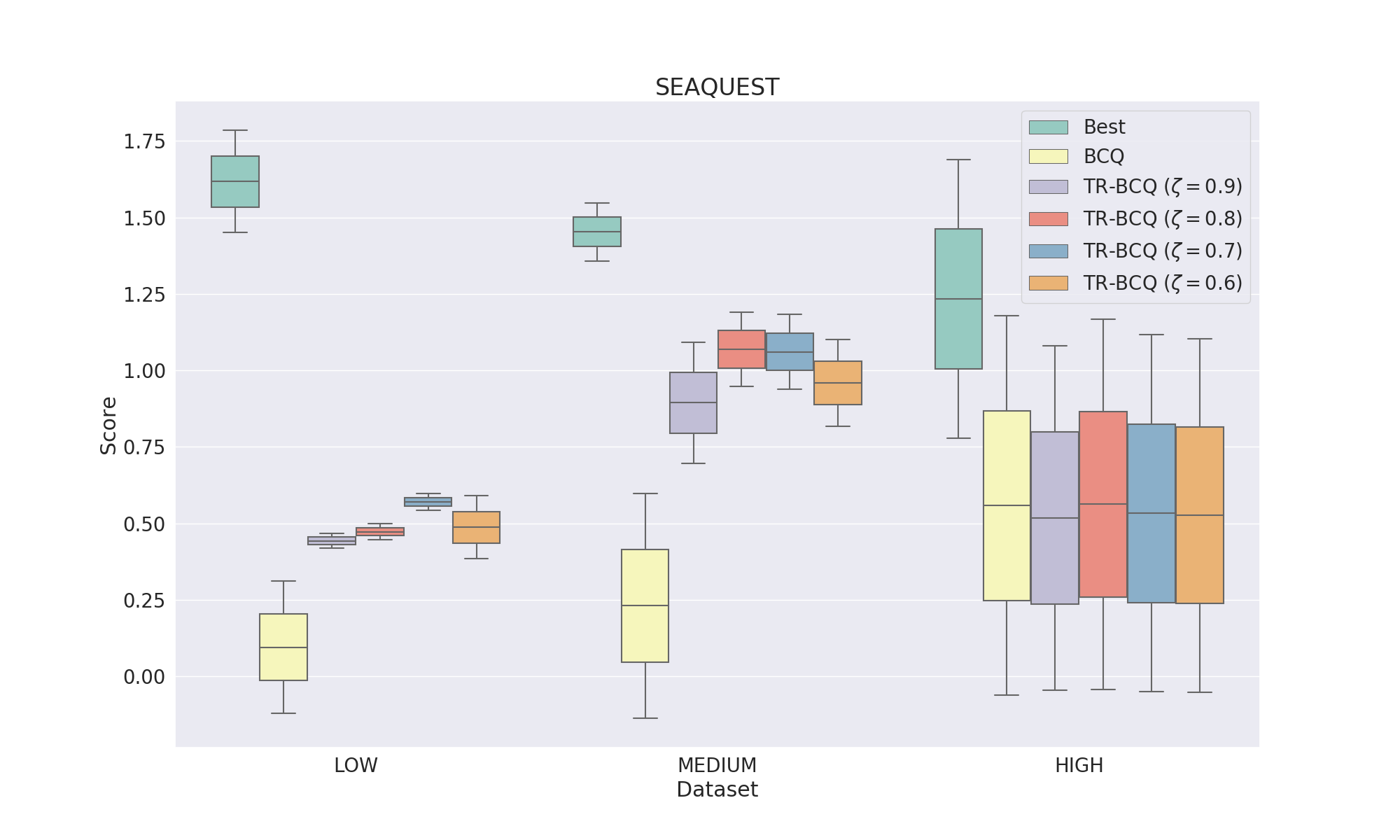}\\
				\vspace{0.01cm}
			\end{minipage}%
		}%
		\subfigure{
			\begin{minipage}[t]{0.333\linewidth}
				\centering
				\includegraphics[width=2.3in]{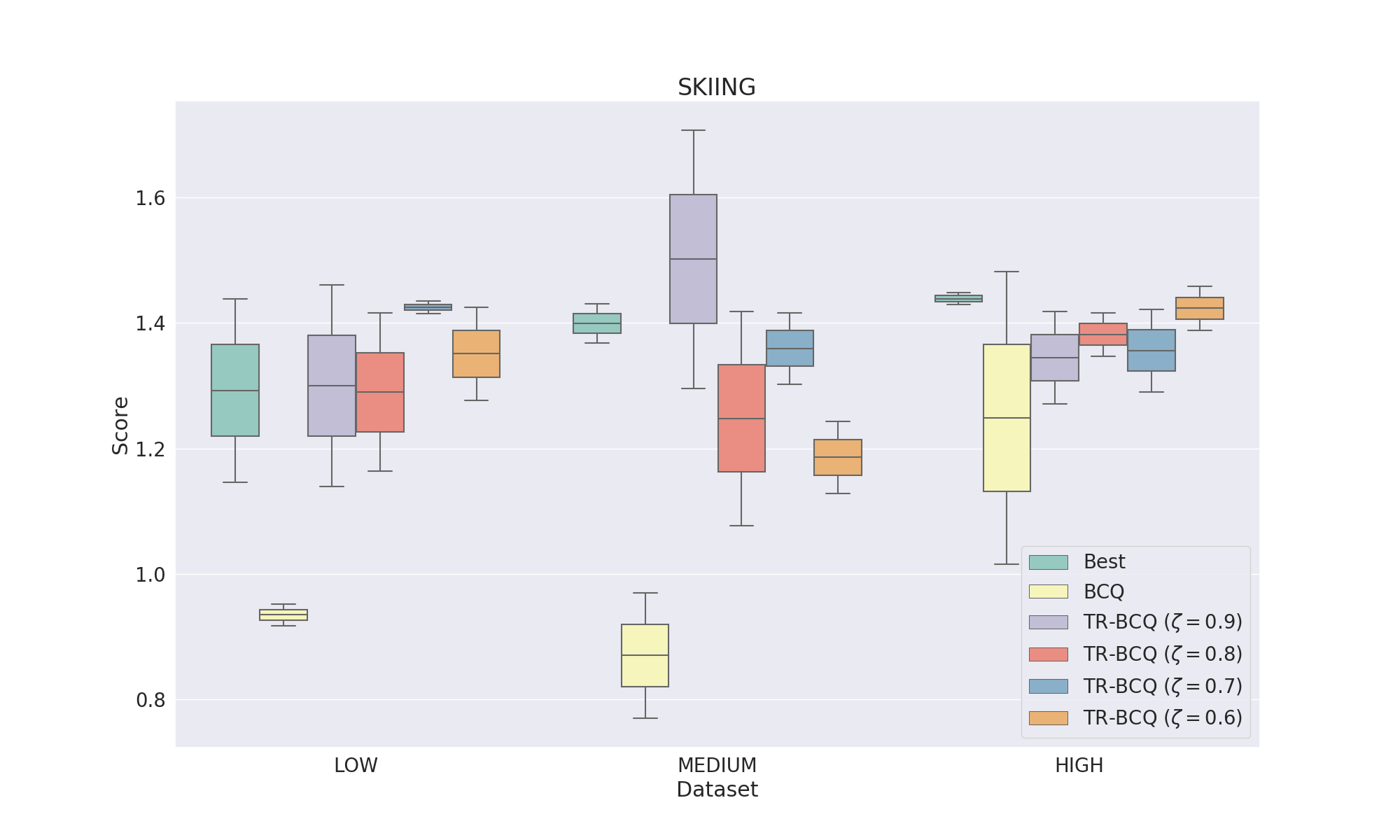}\\
				\vspace{0.01cm}
			\end{minipage}%
		}%

		\subfigure{
			\begin{minipage}[t]{0.333\linewidth}
				\centering
				\includegraphics[width=2.3in]{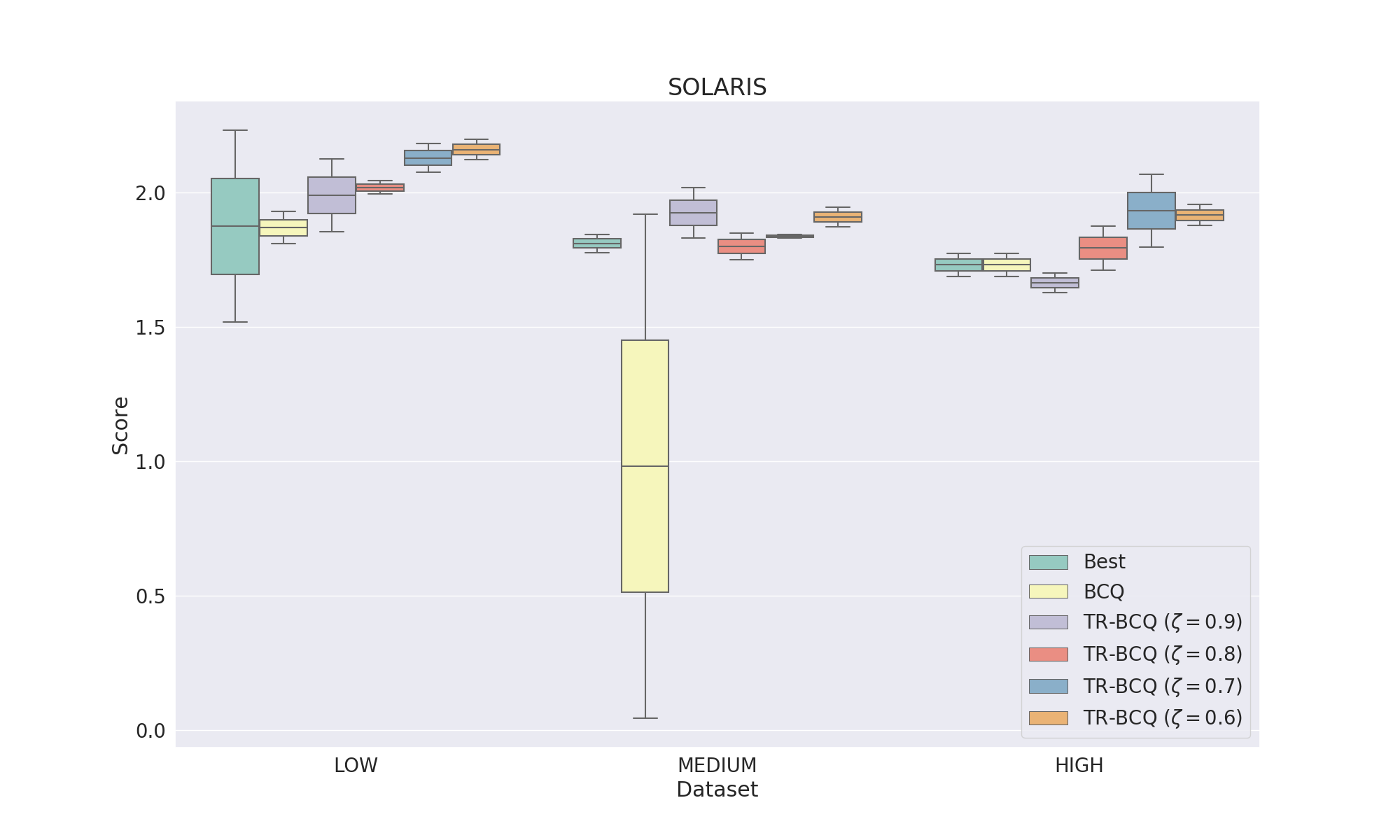}\\
				\vspace{0.01cm}
			\end{minipage}%
		}%
		\subfigure{
			\begin{minipage}[t]{0.333\linewidth}
				\centering
				\includegraphics[width=2.3in]{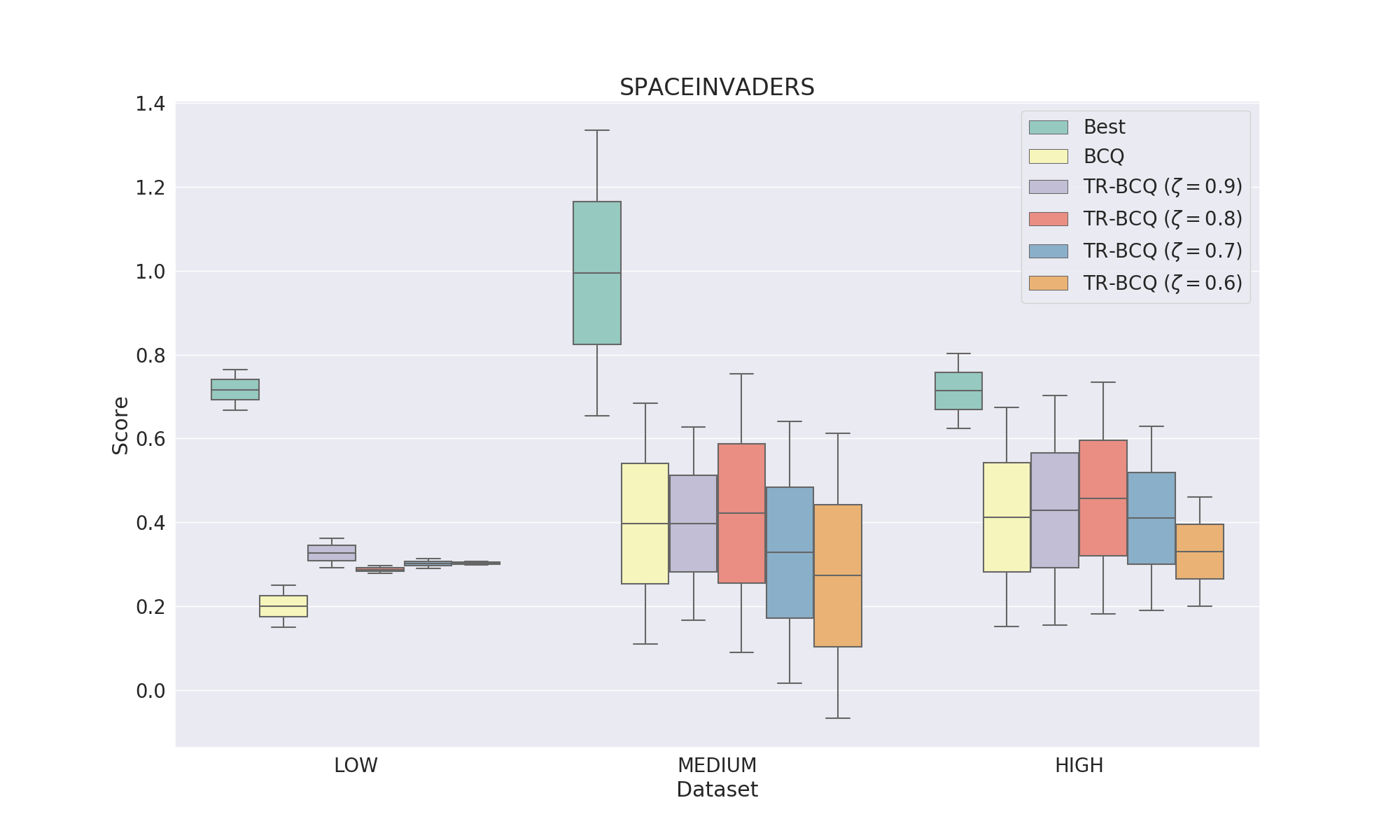}\\
				\vspace{0.01cm}
			\end{minipage}%
		}%
		\subfigure{
			\begin{minipage}[t]{0.333\linewidth}
				\centering
				\includegraphics[width=2.3in]{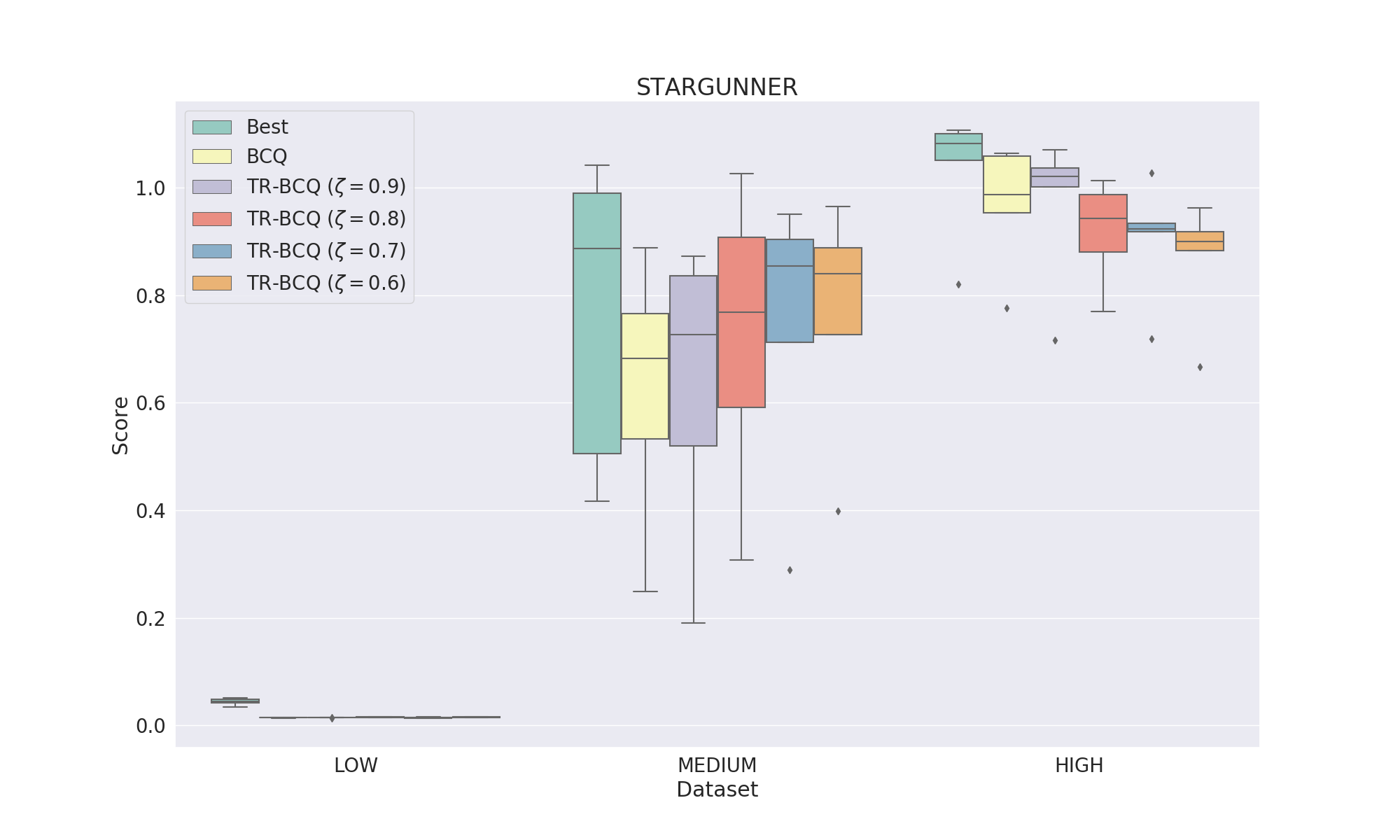}\\
				\vspace{0.01cm}
			\end{minipage}%
		}%

		\subfigure{
			\begin{minipage}[t]{0.333\linewidth}
				\centering
				\includegraphics[width=2.3in]{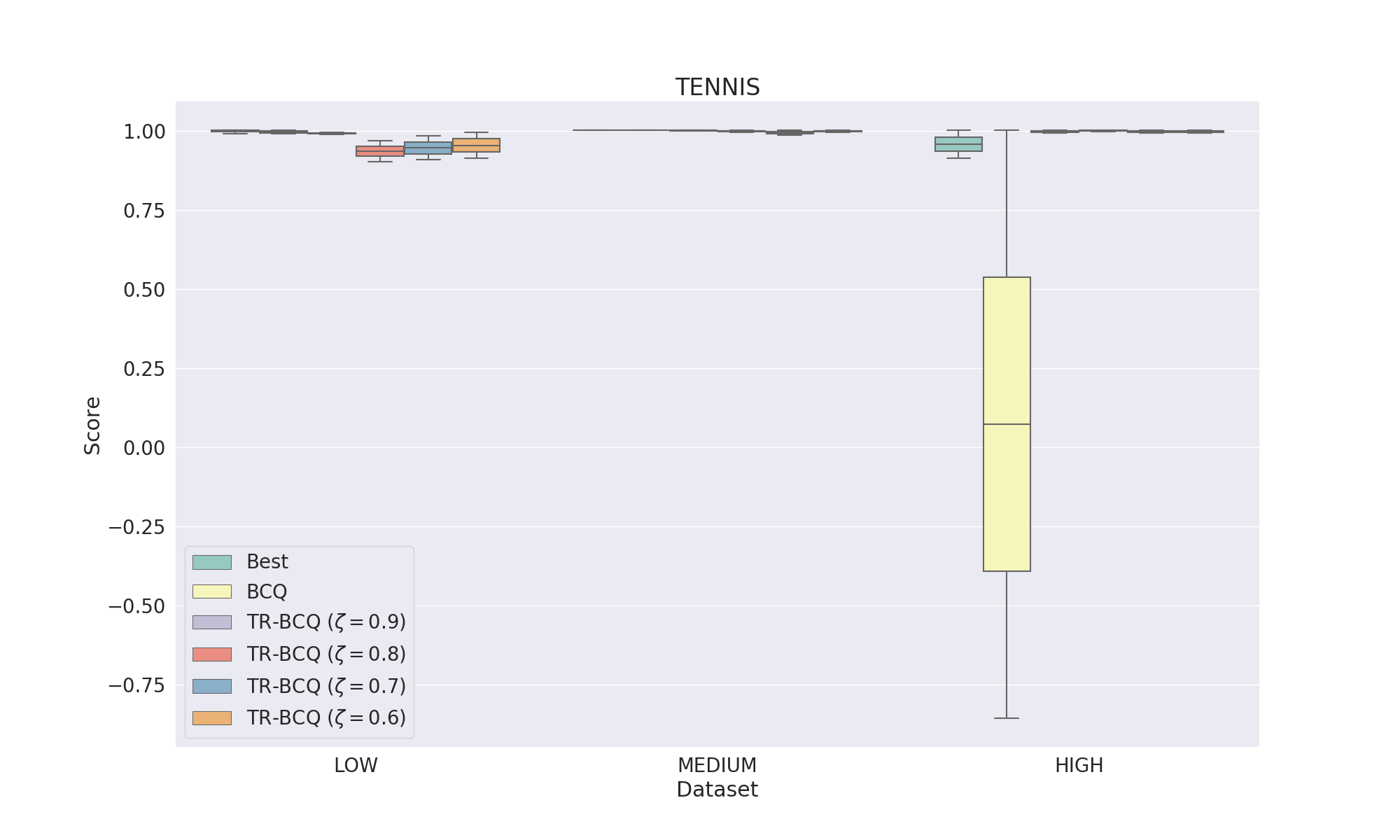}\\
				\vspace{0.01cm}
			\end{minipage}%
		}%
		\subfigure{
			\begin{minipage}[t]{0.333\linewidth}
				\centering
				\includegraphics[width=2.3in]{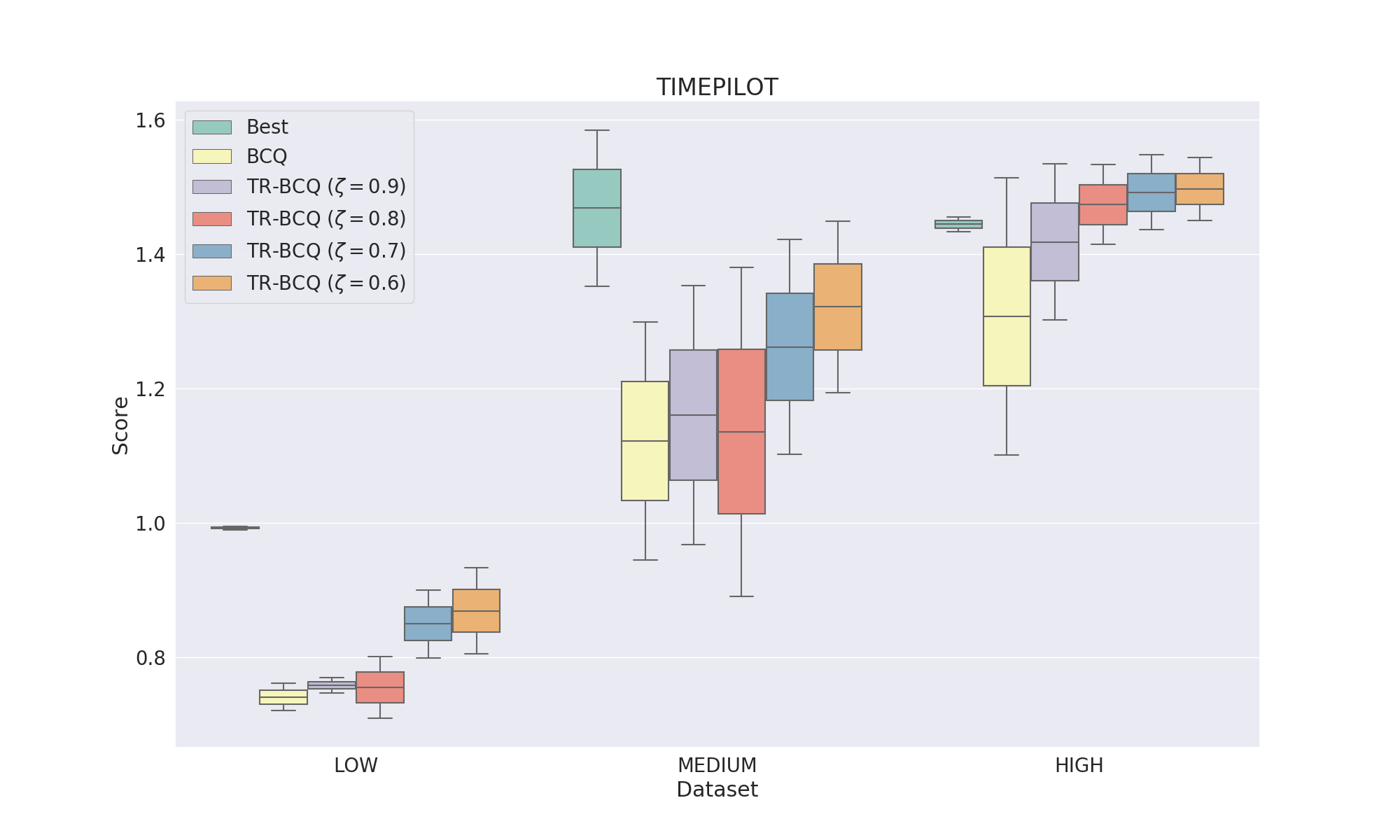}\\
				\vspace{0.01cm}
			\end{minipage}%
		}%
		\subfigure{
			\begin{minipage}[t]{0.333\linewidth}
				\centering
				\includegraphics[width=2.3in]{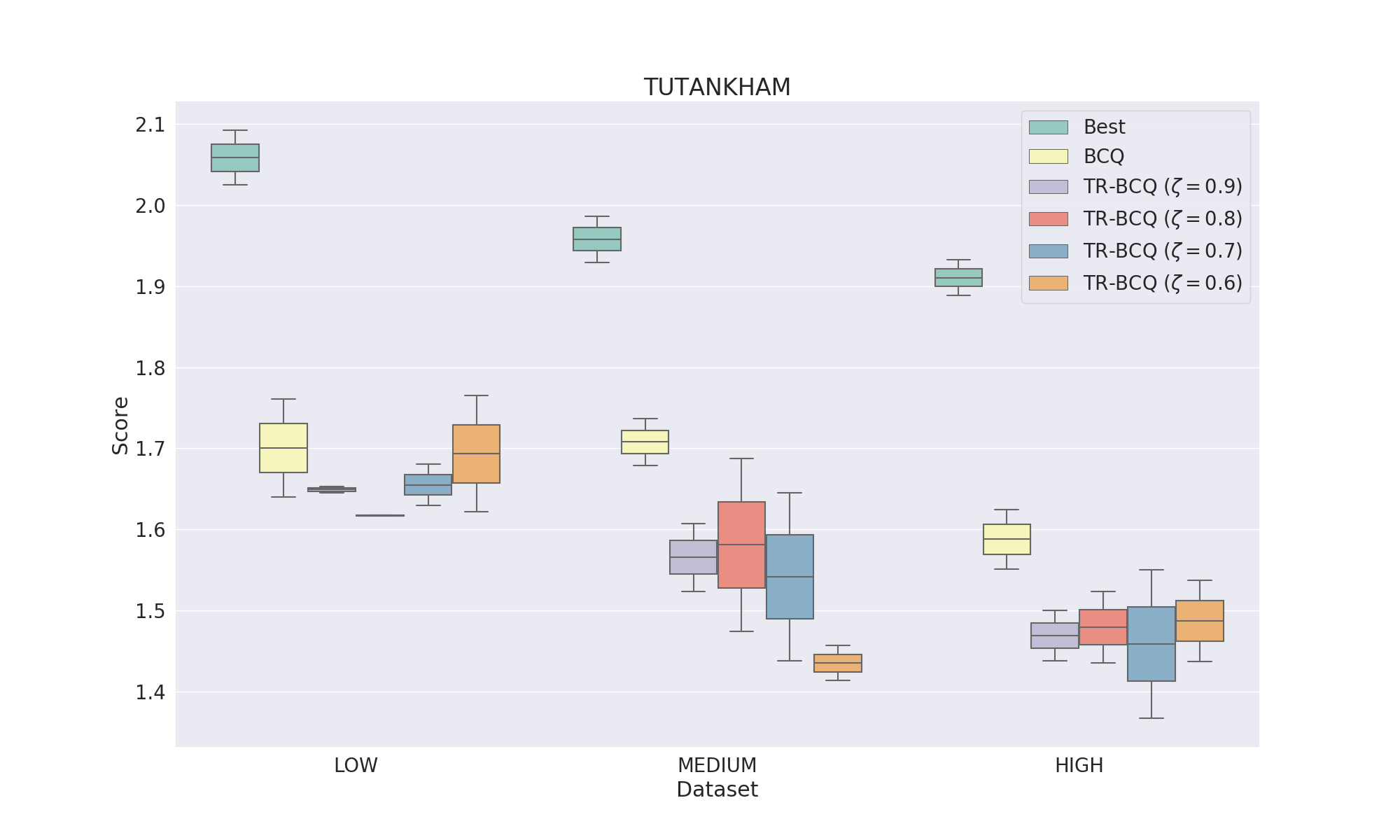}\\
				\vspace{0.01cm}
			\end{minipage}%
		}%

		\centering
		\caption{\textbf{Comparison between  TR-BCQ and best baselines on different datasets from Game NameThisGame to Game Tutankham}}
		\label{fig: Comparison between  TR-BCQ and the best baselines on different datasets from Game NameThisGame to Game Tutankham}
								
	\end{figure*}

	\begin{figure*}[htb]
		\centering

		\vspace{-8cm}

		\subfigure{
			\begin{minipage}[t]{0.333\linewidth}
				\centering
				\includegraphics[width=2.3in]{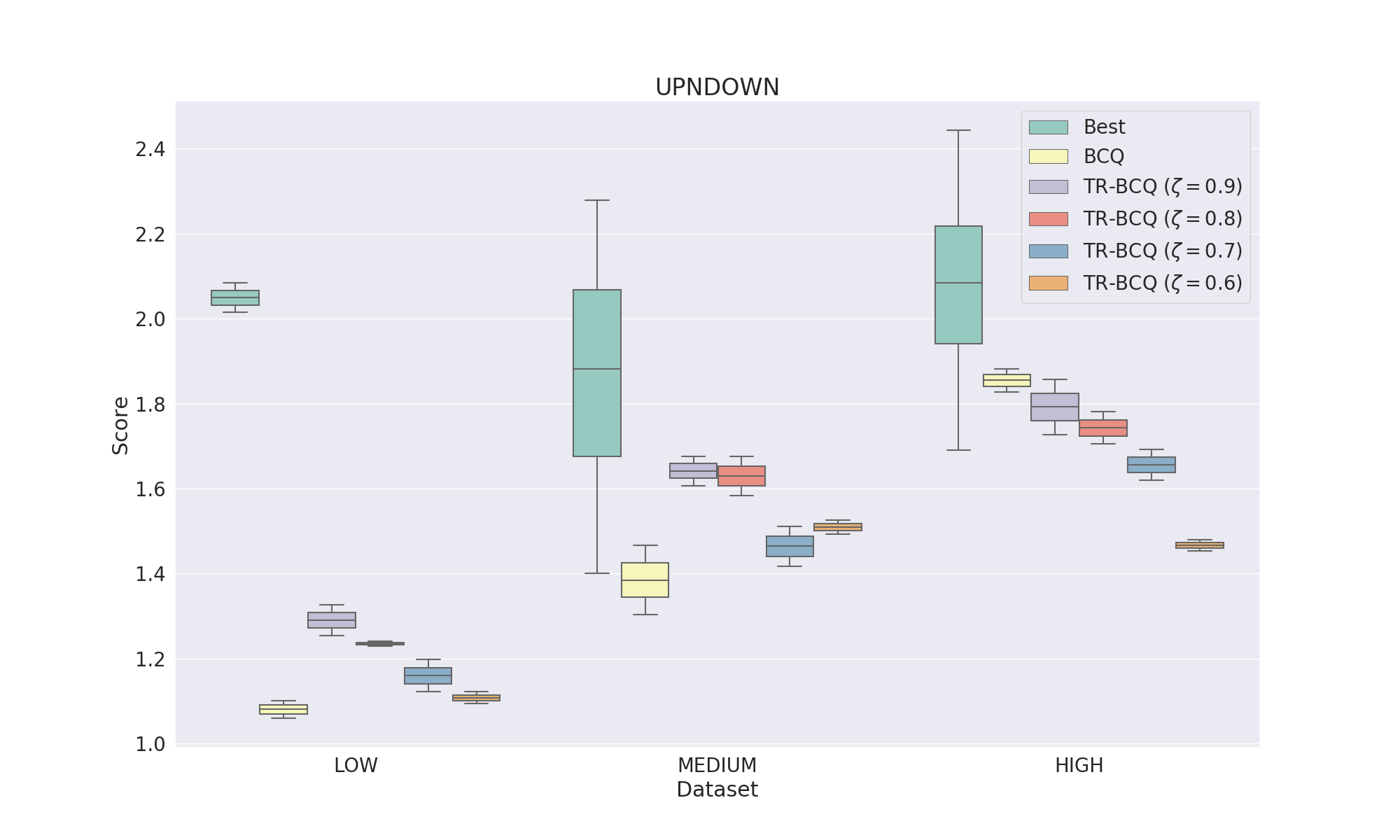}\\
				\vspace{0.01cm}
			\end{minipage}%
		}%
		\subfigure{
			\begin{minipage}[t]{0.333\linewidth}
				\centering
				\includegraphics[width=2.3in]{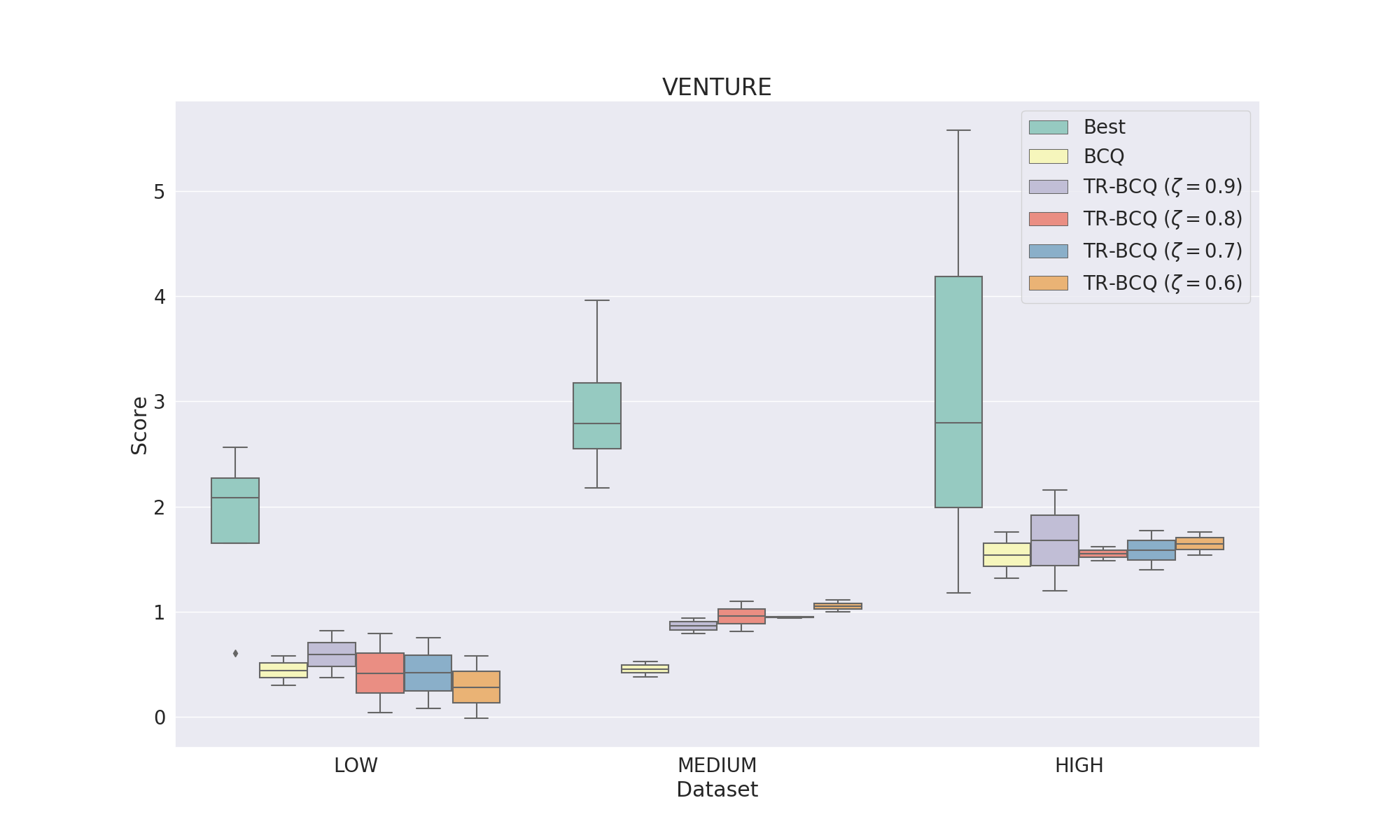}\\
				\vspace{0.01cm}
			\end{minipage}%
		}%
		\subfigure{
			\begin{minipage}[t]{0.333\linewidth}
				\centering
				\includegraphics[width=2.3in]{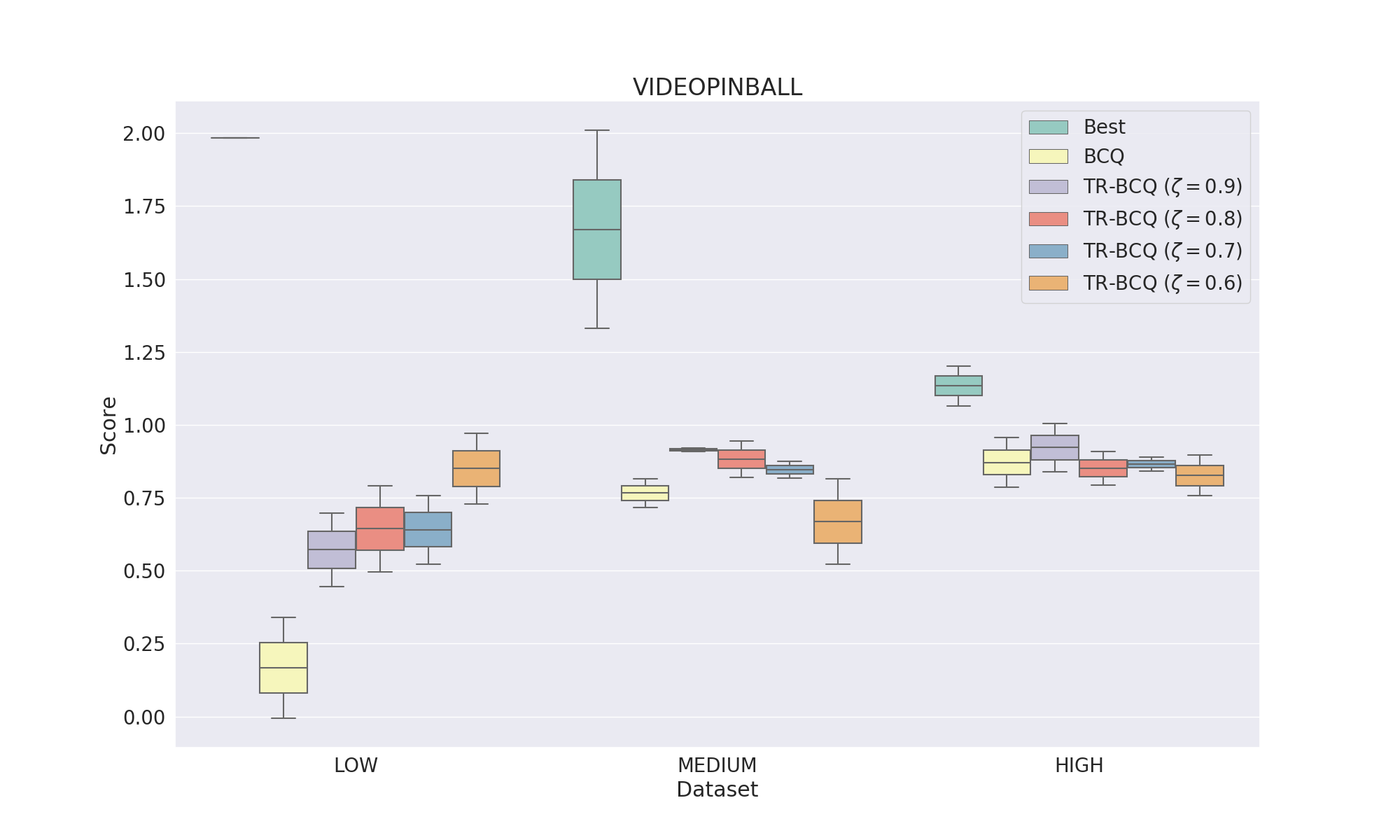}\\
				\vspace{0.01cm}
			\end{minipage}%
		}%

		\subfigure{
			\begin{minipage}[t]{0.333\linewidth}
				\centering
				\includegraphics[width=2.3in]{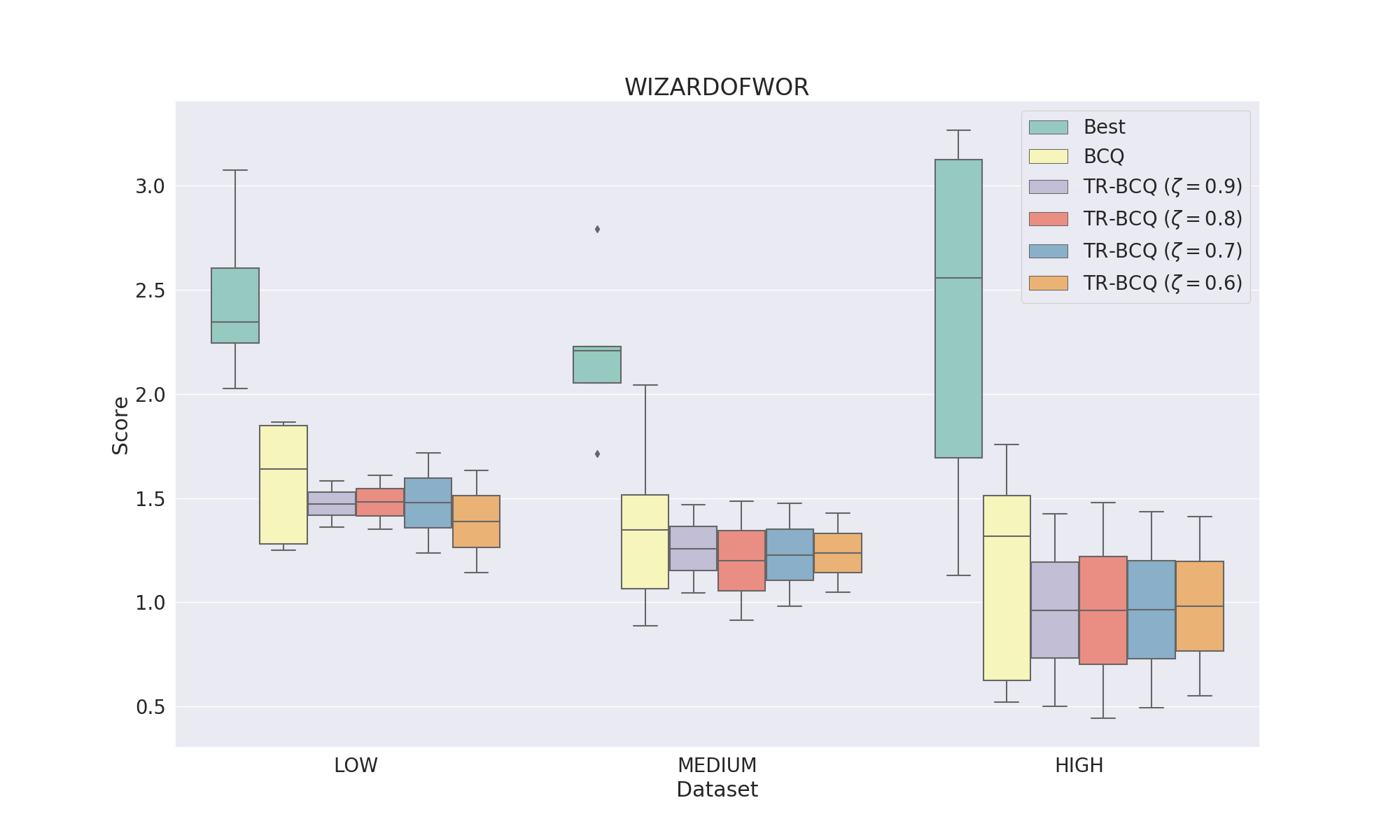}\\
				\vspace{0.01cm}
			\end{minipage}%
		}%
		\subfigure{
			\begin{minipage}[t]{0.333\linewidth}
				\centering
				\includegraphics[width=2.3in]{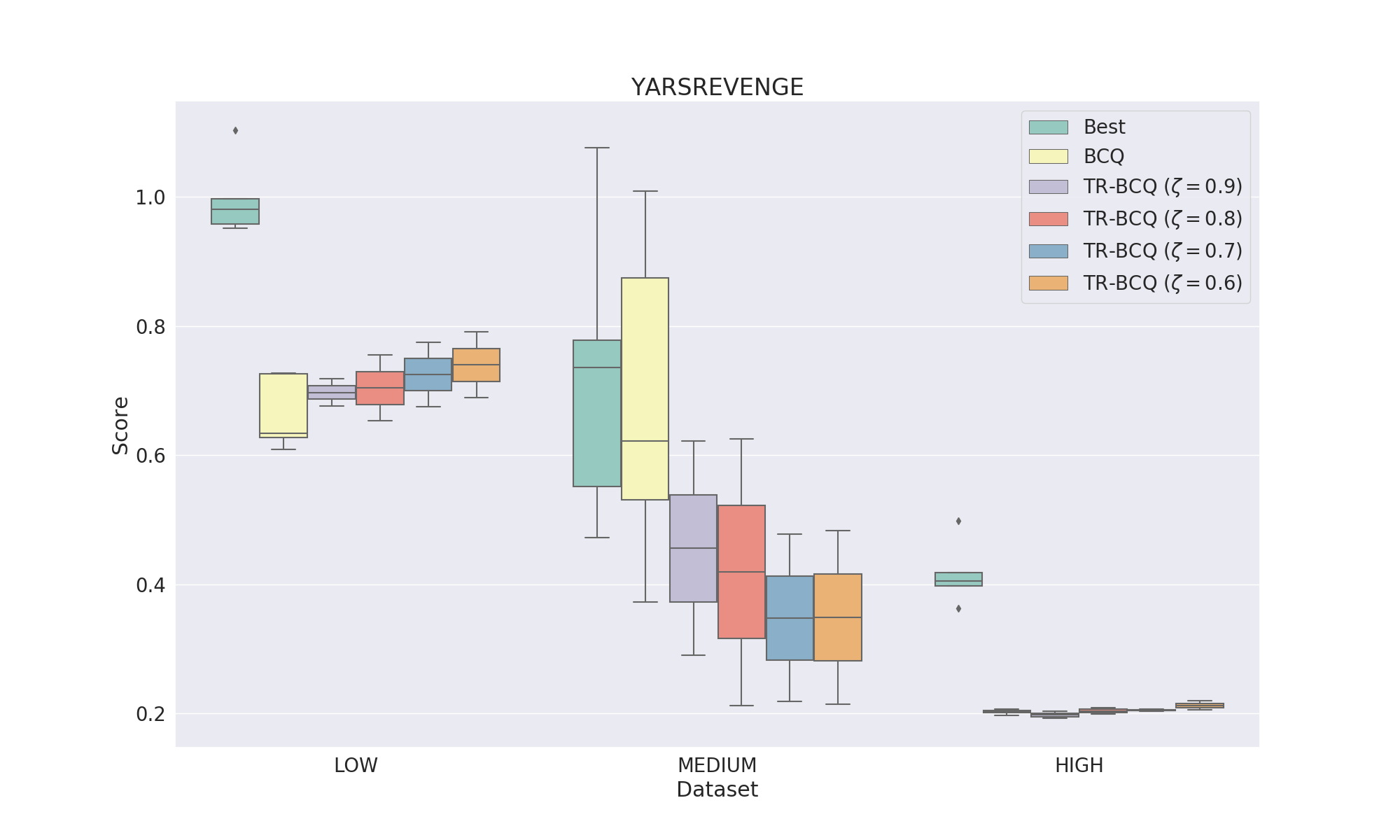}\\
				\vspace{0.01cm}
			\end{minipage}%
		}%
		\subfigure{
			\begin{minipage}[t]{0.333\linewidth}
				\centering
				\includegraphics[width=2.3in]{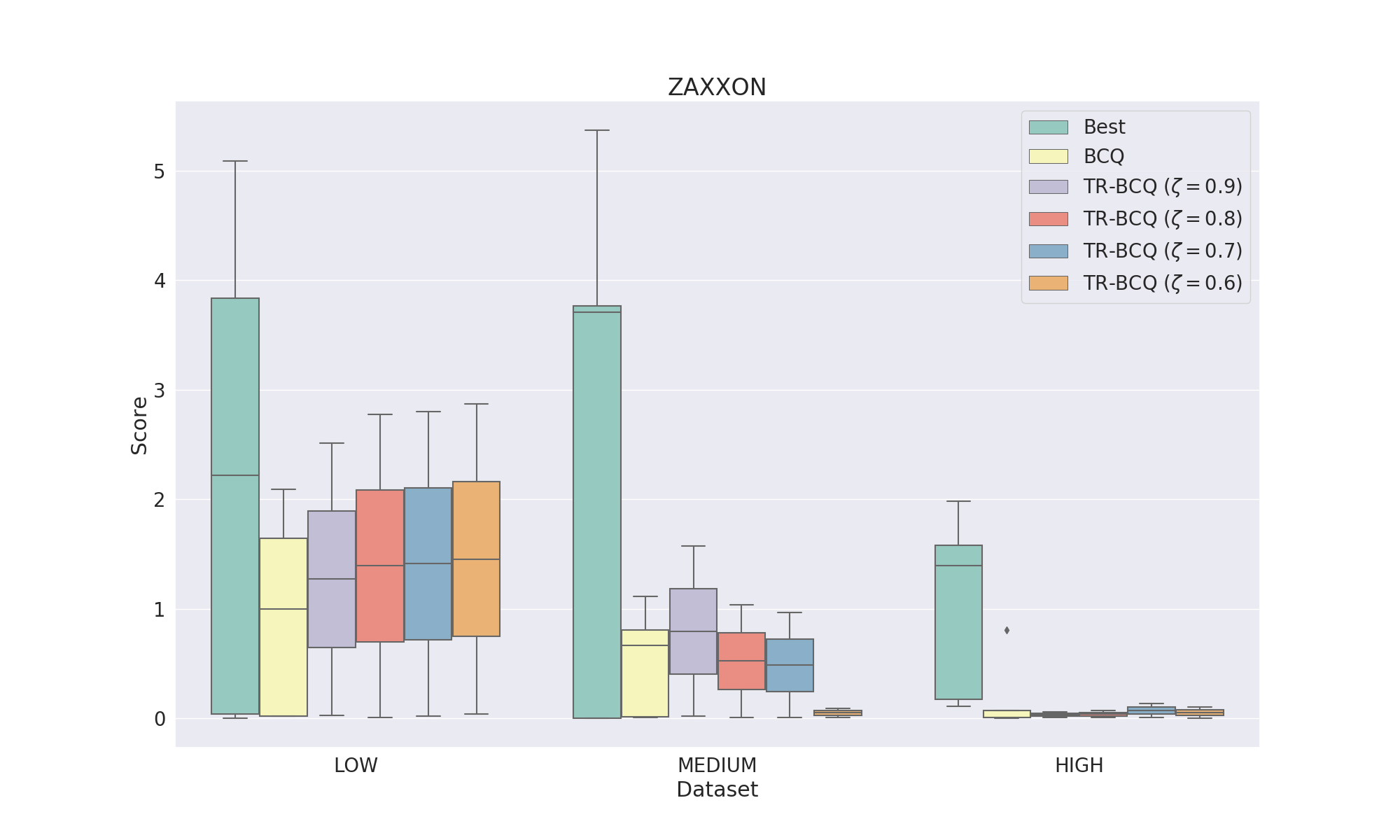}\\
				\vspace{0.01cm}
			\end{minipage}%
		}%

		\centering
		\caption{\textbf{Comparison between  TR-BCQ and best baselines on different datasets from Game UpNDown to Game Zaxxon}}
		\label{fig: Comparison between  TR-BCQ and the best baselines on different datasets from Game UpNDown to Game Zaxxon}
								
	\end{figure*}

\end{document}